%% file: article.tex
\PassOptionsToPackage{pdfauthor={},pdftitle={}}{hyperref}
\documentclass[aop]{imsart}

\setpkgattr{journal}{name}{\empty}
\setpkgattr{author}{prefix}{}

\RequirePackage{amsthm,amsmath,amsfonts,amssymb, yhmath}
\RequirePackage[numbers]{natbib}
\RequirePackage[colorlinks,citecolor=blue,urlcolor=blue]{hyperref}
\RequirePackage{graphicx}

\usepackage{comment}
\usepackage{natbib} 
\usepackage[usenames,dvipsnames]{color}
\definecolor{RefColor}{rgb}{0,0,.85}  

\makeatletter
\def\acks{\gdef\@thefnmark{}\@footnotetext}
\makeatother  

\usepackage{mathtools}
\usepackage[capitalize]{cleveref}
\usepackage{dsfont}
\usepackage{mathdots}
\usepackage{nccmath} 
\usepackage{scalerel}
\usepackage[hang,flushmargin,para]{footmisc}
\usepackage{caption}
\usepackage{titlesec}
\titleformat*{\section}{\large\bfseries}
\titleformat*{\subsection}{\bfseries}

\RequirePackage[OT1]{fontenc}

\usepackage[english]{babel}

\usepackage{enumitem}
\setlist[itemize]{leftmargin=1.5em}

\usepackage{booktabs}
\usepackage{tikz}
\usetikzlibrary{calc}
\usetikzlibrary{decorations.markings,decorations.pathreplacing}
\tikzstyle{mybraces}=[mirrorbrace/.style={
          decoration={brace, mirror},
          decorate},brace/.style={
          decoration={brace},
          decorate}]

\usepackage{amsmath,amssymb,amscd,amsfonts,amsthm,mathtools}
\usepackage{thmtools}

\startlocaldefs

\theoremstyle{plain}
\declaretheoremstyle[postheadspace=.4em,headfont=\bfseries,bodyfont=\itshape,spaceabove=8pt,
spacebelow=10pt]{basic}
\theoremstyle{basic}
\declaretheorem[style=plain,name={Theorem}]{theorem}
\declaretheorem[style=plain,sibling=theorem,name={Lemma}]{lemma}
\declaretheorem[style=plain,sibling=theorem,name={Proposition}]{proposition}
\declaretheorem[style=plain,sibling=theorem,name={Corollary}]{corollary}
\theoremstyle{plain}

\declaretheorem[style=remark,name={Remark}]{remark}
\declaretheorem[style=remark,name={Remark},numbered=no]{remark*}

\declaretheorem[style=plain,name={Assumption}]{assumption}

\makeatletter
\def\thmhead@plain#1#2#3{%
  \thmname{#1}\thmnumber{\@ifnotempty{#1}{ }\@upn{#2}}%
  \thmnote{ {\the\thm@notefont#3}}}
\let\thmhead\thmhead@plain
\makeatother
\theoremstyle{plain}
\declaretheorem[name={Assumption},numbered=no]{nonumassumption}

\crefname{assumption}{Assumption}{Assumptions}

\newenvironment{proplist}{\begin{enumerate}[
    label=(\roman{enumi}),
    ]}
{\end{enumerate}}

\newcommand{\argdot}{{\,\vcenter{\hbox{\tiny$\bullet$}}\,}}

\newcommand{\tagaligneq}{\refstepcounter{equation}\tag{\theequation}}

\newcommand{\msup}{\sup\nolimits}
\newcommand{\minf}{\inf\nolimits}
\newcommand{\mmax}{\max\nolimits}

\newcommand{\Tr}{\text{\rm Tr}}
\newcommand{\ind}{\mathbb{I}}

\usepackage[normalem]{ulem}

\def\bA{\mathbf{A}}
\def\bB{\mathbf{B}}

\def\bH{\mathbf{H}}
\def\bI{\mathbf{I}}

\def\bO{\mathbf{O}}

\def\bR{\mathbf{R}}
\def\bS{\mathbf{S}}
\def\bT{\mathbf{T}}
\def\bU{\mathbf{U}}
\def\bV{\mathbf{V}}
\def\bW{\mathbf{W}}
\def\bX{\mathbf{X}}
\def\bY{\mathbf{Y}}
\def\bZ{\mathbf{Z}}

\def\ba{\mathbf{a}}
\def\bb{\mathbf{b}}

\def\be{\mathbf{e}}

\def\bo{\mathbf{o}}

\def\bt{\mathbf{t}}
\def\bu{\mathbf{u}}
\def\bv{\mathbf{v}}
\def\bw{\mathbf{w}}
\def\bx{\mathbf{x}}
\def\by{\mathbf{y}}

\def\bphi{\mathbf{\phi}}

\def\bpsi{\mathbf{\psi}}

\def\bzero{\mathbf{0}}
\def\bone{\mathbf{1}}

\def\M{\mathbb{M}}
\def\N{\mathbb{N}}

\def\P{\mathbb{P}}

\def\R{\mathbb{R}}

\def\Z{\mathbb{Z}}

\def\q{\mathbb{q}}

\def\cA{\mathcal{A}}
\def\cB{\mathcal{B}}

\def\cD{\mathcal{D}}

\def\cF{\mathcal{F}}

\def\cH{\mathcal{H}}
\def\cI{\mathcal{I}}
\def\cJ{\mathcal{J}}

\def\cL{\mathcal{L}}

\def\cN{\mathcal{N}}
\def\cO{\mathcal{O}}

\def\cS{\mathcal{S}}
\def\cT{\mathcal{T}}

\def\cW{\mathcal{W}}
\def\cX{\mathcal{X}}

\def\cZ{\mathcal{Z}}

\newcommand{\mean}{\mathbb{E}}
\newcommand{\Var}{\text{\rm Var}}
\newcommand{\Cov}{\text{\rm Cov}}

\newcommand{\darrow}{\xrightarrow{\;\text{\tiny\rm d}\;}}

\newcommand{\equdist}{\stackrel{\text{\rm\tiny d}}{=}}

\newcommand{\braces}[1]{{\lbrace #1 \rbrace}}

\DeclareMathOperator{\tsum}{{\textstyle\sum}}

\DeclareMathOperator{\msum}{\medmath\sum}
\DeclareMathOperator{\mprod}{\medmath\prod}
\DeclareMathOperator{\mint}{\scaleobj{.8}{\int}}

\DeclareMathOperator{\argmin}{\text{\rm argmin}}

\newcommand{\tvec}{\text{\rm vec}}

\def\diag{\text{\rm diag}}

\usepackage{accents}

\endlocaldefs

\begin{document}




\begin{frontmatter}
  \title{Gaussian and Non-Gaussian Universality of Data Augmentation}

\begin{aug}
    \author{\fnms{Kevin Han}~\snm{Huang}},
    \author{\fnms{Peter}~\snm{Orbanz}}
    \and    
    \author{\fnms{Morgane}~\snm{Austern}}
    \\[1em]\normalfont\small University of Warwick, University College London and Harvard University
\end{aug}

  

\begin{abstract}
We provide universality results that 
quantify how data augmentation affects the variance and limiting distribution of 
estimates through simple surrogates, and analyze several specific models in detail.
The results 
confirm some observations made in machine learning practice, but also lead to unexpected findings: Data augmentation may increase rather than decrease
the uncertainty of estimates, such as the empirical prediction risk. 
It can act as a regularizer, but
fails to do so in certain high-dimensional problems, and it may shift the double-descent peak of an empirical risk.
Overall, the analysis shows that several properties data augmentation has been attributed with are not either true or false, but rather depend on a combination of factors---notably the data distribution, the properties of the estimator, and the interplay of sample size, number of augmentations, and dimension.   
As our main theoretical tool, we develop an adaptation of Lindeberg's technique for block dependence. The resulting universality regime may be Gaussian or non-Gaussian.
\end{abstract}

  

\end{frontmatter}

\input{article_sec1_intro.tex}

\input{article_sec2_def.tex}

\input{article_sec3_main.tex}

\input{article_sec4_average.tex}

\input{article_sec5_ridge.tex}

\input{article_sec6_ridgeless.tex}

\input{article_sec6_newstuff.tex}

\input{article_sec7_misc.tex}
 \subsection*{Acknowledgements}
 KHH acknowledges funding from the Gatsby Charitable Foundation and the EPSRC grant EP/Y028783/1 (Prob\_AI). PO is supported by the Gatsby Charitable Foundation.

\bibliographystyle{imsart-number}

\bibliography{references}

\newpage
\setcounter{page}{1}
\thispagestyle{empty}
\appendix

\input{appendix}

\end{document}

%% file: article_sec1_intro.tex
\section{Introduction}
\label{sec:intro}

The term \emph{data augmentation} refers to a range of machine learning heuristics
that synthetically enlarge a training
data set: Random transformations are applied to
each training data point, and the transformed points are added to
the training data \citep[e.g.][]{taqi2018impact,shorten2019survey}.
(This meaning of the term data augmentation should not be confused with a separate meaning in statistics, 
which refers to the use of latent variables e.g.\ in the EM algorithm.)
It has quickly become one of the most widely used heuristics in 
machine learning practice, and the scope of the term continues to evolve.
One objective may be to make a
neural network less sensitive to rotations of input images, by
augmenting data with random rotations of training samples \citep[e.g.][]{perez2017effectiveness}.
In other cases, one
may simply reason that ``more data is always better''.

The question how data augmentation affects learning rates remains open. It has been argued that augmentation reduces the variance of estimates
\citep{zhang2021understanding}, that it increases the
effective sample size \citep{balestriero2022data}, and that it acts as a regularizer \citep{balestriero2022effects}, but none of these points have been rigorously established.
Existing analysis studies the bias of estimates 
\citep{balestriero2022effects}, and shows a reduction of variance
for certain \emph{parametric} M-estimators under additional invariance assumptions
\citep{chen2020group}. In the following, we study the limiting behavior of augmentation methods. Two mathematical obstacles are (1) that augmentation makes independently distributed data dependent, and (2) that data may be high-dimensional. One may therefore expect the behavior of augmented estimates to be highly sensitive to the input distribution. We show that, on the contrary, augmented statistics exhibit a form of \emph{universality}: Under general stability conditions, the learning rate of estimates depends on the expectation and covariance matrix of the observations, but is independent of all higher moments (see \cref{thm:main}).

The universality phenomenon is a subject of a fast-growing body of literature \citep{rotar1976limit,rotar1979limit, chatterjee2006generalization,mossel2010,bally2019total}. In statistics and machine learning, it has been applied to various estimators including specific generalized linear models, perceptron models, max-margin estimators and others obtained by empirical risk minimization \cite{montanari2022universality,montanari2023universality,dandi2024universality,gerace2022gaussian,korada2011applications, han2023universality, hu2022universality}. Non-Gaussian generalizations have been established in random matrix theory \citep{arous2008spectrum,deya2014invariance}, and relaxations to weak dependencies are obtained for specific applications \citep{bryson2021marchenko,dudeja2023universality}. 
In contrast to these examples, data augmentation introduces strong dependence that persists asymptotically. The tools we develop allow us to handle this form of dependence, and 
 to analyze specific problems in both Gaussian and non-Gaussian universality regimes. 
The results show that a number of properties commonly attributed to data augmentation --- variance reduction, increase in effective sample size, and regularization ---
each occur in certain cases, but fail in others.

\begin{figure}[t]
\centering
\begin{tikzpicture}
\node[inner sep=0pt] at (0,0) 
    {\includegraphics[width=.48\textwidth]{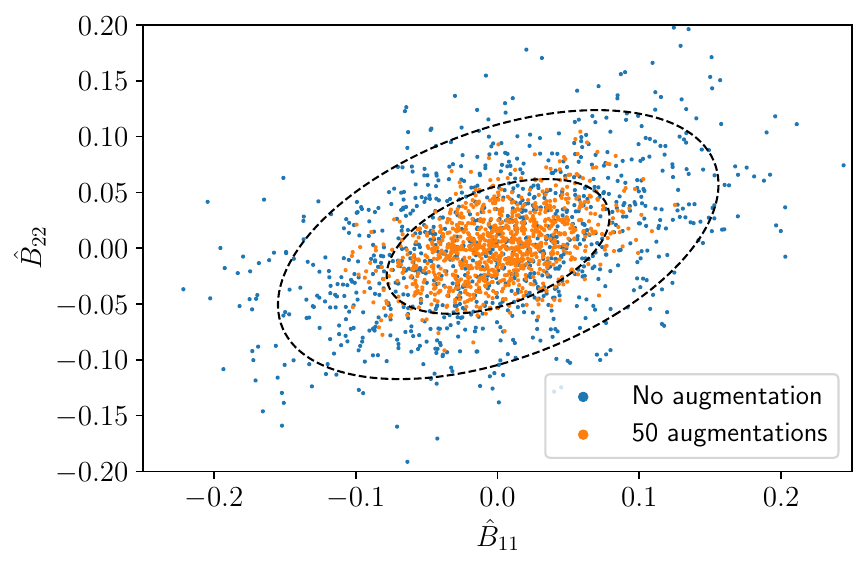}};
\node[inner sep=0pt] at (7.2,0) 
    {\includegraphics[width=.48\textwidth]{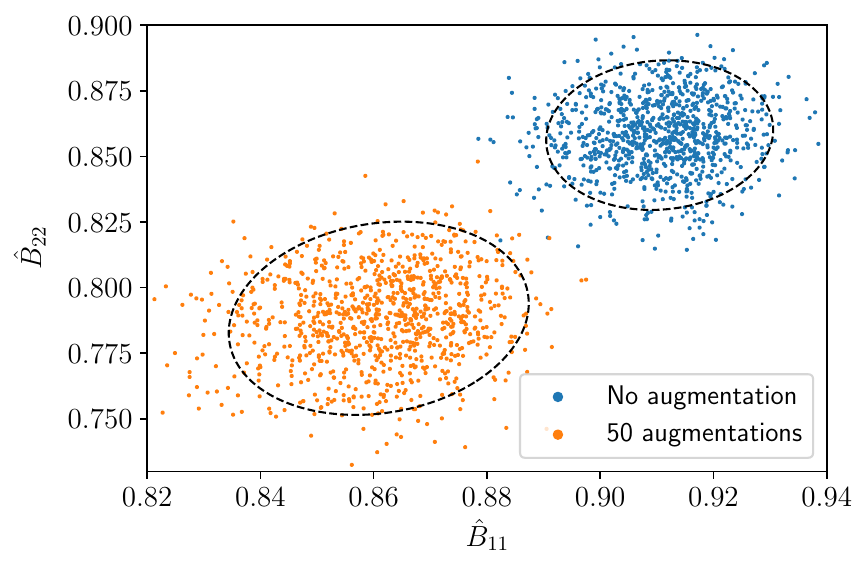}};
\end{tikzpicture}
\centering
\caption{
Effect of augmentation on the variability of estimates. \emph{Left:} On an empirical average. \emph{Right:} On a ridge regression estimator. Each point is an estimate computed from a single simulation experiment, and the dashed lines are the 95\% 2d quantiles of the empirical distribution over 1000 simulations. Augmentation reduces the variability in the left plot, but increases the uncertainty of the estimate in the right plot. See Remark \ref{remark:ridge} in 
Section \ref{sectn:ridge_regression} for details on the plotted experiments.
}
\label{fig:intro_plot}
\end{figure}

\subsection{A non-technical overview}

The remainder of this section sketches our results informally.
Rigorous definitions follow in Section \ref{sec:definitions}.
Our general setup is as follows: Given is a data set, consisting of observations that we assume to be $d$-dimensional i.i.d.\ random vectors
in $\cD \subseteq \mathbb{R}^d$. We are interested in
estimating a quantity $\theta\in\mathbb{R}^q$, for some $q$. This may
be a model parameter, the value of a risk function or a
statistic, and so forth. The data is augmented by applying $k$ randomly
generated transformations to each data point. That yields an 
augmented data set of size $n\cdot k$. An estimator for $\theta$ is
then a function ${f:\cD^{nk}\rightarrow\mathbb{R}^q}$, and
we estimate $\theta$ as
\begin{equation*}
  \text{estimate of }\theta\;=\;f(\text{augmented data})\;.
\end{equation*}
From a statistical perspective, this can be regarded as a form of sample randomization.
As for other randomization techniques, such as the bootstrap or cross-validation, quantitative
analysis of augmentation is complicated by the fact that randomized data points are not independent.
To study such augmented estimates, we rely on the Linderberg's method developed by \cite{chatterjee2006generalization,mossel2010}, and assume that our statistics $f$
satisfy a ``noise stability'' condition (see \cref{sec:definitions}). Informally, noise stability means
that $f$ is not too sensitive to small perturbations of any input coordinate. Examples of noise-stable statistics include sample averages
(such as empirical risks or plug-in estimators), but also overparameterized linear regression, ridge regression, bagged estimators, and
general M-estimators \citep{mei2022generalization,soloff2024bagging,montanari2022universality}.
  Our \cref{thm:main} shows that the distribution of our augmented estimator is identical to the distribution of an estimator trained on some surrogate random variables. More precisely, for
all $h$ in a certain class $\cH$ of smooth functions, we show that 
\begin{equation*}
  \bigl|\,\mean[h(f(\text{augmented data}))]
  \,-\,
  \mean[h(f(\text{generic surrogate variables}))]\,\bigr|
  \;\leq\;
  \tau(n,k)\;.
\end{equation*}
The surrogates are variables completely determined by their mean and variance; depending on the problem, they may be Gaussian (e.g. for sample averages) or non-Gaussian (e.g. for ridge regression).
Under general conditions, ${\tau\rightarrow 0}$, hence the limiting distribution of $f(\text{augmented data})$ is
that of $f(\text{surrogates})$.
In other words, the effect of augmentation on a noise-stable estimator is \emph{completely determined by
two moments} as $n$ grows large. 
The theorem specifies these moments explicitly. 
That allows us to study the limiting estimator and its variance, and to read off the rate of convergence from $\tau$. For sufficiently linear estimators, we can also draw consistent confidence intervals and evaluate their width.
\\[.5em]

\begin{figure}[t]
\centering
\begin{tikzpicture}
    \node[inner sep=0pt] at (3,-5.2)
        {\includegraphics[width=.6\textwidth]{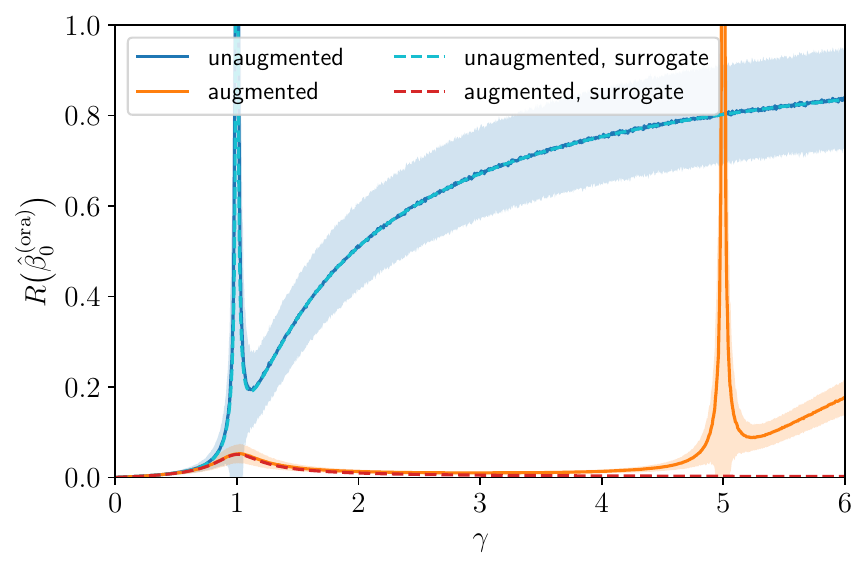}};
\end{tikzpicture}
\centering
\caption{
Effect of an oracle choice of augmentation on the limiting risk of a high-dimensional ridgeless regressor under the asymptotic $d/n \rightarrow \gamma$. A regularization effect is observed around $\gamma = 1$, whereas a new double-descent peak shows up at $\gamma = 5 = k$, the number of augmentations. See \cref{sec:triple:descent:oracle} for the detailed setup. 
}
\label{fig:intro_dd}
\end{figure}

\noindent\textbf{Applications to specific models}.
The function $\tau$ is determined by terms that quantify the noise stability of $f$.
For a given estimator, we can evaluate these terms to verify 
  how fast $\tau$ converges to 0
as
either $n$ or $k$ grows large. 
  This establishes how fast the universality property happens, and we use this to gain insights into the effect of data augmentation for a few different models
:
\\[.2em]
\noindent\textbf{1) Underparameterized models}.
We analyze empirical averages, plug-in estimators, the risk of M-estimators (\cref{sec:average:plugin}) and ridge regression (\cref{sectn:ridge_regression}).
For empirical averages and risks,
we characterize exactly when augmentation reduces variance. 
These results hold more generally for a class of linear sample statistics. For non-linear estimators, the behavior can change significantly:
Augmentation may increase rather than decrease variance. That can occur even in simple models, such as the ridge regression example (see the right plot of \cref{fig:intro_plot}). 
\\[.2em]
\noindent\textbf{2) Overparameterized models}. 
We first analyze the limiting risk of a high-dimensional ridgeless regressor under isotropic noise injection.
Without augmentation, this model is known to exhibit double descent \citep{hastie2022surprises}. We show that the behavior under augmentation 
depends on an interplay of scales: If $d\approx n$, augmentation acts as a regularizer. For higher dimension, namely ${d\approx nk}$, it
causes the risk to diverge to infinity. It can also shift the double-descent peak---see \cref{fig:intro_dd}. We also extend our results to simple neural network models, augmentations beyond noise injection, and bagged estimators of non-linear neural networks. 
%
\\[.5em]
\noindent\textbf{Some key findings 
  about the behavior of data augmentation}.
To place our results in context,
we note three hypotheses generally made in the
existing literature and are either explicitly or implicitly required by proofs \citep[e.g.][]{dao2019kernel,chen2020group,balestriero2022data}: 
(i) Linearity or approximate linearity of the estimator, in the sense that $f$ is linear in contributions of individual data points (typically,
a sample average).
(ii) Invariance of the data source, i.e.\ the transformations used to perform augmentation leave the
data distribution invariant.
(iii) The number of transformations applied to each data point diverges, i.e.\ ${k\rightarrow\infty}$.
In the context of (iii), it is helpful to note that
transformations can be applied once before fitting a model (\emph{offline augmentation}),
or repeatedly during each step of a training algorithm (\emph{online augmentation}). Online augmentation
is feasible if each transformation is computationally cheap (e.g.\ rotations in computer vision). 
Offline augmentation is particularly common in natural language processing, where
more expensive transformations have emerged as useful \citep{feng2021survey}.
The assumption ${k\rightarrow\infty}$ is justified by choosing an online setup and arguing that the
number of steps of the training algorithm is effectively infinite; offline augmentation implies ${k<\infty}$. \cref{thm:main} allows us to drop each of these assumptions, and overall, our results show that
doing so can change the behavior of augmentation decisively. In more detail, our results show the following:
\\[.2em]
\noindent\textbf{1) Augmentation may or may not reduce variance}.
Augmentation is known to reduce variance under assumptions (i)---(iii) above, but
empirical observations by \cite{lyle2020benefits} suggest this may not be true in practice.
\cref{thm:main} allows us to make more detailed statements:
If $f$ is linear, augmentation reduces variance if the transformations do not increase the variance of the data distribution (\cref{sec:plugin}).
If $f$ is non-linear, variance may increase, even if distributional invariance holds (\cref{sec:nonlinear:average} and \cref{sectn:ridge_regression}).
More generally, the effects of augmentation depend not only on the data distribution, but also on the estimator $f$.\\[.2em]
\noindent\textbf{2) Invariance is not essential} for augmentation, regardless of whether $f$ is linear or non-linear. For linear $f$, the relevant criterion for variance reduction is that augmentation does not increase the variance of data variables (\cref{sec:averages}). The invariance assumption (ii) is one way to ensure this, but is not required: Invariance implies all moments are constant under transformation. What matters is that the second moment does not grow. 
\\[.2em]
\noindent\textbf{3) Augmentation and regularization}. It has been argued that data augmentation can be interpreted as a form of regularization \citep[e.g.][]{balestriero2022effects}.
Our results show that augmentation can indeed act as a regularizer, but whether it does depends on details of the application---specifically, on how the sample size $n$, the dimension $d$, and the number $k$ of augmentations per data point grow relative to each other (\cref{sec:ridgeless}).
\\[.2em]
\noindent\textbf{4) Whether augmentation is performed offline or online matters}.
If ${k<\infty}$, data augmentation may not regularize (\cref{sec:ridgeless}). This manifests for ${d\approx nk}$ in the
double-descent peak of the risk in \cref{fig:intro_dd}.
\\[.2em]

\noindent\textbf{In summary},
\cref{thm:main} can be used to derive statistical guarantees for a range of augmented estimators.
Several hypotheses on augmentation considered in machine learning turn out not to be either true or false, but rather depend on the data distribution, the properties of the estimator, and the interplay of sample size, number of augmentations, and dimension. 
The results may also be a step towards making data augmentation a viable technique for statisticians who seek guarantees for the methods they employ.
\\[.2em]
\noindent\textbf{Structure of the article}. \cref{sec:definitions} defines the setup and the concept of noise stability. Theoretical results---the main theorem and a number of consequences---follow in \cref{sec:main:result}. The remaining sections apply these results to linear estimators (\cref{sec:average:plugin}), ridge regression (\cref{sectn:ridge_regression}), an overparameterized models that exhibits double descent~(\cref{sec:triple:descent:oracle,sec:triple:descent:two:stage}), simple neural networks (\cref{sec:network}) and bagged estimators (\cref{sec:bagging}).
All proofs are collected in the~appendix.

%% file: article_sec2_def.tex
\section{Definitions}
\label{sec:definitions}

\textbf{Data and augmentation}.
Throughout, we consider a data set $\cX\coloneqq(\bX_1,\ldots,\bX_n)$, where the $\bX_i$ are i.i.d.\ random elements of some fixed convex subset $\cD \subseteq \mathbb{R}^d$ that contains $\bzero$. The choice of $\bzero$ is for convenience and can be replaced by any other reference point. Let $\cT$ be a set of (measurable)
maps ${\cD \rightarrow \cD}$, and fix some ${k\in\mathbb{N}}$. We generate
$nk$ i.i.d.\ random elements ${\phi_{11},\ldots,\phi_{nk}}$ of $\cT$, and abbreviate
\begin{align*}
  \Phi_i&\coloneqq(\phi_{ij}|j\leq k)
  &
  \Phi&\coloneqq(\phi_{ij}|i\leq n,j\leq k)
  &
  \Phi_i\bX_i&\coloneqq(\phi_{i1}\bX_i,\ldots,\phi_{ik}\bX_i)
  \;.
\end{align*} 
The augmented data is then the ordered list
\begin{equation*}
  \Phi\cX
  \;\coloneqq\;
  (\Phi_1\bX_1,\ldots,\Phi_n\bX_n)
  \;=\;
  (\phi_{11}\bX_{1},\ldots,\phi_{1k}\bX_1,\ldots,\phi_{n1}\bX_n,\ldots,\phi_{nk}\bX_n)\;.
\end{equation*}
Here and throughout, we do not distinguish between a vector and its
transpose, and regard the quantities above
as vectors ${\Phi_i\bX_i\in\cD^k}$
and ${\Phi\cX\in\cD^{nk}}$ where convenient. \\[.5em]
\textbf{Estimates}. An estimate computed from augmented data is the value 
\begin{equation*}
  f(\Phi\cX)\;=\;
  f(\phi_{11}\bX_1,\ldots,\phi_{nk}\bX_n)
\end{equation*}
of a function ${f:\cD^{nk}\rightarrow\mathbb{R}^q}$, for some
${q\in\mathbb{N}}$.
An example is an empirical risk: If $S$ is a regression
function ${\R^d \rightarrow\R}$ (such as a statistic
or a feed-forward neural network), and $C(\hat y, y)$ is
the cost of a prediction $\hat y$ with respect to $y$, 
one might choose $\phi_{ij} = (\pi_{ij}, \tau_{ij})$ as a pair of transformations acting respectively on $\bv \in \R^d$ and $y \in \R$ and $\bX_i = (\bV_i, \bY_i)$, in which case $f(\Phi\cX)$ is the empirical risk
${\frac{1}{nk}\sum_{i\leq n,j\leq k} C(S(\pi_{ij}\bV_i), \tau_{ij} \bY_i  )}$. 
However, we do \emph{not} require that $f$ is
a sum, and other examples are given in Section \ref{sectn:ridge_regression} and \ref{sec:ridgeless}.
\\[.5em]
\textbf{Norms}. Three types of norms appear in what follows: For vectors and tensors,
we use both a ``flattened'' Euclidean norm and its induced operator norm: If
${\bx\in\mathbb{R}^{d_1\times\cdots\times d_m}}$ and ${A\in\mathbb{R}^{d\times d}}$,
\begin{align*}
  \|\bx\|\;&\coloneqq\;\bigl(\,\msum_{i_1\leq d_1,\ldots,i_m\leq
      d_m}|x_{i_1,\ldots,i_m}|^2\bigr)^{1/2}
  &\text{and}&&
  \|A\|_{op}\;\coloneqq\;\sup_{\bv\in\mathbb{R}^d}\mfrac{\|A\bv\|}{\|\bv\|}\;.
\end{align*}
Thus, $\|\bv\|$ is the Euclidean norm of $\bv$ for ${m=1}$, the Frobenius
norm for ${m=2}$, etc. For real-valued random variables $X$, we also use $L_p$-norms, denoted by ${\|X\|_{L_p}\coloneqq\mean[|X|^p]^{1/p}}$. 
\\[.5em]
\textbf{Covariance structure}.
For random vectors $\bY$ and $\bY'$ in $\mathbb{R}^m$, we define 
the ${m\times m}$ covariance matrices 
\begin{align*}
  \Cov[\bY,\bY']\;&\coloneqq\;(\Cov[Y_i,Y'_j])_{i,j\leq m}
  &
  \text{and}&
  &
  \Var[\bY]\;&\coloneqq\;\Cov[\bY,\bY]\;.
\end{align*}
Augmentation introduces dependence: Applying independent random elements $\phi$ and $\psi$ of $\cT$ 
to the same observation $\bX$ results in dependent vectors
$\phi(\bX)$ and $\psi(\bX)$. In the augmented data set,
the entries of each vector $\Phi_i\bX_i$ are hence dependent,
whereas $\Phi_i\bX_i$ and $\Phi_j\bX_j$ are independent if
${i\neq j}$. That partitions the covariance matrix $\Var[\Phi\cX]$ into
${n\times n}$ blocks of size ${kd\times kd}$, and makes it block-diagonal. This block structure is visible in all our results, and
makes Kronecker notation convenient:
For a matrix ${A\in\R^{m\times n}}$ and a matrix $B$ of
arbitrary size, define the Kronecker product
\begin{equation*}
  A\otimes B
  \;\coloneqq\;
  \bigl( A_{ij}B \bigr)_{i\leq m,\,j\leq n}
\end{equation*}
We write ${A^{\otimes k}\coloneqq A\otimes\cdots\otimes A}$
for the $k$-fold product of $A$ with itself.
If $\bv$ and $\bw$ are vectors,
$\bv\otimes\bw=\bv\bw^\top$ is the outer product.
To represent block-diagonal or off-diagonal matrices,
let ${\bI_k}$ be the $k\times k$ identity matrix, and
$\bone_{k\times m}$ a $k\times m$ matrix all of whose entries are
1. Then
\begin{align*}
  \bI_{k}\otimes B
  \;=&\;
  \begin{psmallmatrix}
    B & 0 & 0 & \cdots \\[.1em] 
    0 & B & 0 &  \\[.1em]
    0 & 0 & B &  \\[-.4em]
    {\scriptsize\vdots} & & & {\scriptsize\ddots} 
  \end{psmallmatrix}
  & 
  \text{and}&&
    (\bone_{k\times k}-\bI_{k})\otimes B
    \;=&\;
    \begin{psmallmatrix} 
    0 & B & B & \cdots \\[.1em]
    B & 0 & B &  \\[.1em]
    B & B & 0 &  \\[-.4em]
    {\scriptsize\vdots} & & & {\scriptsize\ddots} 
    \end{psmallmatrix}\;.
\end{align*} 
\textbf{Measuring noise stability}. Our results require a control over the noise stability of $f$ and smoothness of test function $h$, which we define next. 

\vspace{.3em}

Write $\cF_r(\cD^{a},\mathbb{R}^{b})$ for the class of $r$ times differentiable functions
${\cD^a\rightarrow\mathbb{R}^b}$. 
To control how stable a function 
${f\in\cF_r(\cD^{nk},\R^q)}$ is with respect to random perturbation of its 
arguments, we regard it as a function of $n$ arguments
${\bv_1,\ldots,\bv_n\in\cD^k}$. That reflects the block structure above---noise
can only be added separately to components that are independent.
We write $\cL(\cA,\cB)$ as the set of bounded linear functions $\cA \rightarrow \cB$, and denote by $D_i^m$ the $m$th derivative with respect to the $i$th component, 
\begin{equation*}
  D_i^m f(\bv_1,\ldots,\bv_n)
  \;\coloneqq\;
  \mfrac{\partial^m f}{\partial \bv_i^m}\,(\bv_1,\ldots,\bv_n)
  \;\in\;
  \cL\big( (\cD^{k})^m, \R^q \big)
  \;\subseteq\;
  \R^{q \times (dk)^m}
  \;.
\end{equation*}
For instance, if ${q=1}$ and $g$ is the function
${g(\argdot)\coloneqq f(\bv_1,\ldots,\bv_{i-1},\argdot,\bv_{i+1},\ldots,\bv_n)}$,
then ${D^1_if}$ is the transposed gradient $\nabla g^{\top}$, and $D^2_if$ is
the Hessian matrix of $g$.
To measure the sensitivity of $f$ with respect to each of its
$d\times k$ dimensional arguments, we define
\begin{equation*}
    \bW_i(\argdot) \coloneqq (\Phi_1\bX_1,\ldots,\Phi_{i-1}\bX_{i-1},\argdot,\bZ_{i+1},\ldots,\bZ_n)\;,
\end{equation*} 
where $\bZ_j$ are i.i.d.\ surrogate random vectors in $\cD^k$ with first two moments matching those of $\Phi_1\bX_1$: Defining the $d \times d$ matrices 
${\Sigma_{11}\coloneqq\Var[\phi_{11}\bX_1]}$ and
${\Sigma_{12}\coloneqq\Cov[\phi_{11}\bX_1,\phi_{12}\bX_1]}$,
\begin{align} \label{eqn:defn:surrogates}
    \mean \bZ_i
    \;&=\;
    \mathbf{1}_{k\times 1}\otimes\mean[\phi_{11}\bX_1]
    \;
    &\text{and}&&
    \Var \bZ_i
    \;&=\;
    \mathbf{I}_{k}\otimes \Sigma_{11}
    +
    (\mathbf{1}_{k\times k}-\mathbf{I}_k)\otimes \Sigma_{12}\;.
\end{align}
Write $f_s: \cD^{nk} \rightarrow \R$ as the $s$-th coordinate of $f$. Noise stability is measured by
\begin{equation} \label{eqn:defn:alpha_r}
  \alpha_r
  \coloneqq
  \sum_{s\leq q}\max_{i\leq n} \max\bigl\{
  \bigl\|{\sup_{\bw\in[\bzero,\Phi_i\bX_i]}}
  \|D_i^r f_s(\bW_i(\bw))\|\bigr\|_{L_{6}},
    \bigl\|{\sup_{\bw\in[\bzero,\bZ_i]}}
  \|D_i^r f_s(\bW_i(\bw))\|\bigr\|_{L_{6}}
  \bigr\}\;,
\end{equation}
where we have used $[\ba,\bb]$ to represent the set  $\{ c \,\ba + (1-c) \bb : c \in [0,1]\}$. This is a non-negative scalar, and large
values indicate high sensitivity to changes of individual arguments (low noise stability).
Our results also use test functions ${h:\R^q\rightarrow\R}$. For these, we measure smoothness
simply as differentiability, using the scalar quantities
\begin{align*}
  \gamma_r(h)
  \;&\coloneqq\
  \sup
  \braces{ \| \partial^r h(\bv) \| \,|\,\bv \in\R^q}\;,
\end{align*} 
where $\partial^r$ denotes the $r$th differential, i.e.\ $\partial^1h$ is the gradient, $\partial^2h$ the
Hessian, etc.
In the result below, these terms appear in the form of the linear combination
\begin{align} \label{eqn:defn:lambda}
  \lambda(n,k)
  \;\coloneqq\;
  \gamma_3(h) \alpha_1^3 + 3 \gamma_2(h) \alpha_1 \alpha_2 + \gamma_1(h) \alpha_3
  \;.
\end{align}
$\lambda(n,k)$ can then be computed explicitly for specific
models. We note that
the dependence on $n$ and $k$ is via the definition of $\alpha_r$, and that
derivatives appear up to 3rd order and moments up to 6th order. Notably, these conditions require that the effect of changing one data point on the first derivative of $f$ is $o(n^{-1/3})$.\\[.5em]
\textbf{Moment conditions}. Our results also require the following 6th moments on data and the surrogate variables: Write $\bZ_1 = (Z_{1jl})_{j\leq k, l \leq d}$ where $Z_{ijl} \in \R$, and define
\begin{align} \label{eqn:defn:moments}
    &
  c_X\;\coloneqq\;\mfrac{1}{6}\sqrt{\mean\|\phi_{11} \bX_1 \|^6 }
  \quad\text{ and }\quad
    c_Z\;\coloneqq\;\mfrac{1}{6} \sqrt{ \mean\Bigl[ \Bigl(\mfrac{|Z_{111}|^2+\ldots+|Z_{1kd}|^2}{k}\Bigr)^3\Bigr] }\;.
\end{align}

%% file: article_sec3_main.tex
\section{Theoretical results}
\label{sec:main:result}

We now state our main theoretical result and several immediate
consequences. \cref{sec:intro} sketches the main result in
terms of an upper bound $\tau(n,k)$. With the definitions
above, $\tau$ becomes a function measuring noise
stability of $f$ and smoothness of $h$.
\begin{theorem}{\rm (Main result)}
\label{thm:main} Consider i.i.d.\ random elements ${\bX_1,\ldots,\bX_n}$ of
$\cD$, and two functions
${f\in\cF_3(\cD^{nk},\R^q)}$
and
${h\in\cF_3(\R^{q},\R)}$. 
Let ${\phi_{11},\ldots,\phi_{nk}}$ 
be i.i.d.\ random elements of $\cT$ independent of $\cX$, $\lambda(n,k)$ be defined as in \eqref{eqn:defn:lambda}, and moment terms $c_X, c_Z$ be defined as in \eqref{eqn:defn:moments}. Then, for any i.i.d.\,variables ${\bZ_1,\ldots,\bZ_{n}}$ in $\cD^k$ satisfying \eqref{eqn:defn:surrogates},
\begin{equation*}
  \bigl|
  \mean h(f(\Phi\cX))
  -
  \mean h(f(\bZ_1,\ldots,\bZ_n))
  \bigr|
  \;\leq\;
  nk^{3/2}\lambda(n,k)(c_X+c_Z)\;.
\end{equation*}
\end{theorem} 

Hence if $ nk^{3/2}\lambda(n,k)(c_X+c_Z)\rightarrow 0$, this means that the value $\mean h(f(\Phi\cX))$ only asymptotically depends on the mean and variance of the augmented samples. We will see that this implies that the distribution of the augmented estimator is universal. Note that if we choose the test function $h$ appropriately we can for example establish:

\begin{corollary} (Convergence of variance) \label{cor:variance:convergence} Assume the conditions of Theorem \ref{thm:main}. Then
\begin{align*}
    n \big\| \Var[f(\Phi\cX)] - \Var[f(\bZ_1, \ldots, \bZ_n)]  \big\| \;\leq\; 6 n^2 k^{3/2} ( \alpha_0 \alpha_3 + \alpha_1 \alpha_2 ) (c_X + c_Z)\;.
\end{align*}
\end{corollary}

Note that similar derivation can be made for many statistics of $f(\Phi\cX)$ such as the expectation. To compare the distributions on $\mathbb{R}^q$, we use all functions $h$ in a suitable class $\cH$ of test functions.
In the context of the noise stability definitions above, we choose
\begin{equation*}
\cH \coloneqq \{ h: \R^q \rightarrow \R \;|\; h \text{ is thrice-differentiable with } \gamma_1(h), \gamma_2(h), \gamma_3(h) \leq 1 \}\;.
\end{equation*}
The distributions of two random elements ${\bX}$ and $\bY$ of $\mathbb{R}^q$ are then compared by defining
\begin{equation*}
    d_{\cH}( \bX, \bY ) \coloneqq \msup_{h \in \cH} | \mean h(\bX) - \mean h(\bY) |\;,
\end{equation*}
that is, the integral probability metric determined by $\cH$. We note that it metrizes weak convergence.
\begin{lemma} ($d_{\cH}$ metrizes weak convergence)
  \label{lem:d_H}
  Let $\bY$ and ${\bY_1,\bY_2,\ldots}$ be random variables in
  $\R^q$ with ${q\in\mathbb{N}}$ fixed. Then
  ${d_{\cH}(\bY_n,\bY)\rightarrow 0}$ implies weak convergence
  $\smash{\bY_n\overset{d}{\rightarrow} \bY}$.
\end{lemma}
This metric is similar to the generalized Dudley distance of \cite{grigorevskii1976some}, but unlike the latter,
$d_{\cH}$ controls all three derivatives simultaneously. 
\cref{appendix:d_H:compare} compares $d_{\cH}$ to other probability metrics. 
Since ${\mathcal{H}}$ is a subset of ${\mathcal{F}_3(\mathbb{R}^q,\mathbb{R})}$,
replacing $f$ with $\sqrt{n} f$ in Theorem \ref{thm:main} yields:

\begin{corollary} (Convergence in $d_{\cH}$)
\label{cor:d_H:convergence}
Under the conditions of Theorem \ref{thm:main}, 
\begin{align*}
    d_{\cH}( \sqrt{n} f(\Phi\cX), \sqrt{n} f(\bZ_1, \ldots, \bZ_n) ) \;\leq\; n^{3/2}k^{3/2} ( n \alpha_1^3  + 3 n^{1/2} \alpha_1 \alpha_2 + \alpha_3 )(c_X + c_Z) \;.
\end{align*}
\end{corollary}

Thus, Theorem \ref{thm:main} exactly characterizes the asymptotic
variance and distribution of the augmented estimate ${f(\Phi\cX)}$ by showing universality of its distribution,
as summarized in the next corollary.
That allows us, for example, to compute 
consistent quantiles
for $f(\Phi\cX)$.

\begin{corollary} \label{cor:sufficient:cond}  Fix $q$. Assume the
  conditions of Theorem \ref{thm:main} hold, and that the bounds in
  Corollary \ref{cor:variance:convergence} and
  \ref{cor:d_H:convergence} converge to zero as ${n \rightarrow\infty}$. Then \\[-1.4em]
\begin{align*}
    d_{\cH}(\sqrt{n} f(\Phi\cX), \sqrt{n} f(\bZ_1, \ldots, \bZ_n) ) &\rightarrow 0\;
    &\text{and}&&
    n \big\| \Var[f(\Phi\cX)] - \Var[f(\bZ_1, \ldots, \bZ_n)]  \big\| \rightarrow 0\;.
\end{align*}
\end{corollary}

The next lemma simplifies notation throughout---it shows that, if the scaling by $\sqrt{n}$ is dropped, one can still
quantify convergence of both $\mean[f(\Phi\cX)]$ and of the centered estimate. Results can hence be stated without explicitly
centering terms.
\begin{lemma} \label{lem:d_H:centering} Let $\bX$ and $\bY$ be random variables in $\R^q$. Suppose $d_{\cH}(\bX,\bY) \leq \epsilon$ for some constant $\epsilon > 0$. Then $\| \mean \bX - \mean \bY \|  \leq q^{1/2} \epsilon$ and $d_{\cH}( \bX - \mean \bX, \bY - \mean \bY) \leq (1+q^{1/2}) \epsilon$.
\end{lemma}

\begin{remark}(Comments on the main theorem) \label{remark:thm:main}
\emph{(i) Gaussian surrogates}. In most of our examples, the data domain $\cD$ is the entire space $\R^d$. If
so, one may choose the $\bZ_i$ as Gaussian vectors matching the first two moments of $\Phi_1\bX_1$.
\\[.2em]
\emph{(ii) Generalizations}. The proof techniques still apply if some conditions are relaxed.
Generalized results are given in
\cref{appendix:var:cor}, and appear in some of
the applications we study below. For example,
$\bZ_i$ may be matrix-valued (e.g.\ in ridge regression, in Proposition \ref{prop:ridge:regression}).
The range and domain of $\phi_{ij}$ may not
agree (
Theorem 13), and the $\phi_{ij}$ do not have to be
i.i.d. We may also permit $q$ to grow with $n$ and $k$. In \cref{appendix:repeated}, we also include results for the case where the same augmentations are reused across different data points.
\\[.2em]
\emph{(iii) Distributional invariance.}
A common assumption in machine learning is that the data distribution is invariant under $\cT$.
That means that, for all ${\phi\in\cT}$,
\begin{align*}
  \phi\bX_1
  \overset{d}{=} \bX_1
  \qquad\text{ or equivalently }\qquad
  \mean[ f(\bX_1) ] \;=\; \mean[ f(\phi\bX_1) ]
  \quad\text{ for all }f\in\mathbf{L}_1(\bX_1)\;. 
\end{align*}  
From a statistical learning perspective, this is one way to ensure that augmentation does not alter the limiting estimator, although the speed of convergence to that limit may differ. 
In light of \cref{thm:main}, invariance implies that the variance in
\eqref{eqn:defn:surrogates} can be replaced by
\begin{equation*}
  \Var \bZ_i =\bI_k \otimes \mean[ \Var[\phi_{11}\bX_1 | \phi_{11}]] + (\bone_{k \times k} - \bI_k) \otimes \mean[\Cov[\phi_{11}\bX_1, \phi_{12}\bX_1 | \phi_{11}, \phi_{12}]]\;.
\end{equation*}
Note the off-diagonal terms are now covariance matrices that are smaller than those in \eqref{eqn:defn:surrogates} in the Loewner partial order. 
\end{remark}

In conclusion, if the conditions of Theorem \ref{thm:main} hold and the bounds in
  Corollaries \ref{cor:variance:convergence} and
  \ref{cor:d_H:convergence} converge to zero, then the asymptotic distribution of $\sqrt{n}f(\Phi \cX)$ only depends on the mean and covariance of the augmented samples $(\Phi \cX)$. Hence, under general conditions, the effect of data augmentation on the learning rate only depends on how it affects the first few moments of the augmented variables, e.g.~how strong the correlation between the augmented samples is. This universality greatly simplifies the asymptotic analysis of data augmentation. 

%% file: article_sec4_average.tex
\section{Empirical averages and plug-in estimators} \label{sec:average:plugin}

The first class of estimators we consider are functions of the form 
\begin{equation}
  f(\bx_{11}, \ldots, \bx_{nk}) =  g\big( \mfrac{1}{nk} \tsum_{i \leq n, j \leq k} \bx_{ij} \big) \label{eqn:defn:fn:averages}
\end{equation}
for a smooth function $g$. The simplest is an empirical average, which we analyze first. The results we obtain for
such averages still hold if $f$ is approximately linear, in the sense that it can be approximated well by a
first-order Taylor expansion. The risk of an $M$-estimator is an example. The behavior changes if $f$ is non-linear, which
is illustrated by an example in \cref{sec:nonlinear:average}.

\subsection{Comparing limiting variances} \label{sec:no:augmentation}

A natural measure of the effect of data augmentation on the convergence rate is the variance ratio comparing estimates obtained with and without
augmentation. To define a valid baseline for estimates without augmentation,
we must replicate each input vector $k$ times, 
since the number $k$ of augmentations determines the number of arguments of $f$, and also
enters in the upper bound. We denote such $k$-fold replicates by ${\tilde{\bX}_i} \coloneqq (\bX_i, \ldots, \bX_i) \in \cD^k$.
No augmentation then corresponds to the case where 
$\cT$ contains only the identity map of $\cD^{nk}$. By setting each $\phi_{ij}$
to identity in \cref{thm:main}, we can approximate the distribution of $f(\tilde \bX_1, \ldots, \tilde \bX_n)$
by that of $f(\tilde \bZ_1, \ldots, \tilde \bZ_n)$, where $\tilde \bZ_1, \ldots, \tilde \bZ_n$ are any i.i.d.~variables in $\cD^k$ satisfying 
\begin{align} \label{eqn:defn:surrogates:no_aug}
  \mean \bZ_i
  \;&=\;
  \mathbf{1}_{k\times 1}\otimes\mean \bX_1
  \;
  &\text{and}&&
  \Var \bZ_i
  \;&=\;
  \mathbf{1}_{k\times k} \otimes \Var \bX_1\;,
\end{align}
and substituting into \cref{thm:main} shows
\begin{equation*}
  \bigl|
  \mean h(f(\tilde{\bX}_1,\ldots,\tilde{\bX}_n))
  -
  \mean h(f(\tilde\bZ_1,\ldots,\tilde\bZ_n))
  \bigr|
  \;\leq\;
  nk^{3/2}\lambda(n,k)(c_{\tilde X} + c_{\tilde Z})\;. \tagaligneq \label{eq:no_aug}
\end{equation*}
The effect of augmentation versus no
augmentation can now be compared by the ratio
\begin{equation}
    \vartheta(f) \coloneqq 
    \sqrt{\| \Var f(\tilde{\bZ}_1, \ldots, \tilde{\bZ}_n) \|\,/\,\| \Var f(\bZ_1, \ldots, \bZ_n) \|}\;.
    \label{eqn:rate_of_shrinkage}
\end{equation}
If $\vartheta(f) > 1$, augmentation is beneficial in the sense that it speeds up convergence of the estimator (though it may or may not introduce a bias).
If $\vartheta(f) < 1$, it is detrimental, which is possible even if invariance holds.\\[.5em]
\noindent\textbf{Notation}. We write $\Phi\cX$ for
augmented data, and $\cZ\coloneqq\{\bZ_1,\ldots,\bZ_n\}$ for
i.i.d.\,surrogates satisfying \eqref{eqn:defn:surrogates}. 
${\tilde\cX=(\tilde\bX_1,\ldots,\tilde\bX_n)}$ denotes the unaugmented, replicated data 
defined above, and
$\tilde\cZ\coloneqq\{\tilde\bZ_1,\ldots,\tilde\bZ_n\}$ surrogates 
satisfying \eqref{eqn:defn:surrogates:no_aug}. We
refer to $\cZ$ and $\tilde \cZ$ as Gaussian if $\bZ_1, \ldots, \bZ_n$ and
$\tilde\bZ_1, \ldots, \tilde\bZ_n$ are Gaussian vectors in $\R^d$.

\subsection{Empirical averages}
\label{sec:averages}

The arguably most common choice of $f$ is an empirical
average---augmentation is often used with empirical risk minimization,
and the empirical risk is such an average. By Remark \ref{remark:thm:main}(ii) above, empirical estimates of
gradients can also be represented as empirical averages.
An augmented empirical average is of the form
\begin{equation}
  \label {empirical:average}
  f(\bx_{11},\ldots,\bx_{nk}) \coloneqq\mfrac{1}{nk} {\textstyle\sum_{i=1}^n
    \sum_{j=1}^k} \bx_{ij}\;,
\end{equation}
where ${\cD=\R^d}$, and $d$ and $k$ are fixed.
Specializing \cref{thm:main} yields:
\begin{proposition} (Augmenting averages)
  \label{prop:empirical_average}
  Require that $\mean \|\bX_1\|^6$ and $\mean \|\phi_{11}\bX_1\|^6$
  are finite. Let $\cZ$ and $\tilde \cZ$ be Gaussian.
  Then $f$ as above satisfies
  \begin{align*}
    d_{\cH}(\sqrt{n}f(\Phi\cX), \sqrt{n} f(\cZ)) \rightarrow 0
    \quad\text{ and }\quad
    d_{\cH}(\sqrt{n}f(\tilde
    \cX), \sqrt{n} f(\tilde \cZ)) \rightarrow 0\quad\text{ as }n\rightarrow\infty\;.
  \end{align*}\end{proposition} 
  The Gaussian surrogates can be translated into asymptotic quantiles as follows:
  The ratio $\vartheta$ of standard deviations here takes the form
  \begin{align*}
    \vartheta \;=\; \sqrt{ 
      \Big( \mfrac{1}{n} \Var[\bX_1] \Big)
      \,\big /\, 
      \Big(  
        \mfrac{1}{nk}\Var[\phi_{11}\bX_1]
        +
        \mfrac{k-1}{nk}\Cov [\phi_{11} \bX_1, \phi_{12} \bX_1] 
      \Big) 
    } \;.
  \end{align*}
  To keep notation simple, assume ${d=1}$. To obtain $\alpha/2$-th asymptotic quantiles, for ${\alpha\in[0,1]}$,
  denote by $z_{\alpha/2}$ the $(1-{\alpha}/{2})$-percentile of a standard normal.
  Then the lower and upper asymptotic quantiles of $f(\Phi\cX)$ and $f(\tilde \cX)$ are given respectively by 
  \begin{equation*} 
    \mean[\phi_{11}\bX_1]\,\pm\,\mfrac{1}{\sqrt{\vartheta^2 n}}z_{\alpha/2} \sqrt{\Var [\bX_1]}
    \quad\text{ and }\quad
    \mean[\bX_1]\,\pm\,\mfrac{1}{\sqrt{n}}z_{\alpha/2} \sqrt{\Var [\bX_1]}\,\;.
  \end{equation*}
  For empirical averages, the quantiles can be inverted to obtain
  asymptotic ${(1-\alpha)}$-confidence intervals for
  $\mean[\phi_{11} \bX_1]$ and $\mean[\bX_1]$, given by
  \begin{equation*} 
    \Bigl[\, f(\Phi\cX)\,\pm\,\mfrac{1}{\sqrt{\vartheta^2 n}}z_{\alpha/2} \sqrt{\Var [\bX_1]}\,\Bigr] 
    \quad\text{ and }\quad
    \Bigl[\,f(\tilde \cX)\,\pm\,\mfrac{1}{\sqrt{n}}z_{\alpha/2} \sqrt{\Var [\bX_1]}\,\Bigr]
  \end{equation*} 

  \begin{remark}
    We note some implications of \cref{prop:empirical_average}:\\[.2em]
    (i) In terms of confidence region width, computing the empirical average by augmenting 
    $n$ observations is equivalent to averaging over an
    unaugmented data set of size ${\vartheta^2 n}$.\\[.2em]
    (ii) Augmentation is hence beneficial for
    empirical averages if
    ${\|\Var[\phi_{11}\bX_1]\|\leq\|\Var\,\bX_1\|}$. To see this,
    observe that augmentation is beneficial if
    $\vartheta \geq 1$, and that
\begin{equation}
\label{eq:decreasing:variance}
\|\Var f(\cZ)\| =
    \|\mfrac{1}{k}\Var[\phi_{11}\bX_1 ] +\mfrac{k-1}{k}\Cov [\phi_{11} \bX_1, \phi_{12} \bX_1]\|\leq\|\Var[\phi_{11}\bX_1]\|\;.
\end{equation}
(iii) If the data distribution is invariant, in the sense that
  $\phi_{11}\bX_1\overset{d}{=}\bX_1$, augmentation is always
  beneficial, since ${\Var \bX_1 = \Var[\phi_{11}\bX_1] \succeq \Cov [\phi_{11} \bX_1, \phi_{12} \bX_1]}$.
  \end{remark}

\subsection{Parametric plug-in estimators}    \label{sec:plugin}

Most of the observations for empirical averages still hold for plug-in estimators if the dimension is fixed, and
more generally for any approximately linear function of averages, such as the risk of an M-estimator. To see this, note that if we choose 
$g$ in \eqref{eqn:defn:fn:averages} as a sufficiently smooth function, $f$ can be approximated by a first-order Taylor expansion
\begin{equation}
    f^T(\bx_{11}, \ldots, \bx_{nk}) \,\coloneqq\, g(\mean[\phi_{11}\bX_1]) + \partial g(\mean[\phi_{11}\bX_1]) \big( \mfrac{1}{nk} \msum_{i\leq n,j\leq k} \bx_{ij} - \mean[\phi_{11}\bX_1] \big)\;. \label{eqn:defn:first_order_taylor}
\end{equation}
The key observation is that the only random contribution to $f^T$ behaves
exactly like an empirical average. 
Lemma 19 in
the appendix shows that
\begin{align} \label{eqn:plugin}
    &
    d_{\cH}(\sqrt{n} f(\Phi \cX), \sqrt{n} f^T(\cZ) ) \rightarrow 0
    &\;\;\text{ and }\;\;&
    n \, \big(  \|\Var[f(\Phi \cX)]\| - \|\Var[f^{T}(\cZ)] \| \big) \rightarrow 0\;,
\end{align}
provided that $g$ is sufficiently well-behaved and noise stability
holds. That is even true if $d$ grows (not too rapidly) with $n$.

\vspace{.5em}

The variance of $f^T$ now depends additionally on
${\partial g(\mean[\phi_{11}\bX_1])}$. If the data distribution is not invariant
under augmentation, it is possible that ${\|\partial g(\mean[\phi_{11}\bX_1])\|>\|\partial
  g(\mean \bX_1)\|}$.
If so, the overall variance may increase even if
augmentation decreases the variance of the empirical average.
If invariance holds, augmentation reduces variance,
as observed by \cite{chen2020group}. \\[.5em]

\subsection{Non-linear estimators}  \label{sec:nonlinear:average}

\begin{figure}[t]
\centering
\begin{tikzpicture}
\node[inner sep=0pt] at (0,0.03)
    {\includegraphics[width=.51\textwidth]{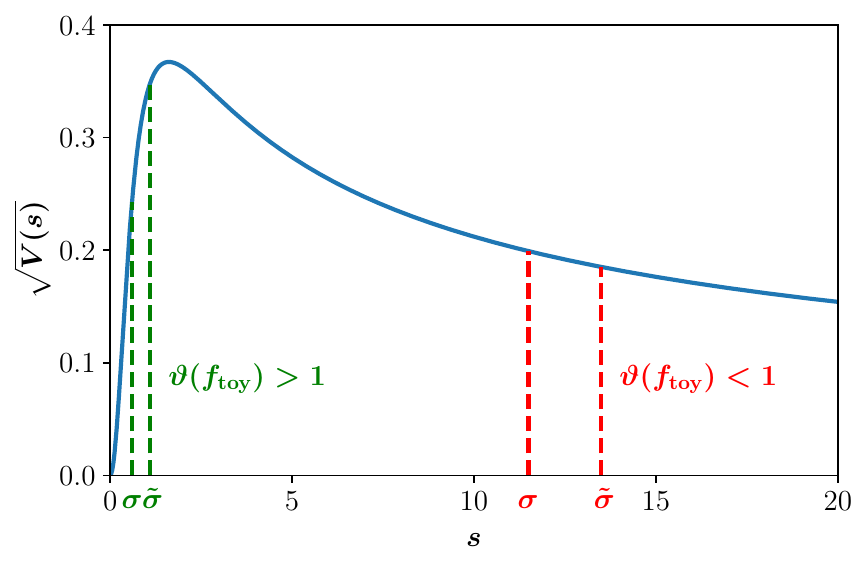}};
\node[inner sep=0pt] at (7.5,0)
    {\includegraphics[width=.5\textwidth]{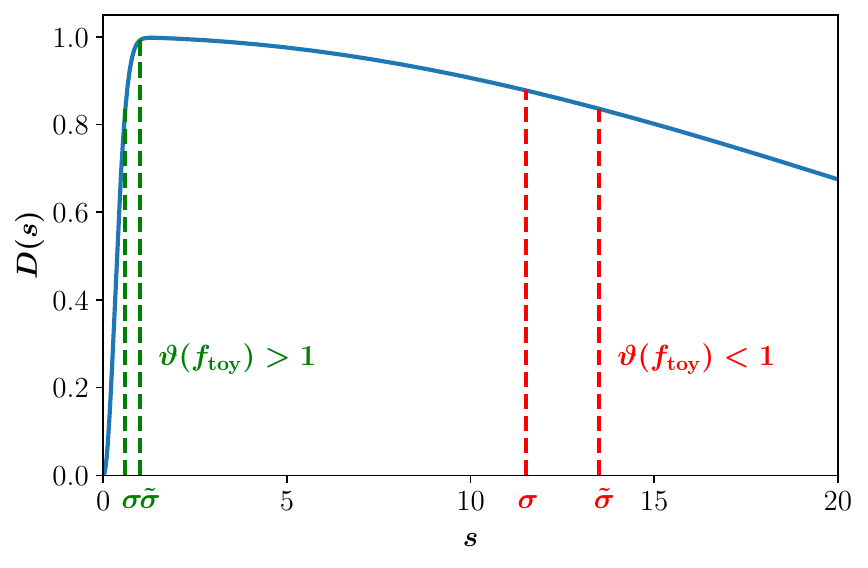}};
\end{tikzpicture}
\centering
\caption{
\emph{Left}: The standard 
deviation $\sqrt{V(s)} \coloneqq \sqrt{\Var[f_{\rm toy}(\cZ)]} = \sqrt{\Var[ g_{\rm toy}( s \xi + \mean[\bX_1])]}$ as a function of $s$. \emph{Right}: The difference $D(s)$ between the $0.025$-th and the $0.975$-th quantiles for $g_{\rm toy}( s \xi + \mean[\bX_1])$ as a function of $s$.
The functions are calculated analytically in 
Proposition 20.
Since neither is monotonic, the parameter space contains regions where data augmentation is
beneficial (green example), and where it is detrimental (red example).
Notably, ${\vartheta(f)<1}$ is possible even if $\sigma$, standard deviation of the augmented average, is smaller than $\tilde{\sigma}$, standard deviation of the unaugmented average.
}
\label{fig:exp_neg_chi_squared_1D}
\end{figure}

We have seen above that, in the linear case, invariance guarantees that augmentation does not increase estimator variance.
If the estimator \eqref{eqn:defn:fn:averages} is not well-approximated by the linearization \eqref{eqn:defn:first_order_taylor},
that need not be true, which can be seen as follows. \cref{thm:main} shows that
\begin{align*}
  \Var[f(\Phi\cX)]
  \;\approx\;
    \Var\Big[ g \Big( \mfrac{\sqrt{ \Var[ \bX_1 ] }}{\sqrt{\vartheta^2 n}} \xi \,+\, \mean[\phi_{11}\bX_1]  \Big) \Big]
    \quad\text{ for }
    \xi \sim \cN(\bzero,\bI_d)\;.
\end{align*}
The same holds, with ${\vartheta=1}$, for the unaugmented variance.
Assume for simplicity that $d=1$ and invariance holds, which implies $\mean[\phi_{11}\bX_1] = \mean[\bX_1]$ and $\vartheta \geq 1$. By a well-known result characterizing the variance of a function of a Gaussian (Proposition 3.1 of \cite{cacoullos1982upper}), we have 
\begin{align*}
  \sigma^2 
  \mean\big[ \partial g( \sigma \xi + \mean[\bX_1 ]) \big]^2
  \leq
  \Var\big[ g\big( \sigma \xi + \mean[\bX_1 ] \big) \big] 
  \leq
  \sigma^2 
  \mean\big[ \partial g( \sigma \xi + \mean[\bX_1 ])^2 \big] 
\end{align*}
for any ${\sigma>0}$.
When $g$ is non-linear, $\partial g$ is not constant, and $\Var\big[ g\big( \sigma \xi + \mean[\bX_1 ] \big) \big] $ is not necessarily monotonic in $\sigma$. Thus, in the non-linear case, invariance of the data distribution does not imply variance reduction.
\cref{fig:exp_neg_chi_squared_1D} illustrates the variance and quantiles for a highly non-linear toy statistic, defined as
\begin{equation} \label{eqn:defn:exp:neg:chi:sq:stat}
  f_{\rm toy}(x_{11},\ldots,x_{nk}) 
  \;\coloneqq\; 
  g_{\rm toy}\Big( \mfrac{1}{nk} \msum_{ij} x_{ij} \Big)
  \;=\; 
  \exp\Big( - \Big( \mfrac{1}{\sqrt{n}k} \msum_{ij} x_{ij} \Big)^2 \Big)\;.
\end{equation} 
In both plots of \cref{fig:exp_neg_chi_squared_1D}, the behavior of augmentation changes from one region of parameter space to another. See
\cref{appendix:toy:results} for formal statements and simulation results.

%% file: article_sec5_ridge.tex
\section{Ridge regression}  \label{sectn:ridge_regression}

This section studies the effect of augmentation on ridge regression in moderate dimensions.
In light of the discussion in the previous section, this is an example of an estimator that is
not approximately linear, which complicates the effect of augmentation on its variance.

In a regression problem, each data point ${\bX_i\coloneqq(\bV_i, \bY_i)}$
consists of a covariate ${\bV_i}$ with values in ${\R^d}$, and a response
${\bY_i}$ in ${\R^b}$. We hence consider pairs of transformations
${(\pi_{ij}, \tau_{ij})}$ as augmentation, where $\pi_{ij}$ acts on
$\bV_i$ and $\tau_{ij}$ acts on $\bY_i$.
A transformed data point is then of the form
${\phi_{ij}\bx_i \coloneqq ((\pi_{ij} \bv_i) (\pi_{ij}
  \bv_{i})^\top,(\pi_{ij} \bv_{i}) (\tau_{ij} \by_{i})^\top)}$,
and hence an element of ${\cD \coloneqq \M^d \times \R^{d \times b}}$, where
$\M^d$ denotes the set of positive semi-definite ${d \times d}$ matrices.
For a fixed $\lambda >0$, the ridge regression estimator on augmented data is therefore
\begin{equation}
    \hat{B}(\phi_{11}\bx_1,\ldots,\phi_{nk}\bx_n) \;\coloneqq\; \Big( \mfrac{1}{nk} \msum_{ij} (\pi_{ij} \bv_{i})(\pi_{ij} \bv_{i})^\top + \lambda \bI_{d} \Big)^{-1} \mfrac{1}{nk} \msum_{ij} (\pi_{ij} \bv_{i}) (\tau_{ij} \by_{i})^\top\;. \label{eq:ridge_est:defn}
\end{equation}
It takes values in $\mathbb{R}^{d\times b}$, and its risk is $R(\hat{B})\coloneqq\mean[\|\bY_{new} - \hat{B}^\top \bV_{new} \|^2_2 \,|\, \hat B]$.

The next result completely characterizes the asymptotic distribution of the risk of a ridge estimator in a moderate-dimensional regime, for any choice of augmentation. In particular, one can study the effect of augmentation on the speed of convergence of the risk to its infinite-data limit,
\begin{proposition} \label{prop:ridge:regression} Suppose
  $\mmax_{l \leq d} \max\{ (\pi_{11} \bV_{1})_l, (\tau_{11}
    \bY_{1})_l \} $ is almost surely bounded by $C d^{-1/2} (\log d)^c$ for some absolute constants $C, c > 0$ and that
  $b=O(d)$. Then there exist i.i.d. surrogate variables
  ${\bZ_1,\ldots,\bZ_n}$ such that
\begin{align*}
    d_{\cH}( \sqrt{n} R^{\Phi\cX}, \sqrt{n} R^Z ) = O(n^{-1/2} d^9)
    \,\,\text{ and }\,\,
    n ( \Var[R^{\Phi\cX}] - \Var[R^Z] ) = O(n^{-1} d^7 (\log d)^{18c})\;,
\end{align*}
where ${R^{\Phi\cX} \coloneqq R(\hat B(\Phi\cX))}$ is the risk of the
estimator trained on augmented data, and $R^Z \coloneqq R(\hat B(\cZ))$ the risk with surrogate variables.
\end{proposition}

\begin{figure}[t]
  \centering
  \begin{tikzpicture}
  \node[inner sep=0pt] at (0,0)
      {\includegraphics[width=.55\textwidth]{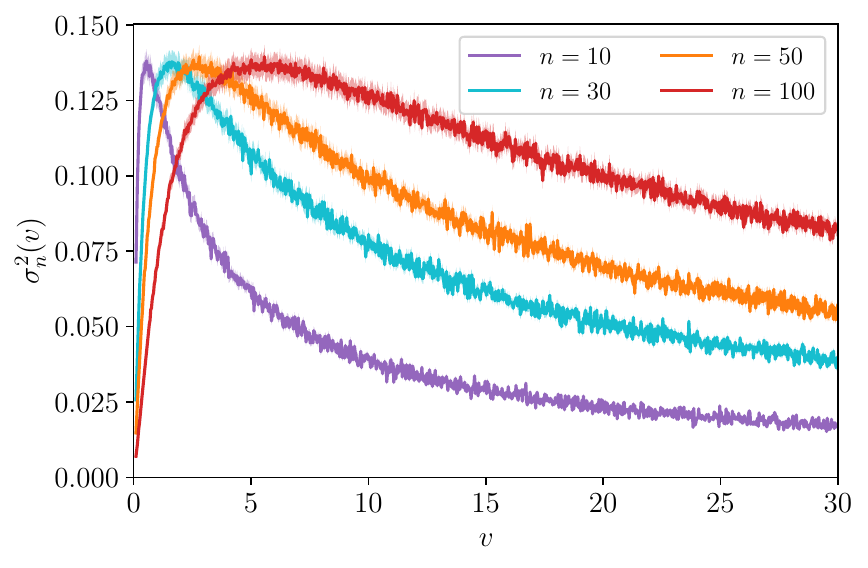}};
  \end{tikzpicture}
  \centering
  \caption{A simple ridge regression example, where variance of the risk is not monotonic in data variance despite invariance.  
  Variance of $R^Z$ in
  Lemma \ref{lem:ridge:toy} is plotted as a function of the augmented covariance $\nu \coloneqq \Cov[(\pi_{11}\bV_1)^2, (\pi_{12}\bV_1)^2]$ for ${\lambda=0.1}$ and ${\mean[\bV_1^2]=0.1}$. As no closed-form formula is available, the plot is generated by a simulation over 10k random seeds.
   }  
  \label{fig:regression:toy}
  \end{figure}

In this case, the surrogate variables $\bZ_i$ are 
random elements of $(\M^d \times \R^{d \times b})^k$, whose first two
moments match those of the augmented data. 
As part of the proof of the proposition,
we also obtain convergence rates for the estimator $\hat B(\Phi\cX)$
(in addition to the rate for its risk above); see 
Lemma 36 
in the appendix.\\[.5em]
\textbf{A detailed analysis of a simple illustrative example}.
We consider a special case in more detail, which illustrates that
unexpected effects of augmentation can occur even in very simple
models: Assume that
\begin{equation} \label{eqn:ridge:toy:model}
  \bY_i \coloneqq \bV_i + \varepsilon_i
  \quad\text{ where }\quad
  \bV_i \stackrel{i.i.d.}{\sim} \cN( \mu \bone_d, \Sigma)
  \quad\text{ and }\quad
  \varepsilon_i \stackrel{i.i.d.}{\sim} \cN( \bzero, c^2 \bI_d)\;.
\end{equation}
This is the setup used in \cref{fig:intro_plot}, where
$d=2$. Detrimental effects of augmentation can occur even in one
dimension, though. To clarify that, we first show the following:

\begin{lemma} \label{lem:ridge:toy}
  Consider the one-dimensional case ($d=1$), with $c=0$ and $\tau_{ij} = \pi_{ij}$. Assume that the
  augmentation leaves the covariate
  distribution invariant, $\smash{\pi_{ij}\bV_i \overset{d}{=}\bV_i}$.
  Write the covariance ${v_{\pi}=\Cov[ (\pi_{11}\bV_1)^2,\, (\pi_{12}\bV_1)^2]}$, and
  generate surrogate variables by drawing
  \begin{equation*}
    \bZ_{111}, \ldots, \bZ_{n11} \;\overset{i.i.d.}{\sim}\; \Gamma\Big( \mfrac{ \mean[\bV_1^2]^2}{v_\pi}, \mfrac{\mean[\bV_1^2]}{v_\pi} \Big)
  \end{equation*}
  and setting ${\bZ_{ijl}\coloneqq\bZ_{i11}}$, for all ${j\leq k}$ and
  ${l=1,2}$. Then
  \begin{equation*}
    d_{\cH}(\sqrt{n} R^{\Phi\cX}, \sqrt{n} R^Z) \rightarrow 0
    \quad\text{ and }\quad
    n(\Var[R^{\Phi\cX}] - \Var[R^Z]) \rightarrow 0
    \qquad\text{ as }n, k \rightarrow \infty\;.
  \end{equation*} 
  Moreover, denoting the Gamma random variable $X_n(v) \sim \Gamma( \frac{n \mean[\bV_1^2]^2}{v}, \frac{n\mean[\bV_1^2]}{v})$, we have 
  \begin{align*}
      \Var[R^Z] \;=\; \sigma_n^2(v_{\pi}) \;=\; \mean[\bV_1^2]^2  \lambda^2 \Var\Big[ \mfrac{1}{(X_n(v_\pi) + \lambda)^2} \Big] \;,
  \end{align*}
  where $\sigma_n$ is a real-valued function that does not depend on the number of augmentations $k$, or on
  the law of the augmentations $\pi_{ij}$.
\end{lemma}

\begin{figure}[t]
  \centering
  \begin{tikzpicture}
  \node[inner sep=0pt] at (-3.9,0)
      {\includegraphics[width=.5\textwidth]{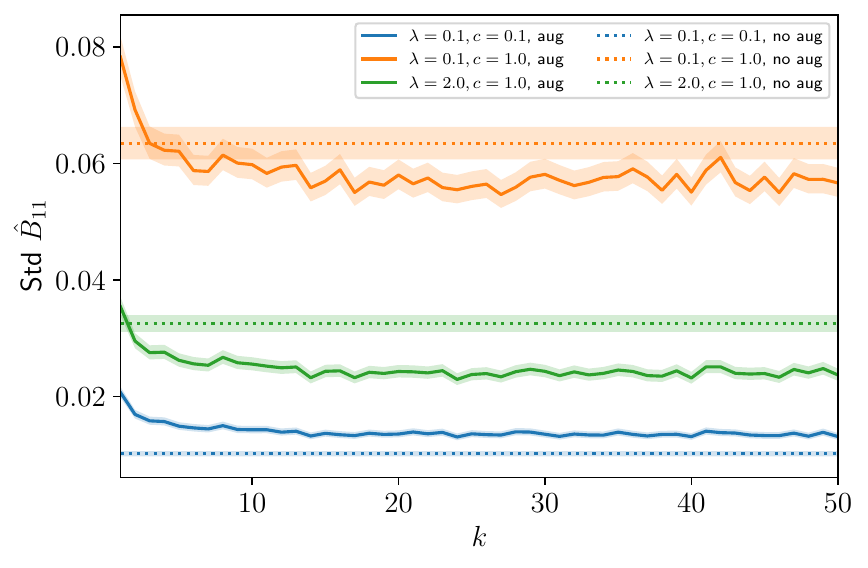}};
  \node[inner sep=0pt] at (3.5,0)
      {\includegraphics[width=.5\textwidth]{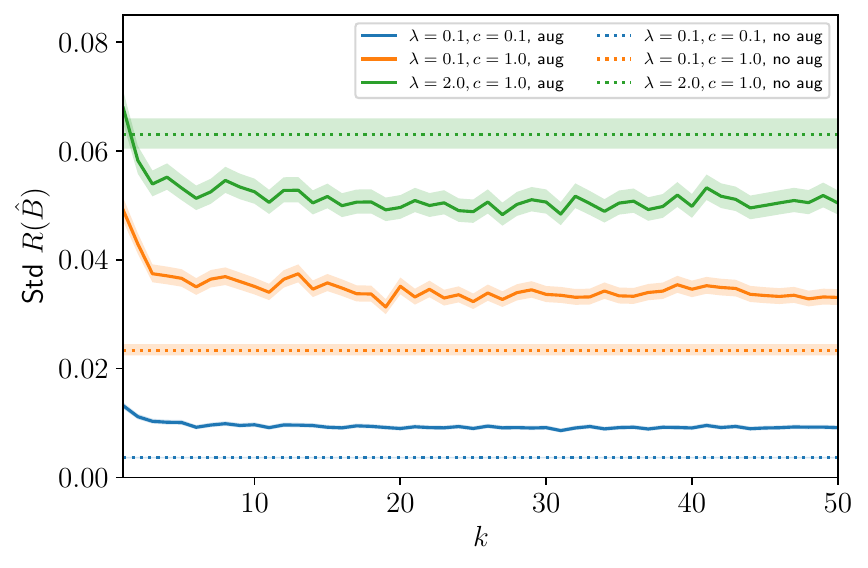}};
  \node[inner sep=0pt] at (-3.9,-4.8)
      {\includegraphics[width=.5\textwidth]{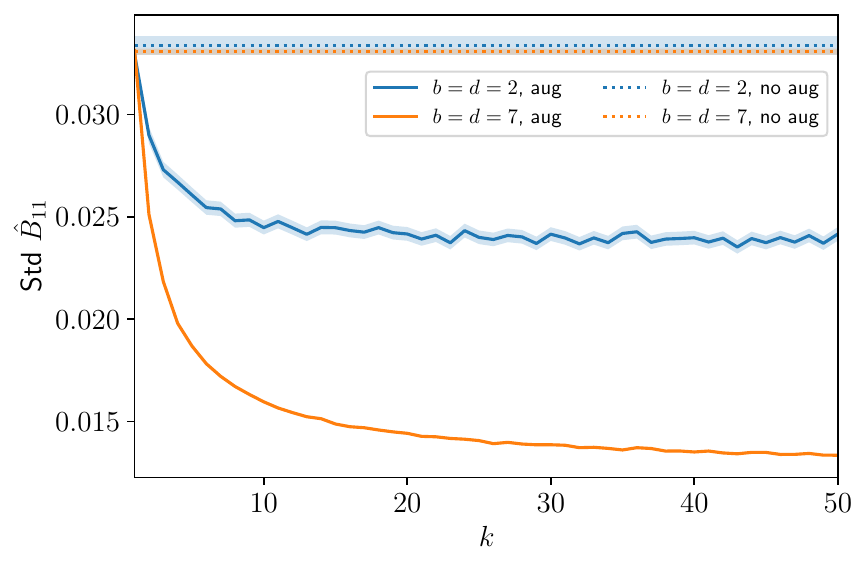}};
  \node[inner sep=0pt] at (3.5,-4.8)
      {\includegraphics[width=.5\textwidth]{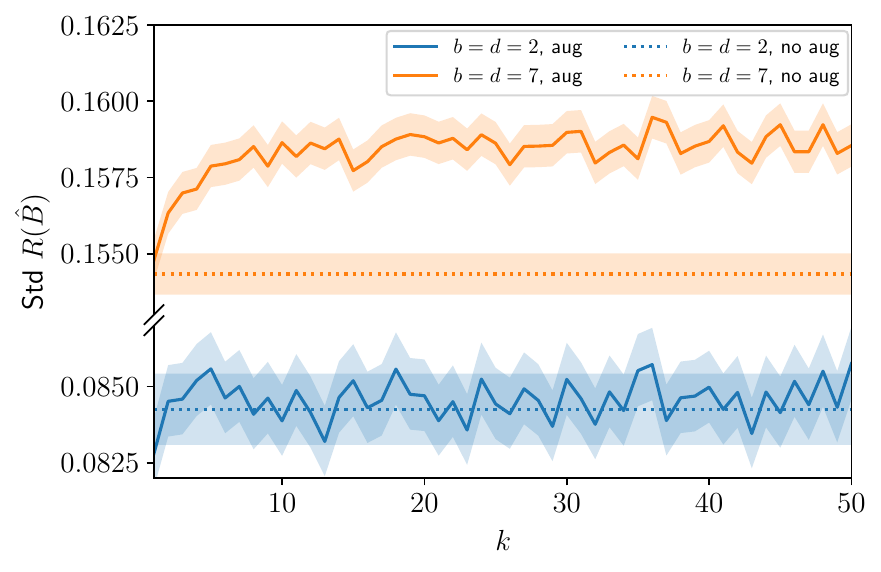}};
  \end{tikzpicture}
  \vspace{-1.8em}
  \centering
  \caption{Augmentation can decrease the variance of an estimator, but at the same time increase the variance of its risk:
    Shown are simulations for ridge regression under \eqref{eqn:ridge:toy:model} with $\mu=0$ and varying $k$. The augmentations on each pair of $\bV_{ij}$ and $\bY_{ij}$ are set to be the same, i.e. $\pi_{ij} = \tau_{ij}$. For random cropping, $n=200$ and
    $\Sigma = \begin{psmallmatrix}
      1 & 0.5 \\
      0.5 & 1
    \end{psmallmatrix}$. For uniform rotations, $n=50$ and $\Sigma = \bI_d$, $c=2$, $\lambda=9$.  \emph{Top Left.} Standard deviation of $(\hat B(\Phi\cX))_{11}$, first coordinate of ridge regression estimate under random cropping. \emph{Top Right.} Standard deviation of $R(\hat B(\Phi\cX))$ under random cropping. \emph{Bottom Left.} $\text{Std }(\hat B(\Phi\cX))_{11}$ under uniform rotations. \emph{Bottom Right.} $\text{Std } R(\hat B(\Phi\cX))$ under uniform rotations.} 
  \label{fig:regression}
\end{figure}

Note the surrogate distribution
can be determined explicitly,
and is non-Gaussian. The main object of interest is the variance
$\sigma_n^2$ of the risk of an augmented ridge regressor. 
For any choice of augmentation, the augmented covariance $\nu_\pi$ is always bounded from above by the unaugmented variance $\Var[(\bV_1)^2]$. This does not generally imply the the augmented ridge regressor is a better estimator---the
simulation in \cref{fig:regression:toy} shows that $\sigma_n$ is non-monotonic, that is, even though augmentation reduces $\nu_\pi$, it may increase the variance of the risk.

\begin{remark}(Details on simulations)
  \label{remark:ridge}
  (i) The simulation in \cref{fig:regression} uses
  the model \eqref{eqn:ridge:toy:model} and two forms of augmentation are both adapted from image analysis:
  \\[.2em]
  (a) Random rotations. We represent the elements of the size-$d$ cyclic group by matrices $C_1, \ldots, C_d$,
  generate random transformations
      \begin{align*}
          \phi_{ij} \;=\; \pi_{ij} \overset{i.i.d.}{\sim} \text{Uniform} \{ C_1 , \ldots, C_d  \}\;,
      \end{align*}
      and set ${\phi_{ij}\bx_i \coloneqq ((\pi_{ij} \bv_i) (\pi_{ij}
        \bv_{i})^\top,(\pi_{ij} \bv_{i}) (\tau_{ij} \by_{i})^\top)}$, i.e.\ we cycle through the
      $d$ coordinates of $\bY_i$ and $\bV_i$ simultaneously. The invariance
      $\smash{(\phi_{11}\bV_1,\phi_{11}\bY_1)\overset{d}{=}(\bV_1,\bY_1)}$ holds.
      \\[.2em]
      (b) Random cropping for $d=2$, where 
      a uniformly chosen coordinate of both $\bY_i$ and $\bV_i$ is set to $0$, i.e.~we have
    \begin{align*}
        \phi_{ij} \;=\; \pi_{ij} \overset{i.i.d.}{\sim} \text{Uniform} \{ C_1 M, \ldots, C_d M \}
        \;\; 
        \text{ where }
        M \;\coloneqq\; \begin{psmallmatrix}
        0 & & & \\ 
        & 1 & & \\
        & & \ddots & \\
        & & & 1
        \end{psmallmatrix}\;.
    \end{align*}
 \\[.2em]
(ii) We can now specify the setting used in Figure
  \ref{fig:intro_plot} in the introduction: It shows the empirical
  average function and the ridge regression estimate computed on the
  random cropping setup in \cref{fig:regression}, for $k=50$ and
  $\lambda=c=0.1$.
\end{remark}

%% file: article_sec6_ridgeless.tex
\section{Limiting risk of a ridgeless regressor in high dimensions} \label{sec:ridgeless}

We next consider the effect of data augmentation on the limiting risk of a ridgeless regressor in high dimensions.
Without augmentation, such regressors are known to exhibit a double-descent phenomenon \citep{hastie2022surprises}.
We show that augmentations can shift the double-descent peak of the risk curve, depending on the number of augmentations (see \cref{fig:intro_dd} in the introduction). Such a shift has been observed empirically by \cite{dhifallah2021inherent}.

\vspace{.5em}

In \Cref{sec:triple:descent:oracle,sec:triple:descent:two:stage}, we first consider the linear model where the univariate response variable $Y_i$ is related to the covariate $\bV_i$ in $\R^d$~by
\begin{align*}
    Y_i \;=\; \bV_i^\top \beta + \epsilon_i \;\; \text{ for } i = 1 ,\ldots, n \,, \tagaligneq \label{eq:model:linear}
\end{align*}
where the variables $\bV_i$ are i.i.d.~mean-zero random (not necessarily Gaussian) vectors,
and the noise variables $\epsilon_i$ are i.i.d.~mean-zero with ${\Var[\epsilon_i] = \sigma_\epsilon^2}$ and a bounded fourth moment.
The dimension $d$ grows linearly with $n$, and the signal $\beta$ and noise variance are assumed non-random with
${\|\beta\| =\Theta(1)}$ and ${\sigma_\epsilon^2=\Theta(1)}$. Following standard assumptions in random matrix theory and for simplicity, we assume the following on the covariates:

\begin{assumption}\label{assumption:covariate} (i) $\bV_i$ has independent coordinates $(V_{il})_{l \leq d}$; (ii) $\mean[V_{il}^3]=0$ and $\mean[V_{il}^4]=3\Var[V_{il}]^2$, i.e.~the first four moments of $V_{il}$ match those of its Gaussian surrogate.
\end{assumption}

\Cref{assumption:covariate}(i) can be relaxed to dependent coordinates; we defer this generalization to \Cref{sec:network}. For \Cref{assumption:covariate}(ii),
a similar assumption was used in \cite{tao2011random} for applying Lindeberg's technique to obtain universality of eigenvalue statistics of large matrices. We expect that the fourth moment condition can be replaced by a sub-exponential tail in view of known results on universality of covariance matrices, but this may require additional proof techniques involving the Dyson Brownian motion (see e.g.~Theorem 5.1 and the subsequent discussion of \cite{pillai2014universality}) and we do not pursue it here. Due to this assumption, we also use a small class of test functions:
\begin{equation*}
\tilde \cH \coloneqq \{ h: \R^q \rightarrow \R \;|\; h \text{ is six-times differentiable with } \gamma_1(h), \ldots, \gamma_6(h) \leq 1 \}\;,
\end{equation*}
which also characterizes weak convergence by a similar proof as \Cref{lem:d_H}. We denote the corresponding integral probability metric as $d_{\tilde \cH}$ and also denote $d_P$ as the Lévy–Prokhorov metric (see \eqref{eq:defn:prokhorov} in \Cref{appendix:auxiliary} for the definition).


\subsection{Double descent shift under oracle augmentation} \label{sec:triple:descent:oracle}

We first consider an oracle setup, where $\beta$ is assumed known. This is a theoretical device, but
we will see that it is informative. The setup is motivated by the fact that, once we have chosen transformations
$\pi_{ij}$ to augment the covariates $\bV_i$, we must also specify a reasonable way to augment the responses $Y_i$.
Since the covariates and responses are related via $\beta$, a known value of $\beta$ allows us to ``pass'' 
transformations from the covariates to the responses according to the model, by defining
\begin{align*}
  \tau_{ij}^{(\rm ora)} Y_i 
  \;\coloneqq\; 
  Y_i + \big( \pi_{ij} \bV_i - \bV_i \big)^\top \beta
  \;=\; 
  (\pi_{ij} \bV_i)^\top \beta 
  +
  \epsilon_i\;.
\end{align*}
If invariance holds for the covariates, it extends to responses,
\begin{equation}
  \label{eq:invariance:double:descent}
  \pi_{ij} \bV_i \overset{d}{=} \bV_i
  \qquad\Longleftrightarrow\qquad
  (\pi_{ij} \bV_i, \tau^{(\rm ora)}_{ij} Y_i) \overset{d}{=} (\bV_i, Y_i)\;.
\end{equation}
The augmented estimator is then 
\begin{align*}
    \hat{\beta}_{\lambda}^{(\rm ora)}
    \;\coloneqq&\
    \Big( \mfrac{1}{nk} \msum_{ij} (\pi_{ij} \bV_{i})(\pi_{ij} \bV_{i})^\top + \lambda \bI_d \Big)^\dagger \mfrac{1}{nk} \msum_{ij} (\pi_{ij} \bV_{i}) \, \tau^{(\rm ora)}_{ij} Y_i
    \;. \tagaligneq \label{eq:oracle:defn}
\end{align*}
This is a ridge estimator for ${\lambda>0}$, and ridgeless for ${\lambda=0}$.
Following \cite{hastie2022surprises}, we study the risk 
\begin{align*}
  \hat L^{(\rm ora)}_{\lambda} 
  \;\coloneqq&\;
  \mean\big[ 
   \big( 
     ( \hat \beta^{(\rm ora)}_{\lambda} - \beta)^\top \bV_{\rm new} 
   \big)^2  
   \,\big| \, \cX \big]
  \qquad\text{ for }\lambda\geq0
  \tagaligneq \label{eqn:interpolator:risk}
\end{align*}
where $\cX = \{ \pi_{ij} \bV_i \}_{i \leq n, j \leq k}$, in the asymptotic regime where 
\begin{align*}
  n,d \rightarrow \infty\,,
  \quad 
  d/n \rightarrow \gamma \in [0, \infty)\,,
  \quad
  d/(kn) \rightarrow \gamma' \in [0, \infty)\,,
  \quad
  k=o(n^{1/4})
  \,, \tagaligneq \label{eq:ridgeless:asymptotic:regime}
\end{align*}
and $k$ is allowed to be fixed or grow with $n$. 
In the unaugmented case, $\smash{\hat \beta^{(\rm ora)}_{\lambda}}$ and $\smash{\hat L^{(\rm ora)}_{\lambda}}$ are precisely the quantities
studied by \cite{hastie2022surprises}, who show that for $\lambda=0$, the risk reproduces the double-descent phenomenon also observed in neural networks. 

\vspace{.5em}

To illustrate the effect of augmentations in a simple model, we focus on the augmentation 

\begin{align*}
  \pi_{ij} \bV_i \;\coloneqq&\; \bV_i + \xi_{ij}  \;,
  \tagaligneq \label{eqn:noise:inject}
\end{align*}
where $(\xi_{ij})_{i,j}$ is a set of i.i.d.~mean-zero noise vectors, each having independent coordinates $(\xi_{ijl})_{l \leq d}$ with $\mean [\xi_{ijl}^3] = 0$ and $\mean[\xi_{ijl}^4] = 3 \Var[\xi_{ijl}]^2$. This form of randomization is also known as \emph{noise injection} in other contexts.

\vspace{.5em}

The main challenge in analyzing the risk is that the augmented risk depends on two strongly correlated high-dimensional sample covariance matrices,
\begin{align*}
  \bar \bX_1 \coloneqq \mfrac{1}{nk} \sum_{i \leq n} \sum_{j \leq k} ( \pi_{ij} \bV_i ) ( \pi_{ij} \bV_i  )^\top
  \;,
  \;\;
  \bar \bX_2 \coloneqq \mfrac{1}{n} \sum_{i \leq n} \bigg( \mfrac{1}{k} \sum_{j \leq k}  ( \pi_{ij} \bV_i )  \bigg) \bigg( \mfrac{1}{k} \sum_{j \leq k} ( \pi_{il} \bV_i  ) \bigg)^\top 
  \;.
\end{align*}
For comparison, $\bar \bX_1 = \bar \bX_2$ in the unaugmented case, and therefore existing analysis of double descent only involves one such matrix (e.g.~\cite{hastie2022surprises}). To address this, we consider the Gaussian surrogate vectors $\bZ_i$'s, where
\begin{align*}
    \mean[\bZ_i] \;=&\; \mean[\pi_{ij}\bV_i]
    &\text{ and }&&
    \Var[\bZ_i] \;=&\; \Var[\pi_{ij}\bV_i]
    \;.
\end{align*}
We denote the corresponding sample covariance matrices by
\begin{align*}
    \bar \bZ_1 \;\coloneqq\; \mfrac{1}{nk} \msum_{i=1}^n \msum_{j=1}^k \bZ_{ij} \bZ_{ij}^\top
    \;,
    \qquad 
    \bar \bZ_2 \;\coloneqq\; \mfrac{1}{n} \msum_{i=1}^n \Big( \mfrac{1}{k} \msum_{j=1}^k \bZ_{ij}  \Big) \Big( \mfrac{1}{k} \msum_{l=1}^k \bZ_{ij} \Big)^\top 
    \;.
\end{align*}
We can now express, for some function $f_\lambda: \R^{d \times d} \times \R^{d \times d} \rightarrow \R$ (see Appendix \ref{appendix:ridgeless:results} for the precise definition),
\begin{align*}
  \hat L_\lambda^{(\rm ora)} \;=\; f_\lambda( \bar \bX_1 \,,\, \bar \bX_2 )\;.
\end{align*}
Applying \cref{thm:main} allows us to approximate $\bar 
\bX_1$ and $\bar \bX_2$ by $\bar \bZ_1$ and $\bar \bZ_2$, whose spectral distributions are in the universality regime of compound Marchenko-Pastur laws \cite{marchenko1967distribution}. This can be used to investigate the limiting risk. The universality result requires several regularity assumptions, which we state next.

\begin{assumption} \label{assumption:ridgeless:universal} The following quantities are $O(1)$:
  \begin{align*}
      \max_{i \leq n, j \leq k, l \leq d} \| (\pi_{ij} \bV_i)_l \|_{L_{10}}
      \;,
      \quad 
      \big\| \| \bar \bX_2 \|_{op} \big\|_{L_{60}}
      \;,
      \quad 
      \big\| \| \bar \bZ_2 \|_{op} \big\|_{L_{60}}
      \;.
  \end{align*}    
\end{assumption}

\begin{assumption} \label{assumption:ridgeless:by:ridge} The following quantities are $O_{\gamma'}(1)$ with probability $1 - o_{\gamma'}(1)$:
  \begin{align*}
      &\;
      \| \bar \bX_1^\dagger \|_{op} 
      \;,\;
      \quad
      \| \bar \bZ_1^\dagger \|_{op} 
      \;,\;
      \quad 
      \| \bar \bX_2 \|_{op}
      \;,\;
      \quad 
      \| \bar \bZ_2 \|_{op} 
      \;,\;
      \\
      &\;
      \msum_{l=1}^d  \ind_{\{ \lambda_l(\bar \bX_1) = 0 \}} \big( v_l(\bar \bX_1)^\top \bar \bX_2 \, v_l(\bar \bX_1) \big)
      \;,\;
      \quad 
      \msum_{l=1}^d  \ind_{\{ \lambda_l(\bar \bZ_1) = 0 \}} \big( v_l(\bar \bZ_1)^\top \bar \bZ_2 \, v_l(\bar \bZ_1) \big) \;,
  \end{align*}
  where $(\lambda_l(A), v_l(A))$ denotes the $l$-th eigenvalue-eigenvector pair of a symmetric matrix $A \in \R^{d \times d}$, and $O_{\gamma'}(\argdot)$ and $o_{\gamma'}(\argdot)$ indicate that the bounding constants are allowed to depend on $\gamma'$.
\end{assumption}
  
\begin{proposition} \label{prop:ridgeless:universality} Fix $\lambda > 0$ and suppose Assumptions \ref{assumption:covariate} and \ref{assumption:ridgeless:universal} hold. Then under the asymptotic regime \eqref{eq:ridgeless:asymptotic:regime}, we have
  \begin{align*}
      d_{\tilde \cH}\big( f_\lambda(\bar \bX_1, \bar \bX_2) \,,\, f_\lambda(\bar \bZ_1 , \bar \bZ_2) \big)
      \;=\;
      O\Big( 
          \mfrac{k^2 \max\{1, \lambda^{-7}\}}{n^{1/2}} 
      \Big)\;.
  \end{align*}
  If additionally \cref{assumption:ridgeless:by:ridge} holds, then 
  \begin{align*}
    d_P\big( f_0(\bar \bX_1, \bar \bX_2) \,,\,  f_0(\bar \bZ_1, \bar \bZ_2)  \big) 
    \;=\; o(1)\;.
  \end{align*}
\end{proposition}

While the assumptions are complicated, 
Lemma 24 in the appendix verifies them for the isotropic Gaussian case. For simplicity, we now focus on the isotropic setup: For some fixed $\sigma_A > 0$, let
\begin{align*}
  \Var[\bV_1] 
  \;=&\; \bI_d
  &\text{ and }&&
  \Var[\xi_{ij}] 
  \;=&\; 
  \sigma_A^2 \bI_d\;.
  \tagaligneq \label{eq:ridgeless:isoG}
\end{align*}
We defer to 
Lemma 23 in the appendix to show that, under \eqref{eq:ridgeless:isoG}, both $\bar \bZ_1$ and $\bar \bZ_2$ are simple functions of the same $d \times nk$ rectangular matrix with i.i.d.~standard Gaussian entries, whose limiting spectral density is the Marchenko-Pastur law. However, the correlations introduced by augmentations mean that, even in the isotropic case \eqref{eq:ridgeless:isoG}, the limiting spectra of $\bar \bZ_1$ and $\bar \bZ_2$ obey some compound Marchenko-Pastur laws --- typically found in the anisotropic setup without augmentation --- and the limiting risk is cumbersome to state, as seen in \cite{hastie2022surprises}. Nevertheless, the Gaussian matrices allow us to derive meaningful surrogates for the risk in settings where the compound Marchenko-Pastur laws do simplify to a simple Marchenko-Pastur law. To specify this surrogate risk, we define, for $\beta \in \R^d$ and $\sigma, \lambda, \gamma > 0$,
\begin{align*}
    R(\beta, \sigma, \lambda, \gamma) 
    \;\coloneqq\; 
    \| \beta \|^2 \lambda^2 \, \partial m_\gamma(-\lambda) 
    +
    \sigma^2 \gamma 
    \, \big( m_\gamma(-\lambda) - \lambda \partial m_\gamma(-\lambda)  \big)
    \;,
\end{align*}
where $m_\gamma(z) \coloneqq \frac{1 - \gamma - z - \sqrt{(1-\gamma-z)^2 - 4\gamma z} }{2 \gamma z}$. For $\lambda = 0$ or $\gamma = 0$, we define the above as the respective limit as $\lambda \rightarrow 0^+$ or $\gamma \rightarrow 0^+$. 
\cite{hastie2022surprises} shows that this is the limiting risk of $\hat \beta^{(\rm ora)}_\lambda$ in the unaugmented case ($k=1$ and $\sigma_A = 0$). The next proposition shows that, under an additional asymptotic constraint, the limiting risk of the augmented estimator can be expressed through $R$. This is possible because the additional constraint allows the risk to be characterized only by $\bar \bZ_2$, the Wishart-distributed surrogate of $\bar \bX_2$; see the proof in 
\cref{appendix:ridgeless:surrogate:proof} for details and for an explicit bound on the approximation.

\begin{proposition} \label{prop:ridgeless:surrogate} Consider the isotropic setup \eqref{eq:ridgeless:isoG} and let $k \geq 2$ and $\sigma^2_A \leq 1$. Write $\lambda_k \coloneqq \frac{(k-1) \sigma^2_A}{k} + \lambda$ and $\sigma^2_k \coloneqq \frac{k+ \sigma_A^2}{k}$. Consider the asymptotic regime \eqref{eq:ridgeless:asymptotic:regime} with $\frac{\sigma^2_A}{ \sqrt{k}} \frac{\sqrt{d}}{\sqrt{n}} = o(1)$ 
and we allow $\lambda \geq 0$. Then
\begin{align*}
  f_\lambda(\bar \bX_1, \bar \bX_2 )
  \;\xrightarrow{\P}&\;
  \lim\,
   R\Big( \mfrac{\lambda}{ \lambda_k} \, \beta,   \mfrac{\sigma_\epsilon}{\sigma_k}, \mfrac{\lambda_k}{\sigma^2_k}, \gamma  \Big)
   \;,
\end{align*}
where $\lim$ denotes the limit under \eqref{eq:ridgeless:asymptotic:regime} with $\frac{\sigma^2_A}{ \sqrt{k}} \frac{\sqrt{d}}{\sqrt{n}} = o(1)$.
\end{proposition}

\cref{prop:ridgeless:surrogate} is meaningful in two regimes: When $\sigma^2_A \rightarrow 0^+$, i.e.~little to no augmentations, or when $\gamma / k \rightarrow 0^+$, i.e.~infinitely many augmentations compared to the dimension-to-sample-size ratio $\gamma=\lim d/n$. When the risk surrogates from \cref{prop:ridgeless:surrogate} are valid, two effects of augmentation are visible: An additional regularization by $(k-1)\sigma_A^2 / k$, and a shrinkage of effective size of $\beta$. The latter can be seen as a debiasing effect, as $\beta$ only plays a role in the bias term of the risk. This mainly arises from the use of oracle augmentation, which introduces additional information on $\beta$. \cref{sec:triple:descent:two:stage} shows that if we additionally need to estimate $\beta$ in the augmentation, a bias term arises.

\begin{figure}[t] 
  \centering
  \begin{tikzpicture}
      \node[inner sep=0pt] at (-4,0)
          {\includegraphics[width=.5\textwidth]{fig6_1_noise_inject_oracle_ridgeless_SURROGATE.pdf}};
      \node[inner sep=0pt] at (3.5,0)
      {\includegraphics[width=.5\textwidth]{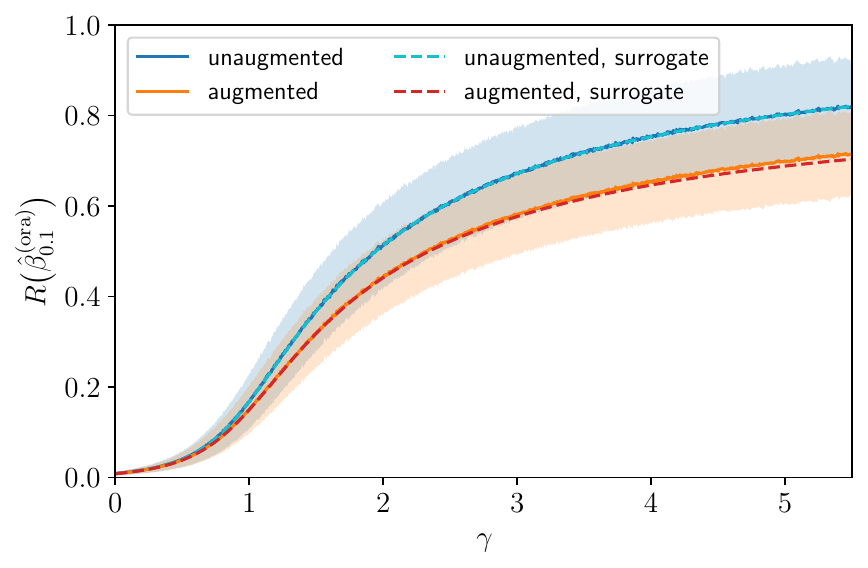}};   
  \end{tikzpicture}
  \caption{\emph{Left.} Risk of the oracle ridgeless estimator $\hat \beta_0^{(\rm ora)}$. \emph{Right.} Risk of the oracle ridge estimator $\hat \beta_\lambda^{(\rm ora)}$ with $\lambda = 0.1$. In both simulations, the data are generated as \eqref{eq:model:linear} with $n=200$, varying $d$, $\| \beta \| = 1$ and $\sigma_\epsilon = 0.1$. The augmentations are noise injections defined in \eqref{eqn:noise:inject} with $k=5$ and $\sigma_A=0.1$. The risk used for simulation is defined in \eqref{eq:practical:risk} while the theoretical risks are obtained from \cref{prop:ridgeless:surrogate}. }   
  \label{fig:noise:inject:oracle} 
\end{figure}

\vspace{.5em}

For the double-descent case $\lambda=0$, the results can be interpreted as follows.
As \cite{hastie2022surprises} explains, whether the unaugmented risk diverges to infinity is determined by the
stability of the pseudoinverse. This stability is measured by the random quantity
\begin{align*}
    \big\| 
      \bar \bX_1^\dagger
    \big\|_{op}
    \;=\; 
    \Big\| 
    \Big( \mfrac{1}{nk} \msum_{ij} (\bV_{i} + \xi_{ij})(\bV_{i} + \xi_{ij})^\top \Big)^\dagger
    \Big\|_{op}
    \;.
\end{align*} 
In the isotropic case, since both Gaussianity and the operator norm are invariant under orthogonal transformations, one may show 
that the
quantity above is distributed as
\begin{align*}
    \Big\| \Big( 
        \mfrac{1}{n} & \msum_{i=1}^n \eta_{i1} \eta_{i1}^\top 
        + 
        \mfrac{\sigma_A^2}{nk} \msum_{i=1}^n \msum_{j=1}^k \eta_{ij} \eta_{ij}^\top \Big)^\dagger 
    \Big\|_{op}
    \;\eqqcolon\;
    \big\| \big( 
        \bW_1
        + 
        \sigma_A^2
        \bW_2 
    \big)^\dagger \big\|_{op}
    \;, \tagaligneq \label{eq:dd:noise:inject:key}
\end{align*}
where $\eta_{ij}$ are i.i.d.~standard Gaussians in $\R^d$ (see 
Lemma 23
in the appendix for the derivation). The two matrices in \eqref{eq:dd:noise:inject:key} are differently
scaled sample covariance matrices, one of $n$ data and another of $nk$ data.
These matrices are correlated through $\{\eta_{i1}\}_{i=1}^n$. The behavior of the risk can then be broken down as follows:
\begin{proplist}
  \item If $\gamma = 1$ (i.e.~$d \approx n$ asymptotically), the pseudoinverse of $\bW_1$ is unstable, whereas since $\gamma' < 1$ (i.e.~$d \lesssim k n $), $\bW_2$ is asymptotically full-ranked
  and close to 
  $\mean \bW_2$. 
  Since 
  $\mean \bW_2$
  is a scaled identity matrix, it acts as a regularization of the pseudoinverse. 
  The regularization effect is evident in \cref{fig:noise:inject:oracle}, where the risk curve of an augmented ridgeless regressor exhibits a small local maximum around $\gamma=1$---similar to what is observed for a ridge regressor in \cite{hastie2022surprises}---instead of the spike towards infinity observed for the unaugmented risk curve. 
  The same regularization effect 
  can be seen from
  the surrogate risk formula from \cref{prop:ridgeless:surrogate}, computed based on the limiting Marchenko-Pastur law  of $\bW_1$; in \cref{fig:noise:inject:oracle}, the surrogate is a good approximation even when $\gamma = 1$ and $k=5$, due to the small noise scale $\sigma_A$ used.
  \item 
  If $\gamma$ exceeds $k$, $\gamma'$ exceeds $1$, and $d$ asymptotically exceeds $kn$. 
  In this case, the sample covariance matrix $\bW_2$ also becomes unstable, and is no longer regularizes $\bW_1$. That causes the risk to diverge, as illustrated in the left plot of \cref{fig:noise:inject:oracle}. 
  The surrogate risk fails to be a good approximation in this regime, as the true risk is now characterized by a compound Marchenko-Pastur law arising from the limiting spectra of $\bW_1+\bW_2$.
  %
  \item As this stability issue does not occur for $\lambda > 0$, no risk spikes are observed for ridge regression. 
  When $\lambda > 0$, the pseudoinverse is also less sensitive to the minimum eigenvalue of the matrices, allowing for the surrogate risk from \cref{prop:ridgeless:surrogate} to serve as a good approximation for larger range of values of $\gamma$. This is evident both in the improved rate of the approximation in \cref{prop:ridgeless:surrogate} and in the right plot of \cref{fig:noise:inject:oracle}.
\end{proplist}

The analysis shows that the interpretation of augmentation as a regularizer suggested in the machine learning literature \citep{dao2019kernel,chen2020group,shorten2019survey,balestriero2022data} depends on the interplay between the number of augmentations $k$, the number of data points $n$ and the dimension $d$. Online augmentation
(where the approximation $k=\infty$ can be justified) behaves like regularizer, as pointed out in previous work.
In offline augmentation (where $k< \infty$), the risk still shows a spike towards infinity that is not
regularized, although this spike now appears around $d \approx nk$ rather than $d \approx n$.

\begin{remark}(Related work)
  (i) The proofs of \cite{hastie2022surprises} use the fact that the random matrices in the unaugmented risk are all rescaled and shifted versions of $\bar \bX_1$, whose eigenspace align. That is a consequence of independence
  between data points, and no longer true if ${k > 1}$.  \\[.2em]
  (ii) Noise injection is studied by \cite{dhifallah2021inherent} for a small $\lambda > 0$, where double-descent is observed in a classification problem with a random feature model but not in regression. Although their work is phrased as a regularization approach, it can be regarded as augmentation. 
  They employ a remarkable proof technique based on tools from convex analysis, and their
  results and ours are complementary: They assume Gaussian data and noise, and
  obtain two separate limiting expressions of the risk
  for an augmented estimator and an unaugmented estimator with a different regularization. 
  Our analysis, on the other hand, 
  shows
  that the shift in double-descent peak is in fact a combination of two effects: A regularization by noise injection around $d \approx n$, and a non-regularized instability around $d \approx nk$. Additionally, our results apply in the non-Gaussian case.
\end{remark}

\subsection{Double and triple descent for sample-splitting estimates} \label{sec:triple:descent:two:stage}

Augmenting the response variables requires knowledge of $\beta$. If we drop the oracle
assumption, we can use a two-stage estimation process with sample splitting, where
an initial estimate $\tilde\beta^{(m)}$ is computed on part of the data.
On the remaining data, this value is used to augment both covariates and responses, and a final estimate $\hat\beta^{(m)}$ is computed.
Consider $m$ i.i.d.~fresh draws of the data $\{\tilde \bV_i, \tilde Y_i\}_{i=1}^m$ obtained e.g.~via data splitting, and form an unaugmented estimator: 
\begin{align*}
  \tilde \beta^{(m)}_{\lambda} 
  \;\coloneqq\; 
  \Big( \mfrac{1}{m} \msum_{i=1}^m \tilde \bV_{i} \tilde \bV_{i}^\top + \lambda \bI_d \Big)^\dagger \mfrac{1}{m} \msum_{i=1}^m \tilde \bV_i \tilde \bY_i\;,
  \qquad 
  \text{ where } \; \lambda \geq 0\;.
\end{align*}
In the case $m=0$, we write $\tilde \beta^{(0)}_{\lambda} = \bzero$. The augmentations applied to $Y_i$'s are given by
\begin{align*}
  \tau^{(m)}_{ij} Y_i 
  \;\coloneqq\; 
  Y_i + \big( \pi_{ij} \bV_i - \bV_i \big)^\top \tilde \beta^{(m)}_{\lambda}
  \;=\; 
  \tau^{(\rm ora)}_{ij} Y_i
  + 
  (\pi_{ij} \bV_i - \bV_i)^\top (\tilde \beta^{(m)}_{\lambda} - \beta )
  \;.
\end{align*}
In this case, invariance of the covariates does not imply invariance of the entire data as in \eqref{eq:invariance:double:descent}.
The final augmented estimator is the two-stage estimator defined with $\tau^{(m)}_{ij}$ as
\begin{align*}
    \hat{\beta}^{(m)}_{\lambda}
    \;\coloneqq&\
    ( \bar \bX_1 + \lambda \bI_d )^\dagger \mfrac{1}{nk} \msum_{ij} (\pi_{ij} \bV_{i}) \, \tau^{(m)}_{ij} Y_i
    \;. \tagaligneq \label{eq:interpolate:defn}
\end{align*}
Thus, $m=0$ corresponds to not augmenting the response variables. 
Observe that the two-stage estimator is related to the oracle estimator by 
\begin{align*}
  \hat{\beta}^{(m)}_{\lambda}
  \;=&\;
  \hat{\beta}^{(\rm ora)}_{\lambda}
  +
  ( \bar \bX_1 + \lambda \bI_d )^\dagger
  \, \bar \bX_\Delta\,
  (\tilde \beta^{(m)}_{\lambda} - \beta )  
  \;,
  \tagaligneq \label{eq:ridgeless:est:relation}
\end{align*}
where the difference arises from the estimation error of the first-stage estimator, $\tilde \beta^{(m)}_{\lambda} - \beta$, as well as the difference arising from augmentation, 
\begin{align*}
  \bar \bX_\Delta \;\coloneqq\; \mfrac{1}{n} \msum_{i=1}^n \Big( \mfrac{1}{k} \msum_{j=1}^k \pi_{ij} \bV_{i} \Big)\Big( \mfrac{1}{k} \msum_{j=1}^k ( \pi_{ij} \bV_{i} - \bV_{i}) \Big)^\top
  \;.
\end{align*}
We consider the risk $R$ defined in \cref{sectn:ridge_regression}, which simplifies under the linear model \eqref{eq:model:linear} as
\begin{align*} 
  R(\hat \beta^{(m)}_{\lambda}) \;=\; 
  \mean[ (Y_{\rm new} - (\hat \beta^{(m)}_{\lambda})^\top \bV_{\rm new} )^2 \,|\, \hat \beta^{(m)}_\lambda]
  \;=\; \| \hat \beta^{(m)}_{\lambda} - \beta \|^2 + \sigma_\epsilon^2
  \;. \tagaligneq \label{eq:practical:risk}
\end{align*}
Note that this risk has an additional $\sigma^2_\epsilon$ not present in \eqref{eqn:interpolator:risk}, which was chosen only for comparison to \cite{hastie2022surprises}. We are again interested in the double-descent case $\lambda =0$. We also allow $m$ to grow with $n$, and write $\rho \coloneqq \lim m/n \in [0,1)$.

\begin{proposition} \label{prop:noise:inject:unaugmented:risk} 
Assume that $\| \bar \bX_1^\dagger \|_{op}$, $\| \bar \bX_2 \|_{op}$, $\| \bar \bX_\Delta \|_{op}$ and $\| \tilde \beta^{(m)}_0 - \beta \|$ are $O(1)$ with probability $1-o(1)$. Then
\begin{align*}
  R(\hat \beta^{(m)}_0)
  -
  \big(
  \sigma_\epsilon^2 
  +
  \hat L_0^{\rm (ora)} 
  + 
  \big\| \bar \bX_{1}^\dagger \bar \bX_\Delta (\tilde \beta^{(m)}_0 - \beta ) \big\|^2
  \big)
  \;\xrightarrow{\P}\; 
  0\;.
\end{align*}
  \end{proposition}

\begin{figure}[t]
  \centering
  \begin{tikzpicture}
      \node[inner sep=0pt] at (-4,0)
          {\includegraphics[width=.48\textwidth]{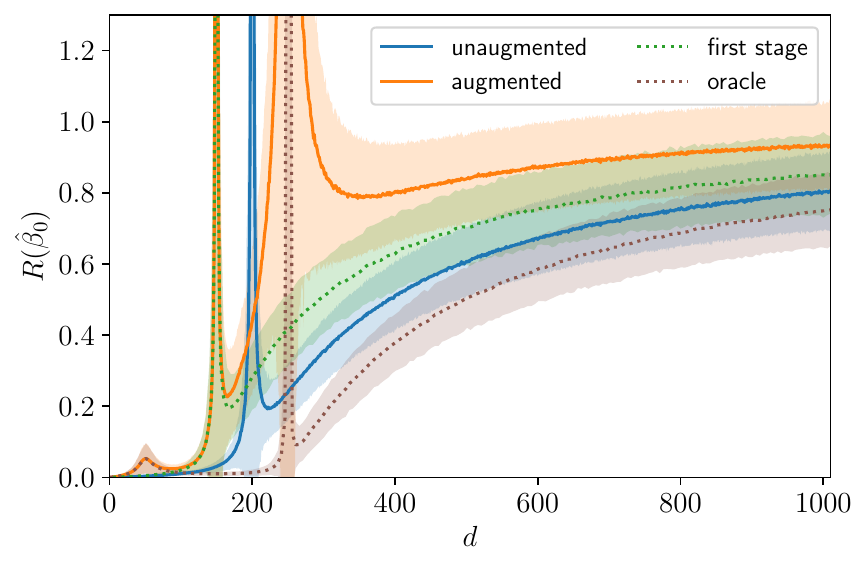}};
      \node[inner sep=0pt] at (3.5,0)
          {\includegraphics[width=.5\textwidth]{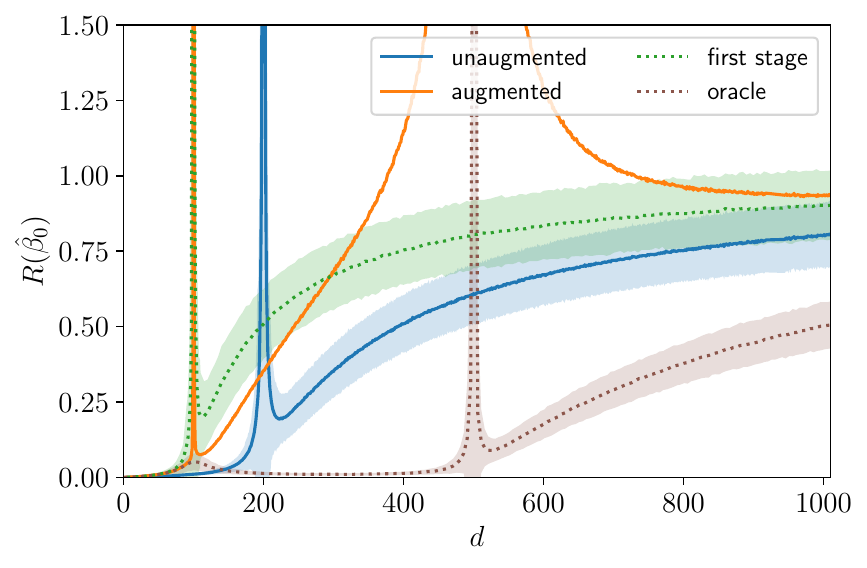}};
      \node[inner sep=0pt] at (-4,-5)
          {\includegraphics[width=.48\textwidth]{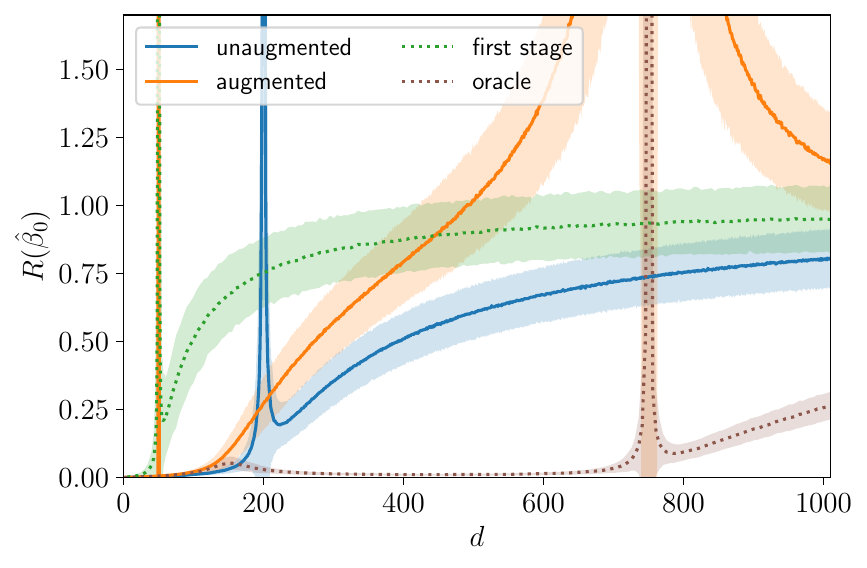}};
      \node[inner sep=0pt] at (3.4,-5)
          {\includegraphics[width=.48\textwidth]{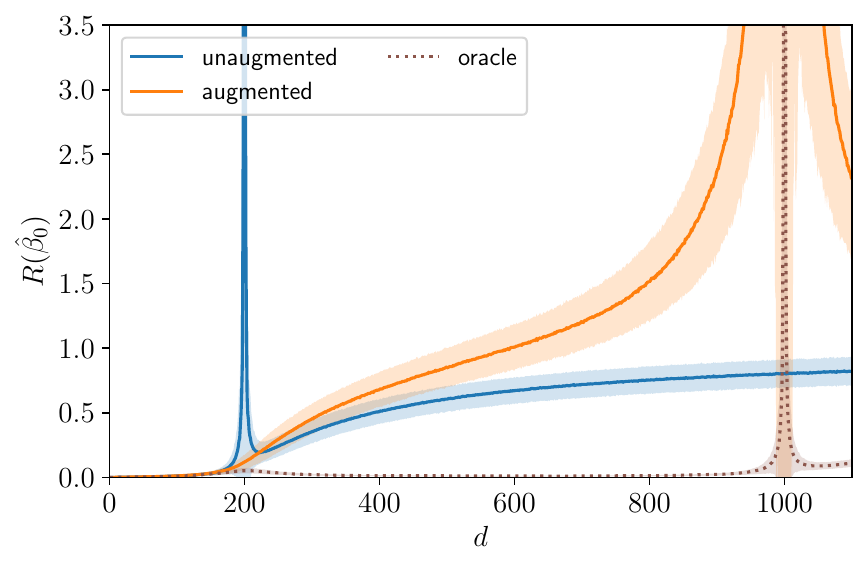}};
  \end{tikzpicture}
  \caption{Risks of the two-stage ridgeless estimator $\hat \beta^{(m)}_0$. In all figures, ${n_{\rm unaug}=200}$ data are used for the unaugmented estimator and ${k=5}$ augmentations are used for the augmented estimator. The number of data used for the two stages of the augmented estimator differ: \emph{Top Left.} 
  ${m=150}$ and ${n_{\rm aug}=50}$; \emph{Top Right.} ${m=n_{\rm aug}=100}$;  \emph{Bottom Left.} ${m=50}$ and ${n_{\rm aug}=150}$; \emph{Bottom Right.} ${m=0}$ and ${n_{\rm aug}=200}$. In each figure, risk of the first-stage unaugmented estimator $\tilde \beta^{(m)}_0$ and risk of the oracle estimator $\hat \beta^{(\rm ora)}_0$ trained on $\{\bV_i\}_{i=1}^{n_{\rm aug}}$ are also plotted for comparison.}
  \label{fig:noise:inject:two:stage} 
\end{figure}

The limiting risk $R(\hat \beta_0^{(m)})$ can be separated into the the risk $\hat L^{(\rm ora)}_0$ of the oracle estimator, the noise $\sigma^2_\epsilon$, and a term $\big\| \bar \bX_{1}^{-1} \bar \bX_\Delta (\tilde \beta^{(m)}_0 - \beta ) \big\|^2$. Our universality result allows one to show that $(\bar \bX_{1}, \bar \bX_\Delta)$ behave like correlated matrices with Gaussian entries, and in the isotropic case, we expect delocalization of the eigenvectors of $\bar \bX_{1}^{-1} \bar \bX_\Delta$ in the sense that 
\begin{align*}
  \big\| \bar \bX_{1}^{-1} \bar \bX_\Delta (\tilde \beta^{(m)}_0 - \beta ) \big\|^2
  \;\approx\;
  \mfrac{1}{d} \Tr\big( \bar \bX_\Delta \bar \bX_{1}^{-2} \bar \bX_\Delta \big)
  \, \| \tilde \beta^{(m)}_0 - \beta \|^2
  \;.
  \tagaligneq \label{eq:ridgeless:conjecture:delocalization}
\end{align*}
A formal justification requires developing anisotropic local laws similar to \cite{knowles2017anisotropic} but for matrices of the form $ \bar \bX_{1}^\dagger \bar \bX_\Delta$, which we leave to future work. Under \eqref{eq:ridgeless:conjecture:delocalization}, the main difference between the two-stage risk $R(\hat \beta_0^{(m)})$ and $\hat L_0^{\rm (ora)}$ is a rescaled risk of the first-stage estimator.
We expect $\hat L^{(\rm ora)}_0$ to diverge near ${\gamma' = 1}$ (i.e.~$d \approx kn$) and $R(\hat \beta_0^{(m)})$  to diverge near ${\gamma / \rho = 1}$ (i.e.~$d \approx m$), leading to \emph{two} spikes in the risk curve of $\hat \beta_0^{(m)}$.
One spike is due to augmentation as discussed in \cref{sec:triple:descent:oracle}, and hence not observed if ${k \rightarrow \infty}$. The other is due to the first-stage, unaugmented regressor on $m$ data, and hence not observed if ${m=0}$.
\cref{fig:noise:inject:two:stage} shows empirical results for fixed $k$ and ${\lambda = 0}$. Both double-descent (for ${m = 0}$) and triple-descent behaviors are clearly visible.

  \begin{remark}
    (i) The results above can be generalized from the ridgeless regressor considered here to two-layer linear networks.
    Indeed, \cite{ba2019generalization} and \cite{chatterji2022interplay} characterize the risk of such a network after
    training in terms of the pseudoinverse in \eqref{eq:dd:noise:inject:key}. Our proof technique can be applied to
    this risk, at the price of more notation.\\[.2em]
    (ii) For simplicity, we have assumed the same value of $\lambda$ is used in both stages, although our approach
    can be extended to distinct values. Since both stages use $\lambda=0$, we see two peaks in the risk, and hence triple-descent.
    If a positive value is used in the first stage instead and ${\lambda=0}$ in the second, one of the peaks would vanish.
  \end{remark}

%% file: article_sec6_newstuff.tex
\subsection{Extensions to simple neural networks and other augmentations} \label{sec:network}

We now consider a linear network model, which has seen wide usage in theoretical analysis \cite{saxe2014exact,arora2019implicit,mixon2022neural,nam2025position} for recovering large-scale empirical phenomena such as neural collapse and grokking; we defer non-linear bagged network models to \Cref{sec:bagging}. Although we only consider the lazy learning regime, where the last layer is trained, the linear network model already introduces significant technical difficulties compared to the linear regression model, as the untrained layers can introduce arbitrary dependence across data coordinates. Moreover, augmentations beyond isotropic noise injection can also introduce data-wise and coordinate-wise dependence. We show that our universality result can accommodate all of these dependencies.

When focusing only on the dependency introduced by augmentations, we observe that, similar to the noise injection case, augmentation shifts the double-descent peak, but the precise effect is now affected by the amount of coordinate-wise dependence augmentations introduce. To quantify this dependency, we introduce an additional notation: Given an $\R^d$ random vector $\eta$, we denote the maximum size of its local dependency neighborhood as $B(\eta) \coloneqq
        \mmax_{l \leq d} \, \big| \inf \{ \cJ \subseteq [d]  \;|\; l \in \cJ \text{ and } (\eta_j)_{j \in \cJ} \text{ is independent of } (\eta_j)_{j \not \in \cJ}   \} \big|
$.

\begin{assumption}(Data) \label{assumption:network:data} Assume that the following conditions hold:
    \begin{proplist}
        \item \textbf{Covariates.} Suppose $\bV_i$'s are i.i.d.~mean-zero and $1$-sub-Gaussian random vectors with $\| \Var[\bV_1] \|_{op} = O(1)$ and with locally dependent coordinates such that $B(\bV_1) =
        o(d^{1/2})$; 
        \item \textbf{Model.} Let $d^{(0)}_0 = d$ and $d^{(0)}_{N_0} = p$. Let $\bW^{(0)}_1, \ldots, \bW^{(0)}_{N_0}$ be independent 
        random matrices 
        such that each $\bW^{(0)}_l$ is $\R^{d^{(0)}_l \times d^{(0)}_{l-1}}$-valued random matrix with i.i.d.~$\cN(0,1/d^{(0)}_{l-1})$ entries, where $d^{(0)}_l$'s grow proportionally to $n$ (see \eqref{eq:network:asymptotic:regime}). As before, fix $\beta \in \R^p$ with $\| \beta \| = O(1)$ and let $\epsilon_i$'s be i.i.d.~mean-zero with $\Var[\epsilon_i] = \sigma^2_\epsilon$. Suppose the true output is generated by 
        \begin{align*}
            Y_i \;=\; \beta^\top \bW^{(0)}_{N_0} \bW^{(0)}_{N_0-1} \ldots \bW^{(0)}_1 \bV_i  + \epsilon_i\;. \tagaligneq \label{eq:model}
        \end{align*}
    \end{proplist}
\end{assumption}

\begin{assumption} (Augmentations)  \label{assumption:network:augmentation} Let the augmentations $\pi_{ij}$'s be i.i.d.~$\R^d \rightarrow \R^d$ transformations, specified as \textbf{one} of the following schemes:
\begin{proplist}
    \item \textbf{Correlated noise injection. } $\pi_{ij}(x) = x + \eta_{ij}$, where $\eta_{ij}$'s are i.i.d.~mean-zero and $1$-sub-Gaussian noise vectors with locally dependent coordinates such that $B( \eta_{11} ) = o(d^{1/2})$;
    \item \textbf{Random cropping. } $\pi_{ij}(x) = (x_l E_{ijl} )_{l \leq d} $, where $E_{ijl}$'s are i.i.d.~Bernoulli variables;
    \item \textbf{Sign-flipping. } $\pi_{ij}(x) = (x_l R_{ijl} )_{l \leq d} $, where $R_{ijl}$'s are i.i.d.~Rademacher variables;
    \item \textbf{Random permutations. } Let $(P_l)_{l \leq N_d}$ be a partition of the index set $[d]$ into $N_d$ subsets and suppose $\sup_{l \leq N_d} | P_l | = O(1)$.  Let $\pi_{ij}$ be i.i.d.~uniformly random permutations of the index set $[d]$ that preserve the partition $(P_l)_{l \leq N_d}$.
\end{proplist}
We also allow the augmentations on labels, $\tau_{ij}$'s, to be one of the following:
\begin{proplist}
    \item \textbf{Oracle. } $\tau_{ij}(Y_i) \coloneqq  \beta^\top \bW^{(0)}_{N_0} \bW^{(0)}_{N_0-1} \ldots \bW^{(0)}_1 \pi_{ij}(\bV_i)  + \epsilon_i$ (c.f.~\Cref{sec:triple:descent:oracle});
    \item \textbf{Identity. } $\tau_{ij}(Y_i) \coloneqq Y_i =  \beta^\top \bW^{(0)}_{N_0} \bW^{(0)}_{N_0-1} \ldots \bW^{(0)}_1 \bV_i  + \epsilon_i$.
\end{proplist}
\end{assumption}

\begin{remark}(Extension to more complicated augmentations) The augmentations in \Cref{assumption:network:augmentation} are chosen for the ease of presentation: (i) The same argument as in \Cref{sec:triple:descent:two:stage} applies for extending $\tau_{ij}$ to the sample-splitting augmentation, where an additional spike is introduced by the first-stage estimator; we omit the details here; (ii) In practice, one may want to crop out or permute a group of coordinates of size $\omega(1)$. We state a much more general setup in \Cref {appendix:locally:dependent:feature}, which allows for any augmentation $\pi_{ij}$'s and $\tau_{ij}$'s such that the augmented data satisfies a local dependency condition. In particular, we are allowed to crop out or permute a group of coordinates of size $\omega(1)$, so long as the original data satisfies a more restrictive dependency condition that $B(\bV_1) = o(d^{r'})$ for some $r' < \frac{1}{2}$.
\end{remark}

Our estimator is given by training the final layer of a pre-trained linear network model with ridge regularization parameter $\lambda > 0$, i.e.
\begin{align*}
    \hat \beta_\lambda(\Phi\cX)
    \;\coloneqq&\; 
    \underset{\tilde \beta \in \R^p}{\argmin}
    \mfrac{1}{nk} \sum_{i \leq n, j \leq k} 
    \big( \tau_{ij}(Y_i) - \tilde \beta^\top W_N W_{N-1} \ldots W_1  \pi_{ij}(\bV_i)  \big)^2
    +
    \lambda \| \tilde \beta\|^2\;,
    \tagaligneq \label{eq:network:ridge}
\end{align*}
where $W_1, \ldots, W_N$ are fixed matrices with $W_l \in \R^{d_l \times d_{l-1}}$, and we again let $d_0 = d$ and $d_N = p$. Note that $N$ does not need to equal $N_0$ and $d_l$ does not need to equal $d^{(0)}_l$, which allows for model misspecification. We again denote the min-norm or ridgeless solution as 
\begin{align*}
    \hat \beta_0(\Phi\cX) \;\coloneqq&\; \textstyle\lim_{\lambda \rightarrow 0^+}
    \hat \beta_\lambda(\Phi\cX)
    \;.
    \tagaligneq \label{eq:network:ridgeless}
\end{align*}
$W_l$'s can be thought of as pre-trained linear layers. Note that in the random neural network literature \cite{saxe2014exact,lee2018deep,arora2019exact}, $W_l$'s are typically taken as random matrices with i.i.d.~Gaussian entries; in that case, the behavior of the network differs depending on whether $N$ is allowed to grow (shallow v.s. deep linear networks) and whether $d_l$'s are fixed or are allowed to grow (narrow v.s. wide networks), as it affects the operator norm of the random matrix product $W_N \ldots W_1$. Here, our risk is not computed over the randomness of the pre-trained layers, and therefore we do not take them to be random. As a result, we do not constrain whether $N$ is fixed or $N$ is allowed to grow, nor how $d_1, \ldots, d_{N-1}$ grows,  
 but instead directly impose a control over the operator norm of the pre-trained layers: 

\begin{assumption} (Non-diverging pre-trained layers) \label{assumption:pretrained:layers} $\| W_N \ldots W_1 \|_{op} \leq C_{op}$ for some absolute constant $C_{op} > 0$ that does not depend on $N$ nor $d_0, d_1, \ldots, d_N$.
\end{assumption}

Analogously to \eqref{eqn:interpolator:risk}, we study the mean-squared test risk
\begin{align*}
  \hat L_{\lambda}(\Phi\cX) 
  \;\coloneqq&\;
  \mean\big[ 
   \big( 
     \hat \beta_{\lambda}(\Phi\cX)^\top W_N \ldots W_1 \bV_{\rm new} - Y_{\rm new}
   \big)^2  
   \,\big| \, \cX, \cW \big]
  \qquad\text{ for }\lambda\geq0\;,
  \tagaligneq \label{eqn:network:risk}
\end{align*}
where we condition on both the input data $\Phi\cX = \{ \pi_{ij} \bV_i \}_{i \leq n, j \leq k}$ and the random weights in the model $\cW = \{ \bW^{(0)}_l \}_{l \leq N_0}$. We also denote the same risk with $\Phi\cX$ replaced by their Gaussian surrogates as $\hat L_{\lambda}(\cZ)$. Analogously to \eqref{eq:ridgeless:asymptotic:regime}, we consider the asymptotic regime where $k, N_0$ are fixed and 
\begin{align*}
  &\;
  n, \, d^{(0)}_0=d_0=d, \, d^{(0)}_1, \, \ldots, \, d^{(0)}_{N_0-1}, \, d^{(0)}_N=d_N=p \,\rightarrow  \, \infty\,,
  \\
  &
  \;\;
  d^{(0)}_l/n 
  \rightarrow \gamma_l  \in[0, \infty)
  \,,\,
  \;
  d^{(0)}_l/(kn) \rightarrow \gamma'_l
  \in[0, \infty)
  \quad 
  \text{ for } 1 \leq l \leq N
  \;.
  \tagaligneq \label{eq:network:asymptotic:regime}
\end{align*}

The next result establishes the universality of $ \hat L_{\lambda}(\Phi\cX)$ for $\lambda > 0$.

\begin{proposition} \label{prop:network} Fix $\lambda > 0$. Under \Cref{assumption:network:data,assumption:network:augmentation,assumption:pretrained:layers} and the asymptotic  \eqref{eq:network:asymptotic:regime},
    \begin{align*} 
        d_P\big( \hat L_{\lambda}(\Phi\cX)\,,\, \hat L_{\lambda}(\cZ)   \big) \;\rightarrow\; 0
        \;.
    \end{align*}
\end{proposition}
 
Similar to \Cref{prop:ridgeless:universality}, the universality of $ \hat L_0(\Phi\cX)$ requires an additional condition analogous to \Cref{assumption:ridgeless:by:ridge}, and we present this result in full in \Cref{appendix:network:ridgeless}.

\vspace{.5em}

As with \Cref{sec:triple:descent:oracle}, universality allows us to reduce the analysis of the double-descent peak to the stability of the pseudoinverse of a Wishart-type matrix $\frac{1}{nk} \sum_{i \leq n, j \leq k} \tilde Z_{ij} \tilde Z_{ij}^\top$, where $\tilde Z_{ij}$ is the Gaussian surrogate for $W_N \ldots W_1 \pi_{ij}(\bV_i)$. While $\tilde Z_{ij}$'s have similar dependence structure across $i \leq n$ and $j \leq k$, the coordinate dependence structure is much more complicated than the isotropic setup in \eqref{eq:ridgeless:isoG}, which is the main hurdle of analysis. To demonstrate how this can be addressed, in \Cref{appendix:peak:correlate}, we include further theoretical analyses, backed by experiments, to show how the different augmentations interact with the double-descent peak in \Cref{sec:triple:descent:oracle} (equivalent to the case $N=0$). The main finding is that, similar to \Cref{sec:triple:descent:oracle}, the double-descent behavior is governed by a sample-covariance matrix of $n$ data and another of $nk$ data; however, since the coordinates of both sample covariance matrices become correlated, the peak is not governed by how the dimension $d$ compares with $n$ and $k$, but by how a notion of ``effective dimension" --- that depends, e.g.~on the ranks of $\Var[\pi_{11}(\bV_1)]$ and $\Cov[\pi_{11}(\bV_1), \pi_{12}(\bV_1)]$ --- compare to $n$ and $k$.

\begin{figure}[t]
  \centering
  \begin{tikzpicture}
      \node[inner sep=0pt] at (-4,0)
          {\includegraphics[width=.48\textwidth]{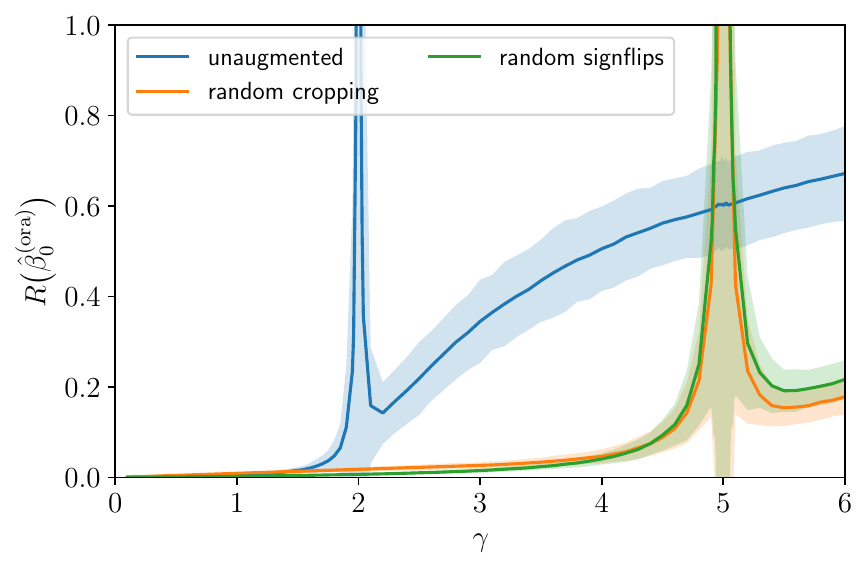}};
      \node[inner sep=0pt] at (3.5,0)
          {\includegraphics[width=.5\textwidth]{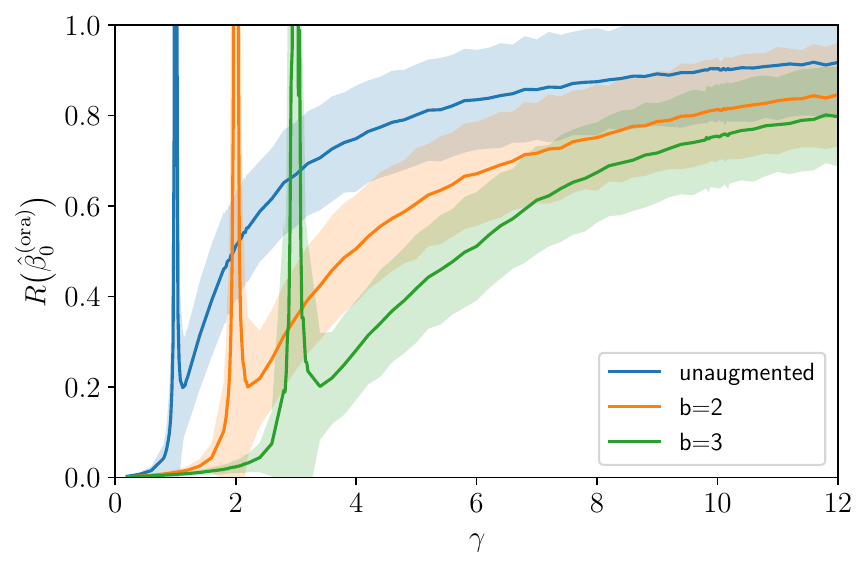}};
  \end{tikzpicture}
  \caption{Risks of the oracle ridgeless estimator $\hat \beta_0^{(\rm ora)}$ defined in \Cref{sec:triple:descent:oracle}. \; {\em Left. } Random cropping and sign-flipping in \Cref{assumption:network:augmentation}. $\bV_i$ is generated such that each coordinate is repeated twice and the rank of $\Var[\bV_i]$ is $d/2$. This shifts the peak of the unaugmented risk to the threshold $d=2n$, but the peak of the augmented risk remains at the threshold $d=kn$. 
  \; {\em Right. } Random permutation in \Cref{assumption:network:augmentation}, where we fix $|P_l| = b$ for all $l \leq N_d$. The augmented risk has a peak at $d=bn$ due to the behavior of a Wishart matrix with $n$ degrees of freedom that is analogous to $\bW_1$ in \eqref{eq:dd:noise:inject:key}. Both effects arise due to the coordinate-wise dependence structure introduced by the data and augmentation choices, and a detailed analysis is included in \Cref{appendix:peak:correlate}. }
  \label{fig:additional:aug} 
\end{figure}

\section{Augmented-and-bagged estimators} \label{sec:bagging}

Bagging \cite{breiman1996bagging}, short for bootstrap aggregating, is an important ensemble algorithm for stabilizing machine learning estimators, and can be applied to a wide range of estimators thanks to its assumption-free stability guarantees \citep{chen2022debiased,soloff2024bagging}. Since our universality result (\Cref{thm:main}) holds under a stability assumption, it can be used to analyze the effects of augmentation on bagged estimators under much more relaxed stability requirements on the base estimator. This notably makes our result applicable to bagged estimators of non-linear networks. 

To formalize how augmentation interacts with bagged estimators, let $f_m: \cD^{mk} \rightarrow \R$ be a thrice-differentiable function that represents a base machine learning estimator trained on $mk$ observations, where $m \leq n$. We shall first augment all $n$ data as before, which yields the $n$ augmented data block $\Phi_1 \bX_1, \ldots, \Phi_n \bX_n$. To form the augmented-and-bagged estimator, we sample $(\upsilon_b)_{b \leq B}$ i.i.d.~uniformly from all permutations of the index set $\{1, \ldots, n\}$, which corresponds to sampling the $n$ data without replacement for a number of $B$ times. The resultant augmented-and-bagged estimator is given by the function $f_m^{(B)}: \cD^{nk} \rightarrow \R$ as
\begin{align*}
    f_m^{(B)}(\Phi \cX ) 
    \;\coloneqq\; 
    \mfrac{1}{B} \msum_{b \leq B} f_m
    \big( \Phi_{\upsilon_b(1)} \bX_{\upsilon_b(1)}, \ldots, \Phi_{\upsilon_b(m)} \bX_{\upsilon_b(m)} \big)
    \;.
\end{align*}
For a generic $f: \cD^{nK} \rightarrow \R$, \Cref{thm:main} says that a sufficient condition for the universality of $f(\Phi \cX )$ is that $f$ is stable, in the sense that the local derivatives from \eqref{eqn:defn:alpha_r} are sufficiently small; recall from the discussion under \eqref{eqn:defn:lambda} that this requires e.g.~the first partial derivative of $f$ to be $o(n^{-1/3})$. Since bagging improves stability, we expect the bagged estimator $f_m^{(B)}(\Phi \cX )$ to exhibit universality with much less stringent requirements on the derivatives of $f_m$. 

Our next set of results show that universality for the bagged estimator only requires the first two partial derivatives of the base estimator $f_m$ to be $O(1)$, and that the third partial derivative is on the order $O(n^{-1/2})$. To formalize this, we define the noise stability term $\alpha_r(f_m^{(B)})$ as in \eqref{eqn:defn:alpha_r}, with the dependence on $f_m^{(B)}$ made explicit. The next result controls $\alpha_r(f_m^{(B)})$ in terms of the stability terms of the base estimator $f_m$, defined as
\begin{align*}
    \alpha^{\rm base}_{r;t}
    \coloneqq
     \max_{i \leq m, \upsilon \in S([m]) }
     \max\big\{
        \big\|
                    \Delta_{i,r,\upsilon}(\Phi_i\bX_i)
        \big\|_{L_{6+t}}
        ,
        \big\|
                \Delta_{i,r,\upsilon}(\bZ_i)
        \big\|_{L_{6+t}}
    \Big\}
\end{align*}
for $r \in \N$ and $t > 0$, where we have denoted $S([m])$ as the set of all permutations on the index set $\{1,\ldots, m\}$, 
$\Delta_{i,r,\upsilon}(\bx) \coloneqq {\sup_{\bw \in[\bzero,\bx]}} \big\| 
                        D_i^r  
                        f_m
                        \big( \bW^{\upsilon}_i(\bw)  \big)
                    \big\|$,
and 
$\bW_{i}^{\upsilon}(\bw) \coloneqq (\Phi_{\upsilon(1)}\bX_{\upsilon(1)},\ldots, \Phi_{\upsilon(i-1)}\bX_{\upsilon(i-1)},\bw,\bZ_{\upsilon(i+1)}, \ldots,\bZ_{\upsilon(m)})$ where  $\upsilon$ permutes the $m$ arguments.

\begin{proposition} \label{prop:bagging} Let $q=1$ and define $(\bX_i)_{i \leq n}$ and $\phi_{ij}$ as in \Cref{thm:main}. If $m=o(\sqrt{n})$ and $B=\Omega(n^{1 - t/(108+18t)} )$ for some fixed $t > 0$, then 
\begin{align*}
    \alpha_r(f_m^{(B)})
    \;=\;
    o\Big( \mfrac{\alpha^{\rm base}_{r;t}}{\sqrt{n}}
    \Big) 
    \qquad 
    \text{ for } r=1,2,3 \;.
\end{align*}
\end{proposition}

Under \Cref{prop:bagging} and \Cref{thm:main}, universality can be established for $f^{(B)}_m$ even though, for instance, the first partial derivative of $f$ is not $o(n^{-1/3})$:

\begin{corollary} \label{cor:bagging:universality} Assume the conditions of \Cref{prop:bagging}. If the moment terms from \Cref{thm:main} satisfy that $c_X, c_Z = O(1)$, and if the stability terms of the base estimator satisfy that $\alpha^{\rm base}_{1;t}, \alpha^{\rm base}_{2;t}  = O(1)$ and $\alpha^{\rm base}_{3;t} = O(n^{-1/2})$, then as $n \rightarrow \infty$,
\begin{align*}
    d_\cH\big( f^{(B)}_m(\Phi\cX), f^{(B)}_m(\bZ_1, \ldots, \bZ_n) \big) \;\rightarrow\; 0\;.
\end{align*}
\end{corollary}

\begin{remark}
In general, we may want to establish universality of $g\big( f_m^{(B)}(\Phi \cX ) \big)$ with respect to some $g: \R^q \rightarrow \R$ that measures a particular property of the estimator, e.g.~the test risk considered in \Cref{sectn:ridge_regression,sec:ridgeless}. A similar result to \Cref{prop:bagging} can be established  for $g \circ f_m^{(B)}$, and we include this generalization in \Cref{appendix:statistics:of:bagging}. 
\end{remark}

The relaxed stability conditions allow us to study augmentations for bagged estimators built on more complicated models. For instance,  we may establish universality for bagged versions of \emph{non-linear} pretrained neural networks of the form 
\begin{align*}
    \argmin_{\tilde \beta \in \R^p} 
    \mfrac{1}{nk} \sum_{i \leq n, j \leq k} 
    \big( \tau_{ij}(Y_i) - \tilde \beta^\top W_N \varphi_{N-1}( W_{N-1} \ldots \varphi_1(  W_1 \pi_{ij}(\bV_i) ) \ldots )  \big)^2
    +
    \lambda \| \tilde \beta\|^2,
\end{align*}
where $W_N, \ldots, W_1$ are the pre-trained layers in \eqref{eq:network:ridge} and $\varphi_N, \ldots, \varphi_1$ are smooth non-linear functions such as pointwise ${\rm tanh}$ activations; for $N=1$, the above can also be viewed as regression with a random feature model. The key to proving universality is to modify the proof of \Cref{prop:network} with \Cref{prop:bagging}. As the setup and the universality results are similar to \Cref{prop:network}, we include their formal statements in \Cref{appendix:bagged:network}.

%% file: appendix.tex
\begin{center}
    \textbf{Appendices}
\end{center}

\noindent
The appendix is organized as follows:
\\[.1em]
\textbf{\cref{appendix:var:cor}} states several generalizations and
additional corrolaries of the main result.
\\[.1em]
\textbf{\cref{appendix:example:results}} states additional results for the toy statistic at the end of 
\cref{sec:averages}, the ridgeless regressor as well as its extensions in 
\cref{sec:ridgeless}, and the bagged estimator in 
\cref{sec:bagging}.
\\[.1em]
\textbf{\cref{appendix:auxiliary}} states and proves auxiliary tools used in subsequent proofs.
\\[.1em]
\textbf{\cref{appendix:proof:main}} proves our main theorem. A proof
overview is given in \cref{proof:overview}.
\\[.1em]
\textbf{\cref{appendix:proof:var:cor}} presents the proofs of the results in \cref{appendix:var:cor}.
\\[.1em]
\textbf{\cref{appendix:der:examples}} proves all results in 
\cref{sec:averages}, \cref{appendix:toy:results} and \cref{sectn:ridge_regression}, 
all of which concern the asymptotic distribution and variance of the estimator.
\\[.1em]
\textbf{\cref{appendix:ridgeless:proof}} proves all results in 
\cref{sec:triple:descent:oracle,sec:triple:descent:two:stage} 
and \cref{appendix:ridgeless:results}, which concern the limiting risk of an overparameterized ridge and ridgeless estimator.
\\[.1em]
\textbf{\cref{appendix:ridgeless:network:proof}} proves all results in \Cref{sec:network} and \Cref{appendix:network:extend}, which concern the limiting risk of an overparamaterized nonlinear feature model and a simple neural network.
\\[.1em]
\textbf{\cref{appendix:proof:bagging}} proves all results in \Cref{sec:bagging} and \Cref{appendix:bagging}, which concern bagged estimators and bagged nonlinear neural networks.
\\[.5em]
\textbf{Notation}. Throughout the appendix, we shorten $\alpha_{r;m}(f)$ to $\alpha_{r;m}$ whenever $f$ is clear from the context, and write
$\cZ^{\delta} \coloneqq \{\bZ^{\delta}_1, \ldots, \bZ^{\delta}_n \} \in \cD^{nk}\;. $

\vspace{.5em}

\appendix

\input{appendix_1_additional_main.tex}

\input{appendix_2_additional_examples.tex}

\input{appendix_3_aux_tools.tex}

\input{appendix_4_proof_main.tex}

\input{appendix_5_proof_add_main.tex}

\input{appendix_6_proof_examples.tex}

\input{appendix_7_ridgeless.tex}

\input{appendix_8_additions.tex}




%% file: appendix_1_additional_main.tex
\section{Variants and corollaries of the main result} \label{appendix:var:cor}

This section provides some additional results. \cref{thm:main:general} below generalizes 
\cref{thm:main} 
such that
(i) transformed data $\phi(x)$ and $x$ are allowed to live in
different domains, and (ii) an additional parameter $\delta$ trades off between a tighter bound and lower variance.
Corresponding generalizations of the corollaries
in 
\cref{sec:main:result} 
follow.
We also provide a formal statement for the convergence of estimates of the
form $g(\text{empirical average})$ discussed in 
\cref{sec:plugin}
(see \cref{lem:plugin:formal}).

\subsection{Generalizations of results in \cref{sec:main:result}
} \label{appendix:general:main:result}

\vspace{.2em}

We first allow the domain and range of elements of $\cT$, i.e. augmentations to differ: Let $\cT'$ be a family of measurable transformations $\cD' \rightarrow \cD$, and the data $\bX_1, \ldots, \bX_n$ be i.i.d.\,random elements of $\cD' \subseteq \R^{d'}$. An example where this formulation is useful is the empirical risk, where we study the empirical average of the following quantities
\begin{align*}
    &
    l(\tau_{11}\bX_1), \ldots, l(\tau_{nk}\bX_n)\;,
    &&
    \text{ for some loss function } l: \cD' \rightarrow \R\;.
\end{align*}
Note that \cref{thm:main:general} remains applicable by setting $\phi_{ij}(\bX_1) \coloneqq l(\tau_{ij} \bX_1)$, with the augmentations used on data are determined through $\tau_{ij}$. 

\vspace{.3em}

Next, we introduce a deterministic parameter $\delta \in [0,1]$, and redefine the moment and mixed smoothness conditions. Recall 
${\Sigma_{11}\coloneqq\Var[\phi_{11}\bX_1]}$ and
${\Sigma_{12}\coloneqq\Cov[\phi_{11}\bX_1,\phi_{12}\bX_1]}$, the $d\times d$ matrices defined in 
\eqref{eqn:defn:surrogates} in the main text.
Consider the following alternative requirements on moments of surrogates $\{\bZ^{\delta}_i\}_{i \leq n}$:
\begin{align} \label{eqn:defn:surrogates:delta}
    \mean \bZ^{\delta}_i
    =
    \mathbf{1}_{k\times 1}\otimes\mean[\phi_{11}\bX_1]
    ,
    \;
    \Var \bZ^{\delta}_i
    =
    \mathbf{I}_{k}\otimes \big( (1-\delta) \Sigma_{11} + \delta \Sigma_{12}\big)
    +
    (\mathbf{1}_{k\times k}-\mathbf{I}_k)\otimes \Sigma_{12}.
\end{align}
Note that when $\delta=0$, this recovers 
 \eqref{eqn:defn:surrogates}.
Write $\bZ^{\delta}_1 = (\bZ^{\delta}_{1j})_{j\leq k}$ where $\bZ^{\delta}_{ij} \in \cD$. In lieu of the moment terms defined in 
\eqref{eqn:defn:moments}
, we consider the moment terms defined by
\begin{align*}
 &
 c_1 \;\coloneqq\; \mfrac{1}{2} \big\| \mean \Var[\phi_{11} \bX_1 | \bX_1 ] \big\|\;,
 &&
 c_X\;\coloneqq\;\mfrac{1}{6}\sqrt{\mean\|\phi_{11} \bX_1 \|^6 }\;,
 &&&
  c_{Z^{\delta}} \;\coloneqq\; 
  \mfrac{1}{6}\sqrt{\mean\bigl[ \|\bZ^{\delta}_{11}\|^6 \bigr]}\;.
\end{align*}
Again when $\delta=0$, the last two moment terms are exactly those defined in 
\eqref{eqn:defn:moments}
. Finally, we also use a tighter moment control on noise stability. Denote $\bW^\delta_i$ as the analogue of $\bW_i$ with $\{\bZ_{i'}\}_{i' > i}$ replaced by $\{\bZ^{\delta}_{i'}\}_{i' > i}$, and define
\begin{align*} 
  \alpha_{r;m}(f)
  \coloneqq&
  \sum_{s \leq q} \max_{i\leq n} \max\biggl\{
  \biggl\|{\sup_{\bw\in[\bzero,\Phi_i\bX_i]}}
  \|D_i^r f_s(\bW^{\delta}_i(\bw))\|\biggr\|_{L_m},
    \biggl\|{\sup_{\bw\in[\bzero,\bZ^{\delta}_i]}}
  \|D_i^r f_s(\bW^{\delta}_i(\bw))\|\biggr\|_{L_m}
  \biggr\}.
\end{align*} 
$\alpha_{r;m}(f)$ is related to $\alpha_r(f)$ defined in 
\eqref{eqn:defn:alpha_r} 
by $\alpha_r(f) = \alpha_{r;6}(f)$ in the case $\delta=0$. The mixed smoothness terms of interest are in turn defined by
\begin{align}
    &\lambda_1(n,k)
    \coloneqq
\gamma_2(h)
    \alpha_{1;2} (f)^2 
    +
    \gamma_1(h) \alpha_{2;1} (f)\;,
    \nonumber
    \\
   \text{and } &\lambda_2(n,k)
   \coloneqq
  \gamma_3(h) \alpha_{1;6} (f)^3 + 3 \gamma_2(h) \alpha_{1;4}(f) \alpha_{2;4}(f) + \gamma_1(h) \alpha_{3;2}(f)\;.
\end{align}
The choice of $L_6$ norm in 
\ref{thm:main} 
is out of simplicity rather than necessity.

\begin{theorem} (Main result, generalized) \label{thm:main:general} Consider i.i.d.\ random elements ${\bX_1,\ldots,\bX_n}$ of
$\cD'$, and two functions
${f\in\cF_3(\cD^{nk},\R^q)}$
and
${h\in\cF_3(\R^{q},\R)}$. 
Let ${\phi_{11},\ldots,\phi_{nk}}$ 
be i.i.d.\ random elements of $\cT'$, independent of $\cX$. Then for any i.i.d.\ variables
${\bZ^{\delta}_1,\ldots,\bZ^{\delta}_{n}}$ in $\cD^k$ satisfying \eqref{eqn:defn:surrogates:delta},
\begin{equation*}
  \bigl|
  \mean h(f(\Phi\cX))
  -
  \mean h(f(\bZ^{\delta}_1,\ldots,\bZ^{\delta}_n))
  \bigr|
  \;\leq\;
  \delta nk^{1/2}\lambda_1(n,k) c_1 +
  nk^{3/2}\lambda_2(n,k)(c_X+c_{Z^{\delta}})\;.
\end{equation*}
\end{theorem}

\vspace{.3em}

The proof of \cref{thm:main:general} is delayed to \cref{appendix:proof:main}. By observing the bound in Theorem \ref{thm:main:general} and the moment condition 
\eqref{eqn:defn:moments}, 
we see that $\delta$ is a parameter that trades off between a tighter bound at the price of higher variances $\Var[\bZ_i]$ (for $\delta=0$), versus an additional term in the bound and smaller variance ($\delta=1$). In particular, setting $\delta=0$ recovers 
\Cref{thm:main}:
\begin{proof}[Proof of 
  \Cref{thm:main}]
In Theorem \ref{thm:main:general}, setting $\cD'=\cD$ recovers $\cT$ from $\cT'$, and setting $\delta=0$ recovers $\{\bZ_i\}_{i \leq n}, c_{Z}$ from $\{\bZ^\delta_i\}_{i \leq n}, c_{Z^\delta}$. Moreover, only the second term remains in the RHS bound. Since for $m \leq 12$ and $\delta=0$, each $\alpha_{r;m}(f)$ is bounded by $\alpha_r(f)$, we have that $\lambda_2(n,k)$ is bounded from above by $\lambda(n,k)$, which recovers the result of 
\Cref{thm:main}.
\end{proof}

Next, we present generalizations of the corollaries in 
\cref{sec:main:result}. 
\cref{cor:variance:convergence} 
concerns convergence of variance, which can be proved by taking $h$ to be the identity function on $\R$, replacing $f$ with coordinates of $f$, $f_r(\argdot)$ and $ f_r (\argdot) f_s(\argdot)$ for $r, s \leq q$, and multiplying across by the scale $n$. We again present a more general result in terms of $\cZ^\delta$ and noise stability terms $\alpha_{r;m}$ defined in
Theorem \ref{thm:main:general},
of which 
\Cref{cor:variance:convergence} 
is then an immediate consequence:

\begin{lemma} (Variance result, generalized) \label{lem:variance:convergence:general} Assume the conditions of Theorem \ref{thm:main:general}, then
\begin{align*}
    n \big\| \Var[f(\Phi\cX)] - \Var[f(\cZ^{\delta})]  \big\| 
    \;\leq&\;
    4 \delta n^2 k^{1/2} ( \alpha_{0;4} \alpha_{2;4} + \alpha_{1;4}^2 ) c_1
    \\
    &\;
    + 6 n^2 k^{3/2} ( \alpha_{0;4} \alpha_{3;4} + \alpha_{1;4} \alpha_{2;4} ) (c_X + c_{Z^{\delta}})\;.
\end{align*}
\end{lemma}

\begin{proof}[Proof of 
  \cref{cor:variance:convergence}] 
Since $\alpha_{r;m}(f) \leq \alpha_r(f)$ for $m \leq 12$ and $\delta=0$, the second term in the bound in Lemma \ref{lem:variance:convergence:general} can be further bounded from above by the desired quantity
$$
 6 n^2 k^{3/2}( \alpha_0 \alpha_3 + \alpha_1 \alpha_2 ) (c_X + c_Z) \;.
$$
Setting $\delta=0$ recovers $\{\bZ_i\}_{i=1}^n$ from $\cZ^{\delta}$ and causes the first term to vanish, which recovers 
\Cref{cor:variance:convergence}.
\end{proof}

\Cref{cor:d_H:convergence} 
concerns convergence in $d_H$. We present a tighter bound below:
\begin{lemma} ($d_{\cH}$ result, generalized) \label{lem:d_H:convergence:general}
Assume the conditions of 
\Cref{thm:main}, 
then
\begin{align*}
    &\;d_{\cH}( \sqrt{n} f(\Phi\cX), \sqrt{n} f(\cZ^{\delta}) ) 
    \\
    \;&\leq\;
    \delta n^{3/2}k^{1/2} c_1 \big( n^{1/2} \alpha_{1;2}^2 + \alpha_{2;1} \big)
    +
    (nk)^{3/2} ( n \alpha_{1;6}^3  + 3 n^{1/2} \alpha_{1;4} \alpha_{2;4} + \alpha_{3;2} )(c_X+c_{Z^{\delta}})\;.
\end{align*}
\end{lemma}

\vspace{.3em}

\begin{proof}[Proof of 
  \cref{cor:d_H:convergence}] 
  We again note that setting $\delta=0$ recovers $\{\bZ_i\}_{i=1}^n$ from $\cZ^{\delta}$ and $c_Z$ from $c_{Z^\delta}$. The required bound is obtained by setting $\delta=0$ and bounding each $\alpha_{r;m}$ term by $\alpha_r$ in the result in Lemma \ref{lem:d_H:convergence:general}:
\begin{align*}
    (nk)^{3/2} ( n (\alpha_{1;6})^3  + 3 n^{1/2} \alpha_{1;4} \alpha_{2;4} + \alpha_{3;2} )
    \;\leq\;
    (nk)^{3/2} ( n \alpha_1^3  + 3 n^{1/2} \alpha_1 \alpha_2 + \alpha_3 )(c_X + c_Z)\;.
\end{align*}
\end{proof}

As discussed in the main text, the result for no augmentations in 
\eqref{eq:no_aug} 
is immediate from setting the augmentations $\phi_{ij}$ to identity almost surely in 
\Cref{thm:main}. 
Equivalent versions of Lemma \ref{lem:variance:convergence:general} and Lemma \ref{lem:d_H:convergence:general} for no augmentation can be obtained similarly, and the statements are omitted here. This means that to compare the case with augmentation versus the case without, we only need to check the conditions of Lemma \ref{lem:variance:convergence:general} and Lemma \ref{lem:d_H:convergence:general} once.

\vspace{-.5em}

\subsection{Results corresponding to 
Remark 1}
\label{appendix:remark:main}

\vspace{.2em}

As mentioned in  
\ref{remark:thm:main}(ii), 
one may allow $q$ to grow with $n$ and $k$. While 
\Cref{cor:variance:convergence} 
still applies if $q$ grows sufficiently slowly, 
\ref{lem:d_H} 
does not apply unless $q$ is fixed. The following lemma is a substitute. As is typical in high-dimensional settings, we focus on studying the convergence of $f_s$, a fixed $s$-th coordinate of $f$ for $s \leq q$. The lemma gives a sufficient condition on $f$ for convergence of variance for $f$ and convergence in $d_{\cH}$ for $f_s$ to hold when $q$ grows with $n$ and $k$.

\begin{lemma} \label{lem:sufficient:cond:vary:q} 
Assume the conditions of 
\Cref{thm:main} 
and fix $s \leq q$. If coordinates of $\phi_{11}\bX_1$ and $\bZ_1$ are $O(1)$ a.s., $\alpha_1 = o( n^{-5/6}(kd)^{-1/2})$, $\alpha_3 = o( (nkd)^{-3/2})$ and $ \alpha_0 \alpha_3, \alpha_1\alpha_2 = o(n^{-2}(kd)^{-3/2})$, either as $n, d, q$ grow with $k$ fixed or as $n, d, q, k$ all grow, then under the same limit,
\begin{align*} 
    &
    d_{\cH}(\sqrt{n} f_s(\Phi\cX), \sqrt{n} f_s(\bZ_1, \ldots, \bZ_n)) \xrightarrow{d} \bzero\;,
    &&
    n \| \Var[f(\Phi\cX)] - \Var[f(\bZ_1, \ldots, \bZ_n)] \| \rightarrow 0 \;.
\end{align*}
\end{lemma}
The proof is a straightforward result from 
Corollary \ref{cor:variance:convergence}, Corollary \ref{cor:d_H:convergence} and Lemma \ref{lem:d_H}. 
In practice, one may want to use Lemma \ref{lem:variance:convergence:general} and Lemma \ref{lem:d_H:convergence:general} directly for tighter controls on moments and noise stability, which is the method we choose for the derivation of examples in \cref{appendix:der:examples}.

\vspace{.3em}

\Cref{remark:thm:main}(iii)
 discusses the setting where data is distributionally invariant to augmentations. In this case, Theorem \ref{thm:main:general} becomes:

\begin{corollary} ($\cT$-invariant data source) \label{cor:thm:main:invariance} Assume the conditions of Theorem \ref{thm:main:general} and $\phi\bX \overset{d}{=}\bX$ for every $\phi \in \cT$. Then
\begin{equation*}
  \bigl|
  \mean h(f(\Phi\cX))
  -
  \mean h(f(\bZ^{\delta}_1,\ldots,\bZ^{\delta}_n))
  \bigr|
  \;\leq\;
  \delta nk^{1/2}\lambda_1(n,k) c_1 + nk^{3/2}\lambda_2(n,k)(c_X+c_{Z^{\delta}})\;,
\end{equation*}
where $\bZ^{\delta}_1, \ldots, \bZ^{\delta}_n$ are i.i.d.\,variables satisfying 
\begin{align*} 
    &
    \mean \bZ^{\delta}_i
    =
    \mathbf{1}_{k\times 1}\otimes\mean[\phi_{11} \bX_1]
    ,
    &&
    \Var \bZ^{\delta}_i
    =
    \mathbf{I}_{k}\otimes \big( (1-\delta) \tilde \Sigma_{11} + \delta \Sigma_{12}\big)
    +
    (\mathbf{1}_{k\times k}-\mathbf{I}_k)\otimes \Sigma_{12},
\end{align*}
where we have denoted
\begin{align*}
    &
    \tilde \Sigma_{11}\;\coloneqq\;\mean \Var[\phi_{11}\bX_1 | \phi_{11}]\;,
    &&
    \Sigma_{12}\;\coloneqq\;\Cov[\phi_{11}\bX_1,\phi_{12}\bX_1]=\mean \Cov[\phi_{11}\bX_1,\phi_{12}\bX_1 | \phi_{11}, \phi_{12}]\;.
\end{align*}
\end{corollary}

\vspace{.2em}

This result is connected to results on central limit theorem under group invariance  \citep{Austern2022}, by observing that when $\cT$ is a group, the distribution of $\bZ$ is described exactly by group averages. We also note that since $\tilde \Sigma_{11} \preceq \Sigma_{11}$, the invariance assumption leads to a reduction in \emph{data} variance, although this does not imply reduction in variance in the estimate $f$. Finally, the invariance assumption implies $\mean[\phi_{11}\bX_1] = \mean[\bX_1]$, in which case the augmented estimate $f(\Phi\cX)$ is a consistent estimate of the unaugmented estimate $f(\tilde \bX_1, \ldots, \tilde \bX_n)$.

\vspace{.3em}

\Cref{remark:thm:main}(iii) 
says that a stricter condition on $f$ that typically requires $k$ to grow recovers a variance structure resembling that observed in \cite{chen2020group}: variance of an conditional average taken over the distribution of augmentations. This is obtained directly by setting $\delta=1$ in Theorem \ref{thm:main:general} and noting that, by Lemma \ref{lem:compare_variance}, $\Cov[\phi_{11}\bX_1, \phi_{12}\bX_1]  = \Var\mean[\phi_{11} \bX_1 | \bX_1]$ :

\begin{corollary} (Smaller data variance) \label{cor:thm:main:smaller:var} Assume the conditions of Theorem \ref{thm:main:general} with $\delta=1$. Then
\begin{equation*}
  \bigl|
  \mean h(f(\Phi\cX))
  -
  \mean h(f(\bZ_1,\ldots,\bZ_n))
  \bigr|
  \;\leq\;
  nk^{1/2}\lambda_1(n,k) c_1 + nk^{3/2}\lambda_2(n,k)(c_X+c_Z)\;,
\end{equation*}
where $\bZ_1, \ldots, \bZ_n$ are i.i.d.\,variables satisfying 
\begin{align*} 
    &
    \mean \bZ_i
    =
    \mathbf{1}_{k\times 1}\otimes\mean[\phi_{11} \bX_1]
    ,
    &&
    \Var \bZ_i
    =
    \mathbf{1}_{k\times k} \otimes \Var \mean[\phi_{11} \bX_1 | \bX_1]\;.
\end{align*}
\end{corollary}

\vspace{.3em}

Note that the data variance is smaller than that in Theorem \ref{thm:main:general} in the following sense: By Lemma \ref{lem:compare_variance}, $ \Var[\phi_{11}\bX_1] \succeq \Cov[\phi_{11}\bX_1, \phi_{12}\bX_1] = \Var\mean[\phi_{11} \bX_1 | \bX_1]$ and therefore we have $ \bI_k \otimes (\Var[\phi_{11}\bX_1] - \Cov[\phi_{11}\bX_1, \phi_{12}\bX_1]) \succeq \bzero$. This implies $\Var \bZ_i$ in Corollary \ref{cor:thm:main:smaller:var} can be compared to that in Theorem \ref{thm:main:general} by
\begin{align*}
    \mathbf{1}_{k\times k} \otimes \Var\mean[\phi_{11} \bX_1 | \bX_1]
    \;=&\;
    \mathbf{1}_{k\times k} \otimes \Cov[\phi_{11}\bX_1, \phi_{12}\bX_1] 
    \\
    \;\preceq&\;
    \mathbf{I}_{k}\otimes \Var[\phi_{11}\bX_1]
    +
    (\mathbf{1}_{k\times k}-\mathbf{I}_k)\otimes \Cov[\phi_{11}\bX_1, \phi_{12}\bX_1] \;.
\end{align*}
The stricter condition on $f$ comes from the fact that, for the bound to decay to zero, on top of requiring $\lambda_2(n,k)$ to be $o(n^{-1}k^{-3/2})$, we also require $\lambda_1(n,k)$ to be $o(nk^{-1/2})$. In the case of empirical average in 
\Cref{prop:empirical_average}, 
one may compute that $\lambda_1(n,k) = \gamma^2(h) n^{-1} k^{-1}$, so a smaller \emph{data} variance is only obtained when we require $k \rightarrow \infty$.

\vspace{-.5em}

\subsection{Plug-in estimates \texorpdfstring{$g(\text{empirical average})$}{g(empirical average)}} \label{appendix:plugin}

\vspace{.2em}

We present convergence results that compare $f(\Phi\cX)\coloneqq g(\text{empirical average})$ to two other statistics. One of them is $f(\cZ^{\delta})$, which is already discussed in Theorem \ref{thm:main:general}, and the other one is the limit discussed in 
\eqref{eqn:plugin}, 
which is the following truncated first-order Taylor expansion:
\begin{equation*}
    f^T(x_{11}, \ldots, x_{nk}) \coloneqq g(\mean[\phi_{11}\bX_1]) + \partial g(\mean[\phi_{11}\bX_1]) \big( \mfrac{1}{nk} \msum_{i=1}^n \msum_{j=1}^k \bx_{ij} - \mean[\phi_{11}\bX_1] \big)\;.
\end{equation*}
Since we need to study the convergence towards a first-order Taylor expansion of $g$, we need to define variants of noise stability terms in terms of $g$. Given $\{\phi_{ij}\bX_i\}_{i\leq n, j\leq k}$ and $\{\bZ^{\delta}_i\}_{i \leq n} \coloneqq \{\bZ^{\delta}_{ij}\}_{i \leq n, j \leq k}$, denote the mean and centered sums
\begin{align*}
    \mu\;\coloneqq\;\mean[\phi_{11}\bX_1]\;,
    \qquad
    \bar \bX \;\coloneqq\; \mfrac{1}{nk} \msum_{i,j} \phi_{ij}\bX_i - \mu\;,
    \qquad
    \bar \bZ^{\delta} \;\coloneqq\; \mfrac{1}{nk} \msum_{i,j} \bZ^{\delta}_{ij} - \mu\;.
\end{align*}
For a function $g: \cD \rightarrow \R^q$ and $s \leq q$, we denote the $s^{\rm th}$ coordinate of $g(\argdot)$ as $g_s(\argdot)$ as before, and define a new noise stability term controlling the noise from first-order Taylor expansion around $\mu$:
\begin{align*}
  \kappa_{r;m}(g) 
  \;\coloneqq\; 
  \msum_{s \leq q} \big\|{\msup_{\bw\in [\bzero, \bar\bX]}}
  \big\| \partial^r g_s\big( \mu + \bw \big) \big\| \big\|_{L_m}\;.
\end{align*}
The first-order Taylor expansion also introduces additional moment terms, which is controlled by Rosenthal's inequality from Corollary \ref{cor:rosenthal:vector} and bounded in terms of: 
\begin{align*}
    \bar c_m 
    \;\coloneqq\;
    \Big( 
    \msum_{s=1}^d  \max\big\{ n^{\frac{2}{m}-1} \big\| \mfrac{1}{k} \msum_{j=1}^k (\phi_{1j}\bX_1 - \mu)_s \big\|^2_{L_m} 
    ,\;
    \big\| \mfrac{1}{k} \msum_{j=1}^k (\phi_{1j}\bX_1 - \mu)_s \big\|^2_{L_2} 
    \big\} 
    \Big)^{1/2}\;.
\end{align*}
Finally, since we will compare $f(\Phi\cX)$ to $f(\cZ^{\delta})$, we consider noise stability terms that resemble $\alpha_{r;m}$ from \cref{thm:main:general} but expressed in terms of $g$:
\begin{align}
  \nu_{r;m}(g)
  \coloneqq&
  \sum_{s \leq q}
  \max_{\substack{i \leq n}}\max\bigg\{
  \Big\|{\sup_{\bw\in[\bzero,\Phi_i\bX_i]}}
  \|\partial^r g_s(\overline\bW^{\delta}_i(\bw))\|\Big\|_{L_m},\,
    \Big\|{\sup_{\bw\in[\bzero,\bZ^{\delta}_i]}}
  \|\partial^r g_s(\overline\bW^{\delta}_i(\bw))\|\Big\|_{L_m}
  \bigg\}
  \nonumber\\
  =& \msum_{s\leq q} \mmax_{i \leq n} \zeta_{i;m}\big( \big\|\partial^r g_s\big(\overline\bW^{\delta}_i(\argdot) \big) \big\| \big) 
  \geq
  \mmax_{i \leq n} \zeta_{i;m}\big( \big\|\partial^r g\big(\overline\bW^{\delta}_i(\argdot) \big) \big\| \big) \;,
  \label{eqn:defn:beta_r_m}
\end{align}
where
$$
 \overline \bW^{\delta}_i(\bw) \;\coloneqq\; \mfrac{1}{nk} \big( \msum_{i'=1}^{i-1} \msum_{j=1}^k \phi_{i'j} \bX_{i'} + \msum_{j=1}^k \bw_j + \msum_{i'=i+1}^n \msum_{j=1}^k \bZ^{\delta}_{i'j} \big)\;. 
$$
We omit $g$-dependence in $\kappa_{r;m}$ and $\nu_{r;m}$ whenever the choice of $g$ is obvious.

\vspace{.3em}

\begin{lemma} (Plug-in estimates) \label{lem:plugin:formal} Assume the conditions of Theorem \ref{thm:main:general}. For $g \in \cF_3(\cD, \R^q)$, define the plug-in estimate $ f(\bx_{11:nk}) =  g\big( \frac{1}{nk} \sum_{i\leq n, j \leq k} \bx_{ij} \big)$ and its Taylor expansion $f^T(\bx_{11:nk})$ as in 
  \eqref{eqn:defn:first_order_taylor}. 
  Then, for any $\cZ^{\delta}$ satisfying the conditions of Theorem \ref{thm:main:general},
\begin{enumerate}
    \item[(i)] the following bounds hold with respect to convergences to $f^T(\cZ^{\delta})$: 
    \begin{align*}
    d_{\cH}\big( \sqrt{n} f(\Phi\cX), \sqrt{n} f^T&(\cZ^{\delta})\big)
    \\
    \;=&\;
    O\big( n^{-1/2} \kappa_{2;3} \, \bar c_3^2 + \delta k^{-1/2} \kappa_{1;1}^2 c_1 +  n^{-1/2}  \kappa_{1;1}^3 (c_X + c_{Z^{\delta}}) \big)\;,
    \\
     n \big\| \Var[f(\Phi \cX)] - \Var\big[&f^T(\cZ^{\delta}) \big] \big\|
    \\
    \;=&\; O\big(  \delta k^{-1} \| \partial g(\mu) \|_2^2 \, c_1^2  +  n^{-1/2} \kappa_{1;1} \kappa_{2;4} \bar c_4^3 + n^{-1} \kappa_{2;6}^2 \bar c_6^4 \big)\;.
    \end{align*}
    \item[(ii)] the following bounds hold with respect to convergences to $f(\cZ^{\delta})$: 
    \begin{align*}
    d_{\cH}( \sqrt{n} f(\Phi \cX), \sqrt{n} f(\cZ^{\delta}) ) 
    \;=&\; 
    O\big( \delta \big( k^{-1/2} \nu_{1;2}^2 + n^{-1/2}k^{-1/2} \nu_{2;1} \big) c_1
    \\
    &
    \quad
    +
    \big(n^{-1/2} \nu_{1;6}^3  + 3 n^{-1} \nu_{1;4} \nu_{2;4} + n^{-3/2} \nu_{3;2} \big) (c_X + c_{Z^{\delta}})
    \big) \;,
    \\
    n \| \Var[f(\Phi \cX)] - \Var[f(\cZ^{\delta})] &\| 
    \;=\; 
    O\big( \delta k^{-1/2} ( \nu_{0;4} \nu_{2;4} + \nu_{1;4}^2 ) c_1
    \\
    &\qquad\qquad +
     n^{-1} ( \nu_{0;4} \nu_{3;4} + \nu_{1;4} \nu_{2;4} ) (c_X +  c_{Z^{\delta}}) \big)\;.
    \end{align*}
\end{enumerate}
\end{lemma}

\begin{remark}
The statement in 
\eqref{eqn:plugin} 
in the main text is obtained from Lemma \ref{lem:plugin:formal}(i) by fixing $q$, setting $\delta=0$ and requiring the bounds to go to $0$, which is a noise stability assumption on $g$ and a constraint on how fast $d$ is allowed to grow. Weak convergence can again be obtained from convergence in $d_{\cH}$ by 
\ref{lem:d_H}.
\end{remark}

\subsection{Repeated augmentation} \label{appendix:repeated}

\vspace{.2em}

In Theorem \ref{thm:main}, each transformation is used once and then discarded. A different strategy is to generate 
only $k$ transformations i.i.d., and  
apply each to all $n$ observations.
That introduces additional dependence: In the notation of Section \ref{sec:definitions}, $\Phi_i\bX_i$ and $\Phi_j\bX_j$ are no longer independent if $i\neq j$. The next result adapts Theorem \ref{thm:main} to this case. We require that $f$ satisfies 
\begin{equation}
  \label{eq:permutation:invariance}
  f(\bx_{11},\ldots,\bx_{1k},\ldots,\bx_{n1},\ldots,\bx_{nk})
  \;=\;
  f(\bx_{1\pi_1(1)},\ldots,\bx_{1\pi_{1}(k)},\ldots,\bx_{n\pi_n(1)},\ldots,\bx_{n\pi_n(k)})
\end{equation}
for any permutations $\pi_1, \hdots, \pi_n$ of $k$ elements. That holds for most statistics
of practical interest, including empirical averages and $M$-estimators.

\begin{theorem}{\rm (Repeated Augmentation)}
\label{thm:repeated} Assume the conditions in Theorem \ref{thm:main} with $\cD = \R^d$ and that $f$ satisfies \eqref{eq:permutation:invariance}. Define $\tilde{\Phi} \coloneqq (\phi_{ij} | i \leq n, j \leq k)$, where $\phi_{1j} = \ldots = \phi_{nj} =: \phi_j$ and $\phi_1, \ldots, \phi_k$ are i.i.d. random elements of $\cT$. Then there are random variables
${\bY_1,\ldots,\bY_{n}}$ in $\R^{kd}$ such that
\begin{align*}
  \big| \mean h ( f( \tilde \Phi \cX) ) &- \mean h ( f( \bY_1, \ldots, \bY_n)
) \big| \\
&\leq 
  n \gamma_1(h) \alpha_1 m_1
  +
  n \omega_2(n,k)( \gamma_2(h) \alpha_1^2 
    + 
    \gamma_1(h) \alpha_2)
  +
  nk^{3/2}\lambda(n,k)(c_X+c_Y)\;.
\end{align*}
Here, $\lambda$, $c_X$ and $\alpha_r$ are defined as in Theorem \ref{thm:main}, and $c_Y$ is defined in a way analogous to $c_Z$:
\begin{align*}
    c_Y\;\coloneqq\;\mfrac{1}{6} \sqrt{ \mean\Bigl[ \Bigl(\mfrac{|Y_{111}|^2+\ldots+|Y_{1kd}|^2}{k}\Bigr)^3\Bigr] }
\end{align*}
The additional constant moment terms are defined by 
$m_1
\coloneqq
\sqrt{2 \Tr \Var \mean[\phi_1 \bX_1 | \phi_1]}$, and
\begin{align*}
m_2
\coloneqq
\sqrt{ \sum_{r,s\leq d} 
\mfrac{\Var \mean\big[(\phi_1 \bX_1)_r (\phi_1 \bX_1)_s \big| \phi_1\big]}{2}},
\;
m_3
\coloneqq
\sqrt{ \sum_{r,s\leq d} 12
\Var \mean\big[(\phi_1 \bX_1)_r (\phi_2 \bX_1)_s \big| \phi_1, \phi_2\big]}\;.
\end{align*}
The variables $\bY_i$ are conditionally i.i.d.\,Gaussian vectors with mean
${\mean[\Psi \bX_1 | \Psi_1 ]}$ and covariance matrix ${\Var[\Psi \bX_1 | \Psi_1 ]}$, conditioning on $\Psi \coloneqq \{ \psi_1, \hdots, \psi_k \}$ i.i.d. distributed as $\{\phi_1, \hdots, \phi_k\}$.
\end{theorem}

The result shows that the additional dependence introduced by using
transformations repeatedly does not vanish as $n$ and $k$ grow. Unlike the Gaussian limit in Theorem \ref{thm:main} (when $\cD$ is taken as $\R^d$), the limit here is characterized by variables $\bY_i$ that are only \emph{conditionally} Gaussian, given an i.i.d.\ copy of the augmentations. 
That further complicates the
effects of augmentation. Indeed, there exist statistics $f$ for which i.i.d.\ augmentation as in Theorem \ref{thm:main} does not affect the variance, but repeated augmentation either increases or decreases it. Lemma \ref{example:repeated_toy} gives such an example: Even when distributional invariance holds, augmentation may increase variance for one statistic and decrease variance for the other.

\begin{lemma} \label{example:repeated_toy} Consider i.i.d.\,random vectors $\bX_1, \bX_2$ in $\R^d$ with mean $\mu$ and $\phi_1, \phi_2 \in \R^{d \times d}$ be i.i.d.\ random matrices such that $\phi_1 \bX_1 \overset{d}{=}\bX_1$. Then for $f_1(\bx_1, \bx_2) \coloneqq \bx_1 + \bx_2$ and $f_2(\bx_1, \bx_2) \coloneqq \bx_1 - \bx_2$,
\begin{proplist}
    \item $\Var f_1( \bX_1, \bX_2) = \Var f_1( \phi_1 \bX_1, \phi_2 \bX_2) \preceq \Var f_1( \phi_1 \bX_1, \phi_1 \bX_2)$, and 
    \item $\Var f_2( \bX_1, \bX_2) = \Var f_2( \phi_1 \bX_1, \phi_2 \bX_2) \succeq \Var f_2( \phi_1 \bX_1, \phi_1 \bX_2)$.
\end{proplist}
\end{lemma}

%% file: appendix_2_additional_examples.tex
\section{Additional results for the examples} \label{appendix:example:results}

\subsection{Results for the toy statistic} \label{appendix:toy:results}

In this section, we present results concerning the toy statistic defined in 
\eqref{eqn:defn:exp:neg:chi:sq:stat}. 
For convenience, we write $f \equiv f_{\rm toy}$. To express variances concisely, we define the function
$V(s)\coloneqq(1+4s^2)^{-1/2} - (1+2s^2)^{-1}$, and write
\begin{align*}
    \tilde{\sigma} 
    \,&\coloneqq\, 
    \sqrt{\Var[\bX_1]}\;
    &&\text{ and }&
    \sigma\,&\coloneqq\, 
    \Bigl(\mfrac{1}{k} \Var[\phi_{11}\bX_1] + \mfrac{k-1}{k} \Cov[\phi_{11}\bX_1, \phi_{12}\bX_1]\Bigr)^{1/2}\;.
\end{align*}

The next result applies 
\cref{thm:main} 
to derive closed-form formula for the quantities plotted in 
\cref{fig:exp_neg_chi_squared_1D}:

\begin{proposition}\label{prop:exp_neg_chi_squared} 
Require that $\mean[\bX_1]=\mean[\phi_{11}\bX_1]=0$, and that
$\mean[|\bX_1|^{12}]$ and $\mean[|\phi_{11}\bX_1|^{12}]$ are finite.
Let $\cZ, \cZ'$ be Gaussian. Then $f \equiv f_{\rm toy}$ defined in 
\eqref{eqn:defn:exp:neg:chi:sq:stat} 
satisfies
\begin{equation*}
  d_{\cH}(f(\Phi \cX), f(\cZ)) \rightarrow 0
  \quad\text{ and }\quad
  \Var[f(\Phi \cX)] - \Var[f(\cZ)] \rightarrow 0
  \quad\;\text{ as }n\rightarrow\infty
\end{equation*}
and the same holds in the unaugmented case where 
$\Phi \cX$ and $\cZ$ are replaced by $\tilde \cX$ and $\tilde \cZ$.
The asymptotic variances are
\begin{equation*}
  \Var f(\cZ)=V(\sigma)
  \quad\text{ and }\quad
  \Var f(\tilde\cZ) = V(\tilde\sigma)
  \quad\text{ and hence }\quad
  \vartheta(f) = \sqrt{V(\tilde\sigma)/V(\sigma)}\;.
\end{equation*}
For any  ${\alpha \in [0,1]}$, the lower and upper $\alpha/2$-th quantiles
for $f(\cZ)$ and $f(\tilde \cZ)$ are given by
     \begin{align*}
    \big(\exp\big(- \sigma^2\pi_u\big),&\,\exp\big(- \sigma^2 \pi_l\big)\big)
&\;\text{and}\;&&
\big(\exp(-\tilde\sigma^2\pi_u),&\,\exp(-\tilde\sigma^2\pi_l)\big)\;,
     \end{align*}
   where $\pi_u$ and  $\pi_l$ are the upper and lower $\alpha/2$-th quantiles of a $\chi_1^2$ random variable.
\label{prop:exp_neg_chi_squared:CI}
\end{proposition}

\begin{figure}[t]
\makebox[\textwidth][c]{
\begin{tikzpicture}
\node[inner sep=0pt] at (0, 0)
    {\includegraphics[width=.5\textwidth]{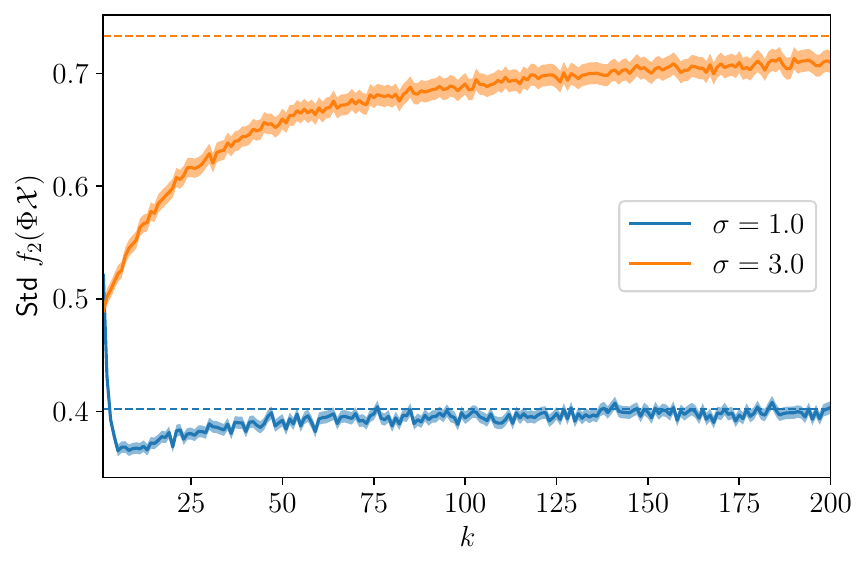}};
\node[inner sep=0pt] at (7.8,0)
    {\includegraphics[width=.5\textwidth]{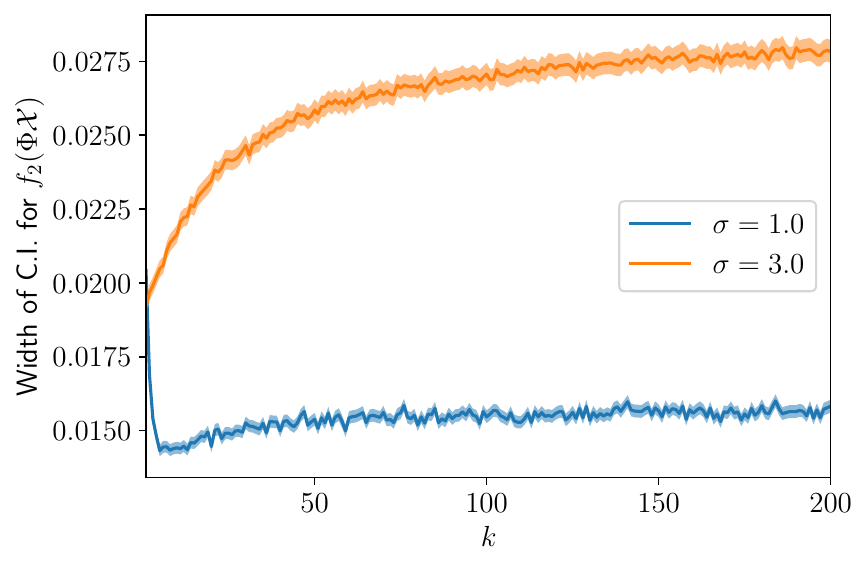}};
\end{tikzpicture}
}
\caption{Simulation for $f_2$ with $n=100$ and varying $k$. \emph{Left}: The standard 
deviation ${\rm Std} \, f_2(\Phi \cX)$. The dotted lines indicate the theoretical value of ${\rm Std} \, f_2(\cZ)$ computed in Lemma \ref{lem:exp_neg_chi_squared_2D}, in which we also verify the convergence of $f_2(\Phi\cX)$ to $f_2(\cZ)$ in $d_{\cH}$. \emph{Right}: Difference between $0.025$-th and $0.975$-th quantiles for $f_2(\Phi \cX)$. In all figures, shaded regions denote 95\% confidence intervals for simulated quantities.}

\label{fig:exp_neg_chi_squared_2D}
\end{figure}

As discussed in the main text, the behavior of $f$ under augmentation is more
complicated than that of averages as both $V(s)$ and ${D(s)\coloneqq\exp(-s^2\pi_l) - \exp(-s^2\pi_u)}$ are not monotonic. This phenomenon persists if we 
extends $f$ to two dimensions, by defining 
\begin{align} \label{eqn:defn:exp:neg:chi:sq:stat:2d} 
    f_2(\bx_{11}, \ldots,
\bx_{nk}) \;\coloneqq\; f(x_{111}, \ldots, x_{nk1})+f(x_{112}, \ldots,
x_{nk2})
    \;.
\end{align} 
Figure \ref{fig:exp_neg_chi_squared_2D}
shows results for 
\begin{align} \label{eqn:exp:neg:chi:sq:simulation:general}
    \bX_i \stackrel{i.i.d.}{\sim} \cN(\bzero, \sigma^2 \begin{psmallmatrix}
 1 & \rho \\ \rho & 1 \end{psmallmatrix})\;, -1 < \rho < 1\;,
    \quad\text{ and }\quad
    \phi_{ij} \stackrel{i.i.d.}{\sim} \text{Uniform}\{ \begin{psmallmatrix} 1 & 0 \\ 0 & 1 \end{psmallmatrix}, \begin{psmallmatrix} 0 & 1 \\ 1 & 0 \end{psmallmatrix} \}
\end{align}
under $\rho=0.5$. In this case, the data distribution is invariant under both possible
transformations. Thus, invariance does not guarantee augmentation to
be well-behaved.
 
\vspace{.5em}

For completeness, we also include \cref{lem:exp_neg_chi_squared_2D}, a result that confirms the applicability of 
 \cref{thm:main} 
 to $f_2$. We also compute an explicit formula for the variances of $f(\cZ)$ and $f(\tilde \cZ)$ under \eqref{eqn:exp:neg:chi:sq:simulation:general} for a general $\rho$.

\begin{lemma} \label{lem:exp_neg_chi_squared_2D} Under the setting \eqref{eqn:exp:neg:chi:sq:simulation:general}, the statistic $f_2$ defined in \eqref{eqn:defn:exp:neg:chi:sq:stat:2d} satisfies
\begin{proplist}
    \item as $n \rightarrow \infty$, $f_2(\Phi \cX) - f_2(\cZ) \xrightarrow{d} 0$ and $\|\Var[f_2(\Phi \cX)] - \Var[f_2(\cZ)]\| \rightarrow 0$, and the same holds with $(\Phi \cX, \bZ)$ replaced by the unaugmented data and surrogates $(\tilde \cX, \tilde \cZ)$;
    \item $\bZ_i$ has zero mean and covariance matrix 
    $$\sigma^2 \bI_k \otimes \begin{psmallmatrix} 1 & \rho \\ \rho & 1\end{psmallmatrix} + \mfrac{(1+\rho) \sigma^2}{2} (\bone_{k \times k} - \bI_k) \otimes \bone_{2 \times 2}\;,$$ 
    while $\tilde{\bZ}_i$ has zero mean and covariance matrix $\sigma^2 \bone_{k \times k} \otimes \begin{psmallmatrix} 1 & \rho \\ \rho & 1 \end{psmallmatrix}$;
    \item write $\sigma^2_- \coloneqq \frac{(1-\rho) \sigma^2}{2}$ and $\sigma^2_+ \coloneqq \frac{(1+\rho) \sigma^2}{2} $, then variance of the augmented data is given by
    \begin{align*}
        \Var[f_2(\cZ)] \;=&\; 2\Big(1+ \mfrac{4 \sigma_-^2}{k} + 4\sigma_+^2 \Big)^{-1/2} + 2 \Big( 1 + \mfrac{4 \sigma_-^2}{k} \Big)^{-1/2} (1+4\sigma_+^2)^{-1/2} 
        \\
        &\; - 4 (1+\mfrac{2 \sigma_-^2}{k} + 2\sigma_+^2)^{-1}\;.
    \end{align*}
    In particular, at $\rho=0.5$, $\lim_{k \rightarrow \infty} \Var[f_2(\cZ)]= 4(1 + 3\sigma^2 )^{-1/2} - 4 \big(1+ \frac{3}{2}\sigma^2 \big)^{-1}$.
\end{proplist}
\end{lemma}

\begin{remark} Note that (i) above only verifies the convergence under $n \rightarrow \infty$ with $k$ fixed. Nevertheless, one may easily check that $f_2$ satisfies the stronger Corollary \ref{cor:thm:main:smaller:var} corresponding to a smaller variance of $\bZ_i$ given by $\mfrac{(1+\rho) \sigma^2}{2} \bone_{2k \times 2k}$ as $n, k \rightarrow \infty$. In that case, the asymptotic variance of the statistic is given exactly by the formula $\lim_{k \rightarrow \infty} \Var[f_2(\cZ)]$ in (iii) above.
\end{remark}

\subsection{Additional results for ridgeless regressor} \label{appendix:ridgeless:results}
This section complements 
\cref{sec:ridgeless} 
and provides tools for simplifying the risk of ridgeless regressors. 

\vspace{1em}

\noindent\textbf{Notation}. For $A, B \in \R^{d \times d}$ symmetric and $\lambda \geq 0$, we denote
\begin{align*}
    f^{(1)}_\lambda(A)
    \;\coloneqq&\;
    \begin{cases} 
        \lambda^2 \beta^\top \big(A  + \lambda \bI_d \big)^{-2} \beta 
        & 
        \text{ for } \lambda > 0
        \\
        \big\| \big( A^\dagger A  - \bI_d \big) \beta \big\|^2
        & 
        \text{ for } \lambda = 0
    \end{cases}
    \;,
    \\
    f^{(2)}_\lambda(A, B)
    \;\coloneqq&\;
        \mfrac{\sigma_\epsilon^2}{n} \,
        \Tr\big( \big(A + \lambda \bI_d \big)^{-2} 
        B \big) 
    \;,
    \qquad 
    f_\lambda(A,B) \;\coloneqq\;  f^{(1)}_\lambda(A) + f^{(2)}_\lambda(A, B)\;,
\end{align*}
where $(\argdot)^{-2}$ is a shorthand for the square of the pseudoinverse $(\argdot)^\dagger$. Observe that by a standard bias-variance decomposition as in \cite{hastie2022surprises}, the risk under the oracle augmentations can be expressed as, for both the case $\lambda > 0$ and the case $\lambda = 0$,
\begin{align*}
    \hat L^{(\rm ora)}_{\lambda}
    \;=&\;
    \big\| \mean\big[ \hat \beta^{(\rm ora)}_{\lambda}(\cX) \big| \cX \big] - \beta \big\|^2 
    + \Tr\big[ \Cov\big[ \hat \beta^{(\rm ora)}_{\lambda}(\cX) \big| \cX \big]  \big]
    \\
    \;=&\; 
    \big\| \big( \big( \bar \bX_1 + \lambda \bI_d \big)^\dagger \bar \bX_1  - \bI_d \big) \beta \big\|^2
    + 
    \mfrac{\sigma_\epsilon^2}{n} \, \Tr\big( 
    \big( \bar \bX_1  + \lambda \bI_d \big)^\dagger
     \bar \bX_2 
    \big( \bar \bX_1 + \lambda \bI_d \big)^\dagger
    \big) 
    \\
    \;=&\;
    f^{(1)}_\lambda(\bar \bX_1) + f^{(2)}_\lambda(\bar \bX_1, \bar \bX_2)
    \;=\;
    f_\lambda(\bar \bX_1, \bar \bX_2)\;.
\end{align*}
Throughout, we write $\be_l$ as the $l$-th standard basis vector of $\R^d$ and denote $X_{ijl}$ as the $l$-th coordinate of $\pi_{ij} V_i$. 

\vspace{1em}

\noindent\textbf{The general case}.  The next lemma approximates $f_\lambda(\bar \bX_1, \bar \bX_2)$ by $f_0(\bar \bX_1, \bar \bX_2)$ in the Lévy–Prokhorov metric $d_P$ defined in \eqref{eq:defn:prokhorov}. The proof exploits the assumption below on the distribution of the extreme eigenvalues of $\bar \bX_1$, $\bar \bX_2$, $\bar \bZ_1$ and $\bar \bZ_2$, as well as the alignment of their zero eigenspace.

\begin{lemma} \label{lem:ridgeless:by:ridge} Under 
     \cref{assumption:ridgeless:by:ridge}, 
     if $d= O(n)$ and $\lambda > 0$, then
    \begin{align*}
        d_P\big(  f^{(1)}_\lambda(\bar \bX_1, \bar \bX_2)  \,,\, f^{(1)}_0(\bar \bX_1, \bar \bX_2)  \big)
        \;=&\; 
        O_{\gamma'} ( \lambda^2  )\;,
        \\
        d_P\big(  f^{(2)}_\lambda(\bar \bX_1, \bar \bX_2)  \,,\, f^{(2)}_0(\bar \bX_1, \bar \bX_2)  \big)
        \;=&\; 
        O_{\gamma'} \Big( \lambda + \mfrac{1}{n \lambda^2}  \Big)\;,
        \\
        d_P \Big( f_\lambda(\bar \bX_1, \bar \bX_2) 
                    \,,\, 
                    f_0(\bar \bX_1, \bar \bX_2) 
              \Big)
       \;=&\;
       O_{\gamma'}\Big( \lambda^2 + \lambda + \mfrac{1}{n \lambda^2}  \Big)\;,
       \\
       d_P \Big( f_\lambda(\bar \bZ_1, \bar \bZ_2) 
                    \,,\, 
                    f_0(\bar \bZ_1, \bar \bZ_2) 
              \Big)
       \;=&\;
       O_{\gamma'}\Big( \lambda^2 + \lambda + \mfrac{1}{n \lambda^2}  \Big)\;,
    \end{align*}
    where $O_{\gamma'}$ indicates that the bounding constant is allowed to depend on $\gamma$.
\end{lemma}

\noindent\textbf{The isotropic case}. In the isotropic case, one may exploit the property of Gaussians to express $\bar \bZ_1$ and $\bar \bZ_2$ explicitly in terms of the same rectangular Gaussian matrix. This allows the risk to be completely characterized by moments and Stieltjes transforms of the Marchenko-Pastur law under appropriate transformations, and simplifies how the two strongly correlated matrices affects the risk. The risk formula then extends to the non-Gaussian case by our universality results. The alternative expression for $\bar \bZ_1$ below also formally justifies 
\eqref{eq:dd:noise:inject:key} 
in the discussion in the main text.

\begin{lemma}(Alternative expression of $\bar \bZ_1$) \label{lem:ridgeless:isoG:alt:rep} Assume 
     \eqref{eq:ridgeless:isoG}. 
    Fix any mutually orthogonal unit vectors $\bv_1, \ldots, \bv_{k-1} \in \R^k$ such that the sum of coordinates of each $\bv_i$ equals zero. Consider the orthogonal matrix $Q_k \in \R^{k \times k}$ and the diagonal matrix $D_k \in \R^{k \times k}$, defined as 
    \begin{align*}
        Q_k 
        \;\coloneqq&\; 
        \begin{psmallmatrix}
            k^{-1/2} & \hdots & k^{-1/2} \\
            \leftarrow  & \bv_1^\top & \rightarrow \\
            & \vdots & \\
            \leftarrow  & \bv_{k-1}^\top & \rightarrow \\
        \end{psmallmatrix}
        &\text{ and }&&
        D_k 
        \;\coloneqq&\;
        \begin{psmallmatrix}
            (k+\sigma_A^2)/k & & \\
            & \sigma_A^2/k & & \\
            & & \ddots & \\
            & & & & \sigma_A^2/k \\
        \end{psmallmatrix}
        \;.
    \end{align*}
    Also define the $\R^{nk \times n}$ matrix
    \begin{align*}
        K \;\coloneqq\; \mfrac{1}{\sqrt{k}} \bI_n \otimes \bone_k \;=\; \mfrac{1}{\sqrt{k}} \begin{psmallmatrix}
            1 & \hdots & 1 & & & & & & & \\
            &        &   & 1 & \hdots & 1 & & & & \\
            &        &   &  &  & & \ddots & & & \\
            &        &   &  &  & &  & 1 & \hdots & 1 \\
        \end{psmallmatrix}^\top\;.
    \end{align*}
    Then almost surely,
    \begin{align*}
        \bar \bZ_1 \;=&\; \mfrac{1}{n} \bH \, \big( \bI_n \otimes D_k \big) \bH^\top
        &\text{ and }&&
        \bar \bZ_2 \;=&\; \mfrac{1}{n} \bH \, \big( \bI_n \otimes D_k^{1/2} Q_k \big) K K^\top \big( \bI_n \otimes Q_k^\top D_k^{1/2} \big) \bH^\top \;,
    \end{align*}
    for some $\bH$ that is an $\R^{d \times nk}$ matrix with i.i.d.~standard Gaussian entries. As a consequence, we have 
    \begin{align*}
        \bar \bZ_1
        \;=\; 
        \mfrac{1}{n} \msum_{i=1}^n \Big( \eta_{i1} \eta_{i1}^\top + \mfrac{\sigma^2_A}{k} \msum_{j=1}^k \eta_{ij} \eta_{ij}^\top \Big)
        \;=\;
        \bar \bZ_2 
        +
        \mfrac{\sigma^2_A}{nk} \msum_{i=1}^n
        \msum_{j=2}^k \eta_{ij} \eta_{ij}^\top
    \end{align*}
    almost surely for some i.i.d.~standard Gaussian vectors $\eta_{ij}$ in $\R^d$.
\end{lemma}

\vspace{.5em}

The next result verifies 
 Assumptions \ref{assumption:ridgeless:universal} and \ref{assumption:ridgeless:by:ridge} 
 for isotropic Gaussian data.

\begin{lemma} \label{lem:ridgeless:isoG:assumption} Suppose $\bX_i \sim \cN(\bzero,\bI_d)$ and $\xi_{ij} \sim \cN(\bzero, \sigma_A^2 \bI_d)$, and consider the asymptotic 
    \eqref{eq:ridgeless:asymptotic:regime} 
    with $\gamma' = \lim d / (kn) \neq 1$. Then xzj
    Assumptions \ref{assumption:ridgeless:universal} and \ref{assumption:ridgeless:by:ridge}
     hold.
\end{lemma}

\subsection{Additional results on nonlinear feature models and simple neural networks in Section \ref{sec:network} }  \label{appendix:network:extend} 

\subsubsection{Locally dependent nonlinear feature model.} 
 \label{appendix:locally:dependent:feature} 

We first present a slight generalization of the universality result of \Cref{prop:network} under a locally dependent nonlinear feature model (\Cref{assumption:network:generalize}).  
The proof of \Cref{prop:network} will then consist of verifying \Cref{assumption:network:generalize} for the specific augmentation schemes used in \Cref{prop:network}. 

\begin{assumption}(Locally dependent nonlinear feature model) \label{assumption:network:generalize} (i) Fix $\beta \in \R^p$ with $\| \beta \| = O(1)$ and let $\epsilon_i$'s be i.i.d.~mean-zero with $\Var[\epsilon_i] = \sigma^2_\epsilon$. Let $(\bV_{ij1}, \bV_{ij0})_{i \leq n, j \leq k}$ be some possibly dependent $\R^d$ random vectors. For some thrice-differentiable function $\varphi_\theta: \R^d \rightarrow \R^{p'}$, parameterized by a random variable $\theta$ in $\R^{b'}$ independent of all other variables, we generate the input vectors
\begin{align*}
     \tilde \bV_{ij} \;\coloneqq&\;  \varphi_\theta \big(  \bV_{ij1} \big)  \;,
\end{align*} 
and for an $\R^{p \times p'}$-valued random matrix  $\bW^{(0)}$  with i.i.d.~$\cN(0, 1/p')$ entries and a thrice-differentiable function $\varphi_{\theta_0}: \R^d \rightarrow \R^{p'}$, parameterized by a random variable $\theta_0 \in \R^{b'}$ independent of all other variables, we generate the output variables
\begin{align*}
    \tilde Y_{ij} \;\coloneqq\; \beta^\top \bW^{(0)} \tilde \bV_{ij}^0 + \epsilon_i\;,
    \quad 
    \tilde \bV_{ij}^0 \;\coloneqq\; \varphi_{\theta_0}(\bV_{ij0})\;.
\end{align*}

\noindent
(ii) The estimator with ridge parameter $\lambda > 0$ is specified as 
\begin{align*}
    \hat \beta_\lambda(\cX)
    \;\coloneqq&\; 
    \argmin_{\tilde \beta \in \R^p} 
    \mfrac{1}{nk} \msum_{i=1}^n  \msum_{j=1}^k
    \big( \tilde Y_{ij} - \tilde \beta^\top \tilde \bV_{ij}  \big)^2
    +
    \lambda \| \tilde \beta\|^2\;,
\end{align*}

 The ridgeless estimator is similarly specified as $\hat \beta_0 = \lim_{\lambda \rightarrow 0^+} \hat \beta_\lambda$;

\noindent 
(iii) \textbf{Block dependence across $i$.} The data blocks $(\bV_{ijr})_{j \leq k, 0 \leq r \leq 1}$ are i.i.d. across $i \leq n$;

\noindent 
(iv) \textbf{Local dependence across coordinates and augmentations.} For $j \leq k$, $0 \leq r \leq 1$ and $l \leq d$, write the dependency neighborhood of the $l$-th coordinate of $ \bV_{1jr}$, $( \bV_{1jr})_l$, as 
\begin{align*}
    \cB_{j,r,l}
    \;\coloneqq\; 
    \inf
    \big\{ 
        \cB \subseteq [k] \times \{0,1\} \times [d] \;\big|\; 
       &\; (j,r,l) \in \cB \text{ and }
        (( \bV_{1j'r'})_{l'})_{(j',r',l') \in \cB} 
        \\
      &\; \text{ is independent of }
        (( \bV_{1j'r'})_{l'})_{(j',r',l') \not\in \cB}
    \big\} 
    \;.
\end{align*}
We assume that the maximum size of the local dependency neighborhood satisfies the following bound:
\begin{align*}
    B_d \;\coloneqq\; \mmax_{j \leq k, \, r\leq 2, \,  l \leq d} \, | \cB_{j,r,l} | \;=\; O(  d^{1/2} ) \;.
\end{align*}

\noindent 
(v) \textbf{Sub-Gaussianity. } We assume that the random vectors $(\tilde \bV_{ij}, \tilde \bV_{ij}^0, \bV_{ijr})_{i \leq n, j \leq k, 0 \leq r \leq 1}$ are all mean-zero and $\sigma_V$-sub-Gaussian for some absolute constant $\sigma_V < \infty$.
\end{assumption}

To specify the test risk, we let $\bV_{\rm new}$ be an $\R^d$ random vector independent of all other variables, and let 
\begin{align*}
    Y_{\rm new} \;\coloneqq\; \beta^\top \bW^{(0)} \varphi_{\theta_0}(\bV_{\rm new}) 
    + \epsilon_{\rm new}
    \;,
\end{align*}
where $\epsilon_{\rm new}$ is an i.i.d.~copy of $\epsilon_1$.
Analogous to \eqref{eqn:network:risk}, we study the risk
\begin{align*}
  \hat L_{\lambda}(\cX) 
  \;\coloneqq&\;
  \mean\big[ 
   \big( 
     \hat \beta_{\lambda}(\cX)^\top \varphi_\theta( \bV_{\rm new} ) 
     -
     Y_{\rm new}
   \big)^2  
   \,\big| \, \cX, \bW^{(0)} \big]
  \qquad\text{ for }\lambda\geq0\;,
  \tagaligneq \label{eqn:network:risk:generalize}
\end{align*}
where we condition on both the input data $\cX = ( \bV_{ijr} )_{i \leq n, j \leq k, 0 \leq r \leq 1}$ and the random weights in the model $\bW^{(0)}$. The risk can be computed explicit as was done in \Cref{appendix:ridgeless:results}, but with respect to sample covariance matrices that are analogous but slightly different from $\bar \bX_1$ and $\bar \bX_2$ in \Cref{sec:triple:descent:oracle}. The next lemma computes this risk. We shall use the following shorthands:
\begin{align*}
    \bar \bX^*_1 
    \;\coloneqq&\; 
    \mfrac{1}{nk} \msum_{i=1}^n \msum_{j=1}^k  \tilde \bV_{ij} (\tilde \bV_{ij})^\top
    \;,
    \qquad
    \bar \bX^*_3
    \;\coloneqq\; 
    \mfrac{1}{nk} \msum_{i=1}^n \msum_{j=1}^k \tilde \bV_{ij} (\tilde \bV_{ij}^0)^\top
    \;,
    \\
    \bar \bX^*_2 
    \;\coloneqq&\;
    \mfrac{1}{n} \msum_{i=1}^n 
    \Big( \mfrac{1}{k} \msum_{j=1}^k  \tilde \bV_{ij} \Big)
    \,
    \Big( \mfrac{1}{k} \msum_{j=1}^k  \tilde \bV_{ij} \Big)^\top 
    \;,
    \;
    \bar \bX^{*;-1}_{1;\lambda}
    \;\coloneqq\;
    \begin{cases}
        (    \bar \bX^*_1 + \lambda \bI_p )^{-1}
        & \text{ for } \lambda > 0 \;,
        \\
        (    \bar \bX^*_1 )^\dagger
        & \text{ for } \lambda = 0 \;.
    \end{cases}
\end{align*}

\begin{lemma} \label{lem:network:risk} Under \Cref{assumption:network:generalize}, we have 
\begin{align*}
    \hat L_\lambda(\cX) 
    \;=&\;
    \beta^\top 
    \bW^{(0)}
    (\bar \bX^*_3)^\top 
    \bar \bX^{*;-1}_{1;\lambda} 
    \,M^{\varphi_\theta} \,
    \bar \bX^{*;-1}_{1;\lambda} 
    (\bar \bX^*_3  ) (\bW^{(0)})^\top 
    \beta  
    \\
    &\;
    +    
    \mfrac{\sigma^2_\epsilon}{n} \,
    \Tr\big(\,
    \bar \bX^{*;-1}_{1;\lambda} 
    \,M^{\varphi_\theta} \,
    \bar \bX^{*;-1}_{1;\lambda} 
    \,
    \bar \bX^*_2
    \,\big)
    \\
    &\;
    -
    2
    \beta^\top 
    \bW^{(0)}
    (\bar \bX^*_3)^\top 
    \bar \bX^{*;-1}_{1;\lambda} 
    \, R^{\varphi_\theta, \varphi_{\theta_0}} \,
    \bW^{(0)} \beta 
    \\
    &\;
    +
    \beta^\top \bW^{(0)} M^{\varphi_{\theta_0}} (\bW^{(0)})^\top \beta
    +
    \sigma_\epsilon^2
    \;,
\end{align*}
where we have defined 
\begin{align*}
    &\;
     M^{\varphi_\theta} \;=\; \mean \big[ \, \varphi_\theta(\bV_{\rm new})  \, \varphi_\theta(\bV_{\rm new})^\top \,  \big]
     \;,
     \quad\quad
     R^{\varphi_\theta, \varphi_{\theta_0}} \;=\; \mean \big[ \, \varphi_\theta(\bV_{\rm new})  \,  \varphi_{\theta_0}(\bV_{\rm new})^\top \,  \big]\;,
     \\
     &\;
     M^{\varphi_{\theta_0}} \;=\; \mean \big[ \, \varphi_{\theta_0}(\bV_{\rm new})  \,  \varphi_{\theta_0}(\bV_{\rm new})^\top \,  \big]
     \;.
\end{align*}
\end{lemma}

From now onwards, we make the following assumption, which implies that the operator norms of $M^{\varphi_\theta}$, $M^{\varphi_{\theta_0}}$ and $R^{\varphi_\theta, \varphi_{\theta_0}}$ are all $O(1)$:

\begin{assumption} \label{assumption:network:bounded:cov} The following quantities are $O(1)$:
\begin{align*}
    \| \mean[\varphi_{\theta_0}(\bV_{\rm new}) \varphi_{\theta_0}(\bV_{\rm new})^\top] \|_{op} 
    \,,\,
    \qquad
    \| \mean[\varphi_\theta(\bV_{\rm new}) \varphi_\theta(\bV_{\rm new})^\top]\|_{op} 
    \;.
\end{align*}
\end{assumption}

Analogously to \eqref{eq:network:asymptotic:regime}, we consider the asymptotic regime where
\begin{align*}
  &\;
  n, \, d, \, p', p \, \rightarrow  \, \infty\,,
  \quad 
  k \text{ is fixed },
  \\
  &\;
  d/n 
  \rightarrow \gamma_0 \in [0, \infty)\,,
  \quad
  d/(kn) \rightarrow \gamma'_0 \in [0, \infty)\,,
  \\
  &\;
  p'/n 
  \rightarrow \gamma_1 \in [0, \infty)\,,
  \quad
  p'/(kn) \rightarrow \gamma'_1 \in [0, \infty)\,,
  \\
  &\;
  p/n 
  \rightarrow \gamma_2 \in [0, \infty)\,,
  \quad
  p/(kn) \rightarrow \gamma'_2 \in [0, \infty)
  \,. \tagaligneq \label{eq:network:asymptotic:regime:generalise}
\end{align*}
We shall show Gaussian universality with respect to the covariates $\cX = ( \bV_{ijr})_{1 \leq i \leq n, 1\leq j \leq k, 0 \leq r \leq 1}$. We denote $\cZ=(\bZ_{ijr})_{i,j,r}$ as the Gaussian surrogates for $( \bV_{ijr})_{i,j,r}$, and write 
\begin{align*}
    \tilde \bZ_{ij} \;\coloneqq&\; \varphi_\theta(\bZ_{ij1}) 
    &\text{ and }&&
    \tilde \bZ_{ij}^0 \;\coloneqq&\; \varphi_{\theta_0}(\bZ_{ij0})
    \quad \text{ for } 1 \leq j \leq k 
    \;.
\end{align*}
In view of the risk formula above, the proof for universality boils down to replacing $\bar \bX^*_1$, $\bar \bX^*_2$, $\bar \bX^*_3$ and $\bar \bX^{*;-1}_{1;\lambda}$ by
\begin{align*}
    \bar \bZ^*_1 
    \;\coloneqq&\; 
    \mfrac{1}{nk} \msum_{i=1}^n \msum_{j=1}^k  \tilde \bZ_{ij} (\tilde \bZ_{ij})^\top
    \;,
    \qquad
    \bar \bZ^*_3
    \;\coloneqq\; 
    \mfrac{1}{nk} \msum_{i=1}^n \msum_{j=1}^k \tilde \bZ_{ij} (\tilde \bZ_{ij}^0)^\top
    \;,
    \\
    \bar \bZ^*_2 
    \;\coloneqq&\;
    \mfrac{1}{n} \msum_{i=1}^n 
    \Big( \mfrac{1}{k} \msum_{j=1}^k  \tilde \bZ_{ij} \Big)
    \,
    \Big( \mfrac{1}{k} \msum_{j=1}^k  \tilde \bZ_{ij} \Big)^\top 
    \;,
    \;\;
    \bar \bZ^{*;-1}_{1;\lambda}
    \;\coloneqq\;
    \begin{cases}
        (    \bar \bZ^*_1 + \lambda \bI_p )^{-1}
        & \text{ for } \lambda > 0 \;,
        \\
        (    \bar \bZ^*_1 )^\dagger
        & \text{ for } \lambda = 0 \;.
    \end{cases}
\end{align*}
The bounds are stated in terms of the following gradient terms of the feature map $\varphi$ and $\varphi_{\theta_0}$: For $r = 1,2,3$, we define 
\begin{align*}
        \gamma_r^\varphi \;\coloneqq&\; \max\big\{ 
        \msup_{\bx \in \R^d} 
        \| \, \| \partial^r \varphi_{\theta_0}(\bx) \|_{op} \, \|_{L_9}
         \,,\,  
        \msup_{\bx \in \R^d} 
        \| \, \| \partial^r \varphi_\theta(\bx) \|_{op}  \,\|_{L_9}
        \big\}\;,
\end{align*}
where for a linear map $T_r: \R^{d^r} \rightarrow \R^{p'}$, we have denoted 
\begin{align*}
    \| T_r \|_{op} \;\coloneqq&\; \sup_{\substack{x_1,\ldots,x_r \in \R^d, y \in \R^{p'} \\ \| x_1 \| = \ldots = \|x_r\| = \|y\| =1 }  }  \big| y^\top T_r(x_1 \otimes \ldots \otimes x_r) \big| \;.
\end{align*}
Note that for $r=1$ and $d=p'$, this recovers the usual operator norm for a symmetric matrix. The next assumption restricts how fast the derivatives of these feature maps are allowed to grow, relative to the maximum size of the local dependency neighborhood $B_d$:

\begin{assumption} \label{assumption:network:feature:gradient} Define $B_d$ as in \Cref{assumption:network:generalize}(iv). We assume the following:
\begin{align*}
    \gamma_1^\varphi \;=\; o\big( B_d^{-1/3} d^{1/6} \big) 
    \,,\, 
    \quad 
    \gamma_2^\varphi \;=\; o\big( B_d^{-2/3} d^{ - 1/6} \big) 
    \,,\,
    \quad 
    \gamma_3^\varphi \;=\; o\big( B_d^{-1}  d^{ - 1/2 } \big) \;.
\end{align*}
\end{assumption}

\begin{remark} Note that the conditions on $\gamma_2^\varphi$ and $\gamma_3^\varphi$ restrict the amount of non-linearity $\varphi$ and $\varphi_{\theta_0}$ can have. In \Cref{appendix:bagged:network}, we show that these conditions can be relaxed by bagging.
\end{remark}

 To relate universality of the ridge estimator ($\lambda > 0$) to that of the ridgeless one ($\lambda \rightarrow 0^+$), for the linear case with noise injection, we have applied \Cref{assumption:ridgeless:by:ridge}. Here, we invoke a similar condition:

\begin{assumption} \label{assumption:network:ridgeless:by:ridge} The following quantities are $O(1)$ with probability $1 - o(1)$:
  \begin{align*}
      &\;
      \| (\bar \bX_1^*)^\dagger \|_{op} 
      \;,\;
      \quad
      \| (\bar \bZ_1^*)^\dagger \|_{op} 
      \;,\;
      \quad 
      \| \bar \bX_2^* \|_{op}
      \;,\;
      \quad 
      \| \bar \bZ_2^* \|_{op} 
      \;,\;
      \quad 
      \| \bar \bX_3^* \|_{op}
      \;,\;
      \quad 
      \| \bar \bZ_3^* \|_{op}
      \;,
      \\
      &\;
      \msum_{l=1}^d  \ind_{\{ \lambda_l(\bar \bX_1^*) = 0 \}} \big( v_l(\bar \bX_1^*)^\top \bar \bX^*_2 \, v_l(\bar \bX_1^*) \big)
      \;,\;
      \quad 
      \msum_{l=1}^d  \ind_{\{ \lambda_l(\bar \bZ_1^*) = 0 \}} \big( v_l(\bar \bZ_1^*)^\top \bar \bZ_2^* \, v_l(\bar \bZ_1^*) \big) \;,
  \end{align*}
  where $(\lambda_l(A), v_l(A))$ denotes the $l$-th eigenvalue-eigenvector pair of a matrix $A \in \R^{d \times d}$, and we have denoted $\| A \|_{op} = \sup_{v \in \R^b, x \in \R^d; \| v \| = \|x \| =1} | v^\top A x |$  for $A \in \R^{b \times d}$. Moreover, the following quantities are $o(1)$ with probability $1-o(1)$:
  \begin{align*}
        \Big\| 
        \msum_{l=1}^d  \ind_{\{ \lambda_l(\bar \bX_1^*) = 0 \}} \bar \bX^*_3 \, v_l(\bar \bX_1^*) v_l(\bar \bX_1^*)^\top 
        \Big\|_{op}
      \;,\;
      \quad 
      \Big\| 
        \msum_{l=1}^d  \ind_{\{ \lambda_l(\bar \bZ_1^*) = 0 \}} \bar \bZ^*_3 \, v_l(\bar \bZ_1^*) v_l(\bar \bZ_1^*)^\top 
        \Big\|_{op}
        \;.
  \end{align*}
\end{assumption}

\begin{remark} Compared to \Cref{assumption:ridgeless:by:ridge}, we additionally require two operator norms to be $o(1)$ with high probability. These norms control the size of $\bar \bX_3^*$ in the zero-eigenspace of $\bar \bX^*_1$. In the unaugmented case as well as the augmentation considered in \Cref{sec:triple:descent:oracle}, $\bar \bX^*_3$ is exactly $\bar \bX^*_1$, which allow these two norms to be exactly zero. We conjecture that this condition is improvable at the expense of more involved techniques for the ridgeless case, and leave it to future work.
\end{remark}

Finally in the result below, we use $\cH^{(4)} \subset \cH$ to denote the class of four-times continuously differentiable function with its first four derivatives uniformly bounded from above by $1$.

\begin{proposition} \label{prop:network:extend:general} Fix $\lambda > 0$. Under \Cref{assumption:network:generalize,assumption:network:bounded:cov,assumption:network:feature:gradient} and the asymptotic \eqref{eq:network:asymptotic:regime:generalise},
\begin{align*}
    d_{\cH^{(4)}}\big( 
        \hat L_\lambda(\cX)
        \,,\,
        \hat L_\lambda(\cZ)
    \big)
    \;=\;
    o 
    \Big( 
        \Big( 
            1
            +
            \mfrac{1}{\lambda^6} 
        \Big)
    \Big)
    \;.
\end{align*}
If additionally \Cref{assumption:network:ridgeless:by:ridge} holds, then
\begin{align*}
    d_P\big( 
        \hat L_0(\cX)
        \,,\,
        \hat L_0(\cZ)
    \big)
    \;=\;
    o(1)
    \;.
\end{align*}
\end{proposition}

\subsubsection{Ridgeless version of \Cref{prop:network} on linear networks.}   \label{appendix:network:ridgeless} 

We follow the notation of \Cref{sec:network} and recall that $\hat L_0(\Phi\cX)$ is the test risk of the augmented ridgeless regressor. The additional condition required to prove universality of $\hat L_0(\Phi\cX)$ is exactly a re-expression of \Cref{assumption:network:ridgeless:by:ridge} above:

\begin{assumption} \label{assumption:network:ridgeless:by:ridge:linearnet} Define $\bW^{(0)}_l$, $W_l$, $\bV_i$, $\pi_{ij}$ and $\tau_{ij}$ as in \Cref{assumption:network:data,assumption:network:augmentation}. Suppose \Cref{assumption:network:ridgeless:by:ridge} holds, where we identify 
\begin{align*}
    \varphi_{\theta_0}( \bv ) \;=\;  \bW^{(0)}_{N_0 -1 } \ldots \bW^{(0)}_1 \bv \;,
    \quad 
    \varphi_\theta( \bv ) \;=\; W_N \ldots W_1 \bv \;,
    \quad 
    \bV_{ij1} \;=\; \pi_{ij}(\bV_i) \;,
\end{align*}
and that $\bV_{ij0} = \bV_i$ if $\tau_{ij}$ is identity a.s.~or $\bV_{ij0} = \pi_{ij}(\bV_i)$ if $\tau_{ij}$ is the oracle augmentation.
\end{assumption}

Since \Cref{prop:network} is proved by verifying the conditions of the first statement of \Cref{prop:network:extend:general} above, the addition of \Cref{assumption:network:ridgeless:by:ridge:linearnet} allows us to conclude the following directly:

\begin{corollary} \label{cor:network:ridgeless} Assume the setup of \Cref{prop:network}. If additionally \Cref{assumption:network:ridgeless:by:ridge:linearnet} holds, then under the asymptotic \eqref{eq:network:asymptotic:regime},
\begin{align*}
    d_P\big( 
        \hat L_0(\Phi\cX)
        \,,\,
        \hat L_0(\cZ)
    \big)
    \;\rightarrow\; 
    0
    \;.
\end{align*}
\end{corollary}

\subsubsection{Analysis of double-descent peak under augmentations beyond isotropic noise injection} \label{appendix:peak:correlate}

In this section, to demonstrate the effect of coordinate dependence on the double descent peaks, we analyze theoretically and numerically the behavior of the oracle ridgeless estimators from \Cref{sec:triple:descent:oracle},
\begin{align*}
    \hat{\beta}_0^{(\rm ora)}
    \;\coloneqq&\
    \Big( \mfrac{1}{nk} \msum_{ij} (\pi_{ij} \bV_{i})(\pi_{ij} \bV_{i})^\top \Big)^\dagger \mfrac{1}{nk} \msum_{ij} (\pi_{ij} \bV_{i}) \, \tau^{(\rm ora)}_{ij} Y_i
    \;.
\end{align*}
under the augmentation schemes in \Cref{assumption:network:augmentation} of \Cref{sec:network}. 
We also analyze numerically the behavior of 
\begin{align*}
    \hat{\beta}_0^{(\rm id)}
    \;\coloneqq&\
    \Big( \mfrac{1}{nk} \msum_{ij} (\pi_{ij} \bV_{i})(\pi_{ij} \bV_{i})^\top  \Big)^\dagger \mfrac{1}{nk} \msum_{ij} (\pi_{ij} \bV_{i}) \, Y_i
    \;.
\end{align*}
The two estimators correspond to the two ways of augmenting $Y_i$'s in \Cref{assumption:network:augmentation}. A theoretical analysis of $\hat{\beta}_0^{(\rm id)}$ is possible but analogous to that of $\hat{\beta}_0^{(\rm ora)}$ with more complicated notation, and hence omitted in this appendix.

\vspace{.5em}

For $\hat{\beta}_0^{(\rm ora)}$, we can deduce from its risk formulas (see e.g.~\Cref{appendix:ridgeless:results} as well as the formulas for the unaugmented case in \cite{hastie2022surprises}) that, the component of the risk that potentially diverges is the variance term
\begin{align*}
    \mfrac{\sigma^2_\epsilon}{n} \, \Tr\Big(   \bar \bZ_1^\dagger \, \bar \bZ_2  \, \bar \bZ_1^\dagger \Big)
    \;,
    \tagaligneq \label{eq:ridgeless:risk:var:term}
\end{align*}
where we have replaced $\bar \bX_1$ and $\bar \bX_2$ by the corresponding Wishart matrices under universality:
\begin{align*}
    \bar \bZ_1 \;=\; \mfrac{1}{nk} \msum_{i \leq n} \msum_{j \leq k} \bZ_{ij} \bZ_{ij}^\top\;,
    \qquad 
    \bar \bZ_2 \;=\; \mfrac{1}{n} \msum_{i \leq n} \Big( \mfrac{1}{k} \msum_{j \leq k} \bZ_{ij} \Big)  \Big( \mfrac{1}{k} \msum_{j \leq k} \bZ_{ij} \Big)^\top\;,
\end{align*}
where $(\bZ_{ij})_{i,j}$ are the Gaussian surrogates for $(\pi_{ij}\bV_i)$. Note that these are the analogues of $\bar \bZ^*_1$ and $\bar \bZ^*_2$ considered in the nonlinear feature model (\Cref {appendix:locally:dependent:feature}) and neural network model (\Cref{appendix:network:ridgeless}) setups before, if we set $\varphi_\theta$ to be the identity map and $N=0$ respectively. Here, we choose to analyze $\bar \bZ_1$ and $\bar \bZ_2$ under the augmentation schemes in \Cref{assumption:network:augmentation}, as it provides the clearest comparison to the isotropic noise injection analysis in \Cref{sec:triple:descent:oracle}. 

\vspace{.5em}

In the discussion in \Cref{sec:triple:descent:oracle}, we have analyzed the double-descent curve by examining the stability of the pseudoinverse $\bar \bZ_1^\dagger$. We will see that for certain augmentations such as random cropping and sign-flipping, a similar analysis of $\bar \bZ_1^\dagger$ suffices, whereas for augmentations that introduce more complicated coordinate-dependence such as correlated noise injection and permutations, a slightly more involved argument is needed to examine the interaction between $\bar \bZ_1$ and $\bar \bZ_2$. Nevertheless, all arguments proceed by analyzing $(\bar \bZ_1, \bar \bZ_2)$ as a linear combination of Wishart matrices, which is made possible by universality.

\vspace{.5em}

The next lemma is analogous to \Cref{lem:ridgeless:isoG:alt:rep} and provides alternative expressions of $\bar \bZ_1$ and $\bar \bZ_2$ for the non-isotropic setup.

\begin{lemma} \label{lem:ridgeless:noniso:alt:rep} (Alternative expressions of $\bar \bZ_1$, non-isotropic setup) Write 
\begin{align*}
    \Sigma_1 
    \;\coloneqq&\;
    \Var[\pi_{11}(\bV_1)] + (k-1)  \Cov[\pi_{11}(\bV_1), \pi_{12}(\bV_1)]
    \;,
    \\
    \Sigma_2
    \;\coloneqq&\;
    \Var[\pi_{11}(\bV_1)] -  \Cov[\pi_{11}(\bV_1), \pi_{12}(\bV_1)]
    \;.
\end{align*}    
Let  $\eta_{ij}$'s be i.i.d.~$\cN(0,I_d)$ vectors. Then $(\bar \bZ_1, \bar \bZ_2)$ is identically distributed as
\begin{align*}
    \Big( 
        &\,
        \mfrac{1}{nk} \msum_{i \leq n} 
        \Sigma_1^{1/2}
        \; \eta_{i1} \eta_{i1}^\top \; 
        \Sigma_1^{1/2}
    +
    \mfrac{1}{nk} \msum_{i \leq n} \msum_{j=2}^k
    \Sigma_2^{1/2} \; \eta_{ij} \eta_{ij}^\top \; \Sigma_2^{1/2}
    \;,
    \\
    &\,
    \mfrac{1}{n} \msum_{i \leq n} 
    \Big( \mfrac{1}{k} \Sigma_1^{1/2} \; \eta_{i1} + \mfrac{1}{k} \msum_{j = 2}^k \Sigma_2^{1/2} \; \eta_{ij}  \Big)
    \Big( \mfrac{1}{k} \Sigma_1^{1/2} \; \eta_{i1} + \mfrac{1}{k} \msum_{j = 2}^k \Sigma_2^{1/2} \; \eta_{ij}  \Big)^\top
    \Big)
    \;.
\end{align*}
\end{lemma}

\begin{remark} Two remarks are in order:
\\[.5em]    
(i) As a sanity check, we recall that in the isotropic setup \eqref{eq:ridgeless:isoG}, $\Var[\pi_{11}(\bV_1)]=(1+\sigma^2_A) \bI_d$ and $\Cov[\pi_{11}(\bV_1), \pi_{12}(\bV_1)]= \bI_d$, so \Cref{lem:ridgeless:noniso:alt:rep} implies 
    \begin{align*}
        \bar \bZ_1 
        \;\overset{d}{=}\;
        \mfrac{1}{n} \msum_{i \leq n} 
        \eta_{i1} \eta_{i1}^\top
        +
        \mfrac{\sigma^2_A}{nk} \msum_{i \leq n, \, j \leq k} \eta_{ij} \eta_{ij}^\top
        \;,
        \tagaligneq \label{eq:dd:noise:inject:key:appendix}
    \end{align*}
which agrees with \eqref{eq:dd:noise:inject:key}. Meanwhile, in the no augmentation case where $\pi_{ij}={\rm id}$ almost surely, we have $\Var[\pi_{11}(\bV_1)]=\Cov[\pi_{11}(\bV_1), \pi_{12}(\bV_1)]=\Var[\bV_1]$, and \Cref{lem:ridgeless:noniso:alt:rep} implies  
\begin{align*}
    \bar \bZ_1 
        \;\overset{d}{=}\;
        \mfrac{1}{n} \msum_{i \leq n} (\Var[\bV_1]^{1/2} \, \eta_{i1})  (\Var[\bV_1]^{1/2} \, \eta_{i1} )^\top
        \tagaligneq \label{eq:Z1:unaugmented}
\end{align*}
as expected. 
\\[.5em]
(ii) \Cref{lem:ridgeless:noniso:alt:rep} expresses $\bar \bZ_1$ as a linear combination of two Wishart matrices of $n$ and $n(k-1)$ degrees of freedom respectively, whereas in \Cref{sec:triple:descent:oracle}, the degrees of freedom are $n$ and $nk$. (i) verifies that the expressions do agree in the isotropic noise injection case due to the special forms of $\Sigma_1$ and $\Sigma_2$, and \Cref{sec:triple:descent:oracle} confirms that $nk$ is the correct parameter to use for analyzing the peak of the augmented double-descent curve. In general, however, our analysis technique does not answer whether $nk$ or $n(k-1)$ should be used other than on a case-by-case basis; a more general and rigorous analysis involves computing the convolution of two Marchenko-Pastur laws, which we do not include in this paper.
\end{remark}

Notice that, similar to \Cref{sec:triple:descent:oracle}, $\Sigma_1$ determines the contribution of a Wishart matrix with $n$ degrees of freedom to $\bar \bZ_1$, whereas $\Sigma_2$ determines the contribution of a Wishart matrix with $n(k-1)$ degrees of freedom to $\bar \bZ_1$.  In the rest of the section, we compute the expression of $(\bar \bZ_1, \bar \bZ_2)$ in \Cref{lem:ridgeless:noniso:alt:rep} under the different augmentations in \Cref{assumption:network:augmentation} and discuss how it corresponds to empirical behaviors. In the calculations below, it is also useful to note that  by \Cref{lem:compare_variance}, $\Cov[\pi_{11}(\bV_1), \pi_{12}(\bV_1)] = \Var \, \mean[\pi_{11}(\bV_1) | \bV_1]$ and
$\Var[\pi_{11}(\bV_1)]-\Cov[\pi_{11}(\bV_1), \pi_{12}(\bV_1)] = \mean\,\Var[\pi_{11}(\bV_1) | \bV_1]$.

\vspace{.5em}

\begin{figure}[t]
  \centering
  \begin{tikzpicture}
      \node[inner sep=0pt] at (-4,0)
          {\includegraphics[width=.48\textwidth]{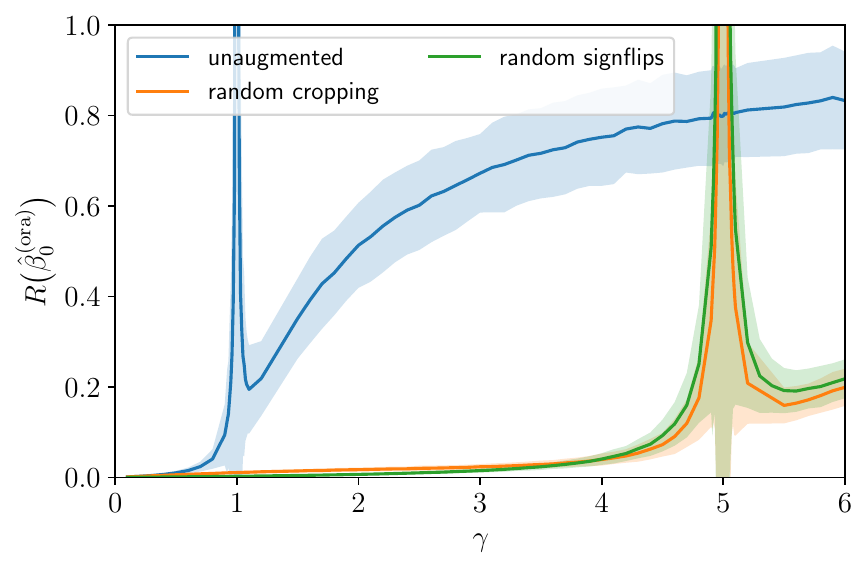}};
      \node[inner sep=0pt] at (3.5,0)
          {\includegraphics[width=.5\textwidth]{crop_signflip_effdim0.5.pdf}};
      \node[inner sep=0pt] at (-4,-5)
          {\includegraphics[width=.48\textwidth]{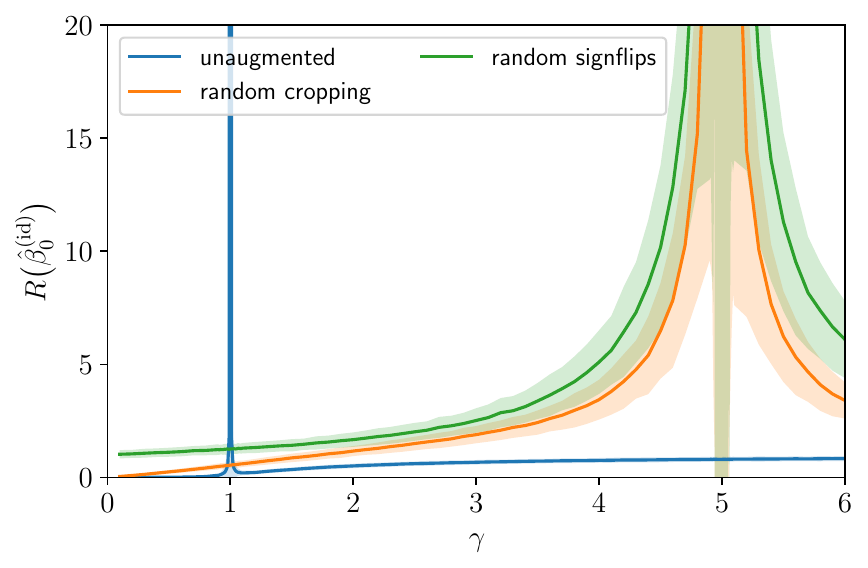}};
      \node[inner sep=0pt] at (3.5,-5)
          {\includegraphics[width=.5\textwidth]{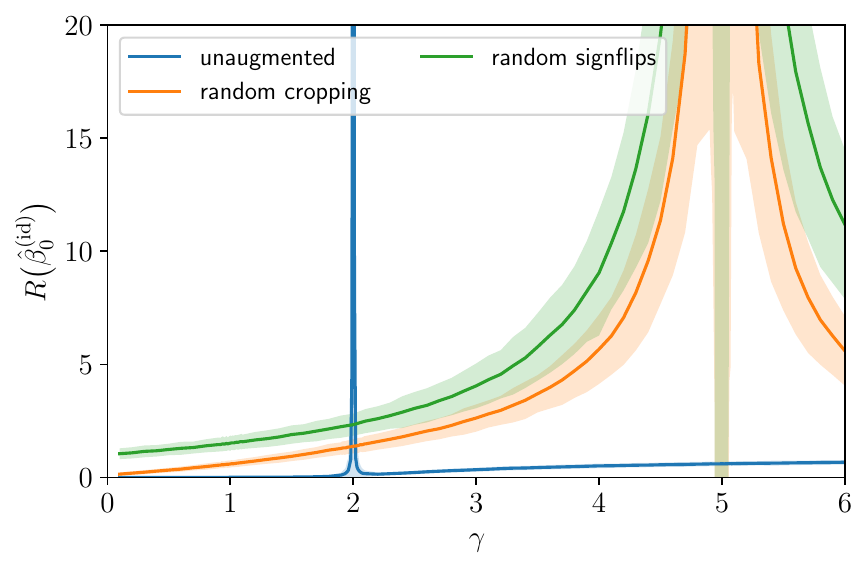}};
  \end{tikzpicture}
  \caption{Risks of  $\hat \beta_0^{(\rm ora)}$  and  $\hat \beta_0^{(\rm id)}$  under random cropping and random sign flipping. In all plots, we have fixed $n=200$, varying $d$, $\| \beta \|=1$, $\sigma_\epsilon=0.1$ and $k=5$.  \emph{Left column.} Data are generated as $\bV_i \overset{i.i.d.}{\sim} \cN(0, \bI_d)$. \emph{Right column.}  Data are generated as $\bV_i \overset{i.i.d.}{\sim} \cN(0, \bI_{d / 2} \otimes \bone_{2 \times 2} )$. The positions of the peak are unaffected by how $Y_i$ is augmented (which differs across rows), but may be affected by $\Var[\bV_1]$ (which differs across columns).}
  \label{fig:crop:flip} 
\end{figure}

\paragraph{Random cropping. } \label{para:random:cropping} By the law of total variance, we can compute
\begin{align*}
    \Var[\pi_{11}(\bV_1)] 
    \;=&\; 
    \Var[ (E_{111} V_{11}, \ldots, E_{11d} V_{1d} )^\top ]
    \\
    \;=&\;
    \Var \, \mean[ (E_{111} V_{11}, \ldots, E_{11d} V_{1d} )^\top | \bV_1 ]
    +
    \mean \, \Var[ (E_{111} V_{11}, \ldots, E_{11d} V_{1d} )^\top | \bV_1 ]
    \\
    \;=&\;
    \mfrac{1}{4} \Var[\bV_1]
    +
    \mean\Big[ \mfrac{1}{4} \diag\{ V_{11}^2, \ldots, V_{1d}^2 \} \Big]
    \\
    \;=&\;
    \mfrac{1}{4} \Var[\bV_1]
    +
    \mfrac{1}{4} \diag\{ \Var[V_{11}], \ldots, \Var[V_{1d}]\}
    \;,
    \\
    \Cov[\pi_{11}(\bV_1), & \,  \pi_{12}(\bV_1)] 
    \;=\;
    \Var\,\mean\big[ 
        (E_{111} V_{11}, \ldots, E_{11d} V_{1d} )^\top
        \,\big|\, \bV_1\big]
    \;=\;
    \mfrac{1}{4} \Var[\bV_1]\;.
\end{align*}
This implies that 
\begin{align*}
    \Sigma_1^{1/2} \;=&\;
    \mfrac{1}{2}\,
    \sqrt{ \diag\{ \Var[V_{11}], \ldots, \Var[V_{1d}]\} 
    +
    k \Var[V_1]
    }
    \;,
    \\ 
    \Sigma_2^{1/2} \;=&\; \mfrac{1}{2} 
    \diag\big\{ \sqrt{\Var[V_{11}]}, \ldots, \sqrt{\Var[V_{1d}]}\big\}
    \;.
\end{align*}
The presence of the diagonal term implies that, provided that every coordinate of $\bV_1$ has positive variance, both matrices remain full-ranked regardless of the structure of $\Var[\bV_1]$. In particular, the Wishart matrix $\frac{1}{nk} \sum_{i \leq n} \sum_{j \leq k} \eta_{ij} \eta_{ij}^\top$ with $nk$ degrees of freedom enters the expression of $\bar \bZ_1$ through a simple positive rescaling, just like how it enters $\bar \bZ_1$ in \eqref{eq:dd:noise:inject:key:appendix} for the isotropic noise injection case. Indeed in \Cref{fig:crop:flip}, we observe that 
for two different choices of $\Var[\bV_1]$, the double descent peak for the ridgeless risk curve under augmentation remains at $\gamma = d/n = k$, just as the isotropic noise injection case in \Cref{fig:noise:inject:oracle} in the main text.

\vspace{.5em}

On the other hand, for the unaugmented risk curve, since the rank of $\Var[\bV_1]$ is halved, the ``effective dimension'' is now $d/2$, as $\frac{1}{n}  \sum_{i \leq n} \eta_{i1} \eta_{i1}^\top$ only enters the expression of $\bar \bZ_1$ in \eqref{eq:Z1:unaugmented} through a $d/2$-dimensional subspace. \Cref{fig:crop:flip} verifies that the double descent peak shifts to the position $\gamma = d/n = 2$, i.e.~where $d/2 = n$. We also remark that in terms of the augmentation on the output $Y_i$, in \Cref{fig:crop:flip},  the choice between an oracle augmentation or the identity only affects the overall risk curve but not the positions of the peak. 

\vspace{.5em}

\paragraph{Random sign-flipping. } We can WLOG identify the Rademacher variables $R_{ijl} = 2  E_{ijl} -1$, where $E_{ijl}$'s are the Bernoulli variables defined in random cropping in \Cref{assumption:network:augmentation}. Therefore by recycling the calculations above, we get that 
\begin{align*}
    \Cov[\pi_{11}(\bV_1), \,  \pi_{12}(\bV_1)] 
    \;=&\;
    \Var\,\mean\big[ 
        (R_{111} V_{11}, \ldots, R_{11d} V_{1d} )^\top
        \,\big|\, \bV_1\big]
    \\
    \;=&\;
    \Var\,\mean\big[ 
        ( (2  E_{111} -1) V_{11}, \ldots,  (2  E_{11d} -1) V_{1d} )^\top
        \,\big|\, \bV_1\big]
    \;=\;
    0\;,
    \\
    \Var[\pi_{11}(\bV_1)]
    \;=&\; 
    \Var[ (R_{111} V_{11}, \ldots, R_{11d} V_{1d} )^\top ]
    \\
    \;=&\;
    \mean \Var\big[ ( (2  E_{111} -1) V_{11}, \ldots,  (2  E_{11d} -1) V_{1d} )^\top  \,\big|\, \bV_1 \big]
    \\
    \;=&\;
    \Var[V_1] + \diag\{ \Var[V_{11}], \ldots, \Var[V_{1d}]\}
    \;.
\end{align*}
This implies that 
\begin{align*}
    \Sigma_1^{1/2}
    \;=\; 
    \Sigma_2^{1/2}
    \;=\;
    \sqrt{ \diag\{ \Var[V_{11}], \ldots, \Var[V_{1d}]\} 
    +
    \Var[\bV_1]
    }
    \;.
\end{align*}
As with the random cropping case, provided that every coordinate of $\bV_1$ has positive variance, both matrices remain full-ranked regardless of the structure of $\Var[\bV_1]$. This is numerically confirmed in \Cref{fig:crop:flip} by the similar behaviors of the two augmented risk curves.

\vspace{.5em}

\begin{figure}[t]
  \centering
  \begin{tikzpicture}
      \node[inner sep=0pt] at (-4,0)
          {\includegraphics[width=.48\textwidth]{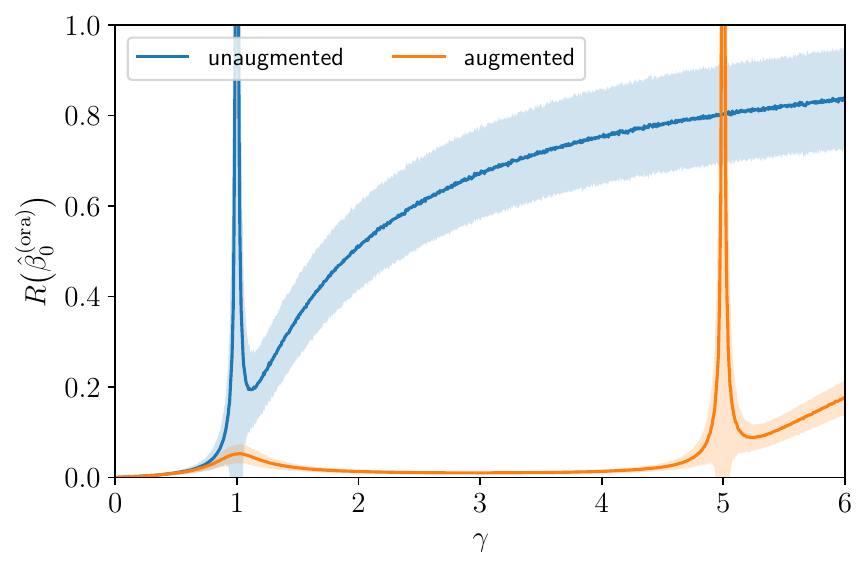}};
    \node[inner sep=0pt] at (-3.6,-2.5){ (i) isotropic noise injection to isotropic data};
      \node[inner sep=0pt] at (3.5,0)
          {\includegraphics[width=.5\textwidth]{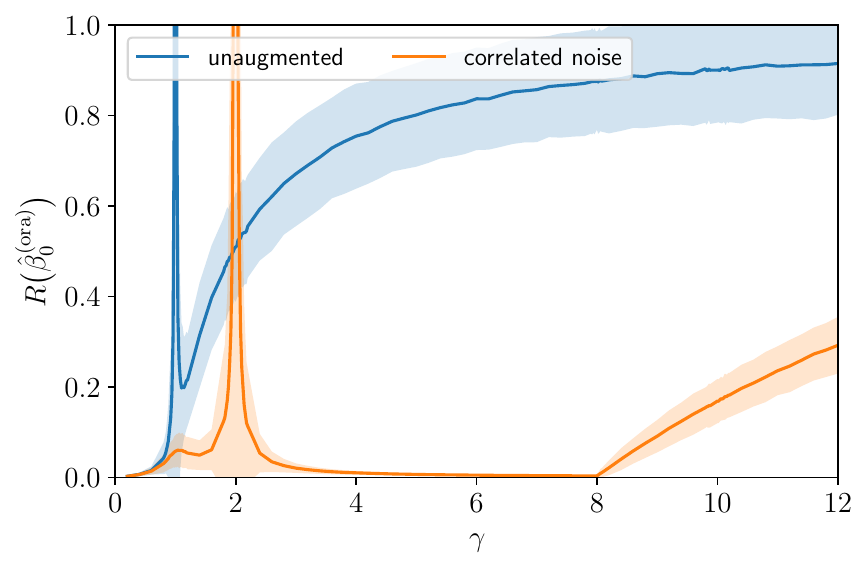}};
    \node[inner sep=0pt] at (3.85,-2.5){ (ii) correlated noise injection to isotropic data};
      \node[inner sep=0pt] at (-4,-5.5)
          {\includegraphics[width=.48\textwidth]{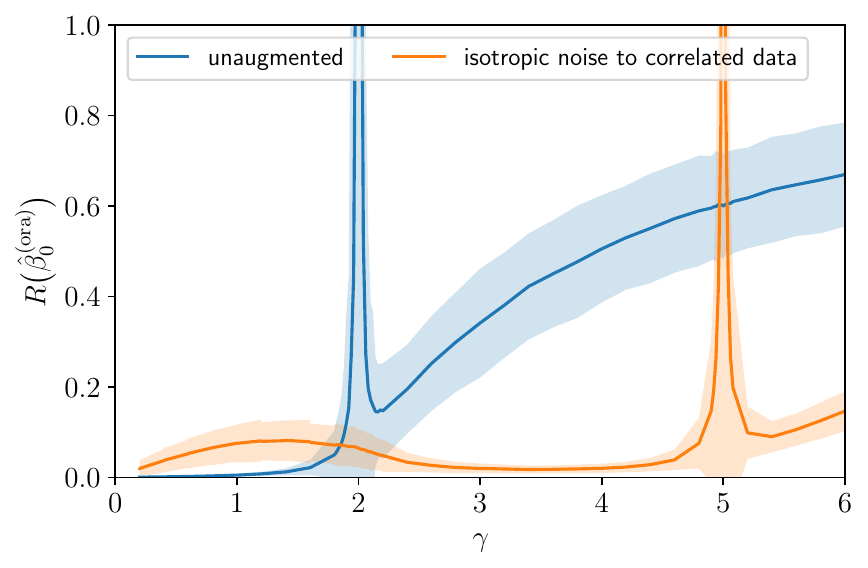}};
    \node[inner sep=0pt] at (-3.6,-8){ (iii) isotropic noise injection to correlated data};
      \node[inner sep=0pt] at (3.5,-5.5)
          {\includegraphics[width=.5\textwidth]{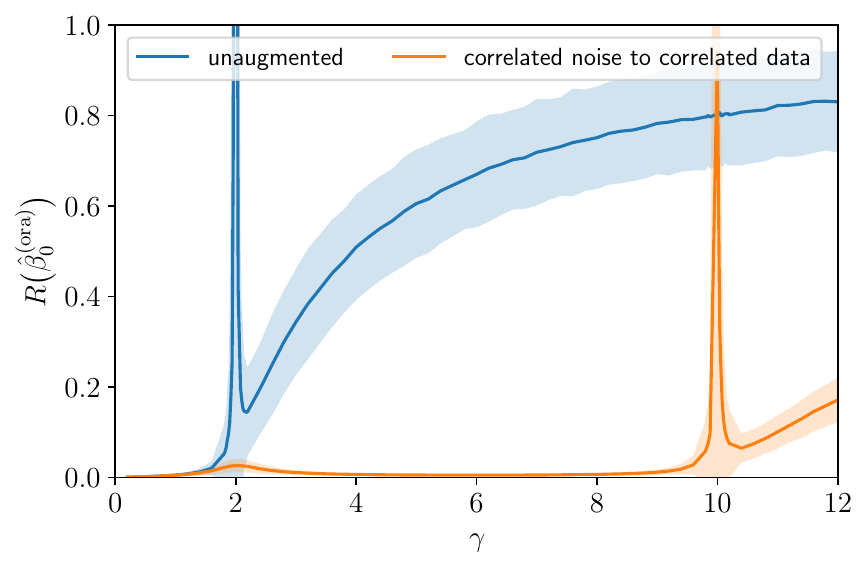}};
    \node[inner sep=0pt] at (3.85,-8){ (iv) correlated noise injection to correlated data};
  \end{tikzpicture}
  \caption{Risks of $\hat \beta^{(\rm ora)}_0$ under noise injection with varying correlation structure in both the data and the noise. (i) is identical to the left plot of \Cref{fig:noise:inject:oracle}, where $\Var[\bV_1] = \bI_d$ and $\Var[\xi_{11}] = 0.01 \bI_d$. (ii) has $\Var[\bV_1] = \bI_d$ and $\Var[\xi_{11}] = 0.01 \bI_{d/2} \otimes \bone_{2 \times 2}$. (iii) has  $\Var[\bV_1] = \bI_{d/2} \otimes \bone_{2 \times 2}$ and $\Var[\xi_{11}] = 0.01 \bI_d$. (iv) has  $\Var[\bV_1] = \bI_{d/2} \otimes \bone_{2 \times 2}$ and $\Var[\xi_{11}] = 0.01 \bI_d \otimes \bone_{2 \times 2}$, i.e.~the data and the noise have the same coordinate-correlation structure. The additional experiments in (ii), (iii) and (iv) were run with $n=100$, varying $d$, $\| \beta \| =1$ and $k=5$.
  }
  \label{fig:correlate:noise} 
\end{figure}

\paragraph{Correlated noise injection. } \label{para:correlated:noise:injection} We now consider injecting noise such that coordinates of the noise vector are allowed to be correlated. For simplicity, we suppose $d=b d'$ for some integers $b$ and $d'$, and consider i.i.d.~noise vectors $\xi_{ij} \sim \cN(0, \frac{\sigma^2_A}{b} I_{d'} \otimes \bone_{b \times b} )$. Then 
\begin{align*}
    &\;
    \Var[\pi_{11}(\bV_1)] 
    \;=\; 
    \Var[\bV_1] + \mfrac{\sigma^2_A}{b} I_{d'} \otimes \bone_{b \times b}
    \quad \text{ and } \quad 
    \Cov[\pi_{11}(\bV_1), \,  \pi_{12}(\bV_1)] 
    \;=\; 
    \Var[\bV_1]
    \;.
\end{align*}
Denote $P_{d'} \coloneqq I_{d'} \otimes \mfrac{1}{b} \bone_{b \times b}$ for simplicity, which is a projection matrix onto a $d'$-dimensional subspace, and write $P_{d'}^\perp = I_d - P_{d'}$. This implies that
\begin{align*}
    \Sigma_1^{1/2}
    \;=&\;
    \sqrt{\sigma^2_A P_{d'} + k   \Var[\bV_1] }
    &\text{ and }&&
    \Sigma_2^{1/2}
    \;=&\;
    \sigma_A \, P_{d'}
    \;.
\end{align*}
We shall use these formulas to analyze \Cref{fig:correlate:noise}, which present experiments that analyze (i) isotropic noise injection to isotropic data, (ii) correlated noise injection to isotropic data, (iii) isotropic noise injection to correlated data, and (iv) correlated noise injection to correlated data. To this end, let  $\cS_{d'}$ and $\cS_{d'}^\perp$ be orthogonal subspaces of $\R^d$ that correspond to $P_{d'}$ and $P_{d'}^\perp$ respectively. We consider two cases depending on the effect of $\Var[\bV_1]$ on the subspace $\cS_{d'}^\perp$:

\vspace{.5em}

\textbf{Case 1: $P_{d'}^\perp \Var[\bV_1] P_{d'}^\perp = 0$.} In this case, the subspace $\cS_{d'}^\perp$ is contained in the zero eigenspace of $\Var[\bV_1]$ and hence also in that of $\Sigma_1$. In other words, the matrix $\bar \bZ_1$ only has non-zero eigenvalues in the subspace $\cS_{d'}$. When restricted to the subspace $\cS_{d'}$, both Wishart matrices in the expression of $\bar \bZ_1$ in \Cref{lem:ridgeless:noniso:alt:rep} enter through a simple rescaling. Therefore the instability of $\bar \bZ_1^\dagger$ can be described by exactly the same argument as the isotropic noise injection case in \Cref{sec:triple:descent:oracle}, except that the dimension $d$ is replaced by the dimension of the smaller subspace $\cS_{d'}$: A regularization effect is expected at $d' = n$, whereas a peak is expected at $d' = nk$. 

\vspace{.5em}

This theoretical analysis is verified by \Cref{fig:correlate:noise}(i), (iii) and (iv): In both (i) and (iii), $d'=d$ and $P_{d'}^\perp = 0$, and a regularization effect is observed near $\gamma=1$ (i.e.~$d'\approx n$) whereas a peak is observed at $\gamma=k$ (i.e.~$d=nk$). In these two settings, the observation holds regardless of the structure of $\Var[\bV_1]$, which only shifted the peak of the unaugmented risk curve (in the same way as discussed in \ref{para:random:cropping} for random cropping). In (iv), $d'=d/2$ and $\Var[\bV_1]$ is chosen to satisfy $P_{d'}^\perp \Var[\bV_1] P_{d'}$. A regularization effect is observed at $\gamma=2$ (i.e.~$d'=d/2=n$), whereas a peak is observed at $\gamma=2k$ (i.e.~$d'=d/2=k$).

\textbf{Case 2: $P_{d'}^\perp \Var[\bV_1] P_{d'}^\perp \neq 0$.} This case includes \Cref{fig:correlate:noise}(ii). In this case, the subspace $\cS_{d'}^\perp$ is not contained in the zero eigenspace of $\Var[\bV_1]$. $P_{d'}^\perp \Sigma_1 P_{d'}^\perp$ is non-zero, whereas $P_{d'}^\perp \Sigma_2 P_{d'} = 0$. For any non-zero vector $v^\perp \in \cS_{d'}^\perp$, the Wishart matrix $\frac{1}{n} \sum_{i \leq n} P_{d'}^\perp \eta_{i1} \eta_{i1}^\top P_{d'}^\perp$ with $n$ degrees of freedom enters the expression for $(v^\perp)^\top \bar \bZ_1^\dagger v^\perp$, whereas the Wishart matrix $\frac{1}{nk} \sum_{i \leq n} \sum_{2 \leq j \leq k} \eta_{ij} \eta_{ij}^\top$ does not.

\vspace{.5em}

Compare this to the isotropic noise injection case: In \Cref{sec:triple:descent:oracle}, we have argued that at $d = n$ when the pseudoinverse  $(\frac{1}{n} \sum_{i \leq n} \eta_{i1} \eta_{i1}^\top)^\dagger$ is unstable, an additional regularisation is provided by the Wishart matrix with $nk$-degrees of freedom. This is no longer the case here, since the Wishart matrix with higher degrees of freedom does not play a role in the subspace $\cS_{d'}^\perp$. Therefore instead of a regularisation ``bump'', we now expect a peak at $n=d-d'$, where $d-d'$ is the dimensionality of the subspace $\cS_{d'}^\perp$. This observation is verified numerically in \Cref{fig:correlate:noise}(ii): There, $d'=d/2$, and a peak is observed at $\gamma = 2$, i.e.~$n=d-d'=d/2$.

\vspace{.5em}

\begin{figure}[t]
  \centering
  \begin{tikzpicture}
      \node[inner sep=0pt] at (0,0)
          {\includegraphics[width=.7\textwidth]{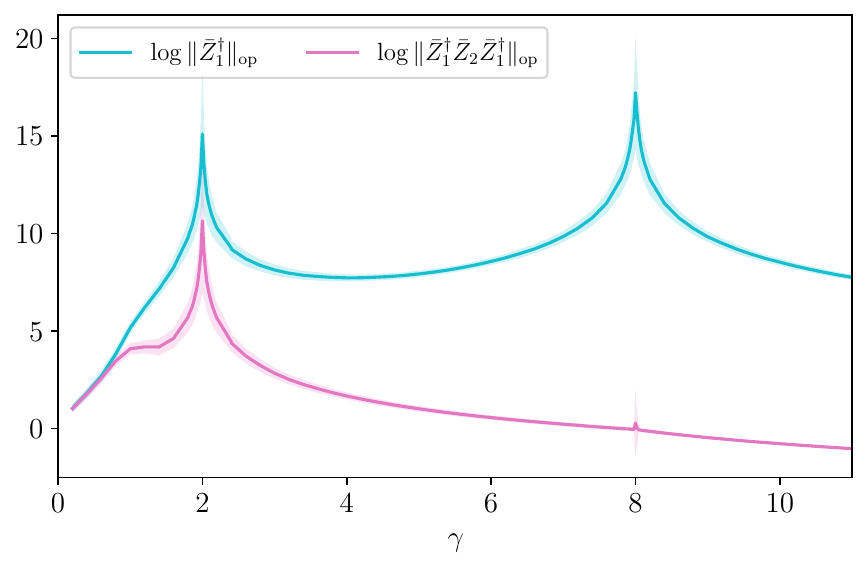}}; 
  \end{tikzpicture}
  \caption{The operator norms of $\bar \bZ_1^\dagger$ and $\bar \bZ_1^\dagger \bar \bZ_2 \bar \bZ_1^\dagger$ in the setup of \Cref{fig:correlate:noise}(ii) on the log scale. Instability of $\|\bar \bZ_1^\dagger\|_{op}$, as evidenced by the wide confidence band, is observed at both $\gamma=2$ (i.e.~$\frac{d}{2} = n$) and $\gamma=2(k-1)=8$ (i.e.~$\frac{d}{2} = n(k-1)$). In contrast, $\|\bar \bZ_1^\dagger \bar \bZ_2 \bar \bZ_1^\dagger\|_{op}$ remains stable at $\gamma=2(k-1)$.}
  \label{fig:correlate:noise:disappearing:peak:analysis} 
\end{figure}

In the subspace $\cS_{d'}$, $\Sigma_2$ is no longer negligible, and we expect $\bar \bZ_1^\dagger$ to be unstable when $d' = n(k-1)$. This is verified numerically in \Cref{fig:correlate:noise:disappearing:peak:analysis}, where we consider the case $d'=d/2$ and observe that $\| \bar \bZ_1^\dagger \|_{op}$ becomes unstable at both $\gamma = 2$ (i.e.~$d-d'=n$) and $\gamma = 2n(k-1)$ (i.e.~$d'=n(k-1)$). However, the corresponding risk curve (top right plot of \Cref{fig:correlate:noise}) does not show a peak at $\gamma = 2(k-1)$, despite a visible non-smooth change in the risk. To explain this, we recall from \eqref{eq:ridgeless:risk:var:term} that the instability of $\bar \bZ_1^\dagger$ enters the risk through the product of dependent matrices $\bar \bZ_1^\dagger \bar \bZ_2 \bar \bZ_1^\dagger$. This matrix product may be analyzed by the following closed-form expression from \Cref{lem:ridgeless:noniso:alt:rep}: For $v \in \cS_{d'}$, we have
\begin{align*}
    v^\top \bar \bZ_1^\dagger \bar \bZ_2 \bar \bZ_1^\dagger v
    \;\overset{d}{=}&\;
    v^\top
    \Big( 
        \mfrac{1}{nk} \msum_{i \leq n} \sqrt{\sigma^2_A + k} \, \eta_{i1} \eta_{i1}^\top M 
        +
        \mfrac{1}{nk} \msum_{i \leq n, 2 \leq j \leq k} \, \sigma^2_A \eta_{ij} \eta_{ij}^\top P_{d'}
    \Big)^\dagger
    \\
    &\;
    \Big( 
        \mfrac{1}{n} \msum_{i \leq n} 
        \Big( \mfrac{M}{k} \eta_{i1} + \mfrac{1}{k} \msum_{j=2}^k \sigma_A P_{d'} \eta_{ij} \Big)
        \Big( \mfrac{M}{k} \eta_{i1} + \mfrac{1}{k} \msum_{j=2}^k \sigma_A P_{d'} \eta_{ij} \Big)^\top
    \Big)
    \\
    &
    \Big( 
        \mfrac{1}{nk} \msum_{i \leq n} \sqrt{\sigma^2_A + k} \, M \eta_{i1} \eta_{i1}^\top 
        +
        \mfrac{1}{nk} \msum_{i \leq n, 2 \leq j \leq k} \, \sigma^2_A P_{d'} \eta_{ij} \eta_{ij}^\top 
    \Big)^\dagger
    v
    \;,
\end{align*}
where we have denoted $M \coloneqq \sqrt{\sigma^2_A + k} \, P_{d'} + \sqrt{k} \, P_{d'}^\perp$. However, since a direct analysis of this matrix product is cumbersome, we have chosen instead to examine it numerically: In the right plot of \Cref{fig:correlate:noise:disappearing:peak:analysis}, we verify numerically that  the product $\bar \bZ_1^\dagger \bar \bZ_2 \bar \bZ_1^\dagger$ remains stable at $\gamma=2n(k-1)$. We conjecture that this arises due to the interactions of $\bar \bZ_1$ and $\bar \bZ_2$ in the subspace $\cS_{d'}$.

\subsubsection{Random permutations. }

\begin{figure}[t]
  \centering
  \begin{tikzpicture}
        \node[inner sep=0pt] at (0,5)
          {\includegraphics[width=.5\textwidth]{permute_n100.pdf}};
      \node[inner sep=0pt] at (-4,0)
          {\includegraphics[width=.48\textwidth]{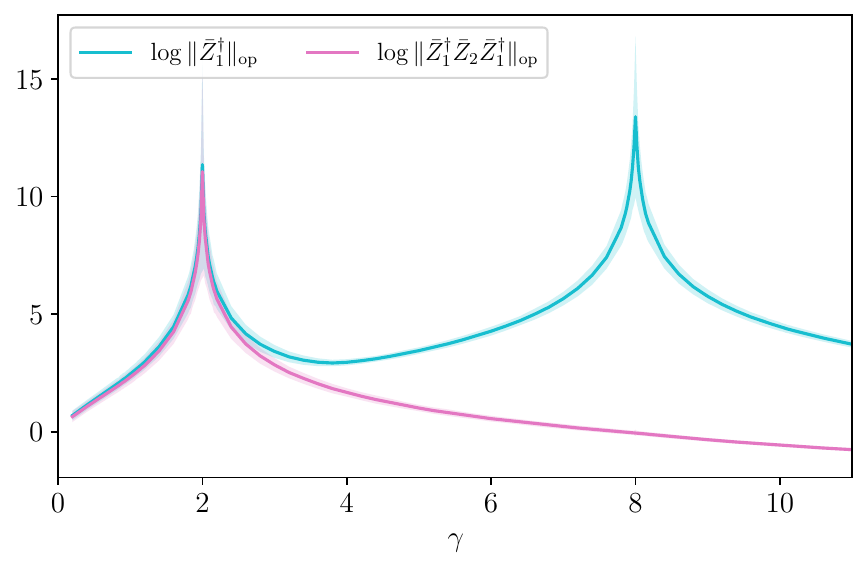}};
      \node[inner sep=0pt] at (-3.9,2.4)
          {\scriptsize  Size of partition $b=2$};
      \node[inner sep=0pt] at (3.5,0)
          {\includegraphics[width=.5\textwidth]{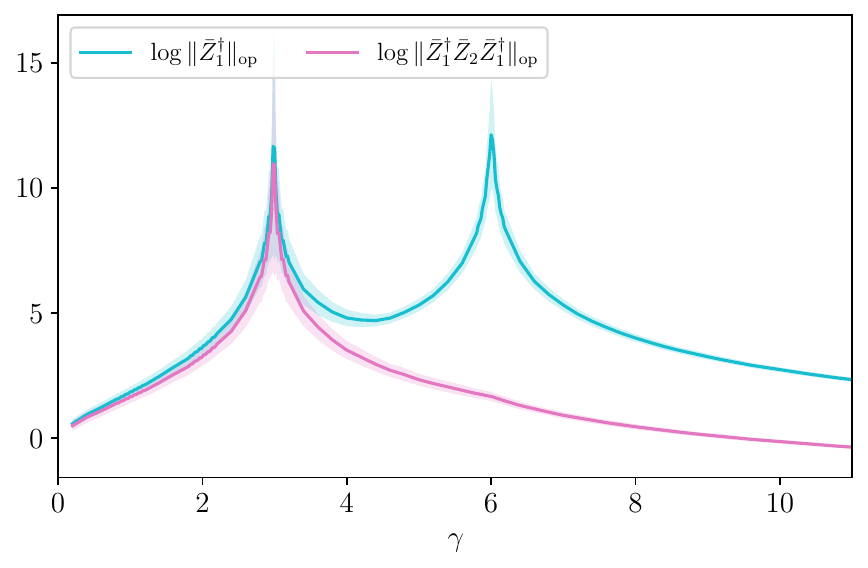}};
      \node[inner sep=0pt] at (3.6,2.45)
          {\scriptsize  Size of partition $b=3$};
  \end{tikzpicture}
  \caption{Risks of $\hat \beta^{(\rm ora)}_0$ and log-operator norms of $\bar \bZ_1^\dagger$ and  $\bar \bZ_1^\dagger \bar \bZ_2 \bar \bZ_1^\dagger$ under random permutations. We have chosen $\beta=\frac{1}{\sqrt{d}}\bone_d$ in this setup, so the oracle estimator $\hat \beta^{(\rm ora)}_0$ equals $\hat \beta^{(\rm id)}_0$. The simulations are performed with $n=100$, varying $d$, varying $b$ (size of partition) and $k=5$. The behaviors mirror that of \Cref{fig:correlate:noise}(ii) and \Cref{fig:correlate:noise:disappearing:peak:analysis}, where $\| \bar \bZ_1^\dagger \|$ becomes unstable at $\gamma=b$ and $\gamma = \frac{b}{b-1}(k-1)$ but only the first instability contributes to a peak in the risk.}
  \label{fig:permute} 
\end{figure}

For simplicity, suppose $d=b N_d$ for some integer $b$ and let the partitions be such that $P_l = \{ (l-1) b +1, \ldots, lb  \}$.  Then we may compute
\begin{align*}
    \Cov[\pi_{11}(\bV_1), \pi_{12}(\bV_1)]
    \;=&\;
    \Var \, \mean[ \pi_{11}(\bV_1) | \bV_1 ]
    \\
    \;=&\;
    \Var\Big[ 
            \Big(     
                \underbrace{\mfrac{1}{b} \msum_{r=1}^b 
                (V_{1r})
                ,\ldots,
                \mfrac{1}{b} \msum_{r=1}^b 
                (V_{1r})}_{\text{repeats $b$ times}}
                ,
                \;\ldots,\;
    \\
    &\;\hspace{4em}
                \underbrace{\mfrac{1}{b} \msum_{r=(N_d-1)b+1}^{N_d b}  
                (V_{1r})
                ,\ldots,
                \mfrac{1}{b} \msum_{r=(N_d-1)b+1}^{N_d b}
                (V_{1r})}_{\text{repeats $b$ times}}
            \Big)^\top
        \Big]
    \\
    \;=&\; 
    \Var\Big[ 
        \Big( 
            \mfrac{1}{b} \msum_{r=1}^b 
                (V_{1r})
                \,,\,\ldots\,,\,
            \mfrac{1}{b} \msum_{r=(N_d-1)b+1}^{N_d b}
                (V_{1r})
        \Big)^\top  
    \Big]
    \otimes 
    \bone_{b \times b}
    \;.
\end{align*}
Meanwhile, writing $\pi_{11}^{P_1}$ as the restriction of $\pi_{11}$ to the partition $P_1$ and $\bV_1^{P_1}$ as the vector of $b$ coordinates of $\bV_1$ restricted to $P_1$, we can compute 
\begin{align*}
    \mean \, \Var\big[ \pi_{11}^{P_1}(\bV_1^{P_1}) \,\big|\, \bV_1  \big] 
    \;=&\;
    \bI_b 
    \,
    \mean\Big[ 
        \mfrac{1}{b} \msum_{r=1}^{b} (V_{1r})^2 
        -
        \Big( \mfrac{1}{b} \msum_{r=1}^{b} V_{1r} \Big)^2
    \Big]
    \\
    &\;
    +
    (\bone_{b \times b} - \bI_b)
    \,
    \mean\Big[
        \mfrac{1}{b(b-1)} \msum_{r \neq s} V_{1r} V_{1s}
        -
        \Big( \mfrac{1}{b} \msum_{r=1}^{b} V_{1r} \Big)^2
    \Big]
    \\
    \;=&\;
    \mfrac{b-1}{b}
    \bI_b 
    \,
    \mean\Big[ 
        \mfrac{1}{b} \msum_{r=1}^{b} (V_{1r})^2 
    \Big]
    - \mfrac{1}{b}
    (\bone_{b \times b} - \bI_b)
    \,
    \mean\Big[ 
        \mfrac{1}{b} \msum_{r=1}^{b} (V_{1r})^2 
    \Big]
    \\
    \;=&\;
    \mean\Big[ 
        \mfrac{1}{b} \msum_{r=1}^{b} (V_{1r})^2 
    \Big]
    \Big( \bI_b - \mfrac{1}{b} \bone_{b \times b} \Big)
    \;,
\end{align*}
which implies
\begin{align*}
    \Sigma_2
    \;=&\;
    \Var[\pi_{11}(\bV_1)]  - \Cov[\pi_{11}(\bV_1), \pi_{12}(\bV_1)]
    \\
    \;=&\; 
    \mean \, \Var[ \pi_{11}(\bV_1) | \bV_1 ]
    \\
    \;=&\;
    \diag\Big\{  
        \mean \, \Var\big[ \pi_{11}^{P_1}(V_1^{P_1}) \,\big|\, V_1  \big] 
        \,,\,\ldots\,,\,
        \mean \, \Var\big[ \pi_{11}^{P_{N_d}}(V_1^{P_{N_d}}) \,\big|\, V_1  \big] 
    \Big\}
    \\
    \;=&\;
    \diag\Big\{  
        \mean\Big[ 
            \mfrac{1}{b} \msum_{r=1}^{b} (V_{1r})^2 
        \Big]
        \,,\,\ldots\,,\,
        \mean\Big[ 
            \mfrac{1}{b} \msum_{r=(N_d-1)b+1}^{N_d b} (V_{1r})^2 
        \Big]
    \Big\}
    \otimes 
    \Big( \bI_b - \mfrac{1}{b} \bone_{b \times b} \Big)
    \;,
\end{align*}
and 
\begin{align*}
    \Sigma_1 
    \;=&\;
    \Sigma_2 + k \Cov[\pi_{11}(\bV_1), \pi_{12}(\bV_1)]
    \\
    \;=&\;
    \diag\Big\{  
        \mean\Big[ 
            \mfrac{1}{b} \msum_{r=1}^{b} (V_{1r})^2 
        \Big]
        \,,\,\ldots\,,\,
        \mean\Big[ 
            \mfrac{1}{b} \msum_{r=(N_d-1)b+1}^{N_d b} (V_{1r})^2 
        \Big]
    \Big\}
    \otimes 
    \Big( \bI_b - \mfrac{1}{b} \bone_{b \times b} \Big)
    \\
    &\;
    +
    \Var\Big[ 
        \Big( 
            \mfrac{1}{b} \msum_{r=1}^b 
                (V_{1r})
                \,,\,\ldots\,,\,
            \mfrac{1}{b} \msum_{r=(N_d-1)b+1}^{N_d b}
                (V_{1r})
        \Big)^\top  
    \Big]
    \otimes 
    \bone_{b \times b}
    \;.
\end{align*}
Notice the similarity with the computations for the correlated noise in \ref{para:correlated:noise:injection}: $\Sigma_2$ is restricted to a subspace $\cS_{d'}$ of dimension $d' \coloneqq d - N_d = d (1-b^{-1})$, whereas $\Sigma_1$ has signals in both $\cS_{d'}$ and its orthogonal complement $\cS_{d'}^\perp$. \Cref{fig:permute} verifies that for permutations, the peaks of the risk curve have similar behaviors as those for the correlated noise injection in \Cref{fig:correlate:noise}(ii): A peak is observed at $\gamma=b$ (i.e.~$d - d' = d b^{-1} = n$) due to the instability of a Wishart matrix with $n$ degrees of freedom in the subspace $\cS_{d'}^\perp$. Meanwhile, while the Wishart matrix with $n(k-1)$ degrees of freedom becomes unstable at $\gamma= \frac{b}{b-1}(k-1) $ (i.e.~$d' = d(1-b^{-1}) n(k-1)$), this does not contribute to another peak.

\subsection{Additional results on bagging in Section \ref{sec:bagging}}  \label{appendix:bagging} 

\subsubsection{Generic statistics of bagged estimators.}  \label{appendix:statistics:of:bagging}

\Cref{prop:bagging} is a result about the stability of the bagged estimator $f_m^{(B)}(\Phi \cX )$. In general, however, we may be interested in specific properties of the estimator $f_m^{(B)}(\Phi \cX )$, such as the test risk. In this section, we study the universality of the composite function $g\big( f_m^{(B)}(\Phi \cX ) \big)$, where $g: \R^q \rightarrow \R$ is some generic function of interest. \Cref{prop:bagging} will then be proved as a special case of our result here. Note that $g$ is set to have univariate output for simplicity, but the same argument can be easily extended to multivariate output with fixed dimensions.

\vspace{.5em}

\Cref{thm:main} says that a sufficient condition for the universality of $g\big( f_m^{(B)}(\Phi \cX ) \big)$ is for the composite function $g \circ f_m^{(B)}$ to be stable, in the sense that the following local derivatives from \eqref{eqn:defn:alpha_r} are sufficiently small:
\begin{align*} 
  \alpha_r^{(B)}
  \;\coloneqq\;
  \max_{i\leq n} \, \max\bigl\{
  &\,
  \bigl\|{\msup_{\bw\in[\bzero,\Phi_i\bX_i]}}
  \|D_i^r (g \circ f_m^{(B)})(\bW_i(\bw))\|\bigr\|_{L_{6}}
  \,,
  \\
  &\;
    \bigl\|{\msup_{\bw\in[\bzero,\bZ_i]}}
  \|D_i^r (g \circ f_m^{(B)})(\bW_i(\bw))\|\bigr\|_{L_{6}}
  \bigr\}\;.
  \tagaligneq \label{eqn:defn:alpha_r:bagged}
\end{align*}
We seek to control these in terms of the following local derivative terms of the base estimator $f^{(B)}_m$:
\begin{align*}
    \alpha_{1;t}^{(m)}
    \;\coloneqq&\;
     \max_{\substack{i \leq n, \, i' \leq m \\ \upsilon \in S([m]) }}
     \max\Big\{\,
    \Big\|
                {\sup_{\bw \in[\bzero,\Phi_i\bX_i]}}
                \Big\|
                \partial g\big( f_m^{(B)}(\bW_i(\bV_i)) \big)
                \,
                    D_{i'}  
                     f_m
                    \big( \bW^{\upsilon}_{i'}(\bw)  \big)
                \Big\|
    \,\Big\|_{L_{6+t}}
    \,,\,
    \\
    &\hspace{7em}
    \Big\|
                {\sup_{\bw \in[\bzero,\bZ_i]}}
                \Big\|
                \partial g\big( f_m^{(B)}(\bW_i(\bV_i)) \big)
                \,
                    D_{i'}  
                     f_m
                    \big( \bW^{\upsilon}_{i'}(\bw)  \big)
                \Big\|
    \,\Big\|_{L_{6+t}}
    \Big\}
    \;,
    \\
    \alpha_{2,1;t}^{(m)}
    \;\coloneqq&\;
     \max_{\substack{i \leq n,  i' \leq m \\ \upsilon \in S([m])}}
     \max\Big\{\,
    \Big\|
                {\sup_{\bw \in[\bzero,\Phi_i\bX_i]}}
                \Big\|
                \partial g\big( f_m^{(B)}(\bW_i(\bV_i)) \big)
                \,
                    D_{i'}^2  
                     f_m
                    \big( \bW^{\upsilon}_{i'}(\bw)  \big)
                \Big\|
    \,\Big\|_{L_{6+t}}
    \,,\,
    \\
    &\hspace{7em}
    \Big\|
                {\sup_{\bw \in[\bzero,\bZ_i]}}
                \Big\|
                \partial g\big( f_m^{(B)}(\bW_i(\bV_i)) \big)
                \,
                    D_{i'}^2  
                     f_m
                    \big( \bW^{\upsilon}_{i'}(\bw)  \big)
                \Big\|
    \,\Big\|_{L_{6+t}}
    \Big\}
    \,,\,
    \\
    \alpha_{2,2;t}^{(m)}
    \;\coloneqq&\;
     \max_{\substack{ 
            i \leq n \\
             \, i', i'' \leq m
             \\
             \upsilon \in S([m])
     }}
     \max\Big\{\,
    \Big\|
                {\sup_{\bw \in[\bzero,\Phi_i\bX_i]}}
                \Big\|
                \partial g^2\big( f_m^{(B)}(\bW_i(\bV_i)) \big)
                \Big(
                    D_{i'}
                     f_m
                    \big( \bW^{\upsilon}_{i'}(\bw)  \big)
                \\
                &\hspace{19em}
                    \otimes 
                    D_{i''}
                     f_m
                    \big( \bW^{\upsilon}_{i''}(\bw)  \big)
                \Big)
                \Big\|
    \,\Big\|_{L_{6+t}}
    \,,\,
    \\
    &\hspace{7em}
    \Big\|
                {\sup_{\bw \in[\bzero,\bZ_i]}}
                \Big\|
                \partial g^2\big( f_m^{(B)}(\bW_i(\bV_i)) \big)
                \Big(
                    D_{i'}
                     f_m
                    \big( \bW^{\upsilon}_{i'}(\bw)  \big)
                \\
                &\hspace{19em}
                    \otimes 
                    D_{i''}
                     f_m
                    \big( \bW^{\upsilon}_{i''}(\bw)  \big)
                \Big)
                \Big\|
    \,\Big\|_{L_{6+t}}
    \Big\}
    \;,
    \\
      \alpha_{3,1;t}^{(m)}
    \;\coloneqq&\;
     \max_{\substack{i \leq n, i' \leq m \\ \upsilon \in S([m])}
     }
     \max\Big\{\,
    \Big\|
                {\sup_{\bw \in[\bzero,\Phi_i\bX_i]}}
                \Big\|
                \partial g\big( f_m^{(B)}(\bW_i(\bV_i)) \big)
                \,
                    D_{i'}^3  
                     f_m
                    \big( \bW^{\upsilon}_{i'}(\bw)  \big)
                \Big\|
    \,\Big\|_{L_{6+t}}
    \,,\,
    \\
    &\hspace{7em}
    \Big\|
                {\sup_{\bw \in[\bzero,\bZ_i]}}
                \Big\|
                \partial g\big( f_m^{(B)}(\bW_i(\bV_i)) \big)
                \,
                    D_{i'}^3  
                     f_m
                    \big( \bW^{\upsilon}_{i'}(\bw)  \big)
                \Big\|
    \,\Big\|_{L_{6+t}}
    \Big\}
    \,,\,
    \\
    \alpha_{3,2;t}^{(m)}
    \;\coloneqq&\;
     \max_{\substack{ 
            i \leq n \\
             \, i', i''\leq m \\ \upsilon \in S([m])
     }}
     \max\Big\{\,
    \Big\|
                {\sup_{\bw \in[\bzero,\Phi_i\bX_i]}}
                \Big\|
                \partial g^2\big( f_m^{(B)}(\bW_i(\bV_i)) \big)
                \Big(
                    D_{i'}
                     f_m
                    \big( \bW^{\upsilon}_{i'}(\bw)  \big)
                \\
                &\hspace{19em}
                    \otimes 
                    D_{i''}^2
                     f_m
                    \big( \bW^{\upsilon}_{i''}(\bw)  \big)
                \Big)
                \Big\|
    \,\Big\|_{L_{6+t}}
    \,,\,
    \\
    &\hspace{7em}
    \Big\|
                {\sup_{\bw \in[\bzero,\bZ_i]}}
                \Big\|
                \partial g^2\big( f_m^{(B)}(\bW_i(\bV_i)) \big)
                \Big(
                    D_{i'}
                     f_m
                    \big( \bW^{\upsilon}_{i'}(\bw)  \big)
                \\
                &\hspace{19em}
                    \otimes 
                    D_{i''}^2
                     f_m
                    \big( \bW^{\upsilon}_{i''}(\bw)  \big)
                \Big)
                \Big\|
    \,\Big\|_{L_{6+t}}
    \Big\} 
    \,,\,
    \\
    \alpha_{3,3;t}^{(m)}
    \;\coloneqq&\;
     \max_{\substack{ 
            i \leq n \\
             \, i', i'', i''' \leq m \\ \upsilon \in S([m])
     }}
     \max\Big\{\,
    \Big\|
                {\sup_{\bw \in[\bzero,\Phi_i\bX_i]}}
                \Big\|
                \partial g^3\big( f_m^{(B)}(\bW_i(\bV_i)) \big)
                \Big(
                    D_{i'}
                     f_m
                    \big( \bW^{\upsilon}_{i'}(\bw)  \big)
                \\
                &\hspace{19em}
                    \otimes 
                    D_{i''}
                     f_m
                    \big( \bW^{\upsilon}_{i''}(\bw)  \big)
                \\
                &\hspace{19em}
                    \otimes 
                    D_{i'''}
                     f_m
                    \big( \bW^{\upsilon}_{i'''}(\bw)  \big)
                \Big)
                \Big\|
    \,\Big\|_{L_{6+t}}
    \,,\,
    \\
    &\hspace{7em}
    \Big\|
                {\sup_{\bw \in[\bzero,\bZ_i]}}
                \Big\|
                \partial g^3\big( f_m^{(B)}(\bW_i(\bV_i)) \big)
                \Big(
                    D_{i'}
                     f_m
                    \big( \bW^{\upsilon}_{i'}(\bw)  \big)
                \\
                &\hspace{19em}
                    \otimes 
                    D_{i''}
                     f_m
                    \big( \bW^{\upsilon}_{i''}(\bw)  \big)
                \\
                &\hspace{19em}
                    \otimes 
                    D_{i'''}
                     f_m
                    \big( \bW^{\upsilon}_{i'''}(\bw)  \big)
                \Big)
                \Big\|
    \,\Big\|_{L_{6+t}}
    \Big\}
    \;,
\end{align*}
where we have denoted $S([m])$ as the set of all permutations on the index set $\{1, \ldots, m\}$ and $\bW_{i'}^{\upsilon}(\bw) \coloneqq (\Phi_{\upsilon(1)}\bX_{\upsilon(1)},\ldots, \Phi_{\upsilon(i'-1)}\bX_{\upsilon(i'-1)},\bw,\bZ_{\upsilon(i'+1)}, \ldots,\bZ_{\upsilon(m)} )$, where $\upsilon$ permutes the $m$ arguments.

\begin{proposition} \label{prop:bagging:general} Let $(\bX_i)_{i \leq n}$ and $\phi_{ij}$ be defined as in \Cref{thm:main}. Suppose $m=o(\sqrt{n})$ and $B=\Omega(n^{1 - t/(108+18t)} )$ for some fixed $t > 0$. Then 
\begin{align*}
    \alpha_1^{(B)} 
    \;=&\;
    o\Big( \mfrac{\alpha^{(m)}_{1;t}}{\sqrt{n}}
    \Big)\;,
    \quad
    \alpha_2^{(B)} 
    \;=\;
    o\Big( \mfrac{ \alpha^{(m)}_{2,1;t}}{\sqrt{n}}
        +
        \mfrac{ \alpha^{(m)}_{2,2;t}}{n}
    \Big)\;,
    \quad
    \alpha_3^{(B)} 
    \;=\;
    o\bigg( 
        \mfrac{ \alpha^{(m)}_{3,1;t}}{\sqrt{n}}
        +
        \mfrac{ \alpha^{(m)}_{3,2;t}}{n}
        +
        \mfrac{ \alpha^{(m)}_{3,3;t}}{n^{3/2}}
    \bigg)\;.
\end{align*}
\end{proposition}

 \Cref{prop:bagging} can then be obtained as a special case of \Cref{prop:bagging:general}. By slightly adapting the proof of \Cref{prop:bagging:general}, we can also obtain an analogous result for a bagged estimator that has quadratic dependence on $\upsilon_b$'s, which is handy for the application in \Cref{appendix:bagged:network}. Fix $q=1$ for simplicity again and write 
\begin{align*}
    f^{\rm quad}(\Phi \cX ) 
    \;\coloneqq\; 
    \mfrac{1}{B^2} \msum_{b, b' \leq B} 
    f^{\rm quad}_m
    \big( 
        &\, \Phi_{\upsilon_b(1)} \bX_{\upsilon_b(1)}, \ldots, \Phi_{\upsilon_b(m)} \bX_{\upsilon_b(m)},
    \\
        &\, \Phi_{\upsilon_{b'}(1)} \bX_{\upsilon_{b'}(1)}, \ldots, \Phi_{\upsilon_{b'}(m)} \bX_{\upsilon_{b'}(m)}, 
    \big)
    \;,
\end{align*}
where the base estimator is given by a thrice-differentiable function $f^{\rm quad}_m: \cD^{2mk} \rightarrow \R$. The next lemma gives a universality bound on $f^{\rm quad}(\Phi \cX )$ in terms of the version of \Cref{thm:main:general} discussed in \Cref{remark:non:iid:no:taylor} and in terms of the following derivative term of $f^{\rm quad}_m$:
\begin{align*}
    \alpha^{\rm quad}_{1;t} 
    \;\coloneqq\;
    \max_{\substack{i \leq n \\ \upsilon, \upsilon' \in S([m]) }} 
        \max \,\Big\{ & \,
        \|  \partial_{\Phi_i\bX_i}  f^{\rm quad}_m( \bW^{\upsilon,\upsilon'}_i(\Theta \Phi_i\bX_i) ) (\Phi_i\bX_i) \|_{L_{3+t}}
        \,,\,
    \\
        &\,
        \|  \partial_{\bZ_i}  f^{\rm quad}_m( \bW^{\upsilon,\upsilon'}_i(\Theta \bZ_i) ) (\bZ_i) \|_{L_{3+t}}
        \Big\} \;.
\end{align*}

\begin{lemma} \label{lemma:bagging:quadratic:universality}  Let $(\bX_i)_{i \leq n}$, $\phi_{ij}$ and $\cZ \coloneqq (\bZ_i)_{i \leq n}$ be defined as in \Cref{thm:main}. Suppose $m=o(\sqrt{n})$ and $B=\Omega(n^{1 - t/(18+6t)} )$ for some fixed $t > 0$.  Then for any differentiable $h: \R \rightarrow \R$ with its  derivative uniformly bounded from above by $1$, we have 
\begin{align*}
    \big| \mean\big[ h(f^{\rm quad}(\Phi \cX )  ) \big] - \mean\big[ h( f^{\rm quad}(\cZ) )  \big] \big|
    \;=\; 
    o\big( \alpha^{\rm quad}_{1;t} \sqrt{n} \big)\;.
\end{align*}
\end{lemma}

\subsubsection{Augmented-and-bagged non-linear networks.}  \label{appendix:bagged:network} In this section, we apply the results on bagging to demonstrate that universality can be established under less stringent stability conditions. 

We first focus on the locally dependent nonlinear feature model in \Cref{appendix:locally:dependent:feature}, and show that we can improve upon the gradient condition on the feature maps $\varphi$ and $\varphi_{\theta_0}$ in \Cref{assumption:network:feature:gradient}. We inherit the notation from  \Cref{appendix:locally:dependent:feature}, and define the bagged version of $\hat \beta_\lambda$ as in \Cref{sec:bagging}:
\begin{align*}
    \hat \beta_\lambda^{\rm bagged} (\cX)
    \;\coloneqq&\; 
    \mfrac{1}{B} \msum_{b \leq B} \hat \beta_{\lambda;m}^{\upsilon_b}(\cX)
    \;,
    \\
    \hat \beta_{\lambda;m}^{\upsilon_b}(\cX)
    \;\coloneqq&\;
    \argmin_{\tilde \beta \in \R^p} 
    \mfrac{1}{mk} \msum_{i=1}^m  \msum_{j=1}^k
    \big( \tilde Y_{\upsilon_b(i)j} - \tilde \beta^\top \tilde \bV_{\upsilon_b(i)j}  \big)^2
    +
    \lambda \| \tilde \beta\|^2\;.
\end{align*}
As with \eqref{eqn:network:risk:generalize}, we study the risk 
\begin{align*}
    \hat L_\lambda^{\rm bagged} (\cX)
     \;\coloneqq\; 
     \mean\big[ 
   \big( 
     (\hat \beta^{\rm bagged}_{\lambda})^\top \varphi( \bV_{\rm new} ) 
     -
     Y_{\rm new}
   \big)^2  
   \,\big| \, \cX, \bW^{(0)} \big]
  \qquad\text{ for }\lambda\geq0\;.
\end{align*}
The risk formula now involve bagged matrices of the form 
\begin{align*}
    \bar \bX^{(b)}_1 
    \;\coloneqq&\; 
    \mfrac{1}{mk} \msum_{i=1}^m \msum_{j=1}^k  \tilde \bV_{\upsilon_b(i)j} (\tilde \bV_{\upsilon_b(i)j})^\top
    \;,
    \quad
    \bar \bX^{(b)}_3
    \;\coloneqq\; 
    \mfrac{1}{mk} \msum_{i=1}^m \msum_{j=1}^k \tilde \bV_{\upsilon_b(i)j} \tilde \bV_0^\top
    \;,
    \\
    \bar \bX^{(b,b')}_2 
    \;\coloneqq&\;
    \mfrac{1}{m^2} \msum_{i,i'=1}^m
    \ind_{\{ \upsilon_b(i) = \upsilon_{b'}(i') \}}
    \Big( \mfrac{1}{k} \msum_{j=1}^k  \tilde \bV_{\upsilon_b(i)j} \Big)
    \,
    \Big( \mfrac{1}{k} \msum_{j=1}^k  \tilde \bV_{\upsilon_{b'}(i')j} \Big)^\top 
    \;,
    \\
    \bar \bX^{(b);-1}_{1;\lambda}
    \;\coloneqq&\;
    \begin{cases}
        (    \bar \bX^{(b)}_1 + \lambda \bI_p )^{-1}
        & \text{ for } \lambda > 0 \;,
        \\
        (    \bar \bX^{(b)}_1 )^\dagger
        & \text{ for } \lambda = 0 \;.
    \end{cases}
\end{align*}
This requires a restatement of \Cref{assumption:network:ridgeless:by:ridge}:  

\begin{nonumassumption}[10$^{(B)}$] \label{assumption:bagging:ridgeless:by:ridge} The following quantities are $O(1)$ with probability $1 - o(1)$:
  \begin{align*}
      &\;
      \| (\bar \bX_1^{(1)})^\dagger \|_{op} 
      \;,\;
      \quad
      \| (\bar \bZ_1^{(1)})^\dagger \|_{op} 
      \;,\;
      \quad 
      \| \bar \bX_3^{(1)} \|_{op}
      \;,\;
      \quad 
      \| \bar \bZ_3^{(1)} \|_{op}
      \;,
      \\
      &\;
      \| \bar \bX_2^{(1,1)} \|_{op}
      \;,\;
      \quad 
      \| \bar \bZ_2^{(1,1)} \|_{op} 
      \;,\;
      \quad
      \| \bar \bX_2^{(1,2)} \|_{op}
      \;,\;
      \quad 
      \| \bar \bZ_2^{(1,2)} \|_{op} 
      \;,\;
      \\
      &
      \sum_{l=1}^d  \ind_{\{ \lambda_l(\bar \bX_1^{(1)}) = 0 \}} \big( v_l(\bar \bX_1^{(1)})^\top \bar \bX^{(1,2)}_2 \, v_l(\bar \bX_1^{(1)}) \big)
      \;,\;
      \sum_{l=1}^d  \ind_{\{ \lambda_l(\bar \bZ_1^{(1)}) = 0 \}} \big( v_l(\bar \bZ_1^{(1)})^\top \bar \bZ_2^{(1,2)} \, v_l(\bar \bZ_1^{(1)}) \big) \;.
  \end{align*}
  Moreover, the following quantities are $o(1)$ with probability $1-o(1)$:
  \begin{align*}
        &\;
        \Big\| 
        \msum_{l=1}^d  \ind_{\{ \lambda_l(\bar \bX_1^{(1)}) = 0 \}} \bar \bX^{(1)}_3 \, v_l(\bar \bX_1^{(1)}) v_l(\bar \bX_1^{(1)})^\top 
        \Big\|_{op}
      \;,\;
      \\
      &\; 
      \Big\| 
        \msum_{l=1}^d  \ind_{\{ \lambda_l(\bar \bZ_1^{(1)}) = 0 \}} \bar \bZ^{(1)}_3 \, v_l(\bar \bZ_1^{(1)}) v_l(\bar \bZ_1^{(1)})^\top 
        \Big\|_{op}
        \;.
  \end{align*}
\end{nonumassumption}

Under bagging, it suffices to have milder assumptions on the feature maps $\varphi$ and $\varphi_{\theta_0}$:

\begin{nonumassumption}[9$^{(B)}$] \label{assumption:bagging:feature:gradient}
Define $B_d$ as in \Cref{assumption:network:generalize}(iv). We assume the following:
\begin{align*}
    \gamma_1^\varphi \;=\; O\big( B_d^{-1/3} d^{1/3} \big) 
    \,,\, 
    \quad 
    \gamma_2^\varphi \;=\; O\big( B_d^{-2/3} d^{ 1/6} \big) 
    \,,\,
    \quad 
    \gamma_3^\varphi \;=\; O\big( B_d^{-1}  \big) \;.
\end{align*}
\end{nonumassumption}

The next result shows that the universality of the test risk for the augmented-and-bagged estimators holds under milder assumption on $\varphi$ and $\varphi_{\theta_0}$. We again recall that $\cH^{(4)} \subset \cH$ to denote the class of four-times continuously differentiable function with its first four derivatives uniformly bounded from above by $1$.

\begin{proposition}  \label{prop:bagged:network}  Fix $\lambda > 0$. Suppose $m=o(\sqrt{n})$ and $B=\Omega(n^{1 - t/(18+6t)} )$ for some fixed $t > 0$. Under Assumptions \ref{assumption:network:generalize}, \ref{assumption:network:bounded:cov} and \hyperref[assumption:bagging:feature:gradient]{9$^{(B)}$} and the asymptotic \eqref{eq:network:asymptotic:regime:generalise},
\begin{align*}
    d_{\tilde \cH^{(4)}} \big( \hat L_\lambda^{\rm bagged}(\cX)  \,,\, \hat L_\lambda^{\rm bagged}(\cX) \big)
    \;=\; 
    o\Big( 1 + 
                \mfrac{1}{\lambda} 
                +
                \mfrac{1}{\lambda^{6}} 
    \Big) 
    \;.
\end{align*}
If additionally Assumption \hyperref[assumption:bagging:ridgeless:by:ridge]{10$^{(B)}$} holds, then
\begin{align*}
    d_P\big( 
        \hat L_0^{\rm bagged}(\cX)
        \,,\,
        \hat L_0^{\rm bagged}(\cZ)
    \big)
    \;=\;
    o(1)
    \;.
\end{align*}
\end{proposition}

The relaxed derivative conditions on $\varphi$ and $\varphi_{\theta_0}$ allow us to, for example, establish universality for the augmented-and-bagged \emph{non-linear} pretrained neural networks:

\begin{assumption} (Bagged non-linear network setup) \label{assumption:bagging:network:nonlinear} Assume the conditions of \Cref{assumption:network:data,assumption:network:augmentation}, except for the following changes:
\begin{proplist}
    \item \textbf{Local dependency. } We require $B(\bV_1) = O(1)$ and, if noise injection in \Cref{assumption:network:augmentation}(i) is chosen, require the noise vectors $\xi_{ij}$ to satisfy $B(\xi_{ij}) = O(1)$;
    \item \textbf{Model. } For $l=1,\ldots, N_0-1$, let $\varphi^{(0)}_l: \R^{d^{(0)}_l} \rightarrow \R^{d^{(0)}_l}$ be some thrice-differentiable functions and suppose the true output is generated instead by 
    \begin{align*}
        Y_i 
        \;=\; 
        \beta^\top \bW^{(0)}_{N_0}  \varphi^{(0)}_{N_0-1} \big( \bW^{(0)}_{N_0-1} \ldots \varphi^{(0)}_1 \big(  \bW^{(0)}_1 \bV_i \big) \ldots \big) 
        + 
        \epsilon_i
        \;.
    \end{align*}
    \item \textbf{Estimator. } For $l=1,\ldots, N-1$, let $\varphi_l: \R^{d_l} \rightarrow \R^{d_l}$ be some thrice-differentiable functions. Instead of the fixed matrices $W_1, \ldots, W_N$ in \eqref{eq:network:ridge}, we now consider independent random matrices $(\bW_l)_{l \leq N}$ such that $N$ is fixed and each $\bW_l$ is $\R^{d_l \times d_{l-1}}$-valued random matrix with i.i.d.~$\cN(0,1/d_{l-1})$ entries. For $\lambda > 0$, we consider the estimator 
    \begin{align*}
        \tilde \beta_\lambda^{\rm bagged} 
        \;\coloneqq&\;
        \mfrac{1}{B} \msum_{b \leq B} \tilde \beta^{\upsilon_b}_{\lambda;m}
        \;,
        \\
        \tilde \beta_{\lambda;m}^{\upsilon_b}
        \;\coloneqq&\;
        \underset{\tilde \beta \in \R^p} {\argmin}
        \mfrac{1}{mk} \msum_{i=1}^m  \msum_{j=1}^k
        \\
        &\hspace{1em} 
        \big( \tau_{\upsilon(i)j}(Y_{\upsilon_b(i)})  \, -\, 
        \tilde \beta^\top
        \bW_N \varphi_{N-1}( \bW_{N-1} \ldots \varphi_1(  \bW_1  (\pi_{\upsilon_b(i)j}(\bV_{\upsilon_b(i)}) ) ) \ldots )
         \big)^2
         \\
        &\hspace{5em}
        +
        \lambda \| \tilde \beta\|^2\;,
    \end{align*}
    where $\upsilon_b$'s are i.i.d.~uniformly drawn from the set of all permutations on $[n]$. We also assume $\max_{l \leq N_0 -1} \| W_l \|_{op} = O(1)$, and 
     denote $\tilde \beta_0^{\rm bagged}  = \lim_{\lambda \rightarrow 0^+} \tilde \beta_\lambda^{\rm bagged}$ as usual.

    \item \textbf{Condition on activation maps. } We assume that 
    \begin{align*}
        \max_{l \leq N_0 -1, \, 1 \leq r \leq 3} \; \msup_{\bx \in \R^{d^{(0)}_l}} \| \partial^r \varphi^{(0)}_l(\bx) \|_{op}
        \;=&\;
        O(1)\;,
        \\
        \max_{l \leq N_0 -1, \, 1 \leq r \leq 3} \; \msup_{\bx \in \R^{d_l}} \| \partial^r \varphi_l(\bx) \|_{op}
        \;=&\;
        O(1)
        \;,
    \end{align*}
    and that the following vectors are mean-zero and $\sigma_V$-sub-Gaussian for some absolute constant $\sigma_V < \infty$:
    \begin{align*}
        &\;
        \tilde \bV_{1j}
        \coloneqq
         \bW_N \varphi_{N-1}( \bW_{N-1} \ldots \varphi_1(  \bW_1  (\pi_{1j}(\bV_1) ) ) \ldots )
         \;,
         \\
        &\; 
        \tilde \bV_{1j}^0 
        \coloneqq
        \begin{cases}
        \bW^{(0)}_{N_0}  \varphi^{(0)}_{N_0-1} \big( \bW^{(0)}_{N_0-1} \ldots \varphi^{(0)}_1 \big(  \bW^{(0)}_1 \pi_{1j}(\bV_1) \big) \ldots \big) 
        &
        \text{ if $\tau_{ij}$ is the oracle} 
        \;,
        \\
        \bW^{(0)}_{N_0}  \varphi^{(0)}_{N_0-1} \big( \bW^{(0)}_{N_0-1} \ldots \varphi^{(0)}_1 \big(  \bW^{(0)}_1 \bV_1 \big) \ldots \big) 
        &
        \text{ if $\tau_{ij}$ is identity a.s.} 
        \;,
        \end{cases}
        \\
        &\;
        \hspace{20em}
         \text{ for } 1 \leq j \leq k \;.
    \end{align*}
\end{proplist}    
\end{assumption}

As before, we denote the test risk corresponding to $\tilde \beta^{\rm bagged}_\lambda$ as $\tilde L^{\rm bagged}_\lambda$.

\begin{corollary} \label{cor:bagging:network:nonlinear}  Fix $\lambda > 0$. Suppose $m=o(\sqrt{n})$ and $B=\Omega(n^{1 - t/(18+6t)} )$ for some fixed $t > 0$. Under \Cref{assumption:bagging:network:nonlinear} and the asymptotic \eqref{eq:network:asymptotic:regime} with $N$ fixed,
\begin{align*}
    d_{\tilde \cH^{(4)}} \big( \tilde L_\lambda^{\rm bagged}(\cX)  \,,\, \tilde L_\lambda^{\rm bagged}(\cX) \big)
    \;=\; 
    o\Big( 1
                +
                \mfrac{1}{\lambda^{6}} 
    \Big) 
    \;.
\end{align*}
\end{corollary}

\begin{remark} Universality for the ridgeless case ($\lambda=0$) holds, if the analogue of Assumption \hyperref[assumption:bagging:ridgeless:by:ridge]{10$^{(B)}$} holds with the setup in \Cref{assumption:bagging:network:nonlinear}.
\end{remark}

We remark that the conditions on the activation maps, \Cref{assumption:bagging:network:nonlinear}(iv), are satisfied, for example, for the following setup:

\begin{lemma} \label{lem:tanh} Consider the setup in \Cref{assumption:bagging:network:nonlinear} except for (iv), and suppose we consider the augmentations (ii)--(iv) in \Cref{assumption:network:augmentation}. Assume that $\bV_1 \overset{d}{=} - \bV_1$. Also suppose that $\varphi_l$'s and $\varphi^{(0)}_l$'s are pointwise ${\rm tanh}$ functions, i.e.~for $x \in \R^{d_l}$ and $x^{(0)} \in \R^{d^{(0)}_l}$,
    \begin{align*}
         \varphi_l(x) \;=&\; (\tanh(x_l))_{l \leq d_l}
         &\text{ and }&&
         \varphi_l(x^{(0)}) \;=&\; (\tanh(x^{(0)}_l))_{l \leq d^{(0)}_l}\;.
    \end{align*}
Then  \Cref{assumption:bagging:network:nonlinear}(iv) holds.
\end{lemma}

%% file: appendix_3_aux_tools.tex
\section{Auxiliary results} \label{appendix:auxiliary}

In this section, we include a collection of results useful for various parts of our proof.

\vspace{-1em}

\subsection{Convergence in \texorpdfstring{$d_{\cH}$}{dH}}
 \label{appendix:conv:dH}
\vspace{.5em}

\subsubsection{The weak convergence lemma}

\vspace{.5em}

\Cref{lem:d_H} 
shows that convergence in $d_{\cH}$ implies weak convergence. The gist of the proof is as follows. Assuming dimension to be one, in Step 1, we construct a thrice-differentiable function in $\cH$ to approximate indicator functions in $\R$. This allows us to bound the difference in probabilities of two random variables $X$ and $Y$ lying in nearby regions by their distance in $d_{\cH}$. In Step 2, we consider a sequence of random variables $Y_n$ converging to $Y$ in $d_{\cH}$, and use Step 1 to bound the probability of $Y_n$ lying in a given region by the probability of $Y$ lying in a nearby region plus $d_{\cH}(Y_n, Y)$, which converges to zero. This allows us to show convergence of the distribution function of $Y_n$ to that of $Y$. Finally, we make use of Cramer-Wold and Slutsky's Lemma to generalize our result to $q \geq 1$ dimensions.
  
\begin{proof}[Proof of 
    Lemma \ref{lem:d_H}]
  {\emph{Step 1}.}
  Assume ${q=1}$. Let ${A\subset\mathbb{R}}$ be a Borel set, and ${\epsilon\in(0,1)}$
  a constant. We will first show that
  \begin{equation}\label{eqn:prob-lower-bound}
    \P(Y\in A_{8\epsilon}) \geq \P(X\in A) - {d_{\cH}(X,Y)}/{\epsilon^4}\;.
  \end{equation}
  where ${A_{\epsilon}\coloneqq \{x\in\mathbb{R} \mid \exists y\in A~{\rm
      s.t.}~|x-y|\le \epsilon\}}$. 
  To this end, define a smoothed approximation of the
  indicator function of $A$ as
  \begin{equation*}
    h_{\epsilon}(x)\coloneqq\frac{1}{\epsilon^4}\int_{x-\epsilon}^x\int_{s-\epsilon}^s\int_{t-\epsilon}^t\int_{y-\epsilon}^y
    \mathbb{I}\braces{z\in A_{4\epsilon}}\,dz\,dy\,dt\,ds\;.
  \end{equation*}
Then $h_{\epsilon}$ is three times differentiable everywhere on
$\mathbb{R}$, and its first
three derivatives are bounded in absolute value by $1/\epsilon^4$. It
follows that ${\epsilon^4 h_{\epsilon}\in\cH}$, and hence that 
\begin{equation*}
  | \mean{h_{\epsilon}(X)}-\mean{h_{\epsilon}(Y)} | \leq d_{\cH}(X, Y)/\epsilon^4\;.
\end{equation*}
Since $h_{\epsilon}=0$ outside $A_{8\epsilon}$ and $h_{\epsilon}=1$ on
$A$, we have  ${\P(Z\in A)\leq\mean[h_{\epsilon}(Z)]\leq\P(Z\in
  A_{8\epsilon})}$
for any random variable $Z$. It follows that
\begin{equation*}
    {\mean{h_{\epsilon}(X)}-\mean{h_{\epsilon}(Y)}}\geq \P(X\in A)-\P(Y\in
    A_{8\epsilon})\;,
\end{equation*}
which implies \eqref{eqn:prob-lower-bound}.
\\[.5em]
{\em{Step 2}.}
To establish weak convergence for ${q=1}$, denote by $F$ the c.d.f of
$Y$. To show ${Y_n\smash{\darrow} Y}$, it
suffices to show that ${\P(Y_n\le b)\rightarrow F(b)}$ at every point
${b\in\mathbb{R}}$ at which $F$ is continuous. For any $\epsilon\in
(0,1)$, we have
\begin{equation*}
  \P(Y\le b+8\epsilon)
  \;\ge\;
  \P(Y_n\le b)-{d_{\cH}(Y_n,Y)}/{\epsilon^4}
  \;\ge\;
  \limsup_n \P(Y_n\le b)\;,
\end{equation*}
where the first inequality uses \eqref{eqn:prob-lower-bound}
and the second ${d_{\mathcal{H}}(Y_n,Y)\rightarrow0}$.
Set ${a=b-8\epsilon}$. Then
\begin{equation*}
  \P(Y_n\le b)=\P(Y_n\le a+8\epsilon)\ge \P(Y\le
  a)-{d_{\cH}(Y_n,Y)}/{\epsilon^4}=\P(Y\le
  b-8\epsilon)-{d_{\cH}(Y_n,Y)}/{\epsilon^4}\;,
\end{equation*}
and hence ${\liminf_n \P(Y_n\le b)\ge \P(Y\le b-8\epsilon)}$. To
summarize, we have
\begin{equation*}
  \P(Y\le b-8\epsilon)
  \;\le\;
  \liminf_n \P(Y_n\le b)
  \;\leq\;
  \limsup_n \P(Y_n\le b)
  \;\leq\;
  \P(Y\le b+8\epsilon)\
\end{equation*}
for any ${\epsilon\in(0,1)}$. Since $F$ is continuous at $b$, we can
choose $\epsilon$ arbitrary small, which shows
$\lim \P(Y_n\le b)=\P(Y\le b)$. Thus, weak convergence holds in $\mathbb{R}$.\\[.5em]
{\em{Step 3}.}
Finally, consider any ${q\in\mathbb{N}}$.
In this case, it is helpful to write $\cH(q)$ for the class $\cH$
of functions with domain $\mathbb{R}^q$.
Recall the Cramer-Wold theorem
\citep[][Corollary 5.5]{Kallenberg:2001}: Weak convergence
${Y_n\smash{\darrow}Y}$ in $\mathbb{R}^q$ holds if, for every vector
${v\in\mathbb{R}^q}$, the scalar products ${v^\top Y_n}$ converge weakly to ${v^\top Y}$.
By Slutsky's lemma, it is sufficient to consider only vectors $v$ with $\|v\|=1$.
Now observe that, if 
${h\in\cH(1)}$ and $\|v\|=1$, the function ${y\mapsto h(v^\top y)}$ is in $\cH(q)$, for every ${v\in\mathbb{R}^q}$. 
It follows that ${d_{\cH(q)}(Y_n,Y)\rightarrow 0}$ implies
${d_{\cH(1)}(v^{\top}Y_n,v^{\top}Y)\rightarrow 0}$ for every
vector $v$, which by Step 2 implies
${v^{\top}Y_n\smash{\darrow}v^{\top}Y}$, and weak convergence in
$\mathbb{R}^q$ holds by Cramer-Wold.
\end{proof}

\vspace{-2em}

\subsubsection{Comparison of \texorpdfstring{$d_{\cH}$}{dH} with known probability metrics} \label{appendix:d_H:compare}
 
\vspace{.2em}

In this section, let $X, Y$ be random variables taking values in $\R$, and define $A^\epsilon$ as in the proof of 
Lemma \ref{lem:d_H}. 
We present a result that helps to build intuitions of $d_{\cH}$ by bounding it with known metrics. Specifically, we consider the Lévy–Prokhorov metric $d_P$ and Kantorovich metric $d_K$, defined respectively as 
\begin{align*}
    d_P(X,Y) \;&=\; \minf_{\epsilon > 0} \{ \epsilon \;|\; \P(X \in A) \leq \P( Y \in A_\epsilon) + \epsilon, 
    \\
    &\qquad\qquad\qquad
    \P(Y \in A) \leq \P( X \in A_\epsilon) + \epsilon \; \text{ for all Borel set } A \subseteq \R \}\;, \tagaligneq \label{eq:defn:prokhorov}
    \\
    d_K(X,Y) \;&=\; \msup \{ \mean[ h(X)] - \mean[ h(Y)] \;|\; h: \R \rightarrow \R \text{ has Lipschitz constant} \leq 1 \} \;.
\end{align*}
The Kantorovich metric is equivalent to the Wasserstein-1 metric when the distributions of $X$ and $Y$ have bounded support. We can compare $d_{\cH}$ to $d_P$ and $d_K$ as follows:

\begin{lemma} \label{lem:compare:d_H} $d_P(X,Y) \leq 8^{4/5} d_{\cH}(X,Y)^{1/5}$ and $d_{\cH}(X,Y) \leq d_K(X,Y)$.
\end{lemma}

\begin{proof} For the first inequality, recall from \eqref{eqn:prob-lower-bound} in the proof of 
    Lemma \ref{lem:d_H} 
that for $\delta > 0$ and any Borel set $A \subset \R$, $\P(Y\in A_{8\delta}) \geq \P(X\in A) - {d_{\cH}(X,Y)}/{\delta^4}$. Setting $\delta = \big( d_{\cH}(\bX, \bY) /8 \big)^{1/5}$ gives
\begin{align*}
    \P( X \in A) \;\leq\; \P \big( Y \in A_{ 8^{4/5} d_{\cH}(X,Y)^{1/5}  } \big) + 8^{4/5} d_{\cH}(X,Y)^{1/5}\;.
\end{align*}
By the definition of $d_P$, this implies that $d_P(X,Y) \leq  8^{4/5} d_{\cH}(X,Y)^{1/5}$. The second inequality $d_{\cH}(X,Y) \leq d_K(X,Y)$ directly follows from the fact that every $h \in \cH$ has its first derivative uniformly bounded above by $1$. 
\end{proof}

\begin{remark} The proof for $d_P(X,Y) \leq 8^{4/5} d_{\cH}(X,Y)^{1/5}$ in Lemma \ref{lem:compare_variance} can be generalized to $\R^q$ so long as $q$ is fixed. Since the inequality says convergence in $d_{\cH}$ implies convergence in $d_P$ and $d_P$ metrizes weak convergence, this gives an alternative proof for 
    Lemma \ref{lem:d_H}.
\end{remark}

\subsubsection{Convergence in $d_\cH$ implies convergence of mean}

\cref{lem:d_H:centering} 
is useful for translating the convergence in $d_{\cH}$ of uncentered quantities to centred versions, and we present the proof below.

\begin{proof}[Proof of 
    \cref{lem:d_H:centering}] 
    The first bound can be proved by noting that each coordinate function that maps an $\R^d$ vector to one of its coordinate in $\R$ belongs to $\cH$:
\begin{align*}
    \| \mean \bX - \mean \bY \| \;=\;  \big( \msum_{l=1}^q | \mean[ X_l] - \mean[Y_l] |^2 \big)^{1/2} \;\leq\; \big( q \, d_{\cH}(\bX, \bY)^2 \big)^{1/2} \;\leq\; q^{1/2} \epsilon \;.
\end{align*}
To prove the second bound, notice that the class of functions $\cH$ is invariant under a constant shift in the argument of the function, which implies $d_{\cH}(\bX - \mean \bX, \bY - \mean \bX) \leq \epsilon$. By a triangle inequality, we have 
    \begin{align*}
        d_{\cH}(\bX - \mean \bX, \bY - \mean \bY) 
        \;\leq&\; 
        \epsilon + d_{\cH}(\bY - \mean \bX, \bY - \mean \bY)
        \\
        \;\leq&\; 
        \epsilon + \msup_{h \in \cH} \big| \mean\big[ h(\bY - \mean \bX) - h(\bY - \mean \bY)\big] \big|
        \\
        \;\overset{(a)}{\leq}&\; 
        \epsilon + \| \mean \bX - \mean \bY \|
        \;\leq\;
        (1+q^{1/2})\epsilon 
        \;.
    \end{align*}
    In $(a)$, we have applied the mean value theorem to $h$ on the interval $[\bY-\mean \bX, \bY-\mean \bY]$ and used $\|\partial h \| \leq 1$. This finishes the proof.
\end{proof}

\subsection{Additional tools}

The following lemma establishes identities for comparing different variances obtained in 
\Cref{thm:main} 
(main result with augmentation), 
\eqref{eq:no_aug} 
(no augmentation) and other variants of the main theorem in Appendix \ref{appendix:remark:main}.

\begin{lemma}
  \label{lem:compare_variance} 
Consider independent random elements $\bphi, \bpsi$ of $\cT$ and $\bX$ of $\cD \subseteq \R^d$, where ${\bphi\equdist\bpsi}$. Then
\begin{proplist}
   \item $\Cov[ \bphi \bX, \bpsi \bX ]= \mean\Cov[\bphi \bX, \bpsi \bX | \bphi, \bpsi] = \Var \mean[\bphi \bX | \bX]$,
   \item $\Var[\bphi \bX]\succeq \mean\Var[\bphi \bX | \bphi] \succeq \Cov[\bphi \bX, \bpsi \bX]$, where $\succeq$ denotes L\"owner's partial order.
  \end{proplist} 
\end{lemma}

\begin{proof}
(i) By independence of $\bphi$ and $\bpsi$, $\Cov\big[\mean[\bphi \bX | \bphi], \mean[ \bpsi \bX | \bpsi] \big] = \bzero .$ By combining this with the law of total covariance, we obtain that
$$   \Cov[ \bphi \bX, \bpsi \bX ]=\mean[   \Cov[ \bphi \bX, \bpsi \bX |\bphi,\bpsi]]+\Cov\big[\mean[\bphi \bX | \bphi], \mean[ \bpsi \bX | \bpsi] \big] =\mean[   \Cov[ \bphi \bX, \bpsi \bX |\bphi,\bpsi]].$$ 
Moreover, independence of $\bphi$ and $\bpsi$ also gives $ \Cov[\bphi \bX, \bpsi \bX | \bX]=0$ almost surely. Therefore by law of total covariance with conditioning performed on $\bX$, we get
\begin{equation*}
    \Cov[ \bphi \bX, \bpsi \bX ] = \Cov\big[ \mean[\bphi \bX | \bX], \mean[\bpsi \bX | \bX] \big]\overset{(a)}{=} \Var\mean[\bphi \bX | \bX]\;,
\end{equation*} where to obtain $(a)$ we used the fact that as $\bphi\overset{d}{=}\bpsi$ we have $\mean[\bphi \bX | \bX]\overset{a.s}{=}\mean[\bpsi \bX | \bX]$.
\\[.3em]
(ii) By the law of total variance we have:
\begin{align}
    \Var[\bphi \bX] & = \mean[\Var[\bphi \bX | \bphi]]+ \Var[\mean[\bphi \bX | \bphi]].
\end{align} We know that $\Var[ \mean[\bphi \bX | \bphi]] \succeq 0$ almost surely, which implies that $ \Var[\bphi \bX] \succeq \mean[\Var[\bphi \bX | \bphi]]$. 
For the second inequality, note that for all deterministic vector $\bv \in \R^d$ we have
$$
\bv^\top \big( \mean\Var[\phi \bX | \phi ] - \mean\Cov[ \phi \bX, \psi \bX | \phi, \psi ] \big) \bv \overset{(b)}{=} \mean\big[\Var[ \bv^\top (\phi \bX) | \phi ] - \Cov[ \bv^\top (\phi \bX), \bv^\top (\psi \bX) | \phi, \psi ] \big]
$$ 
where $(b)$ is obtained by bilinearity of covariance. 
By Cauchy-Schwarz, almost surely,
$$ \Cov[ \bv^\top (\phi \bX), \bv^\top (\psi \bX) | \phi, \psi ]\le \sqrt{\Var[ \bv^\top (\phi \bX) | \phi ]}\sqrt{\Var[ \bv^\top (\psi \bX) | \psi ]}.$$ This implies that $$\bv^\top \big( \mean\Var[\phi \bX | \phi ] - \mean\Cov[ \phi \bX, \psi \bX | \phi, \psi ] \big) \bv \ge 0\;.$$
Therefore we conclude that
$$ \mean\Var[\bphi \bX | \bphi ] \succeq \mean\Cov[\bphi \bX, \bpsi \bX | \bphi, \bpsi] = \Cov[\bphi \bX, \bpsi \bX],$$
where the last inequality is given by (i). This gives the second inequality as desired.
\end{proof}

\vspace{.6em}

The function $\zeta_{i;m}$ defined in the following lemma enters the bound in 
\Cref{thm:main} 
and its variants through the noise stability terms $\alpha_r$ defined in 
\eqref{eqn:defn:alpha_r}
 and $\alpha_{r;m}$ defined in Theorem \ref{thm:main:general}, and will recur throughout the proofs for different examples. We collect useful properties of $\zeta_{i;m}$ into Lemma \ref{lem:zeta_properties} for convenience.

\begin{lemma}
  \label{lem:zeta_properties} For $1 \leq i \leq n$, let $\Phi_1 \bX_i$, $\bZ_i$ be random quantities in $\cD$. For a random function $\bT: \cD^k \rightarrow \R^+_0$, where $\R^+_0$ is the set of non-negative reals, and for $m \in \N$, define 
$$\zeta_{i;m}(\bT) 
\;\coloneqq\; 
\max \Big\{ 
\big\| \msup_{\bw \in [\bzero, \Phi_1 \bX_i]} \bT(\bw) \big\|_{L_m} 
, 
\big\| \msup_{\bw \in [\bzero, \bZ_i]} \bT(\bw) \big\|_{L_m}  
\Big\}\;.$$
Then for any deterministic $\alpha \in \R^+_0$, random functions $\bT_j: \cD^k \rightarrow \R^+_0$, and $s \in \N$, 
\begin{proplist}
\item \textbf{(triangle inequality)} $\zeta_{i;m}(\bT_1 + \bT_2) \;\leq\; \zeta_{i;m}(\bT_1) + \zeta_{i;m}(\bT_2)$,
\item \textbf{(positive homogeneity)} $\zeta_{i;m}(\alpha \bT_1) \;=\; \alpha \zeta_{i;m}(\bT_1)$, 
\item \textbf{(order preservation)} if for all $\bw \in \R^{dk}$, $\bT_1(\bw) \;\leq\; \bT_2(\bw)$ almost surely, then $\zeta_{i;m}(\bT_1) \;\leq\; \zeta_{i;m}(\bT_2)$, 
\item \textbf{(H\"older's inequality)} $\zeta_{i;m}( \prod_{j=1}^s \bT_j ) \;\leq\; \prod_{j=1}^s \zeta_{i;ms}(\bT_j)$, and
\item \textbf{(coordinate decomposition)} if $g: \cD^k \rightarrow \R^q$ is a $r$-times differentiable function and $g_s: \cD^k \rightarrow \R$ denotes the $s$-th coordinate of $g$, then $\zeta_{i;m}(\| \partial^r g(\argdot) \|) \leq \sum_{s \leq q} \zeta_{i;m}(\| \partial^r g_s(\argdot)\|)$. 
\end{proplist}
\end{lemma}

\begin{proof}
(i), (ii) and (iii) are straightforward by properties of $\sup$ and $\max$ and the triangle inequality. To prove (iv), we note that
$$\Big\|\sup_{\bw \in [\bzero, \Phi_1\bX_i]}  \mprod^s_{j=1} \bT(\bw) \Big\|_{L_m} \leq \Big\| \mprod^s_{j=1} \sup_{\bw \in [\bzero, \Phi_1\bX_i]} \bT_j(\bw) \Big\|_{L_m}. $$ By the generalized H\"older's inequality we also have
$$\Big\| \mprod^s_{j=1} \sup_{\bw \in [\bzero, \Phi_1\bX_i]} \bT_j(\bw) \Big\|_{L_m} \leq \mprod^s_{j=1} \Big\| \sup_{\bw \in [\bzero, \Phi_1\bX_i]} \bT_j(\bw) \Big\|_{L_{ms}}\le \mprod^s_{j=1} \zeta_{i;ms}(\bT_j). $$
Similarly we can prove that  $\big\| \sup_{\bw \in [\bzero, \bZ_i]}\mprod^s_{j=1}  \bT_j(\bw) \big\|_{L_m}\le \mprod^s_{j=1} \zeta_{i;ms}(\bT_j) $. This directly implies that $\zeta_{i;m}( \prod_{j=1}^s \bT_j ) \leq \prod_{j=1}^s \zeta_{i;ms}(\bT_j)$. Finally to show (v), note that
\begin{align*}
     \| \partial^r g( \bv ) \| \;=\; \sqrt{\msum_{s \leq q} \| \partial^r g_s(\bv)\|^2 } \; \leq \; \msum_{s \leq q} \| \partial^r g_s(\bv)\|
\end{align*}
for every $\bv \in \cD^k$. By (iii), this implies $\zeta_{i;m}(\| \partial^r g(\argdot) \|) \leq \sum_{s \leq q} \zeta_{i;m}(\| \partial^r g_s(\argdot)\|)$ as required.
\end{proof}

\vspace{.6em}

The following result from Rosenthal \cite{rosenthal1970} is useful for controlling moment terms, and is used throughout the proofs for different examples. We also prove a corollary that extends the result to vectors since we deal with data in $\cD \subseteq \R^d$.

\begin{lemma} (Theorem 3 of Rosenthal \cite{rosenthal1970}) \label{lem:rosenthal} Let $2\leq m<\infty$, and $\bX_1, \ldots, \bX_n$ be independent centred random variables in $\R$ admitting a finite $m$-th moment. Then there exists a constant $K_m$ depending only on $m$ such that
$$ 
\big\| \msum_{i=1}^n \bX_i \big\|_{L_m} 
\;\leq\; 
K_m \max \big\{ \big( \msum_{i=1}^n \| \bX_i \|_{L_m}^m \big)^{1/m}, \big( \msum_{i=1}^n \| \bX_i \|_{L_2}^2 \big)^{1/2} \big\}\;.
$$
\end{lemma}

\begin{corollary} \label{cor:rosenthal:vector}
Let $2\leq m<\infty$, and $\bX_1, \ldots, \bX_n$ be independent centred random vectors in $\R^d$ such that for all $i$, $\| \bX_i\|$ admits a finite $m$-th moment. Denote the $s$-th coordinate of $\bX_i$ by $X_{is}$. Then, there exists a constant $K_m$ depending only on $m$ such that
\begin{align*}
    \Big\| \big\| \msum_{i=1}^n \bX_i \big\| \Big\|_{L_m}
    \;\leq&\;  
    K_m \Big(  \msum_{s=1}^d \max \big\{ \big( \msum_{i=1}^n \| X_{is} \|_{L_m}^m \big)^{2/m}, \msum_{i=1}^n \| X_{is} \|_{L_2}^2 \big\} \Big)^{1/2}
    \;.
\end{align*}
\end{corollary}

\begin{proof}
     By triangle inequality followed by Lemma \ref{lem:rosenthal} applied to $\| \sum_{i=1}^n X_{is} \|_{L_m}$, there exists a constant $K_m$ depending only on $m$ such that
    \begin{align}
    \Big\| \big\| \msum_{i=1}^n \bX_i \big\| \Big\|_{L_m}
    \;&=\;
    \Big( \Big\|  \msum_{s=1}^d \big(\msum_{i=1}^n X_{is} \big)^2 \Big\|_{L_{m/2}} \Big)^{1/2}
    \label{eqn:vector:rosenthal:first:step} \\
    \;&\leq\;
    \Big(  \msum_{s=1}^d \big\| \big(\msum_{i=1}^n X_{is} \big)^2 \big\|_{L_{m/2}} \Big)^{1/2}
    \;=\;
    \Big(  \msum_{s=1}^d \big\| \msum_{i=1}^n X_{is} \big\|_{L_m}^2 \Big)^{1/2}
    \nonumber \\
    \;&\leq\; 
    K_m \Big(  \msum_{s=1}^d \max \big\{ \big( \msum_{i=1}^n \| X_{is} \|_{L_m}^m \big)^{2/m}, \msum_{i=1}^n \| X_{is} \|_{L_2}^2 \big\} \Big)^{1/2}\;.
    \nonumber 
    \end{align}
\end{proof}

\vspace{.2em}

The following lemma bounds the moments of vector norms of a Gaussian random vector in terms of its first two moments, which is useful throughout the proofs.

\begin{lemma} \label{lem:gaussian:moment}
    Consider a random vector $\bX$ in $\R^d$ with bounded mean and variance. Let $\xi$ be a Gaussian vector in $\R^d$ with its mean and variance matching those $\bX$, and write $\| \argdot \|_{\infty}$ as the vector-infinity norm. Then for every integer $m \in \N$,
    \vspace{-.2em}
    \begin{align*}
        \| \, \| \xi \|_{\infty} \, \|_{L_m} \;\leq\;  C_m \| \, \| \bX \|_{\infty} \, \|_{L_2} \sqrt{1+\log d}\;.
    \end{align*}
\end{lemma}

\begin{proof}
    Denote $\Sigma \coloneqq \Var[\bX]$, and write $\xi =  \mean[\bX] + \Sigma^{1/2} \bZ$ where $\bZ$ is a standard Gaussian vector in $\R^d$. First note that by triangle inequality and Jensen's inequality,
    \begin{align*}
        \| \, \| \xi \|_{\infty} \, \|_{L_m}
        \;\leq\; 
        \|  \mean[\bX] \|_{\infty} +\| \, \| \Sigma^{1/2} \bZ \|_{\infty} \, \|_{L_m}
        \;\leq&\;
        \|\, \| \bX\|_{\infty} \, \|_{L_1} + \| \, \| \Sigma^{1/2} \bZ \|_{\infty} \, \|_{L_m}
        \;.
    \end{align*}
    Write $\sigma_l \coloneqq \sqrt{\Sigma_{l,l}}$, the square root of the $(l,l)$-th coordinate of $\Sigma$. If $\sigma_l=0$ for some $l \leq d$, then the $l$-th coordinate of $\Sigma^{1/2}\bZ$ is zero almost surely and does not play a role in $| \Sigma^{1/2} \bZ \|_{\infty}$. We can then remove the $l$-th row and column of $\Sigma$ and consider a lower-dimensional Gaussian vector such that its covariance matrix has strictly positive diagonal entries. If all $\sigma_l$'s are zero, we get the following bound 
    \begin{align*}
        \| \, \| \xi \|_{\infty} \, \|_{L_m}
        \;\leq\; 
        \|\, \| \bX\|_{\infty} \, \|_{L_1}\;,
    \end{align*}
    which implies that $\| \, \| \xi \|_{\infty} \, \|_{L_m}$ satisfies the statement in the lemma. Therefore WLOG we consider the case where $\sigma_l > 0$ for every $l \leq d$. By splitting the integral and applying a union bound, we have that for any $c > 0$,
    \begin{align*}
        \| \, \| &\Sigma^{1/2} \bZ \|_\infty \, \|_{L_m}^m
        \;=\;
        \mean[ \mmax_{l \leq d} | (\Sigma^{1/2} \bZ)_l |^m ]
        \;=\;
        \mint_0^{\infty} \P\big( \mmax_{l \leq d} | (\Sigma^{1/2} \bZ)_l | > x^{1/m} \big) \, dx
        \\
        \;\leq&\;
        c
        +
        d
        \mint_c^{\infty}  \P\big( | (\Sigma^{1/2} \bZ)_l | > x^{1/m} \big) \, dx
        \;=\;
        c
        +
        d
        \mint_c^{\infty}  \P\big( \mfrac{1}{\sigma_l} | (\Sigma^{1/2} \bZ)_l | > \mfrac{1}{\sigma_l} x^{1/m} \big) \, dx
        \\
        \;\overset{(a)}{\leq}&\;
        c
        +
        d
        \mint_c^{\infty}  \mfrac{1}{\sqrt{2 \pi} \frac{1}{\sigma_l} x^{1/m}} \exp\big(- \mfrac{x^{2/m}}{2\sigma_l^2} \big) \, dx
        \\
        \;\leq&\;
        c + \mfrac{d \sigma_l}{\sqrt{2 \pi} c^{1/m}} \mint^{\infty}_c \exp\big(- \mfrac{x^{2/m}}{2\sigma_l^2} \big) \, dx\;. \tagaligneq \label{eqn:moment:gaussian:intermediate}
    \end{align*}
    In $(a)$ we have noted that $\frac{1}{\sigma_l} (\Sigma^{1/2} \bZ)_l \sim \cN(0,1)$, and used the standard lower bound for the c.d.f.\,of a standard normal random variable $Z$ to obtain
    \begin{align*}
        \P( | Z | > u ) \;=\; 2 \P(Z > u) \;\geq\; \mfrac{1}{\sqrt{2\pi} \, x} \exp\big( - \mfrac{x^2}{2} \big)\;.
    \end{align*}
    Choose $c=(2\sigma_l^2 (1+\log d))^{\frac{m}{2}}$. Then by a change of variable, the integral in \eqref{eqn:moment:gaussian:intermediate} becomes
    \begin{align*}
     \mint^{\infty}_c \exp\big(- \mfrac{x^{2/m}}{2\sigma_l^2} \big) \, dx
    \;=&\;
    (2\sigma_l)^{m/2} \mint^{\infty}_{1 + \log d} e^{-y} \, y^{\frac{m}{2}-1} dy
    \\
    \;\leq&\;
    (2\sigma_l)^{m/2} \mint^{\infty}_{1 + \log d} y^{\lfloor \frac{m}{2} \rfloor} e^{-y} \,  dy
    \;=:\; (2\sigma_l)^{m/2} I_{\lfloor \frac{m}{2} \rfloor} 
    \;.
    \end{align*}
    We have denoted $I_k \coloneqq \int^{\infty}_{1 + \log d} y^k e^{-y} dy$. By integration by parts, we get the following recurrence for $k \geq 1$,
    \begin{align*}
        I_k
        =& 
        (1+\log d)^k e^{-1-\log d} + k \, I_{k-1}
        =
        (1+\log d)^k (ed)^{-1} + k \, I_{k-1}\;,
    \end{align*}
    and also $I_0 = (ed)^{-1}$. This implies that there exists some constant $A_m$ depending only on $m$ such that
    \begin{align*}
    \mint^{\infty}_c &\exp\big(- \mfrac{x^{2/m}}{2\sigma_l^2} \big) \, dx
    \;\leq\;
    (2\sigma_l^2)^{m/2}  I_{\lfloor \frac{m}{2} \rfloor} 
    \\
    \;\leq&\;
    (2\sigma_l^2)^{m/2} (ed)^{-1} \lfloor \mfrac{m}{2} \rfloor \,!
    +
     (2\sigma_l^2)^{m/2}  \msum_{k=1}^{\lfloor \frac{m}{2} \rfloor} (ed)^{-(\lfloor \frac{m}{2} \rfloor+1-k)} \mfrac{\lfloor \frac{m}{2} \rfloor!}{k\,!} (1+\log d)^k
    \\
    \;\leq&\;
    A_m d^{-1} \sigma_l^m (1+\log d)^{\lfloor \frac{m}{2} \rfloor}
    \;.
    \end{align*}
    Substituting this and our choice of $c$ into \eqref{eqn:moment:gaussian:intermediate}, while noting that $\sigma_l = \sqrt{\Sigma_{l,l}} \leq \| \Sigma \|_{\infty}^{1/2}$, we get that
    \begin{align*}
        \| \, \| \Sigma^{1/2} \bZ \|_\infty \, \|_{L_m}^m
        \;&\leq\;
        (2\sigma_l^2 (1+\log d))^{\frac{m}{2}}
        +
        \mfrac{d\sigma_l}{\sqrt{2\pi}(2\sigma_l^2 (1+\log d))^{\frac{1}{2}} }
        A_m d^{-1} \sigma_l^m (1+\log d)^{\lfloor \frac{m}{2} \rfloor}
        \\
        \;&\leq\; B_m (\| \Sigma \|_{\infty}(1+ \log d))^{m/2}\;,
    \end{align*}
    for some constant $B_m$ depending only on $m$.
    Finally, by the property of a covariance matrix and Jensen's inequality, we get that
    \begin{align*}
        \| \Sigma \|_{\infty}^{1/2}
        \;\leq\; 
        \mmax_{l \leq d} \Var[X_l]^{1/2}
        \;\leq\; 
        \mmax_{l \leq d} \mean[X_l^2]^{1/2} 
        \;\leq\; 
        \| \mean[\bX \bX^\top] \|_{\infty}^{1/2}
        \;\leq\;
        \| \, \| \bX \|_{\infty} \, \|_{L_2}\;.
    \end{align*}
    These two bounds on $\| \, \| \Sigma^{1/2} \bZ \|_\infty \, \|_{L_m}^m$ and $ \| \Sigma \|_{\infty}^{1/2}$ imply that, for $C_m\coloneqq B_m+1$,
    \begin{align*}
        \| \, \| \xi \|_{\infty} \, \|_{L_m}
        \;\leq&\;
        \|\, \| \bX\|_{\infty} \, \|_{L_1} + \| \, \| \Sigma^{1/2} \bZ \|_{\infty} \, \|_{L_m}
        \\
        \;\leq&\;
        \|\, \| \bX\|_{\infty} \, \|_{L_1} + B_m \| \, \| \bX \|_{\infty} \, \|_{L_2} (1+\log d)^{1/2}
        \\
        \;\leq&\; C_m \| \, \| \bX \|_{\infty} \, \|_{L_2} (1+\log d)^{1/2}\;.
    \end{align*}
\end{proof}

\vspace{.5em}

The next result controls the norm of the largest eigenvalue of a sum of i.i.d.~zero-mean (not necessarily symmetric) matrices.

\begin{lemma} \label{lem:matrix:op:bound} Let $(\bA_i)_{i \leq n}$ be i.i.d.~zero-mean random matrices in $\R^{d \times d}$ and $m \geq 1$. There exists some absolute constant $C > 0$ such that 
    \begin{align*}
        \Big\| \, \Big\| \mfrac{1}{n} & \msum_{i=1}^n \bA_i \Big\|_{op}  \, \Big\|_{L_m}
        \\
        &\;\leq\; 
        \mfrac{C \sqrt{m + \log d \,}}{\sqrt{n}} 
        \;
        \Big(
            \Big\| \,\Big\| \mfrac{1}{n} \msum_{i=1}^n \bA_i \bA_i^\top \Big\|_{op}^{1/2}\, \Big\|_{L_m} 
            +
            \Big\| \,\Big\| \mfrac{1}{n} \msum_{i=1}^n \bA_i^\top \bA_i \Big\|_{op}^{1/2}\, \Big\|_{L_m} 
        \Big)
    \end{align*}
\end{lemma}

\begin{proof}[Proof of \cref{lem:matrix:op:bound}] As $\bA_i$'s are not symmetric, we consider the symmetric matrices 
\begin{align*}
    \bH_i 
    \;\coloneqq\; 
    \begin{pmatrix}
        \bzero & \bA_i \\
        \bA_i^\top & \bzero
    \end{pmatrix}
    \;\in\; \R^{2d \times 2d}
    \;,
\end{align*}
which satisfies the identities
\begin{align*}
    \bH_i^2 
    \;=&\; 
    \begin{pmatrix}
        \bA_i \bA_i^\top & \bzero \\
        \bzero & \bA_i^\top \bA_i
    \end{pmatrix}
    &\text{ and }&&
    \| \bH_i \|_{op} = \| \bA_i \|_{op}\;.
\end{align*}
This allows us to express the quantity of interest in terms of a sum of symmetric matrices 
\begin{align*}
     \Big\| \mfrac{1}{n} \msum_{i=1}^n \bA_i \Big\|_{op}
    \;=\;
    \Big\| \mfrac{1}{n} \msum_{i=1}^n \bH_i \Big\|_{op}\;.
\end{align*}
Let $\varepsilon_1, \ldots, \varepsilon_n$ be i.i.d.~Rademacher variables. By the symmetrization lemma for random vectors (see e.g.~Exercise 6.4.5~of \cite{vershynin2018high}), we have that for $m \geq 1$,
\begin{align*}
    \Big\| \, \Big\| \mfrac{1}{n} \msum_{i=1}^n \bH_i \Big\|_{op}  \, \Big\|_{L_m}
    \;\leq&\; 
    2 \, \Big\| \, \Big\| \mfrac{1}{n} \msum_{i=1}^n \varepsilon_i \bH_i \Big\|_{op}  \, \Big\|_{L_m}
    \\
    \;=&\;
    2 \, \bigg( \mean\bigg[ \; \mean\Big[ \,  \Big\| \mfrac{1}{n} \msum_{i=1}^n \varepsilon_i \bH_i \Big\|_{op}^m  \, \Big| \, (\bH_i)_{i \leq n} \Big] \; \bigg] \bigg)^{1/m}
    \;,
\end{align*}
and by the matrix Khintchine's inequality (see e.g.~Exercise 5.4.13(b) of \cite{vershynin2018high}), there exists some absolute constant $C > 0$ such that almost surely
\begin{align*}
    \mean\Big[ \,  \Big\| \mfrac{1}{n} \msum_{i=1}^n \varepsilon_i \bH_i \Big\|_{op}^m  \, \Big| \, (\bH_i)_{i \leq n} \Big]
    \;\leq\; 
    \bigg( \mfrac{C}{2} \sqrt{ m + \log d} \; \Big\| \mfrac{1}{n^2} \msum_{i=1}^n \bH_i^2 \Big\|_{op}^{1/2} \bigg)^m
    \;.
\end{align*}
Combining the bounds yields
\begin{align*}
    \bigg\| \, \Big\| \mfrac{1}{n} & \msum_{i=1}^n \bA_i \Big\|_{op}  \, \bigg\|_{L_m}
    \;\leq\; 
    C \sqrt{m + \log d} 
    \;\; \bigg\| 
        \, \Big\| \mfrac{1}{n^2} \msum_{i=1}^n \bH_i^2 \Big\|_{op}^{1/2} \,
    \bigg\|_{L_m}
    \\
    \;=&\;
    C \sqrt{m + \log d} 
    \;\; \bigg\| 
        \, \Big\| \mfrac{1}{n^2} \msum_{i=1}^n \begin{pmatrix}
        \bA_i \bA_i^\top & \bzero \\
        \bzero & \bA_i^\top \bA_i
    \end{pmatrix}  \Big\|_{op}^{1/2} \,
    \bigg\|_{L_m}
    \\
    \;\leq&\;
    \mfrac{C \sqrt{m + \log d \,}}{\sqrt{n}} 
    \;
    \Big(
        \Big\| \,\Big\| \mfrac{1}{n} \msum_{i=1}^n \bA_i \bA_i^\top \Big\|_{op}^{1/2}\, \Big\|_{L_m} 
        +
        \Big\| \,\Big\| \mfrac{1}{n} \msum_{i=1}^n \bA_i^\top \bA_i \Big\|_{op}^{1/2}\, \Big\|_{L_m} 
    \Big)
    \;\;.
\end{align*}
\end{proof}

%% file: appendix_4_proof_main.tex
\section{Proof of the main result} \label{appendix:proof:main}

In this section, we prove \cref{thm:main:general}. 
\cref{thm:main}
then follows as a special case. We begin with an outline of the proof technique.

\subsection{Proof overview}
\label{proof:overview}
The main proof idea is based on a technique by \cite{chatterjee2006generalization}, which extends Lindeberg's proof of the central limit theorem
to statistics that are not asymptotically normal.
Chatterjee's approach is as follows:
The goal is to bound the difference
$|\mean[g(\xi_1,\ldots,\xi_n)]-\mean[g(\zeta_1,\ldots,\zeta_n)]|$, for 
independent collections $\xi_1,\ldots,\xi_n$ and
$\zeta_1,\ldots,\zeta_n$ of i.i.d.\ variables and a function $g$. To
this end, abbreviate
$V_i(\argdot)=(\xi_1,\dots,\xi_{i-1},\argdot,\zeta_{i+1},\dots,\zeta_n)$,
and expand into a telescopic sum:
\begin{align*}
    &\mean[g(\xi_1,\ldots,\xi_n)]-    \mean[g(\zeta_1,\ldots,\zeta_n)]
    \;=\;   
    \msum_{i\le n}\mean[ g(V_i(\xi_i)) - g(V_i(\zeta_i))]
    \\&=
    \msum_{i\le n}
    \big(
    \mean[g(V_i(\xi_i)) - g(V_i(0))]
    -\mean[g(V_i(\zeta_i)) - g(V_i(0))] \big)\;.
\end{align*}
By Taylor-expanding the function $g_i(\argdot)\coloneqq g(V_i(\argdot))$ to
third order around $0$, each summand can be represented as 
\begin{align*}
    \mean[ \partial g_i(0) (\xi_i - \zeta_i) ] + \mean[\partial^2 g_i(0) (\xi_i^2 - \zeta_i^2) ] + \mean[\partial^3 g_i(\tilde \xi_i) \xi_i^3 +  \partial^3 g_i(\tilde \zeta_i) \zeta_i^3 ]\;,
\end{align*}
for some $\tilde \xi_i \in[0,\xi_i]$ and $\tilde\zeta_i \in
[0,\zeta_i]$. Since each $(\xi_i, \zeta_i)$ is independent of all
other pairs $\{ (\xi_j, \zeta_j)\}_{j\neq i}$, expectations factorize,
and the expression above becomes
\begin{align}
    \mean[ \partial g_i(0) ] \mean [\xi_i - \zeta_i] + \mean[\partial^2 g_i(0)] \mean[(\xi_i^2 - \zeta_i^2)] + \mean[\partial^3 g_i(\tilde \xi_i) \xi_i^3 +  \partial^3 g_i(\tilde \zeta_i) \zeta_i^3 ]\;. \label{eqn:cartoon:taylor}
\end{align}
The first two terms can then be controlled by matching expectations
and variances of $\xi_i$ and $\zeta_i$. To control the third term, one
imposes boundedness assumptions on $\partial^3 g_i$ and the moments of $\xi_i^3$ and $\zeta_i^3$.

\vspace{.2em}

Proving our result requires some modifications: Since
augmentation induces dependence, the i.i.d.\ assumption above does not
hold. On the other hand, the function $g$ in our problems is of a more
specific form. In broad strokes, our proof proceeds as follows:
\begin{itemize}
    \item We choose $g\coloneqq h \circ f$, where $h$ belongs to the class of
    thrice-differentiable functions with the first three derivatives
    bounded above by 1. Since the statistic $f$ has (by assumption)
    three derivatives, so does $g$.
    \item We group the augmented data into $n$
    independent blocks $\Phi_i\bX_i
    \coloneqq \{\phi_{i1}\bX_i, \ldots, \phi_{ik}\bX_i\}$, for ${i\leq n}$.
    We can then sidestep dependence by applying the technique above to
    each block.
    \item
    To do so, we to take derivates of $g=h\circ f$ with respect to
    blocks of variables. The relevant block-wise version of the chain
    rule is a version of the Fa\`a di Bruno formula.     
    It yields a sum of terms of the form in \eqref{eqn:cartoon:taylor}.
    \item The first two terms in \eqref{eqn:cartoon:taylor} contribute
    a term of order $k$ to the bound: The first expectation vanishes
    by construction. The second also vanishes under the conditions of
    \cref{thm:main},
    and more generally if ${\delta=0}$. If
    ${\delta>0}$, the matrices $\Var[\bZ^{\delta}_i]$ and
    $\Var[\Phi_i\bX_i]$ may differ in their $k$ diagonal entries.
    \item The third term in \eqref{eqn:cartoon:taylor} contributes a
    term of order $k^3$: Here, we use noise stability, which lets us
    control terms involving $\partial^3 g_i$ on the line segments
    $[0,\Phi_i\bX_i]$ and $[0,\bZ_i]$, and moments of
    $(\Phi_i\bX_i)^{\otimes 3}$ and $(\bZ_i)^{\otimes 3}$.
    The moments have dimension of order $k^3$.
    \item Summing over $n$ quantities of the
    form \eqref{eqn:cartoon:taylor} then leads to the bound of the form
    \begin{align*}
        nk \times \text{(second derivative terms)} + nk^3 \times \text{(third derivative terms) }\;.
    \end{align*}
    in \cref{thm:main:general}. In 
    \cref{thm:main}, 
    the first
    term vanishes.
    \end{itemize}

Whether the bound converges depends on the scaling behavior of $f$.
A helpful example is a scaled average
$\sqrt{n} \big(\frac{1}{nk}\sum_{i,j} \phi_{ij}\bX_i\big)$.
Here, the second and third derivatives are respectively of order
$\frac{1}{nk^2}$ and  $\frac{1}{n^{3/2}k^3}$
(see \cref{appendix:empirical_averages} for the exact calculation).
The bound then scales as $\frac{1}{k} + \frac{1}{n^{1/2}}$ for $\delta
> 0$, and as $\frac{1}{n^{1/2}}$ for $\delta = 0$.

\subsection{Proof of Theorem \ref{thm:main:general}} \label{appendix:proof:body:main}
We abbreviate $g \coloneqq h \circ f$, and note that $g$ is a smooth function from $\cD^{nk} $ to $\R$. Recall that we have denoted
\begin{equation*}
  \bW^{\delta}_i(\argdot) 
  \;\coloneqq\;
  (\Phi_1\bX_1,\ldots,\Phi_{i-1}\bX_{i-1},\argdot,\bZ^{\delta}_{i+1},\ldots,\bZ^{\delta}_n)\;.
\end{equation*}
By a telescoping sum argument and the triangle inequality,
\begin{align}
\big| \mean h ( f( \Phi \cX) ) - \mean h ( f( \bZ^{\delta}_1, \ldots, \bZ^{\delta}_n) ) \big| 
&\;=\; 
\big|\mean\msum_{i=1}^n 
 \big[g( \bW^{\delta}_i(\Phi_i \bX_i))  -  g( \bW^{\delta}_i( \bZ^{\delta}_i) ) \big] 
\big| \nonumber
\\&\le \msum_{i=1}^n \big|\mean
 \big[g( \bW^{\delta}_i(\Phi_i \bX_i))  -  g( \bW^{\delta}_i( \bZ^{\delta}_i) ) \big] 
\big| \;.  \label{eqn:thm_main_telescoping_sum}
\end{align}
Each summand can be written as a sum of two terms, 
\begin{equation*}
\big( g( \bW^{\delta}_i(\Phi_i \bX_i))  -  g( \bW^{\delta}_i( \bzero) ) \big)
\,-\,
\big(g( \bW^{\delta}_i( \bZ^{\delta}_i) ) - g(\bW^{\delta}_i(\bzero)) \big)\;.
\end{equation*}
Since $\cD^k$ is convex and contains $\bzero \in \R^{kd}$, we can expand the first term in a Taylor series in the $i^{\text{th}}$ argument of $g$ around $\bzero$ to third order. Then,
\begin{align*}
    &\big| g(\bW^{\delta}_i (\Phi_i \bX_i) ) 
    - 
    g(\bW^{\delta}_i (\bzero) ) 
    - 
    \big(D_i g(\bW^{\delta}_i (\bzero)) \big) (\Phi_i \bX_i) 
    - 
    \mfrac{1}{2}
\big(D_i^2 g(\bW^{\delta}_i (\bzero))\big) \big( (\Phi_i \bX_i) (\Phi_i \bX_i)^\top \big) \big| \\
& \hspace{.5\textwidth} \leq\; \mfrac{1}{6}
\msup_{\bw \in [\bzero, \Phi_i \bX_i] }
\big| D_i^3 g \big(\bW_i (\bw) \big) (\Phi_i \bX_i)^{\otimes 3} \big|
\tagaligneq \label{eq:thm:main:taylor}
\end{align*}
holds almost surely. For the second term, we analogously obtain
\begin{align*}
    &\big| g(\bW^{\delta}_i (\bZ^{\delta}_i) ) 
    - 
    g(\bW^{\delta}_i (\bzero) )
    - 
    \big(D_i g(\bW^{\delta}_i (\bzero)) \big) \bZ^{\delta}_i 
    - 
    \mfrac{1}{2}
\big(D_i^2 g(\bW^{\delta}_i (\bzero))\big) \big( (\bZ^{\delta}_i)(\bZ^{\delta}_i)^\top \big) \big| \\
& \hspace{.5\textwidth} \leq \; \mfrac{1}{6} 
\msup_{\bw \in [\bzero, \bZ^{\delta}_i] }
\big| D_i^3 g \big(\bW^{\delta}_i (\bw) \big) (\bZ^{\delta}_i)^{\otimes 3} \big|
\end{align*} 
almost surely.
Each summand in \eqref{eqn:thm_main_telescoping_sum} is hence bounded above as
\begin{equation} \label{eqn:thm:main:generalized:intermediate}
    \big|\mean\big[g( \bW^{\delta}_i(\Phi_i \bX_i))  -  g( \bW^{\delta}_i( \bZ^{\delta}_i) ) \big]\big|
    \;\leq\;
    |\kappa_{1,i}| + \mfrac{1}{2} |\kappa_{2,i}| + \mfrac{1}{6} | \kappa_{3,i}|\;,
\end{equation}
where
\begin{align*}
&\kappa_{1,i} 
\coloneqq 
\mean \big[
\big(D_i g(\bW^{\delta}_i (\bzero)) \big) 
\big( \Phi_i \bX_i - \bZ^{\delta}_i \big) 
\big] \\
&\kappa_{2,i} 
\coloneqq 
\mean \big[
\big(
D_i^2 g(\bW^{\delta}_i (\bzero)) 
\big) 
\big( 
(\Phi_i \bX_i)(\Phi_i \bX_i)^\top - (\bZ^{\delta}_i) (\bZ^{\delta}_i)^\top
\big) 
\big] \\
&\kappa_{3,i} 
\coloneqq 
\mean \big[
\, \msup_{\bw \in [\bzero, \Phi_i \bX_i] }
\big| 
D_i^3 g \big(\bW^{\delta}_i (\bw) \big) (\Phi_i \bX_i  )^{\otimes 3}
\big| 
+ 
\, \msup_{\bw \in [\bzero, \bZ^{\delta}_i] }
\big| 
D_i^3 g \big(\bW^{\delta}_i (\bw) \big)  (\bZ^{\delta}_i  )^{\otimes 3}
\big|
\big]\;. 
\end{align*}
Substituting into \eqref{eqn:thm_main_telescoping_sum} and applying the triangle inequality shows
\begin{equation*}
\big| \mean h ( f( \Phi \cX) ) - \mean h ( f( \bZ^{\delta}_1, \ldots, \bZ^{\delta}_n) ) \big| 
\;\leq\;
\msum_{i=1}^n \big(|\kappa_{1,i}| + \mfrac{1}{2} |\kappa_{2,i}| + \mfrac{1}{6} | \kappa_{3,i}|\bigr)\;.
\end{equation*}
The next step is to obtain more specific upper bounds for the 
$\kappa_{r,i}$. To this end, first consider
$\kappa_{1,i}$.
Since $(\Phi_i \bX_i, \bZ^{\delta}_i)$ is independent of ${(\Phi_j \bX_j,
  \bZ^{\delta}_j)_{j \neq i}}$, we can factorize the expectation, and obtain
\begin{align*}
\kappa_{1,i} 
\;&=\; 
\mean \big[
D_i g(\bW_i (\bzero)) 
\big] 
\big( 
\mean[ \Phi_i \bX_i] 
- 
\mean[\bZ^{\delta}_i] 
\big)
\;=\;0\;,
\end{align*}
where the second identity holds since
$\mean\bZ^{\delta}_i = \bone_{k \times 1} \otimes \mean[ \phi_{11} \bX_1 ] = \mean[\Phi_i \bX_i]$.
Factorizing the expectation in $\kappa_{2,i}$ shows
\begin{align*}
\kappa_{2,i}
\;&=\; 
\mean \big[
D_i^2 g(\bW_i (\bzero)) 
\big] 
\big( 
\mean \big[ (\Phi_i \bX_i)(\Phi_i \bX_i)^\top \big] - 
\mean \big[ (\bZ^{\delta}_i) (\bZ^{\delta}_i)^\top \big] ) 
\big)\\
 &\;\leq\; 
\big\| 
\mean \big[D_i^2 g(\bW_i (\bzero)) \big]
\big\| 
\big\| 
\mean \big[ (\Phi_i \bX_i)(\Phi_i \bX_i)^\top \big] - \mean \big[ (\bZ^{\delta}_i) (\bZ^{\delta}_i)^\top  \big]  
\big\| \\
&\;\overset{(a)}{=}\; 
\big\| 
\mean \big[D_i^2 g(\bW_i (\bzero)) \big]
\big\| 
\big\| 
\Var[ \Phi_i \bX_i ] - \Var[ \bZ^{\delta}_i ]  
\big\|.
\end{align*}where to obtain (a) we exploited once again the fact that $\mean \bZ^{\delta}_i=\mean[\Phi_i \bX_i]$.
Consider the final norm. Since the covariance matrix of $\Phi_i\bX_i$ is
\begin{align*}
  \Var[ \Phi_i \bX_i ]
  \;=\;
  \bI_{k} \otimes \Var[ \phi_{11} \bX_1 ] + (\bone_{k \times k} -
  \bI_{k}) \otimes \Cov[ \phi_{11} \bX_1, \phi_{12} \bX_1 ]\;,
\end{align*}
the argument of the norm is
\begin{align*}
  \Var[ \Phi_i \bX_i ]
  - \Var[ \bZ^{\delta}_i ]  
  \;=\;
  \delta\bI_k \otimes \big( \Var[ \phi_{11} \bX_1 ] 
- 
\Cov[ \phi_{11} \bX_1, \phi_{12} \bX_1 ] \big)\;.
\end{align*}
Lemma \ref{lem:compare_variance} shows $ \Cov[ \phi_{11} \bX_1, \phi_{12} \bX_1 ] \;=\;
\Var \mean [ \phi_{11} \bX_1 | \bX_1 ]$. It follows that 
\begin{align*}
  \big\|\Var[ \Phi_i \bX_i ] - \Var[ \bZ^{\delta}_i ] \big\|
  &\;=\; 
\delta  
\big\| 
\bI_k \otimes \mean\Var [\phi_{11} \bX_1 | \bX_1] 
\big\|
\;=\; 
2 \delta k^{1/2} c_1\;,
\end{align*}
and hence $\mfrac{1}{2} |\kappa_{2,i}| \;\leq\; \big\| \mean
\big[D_i^2 g(\bW^{\delta}_i (\bzero)) \big] \big\| \delta k^{1/2} c_1$. By applying Cauchy-Scwharz inequality and H\"older's inequality, the term $\kappa_{3,i}$ is upper-bounded by
\begin{align*}
  \kappa_{3,i}
  \;&\leq\;
  \big\| \| \Phi_i \bX_i \|^3 \big\|_{L_2}\,
  \big\|
  {\textstyle\sup_{\bw \in [\bzero, \Phi_i \bX_i] }}
  \| D_i^3 g (\bW^{\delta}_i (\bw) ) \| 
  \big\|_{L_2} \\
\;&+\; 
\big\| \| \bZ^{\delta}_i  \|^3 \big\|_{L_2}
\big\|
{\textstyle\sup_{\bw \in [\bzero, \bZ^{\delta}_i] }}
\| D_i^3 g (\bW^{\delta}_i (\bw) ) \| 
\big\|_{L_2}\;.
\end{align*}
Since the function $x \mapsto x^3$ is convex on $\R_+$, we can apply Jensen's inequality to obtain
\begin{align*}
\big\| \| \Phi_i \bX_i \|^3 \big\|_{L_2}
\;&=\; 
\sqrt{\mean[ \| \Phi_i \bX_i \|^6 ] }
\\
\;&=\;
\sqrt{ \mean \Big[  \big( \msum_{j=1}^k \|\phi_{ij} \bX_i\|^2 \big)^3 \Big] }
\;=\; 
k^{3/2}
\sqrt{  \mean \Big[  \big( \mfrac{1}{k} \msum_{j=1}^k \|\phi_{ij} \bX_i\|^2 \big)^3 \Big] }
\\
\;&\leq\;
k^{3/2}
\sqrt{  \mean \Big[ \mfrac{1}{k} \msum_{j=1}^k \|\phi_{ij} \bX_i\|^6 \Big]  }
    \overset{(a)}{=} 
    k^{3/2} \sqrt{\mean\|\phi_{11} \bX_1 \|^6 }
    = 
    6 k^{3/2} c_X\;,
\end{align*}
where $(a)$ is by noting that for all $i \leq n, j \leq k$, $\phi_{ij} \bX_i$ is identically distributed as $\phi_{11}\bX_1$. On the other hand, by noting that $\bZ^{\delta}_i$ is identically distributed as $\bZ^{\delta}_1$,
\begin{align*}
    \big\| \| \bZ^{\delta}_i  \|^3 \big\|_{L_2} \;=\; \sqrt{ \mean[ \| \bZ^{\delta}_1 \|^6 ] } \;=\; k^{3/2} \sqrt{ \mean\Bigl[ \Bigl(\mfrac{|Z^{\delta}_{111}|^2+\ldots+|Z^{\delta}_{1kd}|^2}{k}\Bigr)^3\Bigr] } \;=\; 6 k^{3/2} c_{Z^{\delta}}\;.
\end{align*}
We can now abbreviate
\begin{equation*}
M_i 
\coloneqq 
\max \braces{ 
\big\|
{\textstyle\sup_{\bw \in [\bzero, \Phi_i \bX_i] }}
\| D_i^3 g \big(\bW^{\delta}_i (\bw) \big) \|\big\|_{L_2} 
,\,
\big\|
{\textstyle\sup_{\bw \in [\bzero, \bZ_i] }}
\| D_i^3 g \big(\bW^{\delta}_i (\bw) \big) \| \big\|_{L_2} 
}\;,
\end{equation*}
and obtain
$\mfrac{1}{6} |\kappa_{3,i}| \;\leq\; k^{3/2} (c_X + c_{Z^{\delta}}) M_i$.
In summary, the right-hand side of
\eqref{eqn:thm_main_telescoping_sum} is hence upper-bounded by
\begin{align*}
  \eqref{eqn:thm_main_telescoping_sum}
  \;&\leq\;
  \delta k^{1/2} c_1
   \Big( 
   \msum_{i=1}^n \big\| \mean \big[D_i^2 g(\bW^{\delta}_i (\bzero)) \big] \big\|
   \Big) 
   \;+\; 
   k^{3/2} (c_X + c_Z) 
   \Big( 
   \msum_{i=1}^n M_i 
   \Big) \\
   \;&\leq\; 
    \delta nk^{1/2} c_1 \, \mmax_{i \leq n} 
   \big\| \mean \big[D_i^2 g(\bW^{\delta}_i (\bzero)) \big] \big\| 
   \;+\; 
   nk^{3/2} (c_X + c_Z) \, \mmax_{i \leq n} M_i\;.
\end{align*}
Lemma \ref{lem:main_thm_chain_rule} below shows that the two maxima
are in turn bounded by
\begin{align} 
&\mmax_{i \leq n} \; \big\| \mean \big[D_i^2 g(\bW^{\delta}_i (\bzero)) \big] \big\|
\;\leq\; 
\gamma_2(h)
    \alpha_{1;2}(f)^2
    +
    \gamma_1(h) \alpha_{2;1}(f) \;=\; 
\lambda_1(n,k), \label{bound:main_thm_second_order} 
\\
&\mmax_{i \leq n} \; M_i 
\;\leq\; 
\gamma_3(h) \alpha_{1;6}(f)^3 + 3 \gamma_2(h) \alpha_{1;4}(f) \alpha_{2;4}(f) + \gamma_1(h) \alpha_{3;2}(f) 
=\lambda_2(n,k). \label{bound:main_thm_third_order} 
\end{align}
That yields the desired upper bound on \eqref{eqn:thm_main_telescoping_sum},
\begin{equation*}
  \big| \mean h ( f( \Phi \cX) ) - \mean h ( f( \bZ^{\delta}_1, \ldots, \bZ^{\delta}_n)
) \big| \leq \delta nk^{1/2} \lambda_1(n,k) c_1 + nk^{3/2}
\lambda_2(n,k) (c_X + c_Z)\;,
\end{equation*}
which finishes the proof.

\begin{remark} \label{remark:main:non:iid} We remark that both 
  \cref{thm:main} 
  and Theorem \ref{thm:main:general} can be generalized directly to independent but not identically distributed vectors $\bX_1, \ldots, \bX_n$. 
  , and that the suprema in the derivative terms can be removed by using a Taylor expansion with integral remainders instead. 
The resultant bound is the following: For some absolute constant $C > 0 $, we have
  \begin{align*}
    &\; 
    \big| \mean h ( f( \Phi \cX) ) - \mean h ( f( \bZ^{\delta}_1, \ldots, \bZ^{\delta}_n)
    ) \big| 
    \\
    &\;\leq\;
    \msum_{i=1}^n 
    \,
    \delta k^{1/2}
    \tilde \chi_1(n,k)
    \,
    \mfrac{\| \mean \Var[ \phi_{i1} \bX_1 | \bX_1 ] \|}{2}
    \,+\,
    \msum_{i=1}^n 
    C k^{3/2}
    \tilde \chi_2(n,k) 
    \mfrac{\sqrt{\mean \| \phi_{i1} \bX_i \|^6 } + \sqrt{\mean \| \bZ_i \|^6 }  }{6}\;,
  \end{align*}
  where 
  \begin{align*}
    &\tilde \chi_1(n,k)
    \coloneqq
  \gamma_2(h)
    \tilde \theta_{1;2} (f)^2 
    +
    \gamma_1(h) \tilde \theta_{2;1} (f)\;,
    \\
    &\tilde \chi_2(n,k)
   \coloneqq
  \gamma_3(h) \tilde \theta_{1;6} (f)^3 + 3 \gamma_2(h) \tilde  \theta_{1;4}(f)  \tilde  \alpha_{2;4}(f) + \gamma_1(h) \tilde  \theta_{3;2}(f)\;,
    \\
    &
      \theta_{r;m}(f)
      \coloneqq
      \sum_{s \leq q} \max_{i\leq n} \max\biggl\{
      \biggl\| 
      \|D_i^r f_s(\bW^{\delta}_i(\Theta \Phi_i\bX_i))\|\biggr\|_{L_m},
        \biggl\|
      \|D_i^r f_s(\bW^{\delta}_i(\Theta \bZ^{\delta}_i))\|\biggr\|_{L_m}
      \biggr\},
\end{align*}
where $\Theta \sim \textrm{Uniform}[0,1]$ is independent of all other random variables and plays the role of the variable to be integrated against in the integral remainders.
\end{remark}

\begin{remark} \label{remark:non:iid:no:taylor} Notice that in the proof of \cref{thm:main:general}, a Cauchy-Schwarz inequality has been taken with respect to the Euclidean norm $\| \argdot \|$ in $\R^d$, which gives a crude upper bound on the dimension dependence. This may not be desirable, e.g.~if the vector inner product involved has a light tail to be exploited, or if the vector product can be rewritten as a sum of $d$ weakly dependent entries. This is the case for the results in \Cref{sec:network}. To obtain a sharper $d$-dependence, instead of \eqref{eq:thm:main:taylor}, we may perform an exact Taylor expansion with the integral remainder without applying the Cauchy-Schwarz inequality to separate $h$ and $f$. In the case with $\delta = 0$ and a first-order Taylor expansion is used, the bound reads 
\begin{align*}
  \big| \mean h ( f( \Phi \cX) ) - \mean h ( f( \bZ_1, \ldots, \bZ_n) ) \big| 
  \;\leq\; 
  \msum_{i=1}^n 
  \big| 
    \;
    \mean\big[  F_{\bW_i, \Theta}(\Phi_i \bX_i)  -  F_{\bW_i, \Theta}(\bZ_i) \big] 
    \;
  \big|
  \;,
\end{align*}
where we have denoted, for $\bx \in \R^{kd}$,
\begin{align*}
  F_{\bW_i, \Theta}( \bx )
  \;\coloneqq&\;
  \partial h( f( \bW_i( \Theta \bx ) ) ) 
  \, 
  \partial_i f( \bW_i( \Theta \bx ) )^\top \bx
  \;,
\end{align*}
and $\Theta \sim \textrm{Uniform}[0,1]$ is independent of all other variables as with \Cref{remark:main:non:iid}.
\end{remark}

\subsection{The remaining bounds}

It remains to establish the bounds in \eqref{bound:main_thm_second_order} and
\eqref{bound:main_thm_third_order}. To this end, we use a
vector-valued version of the generalized chain rule, also known as the
Fa\`a di Bruno formula. Here is a form that is convenient for our
purposes:

\begin{lemma}{\rm [Adapted from Theorem 2.1
  of \cite{constantine1996multivariate}]}
    \label{lem:faa_di_bruno}
    Consider functions ${f\in\cF_3(\cD^{nk},\R^q)}$ and
    ${h\in\cF_3(\R^q,\R)}$, and write ${g\coloneqq h \circ f}$. Then
\begin{align*}
    D_i^2 g (\bu) \;=\;& \partial^2 h\big( f(\bu) \big) \big(D_i f(\bu) \big)^{\otimes 2} + \partial h\big( f(\bu) \big) D_i^2 f(\bu) , \\
    D_i^3 g (\bu) \;=\;& \partial^3 h\big( f(\bu) \big) \big(D_i f(\bu)\big)^{\otimes 3} + 3 \partial^2 h\big( f(\bu) \big) \big(D_i f(\bu) \otimes D^2_i f(\bu)\big) +  \partial h\big( f(\bu) \big)  D_i^3 f(\bu)
\end{align*}
for any ${\bu\in\cD^{nk}}$.
\end{lemma}

We also need the following result for bounding quantities that involve $\zeta_{i;m}$ in terms of noise stability terms $\alpha_{r;m}$ defined in Theorem \ref{thm:main:general}.

\begin{lemma} \label{lem:lower:bound:alpha_r_m}
$ \max_{i \leq n} \zeta_{i;m}(\|D_i^r f(\bW^{\delta}_i(\argdot))\|)  \;\leq\; \alpha_{r;m}(f)\;$.
\end{lemma}
\begin{proof} Note that almost surely
\begin{align*}
    \|D_i^r f(\bW^{\delta}_i(\argdot))\| 
    \;=\; 
    \sqrt{\msum_{s \leq q} \|D_i^r f(\bW^{\delta}_i(\argdot))\|^2} 
    \;\leq\; 
    \msum_{s \leq q} \|D_i^r f_s(\bW^{\delta}_i(\argdot))\|\;.
\end{align*}
Therefore, by triangle inequality of $\zeta_{i;m}$ from Lemma \ref{lem:zeta_properties},
\begin{align*} 
  \alpha_{r;m}(f)
  \coloneqq&
  \sum_{s \leq q} \max_{i\leq n} \max\biggl\{
  \biggl\|{\sup_{\bw\in[\bzero,\Phi_i\bX_i]}}
  \|D_i^r f_s(\bW^{\delta}_i(\bw))\|\biggr\|_{L_m},
    \biggl\|{\sup_{\bw\in[\bzero,\bZ^{\delta}_i]}}
  \|D_i^r f_s(\bW^{\delta}_i(\bw))\|\biggr\|_{L_m}
  \biggr\}
   \\
  =& \msum_{s \leq q} \max_{i \leq n} \zeta_{i;m}(\|D_i^r f_s(\bW^{\delta}_i(\argdot))\|) \;\geq\; \max_{i \leq n} \zeta_{i;m}(\|D_i^r f(\bW^{\delta}_i(\argdot))\|) \;,
\end{align*}
which gives the desired bound.
\end{proof}

\vspace{.3em}

We are now ready to complete the proof for 
\cref{thm:main} 
by proving \eqref{bound:main_thm_second_order} and \eqref{bound:main_thm_third_order}.

\begin{lemma}
\label{lem:main_thm_chain_rule} The bounds \eqref{bound:main_thm_second_order} and \eqref{bound:main_thm_third_order} hold.
\end{lemma}

\begin{proof}
For a random function $\bT: \cD^k \rightarrow \R$, define $\zeta_{i;m}(\bT)$ as in Lemma \ref{lem:zeta_properties} with respect to $\Phi_1 \bX_i$ and $\bZ^{\delta}_i$ from Theorem \ref{thm:main:general},
$$\zeta_{i;m}(\bT) 
\coloneqq 
\max \Big\{ 
\big\| \msup_{\bw \in [\bzero, \Phi_1 \bX_i]} \bT(\bw) \big\|_{L_m} 
, 
\big\| \msup_{\bw \in [\bzero, \bZ^{\delta}_i]} \bT(\bw) \big\|_{L_m}  
\Big\}.$$
We first consider \eqref{bound:main_thm_second_order}. By Lemma \ref{lem:faa_di_bruno}, almost surely,
\begin{align*}
    D_i^2 g (\bW^{\delta}_i(\bzero)) \;=\;& 
    \partial^2 h\big( f(\bW^{\delta}_i(\bzero)) \big) \big(D_i f(\bW^{\delta}_i(\bzero)) \big)^{\otimes 2} 
    + 
    \partial h\big( f(\bW^{\delta}_i(\bzero)) \big) D_i^2 f(\bW^{\delta}_i(\bzero))\;.
\end{align*}
By Jensen's inequality to move $\| \argdot\|$ inside the expectation and Cauchy-Schwarz,
\begin{align*}
    &\;\big\| \mean \big[D_i^2 g(\bW^{\delta}_i (\bzero)) \big]
    \big\|
    \;\leq\;  \zeta_{i;1}\big( \big\| D_i^2 g(\bW^{\delta}_i (\argdot)) \big\|\big) 
    \\
    \leq&\;
    \zeta_{i;1} \big( 
    \big\| \partial^2 h\big(f( \bW^{\delta}_i(\argdot)) \big) \big\| \big\|D_i f(\bW^{\delta}_i(\argdot)) \big\|^2  
    + 
     \big\| \partial h\big( f(\bW^{\delta}_i(\argdot)) \big) \big\|
     \big\|  D^2_i f(\bW^{\delta}_i(\argdot))
      \big\| \big) 
    \\
    \;
    \leq&\;
    \zeta_{i;1} \big(
    \gamma_2(h)
     \|D_i f(\bW^{\delta}_i(\argdot)) \|^2  +
      \gamma_1(h)
      \|  
      D^2_i f(\bW^{\delta}_i(\argdot))
      \| \big) 
      \\
    \;\overset{(a)}{\leq}&\; \gamma_2(h) \; \zeta_2(\| D_i f(\bW^{\delta}_i(\argdot)) \| )^2 
    + 
    \gamma_1(h) \; \zeta_{i;1}(\| D^2_i f(\bW^{\delta}_i(\argdot)) \|) 
    \\
    \;\overset{(b)}{\leq}&\; 
    \gamma_2(h) \,
    \alpha_{1;2}(f)^2 
    +
    \gamma_1(h) \, \alpha_{2;1}(f)
    \;=\; \lambda_1(n,k)\;,
\end{align*} 
where $(a)$ is by H\"older's inequality in \cref{lem:zeta_properties} and $(b)$ is by \cref{lem:lower:bound:alpha_r_m}. Since $\lambda_1(n,k)$ is independent of $i$, we obtain \eqref{bound:main_thm_second_order} as desired:
$$
    \mmax_{1 \leq i \leq n} \big\| \mean \big[D_i^2 g(\bW^{\delta}_i (\bzero)) \big]
    \big\| \;\leq\; \lambda_1(n,k)\;.
$$

\vspace{.3em}

We now want to establish that \eqref{bound:main_thm_third_order} holds. Using Lemma \ref{lem:faa_di_bruno} and the triangle inequality, we obtain that 
 $\| D_i^3 g (\bW^{\delta}_i(\bw)) \| \;\leq\; \bT_{1,i}(\bw) + \bT_{2,i}(\bw) + \bT_{3,i}(\bw),$
where
\begin{align*}
   \bT_{1,i}(\bw) 
   \;&=\; 
   \| \partial^3 h\big( f(\bW^{\delta}_i(\bw)) \big) \| \| D_i f(\bW^{\delta}_i(\bw))\|^3 
   \leq 
   \gamma_3(h) \| D_i f(\bW^{\delta}_i(\bw))\|^3\;, 
   \\
   \bT_{2,i}(\bw) 
   \;&\leq\; 
   3 \gamma_2(h)
   \| D_i f(\bW^{\delta}_i(\bw)) \| \| D^2_i f(\bW^{\delta}_i(\bw)) \|\;, \\
   \bT_{3,i}(\bw) 
   \;&\leq\;  
   \gamma_1(h) 
   \|  D_i^3 f(\bW^{\delta}_i(\bw)) \|\;.
\end{align*}
Then, by triangle inequality of $\zeta_{i;2}$ from Lemma \ref{lem:zeta_properties} (i),
$$
    M_i 
    \;=\; \zeta_{i;2} \big(
    \big\| D_i^3 g \big(\bW^{\delta}_i (\argdot) \big) \big\| \big) 
\;\leq\;
\zeta_{i;2}( \bT_{1,i} ) 
+ 
\zeta_{i;2}( \bT_{2,i} ) 
+ 
\zeta_{i;2}( \bT_{3,i} )\;.
$$
H\"older's inequality of $\zeta_m$ from Lemma \ref{lem:zeta_properties} allows each term to be further bounded as below: 
\begin{align*}
    \zeta_{i;2}(\bT_{1,i}) 
\;\leq&\; 
  \gamma_3(h) \zeta_{i;2} \big( \| D_i f(\bW^{\delta}_i(\argdot))\|^3 \big) 
  \leq 
  \gamma_3(h) \zeta_{i;6} \big( \| D_i f(\bW^{\delta}_i(\argdot))\| \big)^3
  \leq
  \gamma_3(h) \alpha_{1;6}(f)^3 \;,
  \\
  \zeta_{i;2}(\bT_{2,i}) 
  \;\leq&\; 3 \gamma^2(h) \; \zeta_{i;2} \big(   \| D_i f(\bW^{\delta}_i(\argdot)) \| \| D^2_i f(\bW^{\delta}_i(\argdot)) \|
   \big) 
   \\
\;\leq&\; 
   3 \gamma^2(h) \; \zeta_{i;4} \big(   \| D_i f(\bW^{\delta}_i(\argdot)) \| \big) \zeta_{i;4} \big( \| D^2_i f(\bW^{\delta}_i(\argdot)) \|
   \big)
   \;\leq\; 
   3 \gamma_2(h) \alpha_{1;4}(f) \alpha_{2;4}(f) \;,
   \\
   \zeta_{i;2}(\bT_{3,i}) 
    \;\leq&\;
   \gamma_1(h) \; \zeta_{i;2}  \big( \|  D_i^3 f(\bW^{\delta}_i(\argdot)) \|  \big) 
   \;\leq\;
   \gamma_1(h) \alpha_{3;2}(f)\;.
\end{align*}
We have again applied \cref{lem:lower:bound:alpha_r_m} in each of the final inequalities above. Note that all bounds are again independent of $i$. Summing the bounds and taking a maximum recovers \eqref{bound:main_thm_third_order}:
\begin{align*}
    \mmax_{i \leq n} M_i \;&\leq\;
    \gamma_3(h) \, \alpha_{1;6}(f)^3 + 3 \gamma_2(h)  \alpha_{1;4}(f) \alpha_{2;4}(f) + \gamma_1(h)  \alpha_{3;2}(f) \;=\; \lambda_2(n,k)\;.
\end{align*}
\end{proof}

%% file: appendix_5_proof_add_main.tex
\section{Proofs for Appendix \ref{appendix:var:cor}} \label{appendix:proof:var:cor}

\subsection{Proofs for Appendix \ref{appendix:general:main:result}}

The proof for Theorem \ref{thm:main:general} has been discussed in \cref{appendix:proof:main}. In this section we present the proof for \cref{lem:variance:convergence:general} and \cref{lem:d_H:convergence:general}, which shows how Theorem \ref{thm:main:general} can be used to obtain bounds on convergence of variance and convergence in $d_{\cH}$. They are generalizations of 
\cref{cor:variance:convergence} and \cref{cor:d_H:convergence} 
in the main text.

\vspace{.2em}

The main idea in proving Lemma \ref{lem:variance:convergence:general} is to apply the bound on functions of the form $h \circ f$ from Theorem \ref{thm:main:general} with $h$ set to identity and $f$ set to an individual coordinate of $f$ and a product of two individual coordinates of $f$, both scaled up by $\sqrt{n}$. 

\begin{proof}[Proof of Lemma \ref{lem:variance:convergence:general}]
Choose $h(y) \coloneqq y$ for $y \in \R$ and define
\begin{align*}
    &
    f_{rs}(\bx_{11:nk}) \;\coloneqq\; f_r(\bx_{11:nk}) f_s(\bx_{11:nk})\;,
    &&
    \bx_{11:nk} \in \cD^{nk}\;.
\end{align*}
Let $[\argdot]_{r,s}$ denote the $(r,s)$-th coordinate of a matrix. The difference between $f(\Phi\cX)$ and $f(\cZ^{\delta})$ at each coordinate of their covariance matrices can be written in terms of quantities involving $h \circ f_{r_s}$ and $h \circ f_r$:
\begin{align}
    &\; (\Var[ f(\Phi\cX)] )_{r,s} - (\Var[ f(\cZ^{\delta})] )_{r,s} 
    \nonumber\\
    \;=&\;
    \Cov[ f_r(\Phi\cX),  f_s(\Phi\cX)] - \Cov[   f_r(\cZ^{\delta}),  f_s(\cZ^{\delta})]
    \nonumber\\
    \;=&\;
    \mean[h(f_{rs}(\Phi\cX)) - h( f_{rs}(\cZ^{\delta}))] 
    \\
    &\;\;\;- \big(\mean[h(f_r(\Phi\cX))]\mean[h(f_s(\Phi\cX))] - \mean[h(f_r(\cZ^{\delta}))]\mean[h(f_s(\cZ^{\delta}))]\big)
    \nonumber\\
    \;\overset{(a)}{\leq}&\;
    \big|\mean[h(f_{rs}(\Phi\cX)) - h( f_{rs}(\cZ^{\delta}))] \big| 
    + 
    \big|\mean[h(f_r(\Phi\cX)) - h(f_r(\cZ^{\delta}))] \big| \big| \mean [h(f_s(\Phi\cX))] \big|
    \nonumber\\
    &\;+
    \big| \mean [h(f_r(\cZ^{\delta}))] \big| \big|\mean[h(f_s(\Phi\cX)) - h(f_s(\cZ^{\delta}))] \big| 
    \nonumber\\
    \;\overset{(b)}{\leq}&\; T(f_{rs}) + T(f_r) \alpha_{0;1}(f_s) + T(f_s) \alpha_{0;1}(f_r) \;, \label{eqn:conv:variance:intermediate:bound}
\end{align}
In $(a)$, we have added and subtracted $\mean[h(f_r(\cZ))] \mean[h(f_s(\Phi\cX))]$ from the second difference before applying Cauchy-Schwarz inequality. In $(b)$, we have used the noise stability term $\alpha_{r;m}$ defined in Theorem \ref{thm:main:general} and defined the quantity $T(f^*)\coloneqq\mean[h(f^*(\Phi\cX))-h(f^*(\cZ))]$.

\vspace{.3em}

We now proceed to bound $T(f)$ using Theorem \ref{thm:main:general}. First note that $\gamma_1(h)=|\partial h(0)|=1$ and $\gamma_2(h)=\gamma_3(h)=0$. To bound $T(f^*)$ for a given $f^*:\R \rightarrow \R$, making the dependence on $f^*$ explicit, the mixed smoothness terms in Theorem \ref{thm:main:general} is given by
\begin{align*}
    &
    \lambda_1(n,k;f^*) \;=\;\alpha_{2;1}(f^*)\;,
    &&
    \lambda_2(n,k;f^*) \;=\;\alpha_{3;4}(f^*)\;,
\end{align*}
and therefore Theorem \ref{thm:main:general} implies
\begin{align} \label{eqn:var:bound:T_f}
    T(f^*) \;\leq\; \delta nk^{1/2} \alpha_{2;1}(f^*) c_1 + nk^{3/2} \alpha_{3;2}(f^*) (c_X + c_{Z^{\delta}})\;.
\end{align}
Applying \eqref{eqn:var:bound:T_f} to $f_r$ and $ f_s$ allows the last two terms in \eqref{eqn:conv:variance:intermediate:bound} to be bounded as:
\begin{align}
    T(f_r) \alpha_{0;1}(f_s) + T(f_s) &\alpha_{0;1}(f_r) 
    \nonumber \\
    \;\leq&\; 
    \delta nk^{1/2} \big( \alpha_{2;1}(f_r) \alpha_{0;1}(f_s) + \alpha_{2;1}(f_s) \alpha_{0;1}(f_r) \big) c_1
    \nonumber 
    \\
    &\;+
    nk^{3/2} \big(\alpha_{3;2}(f_r) \alpha_{0;1}(f_s) + \alpha_{3;2}(f_s) \alpha_{0;1}(f_r)\big) (c_X+c_{Z^\delta})\;. 
    \label{eqn:conv:variance:linear_term}
\end{align}
To apply \eqref{eqn:var:bound:T_f} to $T(f_{rs})$, we need to compute bounds on the partial derivatives of $ f_{rs}$:
\begin{align*}
    \| D_i  f_{rs}(\bx_{11:nk}) \|
    \;\leq&\;
    | f_r(\bx_{11:nk}) | \, \|\partial f_s(\bx_{11:nk})\| +
    \|\partial f_r(\bx_{11:nk}) \| \; |f_s(\bx_{11:nk}) |
    \;,
    \\
    \| D^2_i  f_{rs}(\bx_{11:nk}) \|
    \;\leq&\;
    | f_r(\bx_{11:nk}) | \, \|\partial^2 f_s(\bx_{11:nk})\|
    +
    2 \|\partial f_r(\bx_{11:nk})\| \, \|\partial f_s(\bx_{11:nk})\|
    \\
    &\; +
    \|\partial^2 f_r(\bx_{11:nk}) \| \; |f_s(\bx_{11:nk}) |
    \;,
    \\
    \| D^3_i  f_{rs}(\bx_{11:nk}) \|
    \;\leq&\;
    | f_r(\bx_{11:nk}) | \, \|\partial^3 f_s(\bx_{11:nk})\|
    +
    3 \|\partial f_r(\bx_{11:nk})\| \, \|\partial^2 f_s(\bx_{11:nk})\|
    \\
    &
    +
    3 \|\partial^2 f_r(\bx_{11:nk})\| \, \|\partial f_s(\bx_{11:nk})\|
    +
    \|\partial^3 f_r(\bx_{11:nk}) \| \; |f_s(\bx_{11:nk}) | 
    \;.
\end{align*}
Since $ f_{rs}$ and $f_r$ both output variables in 1 dimension, recall from \cref{lem:lower:bound:alpha_r_m} that noise stability terms can be rewritten in terms of $\zeta_{i;m}$ in Lemma \ref{lem:zeta_properties}:
\begin{align*}
    \alpha_{R;m}( f_{rs}) = \mmax_{i \leq n} \zeta_{i;m}( \| D^{R}_i  f_{rs} (\bW_i(\argdot)) \|)\;,
    \;
    \alpha_{R;m}(f_r) = \mmax_{i \leq n} \zeta_{i;m}( \| D^{R}_i f_r(\bW_i(\argdot)) \|)\;.
\end{align*}
By triangle inequality, positive homogeneity and H\"older's inequality of $\zeta_m$ from Lemma \ref{lem:zeta_properties}, we get
\begin{align}
    \alpha_{2;1}(&f_{rs} ) \;=\; \mmax_{i \leq n} \zeta_{i;2}( \| D^2_i f_{rs}(\bW_i(\argdot)) \| ) 
    \nonumber \\
    \;\leq&\;
     \mmax_{i \leq n} \Big( \zeta_{i;4}( |f_r(\bW_i(\argdot)) | ) \, \zeta_{i;4}(\|\partial^2 f_s(\bW_i(\argdot))\|) 
    \nonumber \\
    &\;\qquad\qquad + 2 \zeta_{i;4}(\|\partial f_r(\bW_i(\argdot))\| ) \, \zeta_{i;4}( \|\partial f_s(\bW_i(\argdot))\| ) 
    \nonumber \\
    &\;\qquad\qquad + \zeta_{i;4}( \| \partial^2 f_r(\bW_i(\argdot)) \| ) \, \zeta_{i;4}(| f_s(\bW_i(\argdot))|)  \, \Big)
    \nonumber \\
    \;\leq&\; \alpha_{0;4}(f_r) \alpha_{2;4}(f_s) + 2 \alpha_{1;4}(f_r) \alpha_{1;4}(f_s) + \alpha_{2;4}(f_r) \alpha_{0;4}(f_s)  \;, \label{eqn:conv:variance:f_rs_alpha_two}
    \\
    \alpha_{3;2}( &f_{rs}) \;=\; \mmax_{i \leq n} \zeta_{i;2}( \| D^3_i  f_{rs}(\bW_i(\argdot)) \| ) 
    \nonumber \\
    \;\leq&\;  
    \mmax_{i \leq n} \Big( \zeta_{i;4}( |f_r(\bW_i(\argdot)) | ) \, \zeta_{i;4}(\|\partial^3 f_s(\bW_i(\argdot))\|) 
    \nonumber \\
    &\;\qquad\qquad + 3 \zeta_{i;4}(\|\partial f_r(\bW_i(\argdot))\| ) \, \zeta_{i;4}( \|\partial^2 f_s(\bW_i(\argdot))\| ) 
    \nonumber \\
    &\;\qquad\qquad + 3 \zeta_{i;4}(\|\partial^2 f_r(\bW_i(\argdot))\| ) \, \zeta_{i;4}( \|\partial f_s(\bW_i(\argdot))\| ) 
    \nonumber \\
    &\;\qquad\qquad + \zeta_{i;4}( \| \partial^3 f_r(\bW_i(\argdot)) \| ) \, \zeta_{i;4}(| f_s(\bW_i(\argdot))|)  \, \Big)
    \nonumber \\
    \;\leq&\; \alpha_{0;4}(f_r) \alpha_{3;4}(f_s) + 3 \alpha_{1;4}(f_r) \alpha_{2;4}(f_s) + 3  \alpha_{2;4}(f_r) \alpha_{1;4}(f_s) + \alpha_{3;4}(f_r) \alpha_{0;4}(f_s)  \;. \label{eqn:conv:variance:f_rs_alpha_three}
\end{align}
Therefore by \eqref{eqn:var:bound:T_f}, we get
\begin{align*}
        T( f_{rs}) \;\leq\; \delta nk^{1/2} \times \eqref{eqn:conv:variance:f_rs_alpha_two} \times c_1 +  nk^{3/2} \times \eqref{eqn:conv:variance:f_rs_alpha_three} \times (c_X+c_{Z^{\delta}})\;.
\end{align*}
Substitute this and the bound obtained in \eqref{eqn:conv:variance:linear_term} for $T(f_r)$ and $T(f_s)$ into \eqref{eqn:conv:variance:intermediate:bound}, we get
\begin{align*}
    &\;(\Var[ f(\Phi\cX)] )_{r,s} - (\Var[ f(\cZ)] )_{r,s} 
    \\
    \;\leq&\;
    \delta nk^{1/2} c_1 \times 
    \big( \alpha_{2;1}(f_r) \alpha_{0;1}(f_s) + \alpha_{2;1}(f_s) \alpha_{0;1}(f_r) + \alpha_{0;4}(f_r) \alpha_{2;4}(f_s) 
    \\
    &\;\qquad\qquad\quad\;
    + 2 \alpha_{1;4}(f_r) \alpha_{1;4}(f_s) + \alpha_{2;4}(f_r) \alpha_{0;4}(f_s) \big)
    \\
    &\;+ nk^{3/2} (c_X + c_{Z^\delta}) \times \big(
    \alpha_{3;2}(f_r) \alpha_{0;1}(f_s) + \alpha_{3;2}(f_s) \alpha_{0;1}(f_r)
    + \alpha_{0;4}(f_r) \alpha_{3;4}(f_s) 
    \\
    &\;\qquad\qquad\qquad\qquad\quad\; + 3 \alpha_{1;4}(f_r) \alpha_{2;4}(f_s)  + 3  \alpha_{2;4}(f_r) \alpha_{1;4}(f_s) + \alpha_{3;4}(f_r) \alpha_{0;4}(f_s)  
    \big)\;.
\end{align*}
Note that summation of each term above over $1 \leq r,s \leq q$ can be computed as
\begin{align*}
\big( \msum_{r=1}^q \alpha_{R_1;m_1}(f_r) \big) \big( \msum_{s=1}^q \alpha_{R_2;m_2}(f_s) \big)
\;\overset{(a)}{=}\; \alpha_{R_1;m_1}(f) \alpha_{R_2;m_2}(f)\;.
\end{align*}
Therefore,
\begin{align*}
    &\; \| \Var[ f(\Phi\cX)] - \Var[ f(\cZ)]  \| 
    \;\leq\; \msum_{r,s=1}^q \big| [\Var[ f(\Phi\cX)] ]_{r,s} - [\Var[ f(\cZ)] ]_{r,s} \big|
    \\
    \;\leq&\;
    \delta nk^{1/2} c_1 ( 2 \alpha_{2;1}(f) \alpha_{0;1}(f) + 2 \alpha_{0;4}(f) \alpha_{2;4}(f) + 2 \alpha_{1;4}(f) \alpha_{1;4}(f) )
    \\
    &\; +
    nk^{3/2}(c_X + c_{Z^{\delta}}) ( 2 \alpha_{3;2}(f) \alpha_{0;1}(f) + 2 \alpha_{0;4}(f) \alpha_{3;4}(f) + 6\alpha_{1;4}(f) \alpha_{2;4}(f) )
    \\
    \;\overset{(b)}{\leq}&\; 
    4 \delta nk^{1/2} ( \alpha_{0;4} \alpha_{2;4} + \alpha_{1;4}^2 ) c_1
    + 6 nk^{3/2} ( \alpha_{0;4} \alpha_{3;4} + \alpha_{1;4} \alpha_{2;4} ) (c_X + c_{Z^{\delta}})
    \;.
\end{align*}
In $(a)$, we have used \cref{lem:lower:bound:alpha_r_m}. In $(b)$, we have omitted $f$-dependence and used that $\alpha_{2;1} \alpha_{0;1} \leq \alpha_{0;4} \alpha_{2;4}$ and $\alpha_{3;2} \alpha_{0;1} \leq \alpha_{3;4} \alpha_{0;4}$. Multiplying across by $n$ gives the desired result.
\end{proof}

To prove \cref{lem:d_H:convergence:general}, we only need to apply the bound on $h \circ f$ from Theorem \ref{thm:main:general} with $f$ replaced by $\sqrt{n} f$.

\begin{proof}[Proof of \cref{lem:d_H:convergence:general}] Recall that for any $h \in \cH$, $\gamma^1(h), \gamma^2(h), \gamma^3(h) \leq 1$. Moreover, for $\zeta_{i;m}$ defined in Lemma \ref{lem:zeta_properties},
\begin{align*}
    \alpha_{r;m}( \sqrt{n} f) \;&=\; \mmax_{i \leq n} \zeta_{i;m}( \|\sqrt{n} \, D_i^r f(\bW_i(\argdot) ) \|) 
    \\ 
    \;&=\; \sqrt{n}  \mmax_{i \leq n} \zeta_{i;m}( \| \, D_i^r f(\bW_i(\argdot) ) \|) \;=\; \sqrt{n} \alpha_{r;m}(f)\;.
\end{align*}
Therefore, Theorem \ref{thm:main:general} implies that for every $h \in \cH$,
\begin{align*}
  \bigl|
  \mean h( \sqrt{n}  f(\Phi\cX)) &
  -
  \mean h( \sqrt{n} f(\cZ^{\delta}))
  \bigr|
  \\
  \;\leq&\;
  \delta nk^{1/2} c_1 \big( n \alpha_{1;2} (f)^2 + n^{1/2} \alpha_{2;1} (f) \big)
  \\
  &\;
  +
  nk^{3/2} \big( n^{3/2} \alpha_{1;6}(f)^3  + 3  n \alpha_{1;4}(f) \alpha_{2;4}(f) + n^{1/2} \alpha_{3;2}(f) \big)(c_X+c_{Z^{\delta}})\;.
\end{align*}
Taking a supremum over all $h \in \cH$ and omitting $f$-dependence imply that 
\begin{align*}
    &\;
    d_{\cH}(\sqrt{n} f(\Phi\cX), \sqrt{n} f(\cZ^{\delta})) 
    \;=\; 
    \msup_{h \in \cH} \bigl|
  \mean h( \sqrt{n} f(\Phi\cX))
  -
  \mean h( \sqrt{n} f(\cZ^{\delta}))
  \bigr|
  \\ 
  \;\leq&\; 
  \delta n^{3/2}k^{1/2} c_1 \big( n^{1/2} \alpha_{1;2}^2 + \alpha_{2;1} \big)
  +
  (nk)^{3/2} ( n \alpha_{1;6}^3  + 3 n^{1/2} \alpha_{1;4} \alpha_{2;4} + \alpha_{3;2} )(c_X+c_{Z^{\delta}})\;,
\end{align*}
which is the desired bound.
\end{proof}

\subsection{Proofs for \cref{appendix:remark:main}}

\vspace{.2em}

We give the proofs for \cref{lem:sufficient:cond:vary:q}, which concerns convergence when dimension of the statistic $q$ is allowed to grow, and for \cref{cor:thm:main:invariance}, which formulates our main result with the assumption of invariance. Both proofs are direct applications of Theorem \ref{thm:main:general}. The proof of \cref{cor:thm:main:smaller:var} is not stated as it is just obtained by setting $\delta=1$ in \cref{thm:main:general}. 

\begin{proof}[Proof of \cref{lem:sufficient:cond:vary:q}] By assumption, the noise stability terms satisfy
\begin{align*}
    \alpha_1 = o( n^{-5/6}k^{-1/2} d^{-1/2}),
    \;
    \alpha_3 = o( n^{-3/2}k^{-3/2} d^{-3/2}),
    \;
    \alpha_0 \alpha_3, \alpha_1\alpha_2 = o(n^{-2}k^{-3/2}d^{-3/2}).
\end{align*}
Since each coordinate of $\phi_{11}\bX_1$ and $\bZ_1$ is $O(1)$, the moment terms satisfy
\begin{align*}
    c_X \;&=\; \mfrac{1}{6}\big(\mean[\|\phi_{11}\bX_1\|^4]\big)^{3/4} \;=\;   \mfrac{1}{6}\big(\mean\big[ \big( \msum_{s=1}^d (\be_s^\top \phi_{11}\bX_1 )^2 \big)^2 \big] \big)^{3/4} \;=\; O(d^{3/2})\;,
    \\
    c_Z \;&=\; \mfrac{1}{6} \big( \mean\big[ \big( \mfrac{1}{k} \msum_{j \leq k, s\leq d} |Z_{1jd}|^2 \big)^2 \big]\big)^{3/4} \;=\; O(d^{3/2})\;.
\end{align*}
The condition on $\alpha_r$'s imply that the bound in 
\Cref{cor:variance:convergence}, 
with $\delta$ set to $0$, becomes
\begin{align*}
    n \big\| \Var[f(\Phi\cX)] - \Var[f(\bZ_1, \ldots, \bZ_n)]  \big\| 
    \;\leq&\; 6 n^2 k^{3/2}(c_X + c_Z) ( \alpha_0 \alpha_3 + \alpha_1 \alpha_2 ) 
    \;=\; o(1) \;.
\end{align*}
Since $\alpha_r(f_s) \leq \alpha_r(f)$ by definition of $\alpha_r$, the above bounds hold for $\alpha_r(f_s)$. Applying 
\Cref{cor:d_H:convergence} 
to $f_s$ gives
\begin{align*}
    &d_{\cH}( \sqrt{n} f_s(\Phi\cX), \sqrt{n} f_s(\bZ_1, \ldots, \bZ_n) ) 
    \\
    &\;\leq\; n^{3/2}k^{3/2} ( n \alpha_1(f_s)^3  + 3 n^{1/2} \alpha_1(f_s) \alpha_2(f_s) + \alpha_3(f_s) )(c_X + c_Z) \;=\;o(1)\;.
\end{align*}
By 
\Cref{lem:d_H}, 
convergence in $d_{\cH}$ implies weak convergence, which gives the desired result.
\end{proof}

\begin{proof}[Proof of \cref{cor:thm:main:invariance}] By law of total variance,
\begin{align*}
    \Sigma_{11} \;\coloneqq\; \Var[\phi_{11}\bX_1] \;=\; \tilde \Sigma_{11}+ \Var \mean[\phi_{11}\bX_1 | \phi_{11}]\;, 
\end{align*}
and by distributional invariance assumption, almost surely,
\begin{align*}
    \mean[\phi_{11}\bX_1 | \phi_{11}] \;=\; \mean[\phi_{12}\bX_1 | \phi_{12}] \;=\; \mean[\bX_1] \;.
\end{align*}
This implies $\Var \mean[\phi_{11}\bX_1 | \phi_{11}]$ vanishes and therefore $\Sigma_{11} = \tilde \Sigma_{11}$. The equality in $\Sigma_{12}$ is directly from Lemma \ref{lem:compare_variance}.
\end{proof}

\subsection{Proofs for \cref{appendix:plugin}}

We present the proofs for the two results of \cref{lem:plugin:formal} for plug-in estimates. The following lemma is analogous to \cref{lem:lower:bound:alpha_r_m} but for $\kappa_{r;m}$, and will be useful in the proof.

\begin{lemma} \label{lem:lower:bound:kappa_r_m}
    $\big\|{\msup_{\bw\in [\bzero, \bar\bX]}}
        \| \partial^r g( \mu + \bw ) \| \big\|_{L_m} \leq \kappa_{r;m}(g)\;$.
\end{lemma}
\begin{proof} By the definition of $\kappa_{r;m}$ and a triangle inequality,
\begin{align*}
  \kappa_{r;m}(g) \coloneqq \sum_{s \leq q} \big\|{\msup_{\bw\in [\bzero, \bar\bX]}}
  \big\| \partial^r g_s\big( \mu + \bw \big) \big\| \big\|_{L_m}
  \geq
   \big\|{\msup_{\bw\in [\bzero, \bar\bX]}}
  \big\| \partial^r g\big( \mu + \bw \big) \big\| \big\|_{L_m}\;,
\end{align*}
which is the desired bound.
\end{proof}

For the proof of Lemma \ref{lem:plugin:formal}(i), we first compare $g$ to its first-order Taylor expansion. The Taylor expansion only involves an empirical average, whose weak convergence and equality in variance are given by Lemma \ref{lem:variance:convergence:general} and Lemma \ref{lem:d_H:convergence:general} in a similar manner as the proof for 
\Cref{prop:empirical_average}. 
We recall that $\cD$ is assumed to be a convex subset in $\R^d$ containing $\bzero$, which is important for the Taylor expansion argument.

\begin{proof}[Proof of Lemma \ref{lem:plugin:formal}(i)] We first prove the bound in $d_{\cH}$. Using a triangle inequality to separate the bound into two parts, we get
\begin{align}
    d_{\cH}( \sqrt{n} f(\Phi \cX), &\sqrt{n} f^T(\cZ^{\delta}) ) 
    \;=\;
    \msup_{h \in \cH} | \mean [h( \sqrt{n} f(\Phi \cX)) - \mean[h( \sqrt{n} f^T(\cZ^{\delta}) )] | 
    \nonumber \\
    \;&\leq\; d_{\cH}( \sqrt{n} f(\Phi \cX), \sqrt{n} f^T(\Phi\cX) ) + d_{\cH}( \sqrt{n} f^T(\Phi \cX), \sqrt{n} f^T(\cZ^{\delta}) )\;. \label{eqn:lem:plugin:formal:d_H:triangle}
\end{align}
Consider bounding the first term of \eqref{eqn:lem:plugin:formal:d_H:triangle}. Since $f(\Phi \cX) = g\big(\bar\bX + \mu\big)$ and $f^T(\Phi \cX) = g(\mu) + \partial g(\mu) \bar\bX$, a Taylor expansion argument on $g\big(\bar\bX + \mu\big)$ gives
\begin{align*}
   \big\| f(\Phi \cX) - f^T(\Phi\cX) \big\| 
   \;\leq\;\msup_{\bw \in [\bzero, \bar\bX]} \big\|\partial^2 g(\mu+\bw) \big\| \; \big\|\bar\bX\big\|^2\;.
\end{align*}
Recall that $\gamma_1(h)=\sup_{\bw \in \R^q} \{ \| \partial h(\bw) \| \}$. By mean value theorem, the above bound and H\"older's inequality, we get
\begin{align*}
  |
  \mean h(\sqrt{n} f(\Phi\cX))
  - 
  \mean h(\sqrt{n} f^T(\Phi\cX) )
  |
  \leq&\;
  \sqrt{n} \, \gamma_1(h) \, \mean \| f(\Phi\cX) -  f^T(\Phi\cX) \|
  \\
  \;\leq&\; \sqrt{n} \, \gamma_1(h) \, \mean\big[ \msup_{\bw \in [\bzero, \bar\bX]} \| \partial^2 g(\mu+\bw) \| \; \big\|\bar\bX\big\|^2 \big]
  \\
  \;\leq&\;  \sqrt{n} \, \gamma_1(h) \, \big\| \msup_{\bw \in [\bzero, \bar\bX]} \| \partial^2 g(\mu+\bw) \| \big\|_{L_3} \; \big\| \, \|\bar\bX\| \, \big\|_{L_3}^2
  \\
  \;\leq&\; \sqrt{n} \, \gamma_1(h) \, \kappa_{2;3}(g) \, \big\| \, \|\bar\bX\| \, \big\|_{L_3}^2\;.
\end{align*}
In the last inequality we have used \cref{lem:lower:bound:kappa_r_m}.
To control the moment term, we use Rosenthal's inequality for vectors from Corollary \ref{cor:rosenthal:vector}. Since $\phi_{ij}\bX_i$ have bounded 6th moments, for each $2 \leq m \leq 6$, there exists a constant $K_m$ depending only on $m$ such that
\begin{align}
    &\;\big\| \, \big\|\bar\bX\big\| \, \big\|_{L_m}\;=\; \Big\| \Big\| \mfrac{1}{nk}\msum_{i=1}^n \msum_{j=1}^k \phi_{ij} \bX_i - \mu \Big\| \Big\|_{L_m}
    \nonumber \\
    \;\leq&\;  \mfrac{K_m}{n} \bigg( 
    \sum_{s=1}^d  \max\bigg\{ \bigg(\sum_{i=1}^n \bigg\| \mfrac{1}{k} \sum_{j=1}^k (\phi_{ij}\bX_i - \mu)_s \bigg\|^m_{L_m} 
    \bigg)^{2/m},\;
    \sum_{i=1}^n \bigg\| \mfrac{1}{k} \sum_{j=1}^k (\phi_{ij}\bX_i - \mu)_s \big\|^2_{L_2} 
    \bigg\} 
    \bigg)^{1/2}
    \nonumber \\
    \;=&\; \mfrac{K_m}{\sqrt{n}} \Big( 
    \msum_{s=1}^d  \max\big\{ n^{\frac{2}{m}-1} \big\| \mfrac{1}{k} \msum_{j=1}^k (\phi_{1j}\bX_1 - \mu0_s \big\|^2_{L_m} 
    ,\;
    \big\| \mfrac{1}{k} \msum_{j=1}^k (\phi_{1j}\bX_1 - \mu)_s \big\|^2_{L_2} 
    \big\} 
    \Big)^{1/2}
    \nonumber \\
    \;=&\; O(n^{-1/2} \bar c_m) \;. \label{eqn:plugin:rosenthal} 
\end{align}
Substituting this into the bound above, we get
\begin{align*}
  \big|
  \mean h(\sqrt{n} f(\Phi\cX))
  - 
  \mean h\big(\sqrt{n} f^T(\Phi\cX) ) \big)
  \big|
    \;=\; O\big( n^{-1/2} \gamma^1(h) \, \kappa_{2;3}(g) \, \bar c_3^2 \big)\;.
\end{align*}
Since for all $h \in \cH$, $\gamma^1(h) \leq 1$, taking supremum of the above over $h \in \cH$ gives the bound for the first term of \eqref{eqn:lem:plugin:formal:d_H:triangle}: 
\begin{align}
    d_{H}\big( \sqrt{n} f(\Phi\cX), \sqrt{n} f^T(\Phi\cX) \big)
    \;=\; O\big( n^{-1/2} \, \kappa_{2;3}(g) \, \bar c_3^2 \big)\;.
     \label{eqn:plugin:taylor:first_order:d_H_star_bound}
\end{align}
The second term of \eqref{eqn:lem:plugin:formal:d_H:triangle} can be bounded in the usual way by applying Lemma \ref{lem:d_H:convergence:general} to $f^T(\bx_{11:nk}) = g(\mu) + \partial g(\mu) \big( \frac{1}{nk}\sum_{i,j} \bx_{ij} - \mu \big)$. Let $f^T_s$ denote the $s$th coordinate of $f^T$. The partial derivatives are given by:
\begin{align*}
    &
    \Big\| \mfrac{\partial f^T_s(\bx_{11:nk})}{\partial \bx_{ij}} \Big\| 
    = \mfrac{1}{nk} \| \partial g_s(\mu) \|, 
    &&
    \Big\| \mfrac{\partial^2 f^T_s(\bx_{11:nk})}{\partial \bx_{ij_1} \partial \bx_{ij_2}} \Big\| 
    =
    \Big\| \mfrac{\partial^3 f^T_s(\bx_{11:nk})}{\partial \bx_{ij_1} \partial \bx_{ij_2}  \partial \bx_{ij_3}} \Big\| =0.
\end{align*}
This implies that for $s \leq q$,
\begin{align*}
    &
    \| D_i f^T_s(\bx_{11:nk}) \| \;=\; \mfrac{1}{n k^{1/2}} \| \partial g_s(\mu) \|\;,
    &&
    \| D^2_i f^T_s(\bx_{11:nk}) \| \;=\; 
    \| D^3_i f^T_s(\bx_{11:nk}) \| \;=\; 0\;.
\end{align*}
Therefore we have $\alpha_{1;m}\big(f^T\big) = \sum_{s=1}^q n^{-1}k^{-1/2} \| \partial g_s(\mu) \| \leq n^{-1}k^{-1/2} \kappa_{1;1}(g)$ by \cref{lem:lower:bound:kappa_r_m}, and $\alpha_{2;m}\big(f^T\big) = \alpha_{3;m}\big(f^T\big) = 0$. The bound in Lemma \ref{lem:d_H:convergence:general} then becomes
\begin{align*}
    &\;
    \delta n^{3/2}k^{1/2} c_1 \big( n^{1/2} (\alpha_{1;2})^2 + \alpha_{2;1} \big)
    +
    (nk)^{3/2} ( n (\alpha_{1;6})^3  + 3 n^{1/2} \alpha_{1;4} \alpha_{2;4} + \alpha_{3;2} )(c_X+c_{Z^{\delta}})
    \\
    =
    &\;
    O\big( \delta k^{-1/2} \kappa_{1;1}(g)^2 c_1 +  n^{-1/2}  \kappa_{1;1}(g)^3 (c_X + c_{Z^{\delta}}) \big)\;,
\end{align*}
which implies
\begin{align*}
    d_{\cH}( \sqrt{n} f^T(\Phi\cX), \sqrt{n} f^T(\cZ^{\delta}) )\;=\;  O\big(  \delta k^{-1/2} \kappa_{1;1}(g)^2 c_1 +  n^{-1/2}  \kappa_{1;1}(g)^3 (c_X + c_{Z^{\delta}}) \big)\;.
\end{align*}
Substituting this into \eqref{eqn:lem:plugin:formal:d_H:triangle} together with the bound in \eqref{eqn:plugin:taylor:first_order:d_H_star_bound} gives the required bound
\begin{align*}
    d_{\cH}\big( \sqrt{n} f(\Phi\cX), \sqrt{n} & f^T(\cZ^{\delta})\big)
    \;=\;
    O\big( n^{-1/2} \kappa_{2;3} \, \bar c_3^2 + \delta k^{-1/2} \kappa_{1;1}^2 c_1 +  n^{-1/2}  \kappa_{1;1}^3 (c_X + c_{Z^{\delta}}) \big)\;,
\end{align*}
where we have omitted $g$-dependence.

\vspace{.3em} 

Recall that $\Sigma_{11}=\Var[ \phi_{11} \bX_1]$ and $\Sigma_{12}=\Cov[ \phi_{11} \bX_1, \phi_{12} \bX_1]$. For the bound on variance, we first note that by the variance condition on $\bZ^{\delta}_i$ from \eqref{eqn:defn:surrogates:delta}, we get
\begin{align*}
    \Var[\bar \bX] - \Var[\bar \bZ^{\delta}] \;=&\; \mfrac{1}{n} \Big( \mfrac{1}{k} \Sigma_{11}  + \mfrac{k-1}{k} \Sigma_{12} \Big) - \mfrac{1}{n} \Big( \mfrac{1}{k} ((1-\delta)\Sigma_{11} + \delta \Sigma_{12})  + \mfrac{k-1}{k} \Sigma_{12} \Big) 
    \\
    \;=&\; \mfrac{\delta}{nk}(\Sigma_{11} - \Sigma_{12})\;.
\end{align*}
This implies
\begin{align}
   n \|  \Var[f^T(\Phi\cX)] - \Var[ f^T(\cZ)] \| \;=&\; n \|  \Var[g(\mu)+\partial g(\mu) \bar \bX ]  - \Var[g(\mu)+\partial g(\mu) \bar \bZ^{\delta} ] \|
   \nonumber \\
   \;=&\; n \big\| \partial g(\mu) \Var[\bar \bX] \partial g(\mu)^\top  - \partial g(\mu) \Var[\bar \bZ^{\delta}] \partial g(\mu)^\top \big\|
    \nonumber \\
    \;=&\; n \big\| \mfrac{\delta}{nk} \partial g(\mu) (\Sigma_{11} - \Sigma_{12}) \partial g(\mu)^\top \big\| 
    \nonumber \\
    \;\leq&\; \mfrac{4\delta}{k} \| \partial g(\mu) \|_2^2 \, c_1^2 \;, \label{eqn:plugin:taylor:var:first:bound}
\end{align}
where in the inequality we have recalled that $2 c_1 \coloneqq \|  \mean \Var[\phi_{11} \bX_1 | \bX_1 ] \| = \| \Sigma_{11}- \Sigma_{12} \|$ by Lemma \ref{lem:compare_variance}. Next, we bound the quantity
$$  n \| \Var[f(\Phi \cX)] - \Var[f^T(\Phi\cX)] \|\;, $$
for which we use a second-order Taylor expansion on each coordinate of the covariance matrix. For every $s\leq q$, let $f_s(\bx_{11:nk})$ and $g_s(\bx_{11:nk})$ be the $s^{\rm th}$ coordinate of $f(\bx_{11:nk})$ and $g(\bx_{11:nk})$ respectively, i.e. $f_s, g_s$ are both functions $\cD \rightarrow \R$. Then there exists $\tilde \bX^{(s)} \in \big[\bzero, \bar \bX\, \big]$ such that
\begin{equation}
     f_s(\Phi\cX) \;=\; g_s(\mu) + (\partial g_s (\mu))^\top \bar \bX
     +
     \Tr \big(  (\partial^2 g_s (\mu + \tilde \bX^{(s)} ))^\top \,\bar \bX \bar{\bX}^\top \big) \;. \label{eqn:plugin:taylor:each_coordinate}
\end{equation}
Denote for convenience
\begin{align*}
    &
    \bR^1_s \;=\;  (\partial g_s (\mu))^\top \bar \bX \;,
    &&
    \bR^2_s \;=\;\Tr \big(  (\partial^2 g_s (\mu + \tilde \bX^{(s)} ))^\top \,\bar \bX \bar{\bX}^\top \big) \;,
\end{align*}
The Taylor expansion above allows us to control the difference in variance at $(r,s)$-th coordinate:
\begin{align}
    n \big(\Var[f(\Phi \cX)] - \Var\big[ f^T(\Phi\cX) &\big]\big)_{r,s} 
    \;=\;n \big( (\Var[f(\Phi \cX)])_{r,s} -  \big(\Var\big[ g(\mu) + \partial g (\mu) \bar\bX \big]\big)_{r,s}  \big)
    \nonumber \\
    \;=&\; 
    n \big( \Cov[ f_r(\Phi\cX), f_s(\Phi\cX) ] - \Cov\big[ g_r(\mu) + \bR^1_r, \; g_s(\mu) + \bR^1_s  \big] \big)
    \nonumber \\
    \;\overset{(a)}{=}&\;
    n \big( \Cov\big[ \bR^1_r + \bR^2_r,\; \bR^1_s + \bR^2_s \big] - \Cov\big[ \bR^1_r, \; \bR^1_s  \big] \big)
    \nonumber \\
    \;=&\; n \big( \Cov[ \bR^1_r, \bR^2_s] + \Cov[\bR^2_r, \bR^1_s] + \Cov[\bR^2_r, \bR^2_s]  \big)\;. \label{eqn:plugin:taylor:variance:bound:intermediate}
\end{align}
In $(a)$, we have used \eqref{eqn:plugin:taylor:each_coordinate} and the fact that $g_r(\mu)$ and $g_s(\mu)$ are deterministic. To control the first covariance term, by noting that $\mean[\bar\bX]=\bzero$, Cauchy-Schwarz and H\"older's inequality, we get
\begin{align*}
    \Cov[ \bR^1_r, \bR^2_s] \;=&\;
    \mean[ \bR^1_r \bR^2_s] \;=\; \mean\big[  (\partial g_r (\mu))^\top \bar \bX \; \Tr \big(  (\partial^2 g_s (\mu + \tilde \bX^{(s)} ))^\top \,\bar \bX \bar{\bX}^\top \big) \big]
    \\
    \;\leq&\;  \| \partial g_r (\mu) \| \mean\big[ \| \partial^2 g_s (\mu + \tilde \bX^{(s)} ) \| \|\bar\bX\|^3 \big]
    \\
    \;\leq&\; \| \partial g_r (\mu) \| \; \big\| \| \partial^2 g_s (\mu + \tilde \bX^{(s)} ) \| \big\|_{L_4} \; \big\| \|\bar\bX\| \big\|_{L_4}^3
    \\
    \;\overset{(b)}{=}&\; O\big(n^{-3/2} \kappa_{1;1}(g_r) \kappa_{2;4}(g_s) \, \bar c_4^3  \big) \;.
\end{align*}
In $(b)$, we have used the definition of $\kappa^r_m$ and the bound on moments of $\bar \bX$ computed in \eqref{eqn:plugin:rosenthal}. An analogous argument gives
\begin{align*}
    \Cov[ \bR^1_r, \bR^2_s] \;=\;  O\big(n^{-3/2} \kappa_{1;1}(g_s) \kappa_{2;4}(g_r) \, \bar c_4^3  \big) \;,
\end{align*}
and also
\begin{align*}
    \Cov[\bR^2_r, \bR^2_s] 
    \;\leq&\; 
    \big| \mean\big[  \Tr \big(  (\partial^2 g_r (\mu + \tilde \bX^{(r)} ))^\top \,\bar \bX \bar{\bX}^\top \big) \;  \Tr \big(  (\partial^2 g_s (\mu + \tilde \bX^{(s)} ))^\top \,\bar \bX \bar{\bX}^\top \big) \big] \big| 
    \\
    &\;+ 
    \big| \mean\big[  \Tr \big(  (\partial^2 g_r (\mu + \tilde \bX^{(r)} ))^\top \,\bar \bX \bar{\bX}^\top \big) \big] \big| \; \big| \mean \big[ \Tr \big(  (\partial^2 g_s (\mu + \tilde \bX^{(s)} ))^\top \,\bar \bX \bar{\bX}^\top \big) \big] \big|
    \\
    \;\leq&\; 2 \big\| \| \partial^2 g_r (\mu + \tilde \bX^{(r)} ) \| \big\|_{L_6} \; \big\| \| \partial^2 g_s (\mu + \tilde \bX^{(s)} ) \| \big\|_{L_6} \; \big\| \|\bar\bX\| \big\|_{L_6}^4
    \\
    \;=&\; O\big( n^{-2} \kappa_{2;6} (g_r) \kappa_{2;6} (g_s) \, \bar c_6^4 \big) \;.
\end{align*}
Substituting the bounds on each covariance term back into \eqref{eqn:plugin:taylor:variance:bound:intermediate}, we get that
\begin{align*}
    &\;n \big( (\Var[f(\Phi \cX)])_{r,s} -  \big(\Var\big[f^T(\Phi\cX)\big]\big)_{r,s}  \big)
    \\
    \;=&\;
    O\big( n^{-1/2} ( \kappa_{1;1}(g_r) \kappa_{2;4}(g_s) + \kappa_{1;1}(g_s) \kappa_{2;4}(g_r) ) \bar c_4^3 + n^{-1} \kappa_{2;6} (g_r) \kappa_{2;6} (g_s) \, \bar c_6^4 \big)\;.
\end{align*}
Note that by the definition of $\kappa_{R;m}$ in Lemma \ref{lem:plugin:formal},
\begin{align*}
    \msum_{r,s=1}^q \kappa_{R_1;m_1}(g_r) \kappa_{R_2;m_2}(g_s) \;=\; \kappa_{R_1;m_1}(g)\, \kappa_{R_2;m_2}(g)\;,
\end{align*}
so summing the bound above over $r, s \leq q$ gives the bound,
\begin{align*}
    n \big\| \Var[f(\Phi \cX)] - \Var\big[f^T(\Phi \cX) \big] \big\| \;=\; O\big( n^{-1/2} \kappa_{1;1}(g) \kappa_{2;4}(g) \bar c_4^3 + n^{-1} \kappa_{2;6} (g)^2 \bar c_6^4 \big)\;.
\end{align*}
Combine this with the bound from \eqref{eqn:plugin:taylor:var:first:bound} and omitting $g$-dependence gives
\begin{align*}
     n \big\| \Var[f(\Phi \cX)] - \Var\big[f^T(\cZ^{\delta}) \big] \big\|
     = O\big(  \delta k^{-1} \| \partial g(\mu) \|_2^2 \, c_1^2  +  n^{-1/2} \kappa_{1;1} \kappa_{2;4} \bar c_4^3 + n^{-1} \kappa_{2;6}^2 \bar c_6^4 \big)\;.
\end{align*}
\end{proof}

\vspace{.3em}

For Lemma \ref{lem:plugin:formal}(ii), we only need to rewrite the noise stability terms $\alpha_{r;m}(f)$ in Lemma \ref{lem:variance:convergence:general} and \ref{lem:d_H:convergence:general} in terms of $\nu_{r;m}(g)$.

\begin{proof}[Proof of Lemma \ref{lem:plugin:formal}(ii)] We just need to compute the bounds in Lemma \ref{lem:variance:convergence:general} (concerning variance) and Lemma \ref{lem:d_H:convergence:general} (concerning $d_{\cH}$) in terms of $\nu_{r;m}(g)$, which boils down to rewriting $\alpha_{r;m}(f)$ in terms of $\nu_{r;m}(g)$. As usual, we start with computing partial derivatives of $f_s(\bx_{11:nk}) = g\big(\frac{1}{nk} \sum_{i \leq n, j \leq k} \bx_{ij}\big)$:
\begin{align*}
    \mfrac{\partial}{\partial \bx_{ij}} f_s(\bx_{11:nk}) \;&=\; \mfrac{1}{nk} \partial g_s\big( \mfrac{1}{nk} \msum_{i=1}^n \msum_{j=1}^k \bx_{ij} \big) \;, 
    \\
    \mfrac{\partial^2}{\partial \bx_{ij_1} \partial x_{ij_2}} \tilde f_s(\bx_{11:nk})  \;&=\; \mfrac{1}{n^2 k^2} \partial^2 g_s\big( \mfrac{1}{nk} \msum_{i=1}^n \msum_{j=1}^k \bx_{ij} \big) 
    \;, 
    \\
    \mfrac{\partial^3}{\partial \bx_{ij_1} \partial \bx_{ij_2} \partial \bx_{ij_3} } \tilde f_s(\bx_{11:nk})  \;&=\; \mfrac{1}{n^3 k^3} \partial^3 g_s\big( \mfrac{1}{nk} \msum_{i=1}^n \msum_{j=1}^k \bx_{ij} \big)
    \;.
\end{align*}
Norm of the first partial derivative is given by
\begin{align*}
    \|D_i f_s(\bx_{11:nk})\|
    \;=\;
    \sqrt{ \msum_{j=1}^k \Big\| \mfrac{\partial}{\partial \bx_{ij}} f_s(\bx_{11:nk}) \Big\|^2}
    \;=&\;
    \mfrac{1}{n k^{1/2}} \Big\| \partial g_s\Big( \mfrac{1}{nk} \msum_{i=1}^n \msum_{j=1}^k \bx_{ij} \Big)  \Big\|\;,
\end{align*}
and therefore, by the definitions of $\alpha_{r;m}$ from Theorem \ref{thm:main:general} and $\nu_{r;m}$ from \eqref{eqn:defn:beta_r_m},
\begin{align*}
\alpha_{1;m}(f)
  \;\coloneqq&\;
  \msum_{s \leq q} \mmax_{i \leq n} \zeta_{i;m} \big( |D_i f_s(\bW_i(\argdot))| \big)
  \\
  \;=&\; 
  \mfrac{1}{n k^{1/2}}
  \msum_{s \leq q} \mmax_{i \leq n} \zeta_{i;m} \big( |D_i g_s(\overline\bW_i(\argdot)| \big)
  \;=\; \mfrac{1}{n k^{1/2}} \nu_{1;m}(g)\;.
\end{align*} 
Similarly we get $\alpha_{2;m}(f) = \mfrac{1}{n^2 k} \nu_{2;m}(g)$, $\alpha_{3;m}(f) = \mfrac{1}{n^3 k^{3/2}} \nu_{3;m}(g)$ and $\alpha_{0;m}(f) = \nu_{0;m}(g)$.The bound in Lemma \ref{lem:d_H:convergence:general} can then be computed as
\begin{align*}
    &\;
    \delta n^{3/2}k^{1/2} c_1 \big( n^{1/2} \alpha_{1;2}^2 + \alpha_{2;1} \big)
    +
    (nk)^{3/2} ( n \alpha_{1;6}^3  + 3 n^{1/2} \alpha_{1;4} \alpha_{2;4} + \alpha_{3;2} )(c_X+c_{Z^{\delta}})\;.
    \\
    =&\; \delta \big( k^{-1/2} \nu_{1;2}^2 + n^{-1/2}k^{-1/2} \nu_{2;1} \big) c_1
    +
    \big(n^{-1/2} \nu_{1;6}^3  + 3 n^{-1} \nu_{1;4} \nu_{2;4} + n^{-3/2} \nu_{3;2} \big) (c_X + c_{Z^{\delta}}) \;,
\end{align*}
while the bound in Lemma \ref{lem:variance:convergence:general} can be computed as
\begin{align*}
    &\; 
    4 \delta n^2 k^{1/2} ( \alpha_{0;4} \alpha_{2;4} + \alpha_{1;4}^2 ) c_1
    + 6 n^2 k^{3/2} ( \alpha_{0;4} \alpha_{3;4} + \alpha_{1;4} \alpha_{2;4} ) (c_X + c_{Z^{\delta}})
    \\
    =&\;
    O\big( \delta k^{-1/2} ( \nu_{0;4} \nu_{2;4} + \nu_{1;4}^2 ) c_1
    +
     n^{-1} ( \nu_{0;4} \nu_{3;4} + \nu_{1;4} \nu_{2;4} ) (c_X +  c_{Z^{\delta}}) \big)
    \;.
\end{align*}
These give the desired bounds on $d_{\cH}(\sqrt{n} f(\Phi\cX), \sqrt{n} f(\cZ^{\delta}))$ and $n \| \Var[f(\Phi\cX)] - \Var[f(\cZ^{\delta})] \|$.
\end{proof} 

\vspace{-.2em}

\subsection{Proofs for \cref{appendix:repeated}}

\vspace{.2em}

In this section, we first prove \cref{example:repeated_toy}, a toy example showing how repeated augmentation adds additional complexity, and then prove \ref{thm:repeated}, the main result concerning repeated augmentation.

\begin{proof}[Proof of \cref{example:repeated_toy}]
By the invariance $\phi_1 \bX_1 \overset{d}{=}\bX_1$ and the fact that $\bX_1$, $\bX_2$, $\phi_1$ and $\phi_2$ are independent, we get that
\begin{align*}
\Var f_1(\bX_1, \bX_2) 
&\;=\;
\Var f_1(\phi_1 \bX_1, \phi_2 \bX_2)\;,
&&
\Var f_2(\bX_1, \bX_2) 
\;=\; 
\Var f_2(\phi_1 \bX_1, \phi_2 \bX_2)\;.
\end{align*}
For repeated augmentation, notice that for any $\bv \in \R^d$,
\begin{align*}
\bv^\top \Var f_1(\phi_1 \bX_1, \phi_1 \bX_2) \bv
\;=&\;
\bv^\top \Var[\phi_1 \bX_1 + \phi_1 \bX_2 ] \bv
\\
\;=&\;
\bv^\top \Var[ \phi_1 \bX_1 ] \bv + \bv^\top \Var[ \phi_1 \bX_2 ] \bv + 2 \bv^\top \Cov[ \phi_1 \bX_1, \phi_1\bX_2 ] \bv 
\\
\;=&\;
2 \bv^\top \Var [\bX_1]  \bv
+
2 \bv^\top \Cov [\phi_1 \bX_1, \phi_1 \bX_2] \bv
\\
\;=&\; \bv^\top \Var [\phi_1 \bX_1 + \phi_2\bX_2] \bv + 
2 \bv^\top \Cov [\phi_1 \bX_1, \phi_1 \bX_2] \bv  
\\
\;=&\;
\bv^\top \Var f_1(\phi_1 \bX_1, \phi_2\bX_2) \bv + 
2 \bv^\top \Cov [\phi_1 \bX_1, \phi_1 \bX_2] \bv  
\;,
\end{align*}
and similarly
\begin{align*}
\bv^\top \Var f_2(\phi_1 \bX_1, \phi_1 \bX_2) \bv
\;=&\;
\bv^\top \Var f_2(\phi_1 \bX_1, \phi_2\bX_2) \bv
-
2 \bv^\top \Cov [\phi_1 \bX_1, \phi_1 \bX_2] \bv\;.
\end{align*}
Now note that for all $\bv \in \R^d$,
\begin{align*}
\bv^\top \Cov [\phi_1 \bX_1, \phi_1 \bX_2] \bv
\;=&\; 
\mean\big[ (\bX_1^\top \phi_1^\top \bv)^\top (\bX_2^\top \phi_1^\top \bv)  \big] - \mean\big[ \bX_1^\top \phi_1^\top \bv \big]^\top \mean \big[ \bX_2^\top \phi_1^\top \bv \big]
\\
\;=&\;
\mean\big[ (\mu^\top \phi_1^\top \bv)^\top (\mu^\top \phi_1^\top \bv)  \big] - \mean\big[ \mu^\top \phi_1^\top \bv\big]^\top \mean \big[ \mu^\top \phi_1^\top \bv \big]
\\
\;=&\; \Var[ \bv^\top \phi_1 \mu ] \geq 0\;,
\end{align*}
and therefore for all $\bv \in \R^d$,
\begin{align*}
    \bv^\top \Var f_1(\phi \bX_1, \phi_1 \bX_2) \bv \;&\geq\;  \bv^\top \Var f_1(\phi_1 \bX_1, \phi_2\bX_2) \bv\;,
    \\
    \bv^\top \Var f_2(\phi \bX_1, \phi_1 \bX_2) \bv \;&\leq\;  \bv^\top \Var f_2(\phi_1 \bX_1, \phi_2\bX_2) \bv\;,
\end{align*}
which completes the proof.

\end{proof}

The broad stroke idea in proving Theorem \ref{thm:repeated} for repeated augmentation is similar to that of our main result, Theorem \ref{thm:main}, and we refer readers to \cref{appendix:proof:main} for a proof overview. The only difference is that in proving Theorem \ref{thm:main}, we can group data into independent blocks due to i.i.d.\,augmentations being used for different data points. In the proof of Theorem \ref{thm:repeated}, the strategy must be modified: The additional dependence introduced by reusing transformations means moments can no longer be factored off from derivatives, so stronger assumptions on the derivatives are required to control terms. This is achieved by using the symmetry assumption on $f$ from \eqref{eq:permutation:invariance}.

\ 

\begin{proof}[Proof of Theorem \ref{thm:repeated}] Similar to the proof for Theorem \ref{thm:main:general} (a generalized version of Theorem \ref{thm:main}), we abbreviate $g=h \circ f$ and denote
\begin{equation*}
  \bV_i(\argdot) 
  \;\coloneqq\;
  (\tilde{\Phi}_1\bX_1,\ldots,\tilde{\Phi}_{i-1}\bX_{i-1},\argdot,\bY_{i+1},\ldots,\bY_n)\;.
\end{equation*}
The same telescoping sum and Taylor expansion argument follows, yielding
\begin{align}
\big| \mean h ( f( \tilde{\Phi} \cX) ) - \mean h ( f( \bY_1, \ldots, \bY_n) ) \big| 
&\;=\; 
\big|\mean\msum_{i=1}^n 
 \big[g( \bV_i( \tilde{\Phi}_i \bX_i))  -  g( \bV_i( \bY_i) ) \big] 
\big| \nonumber
\\&\le \msum_{i=1}^n \big|\mean
 \big[g( \bV_i( \tilde{\Phi}_i \bX_i))  -  g( \bV_i( \bY_i) ) \big] 
\big|,  \label{eqn:thm_repeated_telescoping_sum}
\end{align}
and each summand is bounded above as
\begin{equation*}
    \big|\mean\big[g( \bV_i(\tilde{\Phi}_i \bX_i))  -  g( \bV_i( \bY_i) ) \big]\big|
    \;\leq\;
    |\tau_{1,i}| + \mfrac{1}{2} |\tau_{2,i}| + \mfrac{1}{6} | \tau_{3,i}|\;,
\end{equation*}
where
\begin{align*}
&\tau_{1,i} 
\coloneqq 
\mean \big[
\big(D_i g(\bV_i (\bzero)) \big) 
\big( \tilde{\Phi}_i \bX_i - \bY_i \big) 
\big] \\
&\tau_{2,i} 
\coloneqq 
\mean \big[
\big(
D_i^2 g(\bV_i (\bzero)) 
\big) 
\big( 
(\tilde{\Phi}_i \bX_i)(\tilde{\Phi}_i \bX_i)^\top - \bY_i \bY_i^\top
\big) 
\big] \\
&\tau_{3,i} 
\coloneqq 
\mean \big[ 
\| \tilde{\Phi}_i \bX_i  \|^3
\sup_{\bw \in [\bzero, \tilde{\Phi}_i \bX_i] }
\big\| 
D_i^3 g \big(\bV_i (\bw) \big) 
\big\| 
+ 
\| \bY_i  \|^3
\sup_{\bw \in [\bzero, \bY_i] }
\big\| 
D_i^3 g \big(\bV_i (\bw) \big) 
\big\|
\big]\;. 
\end{align*}
With a slight abuse of notation, we view $D_i g(\bV_i (\bzero))$ as a function $\R^{dk} \rightarrow \R$ and $D_i^2 g(\bV_i (\bzero))$ as a function $\R^{dk \times dk} \rightarrow \R$. Substituting into \eqref{eqn:thm_repeated_telescoping_sum}, and applying the triangle inequality, shows
\begin{equation*}
\big| \mean h ( f( \tilde{\Phi} \cX) ) - \mean h ( f( \bY_1, \ldots, \bY_n) ) \big| 
\;\leq\;
\msum_{i=1}^n \big(|\tau_{1,i}| + \mfrac{1}{2} |\tau_{2,i}| + \mfrac{1}{6} | \tau_{3,i}|\bigr)\;.
\end{equation*}
The next step is to bound the terms $\tau_{1,i}$, $\tau_{2,i}$ and $\tau_{3,i}$. $\tau_{3,i}$ is  analogous to $\kappa_{3, i}$ in the proof of Theorem \ref{thm:main}. Define
\begin{equation*}
M_i 
\coloneqq 
\max \braces{ 
\big\|
{\textstyle\sup_{\bw \in [\bzero, \tilde{\Phi}_i \bX_i] }}
\| D_i^3 g \big(\bV_i (\bw) \big) \|\big\|_{L_2} 
,\,
\big\|
{\textstyle\sup_{\bw \in [\bzero, \bY_i] }}
\| D_i^3 g \big(\bV_i (\bw) \big) \| \big\|_{L_2} 
}\;,
\end{equation*}
we can handle $\tau_{3,i}$ in the exact same way as in Theorem \ref{thm:main} to obtain
$$\mfrac{1}{6} |\tau_{3,i}| \;\leq\; k^{3/2} (c_X + c_Y) M_i.$$
However, bounding $\tau_{1,i}$ and $\tau_{2,i}$ works differently, since $(\tilde{\Phi}_i \bX_i, \bY_i)$ is no longer independent of ${(\tilde{\Phi}_j \bX_j, \bY_j)_{j \neq i}}$ and therefore not independent of $\bV_i(\bzero)$. To this end, we invoke the permutation invariance assumption \eqref{eq:permutation:invariance} on $f$, which implies the function $g(\bV_i(\argdot)) = h( f(\bV_i(\argdot)) )$ that takes input in $\R^{kd}$ satisfies \eqref{eq:permutation:invariance:partial} in Lemma \ref{lem:perm_inv_implies_same_derivative}. Then Lemma  \ref{lem:perm_inv_implies_same_derivative} shows that, for each $i \leq n$ and for $\bx_{i1}, \ldots, \bx_{ik} \in \R^d$,
\begin{align}
   \mfrac{\partial}{\partial \bx_{i1}} g(\bV_i(\bzero)) &\;=\; \ldots \;=\; \mfrac{\partial}{\partial \bx_{ik}} g(\bV_i(\bzero)), \label{eq:repeated_aug:same_first_derivative}
   \\
    \mfrac{\partial^2}{\partial \bx_{i1}^2} g(\bV_i(\bzero)) &\;=\; \ldots \;=\; \mfrac{\partial^2}{\partial \bx_{ik}^2} g(\bV_i(\bzero)), \label{eq:repeated_aug:same_second_derivative}
    \\
    \mfrac{\partial^2}{\partial \bx_{ir} \partial \bx_{is}} g(\bV_i(\bzero)) & \text{ is the same for all } r \neq s, 1 \leq r, s \leq k.  \label{eq:repeated_aug:same_mixed_derivative}
\end{align}
Consider bounding $\tau_{1,i}$. Rewrite $\tau_{1,i}$ as a sum of $k$ terms and denote $\bY_{ij} \in \R^d$ as $Y_{ij1:ijd}$, the subvector of $\bY_i$ analogous to $\phi_j \bX_i$ in $\tilde \Phi_i\bX_i$. Since that \eqref{eq:repeated_aug:same_first_derivative} allows $\mfrac{\partial}{\partial \bx_{i1}} g(\bV_i(\bzero))$ to be taken outside the summation in $(a)$ below, we get
\begin{align*}
|\tau_{1,i}|
\;&=\; 
\mean \big[ \msum_{j=1}^k \mfrac{\partial}{\partial \bx_{ij}} g(\bV_i(\bzero)) \big( \phi_j \bX_i - \bY_{ij} \big) \big] \\
 &\;\overset{(a)}{=}\;
\mean \big[ \mfrac{\partial}{\partial \bx_{i1}} g(\bV_i(\bzero))  \msum_{j=1}^k \big( \phi_j \bX_i - \bY_{ij} \big) \big]  \\
 &\;\overset{(b)}{=}\;
\mean \Big[ 
\mean \big[ \mfrac{\partial}{\partial \bx_{i1}} g(\bV_i(\bzero)) \big| \tilde{\Phi}, \Psi \big] 
\mean \big[ \msum_{j=1}^k \big( \phi_j \bX_i - \bY_{ij} \big) \big| \tilde{\Phi}, \Psi \big] 
\Big] \\
 &\;\leq\;
\mean \big[ 
\big\|  \mean \big[  \mfrac{\partial}{\partial \bx_{i1}} g(\bV_i(\bzero)) \big| \tilde{\Phi}, \Psi \big] \big\| \; 
\big\|  \mean \big[  \msum_{j=1}^k \big( \phi_j \bX_i - \bY_{ij} \big) \big| \tilde{\Phi}, \Psi \big]  \big\| \big] \\
 &\;\leq\; \Big\| \; \Big\| \mean \big[ \mfrac{\partial}{\partial \bx_{i1}} g(\bV_i(\bzero)) \big| \tilde{\Phi}, \Psi \big] \Big\| \; \Big\|_{L_2} \; \Big\| \; \Big\| \mean \big[ \msum_{j=1}^k \big( \phi_j \bX_i - \bY_{ij} \big) \big| \tilde{\Phi}, \Psi \big] \Big\| \; \Big\|_{L_2}
 \\
 &=:\; (t1i) \, (t2i)\;.
\end{align*} 
where to get $(b)$, we apply conditional independence conditioning on $\Phi$ and $\Psi$, the augmentations for $\cX$ and $\bY_1,\ldots,\bY_n$ respectively, and to obtain the final bound we exploited Cauchy-Schwarz inequality. We will first upper bound $(t2i)$ by the trace of the variance of the augmented $(X_i).$ Moving the summation outside the expectation,
\begin{align}
(t2i)=\Big\| \; \Big\| &\mean \big[ \msum_{j=1}^k \big( \phi_j \bX_i - \bY_{ij} \big) \big| \tilde{\Phi}, \Psi \big] \Big\| \; \Big\|_{L_2} \nonumber \\
\;&=\; 
\sqrt{ \mean \Big[ \mean \Big[\msum_{j_1=1}^k \big( \phi_{j_1} \bX_i - \bY_{ij_1} \big) \Big| \tilde{\Phi}, \Psi \Big]^\top \mean \Big[\msum_{j_2=1}^k \big( \phi_{j_2} \bX_i - \bY_{ij_2} \big) \Big| \tilde{\Phi}, \Psi \Big] \Big]} \nonumber \\
 \;&=\;
\sqrt{ \msum_{j_1, j_2=1}^k \mean \Big[ \mean \big[ \phi_{j_1} \bX_i - \bY_{ij_1} \big| \phi_{j_1}, \psi_{j_1} \big]^\top \mean \big[ \phi_{j_2} \bX_i - \bY_{ij_2} \big| \phi_{j_2}, \psi_{j_2} \big] \Big] }. \label{eq:thm_repeated:first_moment}
\end{align}
In each summand, the expectation is taken over a product of two quantities, which are respectively functions of $\{ \phi_{j_1}, \psi_{j_1} \}$ and $\{ \phi_{j_2}, \psi_{j_2} \}$. For $j_1 \neq j_2$, the two quantities are independent, and are also zero-mean since
\begin{align*}
    \mean \Big[ \mean \big[ \phi_{j} \bX_i - (\bY_i)_{j} \big| \phi_{j}, \psi_{j} \big] \Big] 
    \;&=\; \mean \big[ \mean[ \phi_{j} \bX_i | \phi_{j}] \big] - \mean \big[ \mean[ \bY_{ij} | \psi_{j} \big] \big] \\
    \;&=\; \mean \big[ \mean[ \phi_{j} \bX_i | \phi_{j}] \big] - \mean \big[ \mean[ \psi_{j} \bX_1 | \psi_{j} \big] \big] = \bzero.
\end{align*}
Therefore, summands with $j_1 \neq j_2$ vanish, and \eqref{eq:thm_repeated:first_moment} becomes
\begin{align*}
(t2i)\;&=\;\Big\| \; \Big\| \mean \big[ \msum_{j=1}^k \big( \phi_j \bX_i - \bY_{ij} \big) \big| \tilde{\Phi}, \Psi \big] \Big\| \; \Big\|_{L_2}  \\
 \;&=\;
\sqrt{ \msum_{j=1}^k \mean \Big[ \mean \big[ \phi_{j} \bX_i - \bY_{ij} \big| \phi_j, \psi_j \big]^\top \mean \big[ \phi_j \bX_i - \bY_{ij} \big| \phi_j, \psi_j \big] \Big] } \\
 \;&\overset{(c)}{=}\;
 \sqrt{k} \sqrt{\mean \big[ \mean \big[ \phi_1 \bX_i - \bY_{i1} \big| \phi_1, \psi_1 \big]^\top \mean \big[ \phi_1 \bX_i - \bY_{i1} \big| \phi_1, \psi_1 \big] \big] } \\
 \;&=\;
 \sqrt{k} \sqrt{\Tr \Var \big[ \mean [ \phi_1 \bX_i | \phi_1 ] - \mean [ \bY_{i1} | \psi_1 ] \big] } \\
 \;&\overset{(d)}{=}\;
 \sqrt{k} \sqrt{\Tr \Var \big[ \mean [ \phi_1 \bX_1 | \phi_1 ] - \mean [ \psi_1 \bX_1 | \psi_1 ] \big] } \\
 \;&\overset{(e)}{=}\;
 \sqrt{2k} \sqrt{\Tr \Var \big[ \mean [ \phi_1 \bX_1 | \phi_1 ] \big] } \;=\; \sqrt{k} m_1.
\end{align*}
where we have used that $(\phi_1, \psi_1), \ldots, (\phi_k, \psi_k)$ are i.i.d.\,in $(c)$ and that $\mean [ \phi_1 \bX_1 | \phi_1 ]$ and $\mean [ \psi_1 \bX_1 | \psi_1 ]$ are i.i.d.\,in $(d)$ and $(e)$. Define
$$C_i \coloneqq \Big\| \; \Big\| \mean \big[ \mfrac{\partial}{\partial x_{i11:i1d}} g(\bV_i(\bzero)) \big| \tilde{\Phi}, \Psi \big] \Big\| \; \Big\|_{L_2},$$ we note that $(t1i)\le C_i.$ Therefore 
we obtain
$$ |\tau_{1,i}| \;\leq\; \sqrt{k} m_1 C_i\;. $$
$\tau_{2,i}$ can be bounded similarly by rewriting as a sum of $k^2$ terms and making use of conditional independence. We defer the detailed computation to Lemma \ref{lem:thm_repeated:second_term_bound}. Define 
\begin{align}
E_i &\coloneqq  \Big\| \; \big\| \mean \big[
\mfrac{\partial^2}{\partial \bx_{i1}^2} g(\bV_i(\bzero)) \big| \tilde \Phi, \Psi \big] \big\| \Big\|_{L_2}, &&
F_i \coloneqq \Big\| \; \big\| \mean \big[
\mfrac{\partial^2}{\partial \bx_{i1} \partial \bx_{i2}} g(\bV_i(\bzero)) \big| \tilde \Phi, \Psi \big] \big\| \Big\|_{L_2}. \label{eqn:thm_repeated:Ei_Fi}
\end{align}
Lemma \ref{lem:thm_repeated:second_term_bound} below shows that
\begin{equation}
   \mfrac{1}{2} | \tau_{2,i}| \;\leq\; k^{1/2} m_2 E_i + k^{3/2} m_3 F_i. \label{eq:thm_repeated:second_term_bound}
\end{equation}
In summary, the right hand side of \eqref{eqn:thm_repeated_telescoping_sum} is hence bounded by
\begin{align*}
\eqref{eqn:thm_repeated_telescoping_sum} &\leq \; \msum_{i=1}^n | \tau_{1,i} | + \mfrac{1}{2}  | \tau_{2,i} | + \mfrac{1}{6} | \tau_{1,i} | \\
&\leq n k^{-1/2} m_1 \max_{i \leq n} C_i + k^{1/2} m_2 \max_{i \leq n} E_i + k^{3/2} m_3 \max_{i \leq n} F_i + nk^{3/2}(c_2+c_3) \max_{i \leq n} M_i.
\end{align*}
Lemma \ref{lem:thm_repeated_chain_rule} below shows that the the maximums $\max_{i\le n} E_i, \max_{i\le n} C_i, \max_{i\le n} D_i$ $\max_{i\le n} M_i$ are in turn bounded by
\begin{align} 
&\mmax_{i \leq n} \; C_i
\;\leq\; 
  k^{-1/2} \gamma_1(h) \alpha_1, \label{bound:repeated_thm_first_order} 
\\
&\mmax_{i \leq n} \; E_i
\;\leq\; k^{-1/2}
( \gamma_2(h) \alpha_1^2 
    + 
    \gamma_1(h) \alpha_2), \label{bound:repeated_thm_second_order_1} 
\\
&\mmax_{i \leq n} \; F_i
\;\leq\; k^{-3/2}
( \gamma_2(h) \alpha_1^2 
    + 
    \gamma_1(h) \alpha_2), \label{bound:repeated_thm_second_order_2} 
\\
&\mmax_{i \leq n} \; M_i 
\;\leq\;  
\lambda(n,k). \label{bound:repeated_thm_third_order} 
\end{align}
That yields the desired upper bound on \eqref{eqn:thm_repeated_telescoping_sum},
\begin{align*}
  \big| \mean h ( f( \tilde \Phi \cX) ) &- \mean h ( f( \bY_1, \ldots, \bY_n)
) \big| \\
&\leq 
  n \gamma_1(h) \alpha_1 m_1
  +
  n \omega_2(n,k)( \gamma_2(h) \alpha_1^2 
    + 
    \gamma_1(h) \alpha_2)
  +
  nk^{3/2}\lambda(n,k)(c_X+c_Y)\;.
\end{align*}
which finishes the proof.
\end{proof}

We complete the computation of bounds in Lemma \ref{lem:thm_repeated:second_term_bound} and Lemma \ref{lem:thm_repeated_chain_rule}.

\begin{lemma}\label{lem:thm_repeated:second_term_bound} The bound on $|\tau_{2,i}|$ in \eqref{eq:thm_repeated:second_term_bound} holds.
\end{lemma} 

\begin{proof}
Rewrite $\tau_{2,i}$ as a sum of $k^2$ terms,
\begin{align}
|\tau_{2,i}| 
&= \mean \Big[ 
\msum_{j_1, j_2=1}^k \mfrac{\partial^2}{\partial \bx_{ij_1} \partial \bx_{ij_2}} g(\bV_i(\bzero)) \big( (\phi_{j_1} \bX_i)(\phi_{j_2} \bX_i)^\top - (\bY_{ij_1})(\bY_{ij_2})^\top \big) 
\Big]. \label{eq:thm_repeated:second_term}
\end{align}
Consider the terms with $j_1 = j_2$. \eqref{eq:repeated_aug:same_second_derivative} says that the derivatives are the same for $j_1=1, \ldots, k$ and allows $\mfrac{\partial^2}{\partial x_{ij1:ijd}^2} g(\bV_i(\bzero))$ to be taken out of the following sum,
\begin{align}
& \mean\Big[
\msum_{j=1}^k \mfrac{\partial^2}{\partial \bx_{ij}^2} g(\bV_i(\bzero)) \big( (\phi_j \bX_i)(\phi_j \bX_i)^\top - (\bY_{ij})(\bY_{ij})^\top \big) 
\Big] 
\nonumber \\
&=\; \mean \Big[
\mfrac{\partial^2}{\partial \bx_{i1}^2} g(\bV_i(\bzero)) \msum_{j=1}^k \big( (\phi_j \bX_i)(\phi_j \bX_i)^\top - (\bY_{ij})(\bY_{ij})^\top \big) 
\Big] 
\nonumber \\
&\overset{(a)}{=}\;  \mean \Big[ \mean \big[
\mfrac{\partial^2}{\partial \bx_{i1}^2} g(\bV_i(\bzero)) \big| \tilde \Phi, \Psi \big] \mean \big[ \msum_{j=1}^k \big( (\phi_j \bX_i)(\phi_j \bX_i)^\top - (\bY_{ij})(\bY_{ij})^\top \big) \big| \tilde \Phi, \Psi \big]
\Big] 
\nonumber \\
&\leq\;  \Big\| \; \big\| \mean \big[
\mfrac{\partial^2}{\partial \bx_{i1}^2} g(\bV_i(\bzero)) \big| \tilde \Phi, \Psi \big] \big\| \Big\|_{L_2 }\; \big\| \msum_{j=1}^k \| \bT_{jj} \| \big\|_{L_2} 
\nonumber \\
&= E_i \, \big\| \msum_{j=1}^k \| \bT_{jj} \| \big\|_{L_2},
\label{eqn:thm_repeated:second_term:same_j}
\end{align}
where we have used conditional independence conditioning on $\tilde \Phi$ and $\Psi$ in (a), defined $E_i$ as in \eqref{eqn:thm_repeated:Ei_Fi} and denoted 
\begin{align*}
    \bT_{j_1j_2} 
    \;&\coloneqq\; \mean \big[ (\phi_{j_1} \bX_i)(\phi_{j_2} \bX_i)^\top - (\bY_{ij_1})(\bY_{ij_2})^\top \big| \tilde \Phi, \Psi \big] \\
    \;&=\; \mean \big[ (\phi_{j_1} \bX_i)(\phi_{j_2} \bX_i)^\top | \phi_{j_1}, \phi_{j_2} \big] - \mean \big[ (\bY_{ij_1})(\bY_{ij_2})^\top \big| \psi_{j_1}, \psi_{j_2} \big] 
    \\
    \;&=\; \mean \big[ (\phi_{j_1} \bX_1)(\phi_{j_2} \bX_1)^\top | \phi_{j_1}, \phi_{j_2} \big] - \mean \big[ (\psi_{j_1} \bX_1 )(\psi_{j_2} \bX_1)^\top \big| \psi_{j_1}, \psi_{j_2} \big]\;.
\end{align*}
Consider the terms in \eqref{eq:thm_repeated:second_term} with $j_1 \neq j_2$. \eqref{eq:repeated_aug:same_mixed_derivative} says that the derivatives are the same for $1 \leq j_1, j_2 \leq k$ with $j_1 \neq j_2$, so by a similar argument,
\begin{align}
&\mean\Big[
\msum_{j_1 \neq j_2} \mfrac{\partial^2}{\partial \bx_{ij_1} \partial \bx_{ij_2}} g(\bV_i(\bzero)) \big( (\phi_{j_1} \bX_i)(\phi_{j_2} \bX_i)^\top - (\bY_{ij_1})(\bY_{ij_2})^\top \big) 
\Big] \nonumber\\
&= \mean \Big[
\mfrac{\partial^2}{\partial \bx_{i1}\partial \bx_{i2}} g(\bV_i(\bzero)) \msum_{j_1 \neq j_2} \big( (\phi_{j_1} \bX_i)(\phi_{j_2} \bX_i)^\top - (\bY_{ij_1})(\bY_{ij_2})^\top \big)
\Big] \nonumber\\
&=  \mean \Big[ \mean \big[
\mfrac{\partial^2}{\partial \bx_{i1} \partial \bx_{i2}} g(\bV_i(\bzero)) \big| \tilde \Phi, \Psi \big] \mean \big[ {\msum_{j_1 \neq j_2}}\! \big( (\phi_{j_1} \bX_i)(\phi_{j_2} \bX_i)^\top \!\!-\! (\bY_{ij_1})(\bY_{ij_2})^\top \big)\! \big| \tilde \Phi, \Psi \big]
\Big] \nonumber\\
&\leq  \Big\| \; \big\| \mean \big[
\mfrac{\partial^2}{\partial x_{i11:i1d} \partial x_{i21:i2d}} g(\bV_i(\bzero)) \big| \tilde \Phi, \Psi \big] \big\| \Big\|_{L_2} \; \big\| \msum_{j_1 \neq j_2} \| \bT_{j_1 j_2} \| \big\|_{L_2} \nonumber\\
&= F_i \, \big\| \msum_{j_1 \neq j_2} \| \bT_{j_1 j_2} \| \big\|_{L_2}\;, \label{eqn:thm_repeated:second_term:diff_j}
\end{align}
where we have used $F_i$ defined in \eqref{eqn:thm_repeated:Ei_Fi}. To obtain a bound for \eqref{eqn:thm_repeated:second_term:same_j}and \eqref{eqn:thm_repeated:second_term:diff_j}, we need to bound $\big\| \sum_{j=1}^k \| \bT_{jj} \| \big\|_{L_2}$ and $\big\| \sum_{j_1 \neq j_2} \| \bT_{j_1j_2} \| \big\|_{L_2}$. To this end, we denote
\begin{align*}
    \bA_{\phi} &\coloneqq \tvec\big( \mean \big[ (\phi_1 \bX_1)(\phi_1 \bX_1)^\top | \phi_1 \big] \big), && \; 
    \bA_{\psi} \coloneqq \tvec\big( \mean \big[ (\psi_1 \bX_1 )(\psi_1 \bX_1)^\top \big| \psi_1 \big] \big), \\
    \bB_{\phi} &\coloneqq \tvec\big( \mean \big[ (\phi_1 \bX_1)(\phi_2 \bX_1)^\top | \phi_1, \phi_2 \big] \big), && \; 
    \bB_{\psi} \coloneqq \tvec\big(\mean \big[ (\psi_1 \bX_1 )(\psi_2 \bX_1)^\top \big| \psi_1, \psi_2 \big]\big),
\end{align*}
where $\tvec(\{M_{rs}\}_{i,j \leq d}) = (M_{11}, M_{12}, \ldots, M_{dd}) \in \R^{d^2}$ converts a matrix to its vector representation. Then WLOG we can write $\bT_{11} = \bA_{\phi} - \bA_{\psi}$, $\bT_{12} = \bB_{\phi} - \bB_{\psi}$. Before we proceed, we compute several useful quantities in terms of $\bT$'s. Recall that
\begin{align*}
m_2
\coloneqq
\sqrt{ \msum_{r,s\leq d} 
\mfrac{\Var \mean\big[(\phi_1 \bX_1)_r (\phi_1 \bX_1)_s \big| \phi_1\big]}{2}},
\;
m_3
\coloneqq
\sqrt{ \msum_{r,s\leq d} 12
\Var \mean\big[(\phi_1 \bX_1)_r (\phi_2 \bX_1)_s \big| \phi_1, \phi_2\big]}\;.
\end{align*}
Since $\bA_{\phi}$ and $\bA_{\psi}$ are i.i.d.,
\begin{align}
\mean\big[ \big\|\bT_{jj} \big\|^2 \big]
&=
\mean\big[ \big\|\bT_{11} \big\|^2 \big] 
= 
\mean\big[ \Tr( \bT_{11} \bT_{11}^\top ) \big] 
= 
\Tr \mean\big[ \bT_{11} \bT_{11}^\top \big] \nonumber \\
&= \Tr ( \mean[\bA_{\phi} \bA_{\phi}^\top] - \mean[\bA_{\phi} \bA_{\psi}^\top] - \mean[\bA_{\psi} \bA_{\phi}^\top] + \mean[\bA_{\psi} \bA_{\psi}^\top] ) \nonumber \\ 
&= 
2 \Tr \big( \mean[\bA_{\phi} \bA_{\phi}^\top] - \mean[\bA_{\phi}] \mean [ \bA_{\phi}]^\top) \nonumber \\
&= 2 \msum_{r, s=1}^d \big( \mean[ (\bA_{\phi})_{rs}^2 ] - \mean[ (\bA_{\phi})_{rs} ]^2 \big) \nonumber \\
&= 2 \msum_{r, s=1}^d \Var \mean\big[ (\phi_1 \bX_1)_r (\phi_1 \bX_1)_s \big| \phi_1 \big] = 4 (m_2)^2.\label{eq:thm_repeated:T_same}
\end{align}
Similarly by noting that $\bB_{\phi}$ and $\bB_{\psi}$ are i.i.d., for $j_1 \neq j_2$,
\begin{align}
\mean\big[ \big\|\bT_{j_1j_2} \big\|^2 \big]
&=
\mean\big[ \big\|\bT_{12} \big\|^2 \big] 
=
\Tr \mean\big[ \bT_{12} \bT_{12}^\top \big] \nonumber \\
&= \Tr ( \mean[\bB_{\phi} \bB_{\phi}^\top] - \mean[\bB_{\phi} \bB_{\psi}^\top] - \mean[\bB_{\psi} \bB_{\phi}^\top] + \mean[\bB_{\psi} \bB_{\psi}^\top] ) \nonumber \\ 
&= 2 \msum_{r, s=1}^d \big( \mean[ (\bB_{\phi})_{rs}^2 ] - \mean[ (\bB_{\phi})_{rs} ]^2 \big) \nonumber \\
&= 2 \msum_{r, s=1}^d \Var \mean\big[ (\phi_1 \bX_1)_r (\phi_2 \bX_1)_s \big| \phi_1 \big] = \mfrac{1}{6} (m_3)^2. \label{eq:thm_repeated:T_diff}
\end{align}
On the other hand, by Cauchy-Schwarz with respect to the Frobenius inner product, for $j_1 \neq j_2$ and $l_1 \neq l_2$,
\begin{align} 
\big| \mean[ \Tr (\bT_{j_1 j_2}\bT_{l_1 l_2}^\top) ]) \big| 
\;&\leq\; 
\Big| \mean \Big[ \sqrt{\Tr ( \bT_{j_1 j_2} \bT_{j_1 j_2}^\top)} \sqrt{ \Tr ( \bT_{l_1 l_2} \bT_{l_1 l_2}^\top )} \Big] \Big| \nonumber \\
&\leq\; \sqrt{ \mean \Tr ( \bT_{j_1 j_2} \bT_{j_1 j_2}^\top) } \sqrt{ \mean \Tr ( \bT_{l_1 l_2} \bT_{l_1 l_2}^\top) } \nonumber \\
&=\; \sqrt{ \mean\big[ \big\|\bT_{j_1j_2} \big\|^2 \big] } \sqrt{ \mean\big[ \big\|\bT_{l_1l_2} \big\|^2 \big] }
\;\leq\; \mfrac{1}{6} (m_3)^2\; , \label{eq:thm_repeated:T_bound}
\end{align}
which can be computed using the above relations for each $j_1, j_2, l_1, l_2 \leq k$. Moreover we note that, since $\mean[\bA_{\phi}] = \mean[\bA_{\psi}]$ and $\mean[\bB_{\phi}] = \mean[\bB_{\psi}]$ this directly implies that $\mean[\bT_{11}]=\mean[\bT_{12}]=0$. We are now ready to bound $\big\| \sum_{j=1}^k \| \bT_{jj} \| \big\|_{L_2}$ and $\big\| \sum_{j_1, j_2=1}^k \| \bT_{j_1j_2} \| \big\|_{L_2}$:
\begin{align*}
    \Big\| \; \big\| \msum_{j=1}^k \bT_{jj} \big\| \; \Big\|_{L_2} 
    &\coloneqq 
    \sqrt{ \mean \Big[ \Tr\big( \big(\msum_{j_1=1}^k \bT_{j_1j_1} \big)  \big(\msum_{j_2=1}^k \bT_{j_2j_2} \big)^\top \big) \Big]  } \\
    &= 
    \sqrt{ \msum_{j_1, j_2=1}^k \Tr \mean[ \bT_{j_1 j_1}\bT_{j_2 j_2}^\top ]  } 
    \overset{(a)}{=} 
    \sqrt{ \msum_{j=1}^k \Tr \mean[ \bT_{j_1 j_1} \bT_{j_1 j_1}^\top ] } \overset{(b)}{=} 2 \sqrt{k} m_2,
\end{align*}
where $(a)$ uses the independence of $\bT_{j_1,j_1}$ and $\bT_{j_2,j_2}$, and $(b)$ uses \eqref{eq:thm_repeated:T_same}. On the other hand,
\begin{align}
    \Big\| \; &\big\| \msum_{j_1 \neq j_2} \bT_{j_1j_2} \big\| \; \Big\|_{L_2} 
    \;\coloneqq\; 
    \sqrt{ \msum_{j_1 \neq j_2, l_1 \neq l_2} \Tr \mean[ \bT_{j_1 j_2}\bT_{l_1 l_2}^\top ]  }. \label{eq:thm_repeated:second_mixed_moment}
\end{align}
Consider each summand in \eqref{eq:thm_repeated:second_mixed_moment}. If $j_1, j_2, l_1, l_2$ are all distinct, the summand vanishes since $\bT_{j_1j_2}$ and $\bT_{l_1l_2}$ are independent and zero-mean. Otherwise, we can use \eqref{eq:thm_repeated:T_bound} and \eqref{eq:thm_repeated:T_diff} to upper bound each summand by $\frac{1}{6} (m_3)^2$. The number of non-zero terms is $k^4 - k(k-1)(k-2)(k-3) = 6k^3 - 11k^2 + 6k \leq 6k^3 + 6k \leq 12 k^3$, so \eqref{eq:thm_repeated:second_mixed_moment} can be upper bounded by $2 k^{3/2} m_3$. In summary,
\begin{align*}
    \mfrac{1}{2} | \tau_{2,i}| \;\leq&\; \mfrac{1}{2}
   E_i \big\| \msum_{j=1}^k \| \bT_{jj} \| \big\|_{L_2} + \mfrac{1}{2} F_i \big\|\msum_{j_1 \neq j_2} \| \bT_{j_1 j_2} \| \big\|_{L_2} 
    \;\leq\; 
    k^{1/2}m_2 E_i + k^{3/2} m_3 F_i\;,
\end{align*}
which finishes the proof.
\end{proof}

\vspace{.2em}

\begin{lemma} \label{lem:thm_repeated_chain_rule} The bounds  \eqref{bound:repeated_thm_first_order},  \eqref{bound:repeated_thm_second_order_1}, \eqref{bound:repeated_thm_second_order_2} and \eqref{bound:repeated_thm_third_order} hold.
\end{lemma} 

\begin{proof}
The argument is mostly the same as Lemma \ref{lem:main_thm_chain_rule}, except that we use the permutation invariance assumption \eqref{eq:permutation:invariance} and Lemma \ref{lem:perm_inv_implies_same_derivative} to handle $C_i$, $E_i$ and $F_i$. To obtain \eqref{bound:repeated_thm_first_order}, note that the vector norm $\| \argdot \|$ is a convex function, so by Jensen's inequality,
\begin{align*}
    C_i = \Big\| \; \Big\| \mean \big[ \mfrac{\partial}{\partial \bx_{i1}} g(\bV_i(\bzero)) &\big| \tilde{\Phi}, \Psi \big] \Big\| \; \Big\|_{L_2} \leq \Big\| \;  \mean \Big[ \Big\| \mfrac{\partial}{\partial \bx_{i1}} g(\bV_i(\bzero)) \Big\| \Big| \tilde{\Phi}, \Psi \Big] \; \Big\|_{L_2} \\
    &= \Big\| \; \Big\| \mfrac{\partial}{\partial \bx_{i1}} g(\bV_i(\bzero)) \Big\| \; \Big\|_{L_2} \overset{(a)}{=} k^{-1/2} \Big\| \; \Big\| D_i g(\bV_i(\bzero)) \Big\| \; \Big\|_{L_2}.
\end{align*}
In the last equality (a), we have invoked the permutation invariance assumption on $f$ and Lemma \ref{lem:perm_inv_implies_same_derivative}, which implies that
$$\sqrt{ \mean \Big\| \mfrac{\partial}{\partial \bx_{i1}} g(\bV_i(\bzero)) \Big\|^2} = \sqrt{ \mean \Big( \mfrac{1}{k} \msum_{j=1}^k \Big\| \mfrac{\partial}{\partial \bx_{ij}} g(\bV_i(\bzero)) \Big\|^2 \Big) } = k^{-1/2} \sqrt{ \mean \| D_i g(\bV_i(\bzero))\|}. $$
This allows us to apply a similar argument to that in Lemma \ref{lem:main_thm_chain_rule}. By chain rule, almost surely, $D_i g(\bV_i(\bzero)) = \partial h\big( f(\bV_i(\bzero))\big)(D_i f(\bV_i(\bzero)))$. For a random function $\bT: \R^{dk} \rightarrow \R^+_0$ and $m \in \N$, define 
$$\zeta'_{i;m}(\bT) 
\coloneqq 
\max \Big\{ 
\big\| \msup_{\bw \in [\bzero, \tilde{\Phi}_1 \bX_i]} \bT(\bw) \big\|_{L_m} 
, 
\big\| \msup_{\bw \in [\bzero, \bZ_i]} \bT(\bw) \big\|_{L_m}  
\Big\},$$
which is analogous to the definition of $\zeta_{i;m}$ in Lemma \ref{lem:zeta_properties} and satisfies all the properties in Lemma \ref{lem:zeta_properties}. Then
\begin{align*}
    &\| \; \| D_i g(\bV_i(\bzero)) \| \; \|_{L_2} \;\overset{(a)}{\leq}\; \zeta'_{i;2} \big( \big\| D_i g(\bV_i(\argdot)) \big\| \big) \\
    \;&\leq\;  \zeta'_{i;2} \big( \big\| \partial h\big( f(\bV_i( \argdot) \big) \big\| \big\| D_i f(\bV_i( \argdot)) \big\| \big) \;\leq\;  
    \zeta'_{i;2} \big( \gamma_1(h)
    \big\| D_i f(\bV_i( \argdot)) \big\| \big) \\
    \;&\overset{(b)}{\leq}\;  
    \gamma_1(h) \zeta'_{i;2} ( \| D_i f(\bV_i( \argdot)) \| ) \;\leq\; \gamma_1(h) \alpha_1\;.
\end{align*}
where we have used Lemma \ref{lem:zeta_properties} for (a) and (b). Therefore we obtain the bound \eqref{bound:repeated_thm_first_order} as
$$\mmax_{i \leq n} C_i \leq \mmax_{i \leq n} k^{-1/2} \, \| \; \| D_i g(\bV_i(\bzero)) \| \; \|_{L_2} \leq k^{-1/2} \gamma_1(h) \alpha_1.$$
To obtain \eqref{bound:repeated_thm_second_order_1} for the second partial derivatives, we use Jensen's inequality and Lemma \ref{lem:perm_inv_implies_same_derivative} again to get 
\begin{align*}
E_i \;&=\; \Big\| \; \Big\| \mean \big[
\mfrac{\partial^2}{\partial \bx_{i1}^2} g(\bV_i(\bzero)) \big| \tilde \Phi, \Psi \big] \Big\| \; \Big\|_{L_2} \\
\;&\leq\; 
\Big\| \; \Big\|
\mfrac{\partial^2}{\partial \bx_{i1}^2} g(\bV_i(\bzero)) \Big\| \; \Big\|_{L_2} \;=\; \sqrt{ \mfrac{1}{k} \msum_{j=1}^k \Big\|
\mfrac{\partial^2}{\partial \bx_{ij}^2} g(\bV_i(\bzero)) \Big\|^2 } \\
\;&\leq\; \mfrac{1}{\sqrt{k}}  \sqrt{ \sum_{j_1, j_2=1}^k \Big\|
\mfrac{\partial^2}{\partial \bx_{ij_1}\partial \bx_{ij_2}} g(\bV_i(\bzero)) \Big\|^2 } 
\;=\; \mfrac{1}{\sqrt{k}} \, \| \; \| D_i^2 g( \bV_i( \bzero) ) \| \; \|_{L_2}.
\end{align*}
By the same argument for the mixed the derivatives, in \eqref{bound:repeated_thm_second_order_2},
\begin{align*}
F_i = \Big\| \; \big\| \mean \big[
\mfrac{\partial^2}{\partial \bx_{i1} \partial \bx_{i2}} g(\bV_i(\bzero)) \big| \tilde \Phi, \Psi \big] \big\| \Big\|_{L_2} \;&\leq\; \Big\| \; \big\| 
\mfrac{\partial^2}{\partial \bx_{i1} \partial \bx_{i2}} g(\bV_i(\bzero)) \big\| \Big\|_{L_2} \\
&\leq \; \frac{1}{\sqrt{k(k-1)}} \, \Big\| \; \| D_i^2 g( \bV_i( \bzero) ) \| \; \|_{L_2}\;.
\end{align*}
$\big\| \; \big\| D_i^2 g( \bV_i( \bzero) ) \big\| \; \big\|_{L_2}$ is bounded similarly in Lemma \ref{lem:main_thm_chain_rule} except that we are bounding an $L_2$ norm instead of an $L_1$ norm.
\begin{align*}
    &\big\| \| D_i^2 g(\bV_i (\bzero)) \| \big\|_{L_2}
    \big\| \\
    &\le \zeta'_2\big( \big\| D_i^2 g(\bV_i (\argdot)) \big\|\big) \\
    &\overset{(a)}{\leq}
    \zeta'_2 \big( 
    \big\| \partial^2 h\big(f( \bV_i(\argdot)) \big) \big\| \big\|D_i f(\bV_i(\argdot)) \big\|^2  
    + 
     \big\| \partial h\big( f(\bV_i(\argdot)) \big) \big\|
     \big\|  D^2_i f(\bV_i(\argdot))
      \big\| \big) 
      \\&
    \leq\;
    \zeta'_2 \Big(
    \gamma_2(h) \|D_i f(\bV_i(\argdot)) \|^2  + \gamma_1(h)
      \|  
      D^2_i f(\bV_i(\argdot))
      \| \Big) \\
    \;&\overset{(b)}{\leq}\; \gamma_2(h) \; \zeta'_4(\| D_i f(\bW_i(\argdot)) \| )^2 + \gamma_1(h) \;  \zeta'_2(\| D^2_i f(\bW_i(\argdot)) \|)
    \\
    \;&\leq\; \gamma_2(h) \alpha_1^2 
    + 
    \gamma_1(h) \alpha_2\;,
\end{align*} 
where we used Lemma \ref{lem:faa_di_bruno} to obtain $(a)$ and Lemma \ref{lem:zeta_properties} to get $(b)$. Therefore, the bounds \eqref{bound:repeated_thm_second_order_1} and \eqref{bound:repeated_thm_second_order_2} are obtained as
\begin{align*}
    \mmax_{i \leq n} E_i &\leq k^{-1/2} (  \gamma_2(h) \alpha_1^2 
    + 
    \gamma_1(h) \alpha_2 )\;,
    &&
    \mmax_{i \leq n} F_i \leq k^{-3/2} ( \gamma_2(h) \alpha_1^2 
    + 
    \gamma_1(h) \alpha_2)\;.
\end{align*}
Finally for \eqref{bound:repeated_thm_third_order}, recall that
$$
M_i 
\coloneqq 
\max \braces{ 
\big\|
{\textstyle\sup_{\bw \in [\bzero, \tilde{\Phi}_i \bX_i] }}
\| D_i^3 g \big(\bV_i (\bw) \big) \|\big\|_{L_2} 
,\,
\big\|
{\textstyle\sup_{\bw \in [\bzero, \bY_i] }}
\| D_i^3 g \big(\bV_i (\bw) \big) \| \big\|_{L_2} 
}\;,
$$
and notice that it is the same quantity as $M_i$ from Lemma \ref{lem:main_thm_chain_rule} except that $\bW_i$ is replaced by $\bV_i$, $\Phi_i \bX_i$ is replaced by $\tilde{\Phi}_i \bX_i$ and $\bZ_i$ is replaced by $\bY_i$. The same argument applies to give
$$\mmax_{i \leq n} M_i \;\leq\; \lambda(n,k)\;,$$
which completes the proof.
\end{proof}

\vspace{.2em}

Finally we present the following lemma that describes properties of derivatives of a function satisfying permutation invariance condition:

\begin{lemma} \label{lem:perm_inv_implies_same_derivative} For a function $f \in \cF(\R^{kd}, \R^q)$ that satisfies the permutation invariance assumption
\begin{equation} \label{eq:permutation:invariance:partial}
  f(\bx_1,\ldots,\bx_k)
  \;=\;
  f(\bx_{\pi(1)},\ldots,\bx_{\pi(k)})
\end{equation}
for any permutation $\pi$ of $k$ elements, then at $\bzero \in \R^{kd}$, its derivatives satisfy, for $\bx_1, \ldots, \bx_d \in \R^d$,
\begin{proplist}
    \item $\frac{\partial}{\partial \bx_1} f(\bzero) = \ldots = \frac{\partial}{\partial \bx_d} f(\bzero) $, 
    \item $\frac{\partial^2}{\partial \bx_1^2} f(\bzero) = \ldots = \frac{\partial^2}{\partial \bx_k^2} f(\bzero) $, 
    \item $\frac{\partial^2}{\partial \bx_r \partial \bx_s} f(\bzero)$ is the same for $r \neq s$, $1 \leq r, s \leq k$. 
\end{proplist}
\end{lemma}

\begin{proof}
For $j \leq k, l \leq d$, denote $\be_{jl}$ as the $\big( (j-1)d + l \big)^\text{th}$ basis vector in $\R^{kd}$ and $x_{jl}$ as the $l$th coordinate of $\bx_d$. Without loss of generality we can set $q=1$, because it suffices to prove the results coordinate-wise over the $q$ coordinates.. Consider $\frac{\partial}{\partial \bx_{j}} f(\bzero)$, which exists by assumption and can be written as  $$\mfrac{\partial}{\partial \bx_j} f(\bzero) 
= 
\Big( 
\mfrac{\partial}{\partial x_{j1}} f(\bzero), \ldots, \mfrac{\partial}{\partial x_{jd}} f(\bzero)    
\Big)^\top.$$
For each $l \leq d$, the one-dimensional derivative is defined as
$$\mfrac{\partial}{\partial x_{jl}} f(\bzero) 
\;\coloneqq\; 
{\textstyle \lim}_{\epsilon \rightarrow 0}
\mfrac{ f( \epsilon \be_{jl} ) - f(\bzero)}{\epsilon}
\;\overset{(a)}{=}\;
{\textstyle \lim}_{\epsilon \rightarrow 0}
\mfrac{ f( \epsilon \be_{1l} ) - f(\bzero)}{\epsilon} 
\;=\; 
\mfrac{\partial}{\partial x_{1l}} f(\bzero).
$$
In (a) above, we have used the permutation invariance assumption \eqref{eq:permutation:invariance} across $j \leq k$. This implies $\frac{\partial}{\partial \bx_1} f(\bzero) = \ldots = \frac{\partial}{\partial \bx_k} f(\bzero)$ as required. The second derivative $\frac{\partial^2}{\partial \bx_j^2} f(\bzero)$ is a $\R^{d \times d}$ matrix with the $(l_1, l_2)^{\text{th}}$ coordinate given by $\frac{\partial^2}{\partial x_{jl_1} \partial x_{jl_2}} f(\bzero)$, which is in turn defined by
\begin{align*}
\frac{\partial^2}{\partial x_{jl_1} \partial x_{jl_2}} f(\bzero)
\;&\coloneqq\; 
{\textstyle \lim}_{\delta \rightarrow 0}
\mfrac{ 
\frac{\partial}{\partial x_{jl_1}} f(\delta \be_{jl_2} ) - \frac{\partial}{\partial x_{jl_1}} f(\bzero)}
{\delta} \\
\;&\overset{(b)}{=}\; 
{\textstyle \lim}_{\delta \rightarrow 0} {\textstyle \lim}_{\epsilon \rightarrow 0}
\mfrac{ 
(f(\delta \be_{jl_2} + \epsilon \be_{jl_1}) - f(\delta \be_{jl_2}))  - (f(\epsilon \be_{jl_1}) - f(\bzero))}
{\epsilon \delta} \\
\;&\overset{(c)}{=}\; 
{\textstyle \lim}_{\delta \rightarrow 0} {\textstyle \lim}_{\epsilon \rightarrow 0}
\mfrac{ 
(f(\delta \be_{1l_2} + \epsilon \be_{1l_1}) - f(\delta \be_{1l_2}))  - (f(\epsilon \be_{1l_1}) - f(\bzero))}
{\epsilon \delta} \\
\;&=\;
\frac{\partial^2}{\partial x_{1l_2} \partial x_{1l_1}} f(\bzero).
\end{align*}
We have used the definition for the first derivatives in (b) and assumption \eqref{eq:permutation:invariance} in (c). This implies, as before, $\frac{\partial^2}{\partial \bx_1^2} f(\bzero) = \ldots = \frac{\partial^2}{\partial \bx_k^2} f(\bzero) $. For the mixed derivatives, notice that assumption \eqref{eq:permutation:invariance} implies, for $r \neq s$, $1 \leq r, s, \leq k$ and $1 \leq l_1, l_2 \leq d$,
$$
f(\delta \be_{rl_2} + \epsilon \be_{sl_1})
\;=\;
f(\delta \be_{1l_2} + \epsilon \be_{2l_1}),
$$
by considering a permutation that brings $(r, s)$ to $(1, 2)$. Therefore, by an analogous argument, 
\begin{align*}
\frac{\partial^2}{\partial x_{rl_1} \partial x_{sl_2}} f(\bzero)
\;&\coloneqq\; 
{\textstyle \lim}_{\delta \rightarrow 0}
\mfrac{ 
\frac{\partial}{\partial x_{rl_1}} f(\delta \be_{sl_2} ) - \frac{\partial}{\partial x_{rl_1}} f(\bzero)}
{\delta} \\
\;&=\; 
{\textstyle \lim}_{\delta \rightarrow 0} {\textstyle \lim}_{\epsilon \rightarrow 0}
\mfrac{ 
(f(\delta \be_{sl_2} + \epsilon \be_{rl_1}) - f(\delta \be_{sl_2}))  - (f(\epsilon \be_{rl_1}) - f(\bzero))}
{\epsilon \delta} \\
\;&=\; 
{\textstyle \lim}_{\delta \rightarrow 0} {\textstyle \lim}_{\epsilon \rightarrow 0}
\mfrac{ 
(f(\delta \be_{1l_2} + \epsilon \be_{2l_1}) - f(\delta \be_{1l_2}))  - (f(\epsilon \be_{2l_1}) - f(\bzero))}
{\epsilon \delta} \\
\;&=\;
\frac{\partial^2}{\partial x_{1l_2} \partial x_{1l_1}} f(\bzero).
\end{align*}
This implies $\frac{\partial^2}{\partial \bx_r \partial \bx_s} f(\bzero)$ is the same for $r \neq s$, $1 \leq r, s \leq k$. 
\end{proof}

%% file: appendix_6_proof_examples.tex
\section{Derivation of examples} \label{appendix:der:examples}

Different versions of Gaussian surrogates are used throughout the computation in this section. For clarity, we denote $\bx_{11:nk} \coloneqq \{\bx_{11}, \ldots, \bx_{nk}\}$ and define
\begin{align*}
\bW(\argdot) \;&\coloneqq\; (\Phi_1 \bX_1, \ldots, \Phi_{i-1} \bX_{i-1}, \argdot, \bZ_{i+1}, \ldots, \bZ_n), \\
\tilde{\bW}(\argdot) \;&\coloneqq\; (\tilde{\bX}_1, \ldots, \tilde{\bX}_1, \argdot, \tilde{\bZ}_{i+1}, \ldots, \tilde{\bZ}_n),
\end{align*}
where:
\begin{itemize}
    \item $\Phi_1 \bX_1, \ldots, \Phi_n \bX_n  \in \cD^{k}$ are the augmented data vectors and $\bZ_{1}, \ldots, \bZ_{n} \in \cD^{k}$ are the i.i.d. surrogate vectors, both defined in 
    \Cref{thm:main} 
    (corresponding to $\bZ^{\delta}_i$ defined with $\delta=0$ in Theorem \ref{thm:main:general});  
    \item $\tilde{\bX}_1, \ldots, \tilde{\bX}_n \in \cD^{k}$ are the unaugmented data vectors ($k$-replicate of original data) whereas the surrogate vectors are denoted $\tilde{\bZ}_1, \ldots, \tilde{\bZ}_n \in \cD^{k}$, both defined in 
    \eqref{eq:no_aug}.
\end{itemize}
As before, we write $\Phi\cX = \{\Phi_1\bX_1, \ldots, \Phi_n\bX_n\}$, $\cZ =\{\bZ_1,\ldots, \bZ_n\}$, $\tilde\cX = \{\tilde\bX_1, \ldots, \tilde\bX_n\}$ and $\tilde \cZ = \{\tilde \bZ_1, \ldots, \tilde\bZ_n\}$. In the case $\cZ$ and $\tilde \cZ$ are Gaussian, existence of $\cZ$ and $\tilde \cZ$ is automatic when $\bZ_i$ and $\tilde\bZ_i$ are allowed to take values in $\R^d$ and the only constraints are their respective mean and variance conditions 
\eqref{eqn:defn:surrogates} and \eqref{eqn:defn:surrogates:no_aug}. 
Therefore, we omit existence proof for all examples except for the special case of ridge regression in Appendix \ref{appendix:ridge}. Finally, for functions $f: \cD^{nk} \rightarrow \R^q$ and $g: \cD \rightarrow \R^q$, and for any $s \leq q$, we use $f_s: \cD^{nk} \rightarrow \R$ and $g_s: \cD \rightarrow \R$ to denote the $s$-th coordinate of $f$ and $g$ respectively.

\subsection{Empirical averages} \label{appendix:empirical_averages}

\vspace{.3em}

In this section, we first prove 
\Cref{prop:empirical_average} 
by verifying that for the empirical average, the bounds in Lemma \ref{lem:variance:convergence:general} and \ref{lem:d_H:convergence:general} decay, and by computing the relevant variances and confidence intervals.

\vspace{.3em}

\begin{proof}[Proof of 
    \Cref{prop:empirical_average}] 
We first apply Lemma \ref{lem:d_H:convergence:general} to compare the distance in $d_H$ of $f(\Phi\cX)$ to $f(\cZ)$. To do so, we need to compute the noise stability terms for $f(\bx_{11:nk})=\frac{1}{nk} \sum_{i=1}^n \sum_{j=1}^k \bx_{ij}$. We first compute the derivatives: for any $\bv \in \R^{dk}$, almost surely,
\begin{align*}
    D_i f( \bW_i ( \bv ) ) \;&=\; \mfrac{1}{n k} (\bI_{d}, \ldots, \bI_d)^\top \in \R^{dk \times d}\;,
    &\text{ and }&&
    D^2_i f( \bW_i ( \bv ) )\;=\;\bzero\;.
\end{align*}
Then, for all $m \in \N$ we have
\begin{align*}
\alpha_{1;m} 
  \coloneqq& \sum_{s \leq d}
  \max_{i\leq n}\max\Bigl\{\,
  \Bigl\|{\sup_{\bw\in[\bzero,\Phi_i\bX_i]}}
  \|D_i f_s(\bW_i(\bw))\|\Bigr\|_{L_m},\,
    \Bigl\|{\sup_{\bw\in[\bzero,\bZ_i]}}
  \|D_i f_s(\bW_i(\bw))\|\Bigr\|_{L_m}
  \Bigr\}
  \\
  \;=&\; \msum_{s \leq d} \mfrac{1}{nk} \big\| (\bI_{d}, \ldots, \bI_d)^\top \be_s \big\|
  \;=\;\mfrac{d}{n k^{1/2}},
\end{align*} 
and the noise stability terms associated with higher derivatives are $ \alpha_{2;m} = \alpha_{3;m} = 0$. Since $d$ is fixed and $\phi_{11}\bX_1$ and $\bZ_1$ have bounded 4th moments, we get
\begin{align*}
    &
    c_X \;=\; \mfrac{1}{6} \sqrt{\mean \| \phi_{11}\bX_1 \|^6} = O(1)\;, 
    &&
    c_Z \;=\; \mfrac{1}{6} \sqrt{\mean\Bigl[ \Bigl(\mfrac{1}{k} \msum_{j \leq k, s \leq d} |Z_{1js}|^2 \Bigr)^3\Bigr] }\;=\;O(1)\;.
\end{align*}
Therefore, the bounds in Lemma \ref{lem:d_H:convergence:general} (concerning weak convergence) with $\delta$ set to $0$ become, respectively,
\begin{align}
    (nk)^{3/2} (n(\alpha_{1;6})^3  + 3 n^{1/2} \alpha_{1;4} \alpha_{2;4} + \alpha_{3;2} ) (c_X + c_Z) 
    \;&=\;
    O(n^{-1/2})\;.  \label{eqn:emp:ave:d_H:bound}
\end{align}
Note that while the above calculation uses $\Phi_i\bX_i, \bZ_i, \bW_i$ in the case of augmentation, the same calculation holds for $\tilde\bX_i, \tilde\bZ_i, \tilde\bW_i$ in the case of no augmentation. Therefore, \eqref{eqn:emp:ave:d_H:bound} and Lemma \ref{lem:d_H:convergence:general} lead to the required convergence in (i) that as $n \rightarrow \infty$,
\begin{align*}
    &
    d_{\cH}(\sqrt{n} f(\Phi\cX), \sqrt{n} f(\cZ) ) \xrightarrow{d} 0 \;,
    &&
    d_{\cH}(\sqrt{n} f(\tilde\cX), \sqrt{n}  f(\tilde\cZ) ) \xrightarrow{d} 0 \;.
\end{align*}

\vspace{.3em}

To prove the statements on variances and confidence intervals, we first note that the equality in variance can be directly obtained by noting that moments of $\bZ_i$ match moments of $\Phi_i \bX_i$, which implies
\begin{align*}
    \Var f(\Phi\cX)  \;=\; \mfrac{1}{n} \Var\Big[ \mfrac{1}{k} \msum_{j=1}^k \phi_{1j}\bX_1 \Big] \;=\;  \mfrac{1}{n} \Var\Big[ \mfrac{1}{k} \msum_{j=1}^k \bZ_{1j} \Big] \;=\; \Var f(\cZ)\;.
\end{align*}
The same argument implies $\Var f(\tilde\cX) = \Var f(\tilde \cZ)$. The next step is to obtain the formula for variances and asymptotic confidence intervals. Since $\bZ_i$ is Gaussian in $\R^{dk}$ with mean $\bone_{k\times 1} \otimes \mu$ and variance $\bI_{k} \otimes \Var[\phi_{11}\bX_1] + (\bone_{k \times k} - \bI_k) \otimes \Cov[ \phi_{11} \bX_1, \phi_{12} \bX_1] $, we have
$$
\mfrac{1}{k} \msum_{j=1}^k \bZ_{ij} = \mfrac{1}{k} \underbrace{(\bI_d \ldots \bI_d)}_{k\text{ copies of }\bI_d} \bZ_i \sim \cN\big( \mean[\phi_{11}\bX_1], V \big)\;.$$
where
$$V\coloneqq\mfrac{1}{k} \Var[ \phi_{11} \bX_1]  + \mfrac{k-1}{k} \Cov[ \phi_{11} \bX_1, \phi_{12} \bX_1].$$
We also remark that as the Gaussian vectors $(\bZ_1,\dots,\bZ_n)$ are independent, the empirical averages $\frac{1}{k} \sum_{j=1}^k \bZ_{1j}, \ldots, \frac{1}{k} \sum_{j=1}^k \bZ_{nj}$ are also independent. This directly implies that 
\begin{equation} \label{eqn:emp:ave:augmeted:est:normal}
f(\cZ) 
\;=\; \mfrac{1}{nk} \msum_{i=1}^n \msum_{j=1}^k \bZ_{ij} \sim \cN\big( \mean[\phi_{11}\bX_1], \mfrac{1}{n} V  \big)\;. 
\end{equation}
This gives the desired variance for $f(\cZ)$. On the other hand, since each $\tilde \bZ_i$ is a Gaussian in $\R^{dk}$ with mean $\bone_{k\times 1} \otimes \mean[\bX_1]$ and variance $\bone_{k \times k} \otimes \Var[\bX_1]$, it can be viewed as a $k$-replicate of a Gaussian vector $\tilde\bV_i$ in $\R^d$ with mean $\mean[\bX_1]$ and $\Var[\bX_1]$. By independence of $\tilde{\bZ}_i$'s, $\bV_i$'s are also independent and therefore
\begin{equation} \label{eqn:emp:ave:unaugmeted:est:normal}
f(\tilde \cZ) 
\;=\; 
\mfrac{1}{nk} \msum_{i=1}^n\msum_{j=1}^k \tilde \bZ_{i1} 
\;\overset{d}{=}\; 
\mfrac{1}{n} \msum_{i=1}^n \bV_i 
\sim 
\cN\big( \mean[\bX_1], \mfrac{1}{n} \Var[\bX_1] \big)\;,
\end{equation}
giving the variance expression for $f(\tilde \cZ)$. Finally, for $d=1$, the normal distributions given in \eqref{eqn:emp:ave:augmeted:est:normal} and \eqref{eqn:emp:ave:unaugmeted:est:normal} imply that the lower and upper $\alpha/2$-th quantiles for $f(\cZ)$ and $f(\tilde \cZ)$ are given respectively as
\begin{align*}
     &
     \mean[\phi_{11} \bX_1]\,\pm\,\mfrac{1}{\sqrt{n}}z_{\alpha/2} \sqrt{ V } 
     \;=\;
     \mean[\phi_{11} \bX_1]\,\pm\,\mfrac{1}{\sqrt{\vartheta(f)^2 n}}z_{\alpha/2} \sqrt{\Var [\bX_1]}\;,
     \\
     &
     \mean[\phi_{11} \bX_1]\,\pm\,\mfrac{1}{\sqrt{n}}z_{\alpha/2} \sqrt{\Var [\bX_1]}\;.
\end{align*}
These quantiles are asymptotically valid for $f(\Phi\cX)$ and $f(\tilde\cX)$ respectively since convergence in $d_{\cH}$ implies convergence in distribution by 
\cref{lem:d_H}, 
which finishes the proof.
\end{proof}

\subsection{Exponential of negative chi-squared statistic} \label{appendix:derivation:exp_neg_chi_sq}

\vspace{.2em}

In this section, we prove Proposition \ref{prop:exp_neg_chi_squared} for the one-dimensional statistic defined in 
\eqref{eqn:defn:exp:neg:chi:sq:stat}:
\begin{equation*}
f(x_{11},\ldots,x_{nk}) \coloneqq \exp\big( - \big( \mfrac{1}{\sqrt{n}k} \msum_{i\leq n} \msum_{j\leq k} x_{ij} \big)^2 \big)\;.
\end{equation*}
We also state a 2d generalization of this statistic used in our simulation and prove an analogous lemma that justifies convergences and analytical formula for its confidence regions.

\begin{proof}[Proof of Proposition \ref{prop:exp_neg_chi_squared}] For convergence in $d_{\cH}$ and variance, define $$
g(x) \coloneqq \mfrac{1}{\sqrt{n}} \exp( - n x^2)
\text{ and }
\tilde f(x_{11:nk}) \coloneqq g\big(\mfrac{1}{nk}\msum_{i \leq n, j \leq k} x_{ij}\big)
\;.
$$
Then, the required statistic in 
\eqref{eqn:defn:exp:neg:chi:sq:stat}
satisfies $f(x_{11:nk}) = \sqrt{n} \tilde f(x_{11:nk})$, and applying Lemma \ref{lem:plugin:formal}(ii) with $\delta$ set to $0$ to $\tilde f$ and $g$ will recover the convergences
\begin{align*}
    d_{\cH}(\sqrt{n}\tilde f(\Phi\cX), \sqrt{n} \tilde f(\cZ)) \;&=\; d_{\cH}( f(\Phi\cX), f(\cZ))\;,
    \\
    n ( \Var[ \tilde f(\Phi\cX) ] - \Var [\tilde f(\cZ)] ) \;&=\; \Var[ f(\Phi\cX) ] - \Var [ f(\cZ)]\;.
\end{align*}
It now suffices to compute the noise stability terms $\nu_{r;m}(g)$ used in Lemma \ref{lem:plugin:formal}(ii) defined for $g$. The derivatives for $g$ can be bounded by
\begin{align*}
    \partial g(x) \;&=\; -2 n^{1/2} x \exp( - n x^2)\;,
    \quad
    \partial^2 g(x) \;=\; -2 n^{1/2} \exp( - n x^2) +  4 n^{3/2} x^2 \exp( - n x^2)\;,
    \\
    \partial^3 g(x) \;&=\; 12 n^{3/2} x \exp( - n x^2) - 8 n^{5/2} x^3 \exp( - n x^2)\;.
\end{align*}
Note that $\exp( - n x^2) \in [0,1]$ for all $x \in \R$, so only $x$, $x^2$ and $x^3$ play a role in the bound for $\nu_{1;m}$. The noise stability terms can now be bounded by
\begin{align}
    \nu_{1;m} \;&=\; 
    \max_{\substack{i \leq n}}\max\bigg\{
      \Big\|{\msup_{\bw\in[\bzero,\Phi_i\bX_i]}}
      |\partial g(\overline\bW_i(\bw))|\Big\|_{L_m},\,
        \Big\|{\msup_{\bw\in[\bzero,\bZ_i]}}
      |\partial g(\overline\bW_i(\bw))|\Big\|_{L_m}
      \bigg\}  \nonumber
    \\
    \;&\leq\;
    2 n^{1/2} \max_{\substack{i \leq n}}\max\bigg\{
      \Big\|{\msup_{\bw\in[\bzero,\Phi_i\bX_i]}}
      | \overline\bW_i(\bw)|\Big\|_{L_m},\,
        \Big\|{\msup_{\bw\in[\bzero,\bZ_i]}}
      | \overline\bW_i(\bw)|\Big\|_{L_m}
      \bigg\}
    \nonumber 
    \\
    \;&\leq\; 
    2 n^{1/2} \max_{\substack{i \leq n}}
      \big\|{\msup_{\bw\in[\bzero,\Phi_i\bX_i] \cup [\bzero, \bZ_i]}}
      | \overline\bW_i(\bw)|\big\|_{L_m}
    \;. \label{eqn:exp:neg:chi:sq:intermediate:beta_1_m}
\end{align}
We need to bound the absolute value of $\overline\bW_i(\bw)$. Define $\bA_{i'} \coloneqq \sum_{j=1}^k \phi_{i'j} \bX_{i'} $ and $\bB_{i'} \coloneqq \sum_{j=1}^k \bZ_{i'j} $, and write $\cI \coloneqq [\bzero, \Phi_i\bX_i] \cup [\bzero, \bZ_i]$. Then by triangle inequality, 
\begin{align}
    \Big\| \sup_{\bw \in \cI} \big| \overline\bW_i(\bw)\big| \Big\|_{L_m} &= \mfrac{1}{nk}  \Big\| \, \sup_{\bw \in \cI} \Big| \msum_{i'=1}^{i-1} \msum_{j=1}^k \phi_{i'j} \bX_{i'} + \msum_{j=1}^k \bw_j + \msum_{i'=i+1}^n \msum_{j=1}^k \bZ_{i'j} \Big|\,\Big\|_{L_m}\nonumber
    \\
    &= \mfrac{1}{nk} \Big\| \msup_{\bw \in \cI} \Big| \msum_{i'=1}^{i-1} \bA_{i'} + \msum_{j=1}^k \bw_j + \msum_{i'=i+1}^n  \bB_{i'} \Big| \Big\|_{L_m}\nonumber
    \\
    &\leq \mfrac{1}{nk} \Big\|  \big| \msum_{i'=1}^{i-1} \bA_{i'} \big|  + \max\{ | \bA_i|, |\bB_i| \}  +  \big| \msum_{i'=i+1}^n  \bB_{i'} \big|  \Big\|_{L_m}\;. \label{eqn:exp:neg:chi:sq:intermediate:W_i:bound}
\end{align}
Note that $\bA_1, \ldots, \bA_{i-1}$ are i.i.d.\,random variables with zero mean and finite 12th moments by assumption. Also, for $m \leq 12$, by triangle inequality,
$$
\| \bA_{i'} \|_{L_m} \leq \msum_{j \leq k} \| \phi_{i'j}\bX_i \|_{L_m} \;=\; O(k)\;.
$$
Rosenthal's inequality from Lemma \ref{lem:rosenthal} implies, for $m \leq 12$, there exists a constant $K_m$ depending only on $m$ such that  
\begin{align*}
\big\| \msum_{i' < i} \bA_{i'} \big\|_{L_m} 
\;&\leq\; 
K_m \max \big\{ 
i^{1/m} \big\|\bA_1 \big\|_{L_m}
, 
i^{1/2} \big\|\bA_1 \big\|_{L_2} 
\big\} 
\;=\; 
O(n^{1/2} k)\;.
\end{align*}
The exact same argument applies to $\bB_{i+1}, \ldots, \bB_n$, implying that
\begin{align*}
&
\| \bB_{i} \|_{L_m} \;=\; O(k)\;,
&&
\big\| \msum_{i' > i} \bB_{i'} \big\|_{L_m} 
\;=\; 
O(n^{1/2} k)\;.
\end{align*}
Substituting these results into \eqref{eqn:exp:neg:chi:sq:intermediate:W_i:bound} gives the following control on $\overline\bW_i(\bw)$:
\begin{align*}
    \big\| \msup_{\bw \in \cI} \big| \overline\bW_i(\bw)\big| \big\|_{L_m}
    \;=\; O(n^{-1/2})\;,
\end{align*}
and finally substituting the bound into \eqref{eqn:exp:neg:chi:sq:intermediate:beta_1_m} gives, for $m \leq 12$,
\begin{align*}
    \nu_{1;m} \;=\; O(1)\;.
\end{align*}
The arguments for $\nu_{2;m}$ and $\nu_{3;m}$ are similar, except that $\nu_{2;m}$ involves $x^2$ and $\nu_{3;m}$ involves $x^3$. $\nu_{2;m}$ then requires bounding terms of the form
\begin{align*}
    \big\| \msup_{\bw \in \cI} \big| \overline\bW_i&(\bw)\big|^2 \big\|_{L_m}
    \leq \mfrac{1}{n^2k^2} \Big\|  \big(\big| \msum_{i'=1}^{i-1} \bA_{i'} \big|  + \max\{ | \bA_i|, |\bB_i| \}  +  \big| \msum_{i'=i+1}^n  \bB_{i'} \big|\big)^2  \Big\|_{L_m}
    \\
    =& \mfrac{1}{n^2k^2} \Big\|  \big| \msum_{i'=1}^{i-1} \bA_{i'} \big|  + \max\{ | \bA_i|, |\bB_i| \}  +  \big| \msum_{i'=i+1}^n  \bB_{i'} \big|  \Big\|_{L_{2m}}^{2}
    = O( n^{-1} )\;,
\end{align*}
where the argument proceeds as before but now hold only for $m \leq 6$. $\nu_{3;m}$ similarly requires controlling
\begin{align*}
    \big\| \msup_{\bw \in \cI} \big| \overline\bW_i(\bw)\big|^3 \big\|_{L_m}
    \leq& \mfrac{1}{n^3k^3} \Big\|  \big| \msum_{i'=1}^{i-1} \bA_{i'} \big|  + \max\{ | \bA_i|, |\bB_i| \}  +  \big| \msum_{i'=i+1}^n  \bB_{i'} \big|  \Big\|_{L_{3m}}^{3}
    \\
    =& O( n^{-3/2} )\;,
\end{align*}
which holds now for $m \leq 4$. Therefore,
\begin{align*}
    \nu_{2;m} \;&=\; O( n^{1/2} + n^{3/2} \times n^{-1} ) = O(n^{1/2}) &&\text{ for } m \leq 6\;,
    \\
    \nu_{3;m} \;&=\; O( n^{3/2} \times n^{-1/2} + n^{5/2} \times n^{-3/2} ) = O(n) &&\text{ for } m \leq 4\;.
\end{align*}
Note also that the moment terms $c_X= O(1)$ by assumption and $c_Z = O(1)$ since the 4th moment of a Gaussian random variable with finite mean and variance is bounded. Moreover, $g(x) = \frac{1}{\sqrt{n}} \exp(-nx^2) \in [0, n^{-1/2}]$ and therefore $\nu_{0;m}=O(n^{-1/2})$ for all $m \in \N$. The two bounds in Lemma \ref{lem:plugin:formal}(ii) then become:
\begin{align*}
    \big(n^{-1/2} \nu_{1;6}^3  + n^{-1} \nu_{1;4} \nu_{2;4} + n^{-3/2} \nu_{3;2} \big) (c_X + c_Z)
    \;&=\;
    O( n^{-1/2} )
    \;,
    \\
    n^{-1} ( \nu_{0;4}(g) \nu_{3;4}(g) + \nu_{1;4}(g) \nu_{2;4}(g) ) (c_X + c_Z) 
    \;&=\;
    O(n^{-1/2})\;,
\end{align*}
both of which go to zero as $n \rightarrow \infty$. Applying Lemma \ref{lem:plugin:formal}(ii) to $\tilde f$ then gives the desired convergences that
\begin{align*}
    f(\Phi\cX) - f(\cZ) \;&=\; \sqrt{n} (\tilde f(\Phi\cX) - \tilde f(\cZ) ) \;\xrightarrow{d}\; 0\;,
    \\
    \Var[f(\Phi\cX)] - \Var[f(\cZ)] \;&=\; n ( \Var[\tilde f(\Phi\cX)] - \Var[\tilde f(\cZ)]) \;\xrightarrow{d}\; 0\;.
\end{align*}
The exact same argument works for $\tilde \bX$ and $\tilde \bZ$ by setting $\phi_{ij}$ to identity almost surely and by invoking boundedness of 8th moments of $\bX_i$ and $\mean[\bX_i]=0$. Therefore, the same convergences hold with $(\Phi\cX,\cZ)$ above replaced by $(\tilde \cX, \tilde \cZ)$.

\vspace{.3em}

Next, we prove the formulas for variance and quantiles. Recall the function $V(s)\coloneqq(1+4s^2)^{-1/2} - (1+2s^2)^{-1}$ and the standard deviation terms
\begin{align*}
    \tilde{\sigma} 
    \,&\coloneqq\, 
    \sqrt{\Var[\bX_1]}\;,
    &
    \sigma\,&\coloneqq\, 
    \sqrt{\mfrac{1}{k} \Var[\phi_{11}\bX_1] + \mfrac{k-1}{k} \Cov[\phi_{11}\bX_1, \phi_{12}\bX_1]}\;.
\end{align*}
Recall from \eqref{eqn:emp:ave:augmeted:est:normal} and \eqref{eqn:emp:ave:unaugmeted:est:normal} in the proof of 
\Cref{prop:empirical_average}
(empirical averages) that
\begin{align*}
    \mfrac{1}{nk} \msum_{i=1}^n \msum_{j=1}^k \bZ_{ij} \;&\sim\; \cN \big( \mean[\phi_{11}\bX_1], \mfrac{1}{n} \sigma^2 \big) \;\equiv\; \cN \big( 0, \mfrac{1}{n} \sigma^2 \big)\;,
    \\
    \mfrac{1}{nk} \msum_{i=1}^n \msum_{j=1}^k \tilde \bZ_{ij} \;&\sim\; \cN \big( \mean[\bX_1], \mfrac{1}{n} \tilde \sigma^2 \big) \;\equiv\; \cN \big( 0, \mfrac{1}{n} \tilde\sigma^2 \big)\;.
\end{align*}
Thus, the following quantities are both chi-squared distributed with 1 degree of freedom:
\begin{align}
    - \mfrac{1}{\sigma^2} \log f(\cZ) = \mfrac{1}{\sigma^2} \big(\mfrac{1}{\sqrt{n} k} \msum_{i=1}^n \msum_{j=1}^k \bZ_{ij}\big)^2,
    \quad
    - \mfrac{1}{\tilde \sigma^2}  \log f(\tilde \cZ) = \mfrac{1}{\tilde \sigma^2} \big(\mfrac{1}{\sqrt{n} k} \msum_{i=1}^n \msum_{j=1}^k \tilde \bZ_{ij} \big)^2. \label{eqn:exp:neg:chi:sq:chi_sq_1_distn}
\end{align}
Let $\bU$ be a chi-squared distributed random variable with 1 degree of freedom. We can now use the formula of moment generating functions of $\chi^2_1$ to get
\begin{align*}
    \Var[f(\cZ)] \;=\; \Var[ \exp( - \sigma^2 \bU)] \;&=\; \mean[ \exp( - 2\sigma^2 \bU) ] -  \mean[ \exp( - \sigma^2 \bU) ]^2
    \\
    \;&=\;(1+4\sigma^2)^{-1/2} -(1+2\sigma^2)^{-1} \;=\; V(\sigma)\;,
\end{align*}
as desired. The same argument gives the desired variance for the unaugmented case:
\begin{align*}
    \Var[f(\tilde \cZ)] \;=\; V(\tilde \sigma)\;,
\end{align*}
and the ratio $\vartheta(f)$ defined in 
\eqref{eqn:rate_of_shrinkage} 
can be computed by:
\begin{align*}
    \vartheta(f) \;=\; \sqrt{\Var[f(\tilde \cZ)] / \Var[f(\cZ)] }\;=\; \sqrt{V(\tilde \sigma)/V(\sigma)}\;.
\end{align*}
Finally, notice that $(\pi_l, \pi_u )$ are the lower and upper $\alpha/2$-th quantiles for the quantities in \eqref{eqn:exp:neg:chi:sq:chi_sq_1_distn}. The corresponding quantiles for $f(\cZ)$ and $f(\tilde \cZ)$ then follow by monotonicity of the transforms $x \mapsto \exp(- \sigma^2 x)$ and $x \mapsto \exp(- \tilde\sigma^2 x)$: They are given by
     \begin{align*}
    \big(\exp\big(- \sigma^2\pi_u\big),&\,\exp\big(- \sigma^2 \pi_l\big)\big)
&\;\text{and}\;&&
\big(\exp(-\tilde\sigma^2\pi_u),&\,\exp(-\tilde\sigma^2\pi_l)\big)\;,
    \end{align*}  
as required, and are asymptotically valid for $f(\Phi\cX)$ and $f(\tilde\cX)$ respectively since convergence in $d_{\cH}$ implies convergence in distribution by 
\cref{lem:d_H}.
\end{proof}

We next prove \cref{lem:exp_neg_chi_squared_2D} concerning the 2d generalization of the toy statistic 
 \eqref{eqn:defn:exp:neg:chi:sq:stat}:
\begin{align*} 
    f_2(\bx_{11:nk})
    \;\coloneqq\; 
    \msum_{s=1}^2 \exp\big( - \big( \mfrac{1}{\sqrt{n}k} \msum_{i=1}^n \msum_{j=1}^k x_{ijs} \big)^2 \big)\;.
\end{align*}

\begin{proof}[Proof of Lemma \ref{lem:exp_neg_chi_squared_2D}] The proof for (i) is similar to the 1d case. Recall $g(x) \coloneqq \mfrac{1}{\sqrt{n}} \exp(-nx^2)$ defined in the proof of Proposition \ref{prop:exp_neg_chi_squared}. Define $g_2: \R^2 \rightarrow \R$ and $\tilde f_2: \R^{2nk} \rightarrow \R$ as
\begin{align*}
    g_2(\bx) \;&\coloneqq\; \msum_{s=1}^2 g(x_s)\;,
    &\text{and}&&
    \tilde f_2(\bx_{11:nk}) \;&\coloneqq\; g_2\big( \mfrac{1}{nk} \msum_{i \leq n, j \leq k} \bx_{ij} \big)\;.
\end{align*}
Then as before, $\tilde f_2(\bx_{11:nk}) = \sqrt{n} f_2(\bx_{11:nk})$, and applying Lemma \ref{lem:plugin:formal}(ii) to $\tilde f_2$ and $g_2$ will recover convergences for
\begin{align}
    d_{\cH}(\sqrt{n}\tilde f_2(\Phi\cX), \sqrt{n} \tilde f_2(\cZ)) \;&=\; d_{\cH}( f_2(\Phi\cX), f_2(\cZ))\;, \label{eqn:exp:neg:chi:sq:2d:d_H}
    \\ 
    n ( \Var[ \tilde f_2(\Phi\cX) ] - \Var [\tilde f_2(\cZ)] ) \;&=\; \Var[ f_2(\Phi\cX) ] - \Var [ f_2(\cZ)]\;. \label{eqn:exp:neg:chi:sq:2d:var}
\end{align}
To compute the noise stability terms for $g_2$, recall from the definition in \eqref{eqn:defn:beta_r_m} that
$$
 \overline \bW_i(\bw) \;\coloneqq\; \mfrac{1}{nk} \big( \msum_{i'=1}^{i-1} \msum_{j=1}^k \phi_{i'j} \bX_{i'} + \msum_{j=1}^k \bw_j + \msum_{i'=i+1}^n \msum_{j=1}^k \bZ_{i'j} \big) \in \R^2\;. 
$$
Denote its two coordinates by $\overline \bW_{i1}(\bw)$ and $\overline \bW_{i2}(\bw)$. Then by linearity of differentiation followed by triangle inequality of $\zeta_{i;m}$ from Lemma \ref{lem:zeta_properties},
\begin{align*}
    \nu_{r;m} (g_2) 
    \;=&\; 
    \mmax_{i \leq n} \zeta_{i;m}\big( \big\|\partial^r g_2\big(\overline\bW_i(\argdot) \big) \big\| \big) 
    \\
    \;=&\; 
    \mmax_{i \leq n} \zeta_{i;m}\big( \big\|\partial^r g\big(\overline\bW_{i1}(\argdot) \big) + \partial^r g\big(\overline\bW_{i2}(\argdot) \big) \big\| \big) 
    \\
    \;\leq&\; 
    \mmax_{i \leq n} \zeta_{i;m}\big( \big\|\partial^r g\big(\overline\bW_{i1}(\argdot) \big) \big\| \big) + \mmax_{i \leq n} \zeta_{i;m}\big( \big\| \partial^r g\big(\overline\bW_{i2}(\argdot) \big) \big\| \big)
    \\
    \;=:&\; \nu_{r;m}^{(1)}(g) + \nu_{r;m}^{(2)}(g)\;.
\end{align*}
Note that $\nu_{r;m}^{(1)}(g)$ is $\nu_{r;m}(g)$ defined with respect to the sets of 2d data $\Phi \cX$ and $\cZ$ but restricted to their first coordinates, and $\nu_{r;m}^{(2)}(g)$ with respect to the data restricted to their second. The model \eqref{eqn:exp:neg:chi:sq:simulation:general} ensures existence of all moments, so the same bounds computed for $\nu_{r;m}(g)$ in the 1d case in the proof of Proposition \ref{prop:exp_neg_chi_squared} directly apply to $\nu_{r;m}^{(1)}(g)$, $\nu_{r;m}^{(2)}(g)$ and consequently $\nu_{r;m}(g_2)$. Since we also have $c_x, c_Z = O(1)$, the bounds on \eqref{eqn:exp:neg:chi:sq:2d:d_H} and \eqref{eqn:exp:neg:chi:sq:2d:var} are $O(n^{-1/2})$, exactly the same as the 1d case. Applying Lemma \ref{lem:plugin:formal}(ii) proves the required convergences in (i) as $n \rightarrow \infty$ as before.

\vspace{.3em}

For (ii), by Lemma \ref{lem:compare_variance} and linearity of $\phi_{11}, \phi_{12}$,
\begin{align*}
    \Cov[\phi_{11} \bX_1, \phi_{12} \bX_2] \;&=\; \mean\Cov[ \phi_{11} \bX_1, \phi_{12} \bX_2 | \phi_{11}, \phi_{12}] 
    \\
    \;&=\; \mean[\phi_{11}] \Var[\bX_{11}] \mean[\phi_{12}] \;=\; \mfrac{(1+\rho) \sigma^2}{2} \bone_{2 \times 2}\;.
\end{align*}
Meanwhile, note that $\phi_{ij}\bX_i \overset{d}{=} \bX_i$, which implies that $\Var[\phi_{11}\bX_1] = \Var[\bX_1] = \sigma^2 \begin{psmallmatrix} 1 & \rho \\ \rho & 1\end{psmallmatrix}$ and $\mean[\phi_{11}\bX_1] = \mean[\bX_1] = \bzero$. Substituting these into the formula for moments of $\bZ_i$ from 
\eqref{eqn:defn:surrogates} 
gives the mean and variance required:
\begin{align*}
    &
    \mean \bZ_i \;=\; \bzero\;, 
    &&
    \Var \bZ_i \;=\; \sigma^2 \bI_k \otimes \begin{psmallmatrix} 1 & \rho \\ \rho & 1\end{psmallmatrix} + \mfrac{(1+\rho) \sigma^2}{2} (\bone_{k \times k} - \bI_k) \otimes \bone_{2 \times 2}\;.
\end{align*}
Similarly, substituting the calculations into the formula for moments of $\tilde \bZ_i$ from 
\eqref{eqn:defn:surrogates:no_aug} 
gives $\mean[\tilde \bZ_i] = \bzero$ and $\Var[ \tilde \bZ_i ] = \sigma^2 \bone_{k \times k} \otimes \begin{psmallmatrix} 1 & \rho \\ \rho & 1\end{psmallmatrix} $.

\vspace{.2em}

To compute (iii), first re-express the variance of $\bZ_i$ above as
\begin{align*}
    \Var \bZ_i  
    \;&=\; \sigma^2 \bI_k \otimes \begin{pmatrix} \frac{1-\rho}{2} & \frac{\rho-1}{2} \\ \frac{\rho-1}{2} & \frac{1-\rho}{2}\end{pmatrix} + \mfrac{(1+\rho) \sigma^2}{2} \bone_{2k \times 2k}
    \\
    \;&=\; \mfrac{(1-\rho) \sigma^2}{2} \bI_k \otimes \begin{psmallmatrix} 1 & -1 \\ -1 & 1 \end{psmallmatrix} +  \mfrac{(1+\rho) \sigma^2}{2} \bone_{2k \times 2k} = \sigma^2_- \bI_k \otimes \begin{psmallmatrix} 1 & -1 \\ -1 & 1 \end{psmallmatrix} +  \sigma^2_+ \bone_{2k \times 2k} \;,
\end{align*}
Notice that the structure in mean and variance of $\bZ_i$ allows us to rewrite it as a combination of simple 1d Gaussian random variables. Consider $\bU_{ij} \overset{i.i.d.}{\sim} \cN(0, \sigma_-^2)$ for $i \leq n, j \leq k$ and $\bV_i \overset{i.i.d.}{\sim} \cN(0, \sigma_+^2)$ independent of $\bU_{ij}$'s. Define the random vector in $\R^{2k}$ as
\begin{align*}
    \xi_i \;\coloneqq\; ( \bU_{i;1} + \bV_i \,,\, - \bU_{i;1} + \bV_i\,,\, \bU_{i;2} + \bV_i\,,\, - \bU_{i;2} + \bV_i\,,\, \ldots\,,\,  \bU_{i;k} + \bV_i\,,\, - \bU_{i;k} + \bV_i )^\top\;.
\end{align*}
Since $\mean \bZ_i = \mean \xi_i$ and $\Var \bZ_i = \Var \xi_i$, we have $\xi_i \overset{d}{=} \bZ_i$, which implies
\begin{align*}
    f_2(\cZ) \;\overset{d}{=}&\; f_2(\xi_1, \ldots, \xi_n) 
    \\
    \;=&\; 
    \exp\Big( - \Big( \mfrac{1}{\sqrt{n}k} \msum_{i,j} (\bU_{ij} + \bV_i) \Big)^2 \Big) 
    + \exp\Big( - \Big( \mfrac{1}{\sqrt{n}k} \msum_{i,j} (- \bU_{ij} + \bV_i) \Big)^2 \Big) 
    \\
    \;=:&\; \exp( - \bS_+) + \exp(- \bS_-)\;,
\end{align*}
and therefore
\begin{align}
    \Var[f_2(\cZ)] 
    \;&=\; 
    \Var[ \exp(-\bS_+)] + \Var[\exp(-\bS_-)] + 2\Cov[\exp(-\bS_+),\exp(-\bS_-)]\;. \label{eqn:exp:neg:chi:sq:2d:var:intermediate}
\end{align}
Notice that $\bS_+ \coloneqq \frac{1}{\sqrt{n}k} \sum_{i,j} (\bU_{ij} + \bV_i)$ and $\bS_-\coloneqq\frac{1}{\sqrt{n}k} \sum_{i,j} (-\bU_{ij} + \bV_i)$ are both normally distributed with mean $0$ and variance $\sigma^2_S \coloneqq \frac{\sigma^2_-}{k} + \sigma^2_+$. This means $\frac{\bS_+}{\sigma^2_S}$ and $\frac{\bS_-}{\sigma^2_S}$ are both chi-squared distributed with 1 degree of freedom, and the formula for moment generating function of chi-squared distribution again allows us to compute
\begin{align*}
    \mean [ \exp(- \bS_+) ] \;&=\;  \mean[ \exp(- \bS_-) ]  \;=\; (1+2\sigma^2_S)^{-1/2} \;,
    \\
    \Var [ \exp(- \bS_+) ] \;&=\;  \Var[ \exp(- \bS_-) ] \;=\; (1+4\sigma^2_S)^{-1/2} - (1+2\sigma^2_S)^{-1}\;.
\end{align*}
Moreover, writing $\bar \bU \coloneqq \mfrac{1}{\sqrt{n}k} \msum_{i,j} \bU_{ij} \sim \cN\big(0, \frac{\sigma_-^2}{k}\big)$ and $\bar \bV \coloneqq \mfrac{1}{\sqrt{n}} \msum_{i \leq n} \bV_i \sim \cN(0,\sigma_+^2)$, we have
\begin{align*}
    \mean[ \exp(- \bS_+ - \bS_-)] 
    \;=&\; 
   \mean[ \exp( - (\bar\bU + \bar\bV)^2 - (-\bar\bU + \bar\bV)^2 ]
   \\
   \;=&\;
   \mean[ \exp( - 2 \bar\bU^2 - 2 \bar\bV^2 )] \;=\;\mean[ \exp( - 2 \bar\bU^2)] \mean[\exp(- 2 \bar\bV^2 )]
   \\
   \;=&\; \Big( 1 + \mfrac{4 \sigma_-^2}{k} \Big)^{-1/2} (1+4\sigma_+^2)^{-1/2}\;,
\end{align*}
which implies
\begin{align*}
    \Cov[ \exp(-\bS_+), \exp(-\bS_-)] 
    \;&=\; 
    \mean[ \exp(-\bS_+ -\bS_-)] - \mean[ \exp(-\bS_+)] \mean[\exp( -\bS_-)] 
    \\
    \;&=\;
    \Big( 1 + \mfrac{4 \sigma_-^2}{k} \Big)^{-1/2} (1+4\sigma_+^2)^{-1/2} - (1+2\sigma^2_S)^{-1}\;.
\end{align*}
Substituting the calculations for variances and covariance into \eqref{eqn:exp:neg:chi:sq:2d:var:intermediate}, we obtain
\begin{align*}
    &\;\Var[f_2(\cZ)] 
    \\
    \;=&\; 
    2\big( (1+4\sigma^2_S)^{-1/2} - (1+2\sigma^2_S)^{-1} \big)
    + 2\Big( \Big( 1 + \mfrac{4 \sigma_-^2}{k} \Big)^{-1/2} (1+4\sigma_+^2)^{-1/2} - (1+2\sigma^2_S)^{-1} \Big)
    \\
    \;=&\;
    2(1+4\sigma^2_S)^{-1/2} + 2 \Big( 1 + \mfrac{4 \sigma_-^2}{k} \Big)^{-1/2} (1+4\sigma_+^2)^{-1/2} - 4 (1+2\sigma^2_S)^{-1} 
    \\
    \;=&\;
    2\Big(1+ \mfrac{4 \sigma_-^2}{k} + 4\sigma_+^2 \Big)^{-1/2} + 2 \Big( 1 + \mfrac{4 \sigma_-^2}{k} \Big)^{-1/2} (1+4\sigma_+^2)^{-1/2} - 4 (1+\mfrac{2 \sigma_-^2}{k} + 2\sigma_+^2)^{-1}\;,
\end{align*}
which is the required formula. 
\end{proof}

\subsection{Ridge regression} \label{appendix:ridge}

\vspace{.2em}

In this section, it is useful to define the function $g_B:\M^d \times \R^{d \times b} \rightarrow \R^{d \times b}$:
\begin{align*}
    g_B(\Sigma,A) \;\coloneqq\; \tilde \Sigma^{-1} A \;, \tagaligneq \label{eqn:g_B}
\end{align*}
which allows the ridge estimator to be written as
\begin{align*}
    \hat B^{\Phi\cX} \;\coloneqq\;\hat B(\Phi\cX) \;=\; g_B \Big( \mfrac{1}{nk}\msum_{i,j} (\pi_{ij} \bV_i) (\pi_{ij} \bV_{i})^\top,   \mfrac{1}{nk}\msum_{i,j} (\pi_{ij} \bV_i) (\tau_{ij} \bY_{i})^\top \Big)\;.
\end{align*}
Similarly, we can use $g_B$ to rewrite the estimator with surrogate variables considered in 
\Cref{thm:main} 
and the truncated first-order Taylor version in Lemma \ref{lem:plugin:formal}:
\begin{align*}
    \hat B^Z
    \;&\coloneqq\;
    g_B\big(\mfrac{1}{nk} \msum_{i,j} \bZ_{ij}\big)
    \;
    &\text{and}&&
    \hat B^T
    \;&\coloneqq\;
    g_B(\mu) + \partial g_B(\mu) \big(\mfrac{1}{nk} \msum_{i,j} \bZ_{ij} - \mu \big)\;,
\end{align*}
where $\mu \coloneqq (\mu_1, \mu_2) \coloneqq \big( \mean[(\pi_{11}\bV_1)(\pi_{11}\bV_1)^\top ],  \mean[(\pi_{11}\bV_1)(\tau_{11}\bV_1)^\top ] \big)$. Similarly, consider the function $g_R: \M^d \times \R^{d\times b}  \rightarrow \R$ defined by
\begin{equation} \label{eqn:defn:g_R}
    g_R(\Sigma, A) \;\coloneqq\; \mean[\|\bY_{new} - (\tilde \Sigma^{-1} A)^\top  \bV_{new} \|^2_2]\;.
\end{equation}
This allows us to write the risk as 
$$
R^{\Phi\cX} \;=\; g_R \Big( \mfrac{1}{nk}\msum_{i,j} (\pi_{ij} \bV_i) (\pi_{ij} \bV_{i})^\top,   \mfrac{1}{nk}\msum_{i,j} (\pi_{ij} \bV_i) (\tau_{ij} \bY_{i})^\top \Big)\;,
$$ 
while the estimator considered in 
\Cref{thm:main} 
and the first-order Taylor version in Lemma \ref{lem:plugin:formal} become
\begin{align*}
    R^Z
    \;&\coloneqq\;
    g_R\big(\mfrac{1}{nk} \msum_{i,j} \bZ_{ij}\big)
    \;,\;
    &\text{and}&&
    R^T
    \;&\coloneqq\;
    g_R(\mu) + \partial g_R(\mu) \big(\mfrac{1}{nk} \msum_{i,j} \bZ_{ij} - \mu \big)\;,
\end{align*}
In this section, we first prove
\begin{proplist}
    \item the convergence of $\hat B^{\Phi\cX}$ to $\hat B^Z$ and $\hat B^T$, and the convergence of $R^{\Phi\cX}$ to $R^Z$ and $R^T$, with each convergence rate specified, and
    \item existence of surrogate variables satisfying those convergences.
\end{proplist}

\vspace{.3em}

The proof for (i) follows an argument analogous to previous examples: we compute derivatives of the estimator of interest, and apply variants of 
\Cref{thm:main} 
to obtain convergences. The results are collected in Lemma \ref{lem:ridge:regression:extended} in Appendix \ref{appendix:ridge:proof_conv}. The comment on different convergence rates in 
\Cref{remark:ridge} 
is also clear from Lemma \ref{lem:ridge:regression:extended}. 

\vspace{.3em} 

(ii) is of concern in this setup because the surrogate variables can no longer be Gaussian. Appendix \ref{appendix:ridge:exist_surrogate} states one possible choice from an approximate maximum entropy principle. Combining (i) and (ii) gives the statement in 
\Cref{prop:ridge:regression}.

\vspace{.3em}

Finally, Appendix \ref{appendix:ridge:proof_toy} focuses on the toy model in 
\eqref{eqn:ridge:toy:model}. 
We prove 
\Cref{lem:ridge:toy}, 
which discusses the non-monotonicity of variance of risk as a function of data variance. We also prove Lemma \ref{lem:ridge:toy:non_convergence}, a formal statement of 
\Cref{remark:ridge} 
that $\Var[R^{\Phi\cX}]$ does not converge to $\Var[R^T]$ for sufficiently high dimensions under a toy model.

\vspace{-.3em}

\subsubsection{Proof for convergence of variance and weak convergence} \label{appendix:ridge:proof_conv}

\vspace{.2em}

\begin{lemma} \label{lem:ridge:regression:extended} Assume that $\mmax_{l \leq d} \mmax\{ (\pi_{11} \bV_{1})_l, (\tau_{11} \bY_{1})_l \}$ is a.s.~bounded by $C \tau$ for some $\tau$ to be specified and some absolute constant $C>0$, and that $b=O(d)$. Then, for any i.i.d.\,surrogate variables $\{\bZ_i\}_{i\leq n}$ taking values in $(\M^d \times \R^{d\times b})^k$ matching the first moments of $\Phi_1\bX_1$ with all coordinates uniformly bounded by $C' \tau^2$ a.s.~for some absolute constant $C' > 0$, we have:
\begin{proplist}
    \item assuming $\tau = O(1)$ and fixing $r \leq d$, $s \leq b$, then the $(r,s)$-the coordinate of $\hat B^{\Phi\cX}$ satisfies
\begin{align*}
    d_{\cH}\big( \sqrt{n} \big(\hat B^{\Phi\cX}\big)_{r,s}, \sqrt{n} \big(\hat B^T\big)_{r,s} \big)
    \;&=\; O(n^{-1/2}d^9)\;,
    \\
    d_{\cH}\big( \sqrt{n} \big(\hat B^{\Phi\cX}\big)_{r,s}, \sqrt{n} \big(\hat B^Z\big)_{r,s} \big)
    \;&=\; O(n^{-1/2}d^9)\;;
\end{align*}
    \item assuming $\tau = O(d^{-1/2} (\log d)^c)$ for some absolute constant $c > 0$, then $\hat B^{\Phi\cX}$ satisfies
    \begin{align*}
    n \| \Var[\hat B^{\Phi \cX}] - \Var[\hat B^T] \| 
    \;&=\; O\big( (n^{-1/2} d^7 + n^{-1} d^8)(\log d)^{12c} \big)\;,
    \\
    n \| \Var[\hat B^{\Phi \cX}] - \Var[\hat B^Z] \| 
    \;&=\; O\big( (n^{-1} d^7) (\log d)^{10c} \big)\;;
    \end{align*}
    \item assuming $\tau = O(d^{-1/2} (\log d)^c)$ for some absolute constant $c > 0$, then $R^{\Phi\cX}$ satisfies
    \begin{align*}
        &\,
        d_{\cH}( \sqrt{n} R^{\Phi \cX}, \sqrt{n} R^T ) \;=\; O( (n^{-1/2} d^9) (\log d)^{24c}) \;,
        \\
        &\,
        d_{\cH}( \sqrt{n} R^{\Phi \cX}, \sqrt{n} R^{\cZ} ) \;=\;  O( (n^{-1/2} d^9) (\log d)^{24c}) \;,
        \\
        &
        n ( \Var[R^{\Phi \cX}] - \Var[R^T] ) \;=\; O( (n^{-1/2} d^7 + n^{-1} d^8) (\log d)^{20c} )\;,
        \\
        &
        n ( \Var[R^{\Phi \cX}] - \Var[R^Z] ) \;=\; O(n^{-1} d^7 (\log d)^{18c})\;.
    \end{align*}
\end{proplist}
\end{lemma}

\begin{remark} In the statement of weak convergence of the estimator $\hat B^{\Phi\cX}$, we only consider convergence of one coordinate of $\hat B^{\Phi\cX}$ since we allow dimensions $d, b$ to grow with $n$; this setting was discussed in more details in Lemma \ref{lem:sufficient:cond:vary:q}. The assumption $\tau = O(1)$ for (i) is such that the coordinates we are studying do not go to zero as $n$ grows, while the assumption $\tau=O(d^{-1/2})$ for (ii) and (iii) is such that $\|\pi_{11}\bV_1\|$ and $\|\tau_{11}\bY_1\|$ are $O(1)$ as $n$ grows, which keeps $\| \hat B^{\Phi\cX} \|$ and $R^{\Phi\cX}$ bounded.
\end{remark}

\begin{remark} The difference between the convergence rate of $\Var[R^{\Phi\cX}]$ towards $\Var[R^Z]$ and that towards $\Var[R^T]$ is clear in the additional factor $(n^{1/2} + d)$ in Lemma \ref{lem:ridge:regression:extended}(iii). If we take $d$ to be $\Theta(n^\alpha)$ for $\frac{1}{14} < \alpha < \frac{1}{7}$, we are guaranteed convergence of $\Var[R^{\Phi\cX}]$ to $\Var[R^Z]$ but not necessarily convergence of $\Var[R^{\Phi\cX}]$ to $\Var[R^T]$. Note that the bounds here are not necessarily tight in terms of dimensions, and we discuss this difference in convergence rate in more details in Appendix \ref{appendix:ridge:slower_conv_taylor}.
\end{remark}

\begin{proof}[Proof of Lemma \ref{lem:ridge:regression:extended}(i)] We first prove the weak convergence statements for $(\hat B^{\Phi\cX})_{r,s}$. Let $\be_r$ be the $r$-th basis vector of $\R^d$ and $\bo_s$ be the $s$-th basis vector of $\R^b$. We define the function $ g_{B;rs}: \M^d \times \R^{d \times b} \rightarrow \R$ as
\begin{align*}
    g_{B;rs}(\Sigma,A) \;\coloneqq\; \be_r^\top g_B(\Sigma, A) \bo_s\;=\; \be_r^\top \tilde \Sigma^{-1} A \bo_s \;,
\end{align*}
i.e. the $(r,s)$-th coordinate of $g_B$. The $(r,s)$-th coordinate of $\hat B^{\Phi\cX}$, $\hat B^Z$ and $\hat B^T$ can then be expressed in terms of $g_{B;rs}$ similar to before:
\begin{align*}
    \big(\hat B^{\Phi\cX}\big)_{r,s} 
    \;&=\; 
    g_{B;rs} \Big( \mfrac{1}{nk}\msum_{i,j} (\pi_{ij} \bV_i) (\pi_{ij} \bV_{i})^\top,   \mfrac{1}{nk}\msum_{i,j} (\pi_{ij} \bV_i) (\tau_{ij} \bY_{i})^\top \Big)\;,
    \\
    \big(\hat B^Z\big)_{r,s}
    \;&=\;
    g_{B;rs}\big(\mfrac{1}{nk} \msum_{i,j} \bZ_{ij}\big)
    \;,\;
    \big(\hat B^T\big)_{r,s}
    \;=\;
    g_{B;rs}(\mu) + \partial g_{B;rs}(\mu) \big(\mfrac{1}{nk} \msum_{i,j} \bZ_{ij} - \mu \big)\;.
\end{align*}
To obtain weak convergence of $(\hat B^{\Phi\cX})_{r,s}$ to $(\hat B^Z)_{r,s}$ and $(\hat B^T)_{r,s}$, it suffices to apply the result for the plug-in estimates from Lemma \ref{lem:plugin:formal} with $\delta=0$ to the function $g_{B;rs}$ with respect to the transformed data  $\phi_{ij}\bX_i^* \coloneqq \big(
    (\pi_{ij} \bV_i) (\pi_{ij} \bV_{i})^\top,
    (\pi_{ij} \bV_{i}) (\tau_{11} \bY_{i})^\top
    \big)$.
    
\vspace{.3em}

As before, we start with computing the partial derivatives of $g_{B;rs}(\Sigma, A)$, which can be expressed using $\tilde \Sigma \coloneqq \Sigma + \lambda \bI_d$ and $A$ as:
\begin{align*}
    g_{B;rs}(\Sigma, A)\;&=\; \be_r^\top \tilde \Sigma^{-1} A \bo_s,
    \\
    \mfrac{\partial g_{B;rs}(\Sigma,A)}{\partial \Sigma_{r_1s_1}} \;&=\; - \be_r^\top \tilde \Sigma^{-1} \be_{r_1} \be_{s_1}^\top  \tilde \Sigma^{-1} A \bo_s,
    \quad
    \mfrac{\partial g_{B;rs}(\Sigma,A)}{\partial A_{r_1s_1}} \;=\; \be_r^\top \tilde \Sigma^{-1} \be_{r_1} \ind_{\{s=s_1\}} ,
    \\
    \mfrac{\partial^2 g_{B;rs}(\Sigma,A)}{\partial \Sigma_{r_1s_1} \partial \Sigma_{r_2s_2}}
    \;&=\;
    \msum_{ l_1, l_2 \in \{1,2\};\; l_1 \neq l_2} \be_r
    \tilde \Sigma^{-1}
    \be_{r_{l_1}} \be_{s_{l_1}}^\top 
    \tilde \Sigma^{-1}
    \be_{r_{l_2}} \be_{s_{l_2}}^\top
    \tilde \Sigma^{-1}
    A \bo_s,
    \\
    \mfrac{\partial^2 g_{B;rs}(\Sigma,A)}{\partial \Sigma_{r_1s_1} \partial A_{r_2s_2}}
    \;&=\;
    - 
    \be_r^\top
    \,\tilde \Sigma^{-1}\,
    \be_{r_1} \be_{s_1}^\top
    \,\tilde \Sigma^{-1}\,
    \be_{r_2} \ind_{\{s=s_2\}},
    \qquad \,
    \mfrac{\partial^2 g_{B;rs}(\Sigma,A)}{\partial A_{r_1s_1} \partial A_{r_2s_2}}
    \;=\;
    \bzero,
    \\
    \mfrac{\partial^3 g_{B;rs}(\Sigma,A)}{\partial \Sigma_{r_1s_1} \partial \Sigma_{r_2s_2} \partial \Sigma_{r_3s_3}}
    \;&=\;
    - \msum_{ \substack{l_1, l_2, l_3 \in \{1,2,3\} \\ l_1, l_2, l_3 \text{ distinct}} }
    \be_r^\top \tilde \Sigma^{-1}
    \be_{r_{l_1}} \be_{s_{l_1}}^\top 
    \tilde \Sigma^{-1}
    \be_{r_{l_2}} \be_{s_{l_2}}^\top
    \tilde \Sigma^{-1}
    \be_{r_{l_3}} \be_{s_{l_3}}^\top
    \tilde \Sigma^{-1}
    A \bo_s,
    \\
    \mfrac{\partial^3 g_{B;rs}(\Sigma,A)}{\partial \Sigma_{r_1s_1} \partial \Sigma_{r_2s_2} \partial A_{r_3s_3}}
    \;&=\;
    \msum_{ l_1, l_2 \in \{1,2\};\; l_1 \neq l_2}
    \be_r^\top
    \tilde \Sigma^{-1}
    \be_{r_{l_1}} \be_{s_{l_1}}^\top 
    \tilde \Sigma^{-1}
    \be_{r_{l_2}} \be_{s_{l_2}}^\top
    \tilde \Sigma^{-1}
    \be_{r_3} \ind_{\{s=s_3\}}. \tagaligneq \label{eq:ridge:derivatives}
\end{align*}
To bound the norm of the derivatives, it is useful to have controls over the norms of $\tilde \Sigma^{-1}$ and $A$. Suppose the coordinates of $A$ are uniformly bounded by $C \tau^2$ for some absolute constant $C > 0$, which is the case when we compute the derivatives in $\nu_{r;m}$. Then since $b=O(d)$, we have
\begin{align*}
    \| A \|_{op} \leq \| A \| = O(d \tau^2),\;\;
    \| A \bo_s \| = O( d^{1/2} \tau^2),\;\;
    \| \tilde \Sigma \|_{op} = \|\Sigma + \lambda \bI_d \|_{op} = \mfrac{1}{\sigma_1 + \lambda} = O(1),
\end{align*}
where $\sigma_1 \geq 0$ is the smallest eigenvalue of the positive semi-definite matrix $A$. We also note that for any matrix $M \in \R^{n_1 \times n_2}$ and vectors $\bu\in \R^{n_2}, \bv \in \R^{n_3}$,
\begin{align*}
    &
    \| M \bu \| \leq \| M \|_{op} \| \bu \|\;,
    &&
    \| \bu \bv^\top \|_{op} \leq \| \bu \| \| \bv \|\;.
\end{align*}
Making use of these bounds, we can bound the norms of partial derivatives of $g$ as follows:
\begin{align*}
    \Big\| \mfrac{\partial g_{B;rs}(\Sigma,A)}{\partial \Sigma_{r_1s_1}} \Big\| 
    \;&\leq\; 
    \| \tilde \Sigma^{-1} \be_{r_1} \| \| \be_{s_1}^\top  \tilde \Sigma^{-1} A \| \;\leq\; \|\Sigma^{-1} \|_{op}^2 \|A\|_{op} \;=\; O(d \tau^2)\;.
\end{align*}
We can perform a similar argument for the remaining derivatives. It suffices to count the number of $A$ in each expression and use the bound $\| A \|_{op} \leq \| A \| = O(d \tau^2)$:
\begin{align*}
    \Big\| \mfrac{\partial^2 g_{B;rs}(\Sigma,A)}{\partial A_{r_1s_1} \partial A_{r_2s_2}} \Big\|, \;
    \Big\|  \mfrac{\partial^3 g_{B;rs}(\Sigma,A)}{\partial \Sigma_{r_1s_1} \partial A_{r_2s_2} \partial M_{r_3s_3}} \Big\|,\;
    \Big\|  \mfrac{\partial^3 g_{B;rs}(\Sigma,A)}{\partial A_{r_1s_1} \partial A_{r_2s_2} \partial A_{r_3s_3}} \Big\|
    \;&=\; 0,
    \\
    \Big\| \mfrac{\partial g_{B;rs}(\Sigma,A)}{\partial A_{rs}} \Big\|,\; 
    \Big\| \mfrac{\partial^2 g_{B;rs}(\Sigma,A)}{\partial \Sigma_{r_1s_1} \partial A_{r_2s_2}} \Big\|, \;
    \Big\|  \mfrac{\partial^3 g_{B;rs}(\Sigma,A)}{\partial \Sigma_{r_1s_1} \partial \Sigma_{r_2s_2} \partial A_{r_3s_3}} \Big\|
    \;&=\; O(1),
    \\
    \| g_{B;rs}(\Sigma,A) \|,\;
    \Big\| \mfrac{\partial g_{B;rs}(\Sigma,A)}{\partial \Sigma_{rs}} \Big\|,\; 
    \Big\| \mfrac{\partial^2 g_{B;rs}(\Sigma,A)}{\partial \Sigma_{r_1s_1} \partial \Sigma_{r_2s_2}} \Big\|,\;
    \Big\|  \mfrac{\partial^3 g_{B;rs}(\Sigma,A)}{\partial \Sigma_{r_1s_1} \partial \Sigma_{r_2s_2} \partial \Sigma_{r_3s_3}} \Big\| 
    \;&=\; O(d \tau^2)\;.
\end{align*}
This implies
\begin{align*}
    \big\| \partial g_{B;rs}(\Sigma, A) \big\|
    \;&=\;
    \sqrt{ 
    \msum_{r_1, s_1=1}^d \Big\|  \mfrac{\partial g_{B;rs}(\Sigma, A) }{\partial \Sigma_{r_1s_1}} \Big\|^2 
    +
     \msum_{r_1=1}^d \msum_{s_1=1}^b \Big\|  \mfrac{\partial g_{B;rs}(\Sigma, A)}{\partial A_{r_1s_1}} \Big\|^2
    }
    \\
    \;&=\; O(d^2 \tau^2 + d )\;, 
    \\
    \big\| \partial^2 g_{B;rs}(\Sigma, A)  \big\|
    \;&=\;
    \sqrt{ 
    \sum_{\substack{r_1, r_2, \\ s_1, s_2=1}}^d \Big\|  \mfrac{\partial^2 g_{B;rs}(\Sigma, A)}{\partial \Sigma_{r_1s_1} \partial \Sigma_{r_2s_2}} \Big\|^2 
    +
    \sum_{r_1, s_1, r_2=1}^d \sum_{s_2=1}^b \Big\|  \mfrac{\partial^2 g_{B;rs}(\Sigma, A)}{\partial \Sigma_{r_1s_1} \partial A_{r_2s_2}} \Big\|^2
    } 
    \\
    \;&=\; O(d^3 \tau^2 + d^2 ) , 
    \\
    \big\| \partial^3 g_{B;rs}(\Sigma, A) \big\|
    &=
    \sqrt{
    \sum_{\substack{r_1, r_2, r_3,\\ s_1, s_2, s_3=1}}^d \Big\|  \mfrac{\partial^3 g(\Sigma, A)}{\partial \Sigma_{r_1s_1} \partial \Sigma_{r_2s_2} \partial \Sigma_{r_3s_3}} \Big\|^2 
    +
    \sum_{\substack{r_1, r_2, r_3,\\ s_1, s_2=1}}^d \sum_{s_3=1}^b \Big\|  \mfrac{\partial^3 g(\Sigma, A)}{\partial \Sigma_{r_1s_1} \partial \Sigma_{r_2s_2} \partial A_{r_3s_3}} \Big\|^2 
    }\\
    \;&=\;  O(d^4 \tau^2 + d^3 ) \;.
\end{align*}
Recall that the noise stability terms in Lemma \ref{lem:plugin:formal} are defined by, for $\delta=0$,
\begin{align*}
  \kappa_{t;m}(g) 
  = 
  \sum_{l \leq q} \big\|{\msup_{\bw\in [\bzero, \bar\bX]}}
  \big\| \partial^t g_l\big( \mu + \bw \big) \big\| \big\|_{L_m}, 
  \;
  \nu_{t;m}(g) 
  = 
  \sum_{l\leq q} \max_{i \leq n} \zeta_{i;m}\big( \big\|\partial^t g_l\big(\overline\bW_i(\argdot) \big) \big\| \big),
\end{align*}
where $q = 1$ in the case of $g_{B;rs}$, and the moment terms are defined by
\begin{align*}
    \bar c_m 
    \;&=\;
    \Big( 
    \msum_{l=1}^{d^2+db}  \max\big\{ n^{\frac{2}{m}-1} \big\| \mfrac{1}{k} \msum_{j=1}^k [\phi_{1j}\bX^*_1 - \mu]_l \big\|^2_{L_m} 
    ,\;
    \big\| \mfrac{1}{k} \msum_{j=1}^k [\phi_{1j}\bX^*_1 - \mu]_l \big\|^2_{L_2} 
    \big\} 
    \Big)^{1/2}\;,
    \\
    c_X\;&=\;\mfrac{1}{6}\sqrt{ \mean[\|\phi_{11}\bX^*_1\|^6 }\;,
    \quad
    c_Z\;=\;\mfrac{1}{6}\sqrt{ \mean\Bigl[ \Bigl(\mfrac{|Z_{111}|^2+\ldots+|Z_{1k(d^2+db)}|^2}{k}\Bigr)^3\Bigr] }\;.
\end{align*}
By the bounds on the derivatives of $g_{B;rs}$ from above, we get
\begin{align*}
    &
    \kappa_{0;m}( g_{B;rs} ), \; \nu_{0;m}( g_{B;rs} ) \;=\; O(d \tau^2)\;, 
    &&
    \kappa_{1;m}( g_{B;rs} ), \; \nu_{1;m}( g_{B;rs} ) \;=\; O(d^2 \tau^2 + d)\;, 
    \\
    &
    \kappa_{2;m}( g_{B;rs} ), \; \nu_{2;m}( g_{B;rs} ) \;=\; O(d^3 \tau^2 + d^2)\;,    
    &&
    \kappa_{3;m}( g_{B;rs} ), \; \nu_{3;m}( g_{B;rs} ) \;=\; O(d^4 \tau^2 + d^3)\;,
\end{align*}
and since the coordinates of $\phi_{11}\bX_1$ and $\bZ_1$ are uniformly bounded by $C'' \tau^2$ for $C'' =\max\{C,C'\}$ almost surely,  we get that
\begin{align*}
    &
    \bar c_m \;=\; O(d \tau^2 )\;,
    &&
    c_X, c_Z \;=\; O(d^3 \tau^6)\;.
\end{align*}
Applying Lemma \ref{lem:plugin:formal}(i) to $g_{B;rs}$ with $\delta=0$ and the assumption $\tau=O(1)$ then gives 
\begin{align*}
    d_{\cH}\big( \sqrt{n} \big(\hat B^{\Phi\cX}\big)_{r,s}, \sqrt{n} \big(\hat B^T\big)_{r,s} \big) 
    \;&=\; 
    O\big(n^{-1/2} \kappa_{2;3}(g_{B;rs}) \, \bar c_3^2 + n^{-1/2}  \kappa_{1;1}(g_{B;rs})^3 (c_X + c_Z))\big)
    \\
    \;&=\;
    O(n^{-1/2} d^5 + n^{-1/2} d^6 d^3 ) \;=\; O(n^{-1/2} d^9),
\end{align*}
and applying Lemma \ref{lem:plugin:formal}(ii) with $\delta$ set to $0$ gives
\begin{align*}
    &\;d_{\cH}\big( \sqrt{n} \big(\hat B^{\Phi\cX}\big)_{r,s}, \sqrt{n} \big(\hat B^Z\big)_{r,s} \big)
    \\
    \;=&\; 
    O\big(\big(n^{-1/2} \nu_{1;6}(g_{B;rs})^3  + n^{-1} \nu_{1;4}(g_{B;rs}) \nu_{2;4}(g_{B;rs}) + n^{-3/2} \nu_{3;2}(g_{B;rs}) \big) (c_X + c_Z)\big)
    \\
    \;=&\; 
    O\big( \big(n^{-1/2} d^6 + n^{-1} d^5 + n^{-3/2} d^4 \big) d^3  \big) \;=\; O(n^{-1/2}d^9)\;.
\end{align*}
These are the desired bounds concerning weak convergence of $\big(\hat B^{\Phi\cX}\big)_{r,s}$. $d_H$ indeed metrizes weak convergence here, since $\big(\hat B^{\Phi\cX}\big)_{r,s} \in \R$ and 
\Cref{lem:d_H} 
applies.
\end{proof}

\begin{proof}[Proof of Lemma \ref{lem:ridge:regression:extended}(ii)]
For convergence of variance of $\hat B^{\Phi\cX}$, we need to apply Lemma \ref{lem:plugin:formal} to $g_B$ instead of $g_{B;rs}$. Notice that the noise stability terms of $g_B$ can be computed in terms of those for $g_{B;rs}$ already computed in the proof of (i):
\begin{align*}
    &
    \kappa_{t;m}(g_B) \;=\; \msum_{r=1}^d \msum_{s=1}^b \kappa_{t;m}(g_{B;rs})\;,
    &&
    \nu_{t;m}(g_B) \;=\; \msum_{r=1}^d \msum_{s=1}^b \nu_{t;m}(g_{B;rs})\;.
\end{align*}
This suggests that
\begin{align*}
    &
    \kappa_{0;m}( g_B ), \; \nu_{0;m}( g_B ) \;=\; O(d^3 \tau^2)\;, 
    &&
    \kappa_{1;m}( g_B ), \; \nu_{1;m}( g_B ) \;=\; O(d^4 \tau^2 + d^3)\;, 
    \\
    &
    \kappa_{2;m}( g_B ), \; \nu_{2;m}( g_B ) \;=\; O(d^5 \tau^2 + d^4)\;,    
    &&
    \kappa_{3;m}( g_B ), \; \nu_{3;m}( g_B ) \;=\; O(d^6 \tau^2 + d^5)\;,
\end{align*}
The moment terms are bounded as before: $\bar c_m = O(d \tau^2 )$ and $c_X, c_Z = O(d^3 \tau^6)$. Applying Lemma \ref{lem:plugin:formal} with $\delta=0$ and the assumption $\tau = O(d^{-1/2} (\log d)^c)$ gives
\begin{align*}
    n \| \Var[\hat B^{\Phi \cX}] - \Var[\hat B^T] \| 
    \;&=\; 
    O\big( n^{-1/2} \kappa_{1;1}(g_B) \kappa_{2;4}(g_B) \bar c_4^3 + n^{-1} \kappa_{2;6}(g_B) \kappa_{2;6}(g_B) \, \bar c_6^4 \big)
    \\
    \;&=\; O\big( (n^{-1/2} d^7 + n^{-1} d^8) (\log d)^{12c} \big)\;,
    \\
    n \| \Var[\hat B^{\Phi \cX}] - \Var[\hat B^Z] \| 
    \;&=\;
    O\big(n^{-1} ( \nu_{0;4} (g_B) \nu_{3;4} (g_B) + \nu_{1;4} (g_B) \nu_{2;4} (g_B) ) (c_X + c_Z) \big)
    \\
    \;&=\; O\big( n^{-1} d^7 (\log d)^{10c} \big)\;,
\end{align*}
which are the desired bounds for convergence of variance of $\hat B^{\Phi\cX}$.
\end{proof}

\begin{proof}[Proof of Lemma \ref{lem:ridge:regression:extended}(iii)] We seek to apply Lemma \ref{lem:plugin:formal} to $g_R$. Define
\begin{align*}
    c^Y \;\coloneqq\; \mean[\|\bY_{new}\|^2_2]\;,
    \quad
    C^{VY}_{rs} \;\coloneqq\; \big( \mean[ \bV_{new} \bY_{new}^\top  ] \big)_{rs}\;,
    \quad
    C^V_{rs} \;\coloneqq\; \big(\mean[ \bV_{new} \bV_{new}^\top  ]\big)_{rs}\;,
\end{align*}
This allows us to rewrite $g_R$ as
\begin{align*}
    &\;g_R(\Sigma, A) 
    \;=\; \mean[\|\bY_{new} - g_B(\Sigma, A)^\top  \bV_{new}\|^2_2]
    \\
    \;=&\; \mean[\|\bY_{new}\|^2_2] - 2 \Tr\big( \mean[ \bV_{new} \bY_{new}^\top  ] g_B(\Sigma, A)^\top \big) + \Tr\big( \mean[ \bV_{new} \bV_{new}^\top  ] g_B(\Sigma, A) g_B(\Sigma, A)^\top \big)
    \\
    \;=&\; c^Y - 2 \msum_{r=1}^d \msum_{s=1}^b C^{VY}_{rs} \, g_{B;rs}(\Sigma, A) + \msum_{rs,t=1}^d C^V_{rs}  g_{B;rt}(\Sigma, A)  g_{B;ts}(\Sigma, A)\;.
\end{align*}
As before, we first consider expressing derivatives of $g_R$ in terms of those of $g_{B;rs}$. Omitting the $(\Sigma, A)$-dependence temporarily, we get
\begin{align*}
\partial g_R
\;=&\; 
-2 \msum_{r=1}^d \msum_{s=1}^b C^{VY}_{rs} \, \partial g_{B;rs}
+ \msum_{rs,t=1}^d C^V_{rs} \, \big( \partial g_{B;rt} g_{B;ts} +  g_{B;rt} \partial g_{B;ts} \big), 
\\
\partial^2 g_R
\;=&\; 
-2 \msum_{r=1}^d \msum_{s=1}^b C^{VY}_{rs} \, \partial^2 g_{B;rs}
\\
&\;+ \msum_{rs,t=1}^d C^V_{rs} \, \big( \partial^2 g_{B;rt} g_{B;ts} + 2\partial g_{B;rt} \partial g_{B;ts} +  g_{B;rt} \partial^2 g_{B;ts} \big), 
\\
\partial^3 g_R
\;=&\; 
-2 \msum_{r=1}^d \msum_{s=1}^b C^{VY}_{rs} \, \partial^3 g_{B;rs}
\\
&\;+ \msum_{rs,t=1}^d C^V_{rs} \, \big( \partial^3 g_{B;rt} g_{B;ts} + 3\partial^2 g_{B;rt} \partial g_{B;ts} + 3\partial^ g_{B;rt} \partial^2 g_{B;ts} + g_{B;rt} \partial^3 g_{B;ts} \big)\;.
\end{align*}
Since the noise stability terms of $g_R$ are given by
\begin{align*}
  \kappa_{t;m}(g_R) 
  \;=\; 
  \big\|{\msup_{\bw\in [\bzero, \bar\bX]}}
  \big\| \partial^t g_R\big( \mu + \bw \big) \big\| \big\|_{L_m}
  , 
  \nu_{t;m}(g_R) 
  \;=\; 
  \mmax_{i \leq n}
  \zeta_{i;m}\big( \big\|\partial^t g_R\big(\overline\bW_i(\argdot) \big) \big\| \big),
\end{align*}
they can be bounded in terms of those of $g_{B;rs}$ computed in the proof of (i). With the assumption $\tau=O(d^{-1/2} (\log d)^c)$, the noise stability terms of $g_{B;rs}$ become
\begin{align*}
    &
    \kappa_{0;m}( g_{B;rs} ), \; \nu_{0;m}( g_{B;rs} ) \;=\; O((\log d)^{2c})\;, 
    &&
    \kappa_{1;m}( g_{B;rs} ), \; \nu_{1;m}( g_{B;rs} ) \;=\; O(d (\log d)^{2c} )\;, 
    \\
    &
    \kappa_{2;m}( g_{B;rs} ), \; \nu_{2;m}( g_{B;rs} ) \;=\; O(d^2 (\log d)^{2c})\;,    
    &&
    \kappa_{3;m}( g_{B;rs} ), \; \nu_{3;m}( g_{B;rs} ) \;=\; O(d^3 (\log d)^{2c})\;.
\end{align*}
Also note that $c^Y = O(d \tau^2) = O((\log d)^{2c})$ and $C^{VY}_{r,s}, C^V_{r,s} = O(\tau^2) = O(d^{-1} (\log d)^{2c})$ by assumption. Then, by a triangle inequality followed by H\"older's inequality,
\begin{align*}
    \kappa_{0;m}(g_R)
    \;\leq&\;
    c^Y + 2  \msum_{r=1}^d \msum_{s=1}^b C^{VY}_{r,s}  \kappa_{0;m}(  g_{B;rs} ) +  \msum_{r,s,t=1}^d  C^V_{r,s} \kappa_{0;m}( g_{B;rt} g_{B;ts} )
    \\
    \;\leq&\; c^Y + 2 \msum_{r=1}^d \msum_{s=1}^b C^{VY}_{r,s} \kappa_{0;m}(g_{B;rs}) + \msum_{r,s,t=1}^d C^V_{r,s}  \kappa_{0;2m}(g_{B;rt}) \kappa_{0;2m}(g_{B;ts})
    \\
    \;=&\; O( (1 + d + d^2) (\log d)^{6c} )\;=\; O(d^2 (\log d)^{6c})\;.
\end{align*}
Similarly, by triangle inequality and H\"older's inequality of $\zeta_{i;m}$ in Lemma \ref{lem:zeta_properties}, 
\begin{align*}
    \nu_{0;m}(g_R) 
    \;\leq&\; c^Y + 2 \msum_{r=1}^d \msum_{s=1}^b C^{VY}_{r,s} \nu_{0;m}(g_{B;rs}) + \msum_{r,s,t=1}^d C^V_{r,s}  \nu_{0;2m}(g_{B;rt}) \nu_{0;2m}(g_{B;ts})
    \\
    \;=&\; O( (1 + d + d^2) (\log d)^{6c} )\;=\; O(d^2 (\log d)^{6c})\;.
\end{align*}
The same reasoning allows us to read out other noise stability terms of $g_R$ directly in terms of those of $g_{R;rs}$ and bounds on $C^{VY}_{r,s}$ and $C^V_{r,s}$:
\begin{align*}
    \kappa_{1;m}( g_R ), \; \nu_{1;m}( g_R ) \;&=\; O( (d^2 + d^3) (\log d)^{6c} ) \;=\;O(d^3 (\log d)^{6c}) \;, 
    \\
    \kappa_{2;m}( g_R ), \; \nu_{2;m}( g_R ) \;&=\; O( d^4 (\log d)^{6c} )\;,    
    \qquad
    \kappa_{3;m}( g_R ), \; \nu_{3;m}( g_R ) \;=\; O(d^5 (\log d)^{6c})\;.
\end{align*}
The moment terms are bounded as before: $\bar c_m = O(d \tau^2 ) = O((\log d)^{2c})$ and $c_X, c_Z = O(d^3 \tau^6) = O((\log d)^{6c})$. By Lemma \ref{lem:plugin:formal} with $\delta$ set to $0$, we have
\begin{align*}
    d_{\cH}( \sqrt{n} R^{\Phi \cX}, \sqrt{n} R^T ) 
    \;=&\; 
    O\big(n^{-1/2} \kappa_{2;3}(g_R) \, \bar c_3^2 + n^{-1/2}  \kappa_{1;1}(g_R)^3 (c_X + c_Z) \big)\;,
    \\
    \;=&\; O( (n^{-1/2} d^4 + n^{-1/2} d^9) (\log d)^{24c} ) \;=\; O(n^{-1/2} d^9 (\log d)^{24c})\;,
    \\
    d_{\cH}( \sqrt{n} R^{\Phi \cX}, \sqrt{n} R^{\cZ} ) 
    \;=&\; 
    O\big(\big(n^{-1/2} \nu_{1;6}(g_R)^3  + 3 n^{-1} \nu_{1;4}(g_R) \nu_{2;4}(g_R) + n^{-3/2} \nu_{3;2}(g_R) \big)
    \\
    &\; \qquad \times (c_X + c_Z)\big) 
    \\
    \;=&\;O( ( n^{-1/2} d^9 + n^{-1} d^7 + n^{-3/2} d^5 ) (\log d)^{24c} ) 
    \\
    \;=&\; O(n^{-1/2} d^9 (\log d)^{24c})\;,
\end{align*}
which are the desired bounds in $d_H$, and by Lemma \ref{lem:plugin:formal} with $\delta=0$ again, we have
\begin{align*}
    n ( \Var[R^{\Phi \cX}] - \Var[R^T] )
    \;=&\; 
    O\big( n^{-1/2} \kappa_{1;1}(g_R) \kappa_{2;4}(g_R) \bar c_4^3 + n^{-1} \kappa_{2;6}(g_R) \kappa_{2;6}(g_R) \, \bar c_6^4 \big)
    \\
    \;=&\;
    O\big( (n^{-1/2} d^7 + n^{-1} d^8) (\log d)^{20c} \big) \;, 
    \\
    n ( \Var[R^{\Phi \cX}] - \Var[R^Z] )
    \;=&\;
    O\big(n^{-1} ( \nu_{0;4} \nu_{3;4} + \nu_{1;4} \nu_{2;4} ) (c_X + c_Z) \big)
    \\
    \;=&\;
    O\big( n^{-1} d^7 (\log d)^{18c} \big)\;,
\end{align*}
which are again the desired bounds for variance.
\end{proof}

\vspace{-1em}

\subsubsection{Existence of surrogate variables from a maximum entropy principle} \label{appendix:ridge:exist_surrogate}

\vspace{.3em}

As discussed after 
\cref{prop:ridge:regression}, 
the surrogate variables $\bZ_i \coloneqq \{ \bZ_{ij} \}_{j \leq k} = \{ (\bZ_{ij1}, \bZ_{ij2}) \}_{j \leq k}$ cannot be Gaussian since they take values in $(\M^d \times \R^{d \times b})^k$. Recall that the only restriction we have on $\bZ_i$ is from 
\eqref{eqn:defn:surrogates}: 
$\bZ_i$ should match the first two moments of $\Phi_i\bX^*_i$. A trivial choice is $\Phi_i\bX^*_i$ itself, but is not meaningful because the key of the theorem is that only the first two moments of $\Phi_i\bX^*_i$ matter in the limit.

\vspace{.3em}

The main difficulty is finding a distribution $p_{\M}$ on $\M^d$, the set of $d \times d$ positive semi-definite matrices, such that for $\bZ_{ij1} \sim p_{\M}$,
\begin{align} \label{eqn:pos_def_mat:surrogates}
    \mean [\bZ_{ij1}]
    \;&=\;
    \mean\big[ (\pi_{11}\bV_1) (\pi_{11}\bV_1)^\top \big]
    \;
    &\text{and}&&
    \Var [\bZ_{ij1}]
    \;&=\;
    \Var\big[ (\pi_{11}\bV_1) (\pi_{11}\bV_1)^\top \big]
    \;.
\end{align}
When $d=1$, the problem reduces to finding a distribution on non-negative reals given the first two moments, and one can choose the gamma distribution. When $d > 1$, a natural guess of a distribution on non-negative matrices is the non-central Wishart distribution. Unfortunately, one cannot form a non-central Wishart distribution given any mean and variance on $\M^d$, as illustrated in Lemma \ref{lem:wishart:impossible}. 

\begin{lemma} \label{lem:wishart:impossible} Let $d=1$. There exists random variable $V$\,with 
$\mean V^2 = 1$ and $\Var V^2 = 4$, but there is no non-central Wishart random variable $W$\,with $\mean W = 1$ and $\Var W = 5$. 
\end{lemma}

\begin{proof} Recall that $V \sim \Gamma(\alpha, \nu) $ has $\mean V^2  = \frac{\alpha(\alpha+1)}{\nu^2}$ and $\mean V^4  = \frac{\alpha(\alpha+1)(\alpha+2)(\alpha+3)}{\nu^4}$. Choose $\alpha = \frac{\sqrt{6}}{2}$ and $\nu = \sqrt{\frac{3+\sqrt{6}}{2}}$ gives
\begin{align*}
    &
    \mean V^2  \;=\; \mfrac{\sqrt{6}(\sqrt{6}+2)/4}{(3+\sqrt{6})/2}\;=\;1\;,
    &&
    \mean V^4 \;=\; \mfrac{(\sqrt{6}+4)(\sqrt{6}+6)/4}{(3+\sqrt{6})/2}\;=\;5\;,
\end{align*}
which gives the desired mean and variance for $V^2$. On the other hand, when $d=1$, the non-central Wishart distribution is exactly non-central chi-squared distribution parametrized by the degree of freedom $m$ and mean $\mu$ and variance $\sigma^2$ of the individual Gaussians. We can form the non-central Wishart random variable $W$ by drawing $Z_1, \ldots, Z_m \overset{i.i.d.}{\sim} \cN(0,1)$ and defining
\begin{align*}
    W \;\coloneqq\; \msum_{l=1}^m (\mu+\sigma Z_l)^2\;.
\end{align*}
Suppose $\mean[W] = 1$ and $\Var[W] = 4$. This implies
\begin{align*}
    &
    m (\mu^2 + \sigma^2) \;=\; 1\;,
    &&
    m ( 4 \mu^2 \sigma^2 + 2 \sigma^4 ) \;=\; 4\;.
\end{align*}
Write $x=\sigma^2$ and $\mu^2 = \frac{1}{m} - x$, we get $ m \big( 4 \big(\frac{1}{m} - x \big) x + 2 x^2\big) = 4$, which rearranges to
\begin{align} \label{eqn:wishart:impossible}
    x^2 - \mfrac{2}{m} x + \mfrac{2}{m} \;=\;0\;.
\end{align}
LHS equals $(x - \frac{1}{m})^2 + \frac{2m-1}{m^2}$, which is strictly positive since $m$ is a positive integer. Therefore there is no solution to \eqref{eqn:wishart:impossible} and hence no non-central Wishart random variable $W$\,with $\mean W=1$ and $\Var W =5$. This finishes the proof.
\end{proof}

\vspace{.3em}

The choice $d=1$ for the proof above is for simplicity and not necessity. Wishart distribution fails because of specific structure in its first two moments arisen from the outer product of Gaussian vectors, which may not satisfy the mean and variance required by \eqref{eqn:pos_def_mat:surrogates}. A different approach is to show existence of solution to the problem of moments via maximum entropy principle. In the case $\cD=\R^d$, Gaussian distribution is a max entropy distribution that solves the problem of moments given mean and variance. In the case $\cD$ is a closed subset of $\R^d$, the following result adapted from Ambrozie \cite{ambrozie2013multivariate} studies the problem of moments from an approximate maximum entropy principle:

\begin{lemma} {\rm [Adapted from Corollary 6(a-b)
  of \cite{ambrozie2013multivariate}]} \label{lem:max:approx:entropy} Fix $\epsilon > 0$. Let $T \subseteq \R^d$ be a closed subset and define the multi-index set $I \coloneqq \{ i \in \Z^d_+ \,|\, i_1 + \ldots + i_d \leq 2\}$. Let $(g_i)_{i \in I}$ be a set of reals with $g_{\bzero}=1$. Assume that there exist a probability measure $p_U$ with Lebesgue density function $f_U$ supported on $T$ such that, for every $(i_1, \ldots, i_d) \in I$, 
      \begin{align} \label{eqn:approx:max:entropy:cond}
          \mean_{\bU \sim p_U} [ | U_1^{i_1} \ldots U_d^{i_d} | ] \;&<\; \infty
          &\text{and}&&
          \mean_{\bU \sim p_U} [ U_1^{i_1} \ldots U_d^{i_d} ] \;&=\; g_i\;.
      \end{align}
 Then, there exists a particular solution $p^*_U$ of \eqref{eqn:approx:max:entropy:cond} with Lebesgue density $f^*_U$ that maximizes the $\epsilon$-entropy over all measures $p$ with Lebesgue density $f$,
      \begin{align*}
          H_{\epsilon}(p,f) \;=\; - \mean_{\bU \sim p}[ \log(f) ] - \epsilon \mean_{\bU \sim p}\big[ \| \bU \|^3 \big]\;.
      \end{align*}
\end{lemma}

We can now use Lemma \ref{lem:max:approx:entropy} to construct the surrogate variables $\bZ_i$ in 
\Cref{prop:ridge:regression} 
if the distribution of $\phi_{11}\bX^*_1$ admits a Lebesgue density function.

\begin{proof}[Proof for 
    \Cref{prop:ridge:regression}] 
    Assume first that the distribution of $\phi_{11}\bX^*_1$ admits a Lebesgue density function. Fix $d, b$. Note that $\cD^k$ is closed since $\cD = \M^d \times \R^{d \times b}$ is a product of two closed sets and therefore closed in $\R^{d \times d} \times \R^{d \times b}$. The distribution $p_{X;d,b}$ of $\Phi_1 \bX^*_1$ and its Lebesgue density $f_{X;d,b}$ then satisfy the assumption of Lemma \ref{lem:max:approx:entropy} with $T=\cD^k$ and the condition \eqref{eqn:approx:max:entropy:cond} becoming a bounded moment condition together with
\begin{align} \label{eqn:approx:max:entropy:cond:our:case}
  \mean_{\bU \sim p} [ \bU ] \;&=\; \mean[ \Phi_1\bX^*_1 ]\;
  &\text{and}&&
  \mean_{\bU \sim p} [ \bU^{\otimes 2} ] \;&=\; \mean[ (\Phi_1\bX^*_1)^{\otimes 2} ]\;.
\end{align}
Then by Lemma \ref{lem:max:approx:entropy}, there exists a distribution $p_{Z;d,b}$ with Lebesgue density function $f_{Z;d,b}$ which maximizes the $\epsilon$-entropy in Lemma \ref{lem:max:approx:entropy} while satisfying \eqref{eqn:approx:max:entropy:cond:our:case}. For each fixed $(d,b)$, taking $\bZ_{i;d,b} \sim p_{Z;d,b}$ then gives a choice of the surrogate variables. If the coordinates of $\bZ_{i;d,b}$ are uniformly bounded as $O(d^{-1})$ almost surely as $d$ grows with $b=O(d)$, we can apply Lemma \ref{lem:ridge:regression:extended}(iii) to yield the desired convergences, which finishes the proof. If either $\phi_{11}\bX^*_1$ does not admit a Lebesgue density function or if there is no uniform bound over the coordinates of $\bZ_{i;d,b}$ as $O(d^{-1})$, we take $\bZ_i$ to be an i.i.d.\,copy of $\Phi_i\bX^*_i$ which again gives the desired convergences but in a trivial manner.
\end{proof}

\subsubsection{Simulation and proof for toy example} \label{appendix:ridge:proof_toy}

\vspace{.3em}

In this section we focus on the toy model stated in 
\Cref{lem:ridge:toy}, 
where $d=1$ and
\begin{equation} 
   \label{eqn:ridge:toy:model:lem}
    \bY_i \coloneqq \bV_i
    \;\text{ where }\bV_i \stackrel{i.i.d.}{\sim} \cN( \mu, \sigma^2 ),\;\text{and}\;\pi_{ij} = \tau_{ij} \text{ a.s.}\;.
\end{equation}
Recall that we have taken the surrogate variables to be Gamma random variables. We now prove the convergence of variance and dependence of variance of estimate on the variance of data for the toy example in 
\Cref{lem:ridge:toy}.

\begin{proof}[Proof of 
    \Cref{lem:ridge:toy}] 
    To prove the first convergence statement, note that in 1d, $\M^1$ is the set of non-negative reals, and $\bZ_i = \{ \bZ_{ij1}, \bZ_{ij2} \}_{j \leq k}$ takes values in $(\M^1 \times \R)^k = \cD^k$ which agrees with the domain of data. Moreover, denoting $\mu_{V^2} \coloneqq \mean[(\bV_1)^2]$, the moments of $\bZ_i$ satisfy
\begin{align*}
    \mean[\bZ_i] \;=&\; 
    \bone_{k \times 1} \otimes \begin{psmallmatrix} \mu_{V^2} \\ \mu_{V^2} \end{psmallmatrix}\;,
    \;=\; 
    \bone_{k \times 1} \otimes \begin{psmallmatrix} \mean[(\pi_{11}\bV_1)^2] \\ \mean[(\tau_{11}\bY_1)^2] \end{psmallmatrix}\;,
    \\
    \Var[\bZ_i] \;=&\; \bone_{k \times k} \otimes \begin{psmallmatrix} v_\pi & v_\pi \\ v_\pi & v_\pi \end{psmallmatrix} 
    \;=\; 
    \bone_{k \times k} \otimes 
    \begin{psmallmatrix} 
    \Cov[(\pi_{11}\bV_1)^2, (\pi_{12}\bV_1)^2] & \Cov[(\pi_{11}\bV_1)^2, (\tau_{12}\bY_1)^2] \\
    \Cov[(\tau_{11}\bY_1)^2, (\pi_{12}\bV_1)^2] & \Cov[(\tau_{11}\bY_1)^2, (\tau_{12}\bY_1)^2]
    \end{psmallmatrix}\;.
\end{align*}
This corresponds to the mean and variance of $\bZ^{\delta}_i$ in Lemma \ref{lem:plugin:formal} with $\delta$ set to 1. While the earlier result on ridge regression in Lemma \ref{lem:ridge:regression:extended} does not apply directly, an analogous argument works by computing some additional mixed smoothness terms in Lemma \ref{lem:plugin:formal}(ii). Recall from the proof of Lemma \ref{lem:ridge:regression:extended} that for $d=1$, $\nu_{r;m} = O(1)$ for $0 \leq r \leq 3$. Therefore by Lemma \ref{lem:plugin:formal}(ii) with $\delta=1$, the following convergences hold as $n, k \rightarrow \infty$:
\begin{align*}
    d_{\cH}( \sqrt{n} & f(\Phi \cX), \sqrt{n} f(\bZ_1, \ldots, \bZ_n) ) 
    \\
    \;=&\; 
    O\big( ( k^{-1/2} + n^{-1/2}k^{-1/2} ) c_1
    +
    (n^{-1/2}  + 3 n^{-1}  + n^{-3/2}) (c_X + c_Z)
    \big)
    \;\rightarrow\; 0\;,
    \\
    n \| \Var[f(\Phi \cX)] - &\Var[ f(\bZ_1, \ldots, \bZ_n)] \| 
    \;=\; 
    O(  k^{-1/2} c_1
    +
     n^{-1} (c_X +  c_Z) ) \;\rightarrow\; 0\;.
\end{align*}
For the second statement, we first note that
$$
    S_Z \;\coloneqq\; \mfrac{1}{nk} \msum_{i=1}^n\msum_{j=1}^k \bZ_{ij1} \;=\; \mfrac{1}{nk} \msum_{i=1}^n\msum_{j=1}^k \bZ_{ij2} \;=\; \mfrac{1}{n} \msum_{i=1}^n \bZ_{i11} \sim \Gamma\big( \mfrac{n(\mu_{V^2})^2}{v_{\pi}}, \mfrac{n\mu_{V^2}}{v_{\pi}} \big)\;.
$$
Then we can write the variance of $R^Z$ in terms of $S_Z$:
\begin{align*}
    \Var[R^Z] \;=&\; \Var[ \mean[(\bV_{new} - \hat B^Z \bV_{new} )^2 | \hat B^Z ]  ]
    \\
    \;=&\; \Var[ -2 \mean[\bV_{new}^2 ] \hat B^Z + \mean[\bV_{new}^2] (\hat B^Z)^2 ]
    \\
    \;=&\; (\mu_{V^2})^2 \Var[ -2 \hat B^Z + (\hat B^Z)^2 ]
    \\
    \;=&\; (\mu_{V^2})^2 \Var\Big[ -2 \mfrac{S_Z}{S_Z + \lambda} + \mfrac{(S_Z)^2}{(S_Z + \lambda)^2} \Big]
    \\
    \;=&\; (\mu_{V^2})^2 \Var\Big[ \mfrac{- S_Z^2 - 2 \lambda S_Z}{(S_Z + \lambda)^2} \Big]
    \\
    \;=&\; (\mu_{V^2})^2 \Var\Big[ 1 - \mfrac{ S_Z^2 + 2 \lambda S_Z}{(S_Z + \lambda)^2} \Big]
    \\
    \;=&\;(\mu_{V^2})^2 \lambda^2 \Var\Big[ \mfrac{1}{(S_Z + \lambda)^2} \Big] 
    \\
    \;=&\; 
    \mean[\bV_1^2]^2  \lambda^2 \Var\Big[ \mfrac{1}{(X_n(v) + \lambda)^2} \Big]  \;=\; \sigma^2_n(v)
    \;.
\end{align*}
In the last line, we have denoted the random variable $X_n(v) \sim \Gamma( \frac{n(\mu_{V^2})^2}{v}, \frac{n\mu_{V^2}}{v})$ and recalled the definition of $\sigma_n(\nu)$, which is independent of $k$ and the distribution of $\pi_{ij}$. This completes the proof. 
\end{proof}

\subsection{Departure from Taylor limit at higher dimensions}  \label{appendix:ridge:slower_conv_taylor}

\vspace{.3em}

In Lemma \ref{lem:ridge:regression:extended}, we have shown convergences of the form
\begin{align*}
    n ( \Var[R^{\Phi \cX}] - \Var[R^T] )
    \;&=\; O\big(n^{-1/2} d^7 + n^{-1} d^8 \big)\;,
    \\
    n ( \Var[R^{\Phi \cX}] - \Var[R^Z] )
    \;&=\; O\big( n^{-1} d^7 \big)\;.
\end{align*}
While the bounds are not necessarily tight in terms of dimensions, they hint at different rates of convergences to the two limits. $\Var[R^T]$ has a simple behavior under augmentations as discussed for plugin estimators in 
\Cref{sec:averages}, 
and in particular is reduced when data is invariant under augmentations. On the other hand, $\Var[R^Z]$ has a complex behavior under augmentations as discussed in 
\Cref{sectn:ridge_regression}. 
In the main text, the separation of convergence rates is illustrated by a simulation that shows complex dependence of variance of risk under augmentation at a moderately high dimension. 

\vspace{.2em}

In this section we aim to find evidence for a non-trivial separation of the convergence rates by focusing on the following model: For positive constants $\sigma, \tilde \lambda$ independent of $n$ and $d$, consider
\begin{equation} 
\label{eqn:ridge:toy:model:div}
    \bY_i \coloneqq \bV_i
    \;\text{ where }\bV_i \stackrel{i.i.d.}{\sim} \cN(0, \sigma^2 \bone_{d \times d}),\;\pi_{ij} = \tau_{ij} = \text{id} \text{ a.s.},\;\text{ and } \lambda = d \tilde \lambda\;,
\end{equation}
where ${\rm id}$ is the identity map $\R^d \rightarrow \R^d$ and $\psi$ is an increasing function describing the rate of growth as a function of $d$. The parameter $\lambda$ is chosen to be $O(d)$ instead of $O(1)$ for this model so that the penalty does not vanish and the inverse in ridge regression stays well-defined as $d$ grows to infinity. Focusing on a specific model allows us to have a tight bound in terms of dimensions. The following lemma characterizes the convergence behavior of $\Var[R^{\Phi \cX}]$ to $\Var[ R^T]$ and $\Var[ R^Z]$ in terms of a function depending on $n$. 

\begin{lemma} \label{lem:ridge:toy:non_convergence} Assume the model \eqref{eqn:ridge:toy:model:div}. Let $\{Z_i\}_{i \leq n}$ be i.i.d.\,non-negative random variables with mean $1$, variance $2$ and finite 6th moments, and define $\bZ_i \coloneqq \{ \sigma^2 Z_i \bone_{d\times 1}, \sigma^2 Z_i \bone_{d \times 1} \}_{j \leq k}$. Then 
\begin{proplist}
    \item for $R^T$ defined on $\{\bZ_i\}_{i \leq n}$, 
    $$
    n | \Var[R^{\Phi\cX}] - \Var[R^T] | \;=\; n \Big| d^2\sigma^4 \tilde \lambda^4 \Var\Big[ \mfrac{1}{(\tilde \lambda + \sigma^2 \chi^2_n/n)^2} \Big] - \mfrac{8 d^2 \sigma^8 \tilde \lambda^4}{n (\tilde \lambda + \sigma^2)^6} \Big| \;,
    $$
    where $\chi^2_n$ is a chi-squared distributed random variable with $n$ degrees of freedom;
    \vspace{.2em}
    \item there exist a constant $C_1 > 0$ not depending on $n$ and $d$ and a quantity $C_2 = \Theta(1)$ as $n, d$ grow such that
    $$
     n |\Var[ R^{\Phi \cX}] - \Var[ R^T] |
     \;\geq\;
     nd^2 E(n) C_1 
    - n^{-1} d^2 C_2 \;,$$
    where $E(n)\coloneqq \big| \mean \big[ \frac{(\chi^2_n-n)^3}{n^3 (\tilde \lambda + \sigma^2 \chi^2_\Delta / n)^4} \big] \big|$ and $\chi^2_{\Delta}$ is a random variable between $\chi^2_n$ and $n$;
    \vspace{.2em}
    \item for $R^Z$ defined on $\{\bZ_i\}_{i \leq n}$,  
    $$
    n | \Var[R^{\Phi\cX}] - \Var[R^Z]| \;=\;
    O(n^{-1} d^2)\;
    .
    $$
\end{proplist}
\end{lemma}

\begin{figure}[t]
    \centering
    \includegraphics[width=12cm]{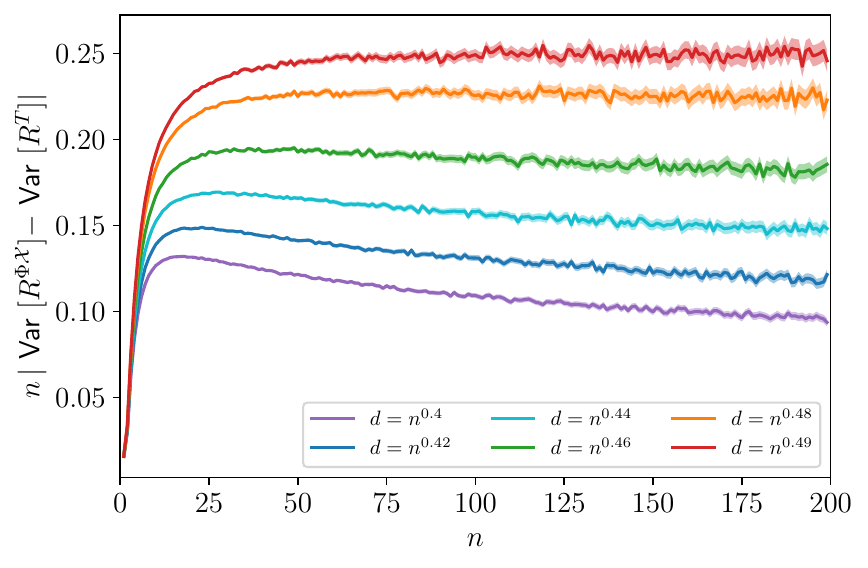}
    \caption{Plot of difference in variances computed in Lemma \ref{lem:ridge:toy:non_convergence}(i) against $n$ for $\tilde\lambda=\sigma=1$.}
    \label{fig:ridge:toy:div:sim_E_n}
\end{figure}

In Lemma \ref{lem:ridge:toy:non_convergence}, while $E(n)$ is a complicated function, if we compare it to $\mean[n^{-3}(\chi^2_n-n)^3]$, we expect the term to be on the order $n^{-3/2}$ as $n$ grows. A natural guess of the order of the first term in Lemma \ref{lem:ridge:toy:non_convergence} is $\Theta(n^{-1/2}d^2)$. This suggests that if $d=n^{\alpha}$ for some $\frac{1}{4} < \alpha < \frac{1}{2}$, we may have $n |\Var[ R^{\Phi \cX}] - \Var[ R^T] |$ not converging to 0 while the convergence of $n | \Var[R^{\Phi\cX}] - \Var[R^Z]|$ still holds due to Lemma \ref{lem:ridge:toy:non_convergence}(iii). A simulation in Figure \ref{fig:ridge:toy:div:sim_E_n} shows that this can indeed be the case in an example parameter regime: if $\{Z_i\}_{i \leq n}$ in Lemma \ref{lem:ridge:toy:non_convergence} are Gamma random variables, $\Var[R^Z] = \Var[R^{\Phi\cX}]$ exactly, whereas no matter how the distribution of $\{Z_i\}_{i \leq n}$ are chosen, the gap between $\Var[R^{\Phi\cX}]$ and $\Var[R^T]$ may not decay to zero as shown in Figure \ref{fig:ridge:toy:div:sim_E_n}. This suggests that for a moderately high dimension, it is most suitable to understand $\Var[R^{\Phi\cX}]$ through $\Var[R^Z]$ instead of $\Var[R^T]$. This completes the discussion from 
\Cref{remark:ridge}. 
It may be of interest to note that in 
\Cref{fig:regression}, 
the regime at which augmentation exhibits complex behavior despite invariance is when $d=7$ and $n=50$, i.e. when $d$ is close to $n^{1/2}$.

\vspace{.3em}

The proof of Lemma \ref{lem:ridge:toy:non_convergence}(i) is by a standard Taylor expansion argument followed by a careful lower bound. The essence of the proof of Lemma \ref{lem:ridge:toy:non_convergence}(ii) is by applying 
\Cref{thm:main}
 while considering the particular structure \eqref{eqn:ridge:toy:model:div}; we spell out the proof in full for clarity.

\begin{proof}[Proof of Lemma \ref{lem:ridge:toy:non_convergence}(i)] Denote $g_1(\Sigma) \coloneqq g_R(\Sigma, \Sigma)$ where $g_R$ is as defined in \eqref{eqn:defn:g_R} and $\mu_S \coloneqq \mean[ (\pi_{ij}\bV_i)(\pi_{ij}\bV_i)^\top ] = \sigma^2 \bone_{d \times d} $. We first seek to simplify the expressions of the variances: 
\begin{align*}
    \Var[ R^T] \;\coloneqq&\;  \Var \Big[ g_R(\mu_S, \mu_S) + \partial g_R(\mu_S, \mu_S) \Big( \mfrac{1}{nk} \msum_{i,j} \bZ_{ij} - (\mu_S, \mu_S) \Big) \Big]
    \\
    \;=&\; \Var \Big[ \partial g_R(\mu_S, \mu_S) \Big( \mfrac{1}{nk} \msum_{i,j} \bZ_{ij} - (\mu_S, \mu_S) \Big) \Big]\;.
\end{align*}
Since $\bZ_i$ matches the two moments of $\Phi_i\bX_i \;=\; \{ (\pi_{ij}\bV_i)(\pi_{ij}\bV_i)^\top, (\pi_{ij}\bV_i)(\pi_{ij}\bV_i)^\top \}_{j \leq k}$ and $\{\bZ_i\}_{i \leq n}$ are i.i.d., we get that
\begin{align*}
    \Var[ R^T]
    \;=&\; \Var \Big[ \partial g_R(\mu_S, \mu_S) \Big( \mfrac{1}{nk} \msum_{i,j} (\pi_{ij}\bV_i)(\pi_{ij}\bV_i)^\top, (\pi_{ij}\bV_i)(\pi_{ij}\bV_i)^\top) - (\mu_S, \mu_S) \Big) \Big]
    \\
    \;=&\; \Var \Big[ \partial g_1(\mu_S) \Big( \mfrac{1}{nk} \msum_{i,j} (\pi_{ij}\bV_i)(\pi_{ij}\bV_i)^\top - \mu_S) \Big) \Big]\;.
\end{align*}
Under \eqref{eqn:ridge:toy:model:div}, we can replace each $\pi_{ij}\bV_i$ by $ \sigma \xi_i \bone_d$ where $\{\xi_i\}_{i \leq n}$ are i.i.d.\,standard normal variables. Denote $\chi^2_n \coloneqq \msum_{i=1}^n \xi_i^2$. Then
\begin{align}
    \Var[R^T]
    \;=&\; \Var \Big[ \partial g_1(\mu_S) \Big( \mfrac{\sigma^2 \chi^2_n}{n} \bone_{d \times d} \Big) \Big]\;. \label{eqn:ridge:div:var:R_T:intermediate}
\end{align}
On the other hand,
\begin{align*}
    R^{\Phi\cX} \;=\; g_1\big( \mfrac{1}{nk}\msum_{i,j} (\pi_{ij}\bV_i)(\pi_{ij}\bV_i)^\top \big) \;=\; g_1\big( \mfrac{\sigma^2 \chi^2_n}{n} \bone_{d\times d} \big)\;.
\end{align*}
Given $\Sigma = x \bone_{d\times d}$ for some $x>0$, the explicit form of $g_1(\Sigma)$ and its derivative are given by Lemma \ref{lem:g1:derivatives} as
\begin{align*}
    g_1(\Sigma) 
    \;&=\;
    \mfrac{d \sigma^2 \lambda^2}{(\lambda + d x)^2}\;,
    &\,&&
    \partial g_1(\Sigma) \bone_{d \times d} 
    \;&=\; - \mfrac{2 d^2 \sigma^2  \lambda^2}{(\lambda + d x)^3}\;,
\end{align*}
This implies
\begin{align*}
    \Var[R^T]\;=\; \Var\Big[- \mfrac{2 d^2 \sigma^2  \lambda^2}{(\lambda + d \sigma^2)^3} \mfrac{\sigma^2 \chi^2_n}{n}  \Big] \;=\; \mfrac{4 d^4 \sigma^8 \lambda^4}{n^2 (\lambda + d \sigma^2)^6} \Var[\chi^2_n] \;=\; \mfrac{8 d^2 \sigma^8 \tilde \lambda^4}{n (\tilde \lambda + \sigma^2)^6}\;,
\end{align*}
where we have used $\Var[\chi^2_n] = 2n$ and $\lambda = d \tilde \lambda$. Moreover
\begin{align*}
    \Var[R^{\Phi\cX}] 
    \;=\; 
    \Var\Big[ \mfrac{d\sigma^2 \lambda^2}{(\lambda + d \sigma^2 \chi^2_n/n)^2} \Big] 
    \;=\; 
    \Var\Big[ \mfrac{d\sigma^2 \tilde \lambda^2}{(\tilde \lambda + \sigma^2 \chi^2_n/n)^2} \Big] 
    \;=\; 
    d^2\sigma^4 \tilde \lambda^4 \Var\Big[ \mfrac{1}{(\tilde \lambda + \sigma^2 \chi^2_n/n)^2} \Big]\;.
\end{align*}
Taking a difference and multiplying by $n$ gives the desired result:
\begin{align*}
    n | \Var[R^{\Phi\cX}] - \Var[R^T] | \;=\; n \Big| d^2\sigma^4 \tilde \lambda^4 \Var\Big[ \mfrac{1}{(\tilde \lambda + \sigma^2 \chi^2_n/n)^2} \Big] - \mfrac{8 d^2 \sigma^8 \tilde \lambda^4}{n (\tilde \lambda + \sigma^2)^6} \Big| \;.
\end{align*}
\end{proof}

\vspace{.2em}

\begin{proof}[Proof of Lemma \ref{lem:ridge:toy:non_convergence}(ii)] Note that a second-order Taylor expansion implies that almost surely there exists $\chi^2_{\Delta} \in [n, \chi^2_n]$ such that
\begin{align*}
    &R^{\Phi\cX} \;=\; g_1\big( \mfrac{1}{nk}\msum_{i,j} (\pi_{ij}\bV_i)(\pi_{ij}\bV_i)^\top \big) \;=\; g_1\big( \mfrac{\sigma^2 \chi^2_n}{n} \bone_{d\times d} \big)
    \\
    &= g_1( \sigma^2 \bone_{d\times d}) 
    + 
    \partial g_1( \sigma^2 \bone_{d\times d})\mfrac{\sigma^2 (\chi^2_n - n)}{n} \bone_{d\times d} 
    + 
    \mfrac{1}{2} \partial^2 g_1\big( \mfrac{\sigma^2 \chi^2_\Delta}{n} \bone_{d\times d}\big) \big( \mfrac{\sigma^4 (\chi^2_n - n)^2}{n^2} \big) (\bone_{d\times d})^{\otimes 2}\;.
\end{align*}
This implies
\begin{align*}
    \Var[ R^{\Phi\cX} ] \;=&\; 
    \Var \Big[ 
    \partial g_1( \mu_S )\mfrac{\sigma^2 \chi^2_n}{n} \bone_{d\times d} 
    + 
    \partial^2 g_1\big( \mfrac{\sigma^2 \chi^2_\Delta}{n} \bone_{d\times d}\big) \big( \mfrac{\sigma^4 (\chi^2_n - n)^2}{2 n^2} \big) (\bone_{d\times d})^{\otimes 2}
    \Big]
    \\
    \;=&\;
    \Var \Big[ 
    \partial g_1( \mu_S )\mfrac{\sigma^2 \chi^2_n}{n} \bone_{d\times d} \Big]
    + 
    \Var \Big[
    \partial^2 g_1\big( \mfrac{\sigma^2 \chi^2_\Delta}{n} \bone_{d\times d}\big) \big( \mfrac{\sigma^4 (\chi^2_n - n)^2}{2 n^2} \big) (\bone_{d\times d})^{\otimes 2}
    \Big]
    \\
    &\; + 2 \Cov \Big[ \partial g_1( \mu_S )\mfrac{\sigma^2 (\chi^2_n - n)}{n} \bone_{d\times d},  \partial^2 g_1\big( \mfrac{\sigma^2 \chi^2_\Delta}{n} \bone_{d\times d}\big) \big( \mfrac{\sigma^4 (\chi^2_n - n)^2}{2 n^2} \big) (\bone_{d\times d})^{\otimes 2} \Big]
    \,
\end{align*}
where the first term equals $\Var[R^T]$ by \eqref{eqn:ridge:div:var:R_T:intermediate}. Therefore by a triangle inequality, the difference in the variances of $R^{\Phi\cX}$ and $R^T$ can be written as
\begin{align}
    & n |\Var[ R^{\Phi \cX}] - \Var[ R^T] |
    \nonumber \\
    \;\geq&\;
    2 n \Big| \Cov \Big[ \partial g_1( \sigma^2 \bone_{d \times d} )\mfrac{\sigma^2 (\chi^2_n - n)}{n} \bone_{d\times d},  \partial^2 g_1\big( \mfrac{\sigma^2 \chi^2_\Delta}{n} \bone_{d\times d}\big)\big( \mfrac{\sigma^4 (\chi^2_n - n)^2}{2 n^2} \big) (\bone_{d\times d})^{\otimes 2} \Big] \Big|
    \label{eqn:ridge:div:R_T:inter:Cov}\\
    &\;
    -
    n \Big| \Var \Big[
    \partial^2 g_1\big( \mfrac{\sigma^2 \chi^2_\Delta}{n} \bone_{d\times d}\big) \big( \mfrac{\sigma^4 (\chi^2_n - n)^2}{2 n^2} \big) (\bone_{d\times d})^{\otimes 2}
    \Big] \Big| \;.  \label{eqn:ridge:div:R_T:inter:Var}
\end{align}
Given $\Sigma = x \bone_{d\times d}$ for some $x>0$, the explicit form of derivatives of $g_1(\Sigma)$ are given by Lemma \ref{lem:g1:derivatives} as
\begin{align*}
    \partial g_1(\Sigma) \bone_{d \times d} 
    \;&=\; - \mfrac{2 d^2 \sigma^2  \lambda^2}{( \lambda + d x)^3}\;,
    &\,&&
    \partial^2 g_1(\Sigma) (\bone_{d \times d})^{\otimes 2} \;&=\;
    \mfrac{6 d^3  \sigma^2  \lambda^2 }{(\lambda + d x)^4}\;.
\end{align*}
Note that $\mean[\chi^2_n - n] = 0$ and $\lambda = d \tilde \lambda$. The covariance term can be computed as
\begin{align*}
    \eqref{eqn:ridge:div:R_T:inter:Cov}
    \;&=\;
    2 n \Big|
    \Cov \Big[ - \mfrac{2 d^2 \sigma^2 \lambda^2}{(\lambda + d \sigma^2 )^3} \mfrac{\sigma^2 (\chi^2_n - n)}{n} ,  \mfrac{6 d^3  \sigma^2  \lambda^2 }{(\lambda + d \sigma^2 \chi^2_\Delta / n)^4}  \mfrac{\sigma^4 (\chi^2_n - n)^2}{2 n^2} \Big] \Big|
    \\
    \;&=\;
    \mfrac{12 n d^5 \sigma^{10} \lambda^4 }{(\lambda + d \sigma^2 )^3} \Big|
    \Cov \Big[ \mfrac{(\chi^2_n-n)}{n}, \mfrac{(\chi^2_n - n)^2}{n^2 (\lambda + d \sigma^2 \chi^2_\Delta / n)^4} \Big] \Big|
    \\
    \;&=\;
    \mfrac{12 n d^5 \sigma^{10} \lambda^4}{(\lambda + d \sigma^2 )^3} \Big|
    \mean \Big[ \mfrac{(\chi^2_n-n)^3}{n^3 (\lambda + d \sigma^2 \chi^2_\Delta / n)^4} \Big] \Big|
    \;=\;
    \mfrac{12 n d^2 \sigma^{10} \tilde \lambda^4}{(\tilde \lambda + \sigma^2 )^3} \Big|
    \mean \Big[ \mfrac{(\chi^2_n-n)^3}{n^3 (\tilde \lambda + \sigma^2 \chi^2_\Delta / n)^4} \Big] \Big|
    \\
    \;&=\;  \mfrac{12 n d^2 \sigma^{10} \tilde \lambda^4}{ ( \tilde\lambda + \sigma^2 )^3}
     E(n)
    \;=\; nd^2 C_1 E(n)
    \;,
\end{align*}
where $C_1 \coloneqq \mfrac{12\sigma^{10}\tilde \lambda^4}{(\tilde \lambda + \sigma^2)^3}$ is a constant not depending on $n$ and $d$ as required. The minus-variance term can be bounded as
\begin{align*}
    \eqref{eqn:ridge:div:R_T:inter:Var}
    \;=&\;
    - n \Big| \Var \Big[
    \mfrac{6 d^3  \sigma^2  \lambda^2 }{(\lambda + d \frac{\sigma^2 \chi^2_\Delta}{n})^4}
    \big( \mfrac{\sigma^4 (\chi^2_n - n)^2}{2 n^2} \big)
    \Big] \Big| 
    \;=\;
    - n \Big| \Var \Big[
    \mfrac{3 d  \sigma^2  \tilde \lambda^2 }{ ( \tilde \lambda + \frac{\sigma^2 \chi^2_\Delta}{n})^4}
    \big( \mfrac{\sigma^4 (\chi^2_n - n)^2}{ n^2} \big)
    \Big] \Big| 
    \\
    \;\overset{(a)}{\geq}&\; 
    - \mean\Big[
    \mfrac{9 n d^2  \sigma^4  \tilde \lambda^4 }{( \tilde \lambda + \frac{\sigma^2 \chi^2_\Delta}{n})^8}
    \big( \mfrac{\sigma^8 (\chi^2_n - n)^4}{ n^4} \big)
    \Big] 
    \\
    \;\overset{(b)}{\geq}&\;
    -
    \mfrac{9 n d^2  \sigma^{12} }{ \tilde \lambda^4} 
    \mean\Big[
     \mfrac{ (\chi^2_n - n)^4}{ n^4} 
    \Big] 
    \\
    \;\overset{(c)}{\geq}&\; 
    - n^{-1} d^2 \mfrac{9 \sigma^{12} }{ \tilde \lambda^4}  \Big( K_4 \max \big\{ n^{-1/4} \big( \| \xi_1^2 - 1 \|_{L_4}^4 \big)^{1/4}, \big( \| \xi_1^2 - 1 \|_{L_2}^2 \big)^{1/2} \big\} \Big)^4 
    \\
    \;=:&\; - n^{-1} d^2 \, C_2 \;.
\end{align*}
where in $(a)$ we have upper bounded variance with a second moment, in $(b)$ we have note that $\chi^2_{\Delta} \geq 0$ and in $(c)$ we have used Rosenthal's inequality from Lemma \ref{lem:rosenthal} to show that there exists a universal constant $K_4$ such that
\begin{align*}
    \mean\Big[
     \mfrac{ (\chi^2_n - n)^4}{ n^4} 
    \Big] \;&=\; \mfrac{1}{n^4} \big\|  \msum_{i=1}^n (\xi_i^2 - 1) \big\|_{L_4}^4
    \\
    \;&\leq\;
   \mfrac{1}{n^4} \Big(K_4 \max \big\{ \big( \msum_{i=1}^n \| \xi_i^2 - 1 \|_{L_4}^4 \big)^{1/4}, \big( \msum_{i=1}^n \| \xi_i^2 - 1 \|_{L_2}^2 \big)^{1/2} \big\}\Big)^4\;.
    \\
    \;&=\;
   \mfrac{1}{n^2} \Big( K_4 \max \big\{ n^{-1/4} \big( \| \xi_1^2 - 1 \|_{L_4}^4 \big)^{1/4}, \big( \| \xi_1^2 - 1 \|_{L_2}^2 \big)^{1/2} \big\} \Big)^4 
   \;=\; \Theta(n^{-2})\;.
\end{align*}
Therefore $C_2$ is $\Theta(1)$ as required, and we obtain the statement in (i) from the bounds on \eqref{eqn:ridge:div:R_T:inter:Cov} and \eqref{eqn:ridge:div:R_T:inter:Var}:
\begin{align*}
     n |\Var[ R^{\Phi \cX}] - \Var[ R^T] |
     \;\geq\;
     nd^2 C_1 
    E(n)
    - n^{-1} d^2 C_2\;.
\end{align*}
\end{proof}

\vspace{.2em} 
\begin{proof}[Proof of Lemma \ref{lem:ridge:toy:non_convergence}(iii)] Write $\omega^2_n \coloneqq \sum_{i=1}^n Z_i$. Note that
\begin{align}
    &\;|\Var[ R^{\Phi \cX}] - \Var[ R^Z] |
    \;=\; \big|\Var\big[ g_1 \big( \mfrac{\sigma^2 \chi^2_n}{n} \bone_{d \times d} \big) \big] - \Var \big[ g_1 \big( \mfrac{\sigma^2 \omega^2_n}{n} \bone_{d \times d} \big) \big] \Big|
    \nonumber \\
    \;&\leq\;
    \Big|
    \mean \big[  g_1 \big( \mfrac{\sigma^2 \chi^2_n}{n} \bone_{d \times d} \big) \big] -
    \mean \big[  g_1 \big( \mfrac{\sigma^2 \omega^2_n}{n} \bone_{d \times d} \big) \big]
    \Big| \, \Big| 
    \mean \big[  g_1 \big( \mfrac{\sigma^2 \chi^2_n}{n} \bone_{d \times d} \big) \big] +
    \mean \big[  g_1 \big(  \mfrac{\sigma^2 \omega^2_n}{n} \bone_{d \times d} \big) \big] \Big|
    \nonumber \\
    \;&\qquad + 
    \big|\mean\big[ g_1 \big(\mfrac{\sigma^2 \chi^2_n}{n} \bone_{d \times d} \big)^2 -g_1 \big(\mfrac{\sigma^2 \omega^2_n}{n} \bone_{d \times d} \big)^2  \big] \big| \;.
    \label{eqn:div:R_Z:intermediate}
\end{align}
We aim to bound \eqref{eqn:div:R_Z:intermediate} by mimicking the proof of 
\Cref{thm:main} 
but use tighter control on dimensions since we know the specific form of the estimator. Write
$$
    \bar W_i( w) \;\coloneqq\; \mfrac{1}{n}\Big( \msum_{i'=1}^{i-1} \xi_{i'}^2 + w + \msum_{i'=i+1}^n Z_{i'} \Big)\;,
$$
and denote $D_i^r g_{1;i}(w) \;\coloneqq\; \partial^r g_1 \big( \frac{\sigma^2}{n} \bar W_i(w) \bone_{d \times d} \big)$ for $r=0,1,2,3$. Then analogous to the proof of 
\Cref{thm:main}, 
by a third-order Taylor expansion around $0$ and noting that the first two moments of $\xi_i^2$ and $\bZ_{ij1}$ match, we obtain that
\begin{align}
    &\Big| \mean \big[  g_1 \big( \mfrac{\sigma^2 \chi^2_n}{n} \bone_{d \times d} \big) \big] -
    \mean \big[  g_1 \big( \mfrac{\sigma^2 \omega^2_n}{n} \bone_{d \times d} \big) \big] \Big|
    \nonumber \\
    \;=&\; \big| 
    \msum_{i=1}^n \mean\big[ g_1( \frac{\sigma^2}{n} \bar W_i( \xi_i^2) \bone_{d\times d} ) - g_1( \frac{\sigma^2}{n} \bar W_i( Z_i) \bone_{d\times d} ) \big] \big|
    \nonumber \\
    \;\leq&\;
    \msum_{i=1}^n \mean \Big[ \sup_{w \in [0, \xi_i^2]} \big| D_i^3 g_{1;i}(w) \mfrac{\sigma^6 (\xi_i^2)^3}{n^3}  (\bone_{d\times d})^{\otimes 3}  \big| +  \sup_{w \in [0, Z_i]}  \big| D_i^3 g_{1;i}(w) \mfrac{\sigma^6 (Z_i)^3}{n^3}  (\bone_{d\times d})^{\otimes 3} \big| \Big]
    \;. \label{eqn:div:R_Z:diff:first:moment}
\end{align}
Similarly,
\begin{align}
    &
    \big|\mean\big[ g_1 \big(\mfrac{\sigma^2 \chi^2_n}{n} \bone_{d \times d} \big)^2 -g_1 \big(\mfrac{\sigma^2 \omega^2_n}{n} \bone_{d \times d} \big)^2  \big] \big|
    \nonumber \\
    \;\leq&\;
    2 \msum_{i=1}^n  \mean \Big[ \sup_{w \in [0, \xi_i^2]} \big| \big(g_{1;i}(w) D_i^3 g_{1;i}(w) + D_i g_{1;i}(w) D_i^2 g_{1;i}(w) \big) \mfrac{\sigma^6 (\xi_i^2)^3}{n^3}  (\bone_{d\times d})^{\otimes 3}  \big|
    \nonumber \\
    &\;\;\;\;\;\;\qquad+
     \sup_{w \in [0, Z_i]}  \big| \big(g_{1;i}(w) D_i^3 g_{1;i}(w) + D_i g_{1;i}(w) D_i^2 g_{1;i}(w) \big)  \mfrac{\sigma^6 (Z_i)^3}{n^3}  (\bone_{d\times d})^{\otimes 3} \big| \Big]
    \;. \label{eqn:div:R_Z:diff:second:moment}
\end{align}
Given $\Sigma = x \bone_{d\times d}$ for some $x>0$, the explicit forms of $g_1(\Sigma)$ and its derivatives from Lemma \ref{lem:g1:derivatives} imply that
\begin{align*}
    g_{1;i}(w) 
    \;&=\;
    \mfrac{d \sigma^2 \lambda^2}{(\lambda + d \frac{\sigma^2 \bar W_i(w)}{n} )^2}\;,
    &\,&&
    D_i g_{1;i}(w) \bone_{d \times d} 
    \;&=\; - \mfrac{2 d^2 \sigma^2  \lambda^2}{ (\lambda + d \frac{\sigma^2 \bar W_i(w)}{n})^3}\;,
    \nonumber \\
    D_i^2 g_{1;i}(w) (\bone_{d \times d})^{\otimes 2} \;&=\;
    \mfrac{6 d^3 \sigma^2  \lambda^2 }{ ( \lambda + d \frac{\sigma^2 \bar W_i(w)}{n})^4}\;,
    &\,&&
    D_i^3 g_{1;i}(w) (\bone_{d \times d})^{\otimes 3} 
    \;&=\;
    - \mfrac{24 d^4 \sigma^2 \lambda^2 }{ ( \lambda + d \frac{\sigma^2 \bar W_i(w)}{n})^5} \;.
\end{align*}
Therefore, by noting $\lambda = d \tilde \lambda$, we get
\begin{align*}
    \eqref{eqn:div:R_Z:diff:first:moment} \;=&\; 24 n^{-3} d^4 \sigma^8 \lambda^2 \msum_{i=1}^n 
    \mean \Big[ 
    \sup_{w \in [0, \xi_i^2]} \Big| \mfrac{(\xi_i^2)^3}{(\lambda + \frac{ d \sigma^2 \bar W_i(w)}{n} )^5}  \Big| 
    +  \sup_{w \in [0, Z_i]}  \Big| \mfrac{ (Z_i)^3}{( \lambda + \frac{ d \sigma^2 \bar W_i(w)}{n} )^5}  \Big| 
    \Big]
    \\
    \;\overset{(a)}{\leq}&\;
    24 n^{-3} d^4 \sigma^8 \lambda^{-3} \msum_{i=1}^n 
    \mean [ (\xi_i^2)^3 + Z_i^3 ]
    \\
    \;=&\;
    24 n^{-2} d \sigma^8 \tilde \lambda^{-3} 
    \mean [ (\xi_1^2)^3 + Z_1^3 ] \;=\; O( n^{-2} d )\;.
\end{align*}
where in $(a)$ we have used that $\bar W_i(w) \geq 0$ almost surely for $w \in [0, \xi_i^2]$ and for $w \in [0, Z_i]$. By the same argument,
\begin{align*}
    \eqref{eqn:div:R_Z:diff:second:moment} 
    \;=&\;
    72 n^{-3} d^5 \sigma^{10} \lambda^4 \msum_{i=1}^n  \mean \Big[ \sup_{w \in [0, \xi_i^2]} \big| \mfrac{(\xi_i^2)^3}{(\lambda + d \frac{\sigma^2 \bar W_i(w)}{n} )^7}   \big| 
    +  
    \sup_{w \in [0, Z_i]} \big| \mfrac{(Z_i^2)^3}{(\lambda + d \frac{\sigma^2 \bar W_i(w)}{n} )^7}   \big| \Big]
    \\
    \;\leq&\; 72 n^{-2} d^2 \sigma^{10} \tilde \lambda^{-3} \mean [ (\xi_i^2)^3 + Z_i^3 ] \;=\; O(n^{-2}d^2)\;.
\end{align*}
Moreover,
\begin{align}
    \Big| 
    \mean \big[  g_1 \big( \mfrac{\sigma^2 \chi^2_n}{n} \bone_{d \times d} \big) \big] +
    \mean \big[  g_1 \big(  \mfrac{\sigma^2 \omega^2_n}{n} \bone_{d \times d} \big) \big] \Big|
    \;=\;
    | \mean[ g_{1;n}(\xi_n^2) + g_{1;1}(Z_1) ] |
    \;=\;
    O( d )\;. \label{eqn:div:R_Z:sum:zero:moment}
\end{align}
Finally the above three bounds imply that
\begin{align*}
    n | \Var[R^{\Phi\cX}] - \Var[R^Z]| \;\leq\; n \, \eqref{eqn:div:R_Z:intermediate}
    \;\leq\;
    n \, \eqref{eqn:div:R_Z:diff:first:moment}  \times \eqref{eqn:div:R_Z:sum:zero:moment}
    +
    n \,\eqref{eqn:div:R_Z:diff:second:moment}
    \;=\;
    O(n^{-1} d^2)\;,
\end{align*}
which is the desired bound.
\end{proof}

\vspace{.2em}

\begin{lemma}\label{lem:g1:derivatives} Consider $\Sigma = x \bone_{d\times d}$ for some $x>0$ and $g_1(\Sigma) \coloneqq g_R(\Sigma, \Sigma)$ where $g_R$ is defined as in \eqref{eqn:defn:g_R} under the model \eqref{eqn:ridge:toy:model:div}. Then, the following derivative formulas hold:
\begin{align*}
    g_1(\Sigma) 
    \;&=\;
    \mfrac{d \sigma^2 \lambda^2}{(\lambda + d x)^2}\;,
    &\,&&
    \partial g_1(\Sigma) \bone_{d \times d} 
    \;&=\; - \mfrac{2 d^2 \sigma^2  \lambda^2}{(\lambda + d x)^3}\;,
    \nonumber \\
    \partial^2 g_1(\Sigma) (\bone_{d \times d})^{\otimes 2} \;&=\;
    \mfrac{6 d^3 \sigma^2  \lambda^2 }{( \lambda + d x)^4}\;,
    &\,&&
    \partial^3 g_1(\Sigma) (\bone_{d \times d})^{\otimes 3} 
    \;&=\;
    - \mfrac{24 d^4 \sigma^2 \lambda^2 }{ ( \lambda + d x)^5} \;.
\end{align*}

\end{lemma}

\begin{proof} 
First note that
\begin{align*}
    \mean[\|\bY_{new}\|^2_2]\;=\; \mean[\|\bV_{new}\|^2_2]\;=\; \sigma^2 d \;,
    \quad
    \mean[ \bV_{new} \bY_{new}^\top  ] \;=\; \mean[ \bV_{new} \bV_{new}^\top  ] \;=\; \sigma^2  \bone_{d \times d} \;,
\end{align*}
which allows us to write
\begin{align*}
    g_1(\Sigma) 
    &= \sigma^2 d - 2 \sigma^2 \msum_{r,s=1}^d \, g_{B;rs}(\Sigma, \Sigma) + \sigma^2 \msum_{r,s,t=1}^d g_{B;rt}(\Sigma, \Sigma)  g_{B;ts}(\Sigma, \Sigma),
\end{align*}
where we have recalled the expression
\begin{align*}
    g_{B;rs}(\Sigma, \Sigma) \;=\; \be_r^\top (\Sigma + \lambda \bI_{d\times d})^{-1} \Sigma \be_s \;.
\end{align*}
Denoting $\tilde \Sigma =  (\Sigma + \lambda \bI_{d\times d})^{-1} =  (\Sigma + \lambda \bI_d)^{-1} $, the partial derivative of $g_{B;rs}$ has been computed in the proof of Lemma \ref{lem:ridge:regression:extended}(i) as
\begin{align*}
    \mfrac{\partial g_{B;rs}(\Sigma,\Sigma)}{\partial \Sigma_{r_1s_1}} 
    &=
    - \be_r^\top \tilde \Sigma^{-1} \be_{r_1} \be_{s_1}^\top  \tilde \Sigma^{-1} \Sigma \be_s + \be_r^\top \tilde \Sigma^{-1} \be_{r_1} \ind_{\{s=s_1\}}
    \\
    \;&=\; \be_r^\top \tilde \Sigma^{-1} \be_{r_1} \be_{s_1}^\top  \big( - \tilde \Sigma^{-1} \Sigma + \tilde \Sigma^{-1} \tilde \Sigma \big) \be_s
    \;=\; \psi(d) \tilde\lambda \be_r^\top \tilde \Sigma^{-1} \be_{r_1} \be_{s_1}^\top \tilde \Sigma^{-1} \be_s \;,
\end{align*}
Similarly
\begin{align*}    
    \mfrac{\partial^2 g_{B;rs}(\Sigma,\Sigma)}{\partial \Sigma_{r_1s_1}\partial\Sigma_{r_2s_2}} 
    \;=&\;
    \msum_{ l_1, l_2 \in \{1,2\};\; l_1 \neq l_2} \Big( \be_r
    \tilde \Sigma^{-1}
    \be_{r_{l_1}} \be_{s_{l_1}}^\top 
    \tilde \Sigma^{-1}
    \be_{r_{l_2}} \be_{s_{l_2}}^\top
    \tilde \Sigma^{-1}
    \Sigma \be_s
    \\
    &\;
    \qquad\qquad\qquad\qquad
    -
    \be_r^\top
    \,\tilde \Sigma^{-1}\,
    \be_{r_{l_1}} \be_{s_{l_1}}^\top
    \,\tilde \Sigma^{-1}\,
    \be_{r_{l_2}} \ind_{\{s=s_{l_2}\}}
    \Big)
    \\
    \;=&\;
    - \psi(d) \tilde\lambda  \msum_{ l_1, l_2 \in \{1,2\};\; l_1 \neq l_2} \be_r
    \tilde \Sigma^{-1}
    \be_{r_{l_1}} \be_{s_{l_1}}^\top 
    \tilde \Sigma^{-1}
    \be_{r_{l_2}} \be_{s_{l_2}}^\top
    \tilde \Sigma^{-1} \be_s\;,
\end{align*}
and
\begin{align*}
    \mfrac{\partial^3 g_{B;rs}(\Sigma,\Sigma)}{\partial \Sigma_{r_1s_1}\partial\Sigma_{r_2s_2}\partial\Sigma_{r_3s_3}}
    \;=&\;
    - \msum_{ \substack{l_1, l_2, l_3 \in \{1,2,3\} \\ l_1, l_2, l_3 \text{ distinct}} }
    \Big(
    \be_r^\top \tilde \Sigma^{-1}
    \be_{r_{l_1}} \be_{s_{l_1}}^\top 
    \tilde \Sigma^{-1}
    \be_{r_{l_2}} \be_{s_{l_2}}^\top
    \tilde \Sigma^{-1}
    \be_{r_{l_3}} \be_{s_{l_3}}^\top
    \tilde \Sigma^{-1}
    \Sigma \be_s
    \\
    &\;\;
    \qquad\qquad\qquad\qquad
    -
    \be_r^\top
    \tilde \Sigma^{-1}
    \be_{r_{l_1}} \be_{s_{l_1}}^\top 
    \tilde \Sigma^{-1}
    \be_{r_{l_2}} \be_{s_{l_2}}^\top
    \tilde \Sigma^{-1}
    \be_{r_{l_3}} \ind_{\{s=s_{l_3}\}}
    \Big)
    \\
    \;=&\;
    \psi(d) \tilde\lambda 
    \msum_{ \substack{l_1, l_2, l_3 \in \{1,2,3\} \\ l_1, l_2, l_3 \text{ distinct}} }
    \be_r^\top \tilde \Sigma^{-1}
    \be_{r_{l_1}} \be_{s_{l_1}}^\top 
    \tilde \Sigma^{-1}
    \be_{r_{l_2}} \be_{s_{l_2}}^\top
    \tilde \Sigma^{-1}
    \be_{r_{l_3}} \be_{s_{l_3}}^\top
    \tilde \Sigma^{-1}
    \be_s\;.
\end{align*}
On the other hand, since $\Sigma=x \bone_{d\times d}$, a calculation gives
\begin{align}
    \tilde \Sigma^{-1} \;=\; (x \bone_{d\times d} +  \lambda \bI_d)^{-1} \;=\; \mfrac{1}{ \lambda (  \lambda+ d x)} \big( ( \lambda + d x) \bI_d  - x \bone_{d \times d} \big)\;, \label{eqn:ridge:div:inv:formula}
\end{align}
in which case, denoting $J_{r,s}(x)\coloneqq ( \ind_{\{r = s\}} ( \lambda + (d-1)x) - \ind_{\{r \neq s\}} x )$, we have
\begin{align*}
    g_{B;rs}(\Sigma, \Sigma) \;=&\;
    \mfrac{x}{ \lambda \,( \lambda+ dx)} \big( (\lambda+dx) - dx) \;=\;
    \mfrac{x}{\lambda+ dx}\;,
    \\
    \mfrac{\partial g_{B;rs}(\Sigma, \Sigma)}{\partial \Sigma_{r_1s_1}}
    \;=&\;
    \mfrac{J_{r,r_1}(x)  J_{s,s_1}(x)}{\lambda \, ( \lambda+ d x)^2} 
    \;,
    \\
    \mfrac{\partial^2 g_{B;rs}(\Sigma, \Sigma)}{\partial \Sigma_{r_1s_1} \partial \Sigma_{r_2s_2}}
    \;=&\;
    - \msum_{ l_1, l_2 \in \{1,2\};\; l_1 \neq l_2}  \mfrac{ J_{r,r_{l_1}}(x)  J_{s_{l_1},r_{l_2}}(x) J_{s_{l_2},s}(x) }{\lambda^2 \,( \lambda+ dx)^3} 
    \;,
    \\
    \mfrac{\partial^3 g_{B;rs}(\Sigma, \Sigma)}{\partial \Sigma_{r_1s_1} \partial \Sigma_{r_2s_2} \partial \Sigma_{r_3s_3}}
    \;=&\; 
    \msum_{ \substack{l_1, l_2, l_3 \in \{1,2,3\} \\ l_1, l_2, l_3 \text{ distinct}} }
    \mfrac{ J_{r,r_{l_1}}(x)  J_{s_{l_1},r_{l_2}}(x) J_{s_{l_2},r_{l_3}}(x) J_{s_{l_3},s}(x)  }{ \lambda^3 \, ( \lambda+d x)^4}
    \;.
\end{align*}
Note that $J_{r,s}(x) = J_{s,r}(x)$ and $\sum_{r=1}^d J_{r,s}(x) = \lambda$. These formulas and the above derivatives imply that
\begin{align*}
    g_1(\Sigma) \;=&\; \sigma^2 d - 2 \sigma^2 \msum_{r,s=1}^d \, g_{B;rs}(\Sigma, \Sigma) + \sigma^2 \msum_{r,s,t=1}^d g_{B;rt}(\Sigma, \Sigma)  g_{B;ts}(\Sigma, \Sigma)
    \\
    \;=&\;
    \sigma^2 d - 2 \mfrac{\sigma^2 x d^2 }{ \lambda + d x} + \mfrac{\sigma^2  x^2 d^3}{(  \lambda + d x)^2}
    \;=\;
    \mfrac{d \sigma^2 \lambda^2}{( \lambda + d x)^2}
    \;,
    \\
    \partial g_1(\Sigma) \bone_{d \times d} \;=&\;
    \msum_{r_1,s_1=1}^d  \mfrac{\partial g_1(\Sigma)}{\partial \Sigma_{r_1 s_1}} 
    \\
    \;=&\;
    - 2 \sigma^2 \msum_{r,s, r_1, s_1} \, \mfrac{\partial g_{B;rs}(\Sigma, \Sigma)}{\partial \Sigma_{r_1 s_1}}  +  2 \sigma^2 \msum_{r,s,t, r_1, s_1} \mfrac{\partial g_{B;rt}(\Sigma, \Sigma)}{\partial \Sigma_{r_1 s_1}}   g_{B;ts}(\Sigma, \Sigma)
    \\
    =&\; -  \mfrac{ 2d^2 \sigma^2 \lambda}{ ( \lambda+ dx)^2} + \mfrac{ 2d^3  \sigma^2 x \lambda}{ ( \lambda + dx)^3} 
    \;=\; - \mfrac{2 d^2 \sigma^2  \lambda^2}{( \lambda + d x)^3}\;,
\end{align*}
\begin{align*}
    &\;\partial^2 g_1(\Sigma) (\bone_{d \times d})^{\otimes 2} \;=\;
    \msum_{r_1, s_1, r_2, s_2=1}^d \mfrac{\partial^2 g_1(\Sigma)}{\partial \Sigma_{r_1 s_1} \partial \Sigma_{r_2 s_2}} 
    \\
    \;=&\;
    - 2 \sigma^2 \msum_{r,s, r_1, s_1, r_2, s_2} \, \mfrac{\partial^2 g_{B;rs}(\Sigma, \Sigma)}{\partial \Sigma_{r_1 s_1} \partial \Sigma_{r_2 s_2}} 
    + 2 \sigma^2 \msum_{r,s,t,r_1, s_1, r_2, s_2} \mfrac{\partial^2 g_{B;rt}(\Sigma, \Sigma)}{\partial \Sigma_{r_1 s_1} \partial \Sigma_{r_2 s_2}}   g_{B;ts}(\Sigma, \Sigma)
    \\
    &\; + 2 \sigma^2 \msum_{r,s,t, r_1, s_1, r_2, s_2} \mfrac{\partial g_{B;rt}(\Sigma, \Sigma)}{ \partial \Sigma_{r_1 s_1}} \mfrac{\partial g_{B;ts}(\Sigma, \Sigma)}{\partial \Sigma_{r_2 s_2}} 
    \\
    \;=&\; 
    \mfrac{4 d^3 \sigma^2 \lambda }{(\lambda + dx)^3}
    - \mfrac{4 d^4 \sigma^2 \lambda x}{( \lambda + d x)^4} + \mfrac{2 d^3  \sigma^2 \lambda^2 }{ (\lambda+ dx)^4}
    \;=\;
    \mfrac{6 d^3 \sigma^2  \lambda^2 }{(\lambda + d x)^4}\;,
\end{align*}
\begin{align*}
    \partial^3 g_1(\Sigma) (\bone_{d \times d})^{\otimes 3} 
    \;=&\;
    - 2 \sigma^2 \msum_{r,s,r_1,s_1,r_2,s_2,r_3,s_3} \, \mfrac{\partial^3 g_{B;rs}(\Sigma, \Sigma)}{\partial \Sigma_{r_1 s_1} \partial \Sigma_{r_2 s_2} \partial \Sigma_{r_3 s_3}} 
    \\
    &\; 
    + 2 \sigma^2 \msum_{r,s,t,r_1,s_1,r_2,s_2,r_3,s_3} \mfrac{\partial^3 g_{B;rt}(\Sigma, \Sigma)}{\partial \Sigma_{r_1 s_1} \partial \Sigma_{r_2 s_2} \partial \Sigma_{r_3 s_3}}   g_{B;ts}(\Sigma, \Sigma)
    \\
    &\; + 6 \sigma^2 \msum_{r,s,t,r_1,s_1,r_2,s_2,r_3,s_3} \mfrac{\partial^2 g_{B;rt}(\Sigma, \Sigma)}{ \partial \Sigma_{r_1 s_1} \partial \Sigma_{r_2 s_2}} \mfrac{\partial g_{B;ts}(\Sigma, \Sigma)}{\partial \Sigma_{r_3 s_3}} 
    \\
    \;=&\;
    - \mfrac{12 d^4 \sigma^2 \lambda }{ ( \lambda + d x)^4} 
    + \mfrac{12 d^5  \sigma^2  \lambda x}{(\lambda+ dx)^5} - \mfrac{12 d^4  \sigma^2  \lambda^2 }{( \lambda+dx)^5}
    \;=\;
    - \mfrac{24 d^4  \sigma^2 \lambda^2 }{ ( \lambda + d x)^5} \;,
\end{align*}
which completes the proof.
\end{proof}

%% file: appendix_7_ridgeless.tex
\section{Proof for Sections \ref{sec:triple:descent:oracle}--\ref{sec:triple:descent:two:stage}
and Appendix \ref{appendix:ridgeless:results}} \label{appendix:ridgeless:proof}

We follow the notation in 
\cref{sec:ridgeless} 
and \cref{appendix:ridgeless:results}. We first prove a list of results on $f^{(1)}_\lambda$ and $f^{(2)}_\lambda$, collected in \cref{lem:ridgeless:tools}, that are useful for subsequent derivations. Section \ref{appendix:ridgeless:results:proof} presents the proofs for results in \cref{appendix:ridgeless:results}, whereas \Cref{appendix:proof:ridgeless:universality,appendix:ridgeless:surrogate:proof,appendix:proof:two:stage:aug,appendix:proof:ridgeless:extend}  present the proofs for 
\cref{sec:ridgeless}.

\vspace{.5em}

Throughout, for a real symmetric matrix $A \in \R^{n \times n}$, we denote $\lambda_1(A) \leq \ldots \leq \lambda_d(A)$ as its eigenvalues and denote the associated eigenvectors as $v_1(A), \ldots, v_d(A)$. 

\begin{lemma} \label{lem:ridgeless:tools} Let $A$, $A'$ and $B$ be $\R^{d \times d}$ symmetric matrices and fix $\lambda \geq 0$.  
\begin{proplist}
    \item The following bounds control the sizes of $f^{(1)}_\lambda$ and $f^{(2)}_\lambda$:
    \begin{align*}
        |  f^{(1)}_\lambda(A) | 
        \;\leq&\; 
            \max_{l \leq d ;\;  \lambda_l(A) \neq - \lambda} 
            \mfrac{\lambda^2}{(\lambda_l(A) + \lambda)^2} 
            \| \beta \|^2 
        \qquad 
        \text{ for } \lambda > 0
        \;,
        \\
        |  f^{(1)}_0(A) | 
        \;\leq&\; 
            \msum_{l=1}^d \ind_{\{ \lambda_l(A) = 0 \}} 
            \| \beta \|^2 
        \;,
        \\
        | f^{(2)}_\lambda(A, B) | 
        \;\leq&\; 
        \max_{ l \leq d ;\;  \lambda_l(A) \neq - \lambda} 
        \; \mfrac{d \, \sigma_\epsilon^2  \, \| B \|_{op} }{n(\lambda_l(A) + \lambda)^2} \;.
    \end{align*}
    \item  The following bounds hold for the approximations of $f^{(1)}_0$ by $f^{(1)}_\lambda$ and $f^{(2)}_0$ by $f^{(2)}_\lambda$, where $\lambda > 0$:
    \begin{align*}
        \big| f^{(1)}_\lambda(A) - f^{(1)}_0(A) \big|
        \;\leq&\;
        \mmax_{l \leq d;\, \lambda_l(A) \not\in \{ 0, - \lambda\} } \mfrac{\lambda^2  \|\beta\|^2}{(\lambda_l(A)+\lambda)^2}
        \;,
        \\
        \big| f^{(2)}_\lambda(A,B) - f^{(2)}_0(A,B) \big|
        \;\leq&\; 
        \mfrac{\sigma_\epsilon^2}{n \lambda^2}  \msum_{l=1}^d  \ind_{\{ \lambda_l(A) \in \{0, - \lambda\} \}} |v_l(A)^\top B \, v_l(A)|
        \\
        & 
        +
        \mfrac{ \lambda d \, \sigma_\epsilon^2  }{n} 
        \,
        \max_{l \leq d ;\, \lambda_l(A) \not\in \{0, - \lambda\} }
        \mfrac{| \lambda + 2 \lambda_l(A)   | \, \| B  \|_{op}}{\lambda_l(A)^2 ( \lambda_l(A) + \lambda)^2}
        \;.
    \end{align*}
\end{proplist}
Now suppose additionally that $\lambda > 0$, $\lambda_1(A) \geq -\lambda /2$ and $\lambda_1(A') \geq -\lambda /2$. Then we have
 \begin{proplist}
    \item[(iii)] the following bounds hold on the effect of perturbing the argument of $f^{(1)}_\lambda$ and $f^{(2)}_\lambda$:
    \begin{align*}
        \big| f^{(1)}_\lambda(A) - f^{(1)}_\lambda(A') \big|
        \;\leq&\;
        4 \, \| \beta \|^2 \| A - A' \|_{op}
        \\
        \big| f^{(2)}_\lambda(A, B) - f^{(2)}_\lambda(A', B) \big|
        \;\leq&\;
        \mfrac{ 16 \, \sigma_\epsilon^2  d }{n \lambda^3}  \| A - A' \|_{op} \| B \|_{op}
        \;.
    \end{align*}
\end{proplist}
\end{lemma}

\vspace{1em}

\begin{proof}[Proof of \cref{lem:ridgeless:tools}] To prove (i), we first note that for $\lambda > 0$,
\begin{align*}
    \big| \, f^{(1)}_\lambda(A) \,\big|
    \;=&\; 
    \lambda^2 
    \, \big| \beta^\top \big(A  + \lambda \bI_d \big)^{-2} \beta 
    \big|
    \\
    \;=&\;
    \msum_{l=1}^d 
    \mfrac{\lambda^2 \, \| \beta \|^2}{( \lambda_l(A) + \lambda)^2} 
    \, \ind_{\{ \lambda_l(A) \neq - \lambda \}}
    \;\leq\;
    \max_{ l \leq d ;\;  \lambda_l(A) \neq - \lambda} 
    \mfrac{\lambda^2 \| \beta \|^2 }{(\lambda_l(A) + \lambda)^2} 
     \;,
\end{align*}
whereas for $\lambda =0$, we have 
\begin{align*}
    \big| \, f^{(1)}_0(A) \,\big|
    \;=&\; 
    \big\| \big( A^\dagger A  - \bI_d \big) \beta \big\|^2
    \\
    \;=&\;
    \msum_{l=1}^d 
    \Big( 
        (0-1)^2 \ind_{\{ \lambda_l(A) = 0 \}}
        +
        (1-1)^2 \ind_{\{ \lambda_l(A) \neq 0 \}}
    \Big)
    \, \| \beta \|^2 
    \\
    \;=&\;
    \msum_{l=1}^d \ind_{\{ \lambda_l(A) = 0 \}} \| \beta \|^2\;.
\end{align*}
Meanwhile for $\lambda \geq 0$, we have
\begin{align*}
    \big| \, f^{(2)}_\lambda(A, B) \,\big|
    \;=&\; 
    \mfrac{\sigma_\epsilon^2}{n} \,
    \big| 
    \Tr\big( \big(A + \lambda \bI_d \big)^{-2} 
    B \big)
    \big|
    \\
    \;\leq&\;
    \mfrac{\sigma_\epsilon^2 \, \| B \|_{op}}{n} \,
    \Big| 
        \msum_{l=1}^d  \mfrac{\ind\{ \lambda_l(A) \neq - \lambda \}}{(\lambda_l(A) + \lambda )^2} \, 
    \Big|
    \;=\;
    \max_{l \leq d ;\;  \lambda_l(A) \neq - \lambda} 
    \; \mfrac{d \, \sigma_\epsilon^2  \, \| B \|_{op} }{n(\lambda_l(A) + \lambda)^2} 
    \;.
\end{align*}
To prove (ii), note that by assumption $\lambda > 0$. The first difference can be bounded as
\begin{align*}
    \big| f^{(1)}_\lambda(A) & - f^{(1)}_0(A) \big|
    \;=\;
    \big| \beta^\top 
    \big( \, 
        \big( A^\dagger A - \bI_d \big)^2
        -
        \lambda^2 \big( A + \lambda \bI_d \big)^{-2}
        \,
    \big)
    \beta \big|
    \\
    \;\leq&\; 
    \|\beta\|^2
    \,
    \big\|   
        \big( A^\dagger A - \bI_d \big)^2
        -
        \lambda^2 \big( A + \lambda \bI_d \big)^{-2}
    \big\|_{op}
    \\
    \;\overset{(a)}{=}&\;
    \|\beta\|^2
    \,
    \max\Big\{
    \Big| (-1)^2 - \mfrac{\lambda^2}{\lambda^2}\Big|  
    \,,\,  
    \mmax_{l \leq d;\, \lambda_l(A) \neq 0}  \Big| 0^2 - \mfrac{\lambda^2 \ind\{\lambda_l(A) \neq - \lambda \} }{(\lambda_l(A)+\lambda)^2} \Big|  \Big\} 
    \\
    \;\leq&\;
    \mmax_{l \leq d;\, \lambda_l(A) \not\in \{ 0, - \lambda\} } \mfrac{\lambda^2  \|\beta\|^2}{(\lambda_l(A)+\lambda)^2}
    \;.
\end{align*}
In $(a)$, we have noted that all matrices involved share the same set of eigenvectors. The second difference can be controlled as 
\begin{align*}
    \big| & f^{(2)}_\lambda(A , B) - f^{(2)}_0(A , B) \big|
    \;=\;
    \mfrac{\sigma_\epsilon^2}{n}  
    \Big|
        \Tr\Big( \, 
            \Big( 
                \big(A + \lambda \bI_d \big)^{-2}
                - 
                A^{-2}
            \Big)
        B \Big)  
    \Big|
    \\
    &\;\leq\;
    \mfrac{\sigma_\epsilon^2 }{n}  
        \msum_{l=1}^d
        \Big| 
            \Big(
                \mfrac{ \ind\{ \lambda_l(A) \neq - \lambda \}}{(\lambda_l(A) + \lambda )^2} 
                -
                \mfrac{ \ind\{ \lambda_l(A) \neq 0 \}}{\lambda_l(A)^2} 
            \Big)
            (v_l(A)^\top B \, v_l(A))
        \Big| 
    \\
    &\;\leq\;
    \mfrac{\sigma_\epsilon^2 }{n}  
    \msum_{l=1}^d 
    \Big( 
        \mfrac{ \ind\{ \lambda_l(A) \in \{0, - \lambda\} \}}{ \lambda^2}
        +
        \ind_{\{ \lambda_l(A) \not\in \{0, - \lambda\} \}} \,
        \mfrac{| \lambda^2 + 2 \lambda \lambda_l(A)   |}{\lambda_l(A)^2 ( \lambda_l(A) + \lambda)^2}
    \Big)
    |v_l(A)^\top B \, v_l(A)|
    \\
    &\;\leq\;
    \mfrac{\sigma_\epsilon^2}{n \lambda^2}  \msum_{l=1}^d  \ind_{\{ \lambda_l(A) \in \{0, - \lambda\} \}} |v_l(A)^\top B \, v_l(A)|
    \\
    &\qquad 
    \,+\,
    \mfrac{ \lambda d \, \sigma_\epsilon^2 \| B \|_{op} }{n} 
    \,
    \max_{l \leq d ;\, \lambda_l(A) \not\in \{0, - \lambda\} }
    \mfrac{| \lambda + 2 \lambda_l(A)   | \, }{\lambda_l(A)^2 ( \lambda_l(A) + \lambda)^2}
    \;.
\end{align*}
To prove (iii), we first note that by assumption, $\lambda_l(A) \geq - \lambda / 2 > - \lambda$ for all $l \leq d$, so the map $\tilde A \mapsto (\tilde A + \lambda \bI_d)^{-1}$ is smooth in the local neighbourhood of the line segment $[0, A]$; the same holds for $A'$. We can now apply the mean value theorem to $f^{(1)}_\lambda$ and $f^{(2)}_\lambda$ by computing their first derivatives: Writing $\tilde A_t = t (A- A') + A'$, we have
\begin{align*}
    \big| f^{(1)}_\lambda(A) - f^{(1)}_\lambda(A') \big|
    &\;\leq\;
    \msup_{t \in [0,1]}
    \Big| 
        \lambda^2 \beta^\top (\tilde A_t + \lambda \bI_d)^{-1} (A - A') 
        \big( \tilde A_t + \lambda \bI_d \big)^{-1} \beta
    \Big|
    \\
    &\;\leq\;
    \mfrac{\lambda^2 \| \beta \|^2 \| A - A' \|_{op} }
    {(\lambda / 2)^2} 
    \;=\; 
    4 \, \| \beta \|^2 \| A - A' \|_{op} \;.
\end{align*}
In the last line, we have noted that all eigenvalues of $t(A- A') + A'$ are bounded from below by $- \lambda / 2$. Similarly we have 
\begin{align*}
    &\;\big| f^{(2)}_\lambda(A, B) - f^{(2)}_\lambda(A', B) \big|
    \\
    &\;\leq\;
    \mfrac{\sigma_\epsilon^2}{n} \,
    \msum_{\substack{q_1, q_2 \in \N\\q_1 + q_2 = 3}}
    \;\,\sup_{t \in [0,1]}
    \big| 
    \Tr\big( (\tilde A_t + \lambda \bI_d)^{-q_1} (A-A') (\tilde A_t + \lambda \bI_d)^{-q_2} 
    B \big)
    \big| 
    \\
    &\;\leq\;
    \mfrac{ 2 \sigma_\epsilon^2 d }{n} 
    \| A - A' \|_{op} \big\|  (\tilde A_t + \lambda \bI_d)^{-1} \big\|_{op}^3 \| B \|_{op}
    \\
    &\;\leq\;
    \mfrac{ 16 \, \sigma_\epsilon^2  d }{n \lambda^3}  \| A - A' \|_{op} \| B \|_{op}\;.
\end{align*}
\end{proof}

\subsection{Proofs for \cref{appendix:ridgeless:results}} \label{appendix:ridgeless:results:proof}

The proof exploits the assumption below on the distribution of the extreme eigenvalues of $\bar \bX_1$, $\bar \bX_2$, $\bar \bZ_1$ and $\bar \bZ_2$, as well as the alignment of their zero eigenspace.

\begin{proof}[Proof of \cref{lem:ridgeless:by:ridge}] First note that by the triangle inequality, almost surely 
\begin{align*} 
    &\;
    \big| f_\lambda(\bar \bX_1, \bar \bX_2)  - f_0(\bar \bX_1, \bar \bX_2)  \big|
    \\
    &\;\leq\; 
    \big| f^{(1)}_\lambda(\bar \bX_1, \bar \bX_2)  - f^{(1)}_0(\bar \bX_1, \bar \bX_2)  \big|
    +
    \big| f^{(2)}_\lambda(\bar \bX_1, \bar \bX_2)  - f^{(2)}_0(\bar \bX_1, \bar \bX_2)  \big|
    \;.
\end{align*}
Applying \cref{lem:ridgeless:tools}(ii), we get that almost surely
\begin{align*}
    \big| f^{(1)}_\lambda(\bar \bX_1, \bar \bX_2)  - f^{(1)}_0(\bar \bX_1, \bar \bX_2)  \big|
    \;\leq\;
    \lambda^2  \|\beta\|^2 \, \max_{l \leq d;\, \lambda_l(\bar \bX_1) \not\in \{ 0, - \lambda\} } \mfrac{1}{(\lambda_l(\bar \bX_1)+\lambda)^2} \;,
\end{align*}
and
\begin{align*}
    \big| f^{(2)}_\lambda(\bar \bX_1, \bar \bX_2)  - f^{(2)}_0( & \bar \bX_1, \bar \bX_2)  \big|
    \;\leq\;
    \mfrac{\sigma_\epsilon^2  }{n \lambda^2}  
    \,
     \msum_{l=1}^d  \ind_{\{ \lambda_l(\bar \bX_1) \in \{0, - \lambda\} \}} \big( v_l(\bar \bX_1)^\top \bar \bX_2 \, v_l(\bar \bX_1) \big)
    \\
    &\;
    +
    \mfrac{ \lambda d \, \sigma_\epsilon^2 \, \| \bar \bX_2 \|_{op}  }{n} 
    \, 
        \max_{l \leq d ;\, \lambda_l(\bar \bX_1) \not\in \{0, - \lambda\} }
        \, \mfrac{| \lambda + 2 \lambda_l(\bar \bX_1)| }{\lambda_l(\bar \bX_1)^2 ( \lambda_l(\bar \bX_1) + \lambda)^2}
    \;.
    \tagaligneq \label{eq:ridgeless:by:ridge:f2}
\end{align*}
The above bound can be simplified by noting that all eigenvalues of $\bar \bX_1$ are non-negative, which implies that almost surely for all $1 \leq l \leq d$,
\begin{align*}
    &\;
    \mfrac{\ind\{\lambda_l(\bar \bX_1) \not\in \{ 0, - \lambda\} \}}{(\lambda_l(\bar \bX_1)+\lambda)^2}
    \;\leq\;
    \mfrac{\ind\{\lambda_l(\bar \bX_1) \neq 0 \}}{\lambda_l(\bar \bX_1)^2}
    \;\leq\;
    \big\| \bar \bX_1^\dagger \big\|_{op}^2 
    \;,\;
    \quad 
    \ind_{\{ \lambda_l(\bar \bX_1) \in \{0, - \lambda\} \}} 
    \;=\;
    \ind_{\{ \lambda_l(\bar \bX_1) = 0 \}} 
    \\
    &\;
    \mfrac{\ind\{\lambda_l(\bar \bX_1) \not\in \{ 0, - \lambda\} \} \times | \lambda + 2 \lambda_l(\bar \bX_1)   |}{\lambda_l(\bar \bX_1)^2 ( \lambda_l(\bar \bX_1) + \lambda)^2}
    \;\leq\;
    \mfrac{2 \, \ind\{\lambda_l(\bar \bX_1) \neq 0 \}}{\lambda_l(\bar \bX_1)^3}
    \;\leq\; 
    2 \big\| \bar \bX_1^\dagger \big\|_{op}^3\;.
\end{align*}
Combining the bounds above and applying 
\cref{assumption:ridgeless:by:ridge} 
gives that
\begin{align*}
    \big| f^{(1)}_\lambda(\bar \bX_1, \bar \bX_2)  - f^{(1)}_0(\bar \bX_1, \bar \bX_2)  \big|
    \;=&\;
    O_{\gamma'}( \lambda^2 )
    \;,
    \\
    \big| f^{(2)}_\lambda(\bar \bX_1, \bar \bX_2)  - f^{(2)}_0(\bar \bX_1, \bar \bX_2)  \big|
    \;=&\;
    O_{\gamma'}\Big( \lambda + \mfrac{1}{n \lambda^2} \Big)
    \;,
    \\
    \big| f_\lambda(\bar \bX_1, \bar \bX_2)  - f_0(\bar \bX_1, \bar \bX_2)  \big|
    \;=&\;
    O_{\gamma'} \Big( \lambda + \lambda^2 + \mfrac{1}{n \lambda^2}  \Big)
\end{align*}
with probability $1 - o_{\gamma'}(1)$. By the definition of the Lévy–Prokhorov metric $d_P$ \eqref{eq:defn:prokhorov}, we obtain 
\begin{align*}
    d_P\big(  f^{(1)}_\lambda(\bar \bX_1, \bar \bX_2)  \,,\, f^{(1)}_0(\bar \bX_1, \bar \bX_2)  \big)
    \;=&\; 
    O_{\gamma'} ( \lambda^2  )\;,
    \\
    d_P\big(  f^{(2)}_\lambda(\bar \bX_1, \bar \bX_2)  \,,\, f^{(2)}_0(\bar \bX_1, \bar \bX_2)  \big)
    \;=&\; 
    O_{\gamma'} \Big( \lambda + \mfrac{1}{n \lambda^2}  \Big)\;,
    \\
    d_P\big(  f_\lambda(\bar \bX_1, \bar \bX_2)  \,,\, f_0(\bar \bX_1, \bar \bX_2)  \big)
    \;=&\; 
    O_{\gamma'} \Big( \lambda + \lambda^2 + \mfrac{1}{n \lambda^2}  \Big)\;,
\end{align*}
which proves the first bound. The second bound follows from applying the same argument with $\bar \bX_1, \bar \bX_2$ replaced by $\bar \bZ_1, \bar \bZ_2$.
\end{proof}

\vspace{1em}

The next proof exploits orthogonal invariance of isotropic Gaussians.

\begin{proof}[Proof of \cref{lem:ridgeless:isoG:alt:rep}] Consider the $\R^{d \times nk}$-valued random matrix
\begin{align*}
    \bU  \;\coloneqq\; 
    \big( \bV_1 + \xi_{11},  \bV_1 + \xi_{12}, \ldots, \bV_n + \xi_{nk}
    \big)\;, 
\end{align*}    
We can then express
\begin{align*}
    \bar \bZ_1 \;=\; \mfrac{1}{nk} \bU \bU^\top \;.
\end{align*}
Notice that under 
\eqref{eq:ridgeless:isoG}, 
$\bU$ have i.i.d.~rows, each of which has a covariance matrix
\begin{align*}
    \bI_n \otimes \big( \bone_{k \times k} + \sigma_A^2 \bI_k \big)  
    \;=\; 
    \bI_n
    \otimes    
    k \, Q_k^\top D_k Q_k
    \;.
\end{align*}
This implies that we can express, for some choice of $\eta'_1, \ldots, \eta'_d \overset{\rm i.i.d.}{\sim} \cN(0, \bI_{nk})$, almost surely
\begin{align*}
    \bU 
    \;=\;
    \sqrt{k}
    \,
    \begin{psmallmatrix}
        \leftarrow\; (\eta'_1)^\top \;\rightarrow
        \\
        \vdots \\
        \leftarrow\; (\eta'_d)^\top \;\rightarrow
    \end{psmallmatrix} 
    ( \bI_n \otimes D_k^{1/2} Q_k )
    \;\eqqcolon\;
    \sqrt{k}
    \,
    \bH
    ( \bI_n \otimes D_k^{1/2} Q_k )
    \;,
\end{align*}
and therefore almost surely we have 
\begin{align*}
    \bar \bZ_1 
    \;=&\; 
    \mfrac{k}{nk} \bH \, \big( \bI_n \otimes D_k^{1/2} Q_k Q_k^\top D_k^{1/2} \big) \bH^\top
    \;=\; 
    \mfrac{1}{n} \bH \, \big( \bI_n \otimes D_k \big) \bH^\top
    \;,
\end{align*}
where $\bH$ is an $\R^{d \times nk}$ matrix with i.i.d.~standard Gaussian entries. Meanwhile, observing that 
\begin{align*}
    \bar \bZ_2 
    \;=&\; \mfrac{1}{n k} \bU K K^\top \bU^\top 
\end{align*}
proves the second statement. The final statement follows by identifying $\eta_{11}, \ldots, \eta_{nk}$ as the column vectors of $\bH$, which yields 
\begin{align*}
    \bar \bZ_1 \;=\; \mfrac{1}{n} \msum_{i=1}^n \Big( \mfrac{k+\sigma^2_A}{k} \eta_{i1} \eta_{i1}^\top + \mfrac{\sigma^2_A}{k} \msum_{j=2}^k \eta_{ij} \eta_{ij}^\top \Big)
    \;.
\end{align*}
By recalling that
\begin{align*}
    Q_k 
    \;\coloneqq&\; 
    \begin{psmallmatrix}
        k^{-1/2} & \hdots & k^{-1/2} \\
        \leftarrow  & \bv_1^\top & \rightarrow \\
        & \vdots & \\
        \leftarrow  & \bv_{k-1}^\top & \rightarrow \\
    \end{psmallmatrix}
    \;,
\end{align*}
and observing that 
\begin{align*}
    \big( \bV_1 + \xi_{11},  \bV_1 + \xi_{12}, \ldots, \bV_n + \xi_{nk}
    \big)
    \;=\;
    \bU 
    \;=\;
    \sqrt{k} \begin{psmallmatrix}
        \uparrow & & \uparrow \\
        \eta_{11} & \cdots & \eta_{nk} 
        \\
        \downarrow & & \downarrow \\
    \end{psmallmatrix}
    (\bI_n \otimes D_k^{1/2} Q_k)
    \;, 
\end{align*}
we obtain that
\begin{align*}
    \eta_{i1} \;=\; \mfrac{1}{k} \msum_{j=1}^k (\bV_i + \xi_{ij}) \times \mfrac{\sqrt{k}}{\sqrt{k+\sigma^2_A}}
\end{align*}
and therefore we can express 
\begin{align*}
    \bar \bZ_1 
    \;=&\; 
    \mfrac{1}{n} \sum_{i=1}^n 
    \Bigg( 
        \Bigg(  \mfrac{1}{k} \sum_{j=1}^k (\bV_i + \xi_{ij}) \Bigg) 
        \Bigg(  \mfrac{1}{k} \sum_{j=1}^k (\bV_i + \xi_{ij}) \Bigg)^\top
        +
        \mfrac{\sigma^2_A}{k}
        \sum_{j=2}^k \eta_{ij} \eta_{ij}^\top
    \Bigg)
    \\
    \;=&\;
    \bar \bZ_2 
    +
    \mfrac{\sigma^2_A}{nk} \msum_{i=1}^n
    \msum_{j=2}^k \eta_{ij} \eta_{ij}^\top
    \;.
\end{align*}
\end{proof}

\vspace{1em}

\begin{proof}[Proof of \cref{lem:ridgeless:isoG:assumption}] We first verify 
    \cref{assumption:ridgeless:by:ridge}. 
Under 
\eqref{eq:ridgeless:isoG}, 
we can apply \cref{lem:ridgeless:isoG:alt:rep} to express
\begin{align*}
    \bar \bZ_1 \;=\; \mfrac{1}{n} \bH \, \big( \bI_n \otimes D_k \big) \bH^\top\;,
\end{align*}
where $D_k \in \R^{k \times k}$ is a positive diagonal matrix with minimum eigenvalue $\sigma_A^2 / k > 0$ and $\bH$ is an $\R^{d \times nk}$ matrix with i.i.d.~standard Gaussian entries. Given a real symmetric matrix $A$, let $\sigma_{\rm min}(A)$ denote its minimum non-zero eigenvalue and $\sigma_{\rm min; > 0}(A)$ denote its minimum non-zero eigenvalue. Then almost surely
\begin{align*}
    \| \bar \bX_1^\dagger \|_{op}
    \;\overset{d}{=}\;
    \| \bar \bZ_1^\dagger \|_{op} 
    \;=&\;
    \big( \sigma_{\rm min; > 0} ( \bar \bZ_1) \big)^{-1}
    \\
    \;=&\;
    \Big( \sigma_{\rm min; > 0} \Big( \mfrac{1}{n} \big(\bI_n \otimes D_k^{1/2} \big) \bH \bH^\top \big(\bI_n \otimes D_k^{1/2} \big) \Big) \Big)^{-1}
    \\
    \;\leq&\;
    \mfrac{1}{\sigma_A^2}
    \Big( \sigma_{\rm min; > 0} \Big( \mfrac{1}{nk} \bH \bH^\top \Big) \Big)^{-1}
    \\
    \;=&\;
    \mfrac{1}{\sigma_A^2}
    \Big( \sigma_{\rm min; > 0} \Big( \mfrac{1}{nk} \msum_{l=1}^d \eta_l \eta_l^\top \Big) \Big)^{-1}
    \;,
\end{align*}
where $\eta_1, \ldots, \eta_d$ are some i.i.d.~standard Gaussian vectors in $\R^{nk}$. Meanwhile, by the minimum singular value bound from Theorem 6.1 of \cite{wainwright2019high}, for any fixed $\epsilon > 0$ and $nk \leq d$,
\begin{align*}
    \P\Big( 
        \sigma_{\rm min} \Big( \mfrac{1}{d} \msum_{l=1}^d \eta_l \eta_l^\top \Big)
        > 
        \Big( (1-\epsilon) - \mfrac{(nk)^{1/2}}{d^{1/2}} \Big)^2 
    \Big) 
    \;\geq\; 
    1 - e^{- d \epsilon^2/2}\;, 
\end{align*}
so if $nk \leq d$ with $\gamma' = \lim d /(kn) \in (1, \infty)$, we get that $\sigma_{\rm min} \big( \frac{1}{nk} \sum_{l=1}^d \eta_l \eta_l^\top \big)$ is bounded from below by some constant $c'_{\gamma'} \in (0, \infty)$ that only depends on $\gamma'$. This is still true if $nk \geq d$ with $\gamma' \in [0,1)$, since in this case 
\begin{align*}
    \sigma_{\rm min; > 0} \Big( \mfrac{1}{nk} \msum_{l=1}^d \eta_l \eta_l^\top \Big)
    \;=&\;  
    \sigma_{\rm min; > 0} \Big( 
    \mfrac{1}{nk} 
    \begin{psmallmatrix}
        \uparrow & & \uparrow \\
        \eta_1 & \hdots & \eta_d \\
        \downarrow & & \downarrow 
    \end{psmallmatrix} 
    \begin{psmallmatrix}
        \leftarrow & \eta_1^\top & \rightarrow \\ 
         & \vdots &  \\ 
        \leftarrow & \eta_d^\top & \rightarrow 
    \end{psmallmatrix} 
    \Big)
    \\
    \;=&\; 
    \sigma_{\rm min} \Big( 
        \mfrac{1}{nk} 
        \begin{psmallmatrix}
            \leftarrow & \eta_1^\top & \rightarrow \\ 
             & \vdots &  \\ 
            \leftarrow & \eta_d^\top & \rightarrow 
        \end{psmallmatrix} 
        \begin{psmallmatrix}
            \uparrow & & \uparrow \\
            \eta_1 & \hdots & \eta_d \\
            \downarrow & & \downarrow 
        \end{psmallmatrix} 
        \Big)
    \;\eqqcolon\;
    \sigma_{\rm min}( \bW_{nk} )
    \;,
\end{align*}
and the same argument applies to the $\R^{d \times d}$ Wishart matrix $\bW_{nk} $. This implies that $\| \bar \bX_1^\dagger \|_{op}$ and $\| \bar \bZ_1^\dagger \|_{op}$ are both $O_{\gamma'}(1)$ with probability $1-o_{\gamma'}(1)$ under the stated assumptions. 

\vspace{.5em}

Meanwhile, by \cref{lem:ridgeless:isoG:alt:rep} again, 
\begin{align*}
    \bar \bX_2 
    \;\overset{d}{=}\;
    \bar \bZ_2 
    \;\overset{a.s.}{=}\; \mfrac{1}{n} \bH \, \big( \bI_n \otimes D_k^{1/2} Q_k \big) K K^\top \big( \bI_n \otimes Q_k^\top D_k^{1/2} \big) \bH^\top 
\end{align*}
where $Q_k \in \R^{k \times k}$ is an orthogonal matrix. Therefore almost surely 
\begin{align*}
    \big\|  \bar \bZ_2  \big\|_{op} 
    \;\leq\; 
    \sigma_{\rm max}\big( K K^\top \big)
    \sigma_{\rm max}\big( \bar \bZ_1 \big)
    \;\leq\; 
    \mfrac{k+\sigma_A^2}{k} \times 
    \sigma_{\rm max} \Big( \mfrac{1}{n} \msum_{l=1}^d \eta_l \eta_l^\top \Big)
    \;,
    \tagaligneq \label{eq:ridgeless:Z2:op:control}
\end{align*}
where we have recalled from the definitions in \cref{lem:ridgeless:isoG:alt:rep} that 
\begin{align*}
    \sigma_{\rm max}(K K^\top) \;=&\; \sigma_{\rm max}\Big( \mfrac{1}{k} \bI_n \otimes \bone_{k \times k} \Big) \;=\; 1
    &\text{ and }&&
    \big\| \bI_n \otimes D_k^{1/2} Q_k \big\| \;\leq\; \sqrt{ \mfrac{k + \sigma_A^2}{k}}\;.
\end{align*}
Applying the maximum singular value bound from Theorem 6.1 of \cite{wainwright2019high} to $\frac{1}{nk} \sum_{l=1}^d \eta_l \eta_l^\top$ implies that $\| \bar \bX_2 \|_{op}$ is $O_{\gamma'}(1)$ with probability $1 - o_{\gamma '}(1)$ provided that $nk \leq d$ with $\gamma' = \lim d/nk > 1$, and by noting again that 
\begin{align*}
    \sigma_{\rm max} \Big( \mfrac{1}{nk} \msum_{l=1}^d \eta_l \eta_l^\top \Big)
    \;=&\; 
    \sigma_{\rm max} (\bW_{nk} )
\end{align*}
for the $\R^{d \times d}$ Wishart matrix $\bW_{nk}$, we get that the same holds when $nk \geq d$ with $\gamma' = \lim d/nk < 1$. This implies that  $\| \bar \bX_2 \|_{op}$ and $\| \bar \bZ_2 \|_{op}$ are both $O_{\gamma'}(1)$ with probability $1-o_{\gamma'}(1)$ under the stated assumptions. 

\vspace{.5em}

The final quantity in 
\cref{assumption:ridgeless:by:ridge} 
can be expressed as
\begin{align*}
    \msum_{l=1}^d  \ind_{\{ \lambda_l(\bar \bX_1) = 0 \}} \big( v_l(\bar \bX_1)^\top \bar \bX_2 v_l(\bar \bX_1) \big)
    \;\overset{d}{=}&\; 
    \msum_{l=1}^d  \ind_{\{ \lambda_l(\bar \bZ_1) = 0 \}} \big( v_l(\bar \bZ_1)^\top \bar \bZ_2 v_l(\bar \bZ_1) \big)
    \\
    \;=&\;
    \msum_{l=1}^d  \ind\Big\{ \lambda_l\Big(\mfrac{1}{nk} \msum_{l=1}^d \eta_l \eta_l^\top\Big) = 0 \Big\}
    \;.
\end{align*}
Since $\bar \bZ_1 = \frac{1}{n} \bH (\bI_n \otimes D_k) \bH^\top$, where $\bI_n \otimes D_k$ is positive-definite, if  $v_l(\bar \bZ_1)$  is a zero eigenvector of $\bar \bZ_1$, then we must have $\bH^\top v_l(\bar \bZ_1) = \bzero$ almost surely. This implies 
\begin{align*}
    v_l(\bar \bZ_1)^\top \bar \bZ_2 \, v_l(\bar \bZ_1) 
    \;=\;
    \mfrac{1}{n} 
    v_l(\bar \bZ_1)^\top  \bH \, \big( \bI_n \otimes D_k^{1/2} Q_k \big) K K^\top \big( \bI_n \otimes Q_k^\top D_k^{1/2} \big) \bH^\top   v_l(\bar \bZ_1) 
    \;=\; 0
\end{align*}
almost surely, and therefore with probability $1 - o(1)$,
\begin{align*}
    \msum_{l=1}^d  \ind_{\{ \lambda_l(\bar \bX_1) = 0 \}} \big( v_l(\bar \bX_1)^\top \bar \bX_2 v_l(\bar \bX_1) \big)
    \;=\; 
    \msum_{l=1}^d  \ind_{\{ \lambda_l(\bar \bZ_1) = 0 \}} \big( v_l(\bar \bZ_1)^\top \bar \bZ_2 v_l(\bar \bZ_1) \big)
    \;=\;
    0
    \;.
\end{align*}
This verifies 
\cref{assumption:ridgeless:by:ridge}.

\vspace{.5em}

To verify 
\cref{assumption:ridgeless:universal}, 
we first note that since the entries of the matrices are all Gaussian,  we automatically have $\max_{i \leq n, j \leq k, l \leq d} \| X_{ijl} \|_{L_{10}} = O(1)$. Meanwhile by \eqref{eq:ridgeless:Z2:op:control},
\begin{align*}
    \big\| \| \bar \bX_2 \|_{op} \big\|_{L_{60}}
    \;=\;
    \big\| \| \bar \bZ_2 \|_{op} \big\|_{L_{60}}
    \;\leq&\;
    \mfrac{k+\sigma_A^2}{k} \,
    \Big\| \Big\| \mfrac{1}{nk} \msum_{l=1}^d \eta_l \eta_l^\top \Big\|_{op} \Big\|_{L_{60}}
    \;=\;
    \mfrac{k+\sigma_A^2}{k} \, \big\| \big\| \bW_{nk} \big\|_{op} \big\|_{L_{60}}
\end{align*}    
where $\bW_{nk}$ is the $\R^{d \times d}$ Wishart matrix defined above. By Theorem 4.6.1 of \cite{vershynin2018high}, there exists some constant $C_1 > 0$ such that, for all $t > 0$,
\begin{align*}
    \P\Big(  \big\| \bW_{nk} - \bI_d \big\|_{op}  > 2 C_1 \mfrac{\sqrt{d} \, + t }{\sqrt{nk}}  + C_1^2 \mfrac{(\sqrt{d} \, + t)^2}{nk}
    \Big) 
    \;\leq\; 
    2 \exp(-t^2)\;.
\end{align*}
Using that $d/(kn) = O(1)$, we get that for every fixed $m \in \N$, there exists some constant $C_m > 0$ depending on $m$ such that
\begin{align*}
    \mean\big[ \| \bar \bZ_2 - \bI_d \|_{op}^m \big] 
    \;\leq\; 
    \mint_0^\infty \P\big(\| \bW_{nk} - \bI_d \|_{op} > s^{1/m} \big) ds
    \;\leq\;
    C_m\;.
\end{align*}
This implies
\begin{align*}
    \big\| \| \bar \bX_2 \|_{op} \big\|_{L_{60}}
    \;=\;
    \big\| \| \bar \bZ_2 \|_{op} \big\|_{L_{60}}
    \;\leq\;
    \big\| \| \bW_{nk} - \bI_d \|_{op} \big\|_{L_{60}} + \| \bI_d \|_{op}
    \;=\; O(1)\;,
\end{align*}
which verifies 
 \cref{assumption:ridgeless:universal}.
\end{proof}

\vspace{1em}

\subsection{Proof of Proposition 10: Universality for oracle augmentation} 
\label{appendix:proof:ridgeless:universality}

The proof adapts the two-moment matching argument from 
\cref{thm:main} 
to utilize the matching of four moments. Write $X_{ijl}$ as the $l$-th coordinate of $\pi_{ij} \bV_i$ for simplicity.  For $1 \leq i \leq n$ and $1 \leq l \leq d$, define the $\R^k$ vectors
    \begin{align*}
        \tilde \bX_{il} \;\coloneqq&\; (X_{i1l} \,,\, \ldots \,,\, X_{ikl} ) 
        &\text{ and }&&
        \tilde \bZ_{il} \;\coloneqq&\; (Z_{i1l} \,,\, \ldots \,,\, Z_{ikl} ) \;.
    \end{align*}
    We also rewrite
    \begin{align*}
        \bar \bX_1
        \;=&\;
        \mfrac{1}{nk} \msum_{i=1}^n \msum_{l_1, l_2=1}^d \tilde \bX_{il_1}^\top \tilde \bX_{il_2} \be_{l_1} \be_{l_2}^\top 
        \;\eqqcolon\;
        S_1( \tilde \bX_{11}, \ldots, \tilde \bX_{nd} )
        \;,
        \\
        \bar \bX_2 
        \;=&\;
        \mfrac{1}{nk^2} \msum_{i=1}^n \msum_{l_1, l_2=1}^d \msum_{j_1, j_2=1}^k  \, X_{ij_1l_1}  \, X_{ij_2l_2}  \, \be_{l_1} \be_{l_2}^\top 
        \;\eqqcolon\;
        S_2( \tilde \bX_{11}, \ldots, \tilde \bX_{nd} )
        \;,
        \\
        \bar \bZ_1
        \;=&\;
        S_1( \tilde \bZ_{11}, \ldots, \tilde \bZ_{nd} )
        \,,
        \qquad
        \bar \bZ_2 
        \;=\;
        S_2( \tilde \bZ_{11}, \ldots, \tilde \bZ_{nd} )
        \;.
    \end{align*}
    As mentioned in \cref{remark:main:non:iid}, 
     \Cref{thm:main} 
     can be directly extended to the independent but non-i.i.d.~case, and we shall use it to replace the sequence of independent vectors $(\tilde \bX_{11}, \ldots, \tilde \bX_{nd})$ by $(\tilde \bZ_{11}, \ldots, \tilde \bZ_{nd})$ (note that in this case, $k$ in 
     \cref{thm:main} 
     is set to $1$). We also seek to exploit the fact that $\tilde \bX_{ij}$ and $\tilde \bZ_{ij}$ matches in the first four moments by assumption. By replacing the third-order Taylor expansion in 
     \cref{thm:main} 
     by a fifth-order Taylor expansion and a fifth-order Fa\`a di Bruno's formula, we obtain that  
    \begin{align*}
        &\;
        d_{\tilde \cH}\big( f_\lambda(\bar \bX_1, \bar \bX_2) \,,\, f_\lambda(\bar \bZ_1 , \bar \bZ_2) \big)
        \\
        &\;\leq\;
        \sum_{i=1}^n \sum_{l=1}^d 
        \mfrac{\sqrt{ \mean \big( \sum_{j=1}^k X_{ijl}^2 \big)^5  } + \sqrt{ \mean \big( \sum_{j=1}^k Z_{ijl}^2 \big)^5  } }{120}
        \\
        &\qquad
        \times 
        \big( 
            \theta_{1;10;X}^5  
            + 10 \theta_{1;8;X}^3\theta_{2;8;X}
            + 10 \theta_{1;6;X}^2\theta_{3;6;X}
            + 15 \theta_{1;6;X} \theta_{2;6;X}^2
            + 10 \theta_{2;4;X} \theta_{3;4;X}
            \\
        & \hspace{4em}
            + 5 \theta_{1;4;X}\theta_{4;4;X}   
            + \theta_{5;2;X}
        \\
        &\hspace{4em}
            +
            \theta_{1;10;Z}^5  
            + 10 \theta_{1;8;Z}^3\theta_{2;8;Z}
            + 10 \theta_{1;6;Z}^2\theta_{3;6;Z}
            + 15 \theta_{1;6;Z} \theta_{2;6;Z}^2
            + 10 \theta_{2;4;Z} \theta_{3;4;Z}
            \\
        & \hspace{4em}
            + 5 \theta_{1;4;Z}\theta_{4;4;Z}   
            + \theta_{5;2;Z}
        \big)
        \;,
    \end{align*}
    where, for $m \geq 2$, $q \in \N$ and $r \in \{1,2\}$, we define
    \begin{align*}
        &\;
        \theta_{q;m;X}
        \;\coloneqq\;
        \max_{i \leq n, l \leq d} \,
            \Big\|
                \,
                \Big\| \partial_{il}^q \, f_\lambda\Big( \bar \bW^{(1)}_{il}(\Theta \tilde \bX_{il}), \bar \bW^{(2)}_{il}( \Theta \tilde \bX_{il}) \Big)
                \Big\|
                \,
            \Big\|_{L_m}
        \;,
        \\
        &\;
        \theta_{q;m;Z}
        \;\coloneqq\;
        \max_{i \leq n, l \leq d} \,
        \Big\| 
            \,
            \Big\| \partial_{il}^q \, f_\lambda\Big( \bar \bW^{(1)}_{il}( \Theta \tilde \bZ_{il}), \bar \bW^{(2)}_{il}( \Theta \tilde \bZ_{il}) \Big)
            \Big\|
            \,
        \Big\|_{L_m}
        \;,
        \\
        &\;
        \bar \bW^{(r)}_{il}(\bx)
        \;\coloneqq\; 
        S_r\big( \tilde \bX_{ \leq il}, \bx, \tilde \bZ_{\geq il} \big)\;.
    \end{align*}
    $\Theta \sim \textrm{Uniform}[0,1]$ is independent of all other random variables, $\tilde \bX_{ \leq il}$ is the sequence formed by $\tilde \bX_{i'l'}$'s such that $(i', l')$ is before $(i,l)$ in the lexicographical order, and $\tilde \bZ_{\geq il}$ corresponds to  $\tilde \bZ_{i'l'}$'s such that $(i', l')$ comes after $(i,l)$. Now note that by the Jensen's inequality, we have 
    \begin{align*}
        \sqrt{ \mean \big( \msum_{j=1}^k X_{ijl}^2 \big)^5  }
        \;=&\;
        k^{5/2} \sqrt{ \mean \big( \mfrac{1}{k} \msum_{j=1}^k X_{ijl}^2 \big)^5  }
        \;\leq\;
        k^{5/2} \max_{j \leq k} \| X_{ijl} \|_{L_{10}}^5
        \;\leq\; 
        k^{5/2} c_0^5
        \;,
    \end{align*}
    where we have used 
     \cref{assumption:ridgeless:universal}
      for the last inequality. Similarly
    \begin{align*}
        \sqrt{ \mean \big( \msum_{j=1}^k X_{ijl}^2 \big)^5  }
        \;\leq\;
        k^{5/2} \max_{j \leq k} \| Z_{ijl} \|_{L_{10}}^5
        \;\leq\; 
        C' k^{5/2} c_0^5
    \end{align*} 
    for some absolute constant $C' > 0$; in the bound above, we have used that $Z_{ijl}$ matches $X_{ijl}$ in the first two moments, the moment formula of a Gaussian and that $\| X_{ijl}\|_{L_1} \leq \| X_{ijl}\|_{L_2} \leq \| X_{ijl}\|_{L_{10}}$. This implies that for some absolute constant $C'' > 0$, we have 
    \begin{align*}
        d_\cH\big( f_\lambda(\bar \bX_1, & \bar \bX_2) \,,\, f_\lambda(\bar \bZ_1 , \bar \bZ_2) \big)
        \\
        \;\leq\;
        C'' n d k^{5/2}&
        \big( 
            \theta_{1;10;X}^5  
            + 10 \theta_{1;8;X}^3\theta_{2;8;X}
            + 10 \theta_{1;6;X}^2\theta_{3;6;X}
            + 15 \theta_{1;6;X} \theta_{2;6;X}^2
            + 10 \theta_{2;4;X} \theta_{3;4;X}
            \\
        &\quad\hspace{-.5em}
            + 5 \theta_{1;4;X}\theta_{4;4;X}   
            + \theta_{5;2;X}
        \\
        &\quad\hspace{-.5em}
            +
            \theta_{1;10;Z}^5  
            + 10 \theta_{1;8;Z}^3\theta_{2;8;Z}
            + 10 \theta_{1;6;Z}^2\theta_{3;6;Z}
            + 15 \theta_{1;6;Z} \theta_{2;6;Z}^2
            + 10 \theta_{2;4;Z} \theta_{3;4;Z}
            \\
        &\quad\hspace{-.5em}
            + 5 \theta_{1;4;Z}\theta_{4;4;Z}   
            + \theta_{5;2;Z}
        \big)
        \;.
        \tagaligneq \label{eq:ridgeless:universality:intermediate}
    \end{align*}
    
    \vspace{.5em}
    
    The remaining proof controls the derivatives. We will perform a detailed calculation of the first derivative, comment on the shared pattern and state the remaining derivatives. We first write $x_{ijl}$ as the $l$-th coordinate of $\bx_{ij}$ and note that
    \begin{align*}
        \mfrac{\partial S_1(\bx_{11}, \ldots, \bx_{nd})}{\partial x_{ijl}}
        \;=&\;
        \mfrac{1}{nk} \msum_{l'=1}^d x_{ijl'} \big( \be_l \be_{l'}^\top + \be_{l'} \be_l^\top \big)
        \;=\;
        \mfrac{1}{nk} \big( \be_l \begin{psmallmatrix}x_{ij1} \\ \vdots \\ x_{ijd} \end{psmallmatrix}^\top + \begin{psmallmatrix}x_{ij1} \\ \vdots \\ x_{ijd} \end{psmallmatrix} \be_l^\top \big)
        \;,
        \\
        \mfrac{\partial^2 S_1(\bx_{11}, \ldots, \bx_{nd})}{\partial x_{ijl}^2}
        \;=&\;
        \mfrac{1}{nk} \big( \be_l \be_{l'}^\top + \be_{l'} \be_l^\top \big)
        \;,
        \qquad 
        \mfrac{\partial^3 S_1(\bx_{11}, \ldots, \bx_{nd})}{\partial x_{ijl}^3}
        \;=\; 
        \bzero\;,
        \\
        \mfrac{\partial S_2(\bx_{11}, \ldots, \bx_{nd})}{\partial x_{ijl}}
        \;=&\;
        \mfrac{1}{nk^2} \msum_{l'=1}^d \msum_{j'=1}^k  x_{ij'l'}  \big( \be_l \be_{l'}^\top + \be_{l'} \be_l^\top \big)
        \\
        \;=&\;
        \mfrac{1}{nk^2} \msum_{j'=1}^k  
        \big( \be_l \begin{psmallmatrix}x_{ij'1} \\ \vdots \\ x_{ij'd} \end{psmallmatrix}^\top + \begin{psmallmatrix}x_{ij'1} \\ \vdots \\ x_{ij'd} \end{psmallmatrix} \be_l^\top \big)
        \;,
        \\
        \mfrac{\partial^2 S_2(\bx_{11}, \ldots, \bx_{nd})}{\partial x_{ijl}^2}
        \;=&\;
        \mfrac{1}{nk^2} \big( \be_l \be_{l'}^\top + \be_{l'} \be_l^\top \big)
        \;,
        \qquad 
        \mfrac{\partial^3 S_2(\bx_{11}, \ldots, \bx_{nd})}{\partial x_{ijl}^3}
        \;=\; 
        \bzero\;.
    \end{align*}
    Meanwhile, since $\bar \bW^{(1)}_{il}(t \tilde \bX_{il})$ is positive semi-definite almost surely for all $t \in [0,1]$, the map $A \mapsto (A + \lambda \bI)^{-1}$ is differentiable in the local neighborhood of the line segment $[0,\bar \bW^{(1)}_{il}(t \tilde \bX_{il})]$ with respect to the Euclidean norm. For positive semi-definite matrix $A \in \R^{d \times d}$ and another matrix $B \in \R^{d \times d}$, denoting $A_\lambda \coloneqq A + \lambda \bI_d$, we can compute 
    \begin{align*}
        \mfrac{\partial f^{(1)}_\lambda(A)}{\partial A_{ij}}
        \;=&\;
        - \msum_{\substack{q_1, q_2 \in \N\\q_1 + q_2 = 3}} \lambda^2 \beta^\top A_\lambda^{-q_1} E_{ij} A_\lambda^{-q_2} \beta
        \;,
        \\
        \mfrac{\partial f^{(2)}_\lambda(A, B)}{\partial A_{ij}}
        \;=&\;
        \mfrac{\sigma_\epsilon^2}{n} \,
        \msum_{\substack{q_1, q_2 \in \N\\q_1 + q_2 = 3}}
        \Tr\big( A_\lambda^{-q_1} E_{ij} A_\lambda^{-q_2} 
        B \big) 
        \;, 
        \qquad
        \mfrac{\partial f^{(2)}_\lambda(A, B)}{\partial B_{ij}}
        \;=\;
        \mfrac{\sigma_\epsilon^2}{n} \,
        \Tr\big( A_\lambda^{-2} 
        E_{ij} \big) \;.
    \end{align*}
    Fix $m \in [2, 10]$. Using a chain rule with the derivatives computed above, we can calculate 
    \begin{align*}
        &\;
        \theta_{1;m;X}
        \;=\;
            \Big\|
                \, 
                \Big\| \partial_{il} \, f_\lambda\Big( \bar \bW^{(1)}_{il}(\Theta \tilde \bX_{il}), \bar \bW^{(2)}_{il}(\Theta \tilde \bX_{il}) \Big)
                \Big\|
                \,
            \Big\|_{L_m}
        \\
        &\;\leq\;
            \Big\|
                \,
                \Big\| \partial_{il} \, f^{(1)}_\lambda\Big( \bar \bW^{(1)}_{il}(\Theta \tilde \bX_{il}) \Big)
                \Big\|
                \,
            \Big\|_{L_m}
            +
            \Big\|
                \,
                \Big\| \partial_{il} \, f^{(2)}_\lambda\Big( \bar \bW^{(1)}_{il}( \Theta \tilde \bX_{il}), \bar \bW^{(2)}_{il}( \Theta \tilde \bX_{il}) \Big)
                \Big\|
                \,
            \Big\|_{L_m}
        \\
        &\;=\;
        \bigg\|
            \,
            \bigg(
                \sum_{j=1}^k
                \Big( 
                    -\lambda^2 
                    \msum_{\substack{q_1, q_2 \in \N\\q_1 + q_2 = 3}}
                    \beta^\top 
                    \big( \bar \bW^{(1)}_{il}(\Theta \tilde \bX_{il}) + \lambda \bI_d \big)^{-q_1}
                    \mfrac{1}{nk} \Theta \big( \be_l (\pi_{ij} \bV_i)^\top + (\pi_{ij} \bV_i)\be_l^\top \big)
        \\
        &\hspace{10em} 
                    \big( \bar \bW^{(1)}_{il}(\Theta \tilde \bX_{il}) + \lambda \bI_d \big)^{-q_2}
                    \beta
                \Big)^2
            \bigg)^{1/2}
            \,
        \bigg\|_{L_m}
        \\
        &\;\qquad
        +
        \bigg\|
        \, 
        \bigg(
            \sum_{j=1}^k
            \Big( 
                \mfrac{\sigma_\epsilon^2}{n} \,
                \msum_{\substack{q_1, q_2 \in \N\\q_1 + q_2 = 3}}
                \Tr\big( \big( \bar \bW^{(1)}_{il}(\Theta \tilde \bX_{il}) + \lambda \bI_d \big)^{-q_1} 
        \\
        &\hspace{5em}
           \times 
           \mfrac{1}{nk} \Theta \big( \be_l (\pi_{ij} \bX_i)^\top + (\pi_{ij} \bX_i) \be_l^\top \big)
           \times 
           \big( \bar \bW^{(1)}_{il}(\Theta \tilde \bX_{il}) + \lambda \bI_d \big)^{-q_2} \bar \bW^{(2)}_{il}(\Theta \tilde \bX_{il}) \big)
        \\
        &\hspace{2.5em}
            +
            \mfrac{\sigma^2_\epsilon}{n} \Tr\Big( 
                \big( \bar \bW^{(1)}_{il}(\Theta \tilde \bX_{il}) + \lambda \bI_d \big)^{-2} 
                \mfrac{1}{nk^2} \sum_{j'=1}^k \Theta \big( \be_l (\pi_{ij'} \bX_i)^\top + (\pi_{ij'} \bX_i) \be_l^\top \big)
            \Big)
        \;
            \Big)^2
            \bigg)^{1/2}
            \,
        \bigg\|_{L_m}
        \\
        &\;\leq\;
        \mfrac{2 \lambda^2 \, \| \beta \|^2}{n k} \,
        \Big\|  \Big\| \big( \bar \bW^{(1)}_{il}(\Theta \tilde \bX_{il}) + \lambda \bI_d \big)^{-1} \Big\|_{op}^3
        \times
        \Big( \msum_{j=1}^k  \| \pi_{ij} \bV_i \|^2 \Big)^{1/2} \Big\|_{L_m}
        \\
        &\;\;
        +
        \mfrac{4 \sigma^2_\epsilon}{n^2 k}
        \, 
        \Big\| 
        \Big\| \big( \bar \bW^{(1)}_{il}(\Theta \tilde \bX_{il}) + \lambda \bI_d \big)^{-1} \Big\|_{op}^3
        \times
        \Big\|  \bar \bW^{(2)}_{il}(\Theta \tilde \bX_{il}) \Big\|_{op} 
        \\
        &
        \hspace{5em}
        \times 
        \Big( \msum_{j=1}^k \| \pi_{ij} \bX_i \|^2 \Big)^{1/2} \, \Big\|_{L_m}
        \\
        &\;\;
        +
        \mfrac{2 \sigma^2_\epsilon}{n^2 k^{3/2}}
        \Big\|  \Big\| \big( \bar \bW^{(1)}_{il}(\Theta \tilde \bX_{il}) + \lambda \bI_d \big)^{-1} \Big\|_{op}^2
        \times 
        \Big( \msum_{j'=1}^k \| \pi_{ij'} \bX_i \| \Big) \Big\|_{L_m}
        \;.
    \end{align*}
    To simplify this bound, notice that since $\bar \bW^{(1)}_{il}(\Theta \tilde \bX_{il})$ is positive semi-definite, almost surely
    \begin{align*}
        \Big\| \big( \bar \bW^{(1)}_{il}(\Theta \tilde \bX_{il}) + \lambda \bI_d \big)^{-1} \Big\|_{op} 
        \;\leq\; 
        \mfrac{1}{\lambda}\;.
    \end{align*}
    Meanwhile since $m \geq 2$, by the Jensen's inequality, 
    \begin{align*}
        \Big\| \Big( \msum_{j=1}^k  \| \pi_{ij} \bV_i \|^2 & \Big)^{1/2} \Big\|_{L_m}
        \;=\;
        k^{1/2} \Big(\mean\Big[ \Big( \mfrac{1}{k} \msum_{j=1}^k  \| \pi_{ij} \bV_i \|^2 \Big)^{m/2} \Big] \Big)^{1/m}
        \\
        \;\leq&\;
        k^{1/2} \,\mmax_{j \leq k} \Big( \mean\Big[ \| \pi_{ij} \bV_i \|^m \Big] \Big)^{1/m}
        \\
        \;=&\;
        d^{1/2} k^{1/2} \,\mmax_{j \leq k} \Big( \mean\Big[ \Big( \mfrac{1}{d} \msum_{l=1}^d X_{ijl}^2 \Big)^{m/2} \Big] \Big)^{1/m}
        \\
        \;\leq&\;
        d^{1/2} 
        k^{1/2}\, \mmax_{i \leq n, j \leq k, l \leq d} \| X_{ijl} \|_{L_{m}}
        =
        O( d^{1/2} k^{1/2})
        \;,
    \end{align*}
    where we have applied 
    \cref{assumption:ridgeless:universal} 
    by noting that $m \leq 12$. Similarly
    \begin{align*}
        \Big\| \, \msum_{j'=1}^k \| \pi_{ij'} \bX_i \| \, \Big\|_{L_m}
        \;=&\;
        O( d^{1/2} )
        \;.
    \end{align*}
    Applying 
    \cref{assumption:ridgeless:universal} 
    again and noting that $|\Theta| \leq 1$ almost surely, we have
    \begin{align*}
        \Big\| \, \Big\|  \bar \bW^{(2)}_{il}(\Theta \tilde \bX_{il}) \Big\|_{op}  \, \Big\|_{L_m}
        \;\leq&\;
        \Big\| 
            \Big\| 
                    \mfrac{1}{n} \msum_{i'=1}^{i-1} 
                    \Big( \mfrac{1}{k} \msum_{j=1}^k (\pi_{i'j} \bV_i)  \Big)
                    \Big( \mfrac{1}{k} \msum_{j=1}^k (\pi_{i'j} \bV_i) \Big)^\top
            \Big\|_{op} 
        \Big\|_{L_m}
        \\
        &\; 
        +
        \Big\| 
            \Big\| 
                \mfrac{\Theta^2}{n}
                \Big( \mfrac{1}{k} \msum_{j=1}^k (\pi_{ij} \bV_i)  \Big)
                \Big( \mfrac{1}{k} \msum_{j=1}^k (\pi_{ij} \bV_i) \Big)^\top
            \Big\|_{op} 
        \Big\|_{L_m}
        \\
        &\; 
        +
        \Big\| 
            \Big\| 
                    \mfrac{1}{n} \msum_{i'=i+1}^{n} 
                    \Big( \mfrac{1}{k} \msum_{j=1}^k \bZ_i  \Big)
                    \Big( \mfrac{1}{k} \msum_{j=1}^k \bZ_i \Big)^\top
            \Big\|_{op} 
        \Big\|_{L_m}
        \\
        \;\leq&\; 
        \mfrac{i-1}{n} c_0 \,+\, \mfrac{1}{n} c_0 \,+\, \mfrac{n-i}{n} c_0 
        \;=\; O(1)
        \;.
    \end{align*} 
    Combining the above calculations and noting additionally that $\| \beta\|=O(1)$, $\sigma_\epsilon = O(1)$ and $d=O(n)$, we get that the first derivative term can be bounded as 
    \begin{align*}
        \theta_{1;m;X}
        \;=\;
        O\Big( 
            \mfrac{d^{1/2} \lambda^{-1}}{nk^{1/2}}
            +
            \mfrac{d^{1/2}(\lambda^{-3} + \lambda^{-2})}{n^2 k^{1/2}}
        \Big)
        \;=\;
        O\Big( 
            \mfrac{\max\{1, \lambda^{-3}\}}{n^{1/2} k^{1/2}}
        \Big)
        \;.
    \end{align*}
    By using the same argument and additionally bounding $\| Z_{ijl} \|_{L_m}$ by $C'' \| X_{ijl} \|_{L_m}$ for some absolute constant $C''$, we also have 
    \begin{align*}
        \theta_{1;m;Z}
        \;=\;
        O\Big( 
            \mfrac{\max\{1, \lambda^{-3}\}}{n^{1/2} k^{1/2}}
        \Big)
        \;.
    \end{align*}
    
    \vspace{.5em}
    
    To handle the higher-order derivative terms up to the fifth order, notice that in the above calculation, differentiating $f^{(1)}_\lambda$ and $f^{(2)}_\lambda$ with respect to $\bar \bW^{(1)}_{il}(\Theta \tilde \bX_{il})$ results in
    \begin{itemize}
        \item an additional $(\bar \bW^{(1)}_{il}(\Theta \tilde \bX_{il}) + \lambda \bI_d )^{-1}$ term, which contributes an $1/\lambda$ factor, and
        \\[.1em]
        \item an additional $\mfrac{\partial \bar \bW^{(1)}_{il}(\Theta \tilde \bX_{il})}{\partial X_{ijl}} = \mfrac{\Theta}{nk}\big( \be_l (\pi_{ij} \bX_i)^\top + (\pi_{ij}\bX_i) \be_l^\top ) $ term, which contributes an $d^{1/2}/nk$ factor,
    \end{itemize}
    whereas differentiating $f^{(2)}_\lambda$ with respect to $\bar \bW^{(2)}_{il}(\Theta \tilde \bX_{il})$ results in 
    \begin{itemize}
        \item an additional $\Big\|  \bar \bW^{(2)}_{il}(\Theta \tilde \bX_{il}) \Big\|_{op}$ term, which is $O(1)$, and
        \item an additional $\mfrac{\partial \bar \bW^{(2)}_{il}(\Theta \tilde \bX_{il})}{\partial X_{ijl}} = \mfrac{\Theta}{nk^2} \sum_{j'=1}^k \big( \be_l (\pi_{ij'} \bX_i)^\top + (\pi_{ij'}\bX_i) \be_l^\top ) $ term, which contributes an $d^{1/2}/nk$ factor.
    \end{itemize}
    We also note a few additional points:
    \begin{itemize}
        \item The initial sizes of $f^{(1)}_\lambda$ and $f^{(2)}_\lambda$ before differentiation are $O(1)$ and $O(n^{-1}\lambda^{-2})$ respectively, and that the norm we compute in $\theta_{q;m;X}$ has a persisting $k^{1/2}$ factor;
        \item The higher derivatives will also involve higher derivatives of  $\bar \bW^{(1)}_{il}(\Theta \tilde \bX_{il})$ and $\bar \bW^{(2)}_{il}(\Theta \tilde \bX_{il})$ with respect to $X_{ijl}$. But since the third derivatives vanish, the only additional terms are their second derivatives, which brings the sizes of the first derivatives down from $O(d^{1/2}/nk)$ down to $O(1/nk)$;
        \item The $q$-th derivative involves at most one copy of $\bar \bW^{(2)}_{il}(\Theta \tilde \bX_{il})$ and $q$ copies of $\pi_{ij} \bV_i$, so the bounding constant involves at most $(q+1)m$-th moments of $\bar \bX_2$, $\bar \bZ_2$ and $\pi_{ij} \bV_i$. As 
        \cref{assumption:ridgeless:universal}
        controls moments up to the order $60 \geq (q+1)m$ for $q \leq 5$, it yields the necessary moment controls for computing up to the fifth derivative.
    \end{itemize}
    One can therefore perform a tedious calculation to verify that each further differentiation brings a multiplicative factor of at most $\max\{1, \lambda^{-1}\} n^{-1/2} $ to the overall upper bound, i.e.~for $1 \leq q \leq 5$, 
    \begin{align*}
        \max\{\theta_{q;m;X}\,,\,\theta_{q;m;Z}\}
        \;=\; 
        O\Big( 
            \mfrac{\max\{1, \lambda^{-2-q}\}}{n^{q/2} k^{1/2}}
        \Big)
        \;.
    \end{align*}
    Plugging the bounds into \eqref{eq:ridgeless:universality:intermediate} implies  
    \begin{align*}
        d_{\tilde \cH}\big( f_\lambda(\bar \bX_1, & \bar \bX_2) \,,\, f_\lambda(\bar \bZ_1 , \bar \bZ_2) \big)
        \\
        \;\leq&\;
        C'' n d k^{5/2} 
        \\
        &
        \times 
        \big( 
            \theta_{1;10;X}^5  
            + 10 \theta_{1;8;X}^3\theta_{2;8;X}
            + 10 \theta_{1;6;X}^2\theta_{3;6;X}
            + 15 \theta_{1;6;X} \theta_{2;6;X}^2
            + 10 \theta_{2;4;X} \theta_{3;4;X}
            \\
        &\quad
            + 5 \theta_{1;4;X}\theta_{4;4;X}   
            + \theta_{5;2;X}
        \\
        &\quad
            +
            \theta_{1;10;Z}^5  
            + 10 \theta_{1;8;Z}^3\theta_{2;8;Z}
            + 10 \theta_{1;6;Z}^2\theta_{3;6;Z}
            + 15 \theta_{1;6;Z} \theta_{2;6;Z}^2
            + 10 \theta_{2;4;Z} \theta_{3;4;Z}
            \\
        &\quad
            + 5 \theta_{1;4;Z}\theta_{4;4;Z}   
            + \theta_{5;2;Z}
        \big)
        \\
        \;=&\;
        O\Big( 
            n d k^{5/2} \times \mfrac{\max\{1, \lambda^{-2-5} \} }{n^{5/2} k^{1/2}}
        \Big)
        \;=\; 
        O\Big( 
            \mfrac{k^2 \max\{1, \lambda^{-7}\}}{n^{1/2}} 
        \Big)\;,
    \end{align*}
    where we have again used $d=O(n)$. This proves the universality statement for $\lambda > 0$ fixed. 

\vspace{.5em}

For the ridgeless case, recall from  \cref{lem:compare:d_H} that $d_P(\argdot, \star) \leq 8^{4/5} d_{\cH}(\argdot,\star)^{1/5}$. By the triangle inequality and Lemma \ref{lem:ridgeless:by:ridge}, we have that for every $\lambda \in (0,1]$,
\begin{align*}
    &\;
    d_P\big( f_0(\bar \bX_1, \bar \bX_2) \,,\,  f_0(\bar \bZ_1, \bar \bZ_2)  \big)
    \\
    &\;\leq\;
    d_P\big( f_0(\bar \bX_1, \bar \bX_2) \,,\, f_\lambda(\bar \bX_1, \bar \bX_2) \big)
    +
    8^{\frac{4}{5}} d_{\cH}\big( f_\lambda(\bar \bX_1, \bar \bX_2) \,,\, f_\lambda(\bar \bZ_1, \bar \bZ_2) \big)^{\frac{1}{5}}
    \\
    & 
    \qquad
    +
    d_P\big( f_\lambda(\bar \bZ_1, \bar \bZ_2 ) \,,\, f_0(\bar \bZ_1, \bar \bZ_2 ) \big)
    \\
    &\;=\;
    O\Big( 
        \lambda + \lambda^2 + \mfrac{1}{n \lambda}  
        +
        \Big(\mfrac{k^2 \max\{1, \lambda^{-7}\} }{n^{1/2}} \Big)^{1/5}
    \Big)\;.
\end{align*}
Since $d=O(n)$ and $1 \leq k^2 = o(n^{1/2})$, setting $\lambda = k^{1/7} n^{-1/28}$ implies that the above bound is $o(1)$, which finishes the proof.

\qed

\subsection{Proof of Proposition 11:
Oracle augmentation via unaugmented risk} \label{appendix:ridgeless:surrogate:proof}

The proof consists of three steps: We first quantify the error of approximating $\bar \bZ_1$ by
\begin{align*}
    \bar \bZ_2 + \mfrac{(k-1)\sigma^2_A}{k} \bI_d 
\end{align*}
in the risk in the case $\lambda >0$. This is followed by a similar approximation for the case $\lambda = 0$. Then we compute the limiting risk by reducing the risk to that of an unaugmented ridge regressor.

\vspace{1em}

\textbf{Step 1: Replace $\bar \bZ_1$ in $f_\lambda(\bar \bZ_1, \bar \bZ_2)$ for $\lambda > 0$.} Recall from \cref{lem:ridgeless:isoG:alt:rep} that 
\begin{align*}
    \bar \bZ_1 \;=\; 
    \bar \bZ_2 
    + 
    \Delta\;,
\end{align*}
where we denote the following rescaled Wishart matrix 
\begin{align*}
    \Delta \;\coloneqq\; \mfrac{\sigma^2_A}{nk} \msum_{i=1}^n
    \msum_{j=2}^k \eta_{ij} \eta_{ij}^\top\;,
\end{align*}
and $\eta_{ij}$'s are i.i.d.~standard Gaussians in $\R^d$. Also note that
\begin{align*}
    \mfrac{(k-1)\sigma^2_A}{k}\, \bI_d \;=\; \mean[\Delta]\;.
\end{align*}
This allows us to control
\begin{align*}
    &\;
    \Big| 
        f^{(1)}_\lambda(\bar \bZ_1)
        -
        f^{(1)}_\lambda\Big(\bar \bZ_2 + \mfrac{(k-1) \sigma^2_A}{k} \bI_d \Big)
    \Big|
    \\
    &\;=\;
    \lambda^2 
    \Big|
        \beta^\top
        \Big( 
            ( \bar \bZ_1 + \lambda \bI_d )^{-2} -
            ( \bar \bZ_2 +  \mean[\Delta] + \lambda \bI_d)^{-2}
        \Big)
        \beta
    \Big|
    \\
    &\;\leq\;
    \lambda^2 \| \beta \|^2 
    \Big\|
            ( \bar \bZ_1 + \lambda \bI_d )^{-2} 
            \, 
            \big( 
                \,
                ( \bar \bZ_2 +  \mean[\Delta] + \lambda \bI_d)^2
                -
                ( \bar \bZ_1 + \lambda \bI_d )^2
                \,
            \big)
            \,
            ( \bar \bZ_2 +  \mean[\Delta] + \lambda \bI_d )^{-2}
    \Big\|_{op} 
    \\
    &\;\leq\;
    \mfrac{\lambda^2 \| \beta \|^2 }{\lambda^2 \big( \frac{k-1}{k} \, \sigma^2_A+ \lambda  \big)^2 }
    \,
    \Big\|
        \bar \bZ_2 + \mfrac{(k-1) \sigma^2_A}{k} \bI_d  + \lambda \bI_d + \bar \bZ_1 +  \lambda \bI_d 
    \Big\|_{op}
    \, 
    \| \mean[\Delta] - \Delta \|_{op}
    \\
    &\;\leq\;
    \mfrac{\| \beta \|^2 }{\big( \frac{k-1}{k} \, \sigma^2_A+ \lambda  \big)^2 }
    \,
    \Big( 
        \| \bar \bZ_2 \|_{op} + \| \bar \bZ_1 \|_{op} + \mfrac{(k-1) \sigma^2_A}{k} + 2 \lambda 
    \Big) 
    \, 
    \| \mean[\Delta] - \Delta \|_{op}
    \;.
\end{align*}
By adapting the proof of \cref{lem:ridgeless:isoG:assumption} and using the maximum singular value bound from Theorem 6.1 of \cite{wainwright2019high}, we see that for any $\epsilon > 0$, with probability $1-\epsilon$ we have
\begin{align*}
        \big\| \bar \bZ_l \big\|_{op}
        \;\leq\; 
        \mfrac{k+\sigma^2_A}{k} \Big( 1 + \sqrt{ \mfrac{2 \log (1/\epsilon)}{n} } + \sqrt{\mfrac{d}{n} \,} \,\Big)
\end{align*}
for both $l=1,2$. Meanwhile, by noting that $\Delta$ is a rescaled sample covariance matrix of $n(k-1)$ i.i.d.~isotropic Gaussians, by Theorem 4.6.1 of \cite{vershynin2018high}, there is some absolute constant $C' > 0$ such that for any $\epsilon > 0$, with probability $1-\epsilon$ we have
\begin{align*}
    \| \Delta - \mean[\Delta] \|_{op}  
    \;\leq\;
    C' \, \mfrac{(k-1)\sigma^2_A}{k}
     \Big(
        \mfrac{\sqrt{d} \, + \sqrt{\log(2/\epsilon)} }{\sqrt{n(k-1)}}  
        +  
        \mfrac{(\sqrt{d} \, +  \sqrt{\log(2/\epsilon)} \; )^2}{n(k-1)}
    \Big) 
    \;.
\end{align*}
Also note that since $k \geq 2$, $\frac{k-1}{k} \in [\frac{1}{2},1]$. This implies that for some absolute constants $C'', C''' > 0$ such that with probability $1 - 3 \epsilon$,
\begin{align*}
    \Big| 
        &f^{(1)}_\lambda(\bar \bZ_1)
        - 
        f^{(1)}_\lambda\Big(\bar \bZ_2 + \mfrac{(k-1) \sigma^2_A}{k} \bI_d \Big)
    \Big|
    \\
    &\;\leq\;
    C''
    \| \beta \|^2 
    \,
    \mfrac{1}{(\sigma^2_A + \lambda)^2} 
    \,
    \Big( 
        \mfrac{k+\sigma^2_A}{k} \Big( 1 + \sqrt{ \mfrac{2 \log (1/\epsilon)}{n} } + \sqrt{\mfrac{d}{n} \,} \,\Big)
        +
        \mfrac{(k-1)\sigma^2_A}{k}
        +
        \lambda
    \Big)
    \, 
    \| \mean[\Delta] - \Delta \|_{op}
    \\
    &\;\leq\;
    \mfrac{C''' \| \beta \|^2}{(\sigma^2_A + \lambda)^2} 
    \,
    \Big( 
        \mfrac{k+\sigma^2_A}{k} \Big( \sqrt{ \mfrac{2 \log (1/\epsilon)}{n} } + \sqrt{\mfrac{d}{n} \,} \,\Big)
        +
        1+\sigma^2_A
        +
        \lambda
    \Big)
    \, 
    \\
    &\qquad 
    \times
    \sigma_A^2
    \Big(
        \mfrac{\sqrt{d} \, + \sqrt{\log(2/\epsilon)} }{\sqrt{n(k-1)}}  
        +  
        \mfrac{(\sqrt{d} \, +  \sqrt{\log(2/\epsilon)} \; )^2}{n(k-1)}
    \Big) 
    \;.
\end{align*}
Notice that by recycling the bound above, we have
\begin{align*}
    &\;
    \Big| 
        f^{(2)}_\lambda(\bar \bZ_1, \bar \bZ_2)
        -
        f^{(2)}_\lambda\Big(\bar \bZ_2 + \mfrac{(k-1) \sigma^2_A}{k} \bI_d, \bar \bZ_2 \Big)
    \Big|
    \\
    &\;=\;
    \mfrac{\sigma_\epsilon^2}{n}
    \Big|
        \Tr\Big( 
            \big( \bar \bZ_1  + \lambda \bI_d \big)^{-2}
            \bar \bZ_2 
            -
            \Big( \bar \bZ_2 + \mfrac{(k-1) \sigma^2_A}{k} \bI_d +  \lambda \bI_d \Big)^{-2}
            \bar \bZ_2 
        \Big) 
    \Big|
    \\
    &\;\leq\;
    \mfrac{\sigma_\epsilon^2 d}{n}
    \,
    \| \bar \bZ_2 \|_{op}
    \,
    \Big\|
            \big( \bar \bZ_1  + \lambda \bI_d \big)^{-2}
            -
            \Big( \bar \bZ_2 + \mfrac{(k-1) \sigma^2_A}{k} \bI_d +  \lambda \bI_d \Big)^{-2}
    \Big\|_{op}
    \\
    &\;\leq\; 
    \mfrac{C''' }{ \lambda^2 (\sigma^2_A + \lambda)^2} 
    \,
    \Big( 
        \mfrac{k+\sigma^2_A}{k} \Big( \sqrt{ \mfrac{2 \log (1/\epsilon)}{n} } + \sqrt{\mfrac{d}{n} \,} \,\Big)
        +
        1+\sigma^2_A
        +
        \lambda
    \Big)^2
    \, 
    \\
    &\qquad 
    \times
    \sigma_A^2
    \Big(
        \mfrac{\sqrt{d} \, + \sqrt{\log(2/\epsilon)} }{\sqrt{n(k-1)}}  
        +  
        \mfrac{(\sqrt{d} \, +  \sqrt{\log(2/\epsilon)} \; )^2}{n(k-1)}
    \Big) 
\end{align*}
for some absolute constant $C''' > 0$ with probability $1 - 3 \epsilon$ for any $\epsilon > 0$. By a union bound, we obtain that there exists some absolute constant $C > 0$ such that for any $\epsilon > 0$,  with probability $1 - 6 \epsilon$, we have 
\begin{align*}
    \Big| 
        f_\lambda(\bar \bZ_1, \bar \bZ_2)
        - &
        f_\lambda\Big(\bar \bZ_2 + \mfrac{(k-1) \sigma^2_A}{k} \bI_d, \bar \bZ_2 \Big)
    \Big|
    \\
    &\;\leq\; 
    C \, \mfrac{1}{(\sigma^2_A + \lambda)^2} 
    \big( \| \beta \|^2 + \lambda^{-2} \big)
    \,
    \Big( 
        \mfrac{k+\sigma^2_A}{k} \Big( \sqrt{ \mfrac{2 \log (1/\epsilon)}{n} } + \sqrt{\mfrac{d}{n} \,} \,\Big)
        +
        1+\sigma^2_A
        +
        \lambda
    \Big)^2
    \, 
    \\
    &\qquad 
    \times
    \sigma^2_A
    \Big(
        \mfrac{\sqrt{d} \, + \sqrt{\log(2/\epsilon)} }{\sqrt{n(k-1)}}  
        +  
        \mfrac{(\sqrt{d} \, +  \sqrt{\log(2/\epsilon)} \; )^2}{n(k-1)}
    \Big) 
    \;.
\end{align*}
In particular this implies that for $\lambda > 0$ fixed, $d=O(n)$, $k \geq 2$ and $\sigma_A^2 \leq 1$, 
\begin{align*}
    \Big| 
        f_\lambda(\bar \bZ_1, \bar \bZ_2)
        - 
        f_\lambda\Big(\bar \bZ_2 + \mfrac{(k-1) \sigma^2_A}{k} \bI_d, \bar \bZ_2 \Big)
    \Big|
    \;=&\; 
    O\Big( \max\Big\{ 1\,,\, \mfrac{d}{n} \Big\}^{3/2} \, \mfrac{\sqrt{d}}{\sqrt{n}} \, \mfrac{\sigma^2_A ( 1 + \lambda^{-2})}{ \sqrt{k}} \Big)
    \\
    \;=&\;
    O\Big(  \mfrac{\sigma^2_A}{ \sqrt{k}} \, \mfrac{\sqrt{d}}{\sqrt{n}} \,\max\Big\{ 1\,,\, \mfrac{d}{n} \Big\}^{3/2} \Big)
\end{align*}
with probability $1 - O(e^{-\min\{d,n\}})$. By the definition of the L\'evy-Prokhorov metric \eqref{eq:defn:prokhorov}, we have
\begin{align*}
    d_P\Big( 
        f_\lambda(\bar \bZ_1, \bar \bZ_2)
        \,,\, 
        f_\lambda\Big(\bar \bZ_2 + \mfrac{(k-1) \sigma^2_A}{k} \bI_d, \bar \bZ_2 \Big)
    \Big)
    \;=\;
    O\Big(  \mfrac{\sigma^2_A}{ \sqrt{k}}  \, \mfrac{\sqrt{d}}{\sqrt{n}} \,\max\Big\{ 1\,,\, \mfrac{d}{n} \Big\}^{3/2}  \Big)\;.
    \tagaligneq \label{eq:ridgeless:surrogate:lamb_pos:approx}
\end{align*}

\vspace{.5em}

\textbf{Step 2: Approximate $f_0(\bar \bZ_1, \bar \bZ_2)$ by $f_\lambda\big(\bar \bZ_2 + \frac{(k-1) \sigma^2_A}{k} \bI_d, \bar \bZ_2 \big)$.} By \cref{lem:ridgeless:isoG:assumption}, we get that the assumptions of \cref{lem:ridgeless:by:ridge} are fulfilled, and in particular in the proof of \cref{lem:ridgeless:isoG:assumption} we have shown that the $1/(n\lambda^2)$ term in fact vanishes. This implies for $\lambda$ small,
\begin{align*}
    d_P\big(  f_\lambda(\bar \bZ_1, \bar \bZ_2) \,,\,  f_0(\bar \bZ_1, \bar \bZ_2) \big)
    \;=\;
    O(\lambda)
    \;.
\end{align*}
Setting $\lambda = \sigma_A^{2/3} k^{-1/6} (d/n)^{1/2}$ and combining this bound with the $d_P$ bound from above, we obtain
\begin{align*}
    d_P\Big(  
        f_0(\bar \bZ_1, \bar \bZ_2) \,,\,  f_{\frac{\sigma_A^{2/3}}{k^{1/6}} \frac{d^{1/2}}{n^{1/2}}}\Big(\bar \bZ_2 + \mfrac{(k-1) \sigma^2_A}{k} \bI_d, \bar \bZ_2 \Big)
    \Big)
    \;=\;
    O\bigg( \mfrac{\sigma_A^{2/3}}{k^{1/6}} \, \mfrac{d^{1/6}}{n^{1/6}} \, \max\Big\{ 1\,,\, \mfrac{d}{n} \Big\}^{1/2} \bigg)
    \;.
    \tagaligneq \label{eq:ridgeless:surrogate:lamb_zero:approx}
\end{align*} 

\vspace{.5em}

\textbf{Step 3: Compute the limiting risk of $f_\lambda\big(\bar \bZ_2 + \frac{(k-1) \sigma^2_A}{k} \bI_d, \bar \bZ_2 \big)$.} Define 
\begin{align*}
    \lambda_k \;\coloneqq\; \mfrac{(k-1) \sigma^2_A}{k} + \lambda\;,
    \qquad
    \sigma^2_k \;\coloneqq\; \mfrac{k+ \sigma_A^2}{k} \;,
    \qquad 
    \tilde \bZ \;\coloneqq\; \mfrac{1}{n} \msum_{i=1}^n \eta_{i1} \eta_{i1}^\top\;,
\end{align*}
where $\eta_{i1}$'s are the i.i.d.~standard Gaussians defined in \cref{lem:ridgeless:isoG:alt:rep}. Recall also that
\begin{align*}
    \bar \bZ_2 
    \;=&\; 
    \mfrac{k+ \sigma_A^2}{k} 
    \, 
    \mfrac{1}{n} \msum_{i=1}^n \eta_{i1} \eta_{i1}^\top
    \;=\;
    \sigma^2_k 
    \, 
    \tilde \bZ
    \;,
\end{align*}
where $\eta_{i1}$'s are i.i.d.~standard Gaussians. 
Observe that 
\begin{align*}
    f_\lambda\Big(\bar \bZ_2 + \mfrac{(k-1) \sigma^2_A}{k} \, \bI_d, \bar \bZ_2 \Big)
    \;=&\;
    \lambda^2 \beta^\top \big( \bar \bZ_2 + \lambda_k \, \bI_d \big)^{-2} \beta 
    +
    \mfrac{\sigma^2_\epsilon}{n}
        \Tr\big( 
            \big(
                \bar \bZ_2 +\lambda_k \, \bI_d 
            \big)^{-2}
            \bar \bZ_2
        \big) 
    \\
    \;=&\;
    \mfrac{\lambda^2}{ \lambda_k^2} \, f^{(1)}_{\lambda_k / \sigma^2_k}(\tilde \bZ)
    +
    \mfrac{1}{\sigma_k^2} \, f^{(2)}_{\lambda_k  / \sigma^2_k}(\tilde \bZ, \tilde \bZ)
    \;.
\end{align*}
Denote the bias and variance parts of the risk defined in \cite{hastie2022surprises} as
\begin{align*}
    R^{(1)}(\beta, \lambda, \gamma)   \;\coloneqq&\;  \| \beta\|^2 \lambda^2 \, \partial m_\gamma(-\lambda) 
    &\text{ and }&&
    R^{(2)}(\sigma, \lambda, \gamma)  \;\coloneqq&\; \sigma^2 \gamma 
    \, \big( m_\gamma(-\lambda) - \lambda \partial m_\gamma(-\lambda)  \big)\;, 
\end{align*}
where we recall 
$  m_\gamma(z) = \frac{1 - \gamma - z - \sqrt{(1-\gamma-z)^2 - 4\gamma z} }{2 \gamma z} $. Now suppose $k$ is fixed and $\lambda > 0$. By Corollary 5 of \cite{hastie2022surprises}, we get that almost surely as $d,n \rightarrow \infty$ with $d/n \rightarrow \gamma$,
\begin{align*}
    f_\lambda\Big(\bar \bZ_2 + \mfrac{(k-1) \sigma^2_A}{k} \, \bI_d, \bar \bZ_2 \Big)
    \;\xrightarrow{a.s.}&\;
    \mfrac{\lambda^2}{ \lambda_k^2} \, R^{(1)}\Big(\beta, \mfrac{\lambda_k}{\sigma^2_k}, \gamma \Big)
    +
    \mfrac{1}{ \lambda_k^2} \, R^{(2)}\Big(\sigma_\epsilon, \mfrac{\lambda_k}{\sigma^2_k}, \gamma \Big)
    \\
    \;=&\;
    R^{(1)}\Big( \mfrac{\lambda}{ \lambda_k} \beta, \mfrac{\lambda_k}{\sigma^2_k}, \gamma \Big)
    +
     R^{(2)}\Big( \mfrac{\sigma_\epsilon}{\sigma_k}, \mfrac{\lambda_k}{\sigma^2_k}, \gamma \Big)
     \\
     \;=&\;
     R\Big( \mfrac{\lambda}{ \lambda_k} \beta,   \mfrac{\sigma_\epsilon}{\sigma_k}, \mfrac{\lambda_k}{\sigma^2_k}, \gamma   \Big)
     \tagaligneq \label{eq:ridgeless:surrogate:hastie:approx}
\end{align*}
for every $k \geq 2 $ and $\sigma^2_A \leq 1$. Note that
\cref{lem:ridgeless:isoG:assumption} shows that
 Assumptions \ref{assumption:ridgeless:universal} and \ref{assumption:ridgeless:by:ridge} 
 both hold under the isotropic setup, so the universality bounds in 
 \cref{prop:ridgeless:universality} 
 hold. In the case $\lambda > 0$, combining the above first with \eqref{eq:ridgeless:surrogate:lamb_pos:approx} under the assumption that $\frac{\sigma^2_A}{ \sqrt{k}} \frac{\sqrt{d}}{\sqrt{n}} = o(1)$ and then with 
  \cref{prop:ridgeless:universality},
   we have 
\begin{align*}
    f_\lambda(\bar \bX_1, \bar \bX_2 )
    \;\xrightarrow{\P}&\;
    \lim 
    \,
    R\Big( \mfrac{\lambda}{ \lambda_k} \beta,   \mfrac{\sigma_\epsilon}{\sigma_k}, \mfrac{\lambda_k}{\sigma^2_k}, \gamma   \Big)
    \;,
\end{align*}
where $\lim$ denotes the limit under 
\eqref{eq:ridgeless:asymptotic:regime} 
with $\frac{\sigma^2_A}{ \sqrt{k}} \frac{\sqrt{d}}{\sqrt{n}} = o(1)$. For the ridgeless case $\lambda = 0$, the same argument applies: 
\cref{prop:ridgeless:universality} 
shows that $f_0(\bar \bX_1, \bar \bX_2)$ and $f_0(\bar \bZ_1, \bar \bZ_2)$ have the same distributional limit under
\eqref{eq:ridgeless:asymptotic:regime}, 
whereas \eqref{eq:ridgeless:surrogate:lamb_zero:approx} shows that $f_0(\bar \bZ_1, \bar \bZ_2)$ and $ f_{\frac{\sigma_A^{2/3}}{k^{1/6}} \frac{d^{1/2}}{n^{1/2}}}\Big(\bar \bZ_2 + \frac{(k-1) \sigma^2_A}{k} \bI_d, \bar \bZ_2 \Big)$ have the same distributional limit under $\frac{\sigma^2_A}{ \sqrt{k}} \frac{\sqrt{d}}{\sqrt{n}} = o(1)$. The distributional limit of $ f_{\frac{\sigma_A^{2/3}}{k^{1/6}} \frac{d^{1/2}}{n^{1/2}}}\Big(\bar \bZ_2 + \frac{(k-1) \sigma^2_A}{k} \bI_d, \bar \bZ_2 \Big)$ under 
\eqref{eq:ridgeless:asymptotic:regime} 
is given by \eqref{eq:ridgeless:surrogate:hastie:approx}, and we note that 
\begin{align*}
     R( \bzero,   \sigma_\epsilon, \sigma^2_A, \gamma)
     \;=\;
     \lim_{\lambda \rightarrow 0^+}
     \,
     R\Big( \mfrac{\lambda}{\lambda + \sigma_A^2} \beta,   \sigma_\epsilon, \lambda + \sigma^2_A, \gamma   \Big)
\end{align*}
exists by continuity as shown in \cite{hastie2022surprises}.

\qed 

\subsection{Proof for Proposition 12:
Two-stage augmentation} \label{appendix:proof:two:stage:aug}

The proof expresses the difference $R(\hat \beta_0^{(m)}) 
- 
\hat L^{(\rm ora)}_0 
-
\big\| \bar \bX_1^\dagger \bar \bX_\Delta (\tilde \beta_0^{(m)} - \beta) \|^2
- 
\sigma^2_\epsilon$ as two quantities involving averages and uses a concentration argument to show that they both converge to zero in probability.

\vspace{.5em}

We first recall from \cref{appendix:ridgeless:results} that
\begin{align*}
    \hat L_0^{\rm (ora)}
    \;=\;
    \beta^\top \big( \bar \bX_1^\dagger \bar \bX_1 - \bI_d \big)^2 \beta 
    +
    \mfrac{\sigma_\epsilon^2}{n} \, \Tr\big( 
        \bar \bX_1^\dagger
         \bar \bX_2 
        \bar \bX_1^\dagger
        \big) 
    \;.
\end{align*}
Meanwhile, recall that  we have defined 
\begin{align*}
    \bar \bX_\Delta \;=\; \mfrac{1}{n} \msum_{i=1}^n \Big( \mfrac{1}{k} \msum_{j=1}^k (\bV_i + \xi_{ij}) \Big) \Big( \mfrac{1}{k} \msum_{j=1}^k \xi_{ij} \Big)^\top 
    \;,
\end{align*}
and denote $\bar \bx_\epsilon = \frac{1}{n} \sum_{i=1}^n \big( \frac{1}{k} \sum_{j=1}^k (\bV_{i} + \xi_{ij}) \big) \epsilon_i$. Then we can express
    \begin{align*}
        \hat{\beta}^{(m)}_0
        \;=&\
        \bar \bX_{1}^\dagger 
        \Big( \mfrac{1}{n} \msum_{i=1}^n \msum_{j=1}^k (\bV_i + \xi_{ij})(\bV_i^\top \beta + \epsilon_i + \xi_{ij}^\top \tilde \beta^{(m)}_0)  \Big)
        \\
        \;=&\;
        \bar \bX_{1}^\dagger \bar \bX_1 \beta 
        +
        \bar \bX_{1}^\dagger \bar \bX_\Delta (\tilde \beta^{(m)}_0 - \beta )
        +
        \bar \bX_{1}^\dagger 
        \bar \bx_\epsilon
        \;,
    \end{align*}
and therefore the risk of interest can be expressed as 
\begin{align*}
    R(\hat \beta^{(m)}_{0})
    \;=&\; 
    \sigma_\epsilon^2  + \| \hat \beta^{(m)}_{0} - \beta \|^2 
    \\
    \;=&\;
    \sigma_\epsilon^2 
    +
    \Big\| 
        \big( \bar \bX_{1}^\dagger \bar \bX_1 - \bI_d \big) 
        \beta 
        +
        \bar \bX_{1}^\dagger \bar \bX_\Delta (\tilde \beta^{(m)}_0 - \beta )
        +
        \bar \bX_{1}^\dagger 
        \bar \bx_\epsilon
    \Big\|^2
    \\
    \;\overset{(a)}{=}&\;
    \sigma_\epsilon^2 
    +
    \beta^\top \big( \bar \bX_{1}^\dagger \bar \bX_1 - \bI_d \big)^2 \beta 
    +
    \big\| \bar \bX_{1}^{-1} \bar \bX_\Delta (\tilde \beta^{(m)}_0 - \beta ) \big\|^2
    \\
    &\;
    +
    2 (\tilde \beta^{(m)}_0 - \beta )^\top \bar \bX_\Delta \bar \bX_{1}^{-2} \bar \bx_\epsilon 
    + 
    \bar \bx_\epsilon^\top \bar \bX_{1}^{-2} \bar \bx_\epsilon
    \\
    \;=&\; 
    \sigma_\epsilon^2 
    +
    \hat L_0^{\rm (ora)} 
    + 
    \big\| \bar \bX_{1}^{-1} \bar \bX_\Delta (\tilde \beta^{(m)}_0 - \beta ) \big\|^2
    \\
    &\; 
    - 
    2 \underbrace{(\tilde \beta^{(m)}_0 - \beta )^\top \bar \bX_\Delta \bar \bX_{1}^{-2} \bar \bx_\epsilon}_{\eqqcolon Q_1}
    -
    \underbrace{\Big( \bar \bx_\epsilon^\top \bar \bX_{1}^{-2} \bar \bx_\epsilon - \mfrac{\sigma^2_\epsilon}{n} \Tr\big( 
        \bar \bX_1^\dagger
         \bar \bX_2 
        \bar \bX_1^\dagger
        \big)  \Big)}_{\eqqcolon Q_2}
    \;.
\end{align*}
In $(a)$, we have noted that $(\bar \bX_1 \bar \bX_1^\dagger - \bI_d) \bar \bX_1^\dagger  = \bzero$ by the property of pseudo-inverse, which allows some cross-terms to vanish. 

\vspace{.5em}

We now prove that $Q_1$ and $Q_2$ converge in probability to zero. By assumption,
$\| \bar \bX_1^\dagger \|_{op} + \| \bar \bX_2 \|_{op} + \| \bar \bX_\Delta \|_{op} +  \| \tilde \beta^{(m)}_0 - \beta \| \leq C$ for some constant $C < \infty$ with probability $1-o(1)$. Define the event
\begin{align*}
    E \;\coloneqq\; \big\{\| \bar \bX_1^\dagger \|_{op} +  \| \bar \bX_2 \|_{op} + \| \bar \bX_\Delta \|_{op} +  \| \tilde \beta^{(m)}_0 - \beta \|  \leq  C\big\}\;.
\end{align*}
By the expression of $\bar \bx_\epsilon$, we can write 
\begin{align*}
    Q_1
     \;=\; (\tilde \beta^{(m)}_0 - \beta )^\top \bar \bX_\Delta \bar \bX_{1}^{-2}  \bar \bx_\epsilon
    \;=\;
    \mfrac{1}{n} \msum_{i=1}^n  (\tilde \beta^{(m)}_0 - \beta )^\top \bar \bX_\Delta \bar \bX_{1}^{-2} \Big( \mfrac{1}{k} \msum_{j \leq k} (\bV_i + \xi_{ij}) \Big) \epsilon_i\;.
\end{align*}
Conditioning on $\tilde \cX = (V_i, \xi_{ij})_{i \leq n, j \leq k}$, we get that almost surely 
\begin{align*}
    \mean[\,Q_1 \,|\, \tilde \cX\,]
    \;=&\; 
    0\;,
    \\
    \Var[\,Q_1 \,|\, \tilde \cX\,]
    \;=&\; 
    \mfrac{\sigma^2_\epsilon}{n^2}
    \sum_{i=1}^n 
    \bigg(
    \Big( \mfrac{1}{k} \sum_{j \leq k} (\bV_i + \xi_{ij}) \Big)^\top
    \bar \bX_{1}^{-2}  \bar \bX_\Delta 
    (\tilde \beta^{(m)}_0 - \beta )
    \\
    &\hspace{5em}
    (\tilde \beta^{(m)}_0 - \beta )^\top \bar \bX_\Delta \bar \bX_{1}^{-2} \Big( \mfrac{1}{k} \sum_{j \leq k} (\bV_i + \xi_{ij}) \Big) 
    \bigg)
    \\
    \;=&\;
    \mfrac{\sigma^2_\epsilon}{n}
    (\tilde \beta^{(m)}_0 - \beta )^\top \bar \bX_\Delta \bar \bX_1^{-2} \bar \bX_2 \bar \bX_1^{-2} \bar \bX_\Delta 
    (\tilde \beta^{(m)}_0 - \beta )
    \\
    \;\leq&\;
    \mfrac{\sigma^2_\epsilon}{n} \| \bar \bX_2 \|_{op} \| \bar \bX_1^\dagger \|_{op}^4 \| \bar \bX_\Delta \|_{op}^2 \| \tilde \beta^{(m)}_0 - \beta \|^2
    \;,
\end{align*}
which is $O(n^{-1})$ on the event $E$. Therefore by splitting the probability according to $E$ and applying the Markov's inequality, we obtain that for any $t > 0$,
\begin{align*}
    \P( |Q_1| > t )
    \;\leq&\;
    \P( |Q_1| > t , E )
    + 
    \P( E^c)
    \\
    \;=&\;
    \mean\big[ \, \P( |Q_1| > t \,|\, \tilde \cX \,) \, \ind_{E} \big]
    +
    o(1)
    \\
    \;\leq&\;
    t^{-2}
    \,
    \mean\big[ \, \Var[ Q_1 \,|\, \tilde \cX ] \, \ind_{E} \big]
    +
    o(1)
    \;=\;
    o(1)\;,
\end{align*}
i.e.~$Q_1$ converges to zero in probability. $Q_2$ can be handled by a similar argument: First note that $\mean[Q_2] = 0$ since 
\begin{align*}
    \mean\big[ 
        \bar \bx_\epsilon^\top \bar \bX_{1}^{-2} \bar \bx_\epsilon 
        \,\big|\, \tilde \cX
    \big] 
    \;=&\;
    \mfrac{1}{n^2} \msum_{i=1}^n \mean\Big[ 
        \epsilon_i  \Big( \mfrac{1}{k} \sum_{j \leq k} (\bV_i + \xi_{ij}) \Big)^\top  \bar \bX_{1}^{-2} 
        \Big( \mfrac{1}{k} \sum_{j \leq k} (\bV_i + \xi_{ij}) \Big)
        \epsilon_i    
    \,\Big| \, \tilde \cX \Big]
    \\
    \;=&\;
    \mfrac{\sigma^2_\epsilon}{n} \Tr\big( 
            \bar \bX_1^\dagger
             \bar \bX_2 
            \bar \bX_1^\dagger
            \big)
    \;.
\end{align*}
While the expression of $\Var\big[ 
    Q_2
    \,\big|\, \tilde \cX
\big] $ involves a complicated expansion of four sums, we note that since $\epsilon_i$ is zero-mean and independent, the only non-vanishing terms are of the form $\epsilon_i^2 \epsilon_{i'}^2$ with $i \neq i'$, with a multiplicity of $O(n^2)$, and $\epsilon_i^4$, with a multiplicity of $O(n)$. Therefore, conditioning on the event $E$, we have that 
\begin{align*}
    \Var\big[ 
        Q_2
        \,\big|\, \tilde \cX
    \big] 
    \;=\; O(n^{-2}) \;=\; o(1)\;,
\end{align*}
and applying the same argument of splitting the probability according to $E$ followed by Markov's inequality gives that $Q_2$ converges to zero in probability. In summary, we have proved the desired statement that 
\begin{align*}
    R(\hat \beta^{(m)}_{0})
    -
    \big(
    \sigma_\epsilon^2 
    +
    \hat L_0^{\rm (ora)} 
    + 
    \big\| \bar \bX_{1}^{-1} \bar \bX_\Delta (\tilde \beta^{(m)}_0 - \beta ) \big\|^2
    \big)
    \;\xrightarrow{\P}\; 
    0\;.
\end{align*}

\qed

%% file: appendix_8_additions.tex
\section{Proofs for Section \ref{sec:network} and Appendix \ref{appendix:network:extend}}\label{appendix:ridgeless:network:proof} 

This appendix collects the proofs for the results for models beyond ridgeless regression and isotropic noise injection:
\begin{itemize}
    \item \Cref{appendix:proof:network:risk} proves \Cref{lem:network:risk} in \Cref{appendix:network:extend}, which computes the risk for the nonlinear feature model;
    \item \Cref{appendix:proof:network:general} proves \Cref{prop:network:extend:general}, the universality result for the nonlinear feature model in \Cref{appendix:network:extend};
    \item \Cref{appendix:proof:ridgeless:extend} proves Proposition 13, the universality result for the linear network model in \Cref{sec:network};
    \item   \Cref{appendix:proof:noniso:alt:rep} proves \Cref{lem:ridgeless:noniso:alt:rep}, which provides the alternative expression of $\bar \bZ^*_1$ for the nonisotropic case.

\end{itemize}

\subsection{Proof of \Cref{lem:network:risk}: Risk computation under \Cref{assumption:network:generalize}.}  \label{appendix:proof:network:risk}
First by taking the expectation over $\bV_{\rm new}$ and $\epsilon_{\rm new}$, we have that for $\lambda \geq 0$,
\begin{align*}
    \hat L_{\lambda}(\cX) 
    \;=&\;
    \mean\big[ 
        \big( 
            \hat \beta_{\lambda}^\top \varphi_\theta( \bV_{\rm new} ) 
            -
            Y_{\rm new}
        \big)^2  
    \,\big| \, \cX, \bW^{(0)} \big]
    \\
    \;=&\;
    \mean\big[ 
        \big( 
            \hat \beta_{\lambda}^\top \varphi_\theta( \bV_{\rm new} ) 
            -
            \beta^\top \bW^{(0)} \varphi_{\theta_0}(\bV_{\rm new})
            -
            \epsilon_{\rm new}
        \big)^2  
    \,\big| \, \cX, \bW^{(0)} \big]
    \\
    \;\overset{(a)}{=}&\;
    \mean\big[ 
        \big( 
            \hat \beta_{\lambda}^\top \varphi_\theta( \bV_{\rm new} ) 
            -
            \beta^\top \bW^{(0)} \varphi_{\theta_0}(\bV_{\rm new})
        \big)^2  
    \,\big| \, \cX, \bW^{(0)} \big]
    +
    \sigma_\epsilon^2
    \\
    \;\overset{(b)}{=}&\;
    \mean\big[ 
    \hat \beta_{\lambda}^\top M^{\varphi_\theta} \hat \beta_{\lambda}
    -
    2 
    \hat \beta_{\lambda}^\top R^{\varphi_\theta, \varphi_{\theta_0}} \bW^{(0)} \beta
    +
    \beta^\top \bW^{(0)} M^{\varphi_{\theta_0}} (\bW^{(0)} )^\top \beta
    \,\big| \, \cX, \bW^{(0)} \big]
    +
    \sigma_\epsilon^2
    \;.
\end{align*}
In $(a)$ we have used that $\epsilon_{\rm new}$ is mean-zero with variance $\sigma_\epsilon^2$; in $(b)$ we have recalled the definition that
\begin{align*}
    &\;
     M^{\varphi_\theta} \;=\; \mean \big[ \, \varphi_\theta(\bV_{\rm new})  \, \varphi_\theta(\bV_{\rm new})^\top \,  \big]
     \;,
     \quad
     R^{\varphi_\theta, \varphi_{\theta_0}} \;=\; \mean \big[ \, \varphi_\theta(\bV_{\rm new})  \,  \varphi_{\theta_0}(\bV_{\rm new})^\top \,  \big]\;,
     \\
     &\;
     M^{\varphi_{\theta_0}} \;=\; \mean \big[ \, \varphi_{\theta_0}(\bV_{\rm new})  \,  \varphi_{\theta_0}(\bV_{\rm new})^\top \,  \big]
     \;.
\end{align*}
Now note that under \Cref{assumption:network:generalize}(ii), we can write the estimator as
\begin{align*}
    \hat \beta_\lambda 
    \;=&\; 
    \bar \bX^{*;-1}_{1;\lambda}
    \,
    \mfrac{1}{nk} \msum_{i=1}^n \msum_{j=1}^k
    \tilde \bV_{ij} \, \tilde Y_{ij}
    \\
    \;=&\;
    \bar \bX^{*;-1}_{1;\lambda} 
    \,
    \Big( \bar \bX^*_3 (\bW^{(0)})^\top \beta  
        +    
        \mfrac{1}{nk} \msum_{i=1}^n \msum_{j=1}^k
        \tilde \bV_{ij} \, \epsilon_i
    \Big)
    \,,
\end{align*}
where we have used the definitions
\begin{align*}
    \bar \bX^*_1 
    \;\coloneqq&\; 
    \mfrac{1}{nk} \msum_{i=1}^n \msum_{j=1}^k  \tilde \bV_{ij} (\tilde \bV_{ij})^\top
    \;,
    \qquad
    \bar \bX^*_3
    \;\coloneqq\; 
    \mfrac{1}{nk} \msum_{i=1}^n \msum_{j=1}^k \tilde \bV_{ij} \tilde \bV_0^\top
    \;,
    \\
    \text{and }\quad &\;
    \bar \bX^{*;-1}_{1;\lambda}
    \;\coloneqq\;
    \begin{cases}
        (    \bar \bX^*_1 + \lambda \bI_{p'} )^{-1}
        & \text{ for } \lambda > 0 \;,
        \\
        (    \bar \bX^*_1 )^\dagger
        & \text{ for } \lambda = 0 \;.
    \end{cases}
\end{align*}
By taking an expectation over $\epsilon_i$ and noting that $\epsilon_i$'s are i.i.d.~zero-mean with variance $\sigma^2_\epsilon$, we get that 
\begin{align*}
    \hat L_\lambda (\cX)
    \;=&\;
    \beta^\top 
    \bW^{(0)}
    (\bar \bX^*_3)^\top 
    \bar \bX^{*;-1}_{1;\lambda} 
    \,M^{\varphi_\theta} \,
    \bar \bX^{*;-1}_{1;\lambda} 
    (\bar \bX^*_3  ) (\bW^{(0)})^\top 
    \beta  
    \\
    &\;
    +    
    \mfrac{\sigma^2_\epsilon}{n} \,
    \Tr\big(\,
    \bar \bX^{*;-1}_{1;\lambda} 
    \,M^{\varphi_\theta} \,
    \bar \bX^{*;-1}_{1;\lambda} 
    \,
    \bar \bX^*_2
    \,\big)
    \\
    &\;
    -
    2
    \beta^\top 
    \bW^{(0)}
    (\bar \bX^*_3)^\top 
    \bar \bX^{*;-1}_{1;\lambda} 
    \, R^{\varphi_\theta, \varphi_{\theta_0}} \,
    (\bW^{(0)})^\top \beta 
    \\
    &\;
    +
    \beta^\top \bW^{(0)} M^{\varphi_{\theta_0}} (\bW^{(0)})^\top \beta
    +
    \sigma_\epsilon^2
    \;,
\end{align*}
where we have recalled the definition 
\begin{align*}
    \bar \bX^*_2 
    \;\coloneqq\;
    \mfrac{1}{n} \msum_{i=1}^n 
    \Big( \mfrac{1}{k} \msum_{j=1}^k  \tilde \bV_{ij} \Big)
    \,
    \Big( \mfrac{1}{k} \msum_{j=1}^k  \tilde \bV_{ij} \Big)^\top 
    \;.
\end{align*}
\qed

\vspace{1em}

\subsection{Proof of \Cref{prop:network:extend:general}: Nonlinear feature model in \Cref{appendix:network:extend}. } \label{appendix:proof:network:general}

We first set up the notation. Let $\Theta \sim \textrm{Uniform}[0,1]$ be independent of all other variables. Also denote 
\begin{align*}
    \tilde \bV_i \coloneqq (\tilde \bV_{ij}, \tilde \bV^0_{ij})_{j \leq k} ,
     \;\;
    \tilde \bZ_i \coloneqq (\tilde \bZ_{ij}, \tilde \bZ_{ij}^0)_{j \leq k},
    \;\;
    \cW_i(\bx)
    \coloneqq
    (\tilde \bV_1, \ldots, \tilde \bV_{i-1}, \bx, \tilde \bZ_{i+1}, \ldots, \tilde \bZ_n)\;.
\end{align*} 
For $\bx = (\bx_j, \bx^0_j)_{j \leq k} \in \R^{2kd}$, we write the sample covariance matrices corresponding to $\cW_i(\bx)$ as 
\begin{align*}
    \bar \bW_{i;1}(\bx) 
    \coloneqq&\;
    \mfrac{1}{nk} 
    \msum_{i' \leq i-1} \msum_{1 \leq j \leq k} \tilde \bV_{i'j} \tilde \bV_{i'j}^\top
    +
    \mfrac{1}{k} \msum_{1 \leq j \leq k} \mfrac{\bx_j}{\sqrt{n}} \mfrac{\bx_j}{\sqrt{n}}^\top
    \\
    &\;
    +
    \mfrac{1}{nk} 
    \msum_{i' \geq i+1} \msum_{1 \leq j \leq k} \tilde \bZ_{i'j} \tilde \bZ_{i'j}^\top
    \;,
    \\
    \bar \bW_{i;2}(\bx)  
    \coloneqq&\; 
    \mfrac{1}{n} 
    \msum_{i' \leq i-1}
    \Big( 
        \mfrac{1}{k} \msum_{1 \leq j \leq k} \tilde \bV_{i'j} \Big) 
    \Big( 
        \mfrac{1}{k} \msum_{1 \leq j \leq k} \tilde \bV_{i'j}
    \Big)^\top
    \\
    &\;
    +
    \Big( \mfrac{1}{k} \msum_{1 \leq j \leq k} \mfrac{\bx_j}{\sqrt{n}} \Big)  
    \Big( \mfrac{1}{k} \msum_{1 \leq j \leq k} \mfrac{\bx_j}{\sqrt{n}} \Big)^\top
    \\
    &\;
    +
    \mfrac{1}{n} 
    \msum_{i' \geq i+1} 
    \Big( \mfrac{1}{k} \msum_{1 \leq j \leq k} \tilde \bZ_{i'j} \Big) 
    \Big(  \mfrac{1}{k} \msum_{1 \leq j \leq k} \tilde \bZ_{i'j}
    \Big)^\top
    \;,
    \\
     \bar \bW_{i;3}(\bx) 
    \coloneqq&\;
    \mfrac{1}{nk} 
    \msum_{i' \leq i-1} \msum_{1 \leq j \leq k} \tilde \bV_{i'j} (\tilde \bV_{i'j}^0)^\top
    +
    \mfrac{1}{k} \msum_{1 \leq j \leq k} \mfrac{\bx_j}{\sqrt{n}} \mfrac{\bx_j^0}{\sqrt{n}}^\top
    \\
    &\;
    +
    \mfrac{1}{nk} 
    \msum_{i' \geq i+1} \msum_{1 \leq j \leq k} \tilde \bZ_{i'j} (\tilde \bZ_{i'j}^0)^\top
    \;.
\end{align*}
We also use the shorthands
\begin{align*}
    \bar \bW_{i;1;\lambda} 
    \;\coloneqq&\; 
    \begin{cases}
        (    \bar \bW_{i;1}( \bzero ) + \lambda \bI_p )^{-1}
        & \text{for } \lambda > 0 \;,
        \\
        (    \bar \bW_{i;1}( \bzero ) )^\dagger
        & \text{for } \lambda = 0 \;,
    \end{cases}
    &\;\text{ and }\;&&
    M_\bx \;\coloneqq&\; \big( \mfrac{\bx_1}{\sqrt{n}}, \ldots, \mfrac{\bx_k}{\sqrt{n}} \big) \in \R^{d \times k}\;.
\end{align*} 
Then by the Woodbury matrix identity,
\begin{align*}
    ( \bar \bW_{i;1}( \bx ) + \lambda \bI_d )^{-1}
    \;=&\; 
    \Big( \bar \bW_{i;1;\lambda} + \mfrac{1}{k} \msum_{1 \leq j \leq k} \mfrac{\bx_j}{\sqrt{n}} \mfrac{\bx_j}{\sqrt{n}}^\top  \Big)^{-1}
    \\
    \;=&\;
    \Big( \bar \bW_{i;1;\lambda} + \mfrac{1}{k} M_\bx M_\bx^\top  \Big)^{-1}
    \\
    \;=&\;
    \bar \bW_{i;1;\lambda}^{-1} 
    -
    \mfrac{1}{k}
    \bar \bW_{i;1;\lambda}^{-1} 
    M_\bx 
    \big( I_k + \mfrac{1}{k} M_\bx^\top \bar \bW_{i;1;\lambda}^{-1} M_\bx  \big)^{-1} 
    M_\bx^\top 
    \bar \bW_{i;1;\lambda}^{-1} 
    \;.
    \tagaligneq \label{eq:ridgeless:extend:woodbury}
\end{align*}

\vspace{.5em}

\textbf{Step 1: Lindeberg over $n$ independent blocks of augmented data. } Recall that we can express 
\begin{align*}
    \hat L_\lambda(\cX) 
    \;=&\; 
    \mfrac{1}{n}
    f_\lambda( \cW_n( \tilde \bV_n ))
    +
    L_0(\bW^{(0)})
    \;,
    \\
    \hat L_\lambda(\cZ) 
    \;=&\; 
    \mfrac{1}{n}
    f_\lambda( \cW_1( \tilde \bZ_1 ))
    +
    L_0(\bW^{(0)})
    \;,
\end{align*}
where 
\begin{align*}
    f_\lambda( \cW_i( \bx  ))
    \;=&\;
    (\sqrt{n} \beta)^\top 
    \bW^{(0)}
    (\bar \bW_{i;3}(\bx))^\top 
    ( \bar \bW_{i;1}( \bx ) + \lambda \bI_d )^{-1}
    \,M^{\varphi_\theta} \,
    \\
    &\hspace{5em} ( \bar \bW_{i;1}( \bx ) + \lambda \bI_d )^{-1}
    (\bar \bW_{i;3}(\bx)) (\bW^{(0)})^\top 
    (\sqrt{n} \beta)
    \\
    &\;
    +    
    \sigma^2_\epsilon \,
    \Tr\big(\,
    ( \bar \bW_{i;1}( \bx ) + \lambda \bI_d )^{-1}
    \,M^{\varphi_\theta} \,
    ( \bar \bW_{i;1}( \bx ) + \lambda \bI_d )^{-1}
    \,
    \bar \bW_{i;2}(\bx)
    \,\big)
    \tagaligneq \label{eq:ridgeless:extend:CS:comment} 
    \\
    &\;
    -
    2
    (\sqrt{n} \beta)^\top 
    \bW^{(0)}
    (\bar \bW_{i;3}(\bx))^\top 
    ( \bar \bW_{i;1}( \bx ) + \lambda \bI_d )^{-1}
    \, R^{\varphi_\theta, \varphi_{\theta_0}} \,
    \bW^{(0)} (\sqrt{n} \beta)
    \;,
\end{align*}
and 
\begin{align*}
    L_0(\bW^{(0)})
    \;\coloneqq\; 
    \beta^\top \bW^{(0)} M^{\varphi_{\theta_0}} (\bW^{(0)}) \beta 
    +
    \sigma^2_\epsilon
    \;.
\end{align*}
Fix $\tilde h \in \cH^{(4)}$, a four-times continuously differentiable function with its first four derivatives uniformly bounded from above by $1$. The first step is to make use of the version of Theorem 1 discussed in \Cref{remark:non:iid:no:taylor} applied to $\tilde \bV_1, \ldots, \tilde \bV_n$ and $\tilde \bZ_1, \ldots, \tilde \bZ_n$ to obtain 
\begin{align*}
    \big| \mean \tilde h ( \hat L_\lambda(\cX) ) - \mean \tilde h ( \hat L_\lambda(\cZ) ) \big| 
    \;\leq\;
    \mfrac{1}{ n }
    \msum_{i=1}^n 
    \big| 
    \;
    \mean\big[  F_i(\tilde \bV_i)  -  F_i(\tilde \bZ_i) \big] 
    \;
    \big|
    \;,
\end{align*}
where, for $\bx = (\bx_j)_{j \leq k} \in \R^{kd}$, we have defined
\begin{align*}
    F_i(\bx)
    \;\coloneqq\;
    \partial \tilde h \Big( 
        \mfrac{1}{n}
        f_\lambda( \cW_i( \Theta \bx ) ) 
        +
        L_0(\bW^{(0)})
    \Big) 
  \, 
   \partial_i f_\lambda( \cW_i( \Theta \bx ) )^\top \bx  
    \;.
\end{align*}
To proceed, we observe that by combining the calculation \eqref{eq:ridgeless:extend:woodbury}, the derivative calculation of $f_\lambda(\cW_i(\bx))$, and the fact that $\tilde h$ is four-times differentiable, $F_i(\bx)$ can be expressed as a three-times continuously differentiable function $\tilde f_{\cW_i(\bzero), L_0(\bW^{(0)})}: \R^{N_k} \rightarrow \R$, which depends on $\cW_i(\bzero)$, of the $N_k \coloneqq  2 k (k+1) $ variables:
    \begin{align*}
        A(\bx_j^0)
        \;\coloneqq&\;
        \beta^\top \bW^{(0)} \bx_j^0
        \quad 
        \text{ for } 1 \leq j \leq k 
        \;,
        \\
        B^{(i)}(\bx_j, \bx_{j'})
        \;\coloneqq&\;
        \Big(\mfrac{\bx_j}{\sqrt{n}} \Big)^\top \bar \bW^{-1}_{i;1;\lambda} 
        \Big(\mfrac{\bx_{j'}}{\sqrt{n}} \Big)
        \quad 
        \text{ for } 1 \leq j, j' \leq k 
        \;,
        \\
        C^{(i)}(\bx_j, \bx_{j'})
        \;\coloneqq&\;
        \Big(\mfrac{\bx_j}{\sqrt{n}} \Big)^\top \bar \bW^{-1}_{i;1;\lambda} 
        \, M^{\varphi_\theta} \,
        \bar \bW^{-1}_{i;1;\lambda} 
        \Big(\mfrac{\bx_{j'}}{\sqrt{n}} \Big)
        \quad 
        \text{ for } 1 \leq j, j' \leq k 
        \;,
        \\
        D^{(i)}(\bx_j)
        \;\coloneqq&\;
        \beta^\top \bW^{(0)}
        (R^{\varphi_\theta, \varphi_{\theta_0}})^\top
        \bar \bW^{-1}_{i;1;\lambda} 
        \bx_j
        \quad 
        \text{ for } 1 \leq j \leq k 
        \;.
    \end{align*}
    Moreover, $\tilde f_{\cW_i(\bzero), L_0(\bW^{(0)})}$ itself and its derivatives are all locally Lipschitz functions, with bounded local Lipschitz constants since $k$ is fixed and $\tilde h$ has four uniformly bounded derivatives. 
     Denote the collection of the $N_k$ variables as 
\begin{align*}
    Q_i(\bx) 
    \;=&\; 
    Q_i(\bx_1, \bx^0_1, \ldots, \bx_k, \bx^0_k)
    \\
    \;=&\; 
    \big(
        \, 
        (A(\bx_j^0))_{j \leq k}
        \,,\, 
        (B^{(i)}(\bx_j, \bx_{j'}) )_{ j, j' \leq k}
        \,,\,
         (C^{(i)}(\bx_j, \bx_{j'}) )_{ j, j' \leq k}
         \,,\,
         (D^{(i)}(\bx_j))_{ j \leq k}
    \big)
    \;.
\end{align*}
This implies that for some constant $L_k > 0$ that only depends on $k$,
\begin{align*}
     &\;
     \big| \mean \tilde h ( \hat L_\lambda(\cX) ) - \mean \tilde h (  \hat L_\lambda(\cZ) ) \big|  
    \\
    &\;\leq\;
    L_k
    \,
    \mmax_{i \leq n}  
    \, 
    \big| 
        \mean[
        \tilde f_{\cW_i(\bzero), L_0(\bW^{(0)})}( Q_i(\tilde \bV_i)  )
        -
        \tilde f_{\cW_i(\bzero), L_0(\bW^{(0)})}( Q_i(\tilde \bZ_i)  )
        ]
    \big|
    \;.
\end{align*}
We remark on how the rest of the proof differs from that of \Cref{prop:ridgeless:universality}. Notice that to control the derivative terms, using the Cauchy-Schwarz inequality naively can yield undesirable dimension-dependence. 
For example,  one of the terms in $\partial_i f_\lambda(\cW_i(\Theta \bx))^\top \bx$ obtained from differentiating the line \eqref{eq:ridgeless:extend:CS:comment} reads 
\begin{align*}
     \sigma^2_\epsilon
     \,
        \Tr \Big( 
            ( \bar \bW_{i;1}( \Theta \bx ) + \lambda \bI_d )^{-1}
            M^{\varphi_\theta}
            ( \bar \bW_{i;1}( \Theta \bx ) + \lambda \bI_d )^{-1}
            \mfrac{\bx_j^0}{\sqrt{n}}  
            \mfrac{(\bx_j^0)}{\sqrt{n}}^\top 
        \Big)
    \;.
\end{align*}
If we are to apply the Cauchy-Schwarz inequality directly, we obtain
\begin{align*}
     \sigma_\epsilon^2
     \mfrac{(\bx_j^0)}{\sqrt{n}}^\top 
     ( \bar \bW_{i;1}( \Theta \bx ) + \lambda \bI_d )^{-1}
            M^{\varphi_\theta}
    ( \bar \bW_{i;1}( \Theta \bx ) + \lambda \bI_d )^{-1}
    \mfrac{\bx_j^0}{\sqrt{n}}  
     \;\leq\;
     \mfrac{2 \sigma_\epsilon^2}{\lambda^2}
     \Big\| \mfrac{\bx_j^0}{\sqrt{n}} \Big\|^2
     \;,
\end{align*}
which is $\Theta(1)$ with high probability. In the proof of \Cref{prop:ridgeless:universality} in \Cref{appendix:proof:ridgeless:universality}, we address this by exploiting four-moment-matching and i.i.d.~coordinate condition of \Cref{assumption:covariate}. In the remainder of this proof, we instead exploit the weak dependence across the coordinates and the sub-Gaussianity condition in \Cref{assumption:network:generalize}.

\vspace{.5em}

\textbf{Step 2: Exploit orthogonal invariance of $\bW^{(0)}$. } 
We now exploit the fact that $\bW^{(0)}$ has i.i.d.~Gaussian entries and is therefore invariant under orthogonal transformations. In particular, let $\bO$ be a uniform draw from the group of $\R^{d \times d}$ orthogonal matrices $\cO(d)$ and independent of all other variables. Then
\begin{align*}
    \bW^{(0)} \;\overset{d}{=}\;\bO  \bW^{(0)}
\end{align*}
and we can replace all occurrences of $\bW^{(0)}$ above by $\bO \bW^{(0)}$. Therefore from now on, with an abuse of notation, we rewrite 
\begin{align*}
    A(\bx_j^0) \;=&\; \beta^\top \bO \bW^{(0)} \bx_j^0
    \quad 
    \text{ for } 1 \leq j \leq k 
    \;,
    \\
    D^{(i)}(\bx_j)
    \;\coloneqq&\;
    \beta^\top \bO \bW^{(0)}
    (R^{\varphi_\theta, \varphi_{\theta_0}})^\top
    \bar \bW^{-1}_{i;1;\lambda} 
    \bx_j
    \quad 
    \text{ for } 1 \leq j \leq k 
    \;.
\end{align*}
Let $\eta \in \cN(0,I_d)$ be independent of all other variables, and write $\tilde \eta \coloneqq \eta / \| \eta \|$, which is uniformly drawn from the unit sphere in $\R^d$. In subsequent calculations, we will be exploiting the property of $\bO$ that for any fixed vector $v \in \R^d$,
\begin{align*}
    \beta^\top
    \bO v 
    \;=&\;
    \beta^\top 
    \, 
    \mfrac{\bO v}{\| v \|} 
    \, 
    \| v\|
    \;\overset{d}{=}\;
    \beta^\top 
    \, 
    \tilde \eta \, \| v \| 
    \;,
\end{align*}
and therefore for a fixed $r \geq 2$, we can compute the $L_r$ norm of $\beta^\top \bO v $ as 
\begin{align*}
    \| \beta^\top \bO v  \|_{L_r}
    \;=\;
    \big\| \beta^\top  \tilde \eta  \big\|_{L_r} \, \| v \|
    \;\overset{(a)}{=}\;
    \mfrac{\| \beta^\top  \eta  \|_{L_r}}{\big\| \| \eta \| \big\|_{L_r} } \, \| v \|
    \;=\;
    O\Big( \mfrac{\| \beta \| \, \| v \|}{d^{1/2}} \Big)
    \;.
    \tagaligneq \label{eq:ortho:contract}
\end{align*}
In $(a)$, we have used that $\tilde \eta$ and $\| \eta \|$ are independent.

\vspace{.5em}

\textbf{Step 3: Approximate $\tilde f \equiv \tilde f_{\cW_i(\bzero), L_0(\bO \bW^{(0)})}$ by a bounded Lipschitz function. } For convenience, we write $\tilde f \equiv \tilde f_{\cW_i(\bzero), L_0(\bO \bW^{(0)})}$ from now on, while noting in particular that $\cW_i(\bzero), L_0(\bO \bW^{(0)})$ are both independent of $\tilde \bV_i$ and $\tilde \bZ_i$. Fix some constant $K > 0$, and define a bounded approximation
\begin{align*}
    \tilde f_K(\bx) \;\coloneqq&\; \tilde f(\bx) \, \ind_{\{ \| \bx \| \leq K \}} \,+\, \tilde f( K \bx / \|\bx\| ) \, \ind_{\{ \| \bx \| > K \}}
    \;.
\end{align*}
Then by the triangle inequality a, we obtain
\begin{align*}
    &\;
    \big| 
       \mean[  \tilde f( Q_i( \tilde \bV_i)  )
        -
        \tilde f( Q_i(\tilde \bZ_i)  )
       ]
    \big|
    \\
    &\qquad\;\leq\;
          \big| 
        \mean[
        \tilde f_K( Q_i(\tilde \bV_i)  )
        -
        \tilde f_K( Q_i(\tilde \bZ_i)  )
        ]
    \big|
    \tagaligneq \label{eq:locally:lipschitz:moment}
    \\
    &\qquad\quad 
    +
    \big| 
        \mean[
        \tilde f( Q_i(\tilde \bV_i)  )
        -
        \tilde f_K( Q_i(\tilde \bV_i)  )
        ]
    \big|
    +
    \big| 
        \mean[
        \tilde f( Q_i(\tilde \bZ_i)  )
        -
        \tilde f_K( Q_i(\tilde \bZ_i)  )
        ]
    \big|
    \;,
    \tagaligneq \label{eq:locally:lipschitz:residual}
\end{align*}
To control \eqref{eq:locally:lipschitz:residual}, we notice that
\begin{align*}
    \tilde f( Q_i(\bx)  )
        -
        \tilde f_K( Q_i(\bx)  ) 
        \neq 0
    \;\Rightarrow\; 
    \| Q_i(\bx) \| > K \;,
\end{align*}
which allows us to bound
\begin{align*}
    \eqref{eq:locally:lipschitz:residual}
    \;\leq&\;
    \big| 
        \mean[
        (\tilde f( Q_i(\tilde \bV_i)  )
        -
        \tilde f_K( Q_i(\tilde \bV_i)  )
        )
        \ind_{\{ \| Q_i(\tilde \bV_i) \| > K  \} }
        ]
    \big|
    \\
    &\;
    +
    \big| 
        \mean[
        (\tilde f( Q_i(\tilde \bZ_i)  )
        -
        \tilde f_K( Q_i(\tilde \bZ_i)  )
        )
        \ind_{\{ \| Q_i(\tilde \bZ_i) \| > K  \} }
        ]
    \big|
    \\
    \;\leq&\;
    \| \tilde f( Q_i(\tilde \bV_i)  ) - \tilde f_K( Q_i(\tilde \bV_i)  ) \|_{L_2}
    \,
    \P( \| Q_i(\tilde \bV_i) \| > K   )^{1/2}
    \\
    &\;
    +
    \|  \tilde f( Q_i(\tilde \bZ_i)  ) - \tilde f_K( Q_i(\tilde \bZ_i)  ) \|_{L_2}
    \,
    \P( \| Q_i(\tilde \bZ_i) \| > K  )^{1/2}
    \;.
\end{align*}
The $L_2$-norms can be verified to be $O(1)$, so it suffices to control the probabilities as $K$ grows. As the argument for $Q_i(\tilde \bZ_i)$ is analogous to that for $Q_i(\tilde \bV_i)$, we present only the one for $Q_i(\tilde \bV_i)$. We shall consider the different components of $Q_i(\tilde \bV_i)$, followed by a union bound. Notice that since $\tilde \bV_{ij}^0$ is mean-zero and sub-Gaussian, by the independence of $\bV_{ij}^0$ from $\bW^{(0)}$, we have that for all $j \leq k$ and any $t > 0$,
\begin{align*}
    \P\big( \, \big| A(\tilde \bV_{ij}^0) \big| > t  \big)
    \;=&\;
    \P\big( \, \big| \beta^\top \bW^{(0)} \, \tilde \bV_{ij}^0 \big| > t \, \big)
    \\
    \;\leq&\;
    2
    \mean\Big[ 
        \exp\Big( - \mfrac{t^2  }{\lambda^2 \| (\bW^{(0)})^\top \beta \|^2  \sigma_V^2 }  \Big)
    \Big]
    \\
    \;\leq&\;
    2
    \mean\Big[ 
        \exp\Big( - \mfrac{t^2  }{\lambda^2 \| \bW^{(0)} \|_{op}  \| \beta \|^2  \sigma_V^2 }  \Big)
    \Big]
    \;,
\end{align*}
where we have denoted the operator norm $\| M \|_{op} \coloneqq \sup_{x \in \cS^{p'-1}} \| M x \|_2$ for an $\R^{p \times p'}$ matrix. Since $\bW^{{(0)}}$ is entrywise i.i.d.~$\cN(0,1/p')$ and $p'/n \rightarrow \gamma_1 \in [0,\infty)$ and $p/n \rightarrow \gamma_2 \in [0,\infty)$ in \eqref{eq:network:asymptotic:regime:generalise}, by standard bounds on the norm of matrix with i.i.d.~Gaussian entries (see e.g.~Theorem 4.4.5 of \cite{vershynin2018high}), there is some absolute constant $C > 0$ such that for all $t > 0$,
\begin{align*}
    \P\Big( \| \bW^{(0)} \|_{op} \,\leq\, C \Big( \mfrac{\sqrt{p}}{\sqrt{p'}} + 1 + \mfrac{t}{\sqrt{p'}} \Big) \Big)
    \;\geq\; 
    1 - 2 \exp( - t^2 )
    \;.
    \tagaligneq \label{eq:gaussian:op:bound}
\end{align*}
Therefore
\begin{align*}
    \P\big( \, \big| A(\tilde \bV_{ij}^0) \big| > t  \big)
    \;\leq&\;
    2
    \exp\Big( - \mfrac{t^2  }{\lambda^2 C \big( \sqrt{p/p'} + 1 + t / \sqrt{p'}  \big)  \| \beta \|^2  \sigma_V^2 }  \Big)
    +
    4 \exp( - t^2 )\;.
\end{align*}
i.e.~the tailed probability decays exponentially in $t$. A similar argument shows that $D^{(i)}(\tilde \bV_{ij})$'s also have exponential tails, by exploiting the sub-Gaussian-ness of $\tilde \bV_{ij}$ and the bound on the operator norm of $\bW^{(0)}$. The only additional argument is to note that 
\begin{align*}
    \| R^{\varphi_\theta, \varphi_{\theta_0}} \|_{op} 
    \;=\; 
    \| \mean[ \varphi( \bV_{\rm new}) \varphi_0(\bV_{\rm new})^\top ]  \|_{op}
    \;=\;
    O(1)
    \tagaligneq \label{eq:R:op:control}
\end{align*}
by \Cref{assumption:network:bounded:cov}. To control the tail of $B^{(i)}(\tilde \bV_{ij}, \tilde \bV_{ij})$, we use that $\tilde \bV_{ij}$ are sub-Gaussian and mean-zero again and apply the generalized Hanson-Wright inequality by \cite{hsu2012tail}: For every $t > 0$ we have 
\begin{align*}
    \P\Big(  
        \,
        | B^{(i)}(\tilde \bV_{ij}, \tilde \bV_{ij}) | 
        > 
        \mfrac{\sigma_V^2}{n} 
        \Big( 
            \Tr( \bar \bW^{-1}_{i;1;\lambda} )  
            + 
            2 \sqrt{\Tr( \bar \bW^{-2}_{i;1;\lambda} ) \, t \, }
            +
            2 \, \| \bar \bW^{-1}_{i;1;\lambda} \|_{op} \, t
        \Big)  
        \, 
    \Big)
    \;\leq\; 
    e^{-t}
    \;,
\end{align*}
which implies 
\begin{align*}
     \P\Big(  
        \,
        | B^{(i)}(\tilde \bV_{ij}, \tilde \bV_{ij}) | 
        > 
        \mfrac{2 \sigma_V^2}{n \lambda} 
        ( \sqrt{d} + \sqrt{t}  )^2
    \Big)
    \;\leq\; 
    e^{-t}
\end{align*}
and therefore 
\begin{align*}
    \P\big(  
        \,
        | B^{(i)}(\tilde \bV_{ij}, \tilde \bV_{ij}) | 
        > 
        t
    \big)
    \;\leq\; 
    \exp\Big( 
        -
            \max\Big\{ 
                \mfrac{ \lambda^{1/2}}{\sqrt{2} \sigma_V}
                \sqrt{t n} 
                -
                \sqrt{d}
                \,,\,
                0
            \Big\}^2        
    \Big)
    \;.
\end{align*}
For $j \neq j'$, by noting that 
\begin{align*}
    | B^{(i)}(\tilde \bV_{ij}, \tilde \bV_{ij'}) | \;\leq\; \max\{ | B^{(i)}(\tilde \bV_{ij}, \tilde \bV_{ij}) |  \,,\, | B^{(i)}(\tilde \bV_{ij'}, \tilde  \bV_{ij'}) |  \} 
\end{align*}
and using a union bound, we obtain 
\begin{align*}
    \P\big(  
        \,
        | B^{(i)}(\tilde \bV_{ij}, \tilde \bV_{ij'}) | 
        > 
        t
    \big)
    \;\leq\; 
    2
    \exp\Big( 
        -
        \max\Big\{ 
                \mfrac{ \lambda^{1/2}}{\sqrt{2} \sigma_V}
                \sqrt{tn} 
                -
                \sqrt{d}
                \,,\,
                0
            \Big\}^2           
    \Big)
    \;.
\end{align*}
A similar bound holds again for $C^{(i)}(\tilde \bV_{ij}, \tilde \bV_{ij'})$ except that we additionally use 
$M^{\varphi_\theta} = \mean \big[ \, \varphi(\bV_{\rm new})  \, \varphi(\bV_{\rm new})^\top \,  \big]$ is bounded.  Combining the bounds and noting that $d/n = O(1)$, we obtain that as $K \rightarrow \infty$, the approximation error of $\tilde f$ by $\tilde f_K$ decays exponentially:
\begin{align*}
    \eqref{eq:locally:lipschitz:residual} \;=\; O( e^{ - \Omega( K) } )\;.
\end{align*}
We are left with handling \eqref{eq:locally:lipschitz:moment}, which measures the difference between $\tilde \bV_i$ and $\tilde \bZ_i$ through a bounded Lipschitz function $\tilde f_K$.

\vspace{.5em}

\textbf{Step 4: Continuous Lindeberg over the weakly dependent coordinates.} We employ the continuous interpolation version of Lindeberg's technique. Let $\| \tilde f_K \|_{\rm Lip}$ denote the Lipschitz constant of $\tilde f_K$. First let $\epsilon > 0$ and consider a smooth approximation of $\tilde f_K$ as 
\begin{align*}
    \tilde f_K^\epsilon( \bx ) 
    \;\coloneqq&\; 
    \mfrac{1}{(2\epsilon)^{3N_k}}
    \,
    \mint_{\bx \pm \epsilon} 
    \,
    \mint_{\bu \pm \epsilon}
    \,
    \mint_{\bt \pm \epsilon}
    \,
    \tilde f_K(\by) 
    \; d \by \, d\bt \, d \bu
    \;,
\end{align*}
where $\bx, \bu, \bt, \by \in \R^{N_k}$ and we have used $\bx \pm \epsilon$ as a shorthand for the hyperrectangle $[x_1 - \epsilon, x_1 + \epsilon] \times \ldots \times [x_{N_k} - \epsilon, x_{N_k} + \epsilon]$. Note that $\tilde f_K^\epsilon$ is thrice differentiable and, as $\tilde f_K$ is Lipschitz, we have 
$$\sup_\bx |\tilde f_K^\epsilon(\bx)-\tilde f_K(\bx)|\leq 3 \epsilon \| \tilde f_K \|_{\rm Lip}  \sqrt{N_k }  \;.$$ 
Now for $i \leq n$, $j \leq k$ and $0 \leq r \leq 1$, we consider the continuous interpolation 
\begin{align*}
    \bX_{ijr}(t)
    \;\coloneqq\;
    \sqrt{t} \, \bV_{ijr}
    \,+\, 
    \sqrt{1-t} \, \bZ_{ijr}
    \;,
\end{align*}
and write
\begin{align*}
    \tilde \bX_{ij}(t)
    =
    \varphi(\bW_j^\top \bX_{ij1}(t))
    \;,
    \quad 
    \tilde \bX_{ij}^0(t)
    =
    \varphi(\bW_j^\top \bX_{ij0}(t))
    \;,
    \quad
    \tilde \bX_i(t)
    =
    \big( 
        \tilde \bX_{ij}(t) 
        \,,\,
        \tilde \bX_{ij}^0(t) 
    \big)_{j \leq k}
    \;.
\end{align*}
Use $\partial_{A_j}$, $\partial_{B_{jj'}}$, $\partial_{C_{jj'}}$ and $\partial_{D_j}$ as the shorthands for the partial derivatives with respect to $A(\tilde \bX_{ij}^0(t) )$, $B^{(i)}(\tilde \bX_{ij}(t), \tilde \bX_{ij'}(t) )$,  $C^{(i)}(\tilde \bX_{ij}(t), \tilde \bX_{ij'}(t) )$ and $D^{(i)}(\tilde \bX_{ij}(t))$ respectively. Then by the fundamental theorem of calculus, we have 
\begin{align*}
    &\;
    \Big| \,
        \mean\big[
        \tilde f^\epsilon_K( Q_i(\tilde \bV_i)  )
        -
        \tilde f^\epsilon_K( Q_i(\tilde \bZ_i)  )
        \big]
    \, \Big|
    \\
    &\;\leq
    \mint_0^1
    \, 
    \big| 
        \mean\big[
        \partial_t
        \tilde f^\epsilon_K( Q_i(\tilde \bX_i(t))  )
        \big]
    \big| 
    \,
    dt
    \\
    &\;\leq
    \mint_0^1
    \;
    \Big| 
        \mean\Big[
        \sum_{\substack{j \leq k \\ l \leq d}}
        \partial_{A_j}
            \tilde f^\epsilon_K \big( Q_i(\tilde \bX_i(t))  \big)
        \,
        \partial A(\tilde \bX_{ij}^0(t))
    \,
        \partial \varphi_0( \bX_{ij0}(t) )
        \,
        \be_l 
            \Big(
                \mfrac{( \bV_{ij0})_l}{2 \sqrt{t}}
                -
                \mfrac{( \bZ_{ij0})_l}{2 \sqrt{1-t}}
            \Big)
        \Big]
    \Big| 
    \,
    dt
    \\
    &\hspace{1em}
    +
    2\,
    \mint_0^1
    \;
    \Big| 
        \mean\Big[
        \sum_{\substack{1 \leq j,j'' \leq k\\ l \leq d}}
        \partial_{B_{jj'}}
            \tilde f^\epsilon_K \big( Q_i(\tilde \bX_i(t))  \big)
        \,
        \partial_1 B^{(i)}(\tilde \bX_{ij}(t), \tilde \bX_{ij'}(t))
    \\
    &\hspace{10em}
        \partial \varphi( \bX_{ij1}(t) )
        \,
        \be_l 
            \Big(
                \mfrac{( \bV_{ij1})_l}{2 \sqrt{t}}
                -
                \mfrac{( \bZ_{ij1})_l}{2 \sqrt{1-t}}
            \Big)
        \Big]
    \Big| 
    \,
    dt
    \\
    &\hspace{1em}
    +
    2\,
    \mint_0^1
    \;
    \Big| 
        \mean\Big[
        \sum_{\substack{1 \leq j,j'' \leq k\\ l \leq d}}
        \partial_{C_{jj'}}
            \tilde f^\epsilon_K \big( Q_i(\tilde \bX_i(t))  \big)
        \,
        \partial_1 C^{(i)}(\tilde \bX_{ij}(t), \tilde \bX_{ij'}(t))
    \\
    &\hspace{10em}
        \partial \varphi( \bX_{ij1}(t) )
        \,
        \be_l 
            \Big(
                \mfrac{( \bV_{ij1})_l}{2 \sqrt{t}}
                -
                \mfrac{( \bZ_{ij1})_l}{2 \sqrt{1-t}}
            \Big)
        \Big]
    \Big| 
    \,
    dt
    \\
    &\hspace{1em}
    +
    \mint_0^1
    \;
    \Big| 
        \mean\Big[
        \sum_{\substack{ j \leq k\\ l \leq d}}
        \partial_{D_j}
            \tilde f^\epsilon_K \big( Q_i(\tilde \bX_i(t))  \big)
        \,
        \partial A(\tilde \bX_{ij}^0(t))
    \,
        \partial \varphi( \bX_{ij0}(t) )
        \,
        \be_l 
            \Big(
                \mfrac{( \bV_{ij0})_l}{2 \sqrt{t}}
                -
                \mfrac{( \bZ_{ij0})_l}{2 \sqrt{1-t}}
            \Big)
        \Big]
    \Big| 
    \,
    dt
    \\
    &\;\eqqcolon\; (\star)
    \;.
\end{align*}
To control the integrals $(\star)$, note that for a fixed $i \leq N$, $\cB_{jrl}$ is the dependency neighborhood of the $l$-th coordinate of $\bV_{ijr}$ in the collection of variables $( (\bV_{ijr})_l )_{j \leq k, 0 \leq r \leq 2, l \leq d}$. Consider the modifications of the variables that leave out $\cB_{jrl}$: For $ j' \leq k$, $0 \leq r' \leq 1$ and $ l' \leq d$,
\begin{align*}
    &\;
    \big(  \bX^{\cB^c_{jrl}}_{ij'r'}(t) \big)_{l'}
    \;\coloneqq\;
    \begin{cases}
        0  & (j',r',l') \in \cB_{jrl}
        \\ 
        (\bX_{ij'r'})_{l'} &  (j',r',l') \not \in \cB_{jrl}
    \end{cases}
    \;,
    \\
    &\;
    \tilde \bX^{\cB^c_{jrl}}_{ij'r'}(t)
    \;=\; 
    \varphi_0(\bW_j^\top \bX^{\cB^c_{jrl}}_{ij'r'}(t)  )
    \;,
    \qquad 
    \tilde \bX^{\cB^c_{jrl}}_{i}(t)
    \;=\; 
    \big( \tilde \bX^{\cB^c_{jrl}}_{ij'0}(t) \,,\, \tilde \bX^{\cB^c_{jrl}}_{ij'1}(t)  \big)_{j' \leq k}
    \;.
\end{align*}
As with the Lindeberg method proof for \Cref{thm:main}, we shall perform a second-order Taylor expansion on the first derivative terms above with respect to $(\, (\bX_{ij'r'}(t))_{l'} \, )_{(j',r', l') \in \cB_{jrl}}$. We then exploit the facts that
\begin{itemize}
    \item $\tilde \bX^{\cB^c_{jrl}}_{i}(t)$ is independent of $( \bV_{ij'r'})_{l'}$ and  $( \bZ_{ij'r'})_{l'}$ for $(j',r',l') \in \cB_{jrl}$,
    \item  $\mean[( \bV_{ij'r'})_{l'}]=0=\mean[( \bZ_{ij'r'})_{l'}]$, which allows us to drop terms linear in $( \bV_{ij'r'})_{l'}$ and $( \bZ_{ij'r'})_{l'}$, 
    and 
    \item $\Var[( \bV_{ij'r'})_{l'}]=\Var[( \bZ_{ij'r'})_{l'}]$, which implies that for any generic function $F: \R^d \rightarrow \R^{d \times d}$ and $j' \leq k$, $r' \in \{0,1\}$,
    \begin{align*}
        \mean\Big[
        \Big( \sqrt{t} \, ( \bV_{ijr})_{l} + \sqrt{1-t} \, ( \bZ_{ijr})_{l}  \Big)^\top 
         F( \bX^{\cB^c_{jrl}}_{ij'r'}(t)) 
           \Big(
                \mfrac{( \bV_{ijr})_l}{2 \sqrt{t}}
                -
                \mfrac{( \bZ_{ijr})_l}{2 \sqrt{1-t}}
            \Big)
        \Big] 
        =
        0\;.
    \end{align*}
\end{itemize}
This allows to keep only the third-order derivative terms. To represent them, we again write $\Theta \sim \textrm{Uniform}[0,1]$ and denote 
 \begin{align*}
    \big(  \bX^{\Theta \cB^c_{jrl}}_{ij'r'}(t) \big)_{l'}
    \;\coloneqq\;
    \begin{cases}
        \Theta  (\bX_{ij'r'})_{l'}   & (j',r',l') \in \cB_{jrl}
        \\ 
        (\bX_{ij'r'})_{l'} &  (j',r',l') \not \in \cB_{jrl}
    \end{cases}
    \;,
    \qquad
    \tilde \bX^{\Theta\cB^c_{jrl}}_{ij'r'}(t)
    \;=\; 
    \varphi(\bW_j^\top \bX^{\Theta\cB^c_{jrl}}_{ij'r'}(t)  )
    \;.
\end{align*}
An explicit enumeration of all the terms in $(\star)$ by product rule is possible but tedious. The key observations to control the derivatives are the following facts:
\begin{itemize}
    \item Since $\tilde f: \R^{N_k} \rightarrow \R$ is a thrice continuously differentiable function, by construction, $\tilde f^\epsilon_K$ is a thrice-differentiable function with each of its $l$-th derivative bounded as $O_K(\epsilon^{-3N_k})$, where the leading constant depends on $K$. Therefore, bounding the derivatives of $\tilde f^\epsilon_K$ introduce terms of the form $\frac{C_K}{\epsilon^{3 N_K}}$, where $(C_K)_{K \in \N}$ is a sequence of constants, independent of $n$, $d$, $p$ and $p'$, such that $C_K \rightarrow \infty$ as $K \rightarrow \infty$;
    \item Since $\bV_{ij0}$, $\bV_{ij1}$, $\bZ_{ij0}$ and $\bZ_{ij1}$ are all uniformly sub-Gaussian, the coordinates $(\bV_{ijr})_l$ and $(\bZ_{ijr})_l$ all have bounded $L_r$ norms for any fixed $r < \infty$;
    \item $A(\bx_j^0)$ is linear in $\bx_j^0 \in \R^d$, so its second and third derivatives vanish. Meanwhile for any fixed $r \geq 2$ and $v \in \R^d$, by \eqref{eq:ortho:contract}, 
    \begin{align*}
        \big\| \partial A(\bx_j^0) v \big\|_{L_r}
        \;=\; 
        \big\| \beta^\top \bW^{(0)} v \big\|_{L_r}
        \;=\;
        O\Big( \mfrac{\| v \|}{d^{1/2}} \Big)
        \;,
    \end{align*}
    whereas we have used $\| \beta \| = O(1)$ and that 
    $\|  \| \bW^{(0)} \|_{op} \|_{L_r} = O(1)$ as a consequence of \eqref{eq:gaussian:op:bound} (see e.g.~the discussion after Theorem 4.4.5 of \cite{vershynin2018high});
    \item $B^{(i)}(\bx_j, \bx_{j'})$ is quadratic in $(\bx_j, \bx_j')$ and its derivatives satisfy that almost surely 
    \begin{align*}
        \| \partial_{\bx_j} B^{(i)}(\bx_j, \bx_{j'})  \| 
        \;=&\;
        \mfrac{1}{n}
        \;
        \| \bar \bW^{-1}_{i;1;\lambda} \, \bx_{j'}  \| 
        \;\leq\;
        \mfrac{d^{1/2}}{n \lambda} \,  \max_{l \leq d}  | (\bx_{j'})_l |
        \;,
        \\
        \| \partial_{\bx_j} \partial_{\bx_{j'}} B^{(i)}(\bx_j, \bx_{j'})  \|_{op}
        \;=&\;
        \mfrac{1}{n} \| \bar \bW^{-1}_{i;1;\lambda} \|_{op} 
        \;\leq\;
        \mfrac{1}{n \lambda} 
        \;;
    \end{align*}
    \item $C^{(i)}(\bx_j, \bx_{j'})$ is quadratic in $(\bx_j, \bx_j')$ and its derivatives satisfy that almost surely 
    \begin{align*}
        \| \partial_{\bx_j} C^{(i)}(\bx_j, \bx_{j'})  \| 
        \;=&\;
        \mfrac{1}{n}
        \;
        \| \bar \bW^{-1}_{i;1;\lambda} M^{\varphi_\theta}  \bar \bW^{-1}_{i;1;\lambda} \, \bx_{j'}^\top  \| 
        \;\leq\;
        \mfrac{d^{1/2}  \, \| M^{\varphi_\theta} \|_{op}}{n \lambda^2} 
          \max_{l \leq d}  | (\bx_{j'})_l |
        \;,
        \\
        \| \partial_{\bx_j} \partial_{\bx_{j'}} C^{(i)}(\bx_j, \bx_{j'})  \|_{op}
        \;\leq&\;
        \mfrac{\| M^{\varphi_\theta} \|_{op}}{n \lambda^2}
        \;,
    \end{align*}
    where we also recall that $\| M^{\varphi_\theta} \|_{op} = O(1)$;
    \item $D^{(i)}(\bx_j)$ is linear in $\bx_j$ and that, by additionally recalling $\| R^{\varphi_\theta, \varphi_{\theta_0}} \|_{op} = O(1)$, we have that for any fixed $r \geq 2$,
    \begin{align*}
        \big\| \, \| \partial D(\bx_j) \| \, \big\|_{L_r}
        \;=\; 
        \big\| \, \big\| \beta^\top \bW^{(0)} (R^{\varphi_\theta, \varphi_{\theta_0}})^\top \bar \bW^{-1}_{i;1;\lambda}  \big\| 
        \, \big\|_{L_r}
        \;=\;
        O(\lambda^{-1} )
        \;;
    \end{align*}
    \item $\gamma_r^\varphi$ provides a uniform bound on the operator norms of the $r$-th derivatives of both $\varphi_0$ and $\varphi$.
\end{itemize}
Recall that $d=O(n)$. In summary, in terms of $n$-dependence, each $r$-th derivative of $A$, $B$, $C$ and $D$ introduces a term that is at most $O(n^{-r/2})$,
whereas in terms of $\lambda$-dependence, we have an overall contribution of at most $O(1 +  \lambda^{-6})$. This implies that for some sequence $C_K \rightarrow \infty$ as $K \rightarrow \infty$, we have 
\begin{align*}
    \Big| \,
        \mean\big[
        \tilde f^\epsilon_K( Q_i(\tilde \bV_i)  )
        -
        \tilde f^\epsilon_K( Q_i(\tilde \bZ_i)  )
        \big]
    \, \Big|
    \;=\; 
    O\Big( 
        \mfrac{C_K}{\epsilon^{3N_K}}
        k^2
         B_d
        (
            1
            +
            \lambda^{-6}
        )
        \Big( 
            \mfrac{(\gamma_1^\varphi)^3}{d^{1/2}}
            +
            \gamma_1^\varphi  \gamma_2^\varphi 
            +
            \gamma_3^\varphi d^{1/2}
        \Big)
    \Big)
    \;.
\end{align*}
Note that $k$ is fixed, and that by \Cref{assumption:network:feature:gradient},
\begin{align*}
    \gamma_1^\varphi \;=\; o( B_d^{-\frac{1}{3}} d^{\frac{1}{6}} ) 
    \,,\, 
    \quad 
    \gamma_2^\varphi \;=\; o( B_d^{-\frac{2}{3}} d^{ - \frac{1}{6}} ) 
    \,,\,
    \quad 
    \gamma_3^\varphi \;=\; o( B_d^{-1}  d^{ - \frac{1}{2}  } ) \;.
\end{align*}
This implies that $B \big( 
            \frac{(\gamma_1^\varphi)^3}{d^{1/2}}
            +
            \gamma_1^\varphi  \gamma_2^\varphi 
            +
            \gamma_3^\varphi d^{1/2}
        \big) = o(1)
$
and therefore 
\begin{align*}
    &\;
    \Big| \,
        \mean\big[
        \tilde f^\epsilon_K( Q_i(\tilde \bV_i)  )
        -
        \tilde f^\epsilon_K( Q_i(\tilde \bZ_i)  )
        \big]
    \, \Big|
    \;=\; 
    o\Big( 
        \mfrac{C_K}{\epsilon^{3N_K}}
        \Big( 
            1
            +
            \mfrac{1}{\lambda^{6}} 
        \Big)
    \Big)
    \;.
\end{align*}

\vspace{.5em}

\textbf{Step 4: Tidying up for the $\lambda >0$ case. } Finally by the triangle inequality and combining the calculations from all four steps, we have
\begin{align*}
    \big| \mean \tilde h ( \hat L_\lambda(\cX) ) - \mean \tilde h (  \hat L_\lambda(\cZ)   ) \big| 
    \;=\;
    O\Big( 
        \,
        e^{-\Omega(K)} 
        +
        3  \epsilon \| \tilde f_K \|_{\rm Lip}  
    \Big)
    +
    o 
    \Big(
        \mfrac{C_K}{\epsilon^{3 N_K}}
        \Big( 
            1
            +
            \mfrac{1}{\lambda^6} 
        \Big)
    \Big)
    \;.
\end{align*}
Recall that we have fixed $\tilde h \in \cH^{(4)}$, a four-times continuously differentiable function with its first four derivatives uniformly bounded from above by $1$, and observe that the bounds above can be stated independently of $\tilde h$. By taking $K \rightarrow \infty$ and $\epsilon \rightarrow 0$ sufficiently slowly,  we obtain
\begin{align*}
    d_{\cH^{(4)}}\big( 
        \hat L_\lambda(\cX)
        \,,\,
        \hat L_\lambda(\cZ) 
    \big)
    \;=\;
    o 
    \Big( 
        \Big( 
            \mfrac{1}{\lambda} 
            +
            \mfrac{1}{\lambda^6} 
        \Big)
    \Big)
    \;.
    \tagaligneq \label{eq:network:ridge:universality}
\end{align*}

\vspace{.5em}

\textbf{Step 5: Take $\lambda \rightarrow 0^+$. } We seek to take $\lambda \rightarrow 0^+$ in  
\begin{align*}
    \hat L_\lambda (\cX)
    \;=&\;
    \beta^\top \bW^{(0)} M^{\varphi_{\theta_0}} (\bW^{(0)})^\top \beta
    +
    \sigma_\epsilon^2
    \\
    &\;
    +
    \beta^\top 
    \bW^{(0)}
    (\bar \bX^*_3)^\top 
    \bar \bX^{*;-1}_{1;\lambda} 
    \,M^{\varphi_\theta} \,
    \bar \bX^{*;-1}_{1;\lambda} 
    (\bar \bX^*_3  ) (\bW^{(0)})^\top 
    \beta  
    \\
    &\;
    +    
    \mfrac{\sigma^2_\epsilon}{n} \,
    \Tr\big(\,
    \bar \bX^{*;-1}_{1;\lambda} 
    \,M^{\varphi_\theta} \,
    \bar \bX^{*;-1}_{1;\lambda} 
    \,
    \bar \bX^*_2
    \,\big)
    \\
    &\;
    -
    2
    \beta^\top 
    \bW^{(0)}
    (\bar \bX^*_3)^\top 
    \bar \bX^{*;-1}_{1;\lambda} 
    \, R^{\varphi_\theta, \varphi_{\theta_0}} \,
    (\bW^{(0)})^\top \beta 
    \\
    \;\coloneqq&\;
    \beta^\top \bW^{(0)} M^{\varphi_{\theta_0}} (\bW^{(0)})^\top \beta
    +
    \sigma_\epsilon^2
    +
    L^1_\lambda
    +
    L^2_\lambda
    +
    L^3_\lambda
    \;.
\end{align*}
We first consider $L^1_\lambda$ and $L^3_\lambda$: Note that 
\begin{align*}
    | L^3_\lambda - L^3_0 |
    \;=&\;
    2 
    \big| 
        \beta^\top \bW^{(0)} (\bar \bX_3^*)^\top   
        \big( ( \bar \bX_1^* + \lambda \bI_{p'})^{-1} - ( \bar \bX_1^* )^\dagger   \big)
            \, R^{\varphi_\theta, \varphi_{\theta_0}} \,
        (\bW^{(0)})^\top \beta 
    \big|
    \\
    \;\leq&\;
    2
    \| \beta^\top \bW^{(0)} (\bar \bX_3^*)^\top   
        \big( ( \bar \bX_1^* + \lambda \bI_{p'})^{-1} - ( \bar \bX_1^* )^\dagger   \big) \|
    \,
    \|  R^{\varphi_\theta, \varphi_{\theta_0}} \,
        (\bW^{(0)})^\top \beta  \|
    \\
    \;=&\;
    O\Big(  \| (\bar \bX_3^*)^\top   
        \big( ( \bar \bX_1^* + \lambda \bI_{p'})^{-1} - ( \bar \bX_1^* )^\dagger   \big) \|_{op} \Big)
    \quad 
    \text{ with probability $1 - o(1)$, }
\end{align*}
where we have noted that $\| R^{\varphi_\theta, \varphi_{\theta_0}}  \|_{op} = O(1)$, $\| \beta \| =O(1)$ and that $\| \bW^{(0)} \|_{op} = O(1)$ with probability $1-o(1)$ by \eqref{eq:gaussian:op:bound}.  Similarly since $\| M^{\varphi_\theta} \|_{op} =O(1)$, almost surely 
\begin{align*}
    | L^1_\lambda - L^1_0 |
    \;=\;
    O\Big( \| (\bar \bX_3^*)^\top   
        \big( ( \bar \bX_1^* + \lambda \bI_{p'})^{-1} - ( \bar \bX_1^* )^\dagger   \big) \|_{op}^2  \Big)
    \;.
\end{align*}
Recall that  $(\lambda_l(A), v_l(A))$ denotes the $l$-th eigenvalue-eigenvector pair of a symmetric matrix $A \in \R^{p' \times p'}$. By the triangle inequality,
\begin{align*}
    &\;
    \| (\bar \bX_3^*)^\top   
        \big( ( \bar \bX_1^* + \lambda \bI_{p'})^{-1} - ( \bar \bX_1^* )^\dagger   \big) \|_{op}
    \\
    &\;\leq\;
    \Big\| 
       \msum_{l \leq p', \lambda_l(\bar \bX^*_1) > 0}
       (\bar \bX_3^*)^\top   
        v_l(\bar \bX_1^*  ) v_l(\bar \bX_1^* )^\top 
        \Big( 
            \mfrac{1}{\lambda_l(\bar \bX_1^*) + \lambda}
            -
            \mfrac{1}{\lambda_l(\bar \bX_1^*)}
        \Big)
    \Big\|_{op}
    \\
    &\hspace{1em}
    +
    \Big\| 
       \msum_{l \leq p', \lambda_l(\bar \bX^*_1) = 0}\,
       (\bar \bX_3^*)^\top   
        v_l(\bar \bX_1^*  ) v_l(\bar \bX_1^* )^\top 
        \Big( \mfrac{1}{\lambda} - 0 \Big)
    \Big\|_{op}
    \\
    &\;\leq\;
    \lambda
    \,
    \Big\| 
       \msum_{l \leq p', \lambda_l(\bar \bX^*_1) > 0}
       (\bar \bX_3^*)^\top   
        v_l(\bar \bX_1^*  ) v_l(\bar \bX_1^* )^\top 
            \mfrac{1 }{\lambda_l(\bar \bX_1^*)^2 }
    \Big\|_{op}
    \\
    &\hspace{1em}
    +
    \mfrac{1}{\lambda}
    \Big\| 
       \msum_{l \leq p', \lambda_l(\bar \bX^*_1) = 0}\,
       (\bar \bX_3^*)^\top   
        v_l(\bar \bX_1^*  ) v_l(\bar \bX_1^* )^\top 
    \Big\|_{op}
    \\
    &\;\leq\; 
    \lambda \| \bar \bX_3^*  \|_{op}  \| (\bar \bX^*_1)^\dagger \|^2_{op}
    + 
    \mfrac{1}{\lambda}
    \Big\| 
       \msum_{l \leq p'}\,
       \ind_{\{ \lambda_l(\bar \bX^*_1) = 0 \}}
       (\bar \bX_3^*)^\top   
        v_l(\bar \bX_1^*  ) v_l(\bar \bX_1^* )^\top 
    \Big\|_{op}
    \\
    &\;=\;
    O(\lambda) + o( \lambda^{-1} )
    \quad 
    \text{ with probability } 1 - o(1)
    \;.
\end{align*}
In the last line, we have used \Cref{assumption:network:ridgeless:by:ridge}. On the other hand, since $\| M_\lambda\|_{op} = O(1)$, $L^2_\lambda$ can be handled in exactly the same way as $f^{(2)}_\lambda$ in \eqref{eq:ridgeless:by:ridge:f2} in the proof of \Cref{lem:ridgeless:by:ridge}, which gives 
\begin{align*}
    | L^2_\lambda - L^2_0 |
    \;=\; O\Big( \lambda +  \mfrac{1}{n \lambda^2} \Big)
    \text{ with probability } 1 - o(1)\;.
\end{align*}
By a union bound, we obtain that for $\lambda \leq 1$, with probability $1-o(1)$,
\begin{align*}
    | \hat L_\lambda(\cX)  - \hat L_0(\cX) | \;=\; O( \lambda ) + o( \lambda^{-2}  )\;.
\end{align*}
 By the definition of the Lévy–Prokhorov metric $d_P$ \eqref{eq:defn:prokhorov}, we obtain 
\begin{align*}
    d_P( \hat L_\lambda(\cX), \hat L_0(\cX) ) \;=\; O( \lambda ) + o( \lambda^{-2}  )\;.
\end{align*}
The same argument applies with $\cX$ replaced by $\cZ$ and gives 
\begin{align*}
    d_P( \hat L_\lambda(\cZ), \hat L_0(\cZ) ) \;=\; O( \lambda ) + o( \lambda^{-2}  )\;.
\end{align*}
Finally as in the last part of the proof of \Cref{prop:ridgeless:universality}, we can modify the argument of \cref{lem:compare:d_H} to show that $d_P$ is bounded from above by $d_{\cH^{(4)}}$ (up to a multiplicative constant and raising to some fractional power). By applying the triangle inequality to \eqref{eq:network:ridge:universality} and taking $\lambda \rightarrow 0^+$, we obtain the desired bound that 
\begin{align*}
    d_P( \hat L_0(\cX), \hat L_0(\cZ) ) 
    \;=\; 
    o(1)\;.
\end{align*}
\qed 

\vspace{1em}

\subsection{Proof of Proposition 13: Simple neural networks} \label{appendix:proof:ridgeless:extend} We seek to apply the first statement of \Cref{prop:network:extend:general}, which requires us to verify \Cref{assumption:network:generalize,assumption:network:bounded:cov,assumption:network:feature:gradient}. We first identify
\begin{align*}
    \bW^{(0)} \;=\; \bW^{(0)}_{N_0} \;,
    \quad 
    \varphi_0( \bv ) \;=\;  \bW^{(0)}_{N_0 -1 } \ldots \bW^{(0)}_1 \bv \;,
    \quad 
    \varphi( \bv ) \;=\; W_N \ldots W_1 \bv \;,
\end{align*}
and identify the data vectors as 
\begin{align*}
    \bV_{ij1} \;=&\; \pi_{ij}(\bV_i) \quad \text{(augmented data)}\;,
    \\
    \bV_{ij0} \;=&\; 
    \begin{cases}
        \bV_i & \text{ if $\tau_{ij}$ is identity (i.e.~do not augment labels) };
        \\
        \pi_{ij}(\bV_i) & \text{ if $\tau_{ij}$ is the oracle augmentation };
    \end{cases}
\end{align*}
 \Cref{assumption:network:generalize}(i)--(iii) are automatically satisfied. Moreover, $\varphi_0$ and $\varphi$ are both linear, which implies that 
 \begin{align*}
    \gamma_2^\varphi \;=\; \gamma_3^\varphi \;=\; 0 \;,
 \end{align*}
 whereas 
 \begin{align*}
    \gamma_1^\varphi 
    \;=\; 
    \max \big\{ 
        \| \bW^{(0)}_{N_0 -1 } \ldots \bW^{(0)}_1 \|_{op} 
        \,,\, 
        \| W_N \ldots W_1 \|_{op} 
    \big\}
    \;\leq\;
    \max \big\{ 
        \| \bW^{(0)}_{N_0 -1 } \ldots \bW^{(0)}_1 \|_{op} 
        \,,\, 
        C_{op}
    \big\} 
 \end{align*}
 where the last inequality follows from \Cref{assumption:pretrained:layers}. Write $\bW^{(0)}_{N_0 -1:1} \coloneqq \bW^{(0)}_{N_0 -1 } \ldots \bW^{(0)}_1$. Then conditioning on the event 
 \begin{align*}
     E \;\coloneqq\; \{ \|\bW^{(0)}_{N_0 -1:1}\|_{op} = O(1) \text{ as } n \rightarrow \infty \} \;,
 \end{align*}
 \Cref{assumption:network:feature:gradient} holds provided that $B_d = o(d^{1/2})$, which we verify later. Moreover by the Jensen's inequality and independence of $\bV_{\rm new}$ from $\bW^{(0)}$ and $ \bW^{(0)}_{N_0 -1:1}$,
 \begin{align*}
    &\;
    \| \mean[ \varphi_0(\bV_{\rm new}) \varphi_0(\bV_{\rm new})^\top | \bW^{(0)}_{N_0 -1:1} ] \|_{op}
    \\
    &\;=\;
    \big\| 
        \mean\big[ 
            \bW^{(0)} \bW^{(0)}_{N_0 -1:1}
            \mean\big[  \bV_{\rm new} \bV_{\rm new}^\top \big]  
            (\bW^{(0)}_{N_0 -1:1})^\top
            ( \bW^{(0)})^\top
        \,\big|\, \bW^{(0)}_{N_0 -1:1} \big]
    \|_{op}
    \\
    &\;\leq\;
     \mean\big[  
            \| \bW^{(0)} \|_{op}^2 
    \,\big|\, E \big]
    \;
    \| \bW^{(0)}_{N_0 -1:1} \|_{op}^2
    \;
    \| \mean\big[  \bV_{\rm new} \bV_{\rm new}^\top \big]   \|_{op} 
    \\
    &\;=\; O(\| \bW^{(0)}_{N_0 -1:1} \|_{op}^2) \;,
 \end{align*}
 where we have used the standard moment bound on $\| \bW^{(0)} \|_{op}$ (see \eqref{eq:gaussian:op:bound}) and the assumption that $\| \Var[\bV_1] \|_{op} = O(1)$. On the other hand,
 \begin{align*}
    &\;
    \| \mean[ \varphi(\bV_{\rm new}) \varphi(\bV_{\rm new})^\top ] \|_{op}
    \\
    &\;=\;
    \| W_N \ldots W_1  \mean\big[  \bV_{\rm new} \bV_{\rm new}^\top \big] (W_1)^\top \ldots (W_N)^\top   \|_{op}
    \;=\; O(1)\;.
 \end{align*}
 Therefore conditioning on $E$, \Cref{assumption:network:bounded:cov} holds. Now to verify the sub-Gaussianity condition in \Cref{assumption:network:generalize}(v), we recall that $\bV_i$'s are mean-zero and $1$-sub-Gaussian, whereas under the different augmentation schemes in \Cref{assumption:network:augmentation},
 \begin{proplist}
    \item \textbf{Noise injection:} $\pi_{ij}(\bV_i)$ is mean-zero and sub-Gaussian since the injected noise is mean-zero and sub-Gaussian;
    \item \textbf{Random cropping, sign-flipping and random permutations:} $\pi_{ij}(\bV_i)$ are mean-zero and $1$-sub-Gaussian conditioning on $\pi_{ij}$, and therefore also mean-zero and $1$-sub-Gaussian marginally.
 \end{proplist}
 This verifies that $\bV_{ij0}$'s and $\bV_{ij1}$'s are mean-zero and sub-Gaussian. Conditioning on $E$, the same holds for $\varphi_0(\bV_{ij0})$ and $\varphi(\bV_{ij1})$ since $\| W_N \ldots W_1 \|_{op} = O(1)$ and $\|\bW^{(0)}_{N_0-1:1} \|_{op} = O(1)$. This implies that \Cref{assumption:network:generalize}(v) holds conditioning on $E$. Finally to verify the local dependence condition in \Cref{assumption:network:generalize}(iv) and the assumption that $B_d = o(d^{1/2})$, we recall that $\bV_i$ are locally dependent with the maximal dependency neighborhood bounded as $o(d^{1/2})$. Under the different augmentation schemes in \Cref{assumption:network:augmentation},
 \begin{proplist}
    \item \textbf{Noise injection:} the additive noise vectors are also locally dependent;
    \item \textbf{Random cropping and sign-flipping:}  the transformations act coordinate-wise and preserve the local dependency neighborhoods;
    \item \textbf{Random permutations:} permutations preserve the partition $(P_l)_{l \leq N_d}$ of the index set $[d]$ with maximum set size satisfying $\sup_{l \leq N_d} | P_l| = O(1)$.
 \end{proplist}
 In all cases, each coordinate of $\pi_{ij}(\bV_i)$ depends on at most $o(d^{1/2})$ of the coordinates of $\pi_{ij}(\bV_i)$ and $o(k d^{1/2})$ of the coordinates of  $\pi_{ij'}(\bV_i)$ for $j' \neq j$. Since $k$ is fixed, we get that the local dependence condition of \Cref{assumption:network:generalize}(iv) is satisfied. Therefore conditioning on $\bW^{(0)}_{N_0-1:1}$ such that $E$ holds, we can apply the first statement of \Cref{prop:network:extend:general} to obtain that for every fixed $\lambda > 0$,
 \begin{align*}
    \msup_{h \in \cH^{(4)}}
    \Big| 
        \, \mean\Big[  h \big( \hat L_{\lambda}(\Phi\cX) \big) \,\big|\, \bW^{(0)}_{N_0-1:1} \Big] \ind_E
        \,-\,
        \, \mean\Big[  h \big( \hat L_{\lambda}(\cZ) \big)  \,\big|\, \bW^{(0)}_{N_0-1:1} \Big] 
        \ind_E
    \Big| 
    \;\rightarrow\; 0\;.
 \end{align*}
 By the discussion at the end of the proof of \Cref{prop:network:extend:general}, convergence in $d_{\cH^{(4)}}$ metrizes convergence in the L\'evy-Prokhorov metric $d_P$. Moreover, since $N_0$ is fixed and since $\bW^{(0)}_{N_0 -1 }, \ldots, \bW^{(0)}_1$  are independent random matrices with i.i.d.~normal entries and with number of rows and columns growing at most linearly in $n$, by a similar argument to \eqref{eq:gaussian:op:bound}, we have
 \begin{align*}
    \P(E) \;=\; 1 - o(1)\;.
 \end{align*}
 By the definition of $d_P$ \eqref{eq:defn:prokhorov}, we can remove the conditioning on $\bW^{(0)}_{N_0-1:1}$ and conclude that
 \begin{align*}
    d_P\big( \hat L_{\lambda}(\Phi\cX)\,,\, \hat L_{\lambda}(\cZ)   \big) \;\rightarrow\; 0\;.
 \end{align*}
\qed

\subsection{Proof of \Cref{lem:ridgeless:noniso:alt:rep}: Alternative expression of the augmented sample covariance matrix, non-isotropic case}  \label{appendix:proof:noniso:alt:rep}

Notice that the $\R^{nkd}$-valued random vector $ \bZ \coloneqq ( \bZ_{11}^\top, \ldots,  \bZ_{nk}^\top)^\top$ can be expressed as 
\begin{align*}
     \bZ \;=\; \sqrt{\Sigma} \, \tilde \eta  
\end{align*}
for a standard $\R^{nkd}$ Gaussian vector $\tilde \eta = (\tilde \eta_{11}^\top, \ldots, \tilde \eta_{nk}^\top)^\top$ and a $\R^{nkd \times nkd}$ covariance matrix $\Sigma$ defined as  
\begin{align*}
    \Sigma 
    \;\coloneqq&\; 
    \begin{psmallmatrix}
        \tilde \Sigma & & \\
        & \ddots & \\
        & & \tilde \Sigma 
    \end{psmallmatrix}
    \;=\;
    \bI_n \otimes \tilde \Sigma 
    \;,
    \\
    \tilde \Sigma 
    \;\coloneqq&\;
    \begin{psmallmatrix}
        V & R & & \cdots & R \\
        R & V &  R &  & \vdots \\
        \vdots & \ddots & \ddots & \ddots &  \\
         &  & R  & V & R \\
        R & \cdots & & R & V \\
    \end{psmallmatrix}
    \;=\;
    \bI_k \otimes (V-R)
    + 
    \bone_{k \times k} \otimes R
    \;\in\;
    \R^{kd \times kd}
    \;,
    \\
    V \;\coloneqq&\; \Var[\pi_{11}(\bV_1)]
    \;,
    \qquad 
    R \;\coloneqq\; \Cov[\pi_{11}(\bV_1)\,,\,\pi_{12}(\bV_1)]
    \;.
\end{align*}
 For $i \leq n$ and $j \leq k$, define the $\R^{d \times nkd}$ projection matrix 
\begin{align*}
    P_{ij}
    \;\coloneqq\;
    ( 0, \ldots, 0, \bI_d, 0, \ldots, 0 )
\end{align*}
where $\bI_d$ appears at the $((i-1)k + j)$-th $d \times d$ block. Then we can express the two matrices of interest as
\begin{align*}
    \bar \bZ_1
    \;=&\;
    \mfrac{1}{nk} \msum_{i \leq n, j \leq k} \bZ_{ij}  \bZ_{ij}^\top 
    \;=\;
    \mfrac{1}{nk} \msum_{i \leq n, j \leq k} (P_{ij}  \bZ) (P_{ij}  \bZ)^\top 
    \\
    \;=&\;
    \mfrac{1}{nk} \msum_{i \leq n, j \leq k} (P_{ij} \sqrt{\Sigma} \, \tilde  \eta) (P_{ij} \sqrt{\Sigma} \, \tilde \eta)^\top 
    \;,
    \\
    \bar \bZ_2 
    \;=&\;
    \mfrac{1}{n} \msum_{i \leq n} 
    \Big(\mfrac{1}{k} \msum_{j \leq k} \bZ_{ij} \Big) 
    \Big(\mfrac{1}{k} \msum_{j \leq k} \bZ_{ij} \Big)^\top 
    \\
    \;=&\;
    \mfrac{1}{n} \msum_{i \leq n} 
    \Big(\mfrac{1}{k} \msum_{j \leq k} P_{ij} \sqrt{\Sigma} \, \tilde  \eta \Big) 
    \Big(\mfrac{1}{k} \msum_{j \leq k} P_{ij} \sqrt{\Sigma} \, \tilde  \eta \Big)^\top 
    \;.
\end{align*}
By noting that $\frac{1}{k} \bone_{k \times k} = \frac{1}{k} \bone_k \bone_k^\top$ is a projection matrix, one can verify that
\begin{align*}
    \sqrt{\Sigma}
    \;=&\;
    \bI_n \otimes 
    \Big( 
        \big(\bI_k - \mfrac{1}{k} \bone_{k \times k} \big)
        \otimes 
        \sqrt{V-R}
        +
        \mfrac{1}{k} \bone_{k \times k} 
        \otimes
        \sqrt{V-R + k R}
    \Big)
    \;,
\end{align*}
where $V-R$ is positive semi-definite by \Cref{lem:compare_variance}. Let $\tilde O_k \in \R^{k \times k}$ be an orthogonal matrix such that the first column vector is $o_1 \coloneqq \frac{1}{\sqrt{k}} \bone_k$ and the remaining column vectors are $o_2, \ldots, o_k$. Then we can write 
\begin{align*}
     \bI_k - \mfrac{1}{k} \bone_{k \times k} 
     \;=&\;
     \tilde O_k^\top 
     \diag\{0,1,\ldots,1\}
     \,
     \tilde O_k
     &\text{ and }&&
     \mfrac{1}{k} \bone_{k \times k} 
     \;=&\;
     \tilde O_k^\top 
     \diag\{1,0,\ldots,0\}
     \,
     \tilde O_k
     \;.
\end{align*} 
Also denote the $\R^{kd}$ random vectors
\begin{align*}
   \tilde \eta_i \;\coloneqq\; (\tilde\eta_{i1}^\top, \ldots, \tilde \eta_{ik}^\top )^\top
\end{align*}
which are independent across $1 \leq i \leq n$. By the orthogonal invariance of the Gaussian distribution, we have 
\begin{align*}
    &\;P_{ij}
    \sqrt{\Sigma} \, \tilde \eta 
    \;=\;
    P_{ij}
    \Big(
    \bI_n \otimes 
    \Big( 
        \big(\bI_k - \mfrac{1}{k} \bone_{k \times k} \big)
        \otimes 
        \sqrt{V-R}
        +
        \mfrac{1}{k} \bone_{k \times k} 
        \otimes
        \sqrt{V-R + k R}
    \Big)
    \Big)
    \,
    \tilde \eta
    \\
    &\;\overset{d}{=}\;
    P_{ij}
    \Big(
    \bI_n \otimes 
    \Big( 
        \big( 
            \tilde O_k^\top 
            \diag\{0,1,\ldots,1\}
        \big)
        \otimes 
        \sqrt{V-R}
    \\
    &\hspace{5em}
        +
        \big( 
            \tilde O_k^\top 
            \diag\{1,0,\ldots,0\}
        \big)
        \otimes
        \sqrt{V-R + k R}
    \Big)
    \Big)
    \,
    \tilde \eta
    \\
    &\;=\;
    P_{ij}
    \Big(
    \bI_n \otimes 
    \Big( 
        \begin{psmallmatrix}
            \leftarrow \bzero^\top \rightarrow 
            \\
            \leftarrow o_2^\top \rightarrow 
            \\
            \vdots 
            \\
            \leftarrow o_k^\top \rightarrow 
        \end{psmallmatrix}
        \otimes 
        \sqrt{V-R}
        +
        \begin{psmallmatrix}
            \leftarrow o_1^\top \rightarrow
            \\
            \leftarrow \bzero^\top \rightarrow 
            \\
            \vdots 
            \\
            \leftarrow \bzero^\top \rightarrow 
        \end{psmallmatrix}
        \otimes
        \sqrt{V-R + k R}
    \Big)
    \Big)
    \,
    \tilde \eta
    \\
    &\;=\;
    \ind_{\{j \neq 1\}}
        \big(\,
        o_j^\top
        \otimes 
        \sqrt{V-R}
        \,\big)
        \,
         \tilde \eta_i
        +
    \ind_{\{j = 1\}}
        \big( \, o_1^\top \otimes \sqrt{V-R + k R} \, \big)
        \,
        \tilde \eta_i
    \\
    &\;\overset{d}{=}\;
    \ind_{\{j \neq 1\}} \, \sqrt{V-R}  \; \eta_{ij}
    +
    \ind_{\{j = 1\}} \, \sqrt{V-R+kR} \; \eta_{ij}
    \\
    &\;=\;
    \ind_{\{j \neq 1\}} \, \Sigma_2^{1/2}  \; \eta_{ij}
    +
    \ind_{\{j = 1\}} \, \Sigma_1^{1/2} \; \eta_{ij}
    \;.
\end{align*}
In the second last line, we have noted that since the $o_j$'s are orthogonal vectors, $(o_j^\top \otimes \sqrt{V-R}) \, \tilde \eta_i$'s live in orthogonal subspaces across $1 \leq j \leq n$ and are thereby independent, and therefore we can re-express them through the i.i.d.~$\cN(0,I_d)$ vectors $\eta_{ij}$. This implies that $(\bar \bZ_1, \bar \bZ_2)$ is identically distributed as
\begin{align*}
    \Big( 
        &\,
        \mfrac{1}{nk} \msum_{i \leq n} 
        \Sigma_1^{1/2}
        \; \eta_{i1} \eta_{i1}^\top \; 
        \Sigma_1^{1/2}
    +
    \mfrac{1}{nk} \msum_{i \leq n} \msum_{j=2}^k
    \Sigma_2^{1/2} \; \eta_{ij} \eta_{ij}^\top \; \Sigma_2^{1/2}
    \;,
    \\
    &\,
    \mfrac{1}{n} \msum_{i \leq n} 
    \Big( \mfrac{1}{k} \Sigma_1^{1/2} \; \eta_{i1} + \mfrac{1}{k} \msum_{j = 2}^k \Sigma_2^{1/2} \; \eta_{ij}  \Big)
    \Big( \mfrac{1}{k} \Sigma_1^{1/2} \; \eta_{i1} + \mfrac{1}{k} \msum_{j = 2}^k \Sigma_2^{1/2} \; \eta_{ij}  \Big)^\top
    \Big)
\end{align*}
as desired.
\qed

\section{Proofs for Section \ref{sec:bagging} and Appendix \ref{appendix:bagging}} \label{appendix:proof:bagging}

This appendix collects the proofs related to bagging of a generic estimator: 
\begin{itemize}
    \item \Cref{appendix:proof:bagging:general} proves \Cref{prop:bagging:general} in \Cref{appendix:statistics:of:bagging}, which concerns the stability of a generic statistic of a bagged estimator;
    \item \Cref{appendix:proof:bagging:quad} proves \Cref{lemma:bagging:quadratic:universality} in \Cref{appendix:statistics:of:bagging}, which concerns the stability of a statistic that has quadratic dependence on the randomization in bagging; 
    \item \Cref{appendix:proof:prop:bagging} proves \Cref{prop:bagging} in \Cref{sec:bagging}, which applies \Cref{prop:bagging:general} to study the stability of a bagged estimator; 
    \item \Cref{appendix:proof:prop:bagged:network} proves \Cref{prop:bagged:network} in \Cref{appendix:bagged:network}, which concerns the universality of a bagged-and-augmented nonlinear feature model;
    \item \Cref{appendix:proof:cor:bagging} proves \Cref{cor:bagging:network:nonlinear} in \Cref{appendix:bagged:network}, which concerns the universality of a bagged-and-augmented nonlinear neural network;
    \item \Cref{appendix:proof:tanh} proves \Cref{lem:tanh}, which verifies the assumptions for a bagged-and-augmented nonlinear neural network with tanh activations.
\end{itemize}

\subsection{Proof of \Cref{prop:bagging:general}: Stability of generic statistics of a bagged estimator.}  \label{appendix:proof:bagging:general}

Fix $i \in [n]$ and, for simplicity, write 
\begin{align*}
    \bV^{(i)}_{i'} \;\coloneqq\; \Phi_I\bX_i \text{ for } i' < i\;,
    \quad 
    \bV^{(i)}_i \;\coloneqq\; \bw\;,
    \quad 
    \bV^{(i)}_{i'} \;\coloneqq\; \bZ_i \text{ for } i > i\;.
\end{align*}

\textbf{Step 1: First derivative.} By the chain rule, we can compute
\begin{align*}
    (\star)_1 &\;\coloneqq\;
    \bigl\|{\sup_{\bw \in[\bzero,\Phi_i\bX_i]}}
    \|D_i (g \circ f_m^{(B)})(\bW_i(\bw))\|\bigr\|_{L_{6}}
    \\
    &\;=\;
    \bigl\|{\sup_{\bV_i\in[\bzero,\Phi_i\bX_i]}}
    \| 
        \partial g\big( f_m^{(B)}(\bW_i(\bV_i)) \big)
        D_i f_m^{(B)}(\bW_i(\bV_i))
    \|\bigr\|_{L_{6}}
    \\
    &\;=\;
    \Big\|{\sup_{\bV_i\in[\bzero,\Phi_i\bX_i]}}
    \Big\| 
        \mfrac{1}{B} \msum_{b \leq B} 
        \Big(
        \underbrace{\partial g\big( f_m^{(B)}(\bW_i(\bV_i)) \big)
    \,
        D_i  
        \, f_m
        \big( \bV^{(i)}_{\upsilon_b(1)}, \ldots, \bV^{(i)}_{\upsilon_b(m)} \big)
        }_{\eqqcolon S^i_b(\bV_i)}
        \Big)
   \Big \|\Big\|_{L_{6}}
   \;.
\end{align*}
Now denote the event  $E^i_b = \{  i \in \{ \upsilon_b(l) \}_{l \leq m} \}$, i.e.~the event where $i$ is included in the $b$-th bagged estimator. Notice that almost surely,
\begin{align*}
    D_i \, f_m
        \big( \bV^{(i)}_{\upsilon_b(1)}, \ldots, \bV^{(i)}_{\upsilon_b(m)} \big)
    \;=\;
    D_i \, f_m
        \big( \bV^{(i)}_{\upsilon_b(1)}, \ldots, \bV^{(i)}_{\upsilon_b(m)} \big)
    \, \ind_{E^i_b}\;.
\end{align*}
Let $\mean_\bV[\argdot] \coloneqq \mean[ \argdot | \bV_1, \ldots, \bV_n]$, i.e.~the conditional expectation is taken over $\upsilon_1, \ldots, \upsilon_B$. Plugging this expression in, applying the triangle inequality and centering the summands with respect tos $\mean_\bV$, we can obtain 
\begin{align*}
    (\star)_1
    \;=&\;
    \Big\|{\sup_{\bV_i\in[\bzero,\Phi_i\bX_i]}}
    \Big\| 
        \mfrac{1}{B} \msum_{b \leq B} 
            S^i_b(\bV_i) \, \ind_{E^i_b}
   \Big \|\Big\|_{L_{6}}
   \\
   \;\leq&\;
   \Big\|
        \mfrac{1}{B} \msum_{b \leq B} 
        {\sup_{\bV_i\in[\bzero,\Phi_i\bX_i]}}
        \big\| 
            S^i_b(\bV_i) \, \ind_{E^i_b}
        \big\|
   \Big\|_{L_{6}}
   \\
   \;\leq&\;
   \Big\|
        \mfrac{1}{B} \msum_{b \leq B} 
        \Big\{ 
            {\sup_{\bV_i\in[\bzero,\Phi_i\bX_i]}}
            \big\| 
                S^i_b(\bV_i) \, \ind_{E^i_b}
            \big\|
            -
            \mean_\bV\big[ 
                \sup_{\bV_i\in[\bzero,\Phi_i\bX_i]}
                \big\| 
                    S^i_b(\bV_i) \, \ind_{E^i_b}
                \big\|
            \big]
        \Big\}
   \Big\|_{L_{6}}
   \\
   &
   +
   \Big\|
        \mean_\bV\big[ 
                \sup_{\bV_i\in[\bzero,\Phi_i\bX_i]}
                \big\| 
                    S^i_1(\bV_i) \, \ind_{E^i_1}
                \big\|
            \big]
   \Big\|_{L_{6}}
   \\
   \;\eqqcolon&\; (\star)_{11} + (\star)_{12}
   \;.
\end{align*}
Conditioning on $\Phi\cX$ and $\cZ$ and focusing purely on the stochasticity of $\upsilon_1, \ldots, \upsilon_B$, the first term is the $L_6$-th norm of a sum of independent and mean-zero quantities, so by \Cref{lem:rosenthal}, there is some absolute constant $C_1 > 0$ such that 
\begin{align*}
    (\star)_{11} 
    \;\leq\; 
    \mfrac{C_1}{\sqrt{B}} 
    \,
    \big\| 
        {\msup_{\bV_i\in[\bzero,\Phi_i\bX_i]}}
            \big\| 
                S^i_1(\bV_i) \, \ind_{E^i_1}
            \big\|
    \big\|_{L_6}
    \;.
\end{align*}
We now need a control on $\P(E^i_1)$. By a union bound, we have
\begin{align*}
    \P(E^i_1)
    \;\leq\;
    \msum_{l=1}^m \P( i = \upsilon_1(l) ) \;=\; \mfrac{m}{n}\;.
\end{align*}
Using this expression and the H\"older inequality, we have that for any fixed $t > 0$,
\begin{align*}
    (\star)_{11} 
    \;\leq&\; 
    \mfrac{C_1}{\sqrt{B}} 
    \,
     \P( E^i_1)^{\frac{t}{36+6t}}
    \,
    \big\| 
        {\msup_{\bV_i\in[\bzero,\Phi_i\bX_i]}}
            \big\| 
                S^i_1(\bV_i)  
            \big\|
    \big\|_{L_{6+t}}
    \\
    \;\leq&\;
    \mfrac{C_1}{\sqrt{B}} \, \mfrac{m^{t/(36+6t)}}{n^{t/(36+6t)}} \,
    \big\| 
        {\msup_{\bV_i\in[\bzero,\Phi_i\bX_i]}}
            \big\| 
                S^i_1(\bV_i)  
            \big\|
    \big\|_{L_{6+t}}
    \;.
\end{align*}
To handle $(\star)_{12}$, notice that $S^i_1(\bV_i) = \bzero$ on the complement event $(E^i_1)^c$ and that $E^i_1$ is independent of $\bV=(\bV_1, \ldots, \bV_n)$. This implies 
\begin{align*}
    (\star)_{12}
    \;=&\;
    \Big\|
        \P_\bV( E^i_1 )
        \;
        \mean_\bV\big[ 
                \sup_{\bV_i\in[\bzero,\Phi_i\bX_i]}
                \big\| 
                    S^i_1(\bV_i)
                \big\|
                \,\big|\, E^i_1
            \big]
   \Big\|_{L_{6}}
   \\
   \;=&\;
   \P(E^i_1)
   \,
   \Big\|
        \,
        \mean_\bV\big[ 
                \sup_{\bV_i\in[\bzero,\Phi_i\bX_i]}
                \big\| 
                    S^i_1(\bV_i)
                \big\|
                \,\big|\, E^i_1
            \big]
   \Big\|_{L_{6}}
   \\
   \;\leq&\;
   \mfrac{m}{n} 
   \, 
   \big\| 
        {\msup_{\bV_i\in[\bzero,\Phi_i\bX_i]}}
            \big\| 
                S^i_1(\bV_i)  
            \big\|
    \big\|_{L_{6+t}}\;,
\end{align*}
where we have used the Jensen's inequality and that $L_6$-norm is bounded from above by $L_{6+t}$-th norm in the last line. Combining the computations gives 
\begin{align*}
    &\;
    \bigl\|{\sup_{\bw \in[\bzero,\Phi_i\bX_i]}}
    \|D_i (g \circ f_m^{(B)})(\bW_i(\bw))\|\bigr\|_{L_{6}}
   \\
   &\;\leq\;
   \Big( \mfrac{C_1}{\sqrt{B}} \mfrac{m^{t/(36+6t)}}{n^{t/(36+6t)}} + \mfrac{m}{n} \Big)
   \, 
   \big\| 
        {\msup_{\bV_i\in[\bzero,\Phi_i\bX_i]}}
            \big\| 
                S^i_1(\bV_i)  
            \big\|
    \big\|_{L_{6+t}}
    \\
    &\;\overset{(a)}{=}\;
    o\Big( \mfrac{\alpha^{(m)}_{1;t}}{\sqrt{n}}
    \Big)
    \;,
\end{align*}
In $(a)$, we have used $m=o(\sqrt{n})$ and $B \gg n^{1 - t/(108+18t)} \gg n^{1 - t/(36+6t)}$; in $(b)$, we have used that 
\begin{align*}
    &\;
    \big\| 
        {\msup_{\bV_i\in[\bzero,\Phi_i\bX_i]}}
            \big\| 
                S^i_1(\bV_i)  
            \big\|
    \big\|_{L_{6+t}}
    \\
    &\;=
    \Big(
        \mean\Big[
        \mean\Big[
                {\sup_{\bV_i\in[\bzero,\Phi_i\bX_i]}}
                \Big\|
                \partial g\big( f_m^{(B)}(\bW_i(\bV_i)) \big)
                \,
                    D_i  
                    \, f_m
                    \big( \bV^{(i)}_{\upsilon_b(1)}, \ldots, \bV^{(i)}_{\upsilon_b(m)} \big)
                \Big\|^{6+t}
        \,\Big|\, \upsilon_b \Big]
    \Big] 
    \Big)^{\frac{1}{6+t}}
    \\
    &\;\leq 
    \max_{\substack{i \leq n \\ i' \leq m \\ \upsilon \in S([m])}}
    \Big(
        \mean\Big[
        \mean\Big[
                {\sup_{\bw \in[\bzero,\Phi_i\bX_i]}}
                \Big\|
                \partial g\big( f_m^{(B)}(\bW_i(\bV_i)) \big)
                \,
                    D_{i'}  
                     f_m
                    \big( \bW^\upsilon_{i'}(\bw)  \big)
                \Big\|^{6+t}
        \,\Big|\, \upsilon_b \Big]
    \Big] 
    \Big)^{\frac{1}{6+t}}
    \\
    &\;=
    \max_{\substack{i \leq n \\ i' \leq m \\ \upsilon \in S([m])}}
    \Big\|
                {\sup_{\bw \in[\bzero,\Phi_i\bX_i]}}
                \Big\|
                \partial g\big( f_m^{(B)}(\bW_i(\bV_i)) \big)
                \,
                    D_{i'}  
                     f_m
                    \big( \bW^\upsilon_{i'}(\bw)  \big)
                \Big\|
    \,\Big\|_{L_{6+t}}
    \\
    &\;\leq \alpha^{(m)}_{1;t}
    \;.
\end{align*}
The same argument applies for all $i \leq n$ and for $\sup_{\bw \in [\bzero, \Phi_i\bX_i]}$ replaced with $\sup_{\bw \in [\bzero, \bZ_i]}$, and therefore 
\begin{align*}
    \alpha_1^{(B)} 
    \;=\;
    o\Big( \mfrac{\alpha^{(m)}_{1;t}}{\sqrt{n}}
    \Big)\;.
\end{align*}

\textbf{Step 2: Second and third derivatives.}  The arguments for the second and third derivatives are similar. For $r=2$, by applying the chain rule we obtain 
\begin{align*}
    (\star)_2
    \;=&\;
    \Big\|{\sup_{\bV_i\in[\bzero,\Phi_i\bX_i]}}
    \| 
        \partial g\big( f_m^{(B)}(\bW_i(\bV_i)) \big)
        D^2_i f_m^{(B)}(\bW_i(\bV_i))
    \\
    &\;
        \hspace{5em}
        +
        \partial^2 g\big( f_m^{(B)}(\bW_i(\bV_i)) \big)
        \big( D_i f_m^{(B)}(\bW_i(\bV_i)) \otimes D_i f_m^{(B)}(\bW_i(\bV_i)) \big)
    \|\Big\|_{L_{6}}
    \\
    \;=&\;
    \Big\|{\sup_{\bV_i\in[\bzero,\Phi_i\bX_i]}}
    \Big\| 
        \mfrac{1}{B} \msum_{b \leq B}
        \partial g\big( f_m^{(B)}(\bW_i(\bV_i)) \big)
        D^2_i f_m\big(\bV^{(i)}_{\upsilon_b(1)}, \ldots, \bV^{(i)}_{\upsilon_b(m)}\big)
    \\
    &\;
        \hspace{5em}
        +
        \mfrac{1}{B^2} \msum_{b,b' \leq B}
        \partial^2 g\big( f_m^{(B)}(\bW_i(\bV_i)) \big)
        \big( 
            D_i f_m\big(\bV^{(i)}_{\upsilon_b(1)}, \ldots, \bV^{(i)}_{\upsilon_b(m)}\big)
    \\
    &\;\hspace{18em}
            \otimes 
            D_i f_m\big(\bV^{(i)}_{\upsilon_{b'}(1)}, \ldots, \bV^{(i)}_{\upsilon_{b'}(m)}\big)
        \big)
    \Big\| 
    \, \Big\|_{L_{6}}
    \\
    \;\leq&\;
    \| \bar Q^{i;1}  \|_{L_6}
    +
    \| \bar Q^{i;2}  \|_{L_6}
   \;,
\end{align*}
where we have defined 
\begin{align*}
    &\;\bar Q^{i;1} \;\coloneqq\; \mfrac{1}{B} \msum_{b \leq B} Q_{b}^{i;1} 
    \;,
    \qquad\qquad 
    \bar Q^{i;2} \;\coloneqq\; \mfrac{1}{B^2} \msum_{b,b' \leq B} Q_{b,b'}^{i;2} \;,
    \\
    &\; Q_{b}^{i;1}  \;\coloneqq\;  {\sup_{\bV_i\in[\bzero,\Phi_i\bX_i]}}
        \Big\| 
            \partial g\big( f_m^{(B)}(\bW_i(\bV_i)) \big)
            D^2_i f_m\big(\bV^{(i)}_{\upsilon_b(1)}, \ldots, \bV^{(i)}_{\upsilon_b(m)}\big)
        \Big\| 
    \;,
    \\  
    &\;
    Q_{b,b'}^{i;2}  \;\coloneqq\;  {\sup_{\bV_i\in[\bzero,\Phi_i\bX_i]}}
        \Big\| 
            \partial^2 g\big( f_m^{(B)}(\bW_i(\bV_i)) \big)
        \Big( 
            D_i f_m\big(\bV^{(i)}_{\upsilon_b(1)}, \ldots, \bV^{(i)}_{\upsilon_b(m)}\big)
    \\
    &\hspace{18em}    
        \otimes 
            D_i f_m\big(\bV^{(i)}_{\upsilon_{b'}(1)}, \ldots, \bV^{(i)}_{\upsilon_{b'}(m)}\big)
        \Big)
        \Big\| 
        \;.
\end{align*}
The argument for controlling $\bar Q^{i;1}$ is identical to the proof for $r=1$ by using the event $E^i_b$, conditioning on the data $\bV_1, \ldots, \bV_n$ and focusing only on the randomness of $\upsilon_1, \ldots, \upsilon_B$. This yields 
\begin{align*}
    \| \bar Q^{i;1}  \|_{L_6}
    \;\leq&\; 
    \max_{b \leq B} \|  \mean_\bV[  Q^{i;1}_b ] \|_{L_6}
    +
    \Big\|  \mfrac{1}{B} \msum_{b \leq B} \big( Q_{b}^{i;1} - \mean_\bV[Q_{b}^{i;1} ])  \Big\|_{L_6}
    \\
    \;\leq&\;
    \max_{b \leq B} \| \P_\bV(E^i_b) \mean_\bV[  Q^{i;1}_b | E^i_b ] \|_{L_6}
    +
    \mfrac{C_1}{\sqrt{B}}  \|  Q_{b}^{i;1} - \mean_\bV[Q_{b}^{i;1} ] \|_{L_6}
    \\
    \;\leq&\;
    \mfrac{m}{n} 
    \| Q^{i;1}_1  \|_{L_6}
    + 
    \mfrac{2 C_1}{\sqrt{B}} 
    \| Q^{i;1}_1  \|_{L_6}
    \\
    \;\leq&\; 
    \Big( \mfrac{m}{n} + \mfrac{2 C_1 \, m^{t/(36+6t)}}{B^{1/2} \, n^{t/36+6t}} \Big) \, \alpha^{(m)}_{2,1;t}
    \;.
\end{align*}
To control $\bar Q^{i;2}$, the main difference is that we now need to handle a double-sum. Instead of applying \Cref{lem:rosenthal}, we make use of Burkholder's bound on the moment of a sum of martingale difference sequences \citep{burkholder1966martingale}, where an explicit constant is given by e.g.~\cite{von1965inequalities}: For a martingale difference sequence $Y_1,\ldots,Y_n$ taking values in $\R$ and $\nu \geq 2$, there exists some constant $C_\nu > 0$ that depends only on $\nu$ such that
\begin{align*}
    \mean \big[ \big| \msum_{i=1}^n Y_i \big|^\nu \big] 
    \;\leq\; 
    C_{\nu} \, n^{\max\{0,\,\nu/2-1\}} \msum_{i=1}^n \mean[ |Y_i |^\nu ]\;.
    \tagaligneq \label{eq:MDS:bound}
\end{align*}
By the triangle inequality followed by applying \eqref{eq:MDS:bound} with respect to $\| \argdot \|_{L_6 | \bV}$, we get that for is some absolute constant $C'_1>0$, 
\begin{align*}
    \big\| \bar Q^{i;2}  \big\|_{L_6}
    \;\leq&\;
    \big\| \mean_\bV\big[  \bar Q^{i;2}  \big]  \big\|_{L_6 }
    +
    \Big\|  \msum_{\tilde b=1}^B 
    \big( 
        \mean_\bV\big[  \bar Q^{i;2} \,\big|\, \upsilon_{\tilde b}, \ldots, \upsilon_1  \big] 
        -
        \mean_\bV\big[  \bar Q^{i;2} \,\big|\, \upsilon_{\tilde b-1}, \ldots, \upsilon_1  \big] 
    \big) \Big\|_{L_6 }
    \\
    \;\overset{\eqref{eq:MDS:bound}}{\leq}&\;
    \big\| \mean_\bV\big[  \bar Q^{i;2}  \big]  \big\|_{L_6 }
    \\
    &\;+
    C'_1  B^{1/2} 
    \Big\|
        \Big( \mfrac{1}{B} \msum_{\tilde b=1}^B 
        \\
    &\hspace{6em}
        \underbrace{\mean_\bV \, \big|
        \mean\big[  \bar Q^{i;2} \,\big|\, \upsilon_{\tilde b}, \ldots, \upsilon_1  \big] 
        -
        \mean\big[  \bar Q^{i;2} \,\big|\, \upsilon_{\tilde b-1}, \ldots, \upsilon_1  \big] 
        \, \big|^6
        }_{ (\Delta)_{\tilde b} }
  \Big)^{1/6}
  \Big\|_{L_6}
  \\
  \;\eqqcolon&\; (\star)_{22} + (\star)_{21}
  \;.
\end{align*}
$(\star)_{22}$ is controlled in a similar way as $(\star)_{12}$ by using $E^i_b$ and $E^i_{b'}$ and noting that $Q^{i;2}_{b,b'} = 0$ on the event $(E^i_b)^c \cup (E^i_b)^c$:
\begin{align*}
    (\star)_{22} 
    \;\leq&\;
    \mfrac{1}{B^2} \, \msum_{b \neq b'}^B
    \big\| \mean_\bV[ Q^{i;2}_{b,b'} ] \big\|_{L_6}
    +
    \mfrac{1}{B^2} \, \msum_{b \leq B}
    \big\| \mean_\bV[ Q^{i;2}_{b,b} ] \big\|_{L_6}
    \\
    \;=&\;
    \max_{\substack{b,b' \\ b \neq b'}}
    \big\| \P_\bV( E^i_b \cap E^i_{b'}  ) \mean_\bV[ Q^{i;2}_{b,b'} \big| E^i_b \cap E^i_{b'} \big] \big\|_{L_6}
    \\
    &\;
    +
    \mfrac{1}{B} \, \max_{b \leq B}
    \big\| \P_\bV( E^i_b   ) \mean_\bV[ Q^{i;2}_{b,b'} \big| E^i_b \big] \big\|_{L_6}
    \\
    \;\leq&\; 
    \mfrac{m^2}{n^2} \, \max_{b \neq b' } \| Q^{i;2}_{b,b'}  \|_{L_6}
    + 
    \mfrac{m}{n B} \, \max_{b} \| Q^{i;2}_{b,b}  \|_{L_6}
    \\
    \;\leq&\;
    \Big( \mfrac{m^2}{n^2}  + \mfrac{m}{n} \mfrac{m^{t/(36+6t)}}{B n^{t/(36+6t)}}  \Big) \alpha^{(m)}_{2,2;t}
    \;.
\end{align*}
To control $(\star)_{21}$, we notice that the only terms involving $\upsilon_{\tilde b}$ appear in the difference $(\Delta)_{\tilde b}$, and therefore
\begin{align*}
    (\star)_{21}
    \;=&\;
    C'_1  B^{1/2} 
    \Big\|
        \Big( \mfrac{1}{B} \msum_{\tilde b=1}^B 
    \\
    & \hspace{5em}
    \mean_\bV
    \,
    \Big|
    \mfrac{2}{B^2} \msum_{b \neq \tilde b} 
    \Big(   
        \mean_\bV\big[ Q^{i;2}_{b,\tilde b} \,\big|\, \upsilon_{\tilde b}, \ldots, \upsilon_1  \big]
        -
        \mean_\bV\big[ Q^{i;2}_{b,\tilde b} \,\big|\, \upsilon_{\tilde b-1}, \ldots, \upsilon_1  \big]
    \Big)
    \\
    & \hspace{7em}
    +
    \mfrac{1}{B^2} 
    \, 
    \mean_\bV\big| 
        \mean_\bV[ Q^{i;2}_{\tilde b, \tilde b} \,\big|\, \upsilon_{\tilde b}, \ldots, \upsilon_1  \big]
        -
        \mean_\bV\big[ Q^{i;2}_{\tilde b,\tilde b} \,\big|\, \upsilon_{\tilde b-1}, \ldots, \upsilon_1  \big]
    \Big|^6
    \\
    & \hspace{30em}
    \Big)^{1/6}
    \Big\|_{L_6}
    \\
    \;\leq&\;
    \mfrac{2 C'_1 }{B^{1/2}}
    \max_{\tilde b \leq B}
    \Big\| 
            \underbrace{
                \mfrac{1}{B} \msum_{b \neq \tilde b}
                \Big(   
                    \mean_\bV\big[ Q^{i;2}_{b,\tilde b} \,\big|\, \upsilon_{\tilde b}, \ldots, \upsilon_1  \big]
                    -
                    \mean_\bV\big[ Q^{i;2}_{b,\tilde b} \,\big|\, \upsilon_{\tilde b-1}, \ldots, \upsilon_1  \big]
                \Big)
            }_{\eqqcolon \bar T^{i,\tilde b}}
    \Big\|_{L_6}
    \\
    &\;
    +
    \mfrac{C'_1}{B^{3/2}} \max_{\tilde b \leq B}
    \Big\| 
         Q^{i;2}_{\tilde b, \tilde b} 
        -
        \mean_\bV\big[ Q^{i;2}_{\tilde b,\tilde b} \big]
    \Big\|_{L_6}
    \\
    \;\leq&\;
    \mfrac{2 C'_1 }{B^{1/2}}
    \max_{\tilde b \leq B}
    \| \bar T^{i, \tilde b} \|_{L_6}
    +
    \mfrac{2 C'_1}{B^{3/2}}  \mfrac{m^{t/(36+6t)}}{n^{t/(36+6t)}} \alpha^{(m)}_{2,2;t}
    \;.
\end{align*}
Denote $\mean_{\bV, \tilde b}[\argdot] \coloneqq \mean[\argdot | \bV_1, \ldots, \bV_n, \upsilon_{\tilde b} ]$. To control $\| \tilde S^{i, \tilde b} \|_{L_6}$, we condition further on $\upsilon_{\tilde b}$ and rewrite again
\begin{align*}
    \bar T^{i, \tilde b}
    \;=\;
    \mean_{\bV,\tilde b}\Big[\bar T^{i, \tilde b} \Big]
    +
    \msum_{b^* \neq \tilde b}^B 
    \Big( 
        \mean_{\bV,\tilde b}\big[ \bar T^{i, \tilde b} \big| \upsilon_b, \ldots, \upsilon_1 \big]
        -
        \mean_{\bV,\tilde b}\big[ \bar T^{i, \tilde b} \big| \upsilon_{b-1}, \ldots, \upsilon_1 \big]
    \Big)
    \;.
\end{align*}
This allows us to apply the same martingale difference sequence bound as before to get 
\begin{align*}
    \|  \bar T^{i, \tilde b} \|_{L_6}
    \;\leq&\;
    \Big\|  \mean_{\bV,\tilde b}\Big[ \bar T^{i, \tilde b} \Big] \Big\|_{L_6}
    \\
    &\; +
    C'_1 B^{1/2} \max_{b^* \neq \tilde b}
    \Big\| 
    \mean_{\bV,\tilde b}\big[ \bar T^{i, \tilde b} \big| \upsilon_{b^*}, \ldots, \upsilon_1 \big]
        -
        \mean_{\bV,\tilde b}\big[ \bar T^{i, \tilde b} \big| \upsilon_{b^*-1}, \ldots, \upsilon_1 \big]
    \Big\|_{L_6}
    \;.
\end{align*}
Moreover by using $E^i_b$ again, we have 
\begin{align*}
    \Big\|  \mean_{\bV,\tilde b}\Big[ \bar T^{i, \tilde b} \Big] \Big\|_{L_6}
    \;\leq&\; 
    \max_{b \neq \tilde b} 
    \Big\|  \mean_\bV\big[ Q^{i;2}_{b,\tilde b} \,\big|\, \upsilon_{\tilde b}  \big]
                    -
                    \mean_\bV\big[ Q^{i;2}_{b,\tilde b} \big]
    \Big\|_{L_6}
    \\
    \;=&\;
    \Big\|  \P_\bV( E^i_b ) \big( \mean_\bV\big[ Q^i_{b,\tilde b} \,\big|\, \upsilon_{\tilde b}, E^i_b  \big]
                    -
                    \mean_\bV\big[ Q^i_{b,\tilde b} \,\big|\, E^i_b  \big]
                \big)
    \Big\|_{L_6}
    \\
    \;\leq&\; 
    \mfrac{2 m}{n} \max_{b \neq \tilde b} \|  Q^{i;2}_{b,\tilde b} \|_{L_6}
    \;\leq\;
    \mfrac{2 m}{n}  \mfrac{m^{t/(36+6t)}}{n^{t/(36+6t)}} \alpha^{(m)}_{2,2;t}
    \;,
\end{align*}
whereas 
\begin{align*}
    &\;
    B^{1/2} \Big\| 
    \mean_{\bV,\tilde b}\big[ \bar T^{i, \tilde b} \big| \upsilon_{b^*}, \ldots, \upsilon_1 \big]
        -
        \mean_{\bV,\tilde b}\big[ \bar T^{i, \tilde b} \big| \upsilon_{b^*-1}, \ldots, \upsilon_1 \big]
    \Big\|_{L_6}
    \\
    &\;\leq\;
    \mfrac{1}{B^{1/2}}
    \Big\| 
        Q^{i;2}_{b^*, \tilde b} - \mean_{\bV, \tilde b}\big[ Q^{i;2}_{b^*,\tilde b}   \big]
    \Big\| 
    \\
    &\;\leq\; \mfrac{2}{B^{1/2}}
    \max_{b \neq  b'} \|  Q^{i;2}_{b, b'} \|_{L_6}
    \\
    &
    \;\leq\; \mfrac{2}{B^{1/2}}  \mfrac{m^{t/(36+6t)}}{n^{t/(36+6t)}} \alpha^{(m)}_{2,2;t}
    \;.
\end{align*}
Combining the above bounds, we obtain that 
\begin{align*}
    (\star)_{21} 
    \;=&\; 
    O\Big(
    \Big(
        \mfrac{m}{n B^{1/2}}
        + 
        \mfrac{1}{B}
        +
        \mfrac{1}{B^{3/2}}  
    \Big)
    \mfrac{m^{t/(36+6t)}}{n^{t/(36+6t)}} 
    \alpha^{(m)}_{2,2;t}
    \Big)
    \\
    \;=&\;
    O\Big(
    \Big(
        \mfrac{m}{n B^{1/2}}
        + 
        \mfrac{1}{B}
    \Big)
    \mfrac{m^{t/(36+6t)}}{n^{t/(36+6t)}} 
    \alpha^{(m)}_{2,2;t}
    \Big)
    \;.
\end{align*}
Combining this with the bound on $(\star)_{22}$, we obtain that 
\begin{align*}
    \| \bar Q^{i;2} \|_{L_6}
    \;=&\; 
    O\Big(
        \Big(
            \mfrac{m^2}{n^2}
            +
            \mfrac{m}{n} \mfrac{1}{B^{1/2}} \mfrac{m^{t/(36+6t)}}{n^{t/(36+6t)}} 
            + 
            \mfrac{1}{B} \mfrac{m^{t/(36+6t)}}{n^{t/(36+6t)}} 
        \Big)
        \alpha^{(m)}_{2,2;t}
    \Big)
    \\
    \;=&\;
    O\Big(
        \Big(
            \mfrac{m}{n}
            +
            \mfrac{1}{B^{1/2}} \mfrac{m^{t/(72+12t)}}{n^{t/(72+12t)}} 
        \Big)^2
        \alpha^{(m)}_{2,2;t}
    \Big)
    \;,
\end{align*}
where we have used that $m/n = o(1)$. Combining this with the bound on $\| \bar Q^{i;1} \|_{L_6}$, we obtain that 
\begin{align*}
    (\star)_2
    \;=&\;
    \Big\|{\sup_{\bV_i\in[\bzero,\Phi_i\bX_i]}}
    \| 
        \partial g\big( f_m^{(B)}(\bW_i(\bV_i)) \big)
        D^2_i f_m^{(B)}(\bW_i(\bV_i))
    \\
    &\;
        \hspace{5em}
        +
        \partial^2 g\big( f_m^{(B)}(\bW_i(\bV_i)) \big)
        \big( D_i f_m^{(B)}(\bW_i(\bV_i)) \otimes D_i f_m^{(B)}(\bW_i(\bV_i)) \big)
    \|\Big\|_{L_{6}}
    \\
    \;=&\;
    O\Big(
        \Big( \mfrac{m}{n} + \mfrac{ m^{t/(36+6t)}}{B^{1/2} \, n^{t/36+6t}} \Big) \, \alpha^{(m)}_{2,1;t}
        +
        \Big(
            \mfrac{m}{n}
            +
            \mfrac{m^{t/(72+12t)}}{B^{1/2} n^{t/(72+12t)}} 
        \Big)^2
        \alpha^{(m)}_{2,2;t}
    \Big)
    \\
    \;=&\;
    o\Big(
        \mfrac{ \alpha^{(m)}_{2,1;t}}{\sqrt{n}}
        +
        \mfrac{ \alpha^{(m)}_{2,2;t}}{n}
    \Big)
   \;.
\end{align*}
In the last line, we have used that  $m=o(\sqrt{n})$ and $B \gg n^{1 - t/(108+18t)} \gg n^{1 - t/(72+12t)} \gg n^{1 - t/(36+6t)}$. The same proof holds for all $i \leq n$ and $\Phi_i\bX_i$ replaced by $\bZ_i$, and therefore 
\begin{align*}
    \alpha_2^{(B)} 
    \;=\;
    o\Big( \mfrac{ \alpha^{(m)}_{2,1;t}}{\sqrt{n}}
        +
        \mfrac{ \alpha^{(m)}_{2,2;t}}{n}
    \Big)\;.
\end{align*}
The proof for the third derivative term is exactly analogous by exploiting $E^i_b$'s and an iterative martingale difference sequence bound, except that we need to handle the following three terms separately:
\begin{align*}
    &\;
     \partial g\big( f_m^{(B)}(\bW_i(\bV_i)) \big)
        D^3_i f_m^{(B)}(\bW_i(\bV_i))
    \;,
    \\
    &\;
     \partial g^2\big( f_m^{(B)}(\bW_i(\bV_i)) \big)
    \big( D_i f_m^{(B)}(\bW_i(\bV_i)) \otimes D^2_i f_m^{(B)}(\bW_i(\bV_i)) \big)
    \;,
    \\
    &\;
     \partial g^3\big( f_m^{(B)}(\bW_i(\bV_i)) \big)
    \big( D_i f_m^{(B)}(\bW_i(\bV_i)) \otimes D_i f_m^{(B)}(\bW_i(\bV_i)) \otimes D_i f_m^{(B)}(\bW_i(\bV_i))  \big)
    \;.
\end{align*}
One may verify that under the condition $m=o(\sqrt{n})$ and $B \gg n^{1 - t/(108+18t)}$,
\begin{align*}
    \alpha_3^{(B)} 
    \;=\;
    o\bigg( 
        \mfrac{ \alpha^{(m)}_{3,1;t}}{\sqrt{n}}
        +
        \mfrac{ \alpha^{(m)}_{3,2;t}}{n}
        +
        \mfrac{ \alpha^{(m)}_{3,3;t}}{n^{3/2}}
    \bigg)\;,
\end{align*}
which finishes the proof.
\qed 

\vspace{1em}

\subsection{Proof of \Cref{lemma:bagging:quadratic:universality}} \label{appendix:proof:bagging:quad} We use the version of \Cref{thm:main:general} discussed in \Cref{remark:non:iid:no:taylor} (i.e.~without taking Cauchy-Schwarz inequality), which gives that for some $\Theta \sim \textrm{Uniform}[0,1]$ independent of all other random variables,
\begin{align*}
    \big| \mean\big[ h(f^{\rm quad}(\Phi \cX )  ) \big] - \mean\big[ f^{\rm quad}(\cZ)  \big] \big|
    \;\leq\; 
    \msum_{i=1}^n 
    \big| 
        \;
        \mean\big[  F_{\bW_i, \Theta}(\Phi_i \bX_i)  -  F_{\bW_i, \Theta}(\bZ_i) \big] 
        \;
    \big|
    \;,
\end{align*}
where we have defined 
\begin{align*}
     F_{\bW_i, \Theta}(\bx)
     \;\coloneqq&\;
    \partial h( f^{\rm quad}( \bW_i( \Theta  \, \bx ) ) ) 
    \, 
    \partial_i f^{\rm quad}( \bW_i( \Theta \bx )^\top  \bx
    \;.
\end{align*}
Since the derivative of $h$ is uniformly bounded from above by $1$, by the triangle inequality, we have 
\begin{align*}
    &\;
    \big| \mean\big[ h(f^{\rm quad}(\Phi \cX )  ) \big] - \mean\big[ f^{\rm quad}(\cZ)  \big] \big|
    \\
    &\;\leq\;
    \msum_{i \leq n}
    \big( 
        \mean \, \big| 
            D_i f^{\rm quad}( \bW_i( \Theta \Phi_i\bX_i ) ) (\Phi_i\bX_i) \big|^3
        +
        \mean \, \big| 
            D_i f^{\rm quad}( \bW_i( \Theta \bZ_i ) ) (\bZ_i) \big|^3
    \big)
    \;.
    \tagaligneq \label{eq:bag:quadratic:lindeberg}
\end{align*}
We first use the definition of $f^{\rm quad}$ to express
\begin{align*}
    &\;\| D_i f^{\rm quad}( \bW_i( \Theta \Phi_i\bX_i ) ) (\Phi_i\bX_i) \|_{L_3}
    \\
    &\;=\; 
    \Big\| \mfrac{1}{B^2} \msum_{b,b' \leq B} 
        D_i f_m^{\rm quad}( \bW^{\upsilon_b, \upsilon_{b'}}_i( \Theta \Phi_i\bX_i ) ) 
        (\Phi_i\bX_i)
    \Big\|_{L_3}
    \;,
\end{align*}
where we have denoted
\begin{align*}
    &\;
    \bW^{\upsilon_b, \upsilon_{b'}}_i( \bx )  
    \;\coloneqq\; 
    ( \eta_{\upsilon_b(1)}( \bx )  , \ldots, \eta_{\upsilon_b(m)} ( \bx )  , \eta_{\upsilon_{b'}(1)}( \bx )  , \ldots, \eta_{\upsilon_{b'}(m)}( \bx )     )
    \\
    &\;
    \eta_{i'}(\bx) \;\coloneqq\; \begin{cases}
        \Phi_i\bX_i & \text{ for } i' < i\;,
        \\
        \bx & \text{ for } i' = i\;,
        \\
        \bZ_i & \text{ for } i' > i\;.
    \end{cases}
\end{align*}
Define the event  $E^i_b = \{  i \in \{ \upsilon_b(l) \}_{l \leq m} \}$ as in the proof of \Cref{prop:bagging:general}. By the triangle inequality, we have that 
\begin{align*}
    &\;\| D_i f^{\rm quad}( \bW_i( \Theta \Phi_i\bX_i ) ) (\Phi_i\bX_i)^{\otimes s} \|_{L_3}
    \\
    &\;\leq\; 
    \Big\| \mfrac{1}{B^2}  \msum_{b' \leq B} \msum_{b \neq b'}   
        \partial_{\Phi_i\bX_i} f_m^{\rm quad}( \bW^{\upsilon_b, \upsilon_{b'}}_i( \Theta \Phi_i\bX_i ) ) 
        (\Phi_i\bX_i)
        \,
        \ind_{E^i_b \cap (E^i_{b'})^c}
    \Big\|_{L_3}
    \\
    &\qquad 
    +
    \Big\| \mfrac{1}{B^2}     \msum_{b \leq B} \msum_{b' \neq b}  
        \partial_{\Phi_i\bX_i} f_m^{\rm quad}( \bW^{\upsilon_b, \upsilon_{b'}}_i( \Theta \Phi_i\bX_i ) ) 
        (\Phi_i\bX_i)
        \,
        \ind_{ (E^i_b)^c \cap E^i_{b'}}
    \Big\|_{L_3}
    \\
    &\qquad 
    +
    \Big\| \mfrac{1}{B^2}   \msum_{b,b' \leq B} 
        \partial_{\Phi_i\bX_i} f_m^{\rm quad}( \bW^{\upsilon_b, \upsilon_b}_i( \Theta \Phi_i\bX_i ) ) 
        (\Phi_i\bX_i)
        \,
        \ind_{ E^i_b \cap E^i_{b'}}
    \Big\|_{L_3}
    \\
    &\;\leq\; 
     \max_{b' \leq B}
    \Big\| \mfrac{1}{B} \msum_{b \neq b'}  
        \partial_{\Phi_i\bX_i} f_m^{\rm quad}( \bW^{\upsilon_b, \upsilon_{b'}}_i( \Theta \Phi_i\bX_i ) ) 
        (\Phi_i\bX_i)
        \,
        \ind_{(E^i_{b'})^c}
    \Big\|_{L_3}
    \\
    &\qquad 
    +
    \max_{b \leq B}
    \Big\| \mfrac{1}{B} \msum_{b' \neq b}  
        \partial_{\Phi_i\bX_i} f_m^{\rm quad}( \bW^{\upsilon_b, \upsilon_{b'}}_i( \Theta \Phi_i\bX_i ) ) 
        (\Phi_i\bX_i)
        \,
        \ind_{(E^i_{b})^c}
    \Big\|_{L_3}
    \\
    &\qquad  
    +
    \Big\| \mfrac{1}{B^2}   \msum_{b,b' \leq B} 
        \partial_{\Phi_i\bX_i} f_m^{\rm quad}( \bW^{\upsilon_b, \upsilon_b}_i( \Theta \Phi_i\bX_i ) ) 
        (\Phi_i\bX_i)
        \,
        \ind_{ E^i_b \cap E^i_{b'}}
    \Big\|_{L_3}
    \\
    &\;=\;
    \max_{b' \leq B}  \Big\|  D_i f_1^{b', \upsilon_{b'}}( \bW_i( \Theta \Phi_i\bX_i ) ) (\Phi_i \bX_i)
    \Big\|_{L_3}
    \\
    &\qquad 
    +
    \max_{b \leq B}  \Big\|  D_i f_2^{b, \upsilon_b}( \bW_i( \Theta \Phi_i\bX_i ) ) (\Phi_i \bX_i)
    \Big\|_{L_3}
    \\
    &\qquad 
    +
    \Big\|  D_i  f_3( \bW_i( \Theta \Phi_i\bX_i ) )  (\Phi_i \bX_i)
    \Big\|_{L_3}
    \;,
\end{align*}
where we have defined, for $\bv_1, \ldots, \bv_n \in \cD^k$,
\begin{align*}
    f_1^{b', \upsilon_{b'}}( \bv_1, \ldots, \bv_n  )
    \;=&\;
    \mfrac{1}{B} \msum_{\substack{b \leq B  \\ b \neq b'}}
     f_m^{\rm quad}( \bv_{\upsilon_b(1)}, \ldots, \bv_{\upsilon_b(m)}, \bv_{\upsilon_{b'}(1)}, \ldots, \bv_{\upsilon_{b'}(m)} ) 
    \ind_{(E^i_{b'})^c }
     \;,
     \\
     f_2^{b, \upsilon_b}( \bv_1, \ldots, \bv_n  )
    \;=&\;
    \mfrac{1}{B} \msum_{\substack{b' \leq B  \\ b' \neq b}}
     f_m^{\rm quad}( \bv_{\upsilon_b(1)}, \ldots, \bv_{\upsilon_b(m)}, \bv_{\upsilon_{b'}(1)}, \ldots, \bv_{\upsilon_{b'}(m)} ) 
    \ind_{(E^i_b)^c }
    \;,
     \\
     f_3( \bv_1, \ldots, \bv_n  )
    \;=&\;
    \mfrac{1}{B^2}   \msum_{b,b' \leq B} 
        f_m^{\rm quad}( \bv_{\upsilon_b(1)}, \ldots, \bv_{\upsilon_b(m)}, \bv_{\upsilon_{b'}(1)}, \ldots, \bv_{\upsilon_{b'}(m)} ) 
        \,
        \ind_{ E^i_b \cap E^i_{b'}}
     \;.
\end{align*}
By construction, $\bv_i$ can only appear in $f_1^{b', \upsilon_{b'}}( \bv_1, \ldots, \bv_n )$ through the first $m$ arguments of $f^{\rm quad}_m$, on which the permutations $\upsilon_b$ act, and similarly $\bv_i$ can only appear in $f_2^{b', \upsilon_{b'}}( \bv_1, \ldots, \bv_n )$ through the last $m$ arguments of $f^{\rm quad}_m$, on which the permutations $\upsilon_{b'}$ are act. 
Therefore, $f_1^{b', \upsilon_{b'}}$ and $f_2^{b, \upsilon_b}$ are exactly in the form of $f^{(B)}_m$ considered in \Cref{prop:bagging}. In particular, their derivatives are in the form of $\frac{1}{B} \sum_{b \leq B} S^i_b$ in \textbf{Step 1} in the proof of \Cref{prop:bagging:general} (the generalization of \Cref{prop:bagging}), where each $S^i_b$ vanishes on the event $E^i_b$. The only differences are that
\begin{proplist}
    \item  In both \Cref{prop:bagging} and \Cref{prop:bagging:general}, we have stated a control in terms of the norms of the derivatives of $ D^s_i f^{(B)}_m$, but observe that the exact same proof applies to quantities of the form  $D^s_i f^{(B)}_m(\bW_i(\Theta \Phi_i \bX_i)) (\Phi_i \bX_i)^{\otimes s}$;
    \item We can use the H\"older inequality with respect to $L_3$ norm instead of the $L_6$ norm.
\end{proplist}
Therefore by the same argument as \textbf{Step 1} in the proof of \Cref{prop:bagging} , we obtain
\begin{align*}
    \Big\|  D_i f_1^{b', \upsilon_{b'}}( \bW_i( \Theta \Phi_i\bX_i ) ) (\Phi_i \bX_i)
    \Big\|_{L_3}
    \;=&\;
    O\Big( 
        \Big( \mfrac{1}{B^{1/2}} \mfrac{m^{t/(9+3t)}}{n^{t/(9+3t)}} + \mfrac{m}{n}\Big) 
        \,
        \alpha^{\rm quad}_{1;t} 
    \Big)
    \\
    \;=&\;
    o\Big( 
        \mfrac{\alpha^{\rm quad}_{1;t} }{\sqrt{n}}
    \Big)
\end{align*}
where we have used $m=o(n^{1/2})$ and $B\gg n^{1-t/(18+6t)} \gg n^{1-t/(9+3t)} $ and recalled the definition
\begin{align*}
    \alpha^{\rm quad}_{1;t} 
    \;\coloneqq\;
    \max_{\substack{i \leq n \\ \upsilon, \upsilon' \in S([m]) }} 
        \max \,\Big\{ & \,
        \|  \partial_{\Phi_i\bX_i}  f^{\rm quad}_m( \bW^{\upsilon,\upsilon'}_i(\Theta \Phi_i\bX_i) ) (\Phi_i\bX_i) \|_{L_{3+t}}
        \,,\,
    \\
        &\,
        \|  \partial_{\bZ_i}  f^{\rm quad}_m( \bW^{\upsilon,\upsilon'}_i(\Theta \bZ_i) ) (\bZ_i) \|_{L_{3+t}}
        \Big\} \;.
\end{align*}
Similarly,
\begin{align*}
    \Big\|  D_i f_2^{b', \upsilon_{b'}}( \bW_i( \Theta \Phi_i\bX_i ) ) (\Phi_i \bX_i)
    \Big\|_{L_3}
    \;=\;
    o\Big( 
        \mfrac{\alpha^{\rm quad}_{1;t} }{\sqrt{n}}
    \Big)
    \;.
\end{align*}
To handle $f_3$, which involves a double-sum, we notice that each summand vanishes on the event $(E^i_b)^c \cup (E^i_{b'})^c$. Indeed, its derivatives are exactly in the form of $\bar Q^{i;2} = \frac{1}{B^2} \sum_{b,b' \leq B} Q^{i;2}_{b,b'}$ in \textbf{Step 2} in the proof of \Cref{prop:bagging:general}, which makes the same proof applicable, and therefore 
\begin{align*}
    \Big\|  D_i f_3( \bW_i( \Theta \Phi_i\bX_i ) ) (\Phi_i \bX_i)
    \Big\|_{L_3}
    \;=&\;
    O\Big( 
        \Big( \mfrac{m}{n} + \mfrac{1}{B^{1/2}} \mfrac{m^{t/(18+6t)}}{n^{t/(18+6t)}} \Big)^2 
        \,
        \alpha^{\rm quad}_{1;t}
    \Big)
    \\
    \;=&\;
    o\Big( 
        \mfrac{\alpha^{\rm quad}_{1;t} }{\sqrt{n}}
    \Big)
    \;,
\end{align*}
where we have again used $m=o(n^{1/2})$ and $B =\Omega(n^{1-t/(18+6t)})$. The same argument applies with $\Phi_i\bX_i$ replaced by $\bZ_i$ too. Applying the H\"older's inequality to \eqref{eq:bag:quadratic:lindeberg} followed by using the above derivative bounds, we obtain that
\begin{align*}
    \big| \mean\big[ h(f^{\rm quad}(\Phi \cX )  ) \big] - \mean\big[ f^{\rm quad}(\cZ)  \big] \big|
    \;=\;
    o\Big( n \times 
        \Big( 
            \mfrac{ \alpha^{\rm quad}_{1;t} }{\sqrt{n}} 
        \Big)
    \Big)
    \;=\;
    o\big( \alpha^{\rm quad}_{1;t} \sqrt{n} \big)
\end{align*}
as desired.
\qed

\vspace{1em}

\subsection{Proof of \Cref{prop:bagging}: Stability of a bagged estimator.} \label{appendix:proof:prop:bagging} We seek to apply \Cref{prop:bagging:general}. By setting $q=1$ and identifying $g: \R \rightarrow \R$ as the identity function, higher derivatives of $g$ vanish, which yields the desired bounds that 
\begin{align*}
    \alpha_r(f_m^{(B)})
    \;=\;
    o\Big( \mfrac{\alpha^{\rm base}_{r;t}}{\sqrt{n}}
    \Big) 
    \qquad 
    \text{ for } r=1,2,3 \;.
\end{align*}
\qed

\subsection{Proof of \Cref{prop:bagged:network}: Universality of augmented-and-bagged locally dependent nonlinear feature models}  \label{appendix:proof:prop:bagged:network}

By an analogous argument to the proof of \Cref{lem:network:risk} in \Cref{appendix:proof:network:risk}, except that $\hat \beta_\lambda$ is replaced by $\hat \beta_\lambda^{\rm bagged}  = \frac{1}{B} \sum_{b \leq B} \hat \beta_{\lambda;m}^{\upsilon_b}$, we can compute 
\begin{align*}
    \hat L_\lambda^{\rm bagged}(\cX)
     \;=&\;
    \mfrac{1}{B^2} \msum_{b,b' \leq B}
    \beta^\top 
    \bW^{(0)}
    (\bar \bX^{(b)}_3)^\top 
    \bar \bX^{{(b)};-1}_{1;\lambda} 
    \,M^{\varphi_\theta} \,
    \bar \bX^{{(b')};-1}_{1;\lambda} 
    (\bar \bX^{(b')}_3  ) (\bW^{(0)})^\top 
    \beta  
    \\
    &\;
    +    
    \mfrac{\sigma^2_\epsilon}{n} \,
    \mfrac{1}{B^2} \msum_{b,b' \leq B}
    \,
    \Tr\big(\,
    \bar \bX^{{(b)};-1}_{1;\lambda} 
    \,M^{\varphi_\theta} \,
    \bar \bX^{{(b')};-1}_{1;\lambda} 
    \,
    \bar \bX^{(b,b')}_2
    \,\big)
    \\
    &\;
    -
    \mfrac{2}{B} \msum_{b \leq B}
    \beta^\top 
    \bW^{(0)}
    (\bar \bX^{(b)}_3)^\top 
    \bar \bX^{{(b)};-1}_{1;\lambda} 
    \, R^{\varphi_\theta, \varphi_{\theta_0}} \,
    \bW^{(0)} \beta 
    \\
    &\;
    +
    \beta^\top \bW^{(0)} M^{\varphi_{\theta_0}} (\bW^{(0)})^\top \beta
    +
    \sigma_\epsilon^2
    \;,
\end{align*}
where we have denoted 
\begin{align*}
    \bar \bX^{(b)}_1 
    \;\coloneqq&\; 
    \mfrac{1}{mk} \msum_{i=1}^m \msum_{j=1}^k  \tilde \bV_{\upsilon_b(i)j} (\tilde \bV_{\upsilon_b(i)j})^\top
    \;,
    \quad
    \bar \bX^{(b)}_3
    \;\coloneqq\; 
    \mfrac{1}{mk} \msum_{i=1}^m \msum_{j=1}^k \tilde \bV_{\upsilon_b(i)j} \tilde \bV_0^\top
    \;,
    \\
    \bar \bX^{(b,b')}_2 
    \;\coloneqq&\;
    \mfrac{1}{m^2} \msum_{i,i'=1}^m
    \ind_{\{ \upsilon_b(i) = \upsilon_{b'}(i') \}}
    \Big( \mfrac{1}{k} \msum_{j=1}^k  \tilde \bV_{\upsilon_b(i)j} \Big)
    \,
    \Big( \mfrac{1}{k} \msum_{j=1}^k  \tilde \bV_{\upsilon_{b'}(i')j} \Big)^\top 
    \;,
    \\
    \bar \bX^{(b);-1}_{1;\lambda}
    \;\coloneqq&\;
    \begin{cases}
        (    \bar \bX^{(b)}_1 + \lambda \bI_p )^{-1}
        & \text{ for } \lambda > 0 \;,
        \\
        (    \bar \bX^{(b)}_1 )^\dagger
        & \text{ for } \lambda = 0 \;.
    \end{cases}
\end{align*}
Now for $b,b' \leq B$, define  
\begin{align*}
    \hat L_\lambda^{(b,b')}(\cX)
     \;=&\;
    \beta^\top 
    \bW^{(0)}
    (\bar \bX^{(b)}_3)^\top 
    \bar \bX^{{(b)};-1}_{1;\lambda} 
    \,M^{\varphi_\theta} \,
    \bar \bX^{{(b')};-1}_{1;\lambda} 
    (\bar \bX^{(b')}_3  ) (\bW^{(0)})^\top 
    \beta  
    \\
    &\;
    +    
    \mfrac{\sigma^2_\epsilon}{n} \,
    \Tr\big(\,
    \bar \bX^{{(b)};-1}_{1;\lambda} 
    \,M^{\varphi_\theta} \,
    \bar \bX^{{(b')};-1}_{1;\lambda} 
    \,
    \bar \bX^{(b,b')}_2
    \,\big)
    \\
    &\;
    -
    2
    \beta^\top 
    \bW^{(0)}
    (\bar \bX^{(b)}_3)^\top 
    \bar \bX^{{(b)};-1}_{1;\lambda} 
    \, R^{\varphi_\theta, \varphi_{\theta_0}} \,
    \bW^{(0)} \beta 
    \\
    &\;
    +
    \beta^\top \bW^{(0)} M^{\varphi_{\theta_0}} (\bW^{(0)})^\top \beta
    +
    \sigma_\epsilon^2
    \;,
\end{align*}
which allows us to write 
\begin{align*}
    \hat L_\lambda^{\rm bagged}(\cX) \;=\; \mfrac{1}{B^2} \msum_{b,b' \leq B} \hat L_\lambda^{(b,b')}(\cX)
    \;.
\end{align*}
This allows us to apply \Cref{lemma:bagging:quadratic:universality} and obtain that
\begin{align*}
    &\;
    d_{\tilde \cH^{(4)}} \big( \hat L_\lambda^{\rm bagged}(\cX)  \,,\, \hat L_\lambda^{\rm bagged}(\cX) \big)
    \\
    &\;=\;
    o\Big( 
        \mfrac{1}{\sqrt{n}} 
        \, 
        \max_{\substack{i \leq n \\ \upsilon, \upsilon' \in S([m]) }} 
        \max \,\Big\{ \,
        \|  n \,  \partial_{\Phi_i\bX_i}  \hat L^{(b,b')}_\lambda( \bW^{\upsilon,\upsilon'}_i(\Theta \Phi_i\bX_i) ) (\Phi_i\bX_i) \|_{L_{3+t}}
            \,,\,
    \\
            &\,\hspace{12em} 
            \|  n \, \partial_{\bZ_i} \hat L^{(b,b')}_\lambda( \bW^{\upsilon,\upsilon'}_i(\Theta \bZ_i) ) (\bZ_i) \|_{L_{3+t}}
            \Big\} 
    \Big)
    \;.
\end{align*}
Now notice that $\partial_{\bZ_i} \hat L^{(b,b')}_\lambda$ is almost identical to $\hat L_\lambda(\cX)$ except that the matrices $\bar \bX^{*;-1}_{1;\lambda} $, $\bar \bX^{*}_2$ and $\bar \bX^{*}_3$ have been replaced by their bagged analogues. In particular, without applying the bounds on $\gamma_1^\varphi, \gamma_2^\varphi, \gamma_3^\varphi$, \textbf{Step 1 -- 4} of the proof of \Cref{prop:network:extend:general} can be recycled to show that 
\begin{align*}
     &\;
    d_{\tilde \cH^{(4)}} \big( \hat L_\lambda^{\rm bagged}(\cX)  \,,\, \hat L_\lambda^{\rm bagged}(\cX) \big)
    \\
    &\;=\;
    o\Big( 
        \mfrac{1}{\sqrt{n}} 
        \Big(  
            e^{-\Omega(K)}
            +
            3 \epsilon \| \tilde f_K \|_{\rm Lip}
            +
            \mfrac{C_K}{\epsilon^{3N_K}}
            B_d
            \Big( 
                1
                +
                \mfrac{1}{\lambda^{6}} 
            \Big)
            \Big( 
                \mfrac{(\gamma_1^\varphi)^3}{d^{1/2}}
                +
                \gamma_1^\varphi  \gamma_2^\varphi 
                +
                \gamma_3^\varphi d^{1/2}
            \Big)
        \Big)
    \Big)
    \;.
\end{align*}
We now apply Assumption \hyperref[assumption:bagging:feature:gradient]{9$^{(B)}$} instead of \Cref{assumption:network:feature:gradient}:
\begin{align*}
    \gamma_1^\varphi \;=\; O\big( B_d^{-1/3} d^{1/3} \big) 
    \,,\, 
    \quad 
    \gamma_2^\varphi \;=\; O\big( B_d^{-2/3} d^{ 1/6} \big) 
    \,,\,
    \quad 
    \gamma_3^\varphi \;=\; O\big( B_d^{-1}  \big) \;,
\end{align*}
and fix $K > 0$ and $\epsilon > 0$. We then obtain the desired bound
\begin{align*}
    d_{\tilde \cH^{(4)}} \big( \hat L_\lambda^{\rm bagged}(\cX)  \,,\, \hat L_\lambda^{\rm bagged}(\cX) \big)
    \;=\; 
    o\Big( 1 
                +
                \mfrac{1}{\lambda^{6}} 
    \Big) 
    \;.
\end{align*}
The proof for the $\lambda=0$ case is exactly the same as \textbf{Step 5} of the proof of \Cref{prop:network:extend:general}, except that the covariance matrices need to be replaced by the corresponding bagged versions and \Cref{assumption:network:ridgeless:by:ridge} is replaced by Assumption \hyperref[assumption:bagging:ridgeless:by:ridge]{10$^{(B)}$}. This finishes the proof.
\qed

\vspace{1em}

\subsection{Proof of \Cref{cor:bagging:network:nonlinear}: Universality of augmented-and-bagged nonlinear networks} \label{appendix:proof:cor:bagging}

We seek to apply \Cref{prop:bagged:network}. The proof is largely similar to that for the linear network case (\Cref{prop:network}): \Cref{assumption:network:generalize}(i)-(iii) are automatically satisfied, whereas  \Cref{assumption:network:generalize}(iv) and the part of \Cref{assumption:network:generalize}(v) that concerns $\bV_{ij0}$ and $\bV_{ij1}$ are verified in the same way as that in the proof of \Cref{prop:network} in \Cref{appendix:proof:ridgeless:extend}. The mean-zero and sub-Gaussianity of $\tilde \bV_{ij}$ and $\tilde \bV_{ij}^0$ follow directly from the activation map conditions in \Cref{assumption:bagging:network:nonlinear}(iv). Verifying \Cref{assumption:network:bounded:cov} in the proof of \Cref{prop:network} rests on using that $\| \bW^{(0)}_{N_0-1} \ldots \bW^{(0)}_1 \|_{op} = O(1)$ with high probability and that $\| W_N \ldots W_1 \|_{op} = O(1)$; here, \Cref{assumption:network:bounded:cov}  can be verified directly with the additional operator norm controls in  \Cref{assumption:bagging:network:nonlinear}(iv) and the fact that $N, N_0$ are both fixed. 

\vspace{.5em}

We are left with verifying Assumption \hyperref[assumption:bagging:feature:gradient]{9$^{(B)}$}. By the $O(1)$-local dependency condition in \Cref{assumption:bagging:network:nonlinear} and noting that the augmentations considered do not increase the asymptotic size of the local dependency neighborhood (as verified in \Cref{appendix:proof:ridgeless:extend}), we have that $B_d = \Theta(1)$, so to verify Assumption \hyperref[assumption:bagging:feature:gradient]{9$^{(B)}$}, it suffices to show that $\gamma^\varphi_1, \gamma^\varphi_2, \gamma^\varphi_3$ are all $O(1)$. This again follows directly from the operator norm controls in  \Cref{assumption:bagging:network:nonlinear}(iv) and the fact that $N, N_0$ are both fixed. Therefore \Cref{prop:bagged:network} applies to give the desired result.

\qed  

\vspace{1em}

\subsection{Proof of \Cref{lem:tanh}: Verification of activation map conditions for pointwise tanh} \label{appendix:proof:tanh}

We first control the operator norms:
\begin{align*}
    \msup_{\bx \in \R^{d_l}}
    \| \partial^r \varphi_l(\bx) \|_{op} 
    \;=&\; 
    \sup_{\substack{\tilde \bx^{(1)},\ldots,\tilde \bx^{(r)}, \by \in \R^{d_l} \\ \| \tilde \bx^{(1)} \| = \ldots = \|\tilde \bx^{(r)}\| = \|\by\| =1 }  }  
    \big| \by^\top \partial^r \varphi_l(\bx) (\tilde \bx^{(1)} \otimes \ldots \otimes \tilde \bx^{(r)}) \big| 
    \\
    \;=&\;
    \sup_{\substack{\bx'_1,\ldots,\bx'_r, \by \in \R^{d_l} \\ \| \bx'_1 \| = \ldots = \|\bx'_r\| = \|\by\| =1 }  }  
    \big| 
        \msum_{s=1}^{d_l}
            y_s 
            \, 
            \partial^r \tanh(x_s) 
            \, 
            \tilde \bx^{(1)}_s \ldots \tilde \bx^{(r)}_s
    \big| 
    \\
    \;\leq&\;
    \msup_{x \in \R} | \partial^r \tanh(x)| \;=\; O(1)\;.
\end{align*}
The same argument applies to all $1 \leq l \leq N_0-1$ and to $\varphi_0$ as well, which verifies the operator norm bounds in  \Cref{assumption:bagging:network:nonlinear}(iv). Now note that $\bV_1 \overset{d}{=} - \bV_1$ by assumption, and that all augmentations considered in \Cref{assumption:network:augmentation} (and therefore in \Cref{assumption:bagging:network:nonlinear}) also satisfy that $\pi_{11}(\bV_1) \overset{d}{=} - \pi_{11}(\bV_1)$.  Since $\tanh(-x) = - \tanh(x)$, we have
\begin{align*}
    \tilde \bV_{1j} 
    \;=&\; 
     \bW_N \varphi_{N-1}( \bW_{N-1} \ldots \varphi_1(  \bW_1  (\pi_{1j}(\bV_1) ) ) \ldots )
     \\
     \;\overset{d}{=}&\;
     \bW_N \varphi_{N-1}( \bW_{N-1} \ldots \varphi_1(  \bW_1  ( - \pi_{1j}(\bV_1) ) ) \ldots )
     \\
     \;=&\;
     - \bW_N \varphi_{N-1}( \bW_{N-1} \ldots \varphi_1(  \bW_1  ( \pi_{1j}(\bV_1) ) ) \ldots )
     \;=\; - \tilde \bV_{1j} \;,
\end{align*}
which proves that $\tilde \bV_{1j} $'s are zero-mean. By the same argument, $\tilde \bV_{1j}^0$'s are zero-mean. Finally to verify sub-Gaussianity, we recall that $\bW_N$ is entrywsie i.i.d.~$\cN(0, 1/d_{N-1})$ and that $\varphi_{N-1}$ is the pointwise tanh activation function. Write 
\begin{align*}
    \bv_{N-1} \;\coloneqq\; \varphi_{N-1}( \bW_{N-1} \ldots \varphi_1(  \bW_1  (\pi_{1j}(\bV_1) ) ) \ldots )\;,
\end{align*}
which is bounded pointwise. Then for any $v \in \R^{p'}$ with $\| v \| = 1$, 
\begin{align*}
    v^\top \tilde \bV_{1j} 
    \;=&\; 
     v^\top 
     \bW_N 
     \bv_{N-1}\;,
\end{align*}
which, conditioning on $\bv_{N-1}$, is normal distributed with zero mean and a variance of 
\begin{align*}
    \mfrac{\| v \|^2 \| \bv_{N-1}  \|^2}{d_{N-1}}
    \;=\;
    \mfrac{1 \times \sum_{l \leq d_{N-1}} (\bv_{N-1})_l^2 }{d_{N-1}}
    \;\leq\;
    1
\end{align*}
almost surely. This implies that $\bv^\top \tilde \bV_{1j}$ is $1$-sub-Gaussian for all $\bv$ and therefore so is $\tilde \bV_{1j}$. The same argument applies to show that $\tilde \bV_{1j}^0$ are also sub-Gaussian, which concludes the proof. \qed